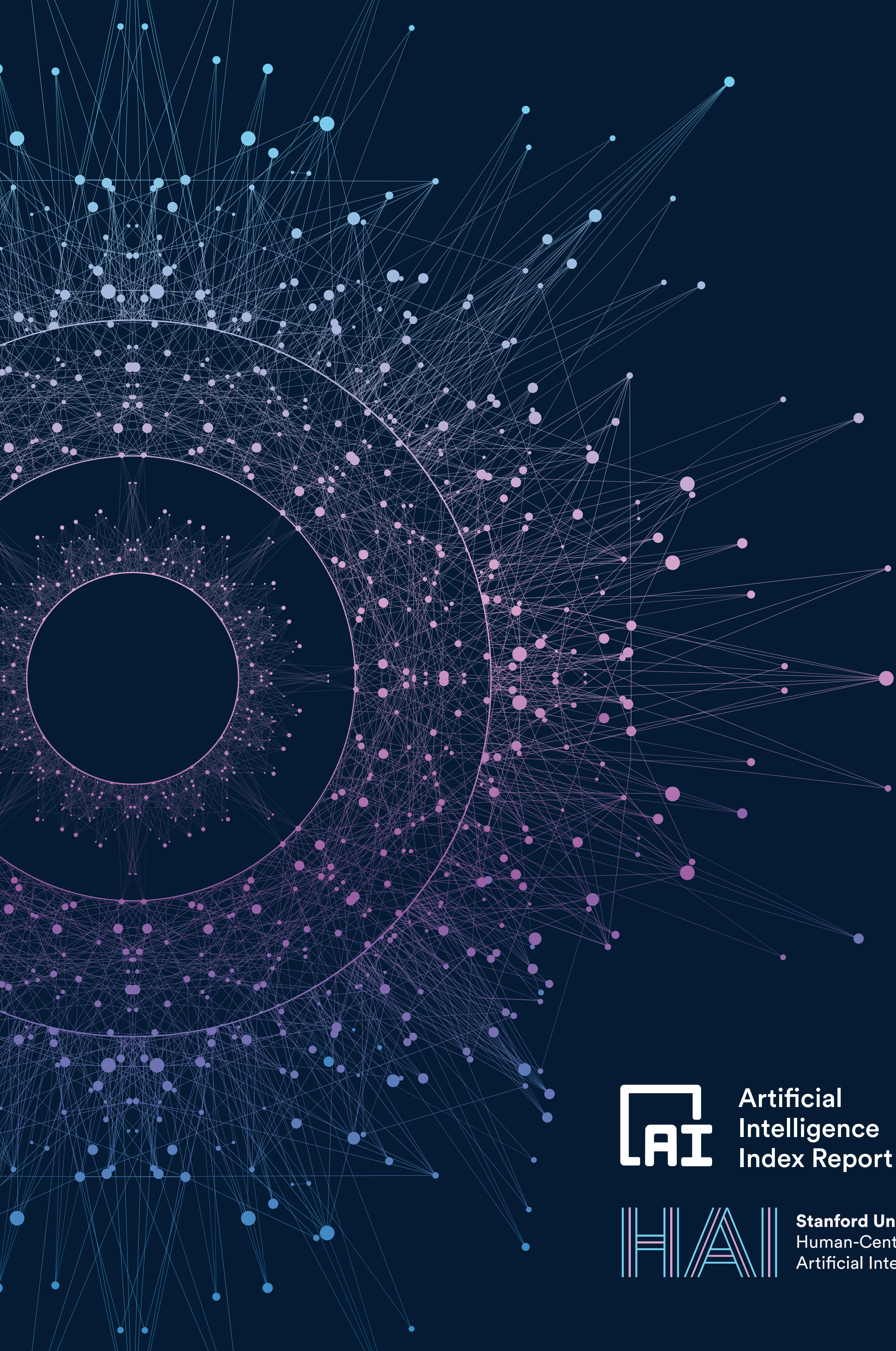

# Artificial Intelligence Index Report 2021

Stanford University
Human-Centered Artificial Intelligence

# INTRODUCTION TO THE 2021 AI INDEX REPORT

Welcome to the fourth edition of the AI Index Report! This year we significantly expanded the amount of data available in the report, worked with a broader set of external organizations to calibrate our data, and deepened our connections with Stanford's Institute for Human-Centered Artificial Intelligence (HAI).

The AI Index Report tracks, collates, distills, and visualizes data related to artificial intelligence. Its mission is to provide unbiased, rigorously vetted, and globally sourced data for policymakers, researchers, executives, journalists, and the general public to develop intuitions about the complex field of AI. The report aims to be the world's most credible and authoritative source for data and insights about AI.

## COVID AND AI

The 2021 report shows the effects of COVID-19 on AI development from multiple perspectives. The Technical Performance chapter discusses how an AI startup used machine-learning-based techniques to accelerate COVID-related drug discovery during the pandemic, and our Economy chapter suggests that AI hiring and private investment were not significantly adversely influenced by the pandemic, as both grew during 2020. If anything, COVID-19 may have led to a higher number of people participating in AI research conferences, as the pandemic forced conferences to shift to virtual formats, which in turn led to significant spikes in attendance.

## CHANGES FOR THIS EDITION

In 2020, we surveyed more than 140 readers from government, industry, and academia about what they found most useful about the report and what we should change. The main suggested areas for improvement were:

- **Technical performance:** We significantly expanded this chapter in 2021, carrying out more of our own analysis.
- **Diversity and ethics data:** We gathered more data for this year's report, although our investigation surfaced several areas where the AI community currently lacks good information.
- **Country comparisons:** Readers were generally interested in being able to use the AI Index for cross-country comparisons. To support this, we:
  - gathered more data to allow for comparison among countries, especially relating to economics and bibliometrics; and
  - included a thorough summary of the various AI strategies adopted by different countries and how they evolved over time.

## PUBLIC DATA AND TOOLS

The AI Index 2021 Report is supplemented by raw data and an interactive tool. We invite each member of the AI community to use the data and tool in a way most relevant to their work and interests.

- **Raw data and charts:** The public data and high-resolution images of all the charts in the report are available on Google Drive.
- **Global AI Vibrancy Tool:** We revamped the Global AI Vibrancy Tool this year, allowing for better interactive visualization when comparing up to 26 countries across 22 indicators. The updated tool provides transparent evaluation of the relative position of countries based on users' preference; identifies relevant national indicators to guide policy priorities at a country level; and shows local centers of AI excellence for not just advanced economies but also emerging markets.
- **Issues in AI measurement:** In fall 2020, we published "Measurement in AI Policy: Opportunities and Challenges," a report that lays out a variety of AI measurement issues discussed at a conference hosted by the AI Index in fall 2019.

# Table of Contents

ACCESS THE PUBLIC DATA

# TOP 9 TAKEAWAYS

1 **AI investment in drug design and discovery increased significantly:** "Drugs, Cancer, Molecular, Drug Discovery" received the greatest amount of private AI investment in 2020, with more than USD 13.8 billion, 4.5 times higher than 2019.

2 **The industry shift continues:** In 2019, 65% of graduating North American PhDs in AI went into industry—up from 44.4% in 2010, highlighting the greater role industry has begun to play in AI development.

3 **Generative everything:** AI systems can now compose text, audio, and images to a sufficiently high standard that humans have a hard time telling the difference between synthetic and non-synthetic outputs for some constrained applications of the technology.

4 **AI has a diversity challenge:** In 2019, 45% new U.S. resident AI PhD graduates were white—by comparison, 2.4% were African American and 3.2% were Hispanic.

5 **China overtakes the US in AI journal citations:** After surpassing the United States in the total number of journal publications several years ago, China now also leads in journal citations; however, the United States has consistently (and significantly) more AI conference papers (which are also more heavily cited) than China over the last decade.

6 **The majority of the US AI PhD grads are from abroad—and they're staying in the US:** The percentage of international students among new AI PhDs in North America continued to rise in 2019, to 64.3%—a 4.3% increase from 2018. Among foreign graduates, 81.8% stayed in the United States and 8.6% have taken jobs outside the United States.

7 **Surveillance technologies are fast, cheap, and increasingly ubiquitous:** The technologies necessary for large-scale surveillance are rapidly maturing, with techniques for image classification, face recognition, video analysis, and voice identification all seeing significant progress in 2020.

8 **AI ethics lacks benchmarks and consensus:** Though a number of groups are producing a range of qualitative or normative outputs in the AI ethics domain, the field generally lacks benchmarks that can be used to measure or assess the relationship between broader societal discussions about technology development and the development of the technology itself. Furthermore, researchers and civil society view AI ethics as more important than industrial organizations.

9 **AI has gained the attention of the U.S. Congress:** The 116th Congress is the most AI-focused congressional session in history with the number of mentions of AI in congressional record more than triple that of the 115th Congress.

# AI Index Steering Committee

### Co-Directors

Jack Clark
OECD, GPAI

Raymond Perrault
SRI International

### Members

Erik Brynjolfsson
Stanford University

John Etchemendy
Stanford University

Deep Ganguli
Stanford University

Barbara Grosz
Harvard University

Terah Lyons
Partnership on AI

James Manyika
McKinsey Global Institute

Juan Carlos Niebles
Stanford University

Michael Sellitto
Stanford University

Yoav Shoham (Founding Director)
Stanford University, AI21 Labs

# AI Index Staff

### Research Manager and Editor in Chief

Daniel Zhang
Stanford University

### Program Manager

Saurabh Mishra
Stanford University

## How to Cite This Report

Daniel Zhang, Saurabh Mishra, Erik Brynjolfsson, John Etchemendy, Deep Ganguli, Barbara Grosz, Terah Lyons, James Manyika, Juan Carlos Niebles, Michael Sellitto, Yoav Shoham, Jack Clark, and Raymond Perrault, "The AI Index 2021 Annual Report," AI Index Steering Committee, Human-Centered AI Institute, Stanford University, Stanford, CA, March 2021.

The AI Index is an independent initiative at Stanford University's Human-Centered Artificial Intelligence Institute (HAI).

Artificial Intelligence Index

Stanford University
Human-Centered
Artificial Intelligence

The AI Index was conceived within the One Hundred Year Study on AI (AI100).

We thank our supporting partners

McKinsey & Company · Google · OpenAI · genpact · AI21labs · pwc

We welcome feedback and new ideas for next year.
Contact us at AI-Index-Report@stanford.edu.

# Acknowledgments

We appreciate the following organizations and individuals who provided data, analysis, advice, and expert commentary for inclusion in the AI Index 2021 Report:

## Organizations

**arXiv**
Jim Entwood, Paul Ginsparg, Joe Halpern, Eleonora Presani

**AI Ethics Lab**
Cansu Canca, Yasemin Usta

**Black in AI**
Rediet Abebe, Hassan Kane

**Bloomberg Government**
Chris Cornillie

**Burning Glass Technologies**
Layla O'Kane, Bledi Taska, Zhou Zhou

**Computing Research Association**
Andrew Bernat, Susan Davidson

**Elsevier**
Clive Bastin, Jörg Hellwig, Sarah Huggett, Mark Siebert

**Intento**
Grigory Sapunov, Konstantin Savenkov

**International Federation of Robotics**
Susanne Bieller, Jeff Burnstein

**Joint Research Center, European Commission**
Giuditta De Prato, Montserrat López Cobo, Riccardo Righi

**LinkedIn**
Guy Berger, Mar Carpanelli, Di Mo, Virginia Ramsey

**Liquidnet**
Jeffrey Banner, Steven Nichols

**McKinsey Global Institute**
Brittany Presten

**Microsoft Academic Graph**
Iris Shen, Kuansan Wang

**National Institute of Standards and Technology**
Patrick Grother

**Nesta**
Joel Klinger, Juan Mateos-Garcia, Kostas Stathoulopoulos

**NetBase Quid**
Zen Ahmed, Scott Cohen, Julie Kim

**PostEra**
Aaron Morris

**Queer in AI**
Raphael Gontijo Lopes

**State of AI Report**
Nathan Benaich, Ian Hogarth

**Women in Machine Learning**
Sarah Tan, Jane Wang

## Individuals

**ActivityNet**
Fabian Caba (Adobe Research); Bernard Ghanem (King Abdullah University of Science and Technology); Cees Snoek (University of Amsterdam)

**AI Brain Drain and Faculty Departure**
Michael Gofman (University of Rochester); Zhao Jin (Cheung Kong Graduate School of Business)

**Automated Theorem Proving**
Geoff Sutcliffe (University of Miami); Christian Suttner (Connion GmbH)

**Boolean Satisfiability Problem**
Lars Kotthoff (University of Wyoming)

**Corporate Representation at AI Research Conferences**
Nuruddin Ahmed (Ivey Business School, Western University); Muntasir Wahed (Virginia Tech)

**Conference Attendance**
Maria Gini, Gita Sukthankar (AAMAS); Carol Hamilton (AAAI); Dan Jurafsky (ACL); Walter Scheirer, Ramin Zabih (CVPR); Jörg Hoffmann, Erez Karpas (ICAPS); Paul Oh (IROS); Pavlos Peppas, Michael Thielscher (KR)

**Ethics at AI Conferences**
Pedro Avelar, Luis Lamb, Marcelo Prates (Federal University of Rio Grande do Sul)

**ImageNet**
Lucas Beyer, Alexey Dosovitskiy, Neil Houlsby (Google)

**MLPerf/DAWNBench**
Cody Coleman (Stanford University), Peter Mattson (Google)

**Molecular Synthesis**
Philippe Schwaller (IBM Research–Europe)

**Visual Question Answering**
Dhruv Batra, Devi Parikh (Georgia Tech/FAIR); Ayush Shrivastava (Georgia Tech)

**You Only Look Once**
Xiang Long (Baidu)

## Advice and Expert Commentary

Alexey Bochkovskiy; Baidu's PaddlePaddle Computer Vision Team; Chenggang Xu (Cheung Kong Graduate School of Business); Mohammed AlQuraishi (Columbia University); Evan Schnidman (EAS Innovation); Fanghzhen Lin (Hong Kong University of Science and Technology); David Kanter (MLCommons); Sam Bowman (New York University); Maneesh Agrawala, Jeannette Bohg, Emma Brunskill, Chelsea Finn, Aditya Grover, Tatsunori Hashimoto, Dan Jurafsky, Percy Liang, Sharon Zhou (Stanford University); Vamsi Sistla (University of California, Berkeley); Simon King (University of Edinburgh); Ivan Goncharov (Weights & Biases)

## Graduate Researchers

Ankita Banerjea, Yu-chi Tsao (Stanford University)

## Report and Website

Michi Turner (report graphic design and cover art); Nancy King (report editor); Michael Taylor (report data visualization); Kevin Litman-Navarro (Global AI Vibrancy Tool design and development); Travis Taylor (AI Index website design); Digital Avenues (AI Index website development)

# REPORT HIGHLIGHTS

## CHAPTER 1: RESEARCH & DEVELOPMENT

- The number of AI journal publications grew by 34.5% from 2019 to 2020—a much higher percentage growth than from 2018 to 2019 (19.6%).
- In every major country and region, the highest proportion of peer-reviewed AI papers comes from academic institutions. But the second most important originators are different: In the United States, corporate-affiliated research represents 19.2% of the total publications, whereas government is the second most important in China (15.6%) and the European Union (17.2%).
- In 2020, and for the first time, China surpassed the United States in the share of AI journal citations in the world, having briefly overtaken the United States in the overall number of AI journal publications in 2004 and then retaken the lead in 2017. However, the United States has consistently (and significantly) more cited AI conference papers than China over the last decade.
- In response to COVID-19, most major AI conferences took place virtually and registered a significant increase in attendance as a result. The number of attendees across nine conferences almost doubled in 2020.
- In just the last six years, the number of AI-related publications on arXiv grew by more than sixfold, from 5,478 in 2015 to 34,736 in 2020.
- AI publications represented 3.8% of all peer-reviewed scientific publications worldwide in 2019, up from 1.3% in 2011.

## CHAPTER 2: TECHNICAL PERFORMANCE

- **Generative everything:** AI systems can now compose text, audio, and images to a sufficiently high standard that humans have a hard time telling the difference between synthetic and non-synthetic outputs for some constrained applications of the technology. That promises to generate a tremendous range of downstream applications of AI for both socially useful and less useful purposes. It is also causing researchers to invest in technologies for detecting generative models; the DeepFake Detection Challenge data indicates how well computers can distinguish between different outputs.
- **The industrialization of computer vision:** Computer vision has seen immense progress in the past decade, primarily due to the use of machine learning techniques (specifically deep learning). New data shows that computer vision is industrializing: Performance is starting to flatten on some of the largest benchmarks, suggesting that the community needs to develop and agree on harder ones that further test performance. Meanwhile, companies are investing increasingly large amounts of computational resources to train computer vision systems at a faster rate than ever before. Meanwhile, technologies for use in deployed systems—like object-detection frameworks for analysis of still frames from videos—are maturing rapidly, indicating further AI deployment.

- **Natural Language Processing (NLP) outruns its evaluation metrics:** Rapid progress in NLP has yielded AI systems with significantly improved language capabilities that have started to have a meaningful economic impact on the world. Google and Microsoft have both deployed the BERT language model into their search engines, while other large language models have been developed by companies ranging from Microsoft to OpenAI. Progress in NLP has been so swift that technical advances have started to outpace the benchmarks to test for them. This can be seen in the rapid emergence of systems that obtain human level performance on SuperGLUE, an NLP evaluation suite developed in response to earlier NLP progress overshooting the capabilities being assessed by GLUE.
- **New analyses on reasoning:** Most measures of technical problems show for each time point the performance of the best system at that time on a fixed benchmark. New analyses developed for the AI Index offer metrics that allow for an evolving benchmark, and for the attribution to individual systems of credit for a share of the overall performance of a group of systems over time. These are applied to two symbolic reasoning problems, Automated Theorem Proving and Satisfiability of Boolean formulas.
- **Machine learning is changing the game in healthcare and biology:** The landscape of the healthcare and biology industries has evolved substantially with the adoption of machine learning. DeepMind's AlphaFold applied deep learning technique to make a significant breakthrough in the decades-long biology challenge of protein folding. Scientists use ML models to learn representations of chemical molecules for more effective chemical synthesis planning. PostEra, an AI startup used ML-based techniques to accelerate COVID-related drug discovery during the pandemic.

## CHAPTER 3: THE ECONOMY

- "Drugs, Cancer, Molecular, Drug Discovery" received the greatest amount of private AI investment in 2020, with more than USD 13.8 billion, 4.5 times higher than 2019.
- Brazil, India, Canada, Singapore, and South Africa are the countries with the highest growth in AI hiring from 2016 to 2020. Despite the COVID-19 pandemic, the AI hiring continued to grow across sample countries in 2020.
- More private investment in AI is being funneled into fewer startups. Despite the pandemic, 2020 saw a 9.3% increase in the amount of private AI investment from 2019—a higher percentage increase than from 2018 to 2019 (5.7%), though the number of newly funded companies decreased for the third year in a row.
- Despite growing calls to address ethical concerns associated with using AI, efforts to address these concerns in the industry are limited, according to a McKinsey survey. For example, issues such as equity and fairness in AI continue to receive comparatively little attention from companies. Moreover, fewer companies in 2020 view personal or individual privacy risks as relevant, compared with in 2019, and there was no change in the percentage of respondents whose companies are taking steps to mitigate these particular risks.
- Despite the economic downturn caused by the pandemic, half the respondents in a McKinsey survey said that the coronavirus had no effect on their investment in AI, while 27% actually reported increasing their investment. Less than a fourth of businesses decreased their investment in AI.
- The United States recorded a decrease in its share of AI job postings from 2019 to 2020—the first drop in six years. The total number of AI jobs posted in the United States also decreased by 8.2% from 2019 to 2020, from 325,724 in 2019 to 300,999 jobs in 2020.

## CHAPTER 4: AI EDUCATION

- An AI Index survey conducted in 2020 suggests that the world's top universities have increased their investment in AI education over the past four years. The number of courses that teach students the skills necessary to build or deploy a practical AI model on the undergraduate and graduate levels has increased by 102.9% and 41.7%, respectively, in the last four academic years.
- More AI PhD graduates in North America chose to work in industry in the past 10 years, while fewer opted for jobs in academia, according to an annual survey from the Computing Research Association (CRA). The share of new AI PhDs who chose industry jobs increased by 48% in the past decade, from 44.4% in 2010 to 65.7% in 2019. By contrast, the share of new AI PhDs entering academia dropped by 44%, from 42.1% in 2010 to 23.7% in 2019.
- In the last 10 years, AI-related PhDs have gone from 14.2% of the total of CS PhDs granted in the United States, to around 23% as of 2019, according to the CRA survey. At the same time, other previously popular CS PhDs have declined in popularity, including networking, software engineering, and programming languages. Compilers all saw a reduction in PhDs granted relative to 2010, while AI and Robotics/Vision specializations saw a substantial increase.
- After a two-year increase, the number of AI faculty departures from universities to industry jobs in North America dropped from 42 in 2018 to 33 in 2019 (28 of these are tenured faculty and five are untenured). Carnegie Mellon University had the largest number of AI faculty departures between 2004 and 2019 (16), followed by the Georgia Institute of Technology (14) and University of Washington (12).
- The percentage of international students among new AI PhDs in North America continued to rise in 2019, to 64.3%—a 4.3% increase from 2018. Among foreign graduates, 81.8% stayed in the United States and 8.6% have taken jobs outside the United States.
- In the European Union, the vast majority of specialized AI academic offerings are taught at the master's level; robotics and automation is by far the most frequently taught course in the specialized bachelor's and master's programs, while machine learning (ML) dominates in the specialized short courses.

## CHAPTER 5: ETHICAL CHALLENGES OF AI APPLICATIONS

- The number of papers with ethics-related keywords in titles submitted to AI conferences has grown since 2015, though the average number of paper titles matching ethics-related keywords at major AI conferences remains low over the years.
- The five news topics that got the most attention in 2020 related to the ethical use of AI were the release of the European Commission's white paper on AI, Google's dismissal of ethics researcher Timnit Gebru, the AI ethics committee formed by the United Nations, the Vatican's AI ethics plan, and IBM's exiting the facial-recognition businesses.

## CHAPTER 6: DIVERSITY IN AI

- The percentages of female AI PhD graduates and tenure-track computer science (CS) faculty have remained low for more than a decade. Female graduates of AI PhD programs in North America have accounted for less than 18% of all PhD graduates on average, according to an annual survey from the Computing Research Association (CRA). An AI Index survey suggests that female faculty make up just 16% of all tenure-track CS faculty at several universities around the world.
- The CRA survey suggests that in 2019, among new U.S. resident AI PhD graduates, 45% were white, while 22.4% were Asian, 3.2% were Hispanic, and 2.4% were African American.
- The percentage of white (non-Hispanic) new computing PhDs has changed little over the last 10 years, accounting for 62.7% on average. The share of Black or African American (non-Hispanic) and Hispanic computing PhDs in the same period is significantly lower, with an average of 3.1% and 3.3%, respectively.
- The participation in Black in AI workshops, which are co-located with the Conference on Neural Information Processing Systems (NeurIPS), has grown significantly in recent years. The numbers of attendees and submitted papers in 2019 are 2.6 times higher than in 2017, while the number of accepted papers is 2.1 times higher.
- In a membership survey by Queer in AI in 2020, almost half the respondents said they view the lack of inclusiveness in the field as an obstacle they have faced in becoming a practitioner in the AI/ML field. More than 40% of members surveyed said they have experienced discrimination or harassment at work or school.

## CHAPTER 7: AI POLICY AND NATIONAL STRATEGIES

- Since Canada published the world's first national AI strategy in 2017, more than 30 other countries and regions have published similar documents as of December 2020.
- The launch of the Global Partnership on AI (GPAI) and Organisation for Economic Co-operation and Development (OECD) AI Policy Observatory and Network of Experts on AI in 2020 promoted intergovernmental efforts to work together to support the development of AI for all.
- In the United States, the 116th Congress was the most AI-focused congressional session in history. The number of mentions of AI by this Congress in legislation, committee reports, and Congressional Research Service (CRS) reports is more than triple that of the 115th Congress.

# CHAPTER 1:
# Research & Development

Artificial Intelligence
Index Report 2021

## CHAPTER 1:

# Chapter Preview

**ACCESS THE PUBLIC DATA**

# Overview

The report opens with an overview of the research and development (R&D) efforts in artificial intelligence (AI) because R&D is fundamental to AI progress. Since the technology first captured the imagination of computer scientists and mathematicians in the 1950s, AI has grown into a major research discipline with significant commercial applications. The number of AI publications has increased dramatically in the past 20 years. The rise of AI conferences and preprint archives has expanded the dissemination of research and scholarly communications. Major powers, including China, the European Union, and the United States, are racing to invest in AI research. The R&D chapter aims to capture the progress in this increasingly complex and competitive field.

This chapter begins by examining AI publications—from peer-reviewed journal articles to conference papers and patents, including the citation impact of each, using data from the Elsevier/Scopus and Microsoft Academic Graph (MAG) databases, as well as data from the arXiv paper preprint repository and Nesta. It examines contributions to AI R&D from major AI entities and geographic regions and considers how those contributions are shaping the field. The second and third sections discuss R&D activities at major AI conferences and on GitHub.

## CHAPTER HIGHLIGHTS

- The number of AI journal publications grew by 34.5% from 2019 to 2020—a much higher percentage growth than from 2018 to 2019 (19.6%).
- In every major country and region, the highest proportion of peer-reviewed AI papers comes from academic institutions. But the second most important originators are different: In the United States, corporate-affiliated research represents 19.2% of the total publications, whereas government is the second most important in China (15.6%) and the European Union (17.2%).
- In 2020, and for the first time, China surpassed the United States in the share of AI journal citations in the world, having briefly overtaken the United States in the overall number of AI journal publications in 2004 and then retaken the lead in 2017. However, the United States has consistently (and significantly) more cited AI conference papers than China over the last decade.
- In response to COVID-19, most major AI conferences took place virtually and registered a significant increase in attendance as a result. The number of attendees across nine conferences almost doubled in 2020.
- In just the last six years, the number of AI-related publications on arXiv grew by more than sixfold, from 5,478 in 2015 to 34,736 in 2020.
- AI publications represented 3.8% of all peer-reviewed scientific publications worldwide in 2019, up from 1.3% in 2011.

AI publications include peer-reviewed publications, journal articles, conference papers, and patents. To track trends among these publications and to assess the state of AI R&D activities around the world, the following datasets were used: the Elsevier/Scopus database for peer-reviewed publications; the Microsoft Academic Graph (MAG) database for all journals, conference papers, and patent publications; and arXiv and Nesta data for electronic preprints.

# 1.1 PUBLICATIONS

## PEER-REVIEWED AI PUBLICATIONS

This section presents data from the Scopus database by Elsevier. Scopus contains 70 million peer-reviewed research items curated from more than 5,000 international publishers. The 2019 version of the data shown below is derived from an entirely new set of publications, so figures of all peer-reviewed AI publications differ from those in previous years' AI Index reports. Due to changes in the methodology for indexing publications, the accuracy of the dataset increased from 80% to 84% (see the Appendix for more details).

### Overview

Figure 1.1.1a shows the number of peer-reviewed AI publications, and Figure 1.1.1b shows the share of those among all peer-reviewed publications in the world. The total number of publications grew by nearly 12 times between 2000 and 2019. Over the same period, the percentage of peer-reviewed publications increased from 0.82% of all publications in 2000 to 3.8% in 2019.

### By Region[1]

Among the total number of peer-reviewed AI publications in the world, East Asia & Pacific has held the largest share since 2004, followed by Europe & Central Asia, and North America (Figure 1.1.2). Between 2009 and 2019, South Asia and sub-Saharan Africa experienced the highest growth in terms of the number of peer-reviewed AI publications, increasing by eight- and sevenfold, respectively.

NUMBER of PEER-REVIEWED AI PUBLICATIONS, 2000-19

Source: Elsevier/Scopus, 2020 | Chart: 2021 AI Index Report

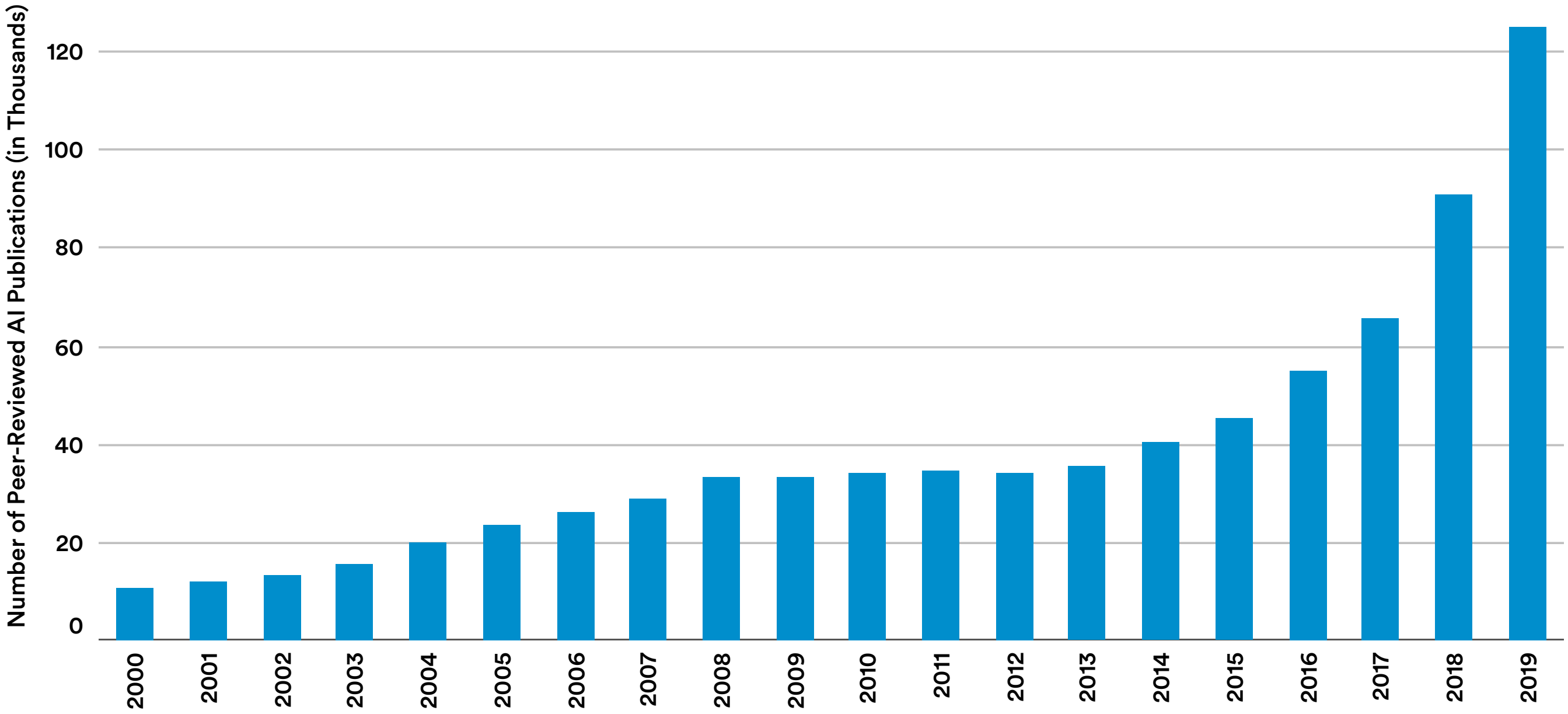

Figure 1.1.1a

1 Regions in this chapter are classified according to the World Bank analytical grouping.

## PEER-REVIEWED AI PUBLICATIONS (% of TOTAL), 2000-19

Source: Elsevier/Scopus, 2020 | Chart: 2021 AI Index Report

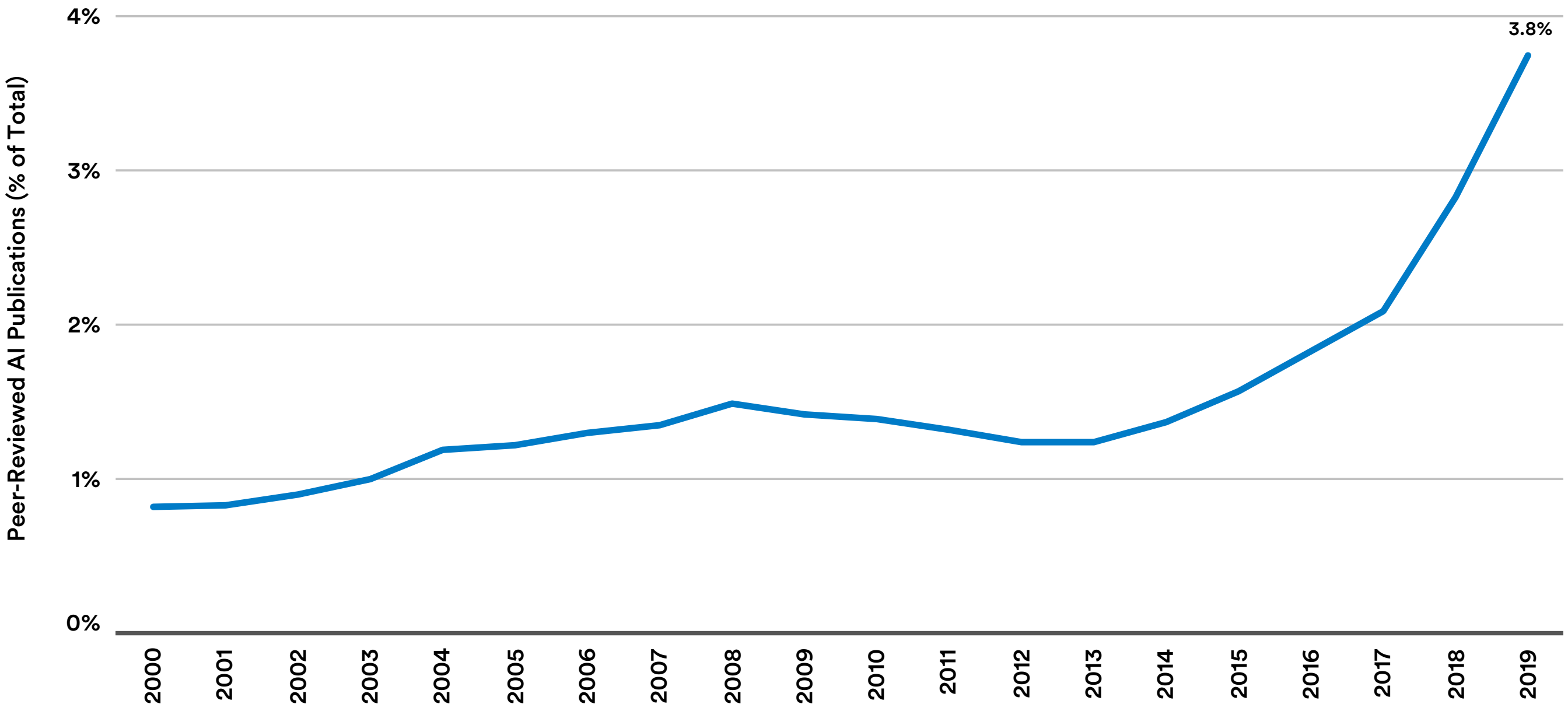

Figure 1.1.1b

## PEER-REVIEWED AI PUBLICATIONS (% of TOTAL) by REGION, 2000-19

Source: Microsoft Academic Graph, 2020 | Chart: 2021 AI Index Report

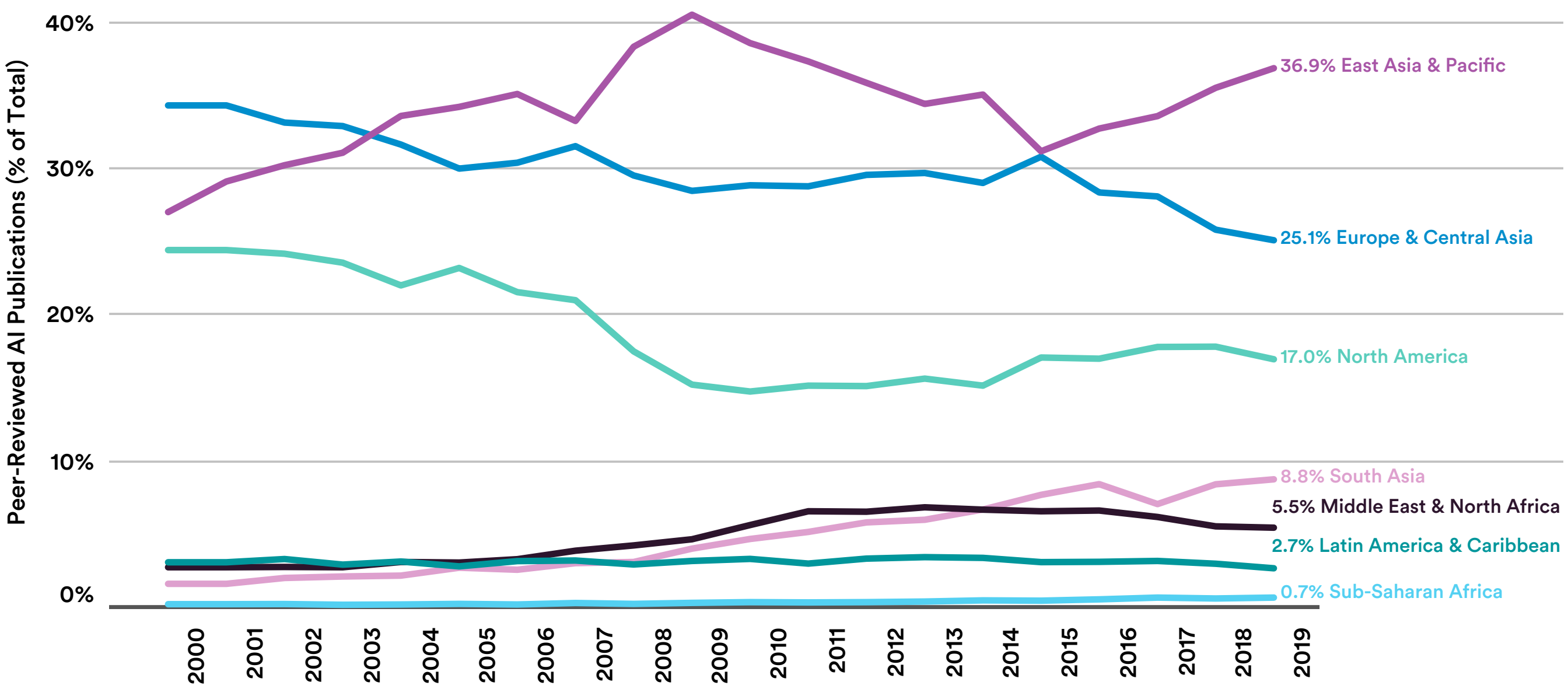

Figure 1.1.2

## By Geographic Area

To compare the activity among the world's major AI players, this section shows trends of peer-reviewed AI publications coming out of China, the European Union, and the United States. As of 2019, China led in the share of peer-reviewed AI publications in the world, after overtaking the European Union in 2017 (Figure 1.1.3). It published 3.5 times more peer-reviewed AI papers in 2019 than it did in 2014—while the European Union published just 2 times more papers and the United States 2.75 times more over the same period.

PEER-REVIEWED AI PUBLICATIONS (% of WORLD TOTAL) by GEOGRAPHIC AREA, 2000-19
Source: Elsevier/Scopus, 2020 | Chart: 2021 AI Index Report

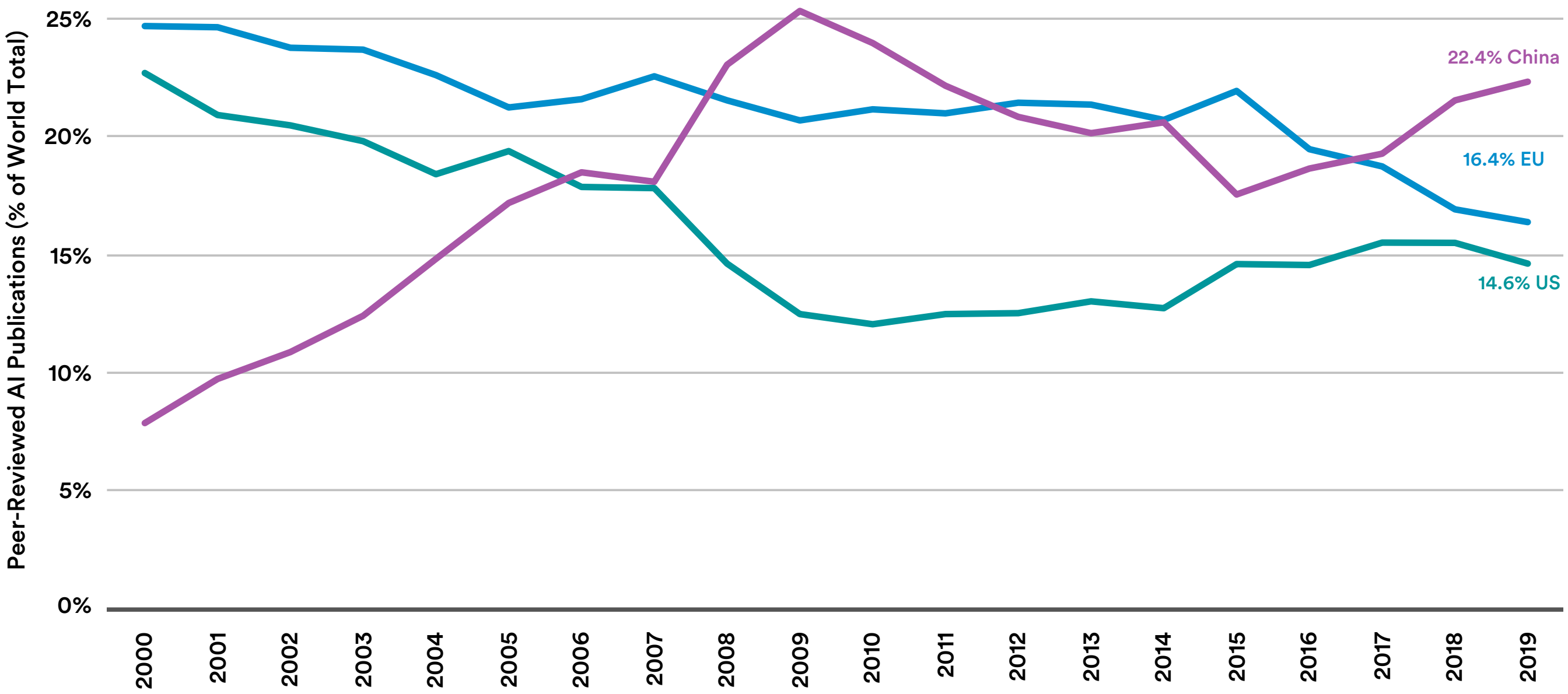

Figure 1.1.3

## By Institutional Affiliation

The following charts show the number of peer-reviewed AI publications affiliated with corporate, government, medical, and other institutions in China (Figure 1.1.4a), the European Union (Figure 1.1.4b), and the United States (Figure 1.1.4c).[2] In 2019, roughly 95.4% of overall peer-reviewed AI publications in China were affiliated with the academic field, compared with 81.9% in the European Union and 89.6% in the United States. Those affiliation categories are not mutually exclusive, as some authors could be affiliated with more than one type of institution.

The data suggests that, excluding academia, government institutions—more than those in other categories—consistently contribute the highest percentage of peer-reviewed AI publications in both China and the European Union (15.6% and 17.2 %, respectively, in 2019), while in the United States, the highest portion is corporate-affiliated (19.2%).

NUMBER of PEER-REVIEWED AI PUBLICATIONS in CHINA by INSTITUTIONAL AFFILIATION, 2000-19

Source: Elsevier/Scopus, 2020 | Chart: 2021 AI Index Report

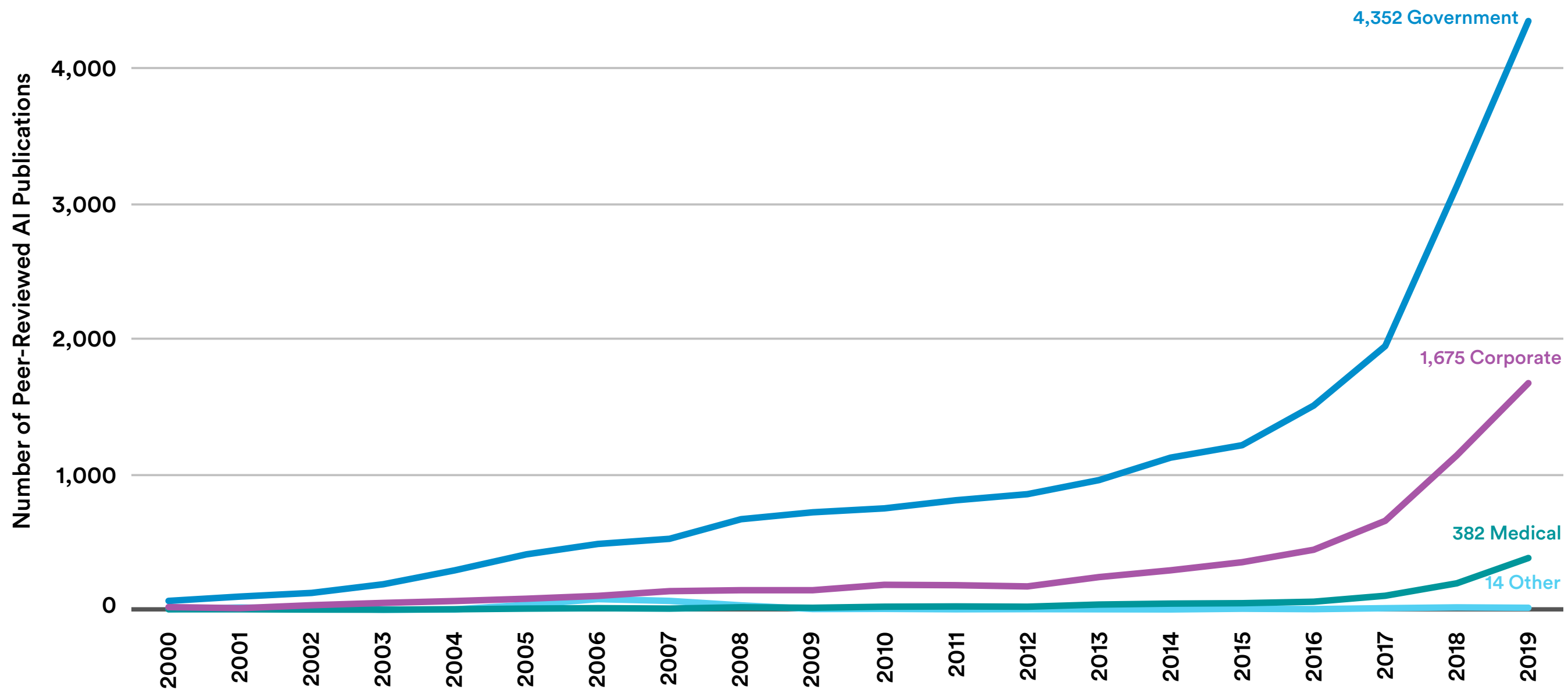

Figure 1.1.4a

2 Across all three geographic areas, the number of papers affiliated with academia exceeds that of government-, corporate-, and medical-affiliated ones; therefore, the academia affiliation is not shown, as it would distort the graphs.

NUMBER of PEER-REVIEWED AI PUBLICATIONS in the EUROPEAN UNION by INSTITUTIONAL AFFILIATION, 2000-19
Source: Elsevier/Scopus, 2020 | Chart: 2021 AI Index Report

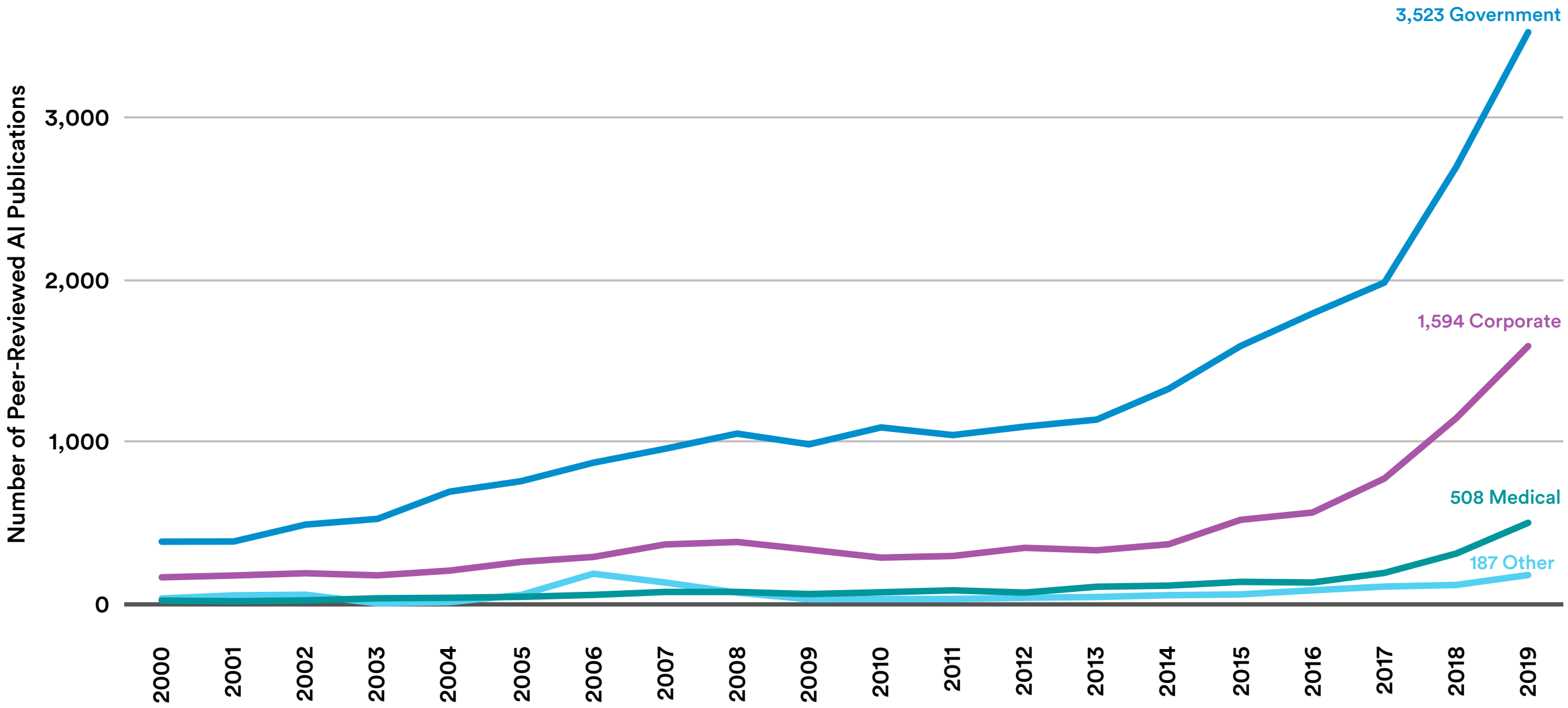

Figure 1.1.4b

NUMBER of PEER-REVIEWED AI PUBLICATIONS in the UNITED STATES by INSTITUTIONAL AFFILIATION, 2000-19
Source: Elsevier/Scopus, 2020 | Chart: 2021 AI Index Report

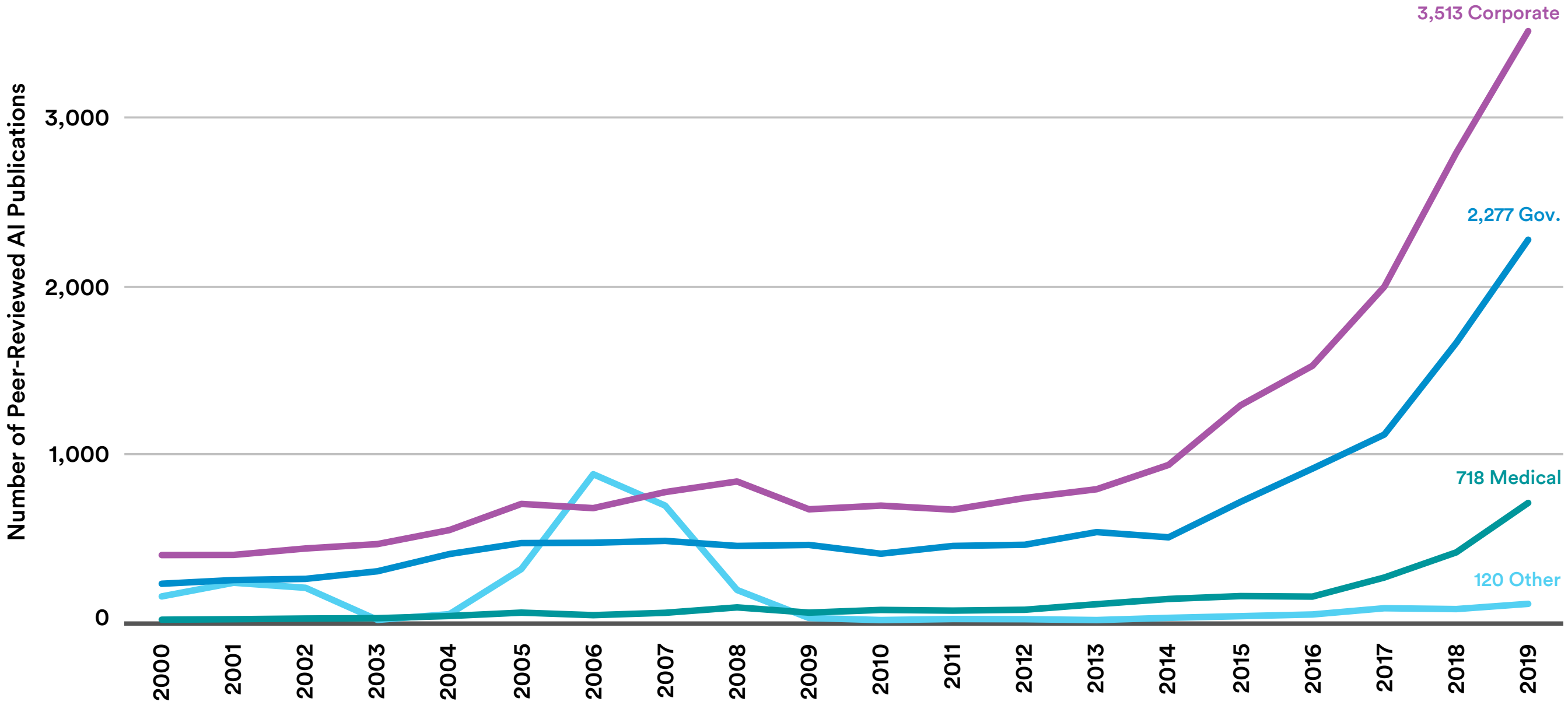

Figure 1.1.4c

## Academic-Corporate Collaboration

Since the 1980s, the R&D collaboration between academia and industry in the United States has grown in importance and popularity, made visible by the proliferation of industry-university research centers as well as corporate contributions to university research. Figure 1.1.5 shows that between 2015 and 2019, the United States produced the highest number of hybrid academic-corporate, co-authored, peer-reviewed AI publications—more than double the amount in the European Union, which comes in second, followed by China in third place.

**NUMBER of ACADEMIC-CORPORATE PEER-REVIEWED AI PUBLICATIONS by GEOGRAPHIC AREA, 2015-19 (SUM)**
Source: Elsevier/Scopus, 2020 | Chart: 2021 AI Index Report

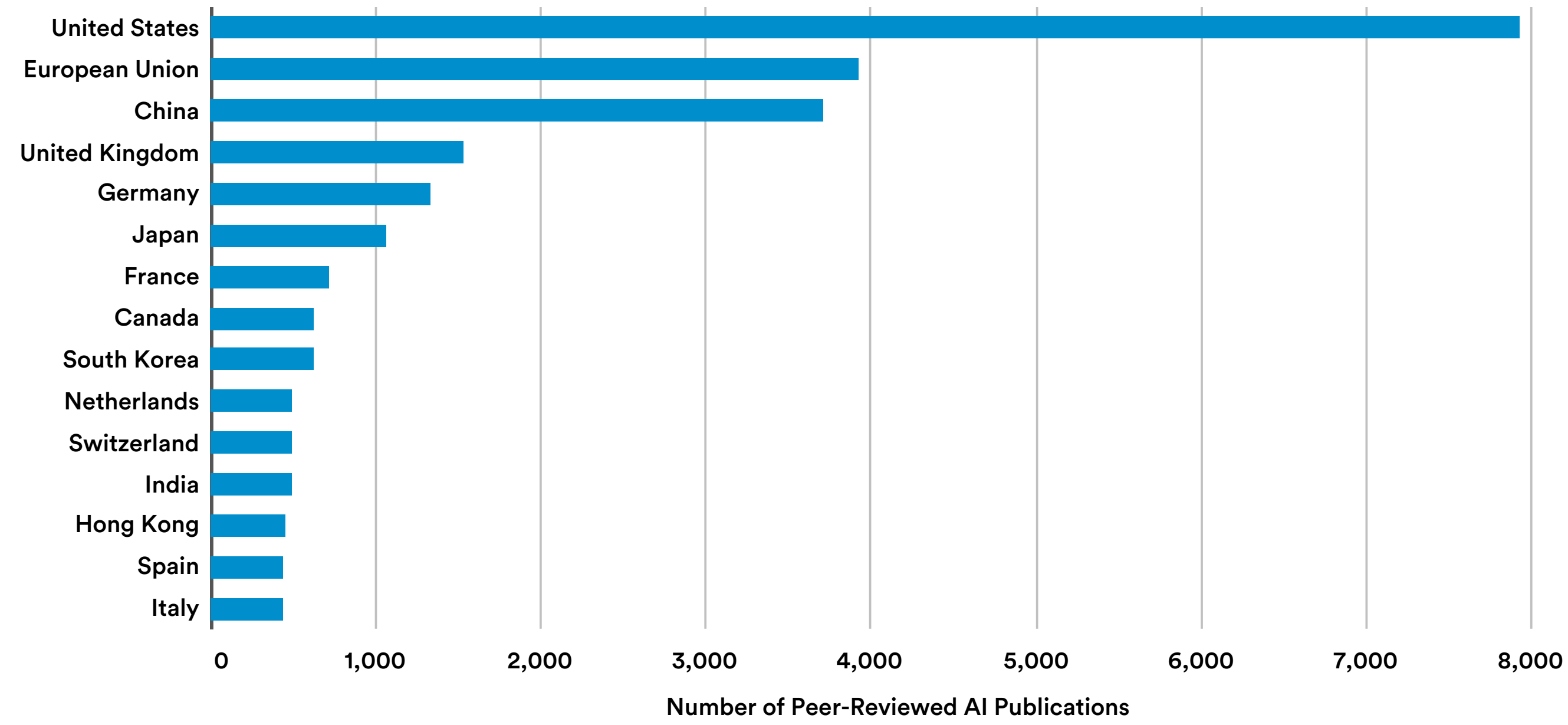

Figure 1.1.5

To assess how academic-corporate collaborations impact the Field-Weighted Citation Impact (FWCI) of AI publications from different geographic regions, see Figure 1.1.6. FWCI measures how the number of citations received by publications compares with the average number of citations received by other similar publications in the same year, discipline, and format (book, article, conference paper, etc.). A value of 1.0 represents the world average. More than or less than 1 means publications are cited more or less than expected, according to the world average. For example, an FWCI of 0.75 means 25% fewer citations than the world average.

The chart shows the FWCI for all peer-reviewed AI publications on the y-axis and the total number (on a log scale) of academic-corporate co-authored publications on the x-axis. To increase the signal-to-noise ratio of the FWCI metric, only countries that have more than 1,000 peer-reviewed AI publications in 2020 are included.

**PEER-REVIEWED AI PUBLICATIONS' FIELD-WEIGHTED CITATION IMPACT and NUMBER of ACADEMIC-CORPORATE PEER-REVIEWED AI PUBLICATIONS, 2019**

Source: Elsevier/Scopus, 2020 | Chart: 2021 AI Index Report

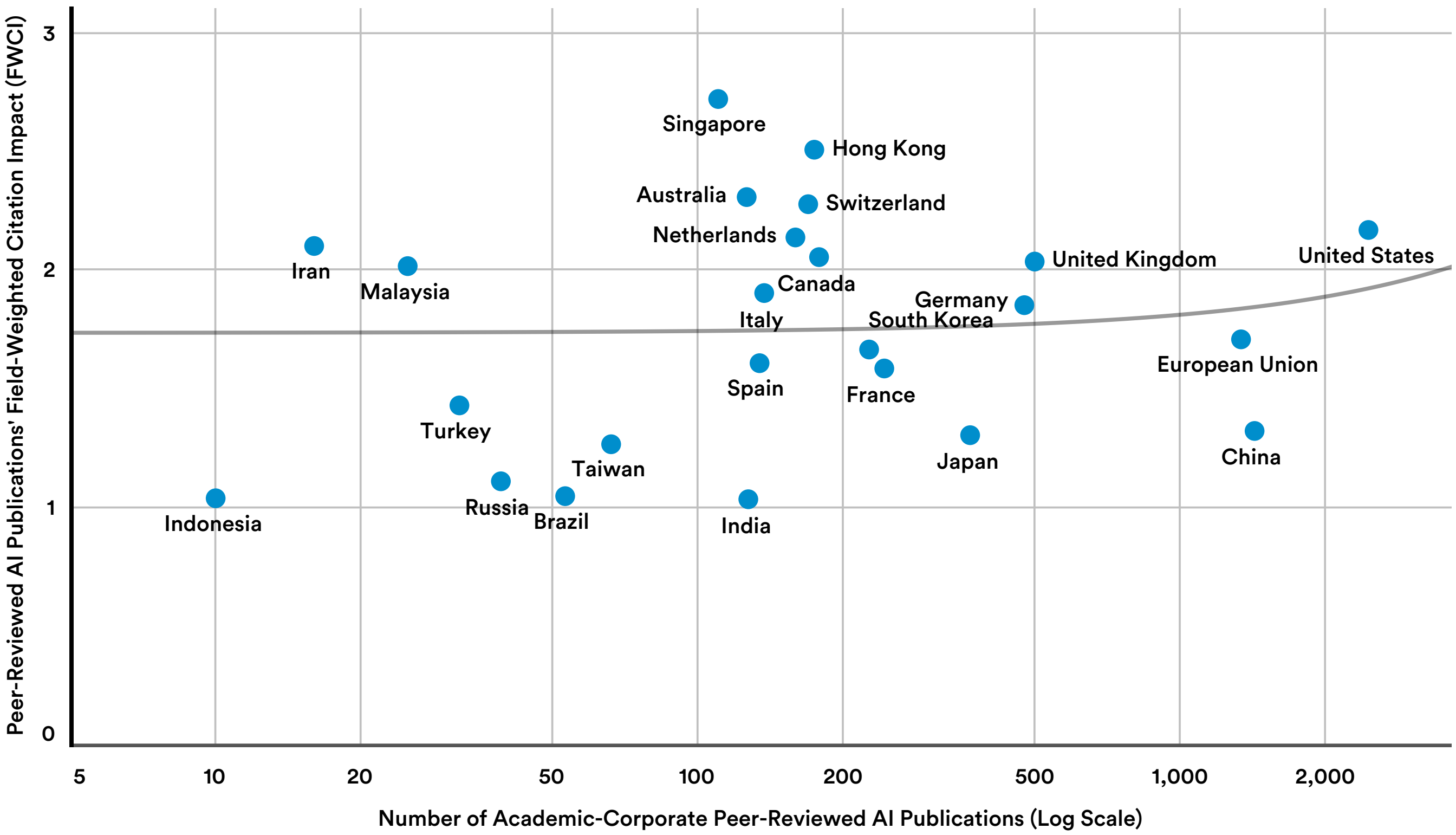

Figure 1.1.6

## AI JOURNAL PUBLICATIONS

The next three sections chart the trends in the publication of AI journals, conference publications, and patents, as well as their respective citations that provide a signal for R&D impact, based on data from Microsoft Academic Graph. MAG[3] is a knowledge graph consisting of more than 225 million publications (at the end of November 2019).

### Overview

Overall, the number of AI journal publications in 2020 is 5.4 times higher than it was in 2000 (Figure 1.1.7a). In 2020, the number of AI journal publications increased by 34.5% from 2019—a much higher percentage growth than from 2018 to 2019 (19.6%). Similarly, the share of AI journal publications among all publications in the world has jumped by 0.4 percentage points in 2020, higher than the average of 0.03 percentage points in the past five years (Figure 1.1.7b).

NUMBER of AI JOURNAL PUBLICATIONS, 2000-20

Source: Microsoft Academic Graph, 2020 | Chart: 2021 AI Index Report

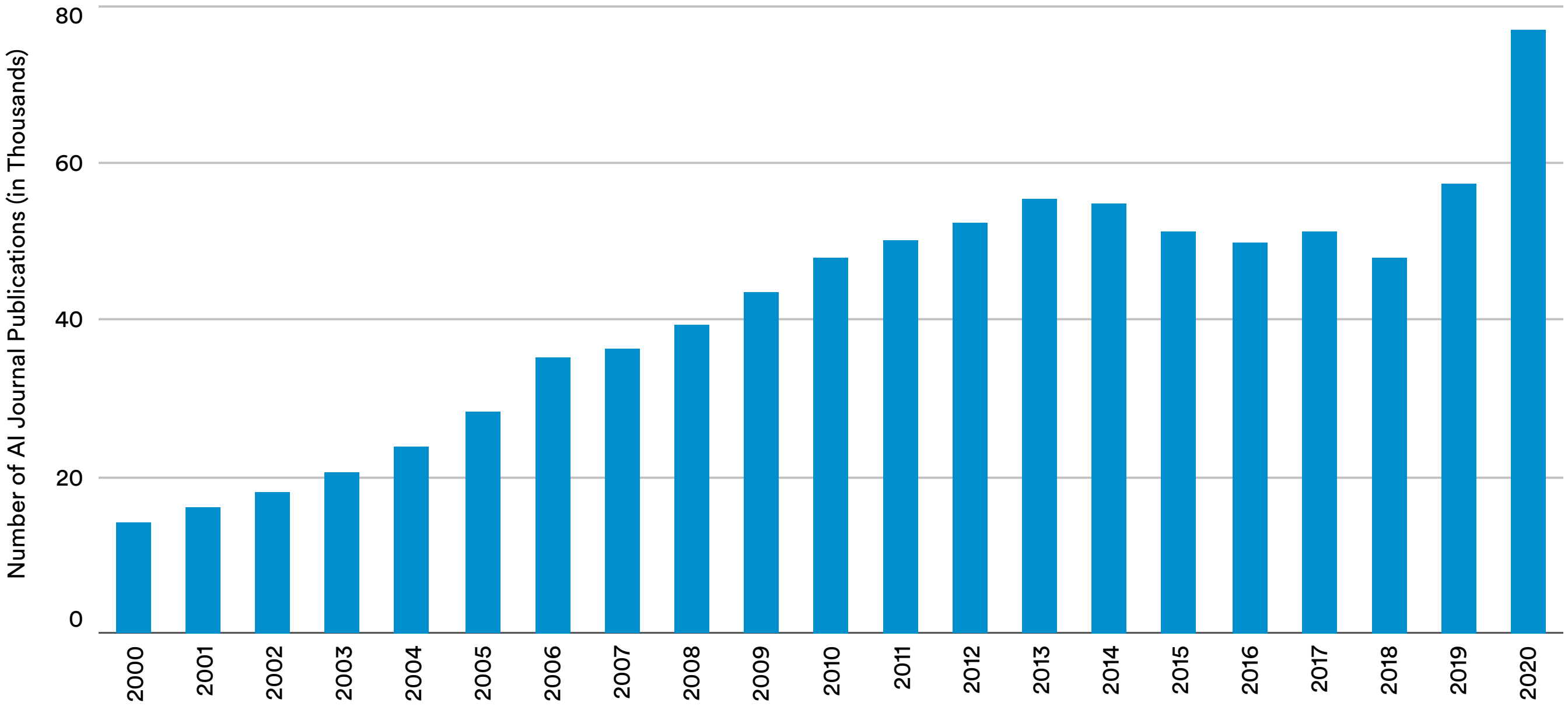

Figure 1.1.7a

AI JOURNAL PUBLICATIONS (% of ALL JOURNAL PUBLICATIONS), 2000-20

Source: Microsoft Academic Graph, 2020 | Chart: 2021 AI Index Report

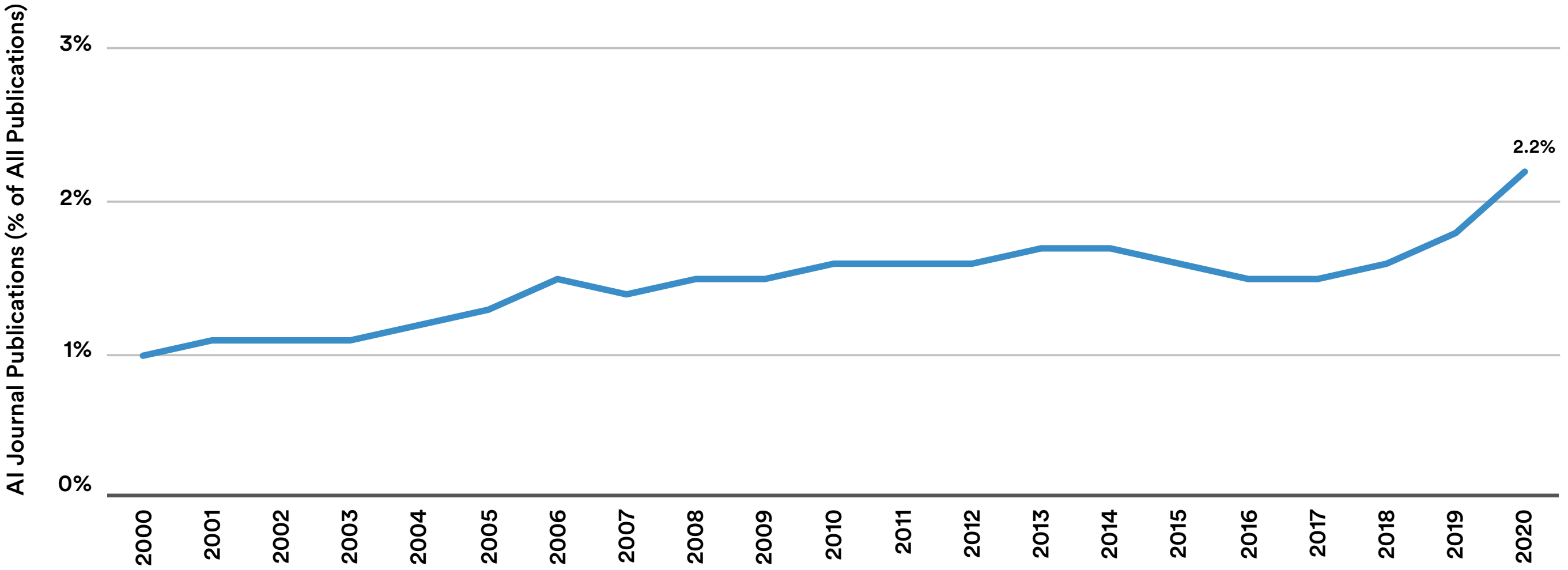

Figure 1.1.7b

3 See “An Overview of Microsoft Academic Service (MAS) and Applications” and “A Review of Microsoft Academic Services for Science of Science Studies” for more details.

## By Region

Figure 1.1.8 shows the share of AI journals—the dominant publication entity in terms of numbers in the MAG database—by region between 2000 and 2020. East Asia & Pacific, Europe & Central Asia, and North America are responsible for the majority of AI journal publications in the past 21 years, while the lead position among the three regions changes over time. In 2020, East Asia & Pacific held the highest share (26.7%), followed by Europe & Central Asia (13.3%) and North America (14.0%). Additionally, in the last 10 years, South Asia, and Middle East & North Africa saw the most significant growth, as the number of AI journal publications in those two regions grew six- and fourfold, respectively.

AI JOURNAL PUBLICATIONS (% of WORLD TOTAL) by REGION, 2000-20

Source: Microsoft Academic Graph, 2020 | Chart: 2021 AI Index Report

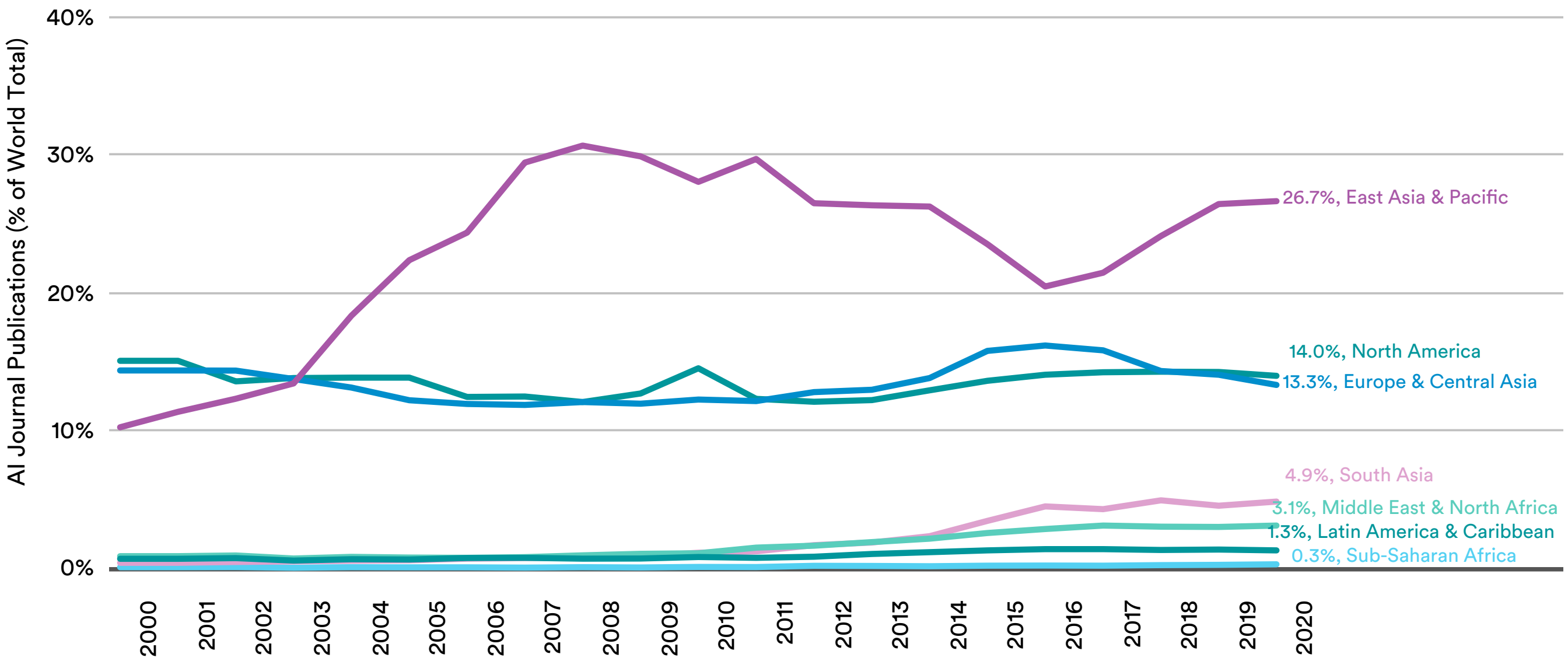

Figure 1.1.8

## By Geographic Area

Figure 1.1.9 shows that among the three major AI powers, China has had the largest share of AI journal publications in the world since 2017, with 18.0% in 2020, followed by the United States (12.3%) and the European Union (8.6%).

### AI JOURNAL PUBLICATIONS (% of WORLD TOTAL) by GEOGRAPHIC AREA, 2000-20

Source: Microsoft Academic Graph, 2020 | Chart: 2021 AI Index Report

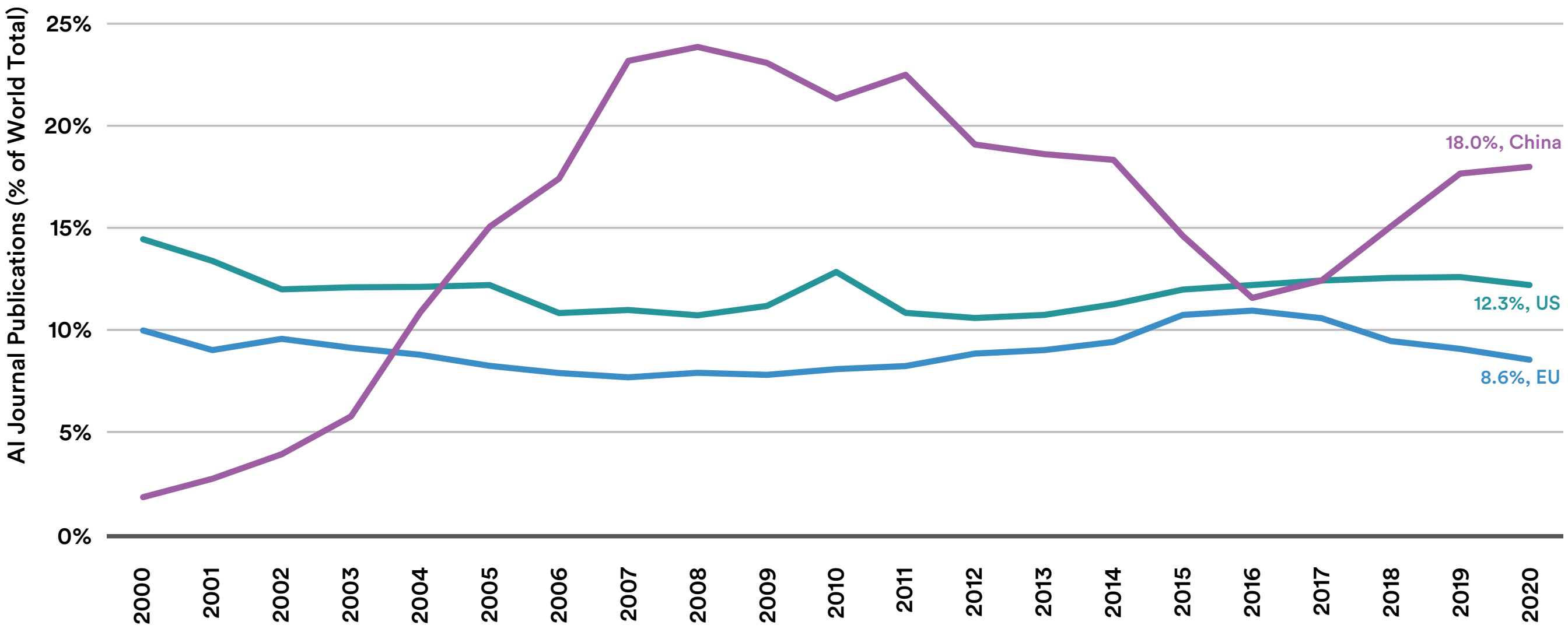

Figure 1.1.9

## Citation

In terms of the highest share of AI journal citations, Figure 1.1.10 shows that China (20.7%) overtook the United States (19.8%) in 2020 for the first time, while the European Union continued to lose overall share.

### AI JOURNAL CITATIONS (% of WORLD TOTAL) by GEOGRAPHIC AREA, 2000-20

Source: Microsoft Academic Graph, 2020 | Chart: 2021 AI Index Report

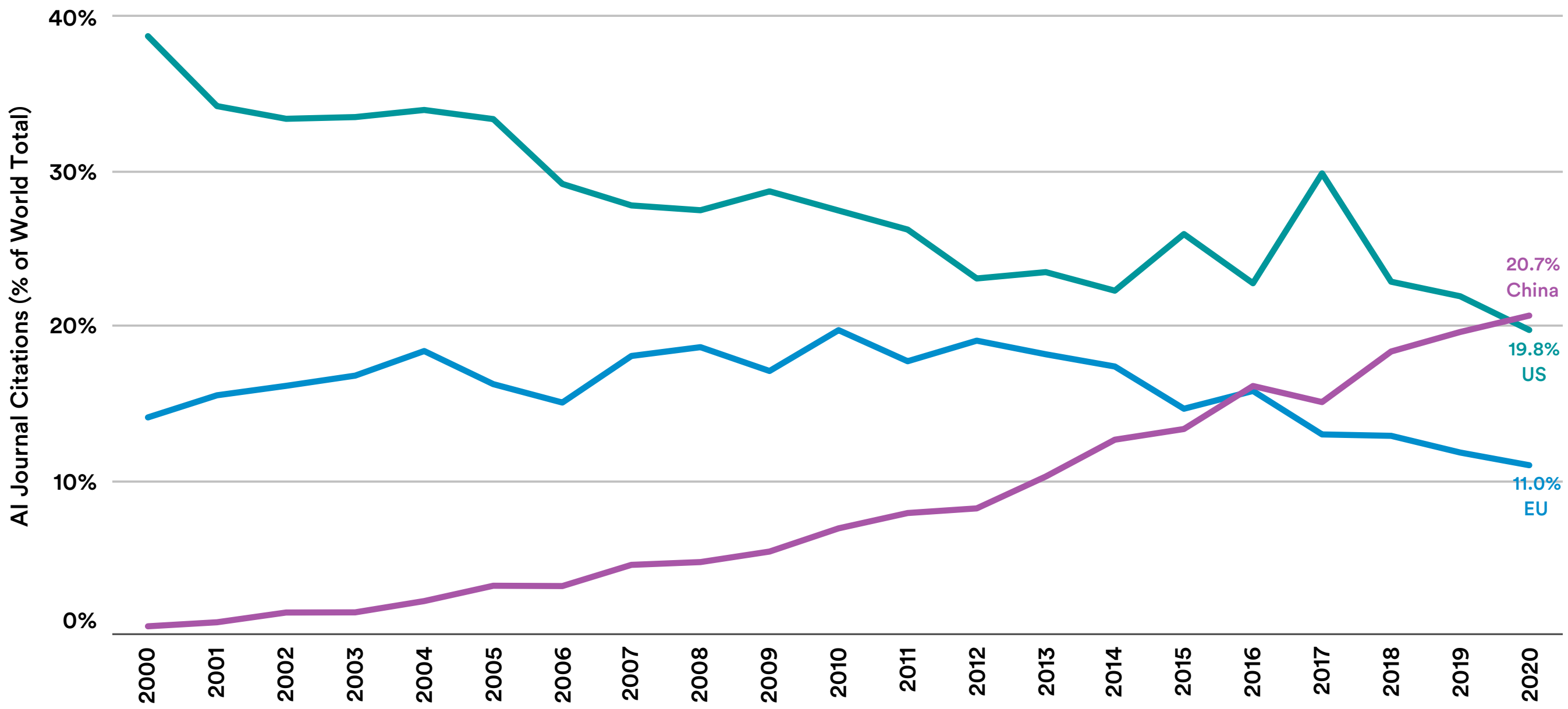

Figure 1.1.10

## AI CONFERENCE PUBLICATIONS

### Overview

Between 2000 and 2019, the number of AI conference publications increased fourfold, although the growth flattened out in the past ten years, with the number of publications in 2019 just 1.09 times higher than the number in 2010.[4]

**NUMBER of AI CONFERENCE PUBLICATIONS, 2000-20**

Source: Microsoft Academic Graph, 2020 | Chart: 2021 AI Index Report

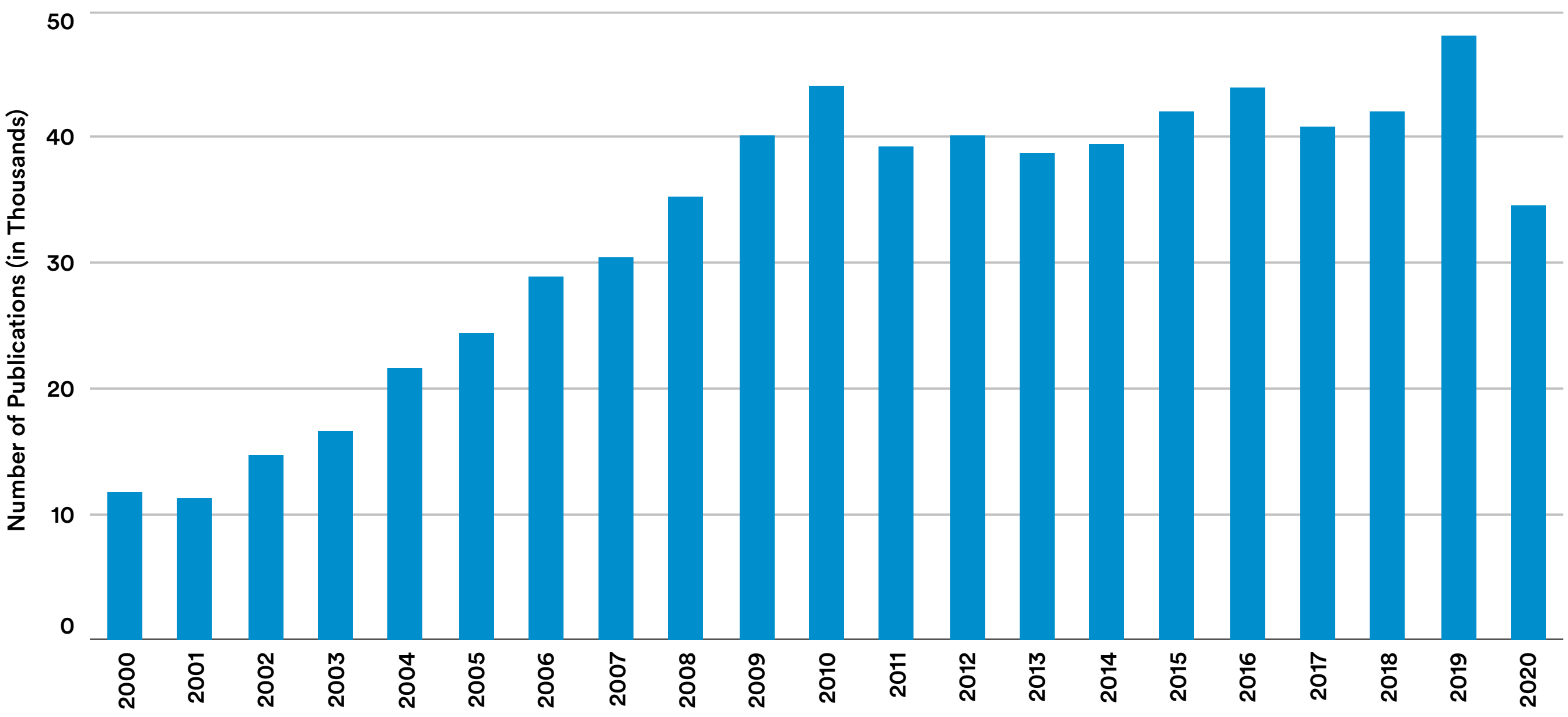

Figure 1.1.11a

**AI CONFERENCE PUBLICATIONS (% of ALL CONFERENCE PUBLICATIONS), 2000-20**

Source: Microsoft Academic Graph, 2020 | Chart: 2021 AI Index Report

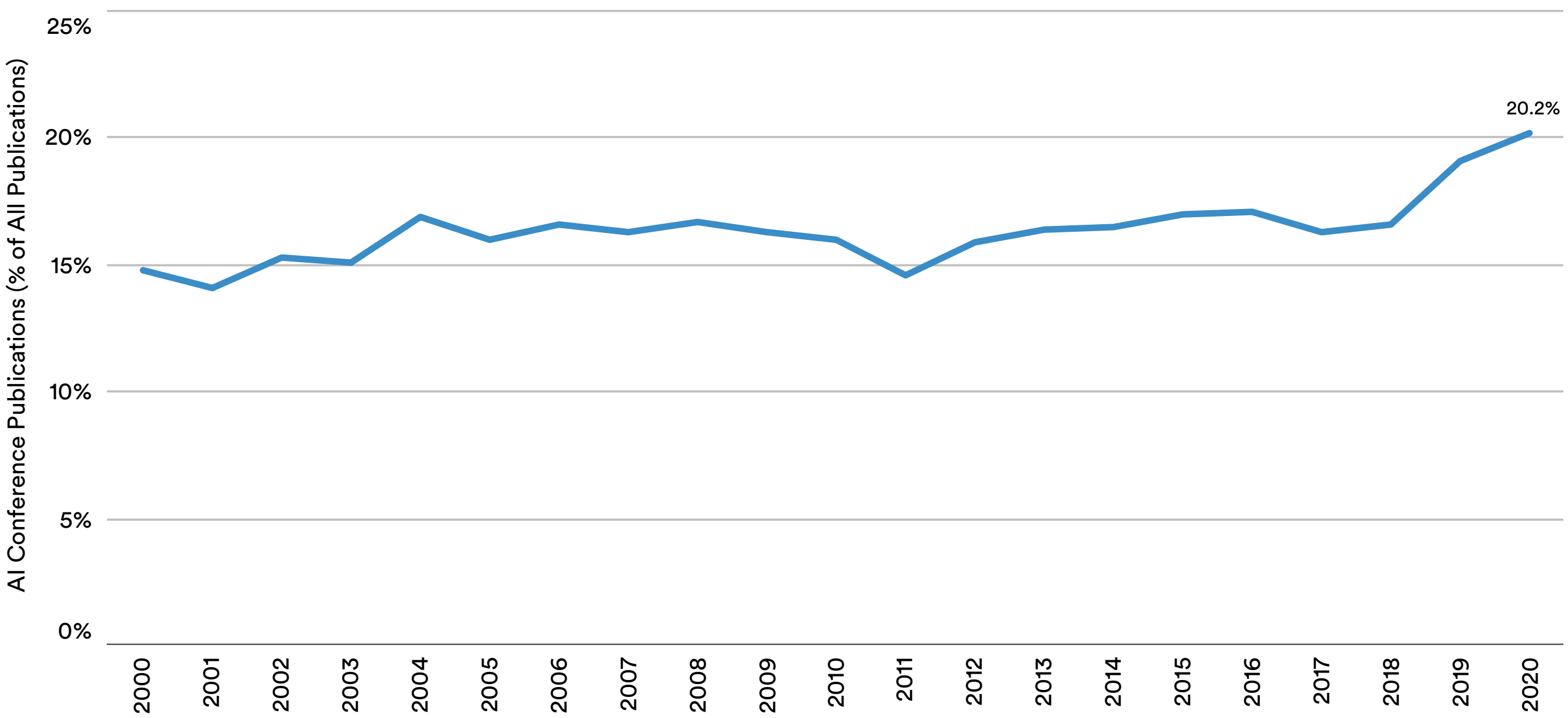

Figure 1.1.11b

4 Note that conference data in 2020 on the MAG system is not yet complete. See the Appendix for details.

## By Region

Figure 1.1.12 shows that, similar to the trends in AI journal publication, East Asia & Pacific, Europe & Central Asia, and North America are the world's dominant sources for AI conference publications. Specifically, East Asia & Pacific took the lead starting in 2004, accounting for more than 27% in 2020. North America overtook Europe & Central Asia to claim second place in 2018, accounting for 20.1%, followed by 21.7% in 2020.

### AI CONFERENCE PUBLICATIONS (% of WORLD TOTAL) by REGION, 2000-20

Source: Microsoft Academic Graph, 2020 | Chart: 2021 AI Index Report

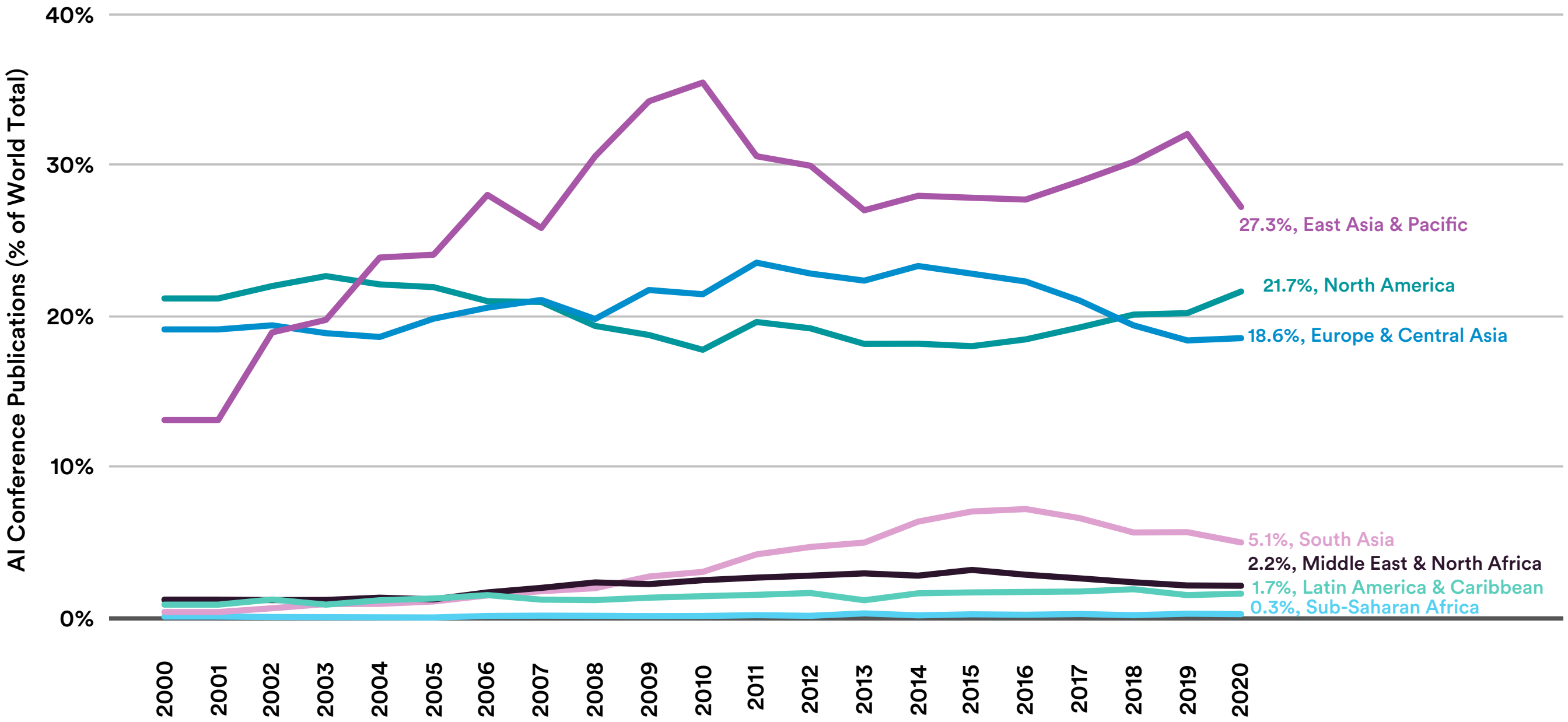

Figure 1.1.12

## By Geographic Area

China overtook the United States in the share of AI conference publications in the world in 2019 (Figure 1.1.13). Its share has grown significantly since 2000. China's percentage of AI conference publications in 2019 is almost nine times higher than it was in 2000. The share of conference publications for the European Union peaked in 2011 and continues to decline.

**AI CONFERENCE PUBLICATIONS (% of WORLD TOTAL) by GEOGRAPHIC AREA, 2000-20**

Source: Microsoft Academic Graph, 2020 | Chart: 2021 AI Index Report

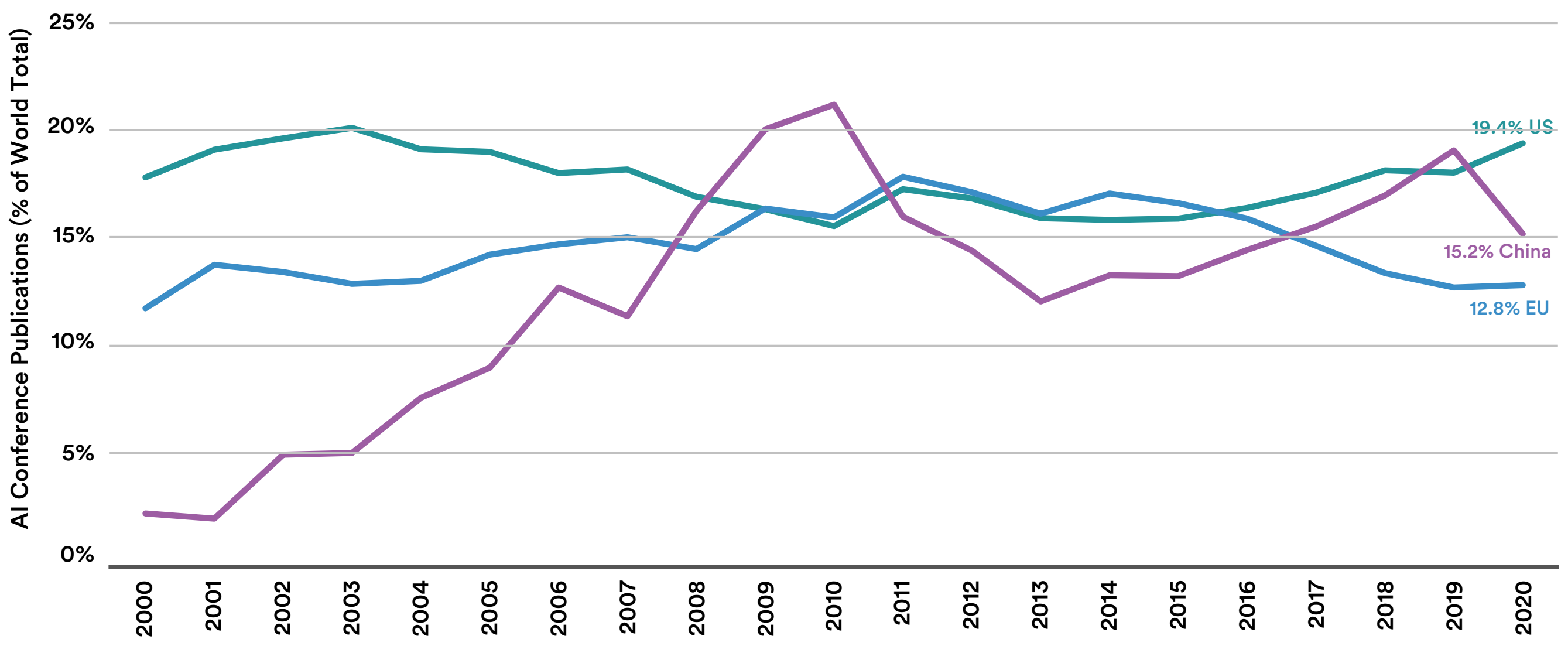

Figure 1.1.13

## Citation

With respect to citations of AI conference publications, Figure 1.1.14 shows that the United States has held a dominant lead among the major powers over the past 21 years. The United States tops the list with 40.1% of overall citations in 2020, followed by China (11.8%) and the European Union (10.9%).

**AI CONFERENCE CITATIONS (% of WORLD TOTAL) by GEOGRAPHIC AREA, 2000-20**

Source: Microsoft Academic Graph, 2020 | Chart: 2021 AI Index Report

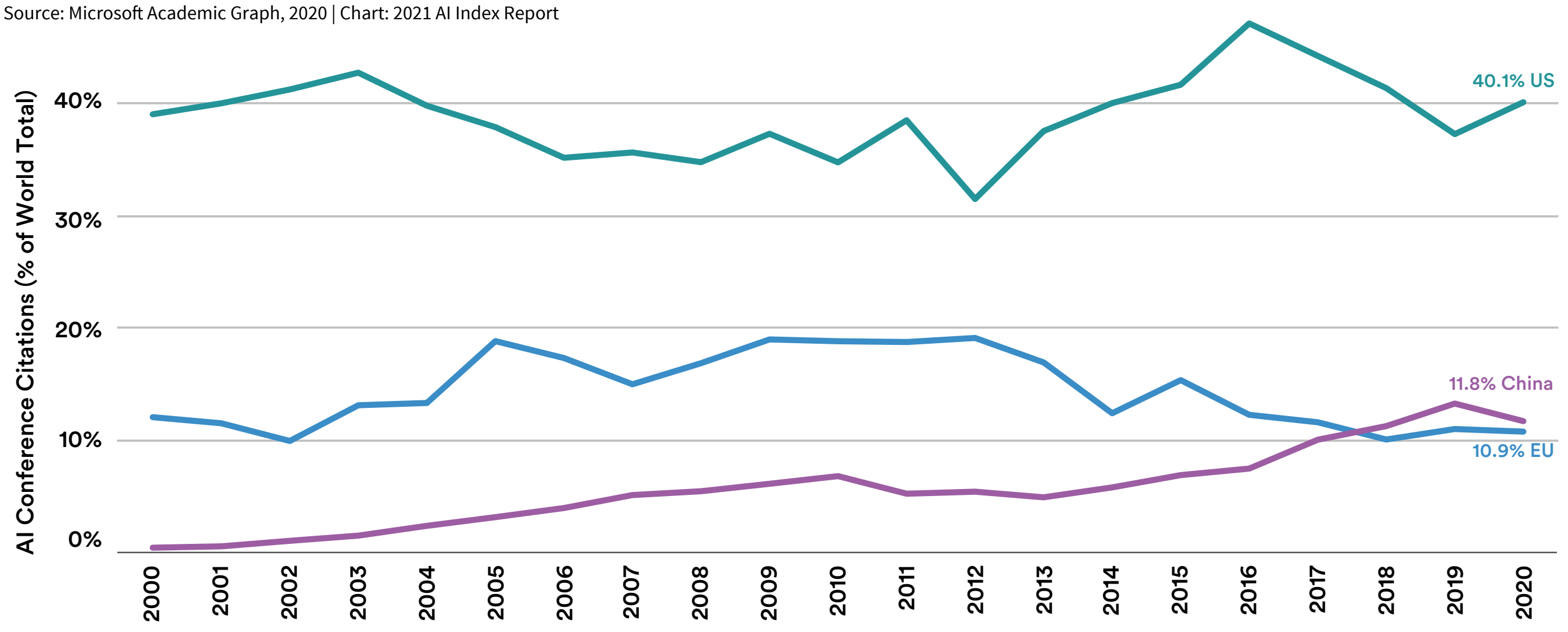

Figure 1.1.14

# AI PATENTS

## Overview

The total number of AI patents published in the world has been steadily increasing in the past two decades, growing from 21,806 in 2000 to more than 4.5 times that, or 101,876, in 2019 (Figure 1.1.15a). The share of AI patents published in the world exhibits a lesser increase, from around 2% in 2000 to 2.9% in 2020 (Figure 1.1.15b). The AI patent data is incomplete—only 8% of the dataset in 2020 includes a country or regional affiliation. There is reason to question the data on the share of AI patent publications by both region and geographic area, and it is therefore not included in the main report. See the Appendix for details.

**NUMBER of AI PATENT PUBLICATIONS, 2000-20**

Source: Microsoft Academic Graph, 2020 | Chart: 2021 AI Index Report

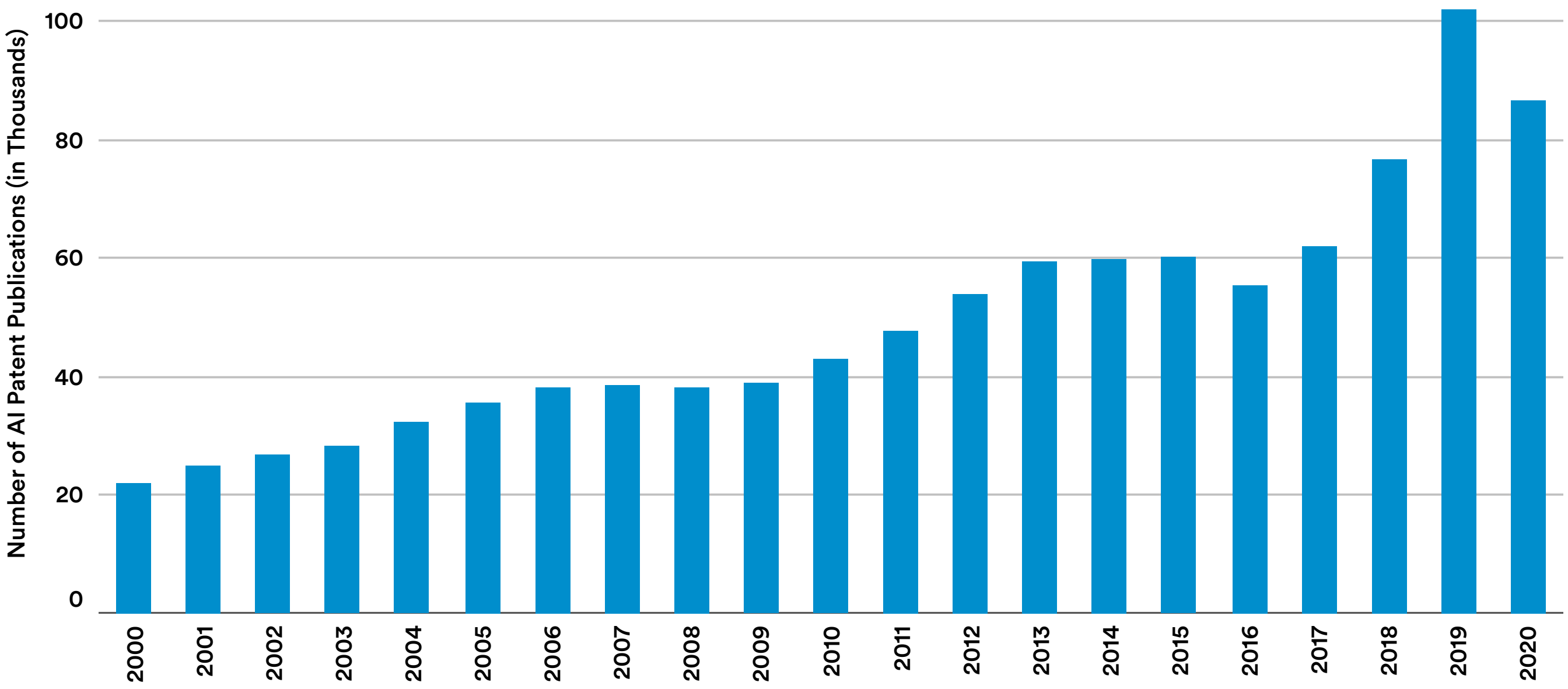

Figure 1.1.15a

**AI PATENT PUBLICATIONS (% of ALL PATENT PUBLICATIONS), 2000-20**

Source: Microsoft Academic Graph, 2020 | Chart: 2021 AI Index Report

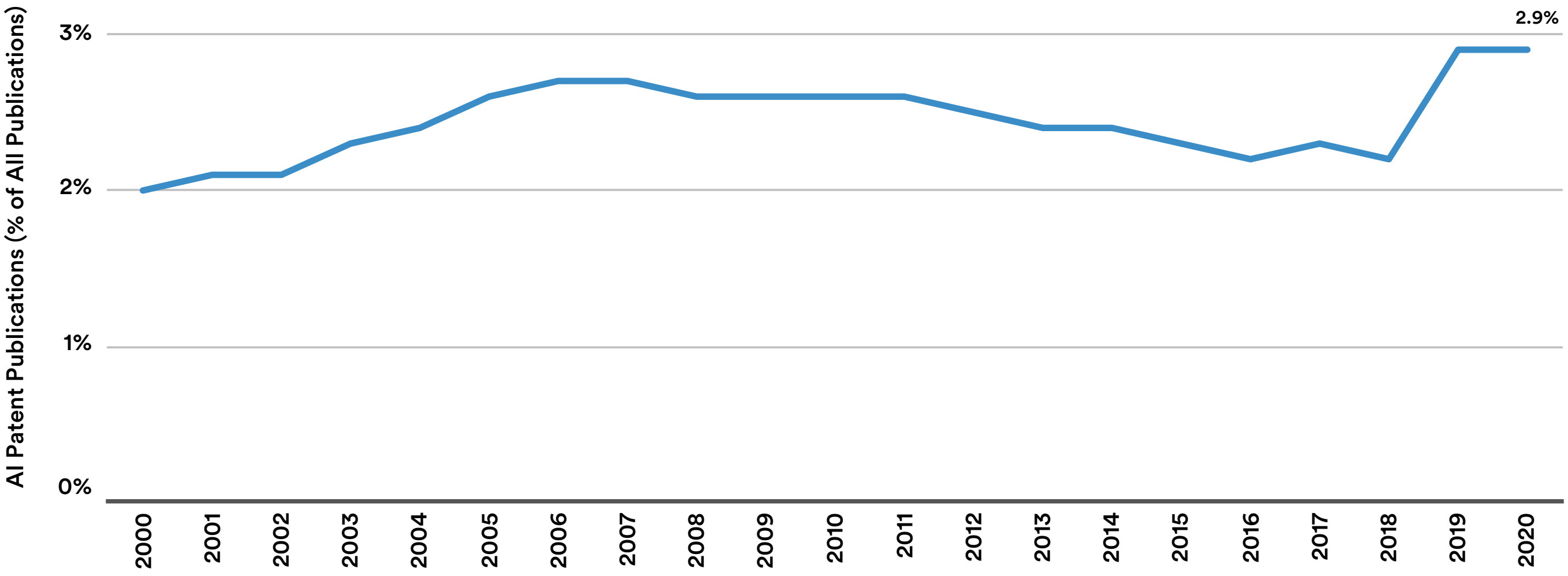

Figure 1.1.15b

## ARXIV PUBLICATIONS

In addition to the traditional avenues for publishing academic papers (discussed above), AI researchers have embraced the practice of publishing their work (often pre–peer review) on arXiv, an online repository of electronic preprints. arXiv allows researchers to share their findings before submitting them to journals and conferences, which greatly accelerates the cycle of information discovery and dissemination. The number of AI-related publications in this section includes preprints on arXiv under cs.AI (artificial intelligence), cs.CL (computation and language), cs.CV (computer vision), cs.NE (neural and evolutionary computing), cs.RO (robotics), cs.LG (machine learning in computer science), and stat.ML (machine learning in statistics).

### Overview

In just six years, the number of AI-related publications on arXiv grew more than sixfold, from 5,478 in 2015 to 34,736 in 2020 (Figure 1.1.16).

NUMBER of AI-RELATED PUBLICATIONS on ARXIV, 2015-20

Source: arXiv, 2020 | Chart: 2021 AI Index Report

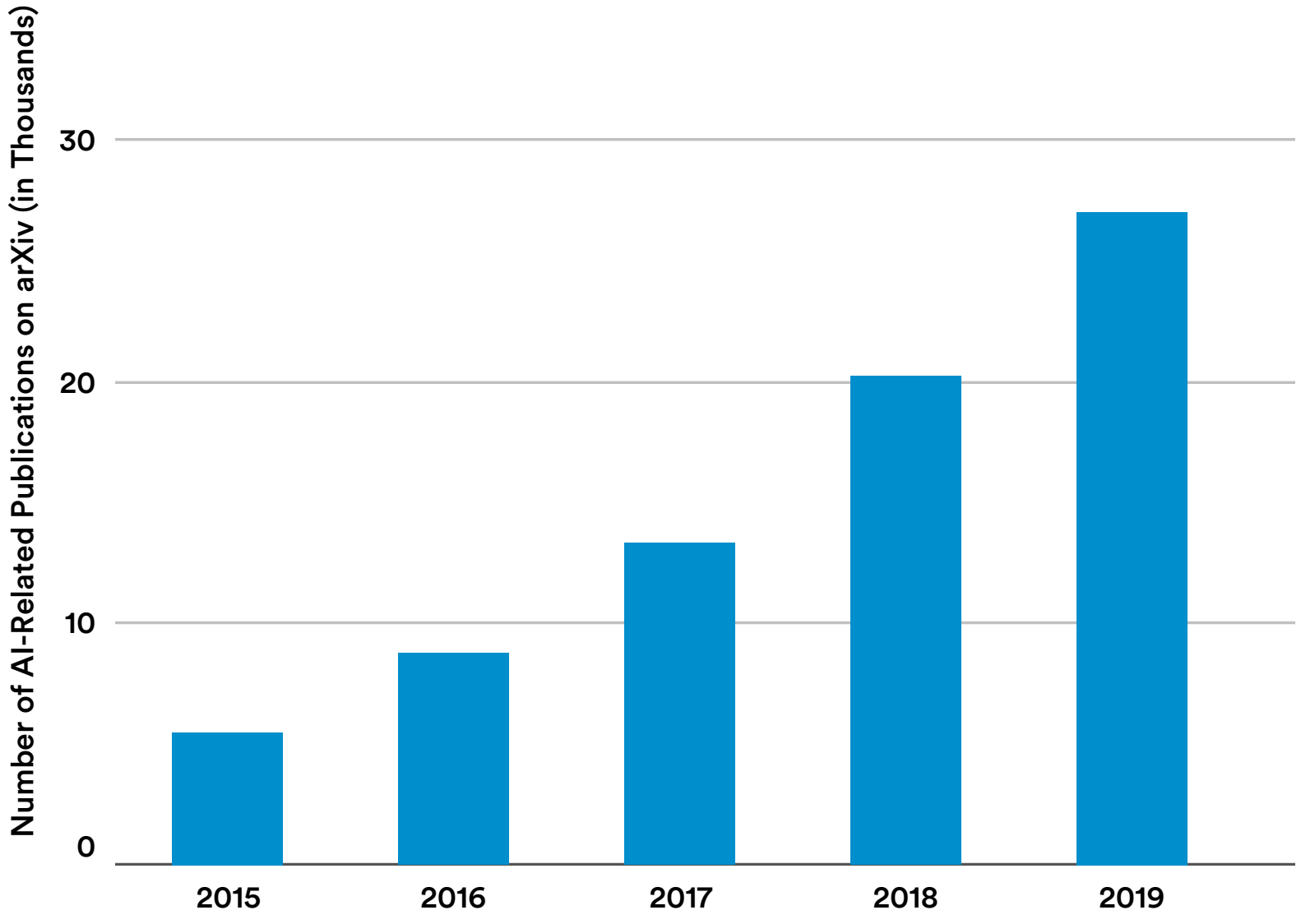

Figure 1.1.16

### By Region

The analysis by region shows that while North America still holds the lead in the global share of arXiV AI-related publications, its share has been decreasing—from 41.6% in 2017 to 36.3% in 2020 (Figure 1.1.17). Meanwhile, the share of publications in East Asia & Pacific has grown steadily in the past five years—from 17.3% in 2015 to 26.5% in 2020.

ARXIV AI-RELATED PUBLICATIONS (% of WORLD TOTAL) by REGION, 2015-20

Source: arXiv, 2020 | Chart: 2021 AI Index Report

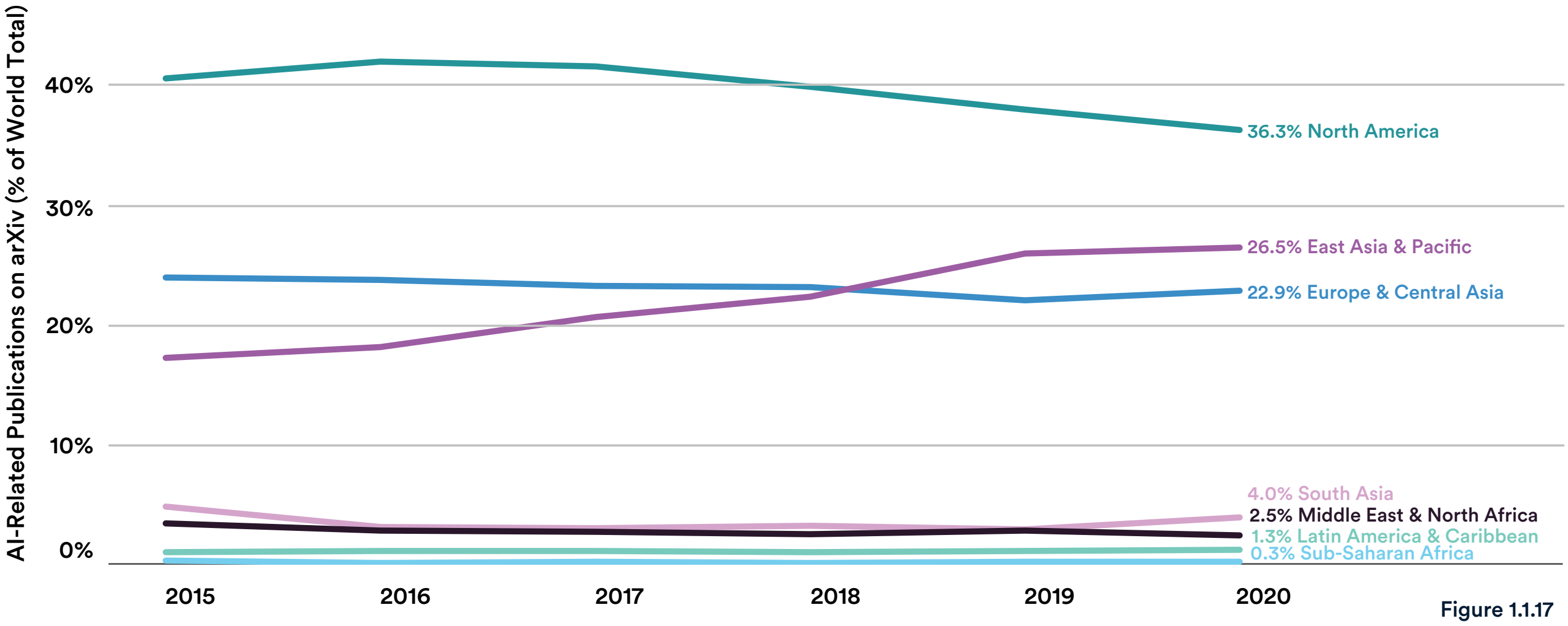

Figure 1.1.17

## By Geographic Area

While the total number of AI-related publications on arXiv is increasing among the three major AI powers, China is catching up with the United States (Figure 1.1.18a and Figure 1.1.18b). The share of publication counts by the European Union, on the other hand, has remained largely unchanged.

NUMBER of AI-RELATED PUBLICATIONS on ARXIV by GEOGRAPHIC AREA, 2015-20

Source: arXiv, 2020 | Chart: 2021 AI Index Report

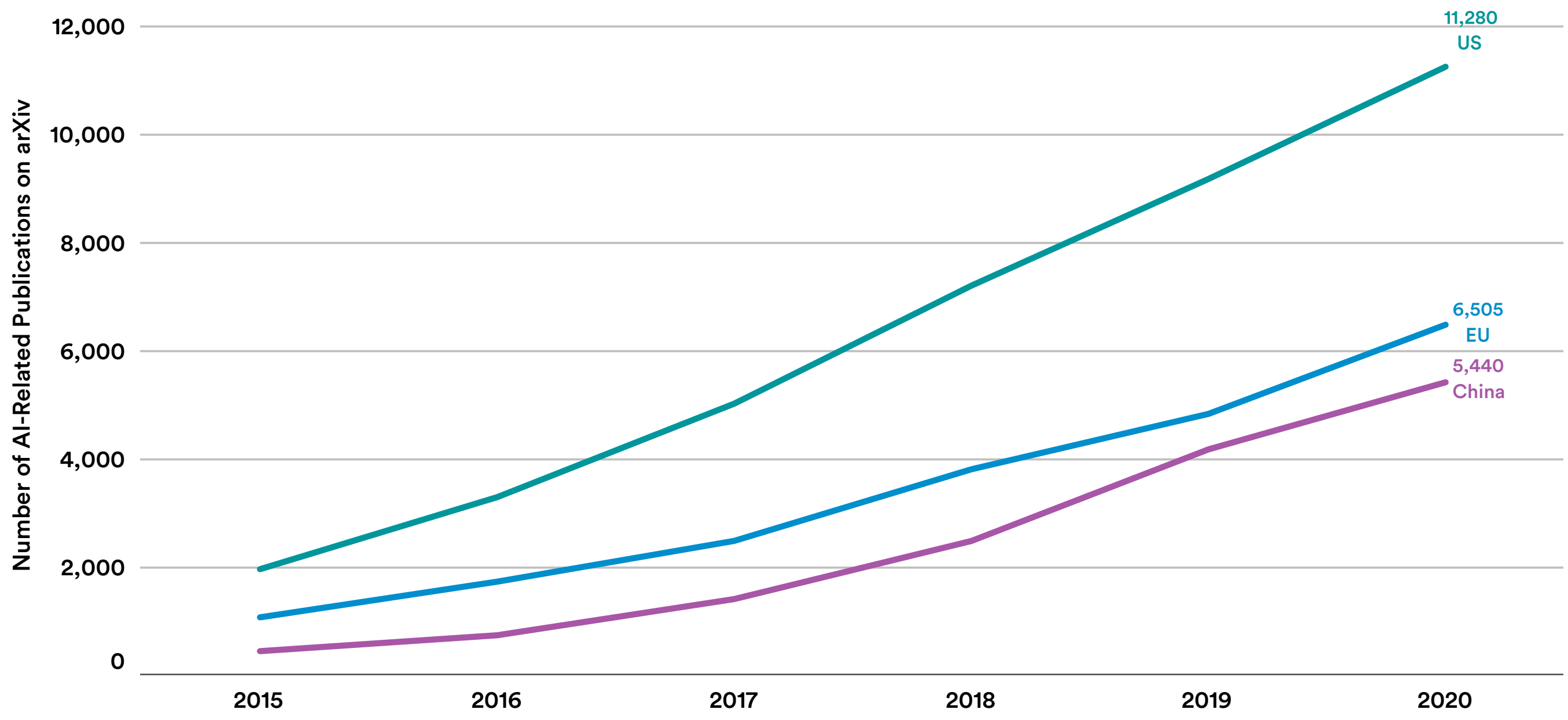

Figure 1.1.18a

ARXIV AI-RELATED PUBLICATIONS (% of WORLD TOTAL) by GEOGRAPHIC AREA, 2015-20

Source: arXiv, 2020 | Chart: 2021 AI Index Report

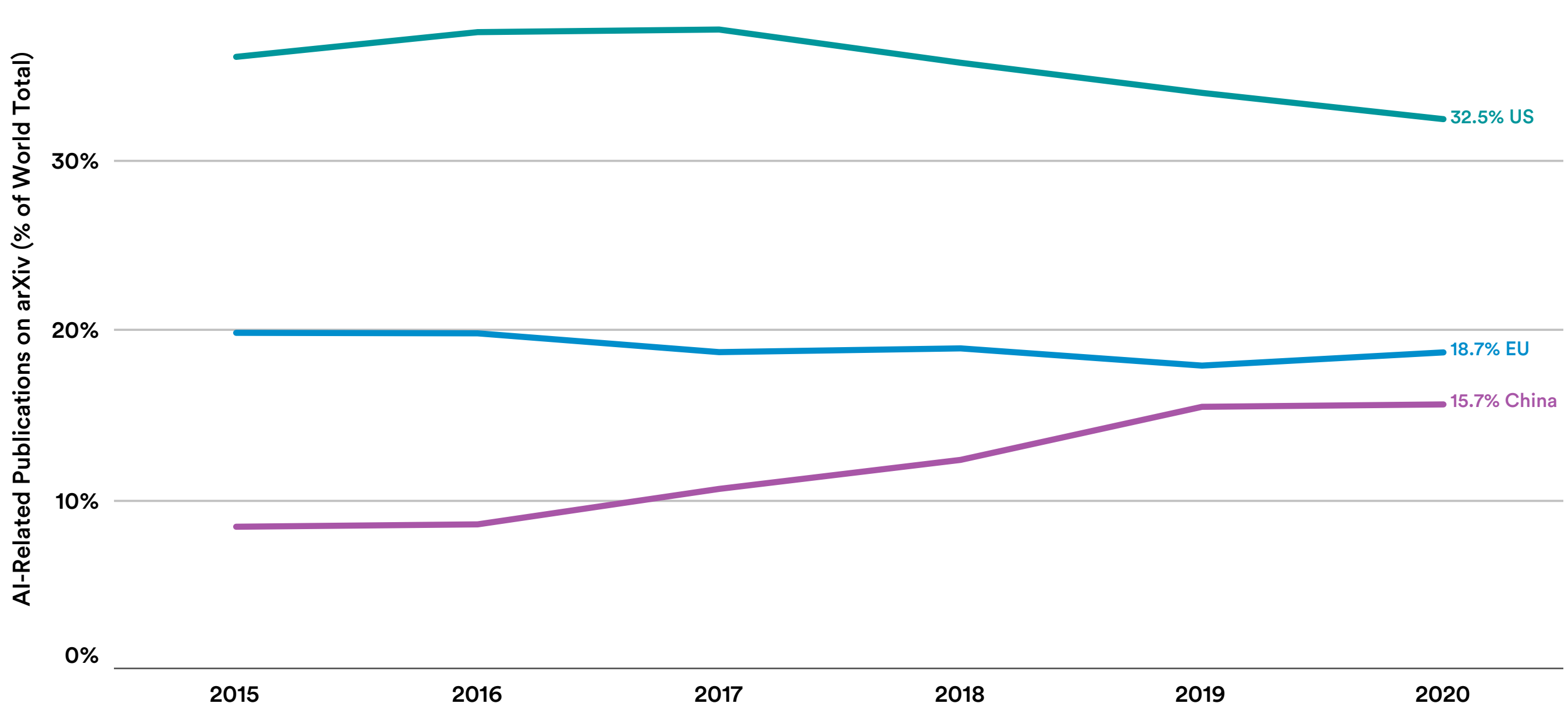

Figure 1.1.18b

## By Field of Study

Among the six fields of study related to AI on arXiv, the number of publications in Robotics (cs.RO) and Machine Learning in computer science (cs.LG) have seen the fastest growth between 2015 and 2020, increasing by 11 times and 10 times respectively (Figure 1.1.19). In 2020, cs.LG and Computer Vision (cs.CV) lead in the overall number of publications, accounting for 32.0% and 31.7%, respectively, of all AI-related publications on arXiv. Between 2019 and 2020, the fastest-growing categories of the seven studied here were Computation and Language (cs.CL), by 35.4%, and cs.RO, by 35.8%.

NUMBER of AI-RELATED PUBLICATIONS on ARXIV by FIELD of STUDY 2015-20
Source: arXiv, 2020 | Chart: 2021 AI Index Report

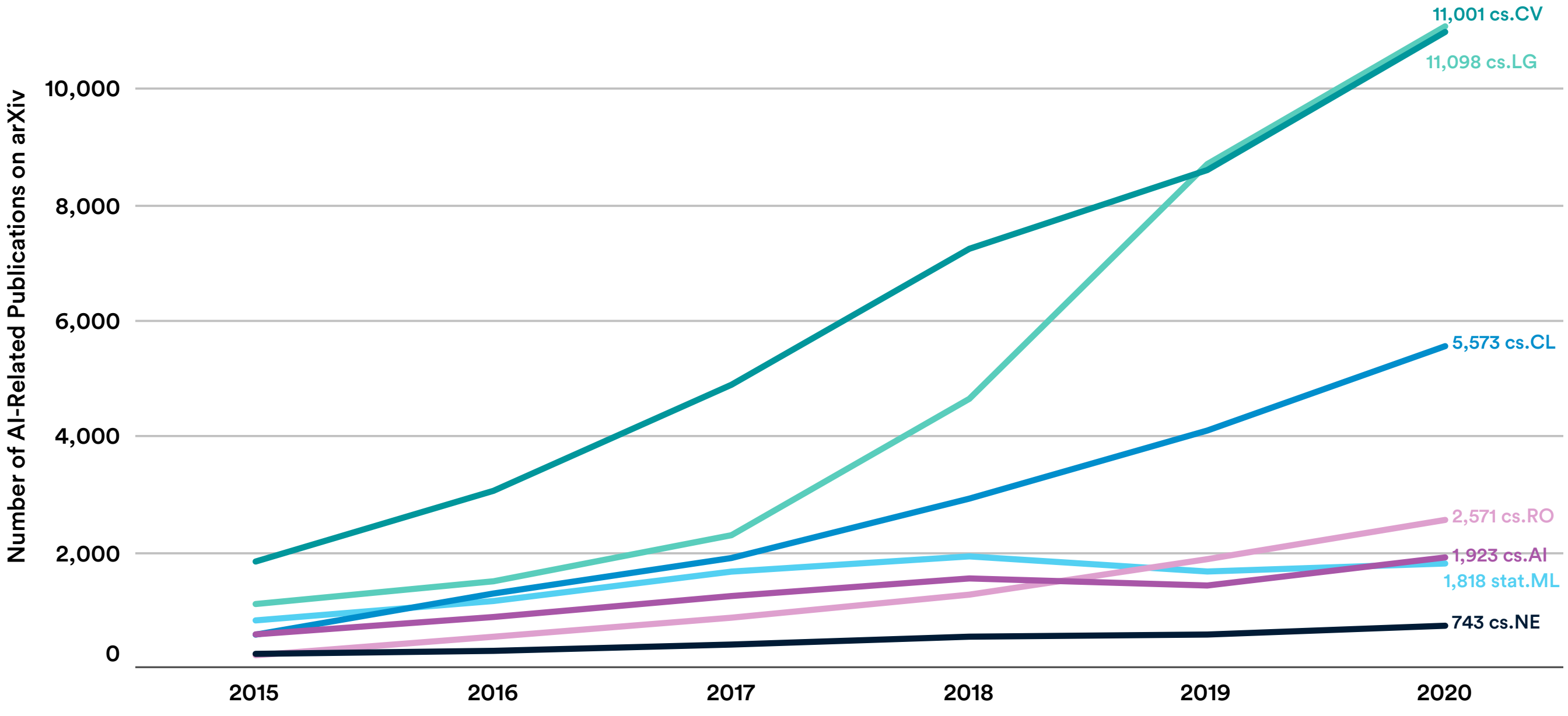

Figure 1.1.19

# Deep Learning Papers on arXiv

With increased access to data and significant improvements in computing power, the field of deep learning (DL) is growing at breakneck speed. Researchers from Nesta used a topic modeling algorithm to identify the deep learning papers on arXiv by analyzing the abstract of arXiv papers under the Computer Science (CS) and Machine Learning in Statistics (state.ML) categories. Figure 1.1.20 suggests that in the last five years alone, the overall number of DL publications on arXiv grew almost sixfold.

NUMBER of DEEP LEARNING PUBLICATIONS on ARXIV, 2010-19

Source: arXiv/Nesta, 2020 | Chart: 2021 AI Index Report

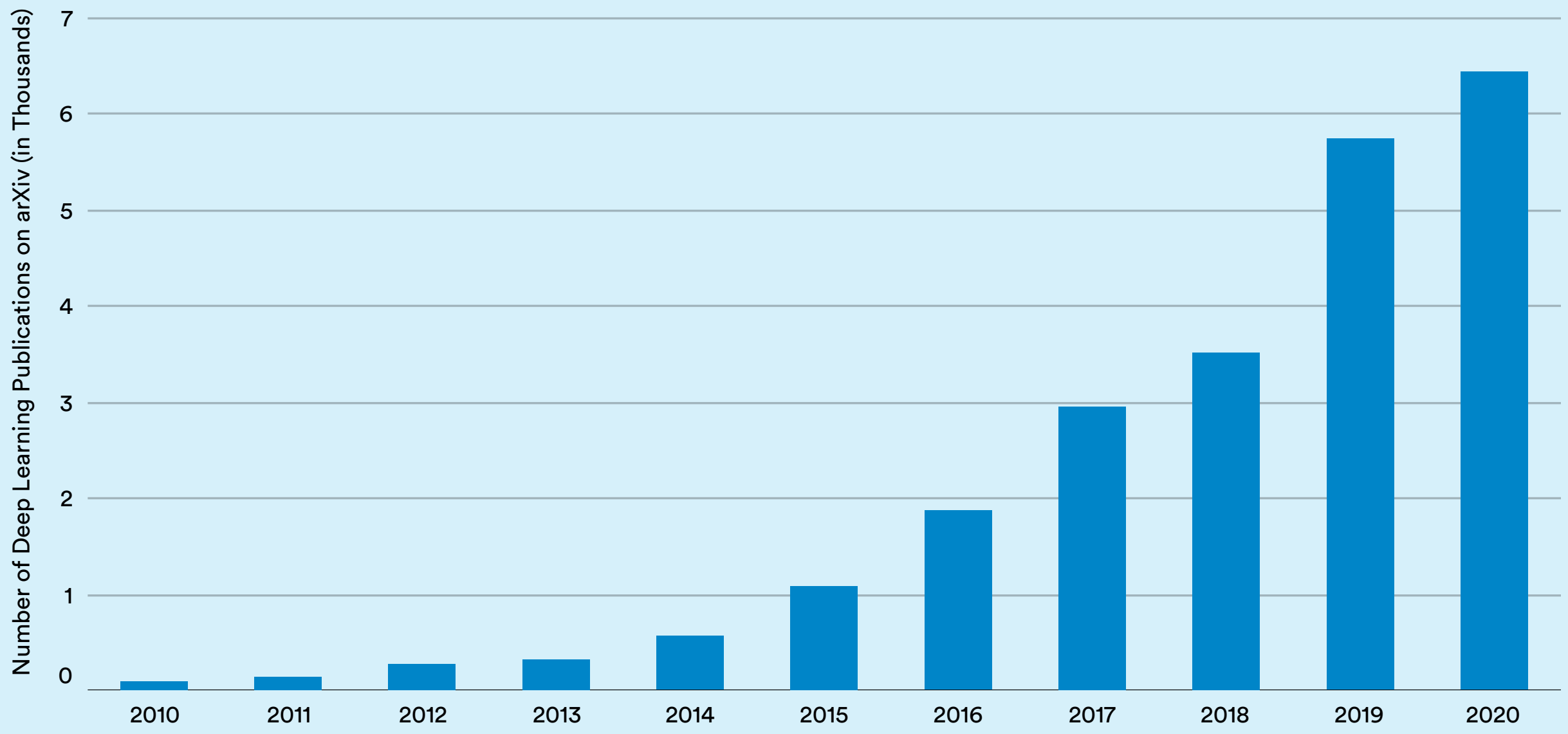

Figure 1.1.20

Conference attendance is an indication of broader industrial and academic interest in a scientific field. In the past 20 years, AI conferences have grown not only in size but also in number and prestige. This section presents data on the trends in attendance at and submissions to major AI conferences.

# 1.2 CONFERENCES

## CONFERENCE ATTENDANCE

Last year saw a significant increase in participation levels at AI conferences, as most were offered through a virtual format. Only the 34th Association for the Advancement of Artificial Intelligence (AAAI) Conference on Artificial Intelligence was held in person in February 2020. Conference organizers report that a virtual format allows for higher attendance of researchers from all over the world, though exact attendance numbers are difficult to measure.

Due to the atypical nature of 2020 conference attendance data, the 11 major AI conferences in 2019 have been split into two categories based on 2019 attendance data: large AI conferences with over 3,000 attendees and small AI conferences with fewer than 3,000 attendees. Figure 1.2.1 shows that in 2020, the total number of attendees across nine conferences almost doubled.[5] In particular, the International Conference on Intelligent Robots and Systems (IROS) extended the virtual conference to allow users to watch events for up to three months, which explains the high attendance count. Because the International Joint Conference on Artificial Intelligence (IJCAI) was held in 2019 and January 2021—but not in 2020—it does not appear on the charts.

5 For the AAMAS conference, the attendance in 2020 is based on the number of users on site reported by the platform that recorded the talks and managed the online conference; For the KR conference, the attendance in 2020 is based on the number of registrations; For the ICPAS conference, the attendance of 450 in 2020 is an estimate as some participants may have used anonymous Zoom accounts.

## ATTENDANCE at LARGE AI CONFERENCES, 2010-20
Source: Conference Data | Chart: 2021 AI Index Report

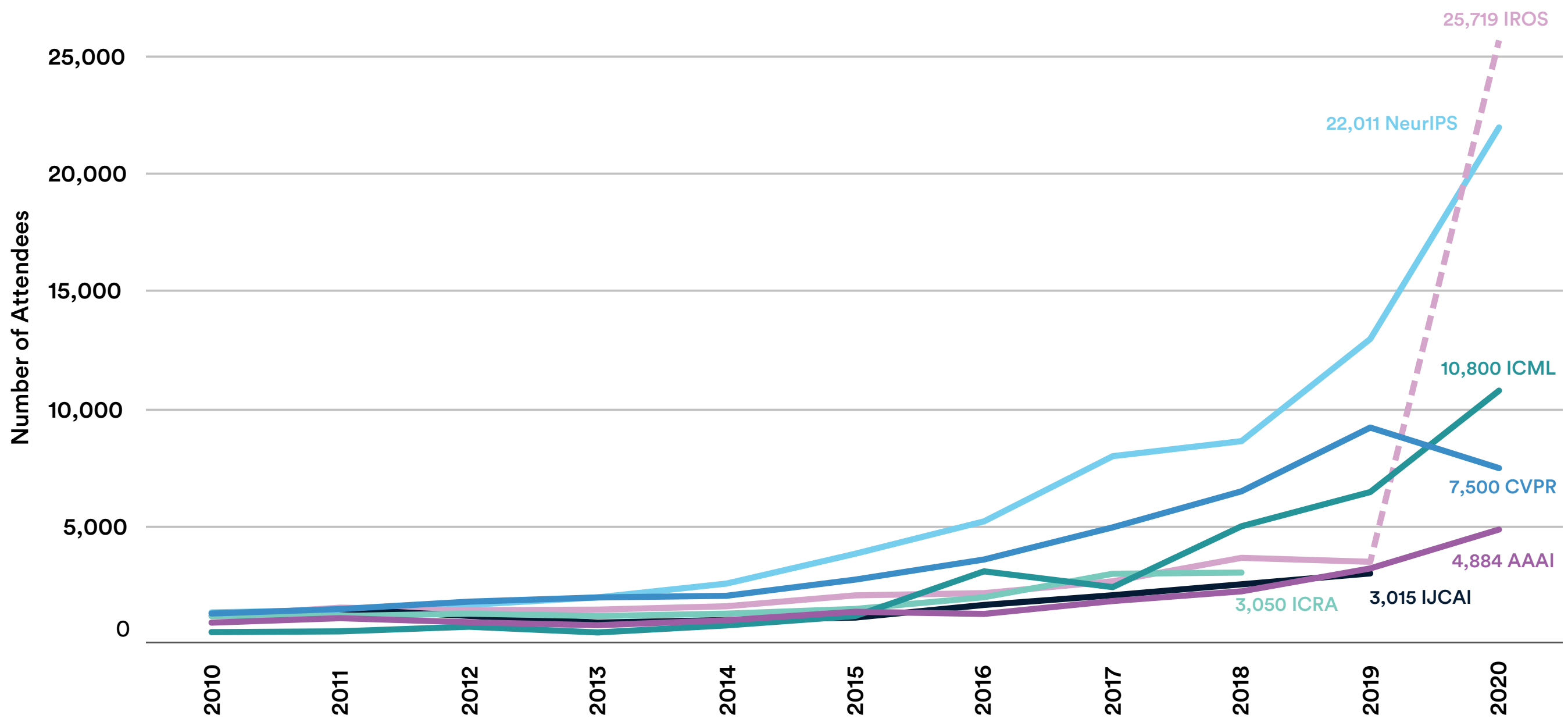

Figure 1.2.1

## ATTENDANCE at SMALL AI CONFERENCES, 2010-20
Source: Conference Data | Chart: 2021 AI Index Report

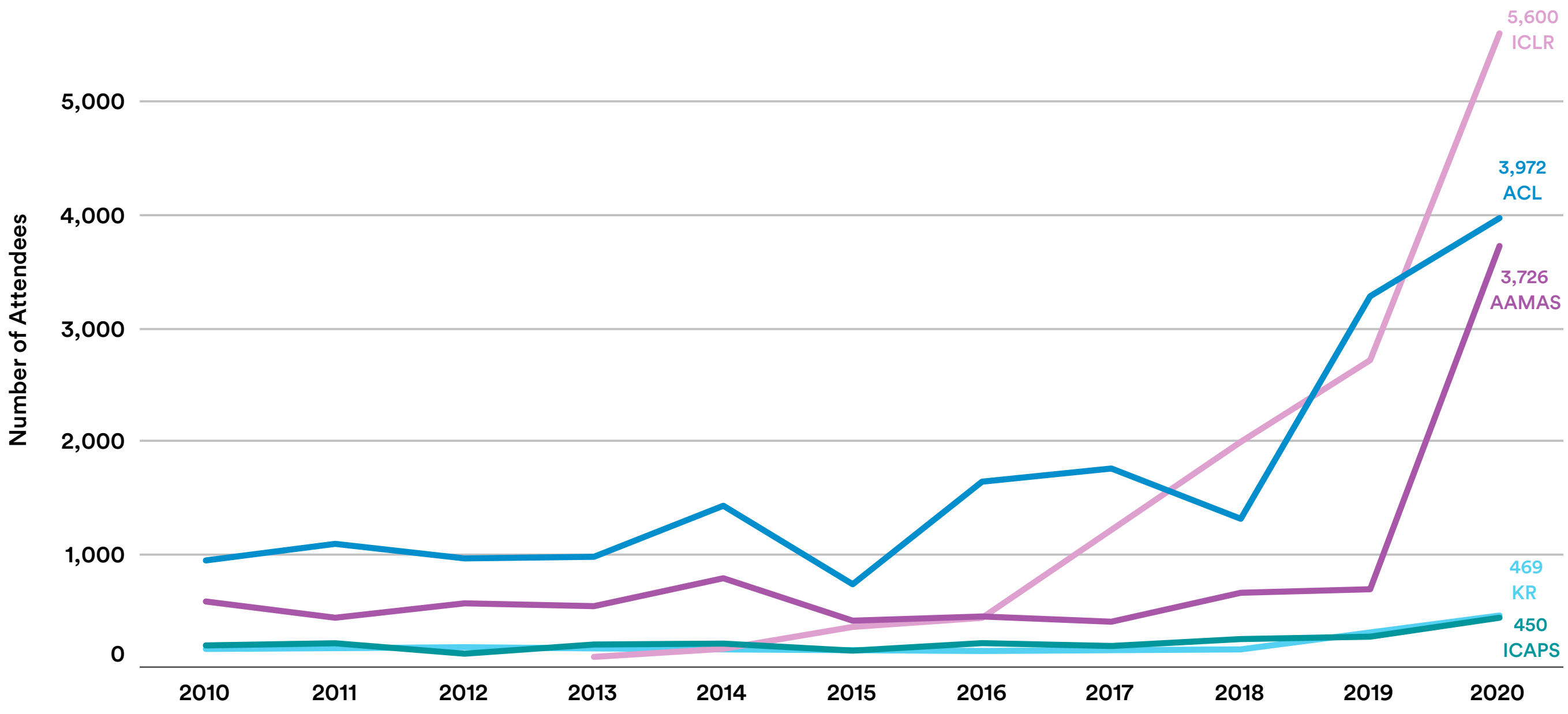

Figure 1.2.2

## Corporate Representation at AI Research Conferences

Researchers from Virginia Tech and Ivey Business School, Western University found that large technology firms have increased participation in major AI conferences. In their paper, titled "The De-Democratization of AI: Deep Learning and the Compute Divide in Artificial Intelligence Research," the ressearchers use the share of papers affiliated with firms over time at AI conferences to illustrate the increased presence of firms in AI research. They argue that the unequal distribution of compute power in academia, which they refer to as the "compute divide," is adding to the inequality in the era of deep learning. Big tech firms tend to have more resources to design AI products, but they also tend to be less diverse than less elite or smaller institutions. This raises concerns about bias and fairness within AI. All 10 major AI conferences displayed in Figure 1.2.3 show an upward trend in corporate representation, which further extends the compute divide.

SHARE of FORTUNE GLOBAL 500 TECH-AFFILIATED PAPERS
Source: Ahmed & Wahed, 2020 | Chart: 2021 AI Index Report

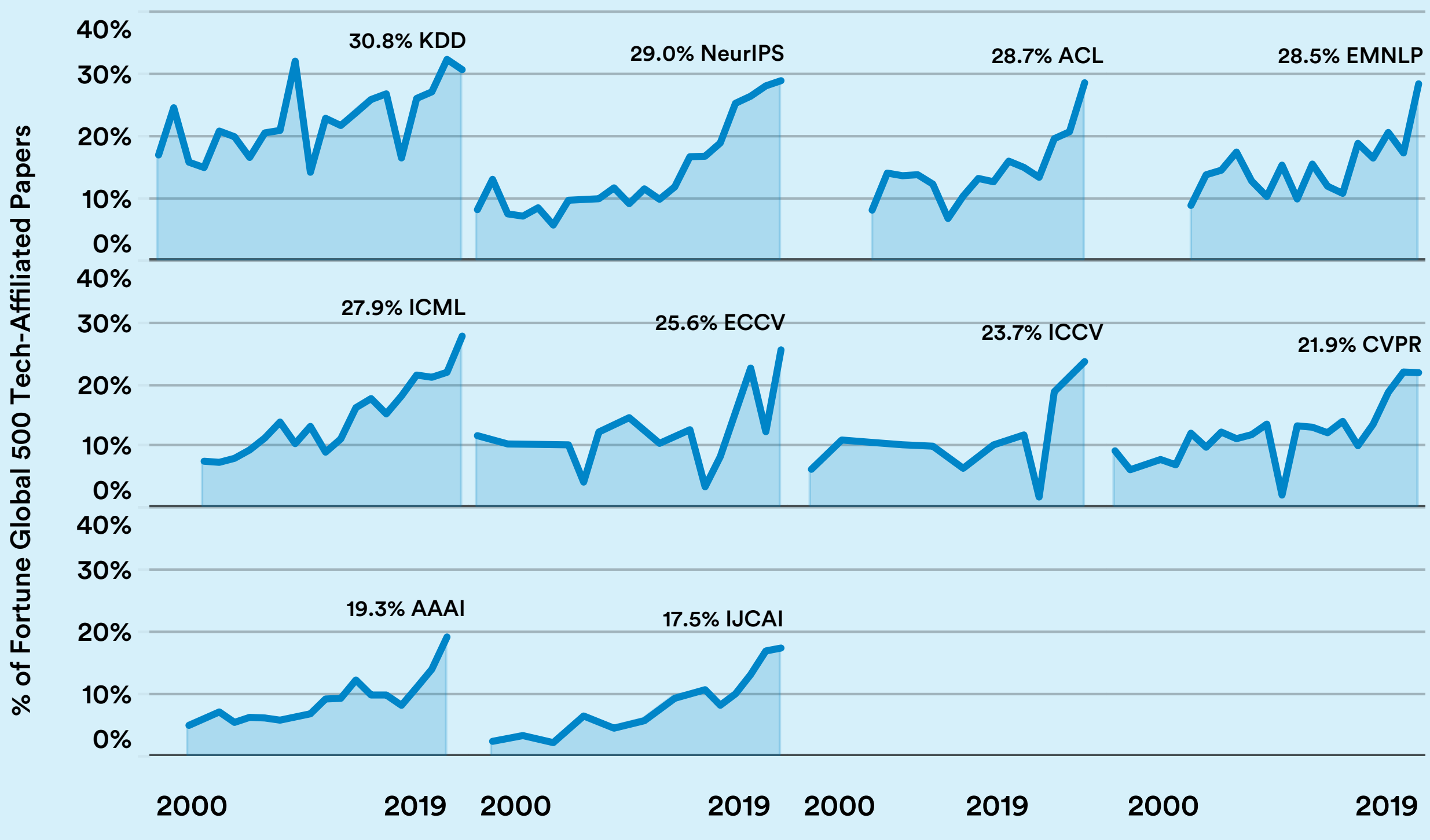

Figure 1.2.3

A software library is a collection of computer code that is used to create applications and products. Popular AI-specific software libraries—such as TensorFlow and PyTorch—help developers create their AI solutions quickly and efficiently. This section analyzes the popularity of software libraries through GitHub data.

# 1.3 AI OPEN-SOURCE SOFTWARE LIBRARIES

## GITHUB STARS

GitHub is a code hosting platform that AI researchers and developers frequently use to upload, comment on, and download software. GitHub users can "star" a project to save it in their list, thereby expressing their interests and likes—similar to the "like'' function on Twitter and other social media platforms. As AI researchers upload packages on GitHub that mention the use of an open-source library, the "star" function on GitHub can be used to measure the popularity of various AI programming open-source libraries.

Figure 1.3.1 suggests that TensorFlow (developed by Google and publicly released in 2017) is the most popular AI software library. The second most popular library in 2020 is Keras (also developed by Google and built on top of TensorFlow 2.0). Excluding TensorFlow, Figure 1.3.2 shows that PyTorch (created by Facebook) is another library that is becoming increasingly popular.

**TensorFlow (developed by Google and publicly released in 2017) is the most popular AI software library. The second most popular library in 2020 is Keras (also developed by Google and built on top of TensorFlow 2.0).**

## NUMBER of GITHUB STARS by AI LIBRARY, 2014-20

Source: GitHub, 2020 | Chart: 2021 AI Index Report

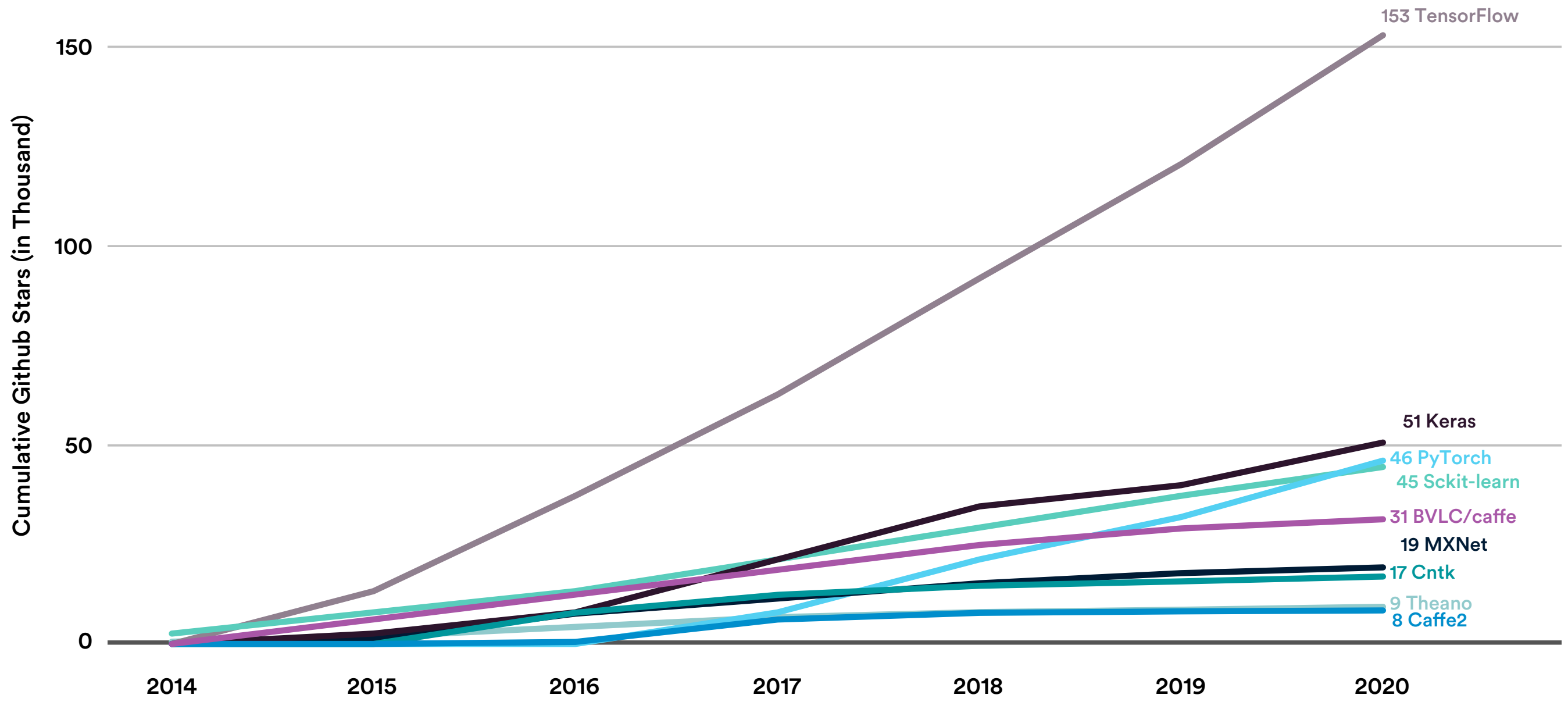

Figure 1.3.1

## NUMBER of GITHUB STARS by AI LIBRARY (excluding TENSORFLOW), 2014-20

Source: GitHub, 2020 | Chart: 2021 AI Index Report

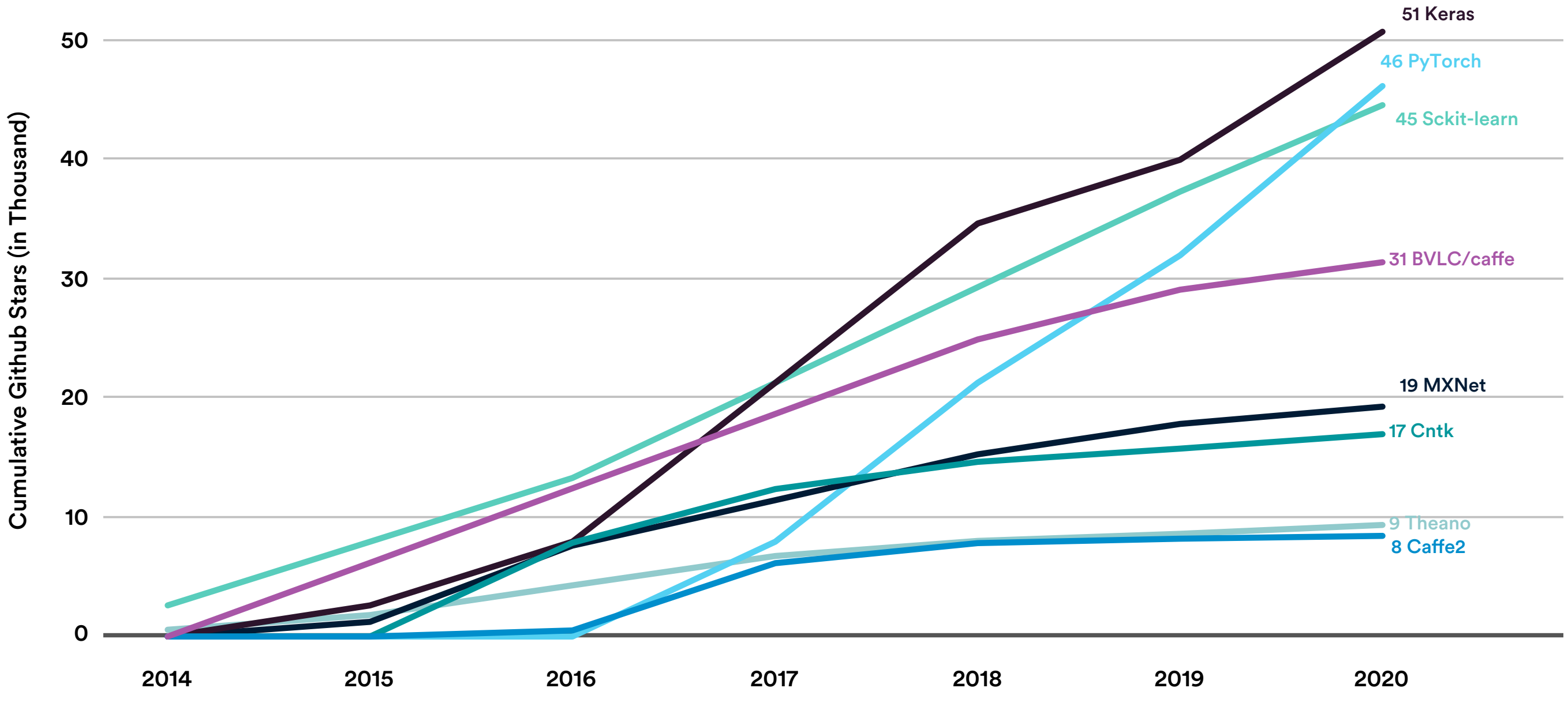

Figure 1.3.2

# CHAPTER 2:
# Technical Performance

Artificial Intelligence
Index Report 2021

## CHAPTER 2:
# Chapter Preview

ACCESS THE PUBLIC DATA

# Overview

This chapter highlights the technical progress in various subfields of AI, including computer vision, language, speech, concept learning, and theorem proving. It uses a combination of quantitative measurements, such as common benchmarks and prize challenges, and qualitative insights from academic papers to showcase the developments in state-of-the-art AI technologies.

While technological advances allow AI systems to be deployed more widely and easily than ever, concerns about the use of AI are also growing, particularly when it comes to issues such as algorithmic bias. The emergence of new AI capabilities such as being able to synthesize images and videos also poses ethical challenges.

## CHAPTER HIGHLIGHTS

- **Generative everything:** AI systems can now compose text, audio, and images to a sufficiently high standard that humans have a hard time telling the difference between synthetic and non-synthetic outputs for some constrained applications of the technology. That promises to generate a tremendous range of downstream applications of AI for both socially useful and less useful purposes. It is also causing researchers to invest in technologies for detecting generative models; the DeepFake Detection Challenge data indicates how well computers can distinguish between different outputs.

- **The industrialization of computer vision:** Computer vision has seen immense progress in the past decade, primarily due to the use of machine learning techniques (specifically deep learning). New data shows that computer vision is industrializing: Performance is starting to flatten on some of the largest benchmarks, suggesting that the community needs to develop and agree on harder ones that further test performance. Meanwhile, companies are investing increasingly large amounts of computational resources to train computer vision systems at a faster rate than ever before. Meanwhile, technologies for use in deployed systems—like object-detection frameworks for analysis of still frames from videos—are maturing rapidly, indicating further AI deployment.

- **Natural Language Processing (NLP) outruns its evaluation metrics:** Rapid progress in NLP has yielded AI systems with significantly improved language capabilities that have started to have a meaningful economic impact on the world. Google and Microsoft have both deployed the BERT language model into their search engines, while other large language models have been developed by companies ranging from Microsoft to OpenAI. Progress in NLP has been so swift that technical advances have started to outpace the benchmarks to test for them. This can be seen in the rapid emergence of systems that obtain human level performance on SuperGLUE, an NLP evaluation suite developed in response to earlier NLP progress overshooting the capabilities being assessed by GLUE.

- **New analyses on reasoning:** Most measures of technical problems show for each time point the performance of the best system at that time on a fixed benchmark. New analyses developed for the AI Index offer metrics that allow for an evolving benchmark, and for the attribution to individual systems of credit for a share of the overall performance of a group of systems over time. These are applied to two symbolic reasoning problems, Automated Theorem Proving and Satisfiability of Boolean formulas.

- **Machine learning is changing the game in healthcare and biology:** The landscape of the healthcare and biology industries has evolved substantially with the adoption of machine learning. DeepMind's AlphaFold applied deep learning technique to make a significant breakthrough in the decades-long biology challenge of protein folding. Scientists use ML models to learn representations of chemical molecules for more effective chemical synthesis planning. PostEra, an AI startup used ML-based techniques to accelerate COVID-related drug discovery during the pandemic.

# Computer Vision

Introduced in the 1960s, the field of computer vision has seen significant progress and in recent years has started to reach human levels of performance on some restricted visual tasks. Common computer vision tasks include object recognition, pose estimation, and semantic segmentation. The maturation of computer vision technology has unlocked a range of applications: self-driving cars, medical image analysis, consumer applications (e.g., Google Photos), security applications (e.g., surveillance, satellite imagery analysis), industrial applications (e.g., detecting defective parts in manufacturing and assembly), and others.

# 2.1 COMPUTER VISION—IMAGE

## IMAGE CLASSIFICATION

In the 2010s, the field of image recognition and classification began to switch from classical AI techniques to ones based on machine learning and, specifically, deep learning. Since then, image recognition has shifted from being an expensive, domain-specific technology to being one that is more affordable and applicable to more areas—primarily due to advancements in the underlying technology (algorithms, compute hardware, and the utilization of larger datasets).

### ImageNet

Created by computer scientists from Stanford University and Princeton University in 2009, ImageNet is a dataset of over 14 million images across 200 classes that expands and improves the data available for researchers to train AI algorithms. In 2012, researchers from the University of Toronto used techniques based on deep learning to set a new state of the art in the ImageNet Large Scale Visual Recognition Challenge.

Since then, deep learning techniques have ruled the competition leaderboards—several widely used techniques have debuted in ImageNet competition entries. In 2015, a team from Microsoft Research said it had surpassed human-level performance on the image classification task[1] via the use of "residual networks"—an innovation that subsequently proliferated into other AI systems. Even after the end of the competition in 2017, researchers continue to use the ImageNet dataset to test and develop computer vision applications.

The image classification task of the ImageNet Challenge asks machines to assign a class label to an image based on the main object in the image. The following graphs explore the evolution of the top-performing ImageNet systems over time, as well as how algorithmic and infrastructure advances have allowed researchers to increase the efficiency of training image recognition systems and reduce the absolute time it takes to train high-performing ones.

**Image recognition has shifted from being an expensive, domain-specific technology to being one that is more affordable and applicable to more areas—primarily due to advancements in the underlying technology.**

### ImageNet: Top-1 Accuracy

Top-1 accuracy tests for how well an AI system can assign the correct label to an image, specifically whether its single most highly probable prediction (out of all possible labels) is the same as the target label. In recent years, researchers have started to focus on improving performance on ImageNet by pre-training their systems on extra training data, for instance photos from Instagram or other social media sources. By pre-training on these datasets, they're able to more effectively use ImageNet data, which further improves performance. Figure 2.1.1 shows that recent systems with extra training data make 1 error out of every 10 tries on top-1 accuracy, versus 4 errors out of every 10 tries in December 2012. The model from the Google Brain team achieved 90.2% on top-1 accuracy in January 2021.

1 Though it is worth noting that the human baseline for this metric comes from a single Stanford graduate student who took roughly the same test as the AI systems took.

IMAGENET CHALLENGE: TOP-1 ACCURACY
Source: Papers with Code, 2020; AI Index, 2021 | Chart: 2021 AI Index Report

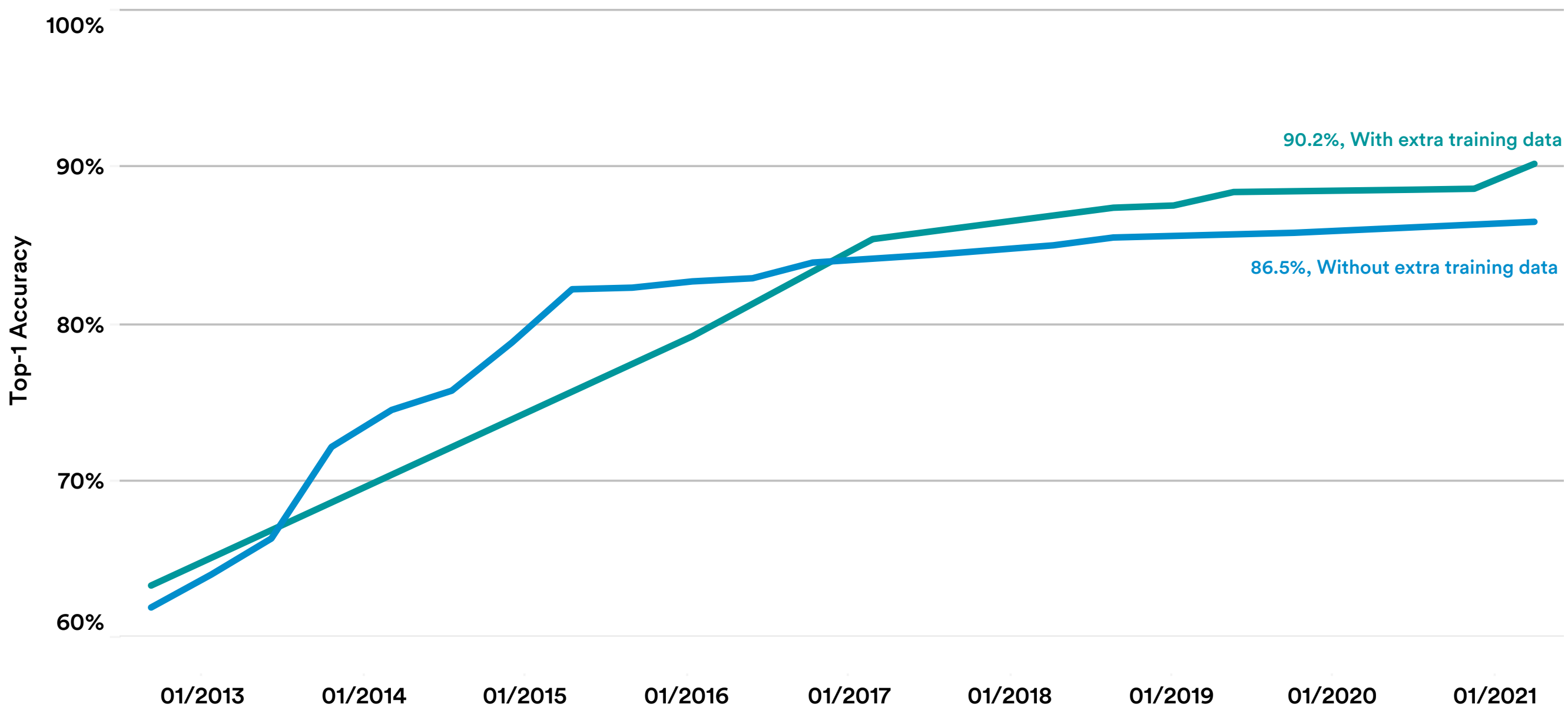

Figure 2.1.1

## ImageNet: Top-5 Accuracy

Top-5 accuracy asks whether the correct label is in at least the classifier's top five predictions. Figure 2.1.2 shows that the error rate has improved from around 85% in 2013 to almost 99% in 2020.[2]

IMAGENET CHALLENGE: TOP-5 ACCURACY
Source: Papers with Code, 2020; AI Index, 2021 | Chart: 2021 AI Index Report

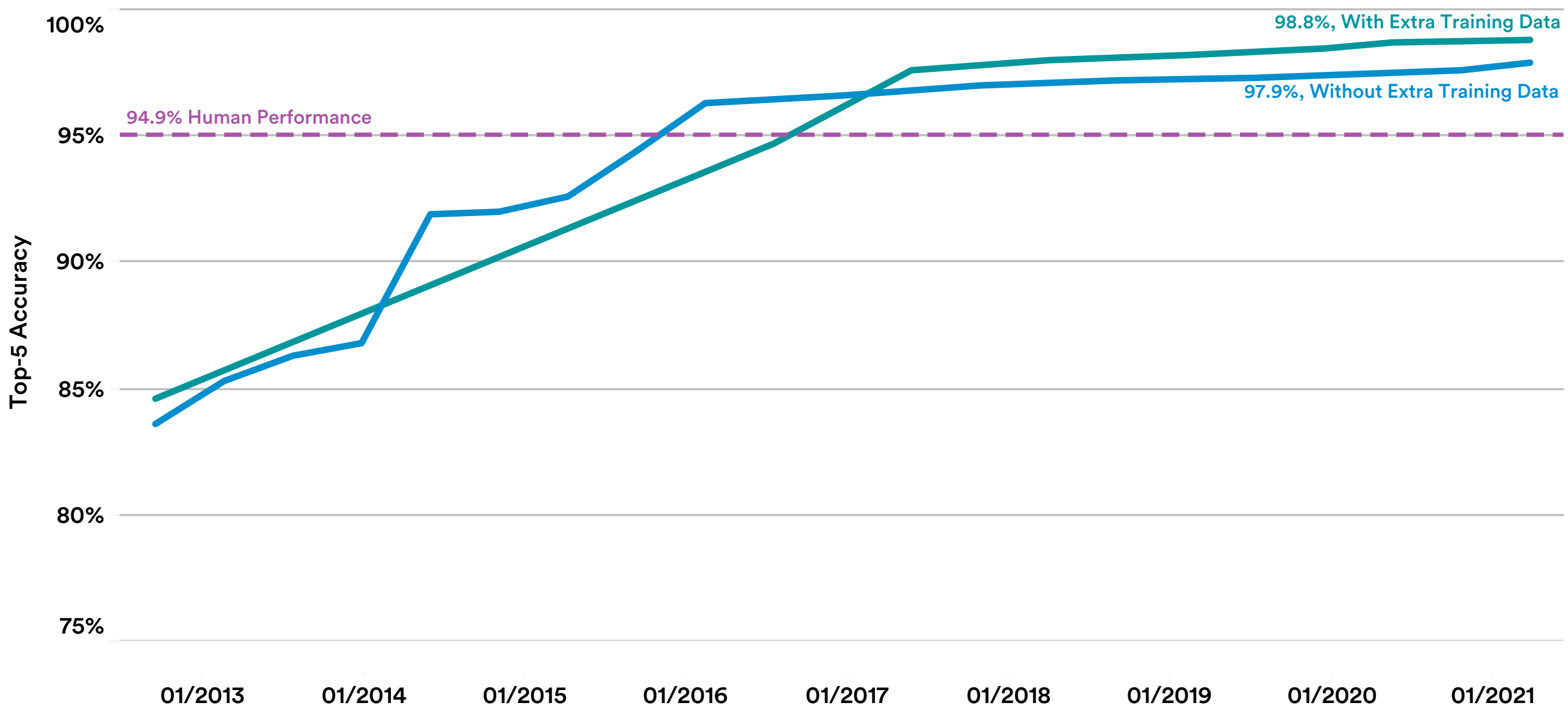

Figure 2.1.2

2 Note: For data on human error, a human was shown 500 images and then was asked to annotate 1,500 test images; their error rate was 5.1% for Top-5 classification. This is a very rough baseline, but it gives us a sense of human performance on this task.

## ImageNet: Training Time

Along with measuring the raw improvement in accuracy over time, it is useful to evaluate how long it takes to train image classifiers on ImageNet to a standard performance level as it sheds light on advances in the underlying computational infrastructure for large-scale AI training. This is important to measure because the faster you can train a system, the more quickly you can evaluate it and update it with new data. Therefore, the faster ImageNet systems can be trained, the more productive organizations can become at developing and deploying AI systems. Imagine the difference between waiting a few seconds for a system to train versus waiting a few hours, and what that difference means for the type and volume of ideas researchers explore and how risky they might be.

What follows are the results from MLPerf, a competition run by the MLCommons organization that challenges entrants to train an ImageNet network using a common (residual network) architecture, and then ranks systems according to the absolute "wall clock" time it takes them to train a system.[3]

As shown in Figure 2.1.3, the training time on ImageNet has fallen from 6.2 minutes (December 2018) to 47 seconds (July 2020). At the same time, the amount of hardware used to achieve these results has increased dramatically; frontier systems have been dominated by the use of "accelerator" chips, starting with GPUs in the 2018 results, and transitioning to Google's TPUs for the best-in-class results from 2019 and 2020.

**Distribution of Training Time:** MLPerf does not just show the state of the art for each competition period; it also makes available all the data behind each entry in each competition cycle. This, in turn, reveals the distribution of training times for each period (Figure 2.1.3). (Note that in each MLPerf competition, competitors typically submit multiple entries that use different permutations of hardware.)

Figure 2.1.4 shows that in the past couple of years, training times have shortened, as has the variance between MLPerf entries. At the same time, competitors have started to use larger and larger numbers of accelerator chips to speed training times. This is in line with broader trends in AI development, as large-scale training becomes better understood, with a higher degree of shared best practices and infrastructure.

3 The next MLPerf update is planned for June 2021.

IMAGENET: TRAINING TIME and HARDWARE of the BEST SYSTEM
Source: MLPerf, 2020 | Chart: 2021 AI Index Report

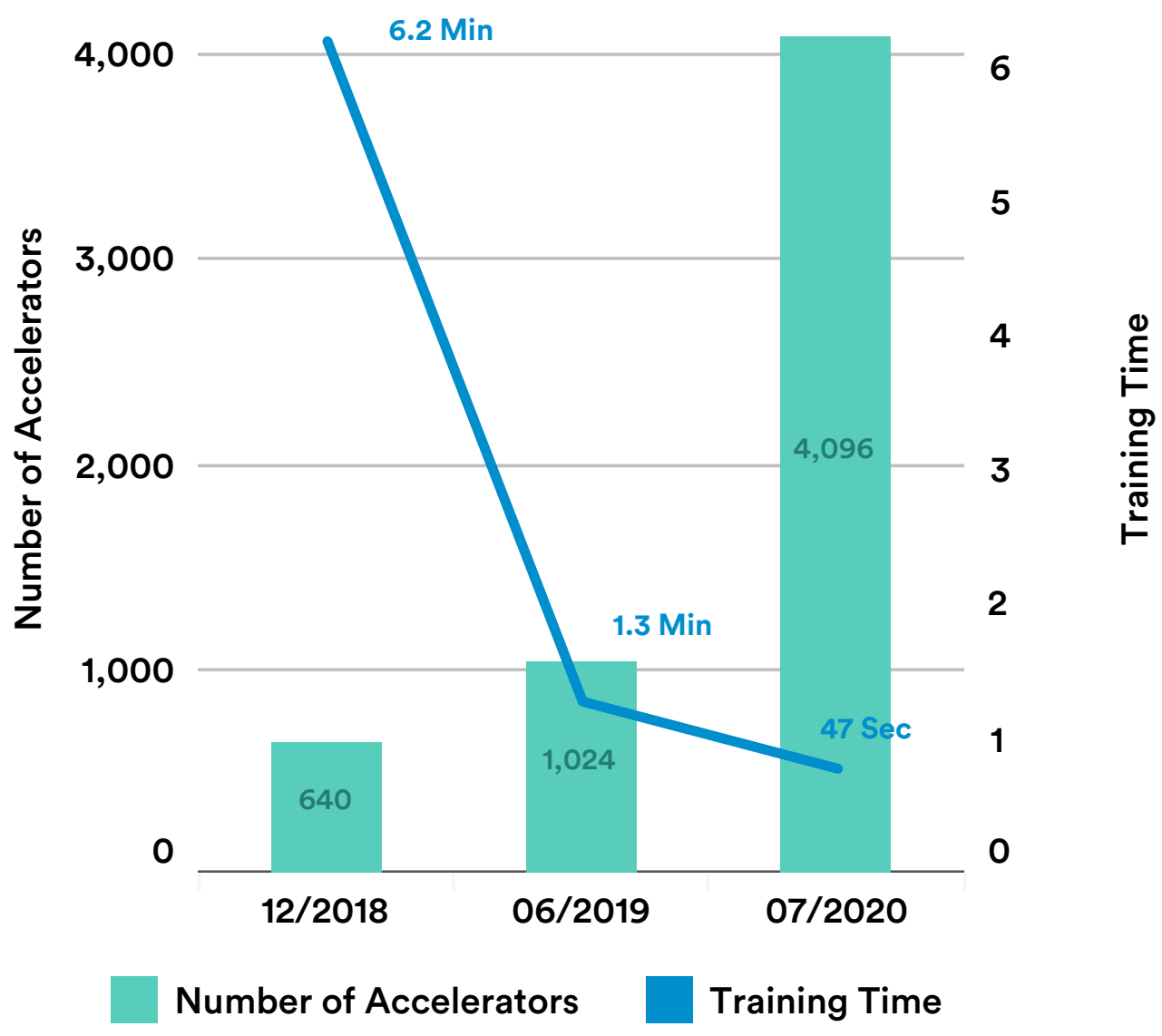

Figure 2.1.3

IMAGENET: DISTRIBUTION of TRAINING TIME
Source: MLPerf, 2020 | Chart: 2021 AI Index Report

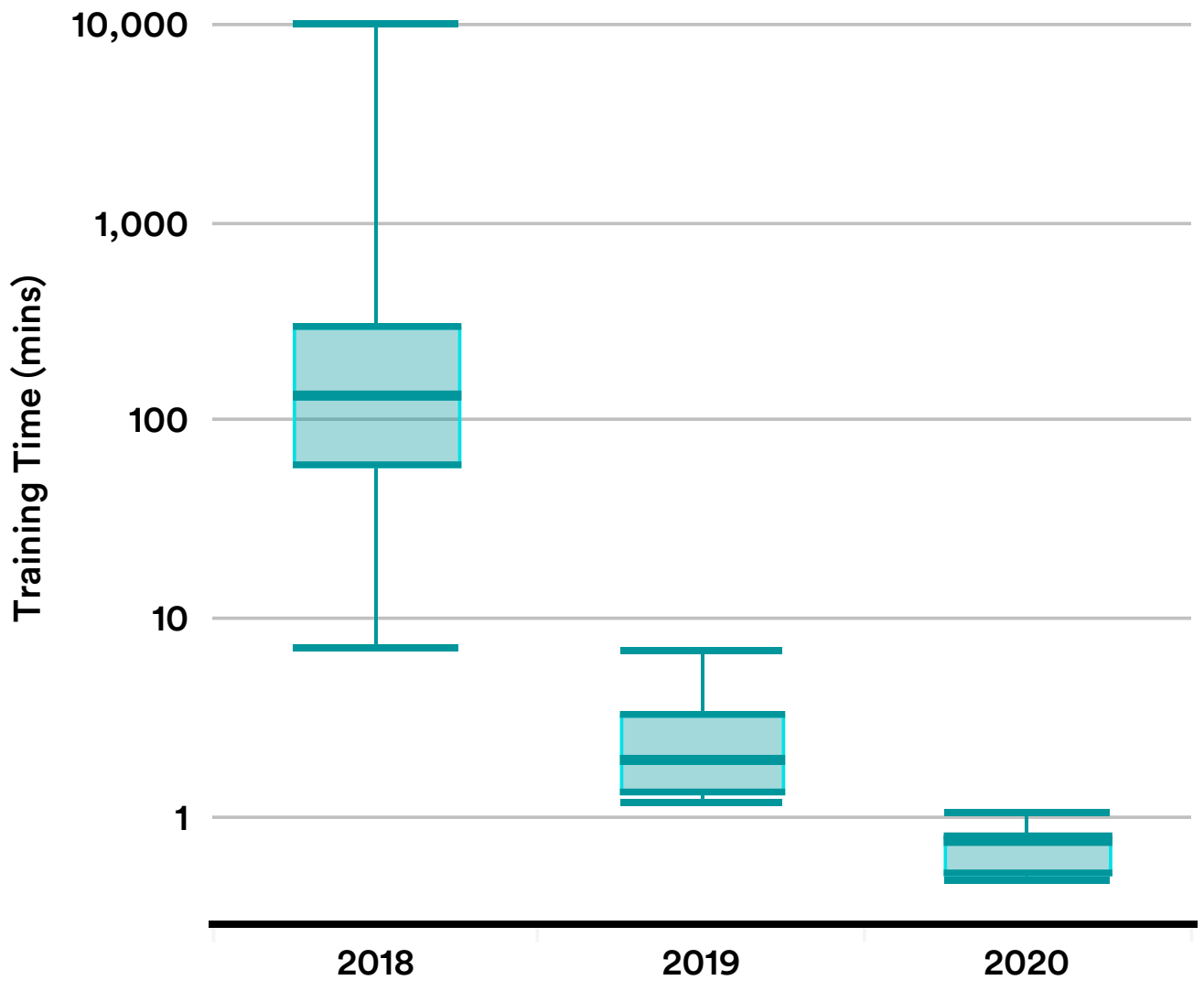

Figure 2.1.4

## ImageNet: Training Costs

How much does it cost to train a contemporary image-recognition system? The answer, according to tests run by the Stanford DAWNBench team, is a few dollars in 2020, down by around 150 times from costs in 2017 (Figure 2.1.5). To put this in perspective, what cost one entrant around USD 1,100 to do in October 2017 now costs about USD 7.43. This represents progress in algorithm design as well as a drop in the costs of cloud-computing resources.

IMAGENET: TRAINING COST (to 93% ACCURACY)
Source: DAWNBench, 2020 | Chart: 2021 AI Index Report

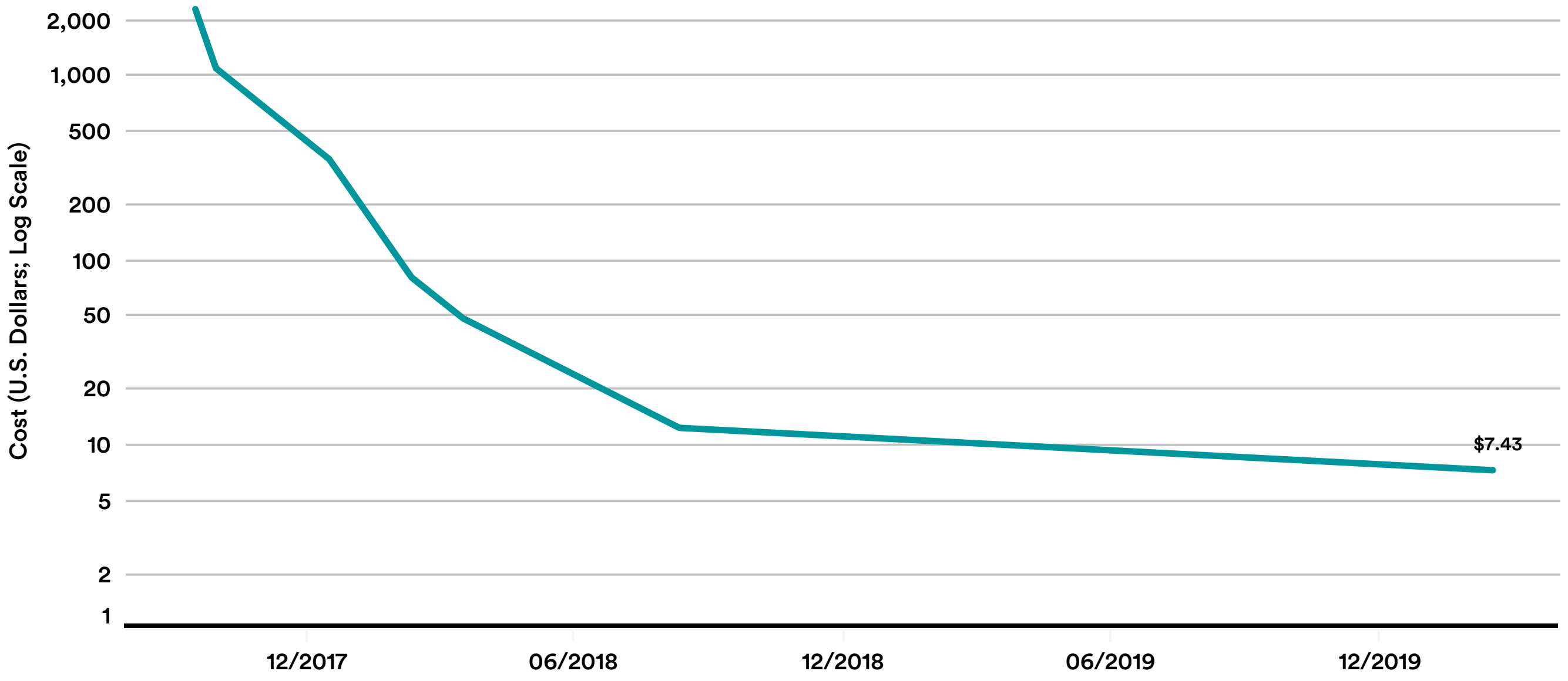

Figure 2.1.5

## Harder Tests Beyond ImageNet

In spite of the progress in performance on ImageNet, current computer vision systems are still not perfect. To better study their limitations, researchers have in recent years started to develop more challenging image classification benchmarks. But since ImageNet is already a large dataset, which requires a nontrivial amount of resources to use, it does not intuitively make sense to simply expand the resolution of the images in ImageNet or the absolute size of the dataset—as either action would further increase the cost to researchers when training systems on ImageNet. Instead, people have tried to figure out new ways to test the robustness of image classifiers by creating custom datasets, many of which are compatible with ImageNet (and are typically smaller). These include

**IMAGENET ADVERSARIAL:** This is a dataset of images similar to those found in ImageNet but incorporating natural confounders (e.g., a butterfly sitting on a carpet with a similar texture to the butterfly), and images that are persistently misclassified by contemporary systems. These images "cause consistent classification mistakes due to scene complications encountered in the long tail of scene configurations and by exploiting classifier blind spots," according to the researchers. Therefore, making progress on ImageNet Adversarial could improve the ability of models to generalize.

**IMAGENET-C:** This is a dataset of common ImageNet images with 75 visual corruptions applied to them (e.g., changes in brightness, contrast, pixelations, fog effects, etc.). By testing systems against this, researchers can provide even more information about the generalization capabilities of these models.

**IMAGENET-RENDITION:** This tests generalization by seeing how well ImageNet-trained models can categorize 30,000 illustrations of 200 ImageNet classes. Since ImageNet is designed to be built out of photos, generalization here indicates that systems have learned something more subtle about what they're trying to classify, because they're able to "understand" the relationship between illustrations and the photographed images they've been trained on.

**What is the Time Table for Tracking This Data?** As these benchmarks are relatively new, the plan is to wait a couple of years for the community to test a range of systems against them, which will generate the temporal information necessary to make graphs tracking progress overtime.

## IMAGE GENERATION

Image generation is the task of generating images that look indistinguishable from "real" images. Image generation systems have a variety of uses, ranging from augmenting search capabilities (it is easier to search for a specific image if you can generate other images like it) to serving as an aid for other generative uses (e.g., editing images, creating content for specific purposes, generating multiple variations of a single image to help designers brainstorm, and so on).

In recent years, image generation progress has accelerated as a consequence of the continued improvement in deep learning–based algorithms, as well as the use of increased computation and larger datasets.

### STL-10: Fréchet Inception Distance (FID) Score

One way to measure progress in image generation is via a technique called Fréchet Inception Distance (FID), which roughly correlates to the difference between how a given AI system "thinks" about a synthetic image versus a real image, where a real image has a score of 0 and synthetic images that look similar have scores that approach 0.

Figure 2.1.6 shows the progress of generative models over the past two years at generating convincing synthetic images in the STL-10 dataset, which is designed to test how effective systems are at generating images and gleaning other information about them.

STL-10: FRÉCHET INCEPTION DISTANCE (FID) SCORE
Source: Papers with Code, 2020 | Chart: 2021 AI Index Report

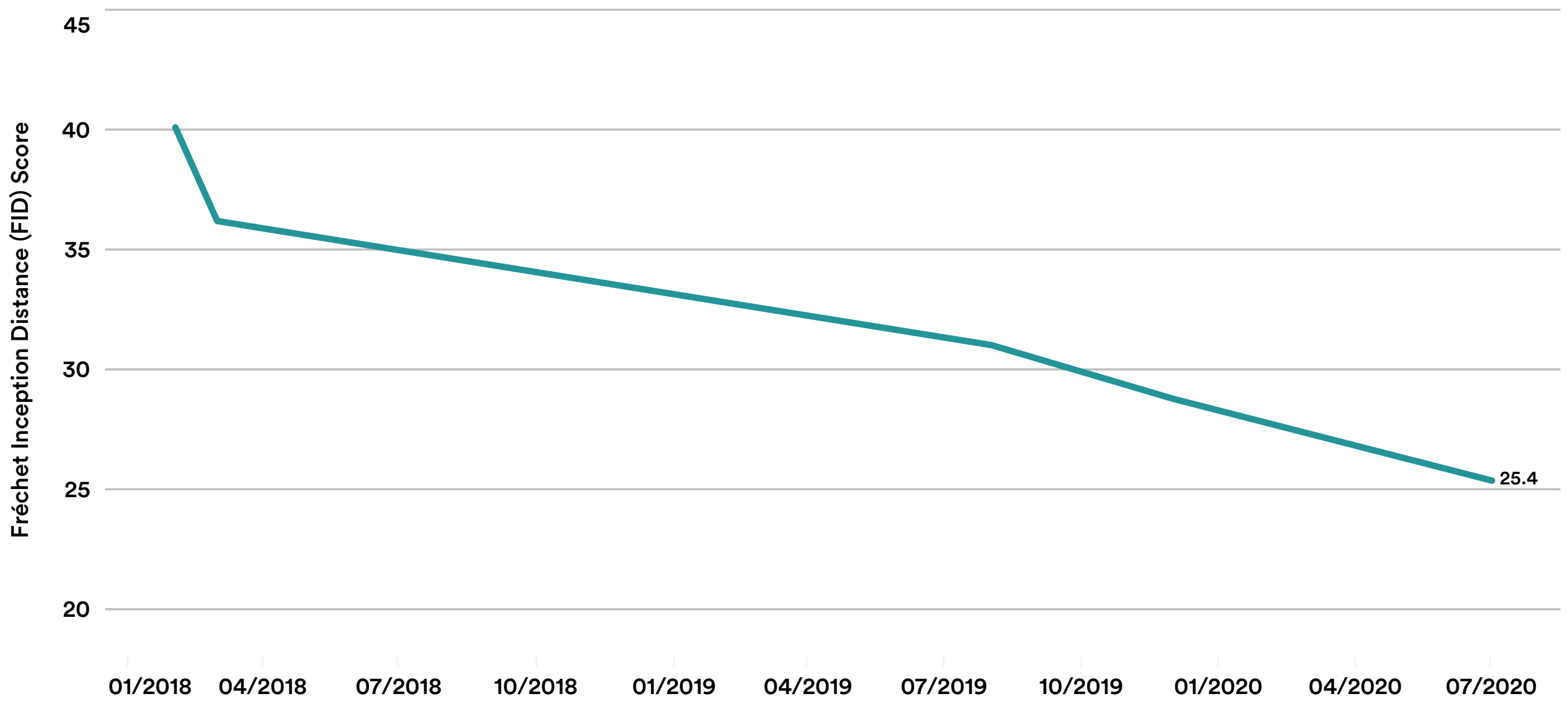

Figure 2.1.6

## FID Versus Real Life

FID has drawbacks as an evaluation technique—specifically, it assesses progress on image generation via quantitative metrics that use data from the model itself, rather than other evaluation techniques. Another approach is using teams of humans to evaluate the outputs of these models; for instance, the Human eYe Perceptual Evaluation (HYPE) method tries to judge image quality by showing synthetically generated images to humans and using their qualitative ratings to drive the evaluation methodology. This approach is more expensive and slower to run than typical evaluations, but it may become more important as generative models get better.

**Qualitative Examples:** To get a sense of progress, you can look at the evolution in the quality of synthetically generated images over time. In Figure 2.1.7, you can see the best-in-class examples of synthetic images of human faces, ordered over time. By 2018, performance of this task had become sufficiently good that it is difficult for humans to easily model further progress (though it is possible to train machine learning systems to spot fakes, it is becoming more challenging). This provides a visceral example of recent progress in this domain and underscores the need for new evaluation methods to gauge future progress. In addition, in recent years people have turned to doing generative modeling on a broader range of categories than just images of people's faces, which is another way to test for generalization.

### GAN PROGRESS ON FACE GENERATION

Source: Goodfellow et al., 2014; Radford et al., 2016; Liu & Tuzel, 2016; Karras et al., 2018; Karras et al., 2019; Goodfellow, 2019; Karras et al., 2020; AI Index, 2021

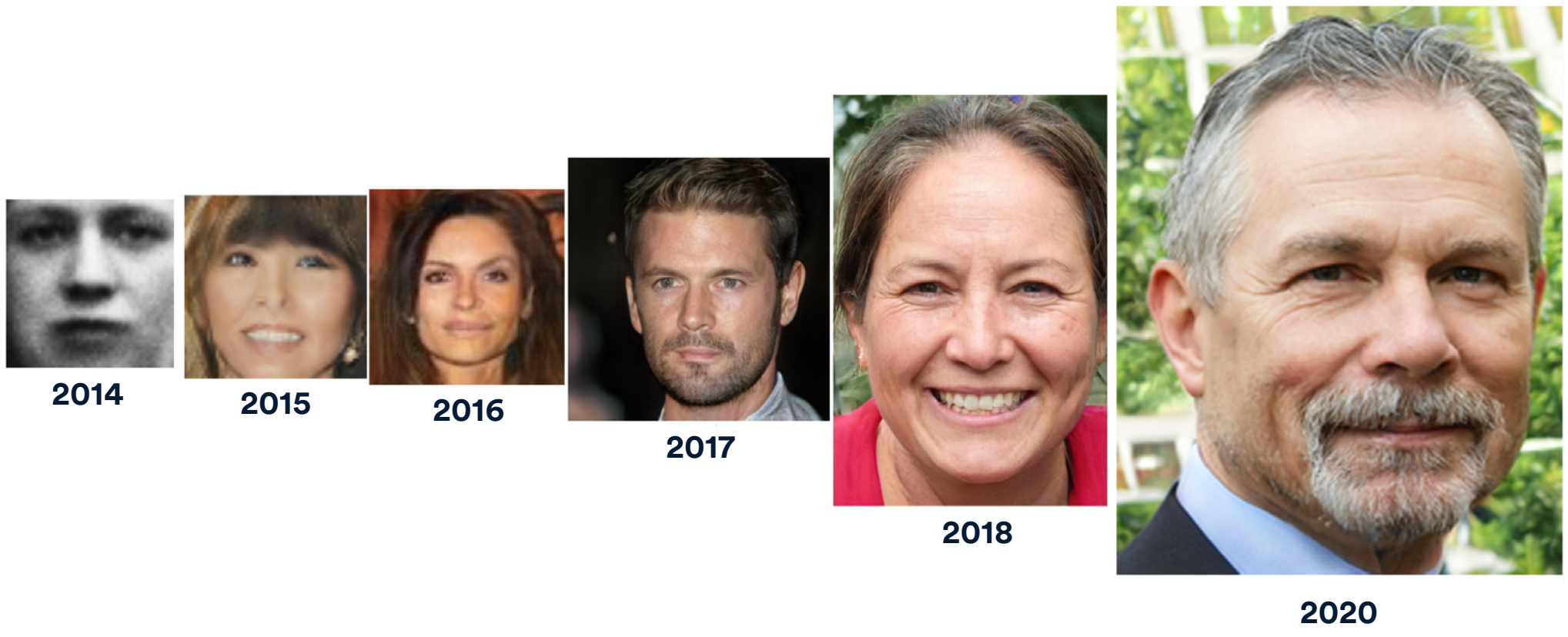

Figure 2.1.7

## DEEPFAKE DETECTION

Advances in image synthesis have created new opportunities as well as threats. For instance, in recent years, researchers have harnessed breakthroughs in synthetic imagery to create AI systems that can generate synthetic images of human faces, then superimpose those faces onto the faces of other people in photographs or movies. People call this application of generative technology a "deepfake." Malicious uses of deepfakes include misinformation and the creation of (predominantly misogynistic) pornography. To try to combat this, researchers are developing deepfake-detection technologies.

### Deepfake Detection Challenge (DFDC)

Created in September 2019 by Facebook, the Deepfake Detection Challenge (DFDC) measures progress on deepfake-detection technology. A two-part challenge, DFDC asks participants to train and test their models from a public dataset of around 100,000 clips. The submissions are scored on log loss, a classification metric based on probabilities. A smaller log loss means a more accurate prediction of deepfake videos. According to Figure 2.1.8, log loss dropped by around 0.5 as the challenge progressed between December 2019 and March 2020.

DEEPFAKE DETECTION CHALLENGE: LOG LOSS

Source: Kaggle, 2020 | Chart: 2021 AI Index Report

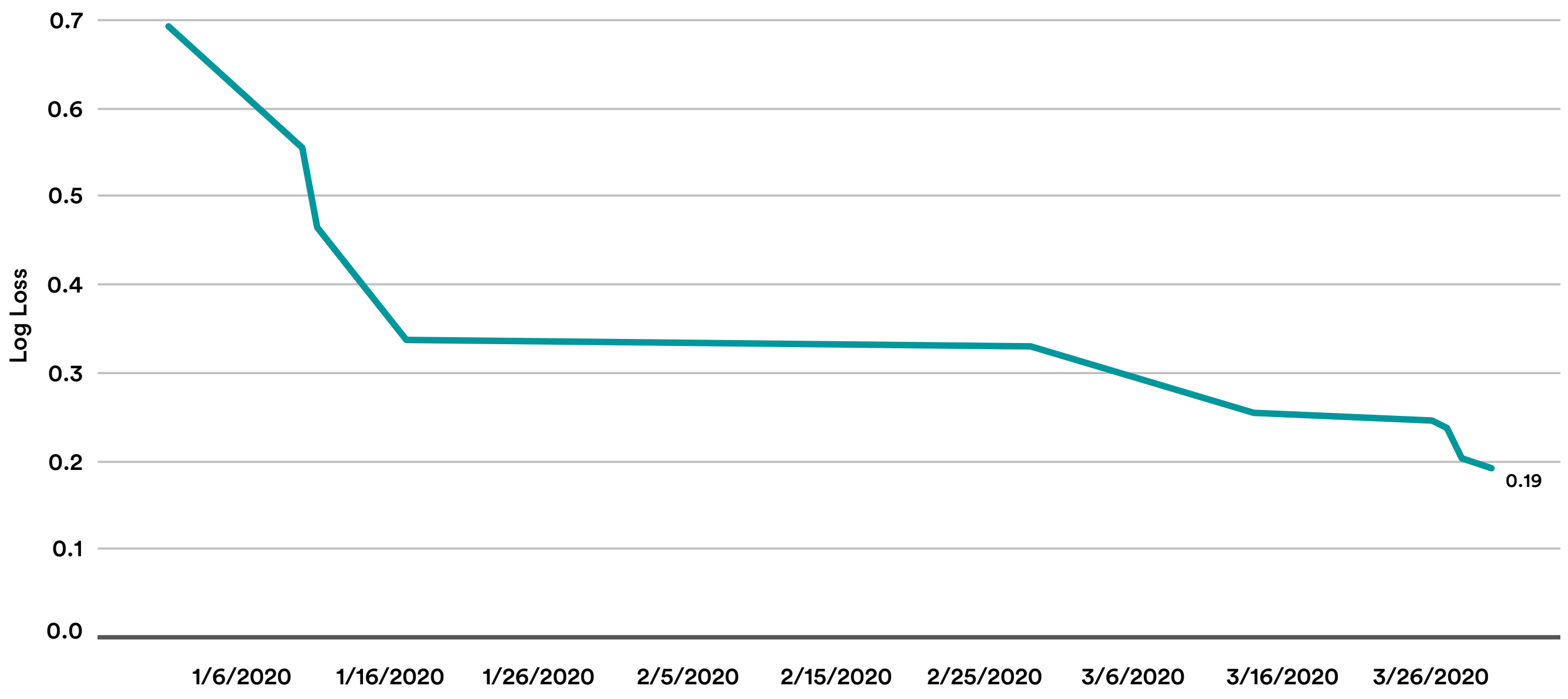

Figure 2.1.8

## HUMAN POSE ESTIMATION

Human pose estimation is the problem of estimating the positions of human body parts or joints (wrists, elbows, etc.) from a single image. Human pose estimation is a classic "omni-use" AI capability. Systems that are good at this task can be used for a range of applications, such as creating augmented reality applications for the fashion industry, analyzing behaviors gleaned from physical body analysis in crowds, surveilling people for specific behaviors, aiding with analysis of live sporting and athletic events, mapping the movements of a person to a virtual avatar, and so on.

### Common Objects in Context (COCO): Keypoint Detection Challenge

Common Objects in Context (COCO) is a large-scale dataset for object detection, segmentation, and captioning with 330,000 images and 1.5 million object instances. Its Keypoint Detection Challenge requires machines to simultaneously detect an object or a person and localize their body keypoints—points in the image that stand out, such as a person's elbows, knees, and other joints. The task evaluates algorithms based on average precision (AP), a metric that can be used to measure the accuracy of object detectors. Figure 2.1.9 shows that the accuracy of algorithms in this task has improved by roughly 33% in the past four years, with the latest machine scoring 80.8% on average precision.

**COCO KEYPOINT CHALLENGE: AVERAGE PRECISION**
Source: COCO Leaderboard, 2020 | Chart: 2021 AI Index Report

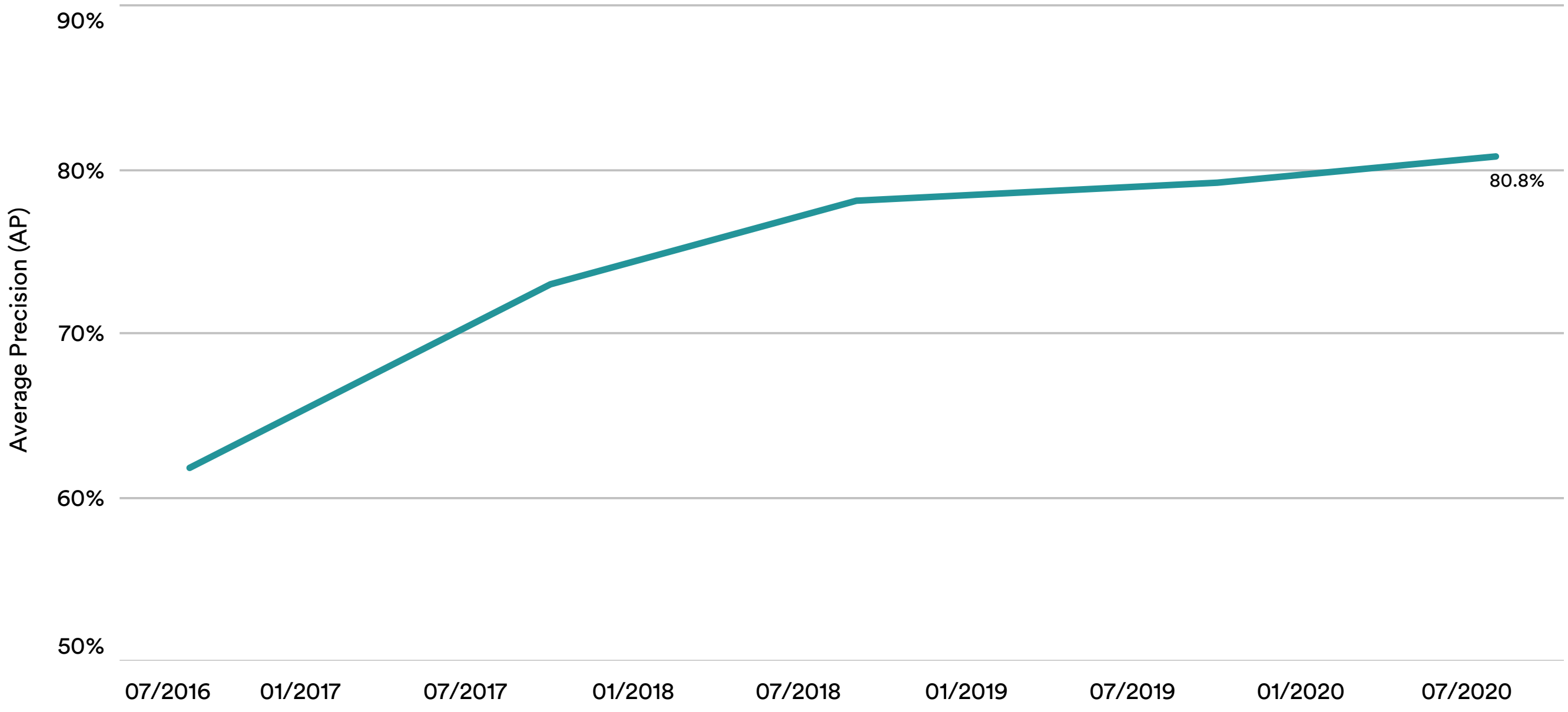

Figure 2.1.9

## Common Objects in Context (COCO): DensePose Challenge

DensePose, or dense human pose estimation, is the task of extracting a 3D mesh model of a human body from a 2D image. After open-sourcing a system called DensePose in 2018, Facebook built DensePose-COCO, a large-scale dataset of image-to-surface correspondences annotated on 50,000 COCO images. Since then, DensePose has become a canonical benchmark dataset.

The COCO DensePose Challenge involves tasks of simultaneously detecting people, segmenting their bodies, and estimating the correspondences between image pixels that belong to a human body and a template 3D model. The average precision is calculated based on the geodesic point similarity (GPS) metric, a correspondence matching score that measures the geodesic distances between the estimated points and the true location of the body points on the image. The accuracy has grown from 56% in 2018 to 72% in 2019 (Figure 2.1.10).

COCO DENSEPOSE CHALLENGE: AVERAGE PRECISION
Source: arXiv & CodaLab, 2020 | Chart: 2021 AI Index Report

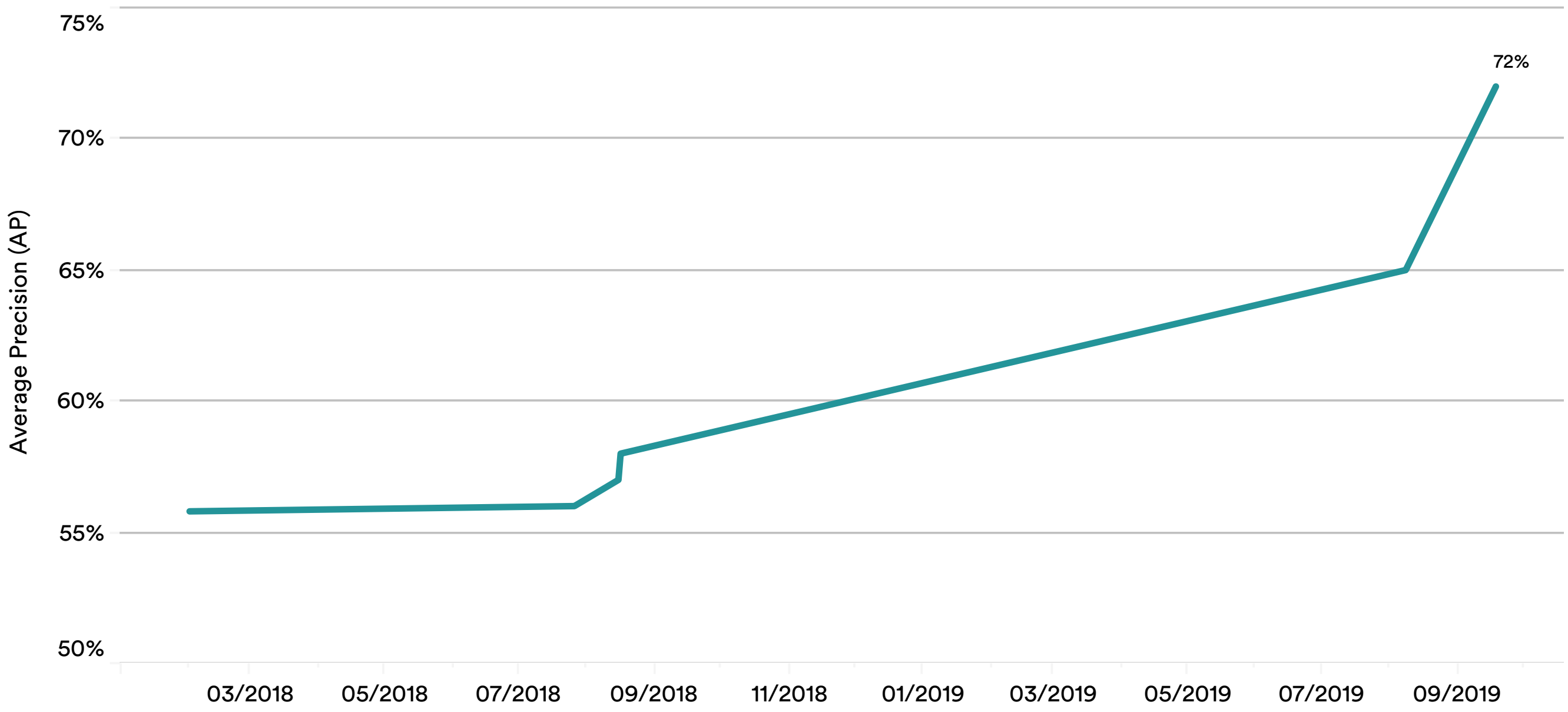

Figure 2.1.10

## SEMANTIC SEGMENTATION

Semantic segmentation is the task of classifying each pixel in an image to a particular label, such as person, cat, etc. Where image classification tries to assign a label to the entire image, semantic segmentation tries to isolate the distinct entities and objects in a given image, allowing for more fine-grained identification. Semantic segmentation is a basic input technology for self-driving cars (identifying and isolating objects on roads), image analysis, medical applications, and more.

### Cityscapes

Cityscapes is a large-scale dataset of diverse urban street scenes across 50 different cities recorded during the daytime over several months (during spring, summer, and fall) of the year. The dataset contains 5,000 images with high-quality, pixel-level annotations and 20,000 weekly labeled ones. Semantic scene understanding, especially in the urban space, is crucial to the environmental perception of autonomous vehicles. Cityscapes is useful for training deep neural networks to understand the urban environment.

One Cityscapes task that focuses on semantic segmentation is the pixel-level semantic labeling task. This task requires an algorithm to predict the per-pixel semantic labeling of the image, partitioning an image into different categories, like cars, buses, people, trees, and roads. Participants are evaluated based on the intersection-over-union (IoU) metric. A higher IoU score means a better segmentation accuracy. Between 2014 and 2020, the mean IoU increased by 35% (Figure 2.1.11). There was a significant boost to progress in 2016 and 2017 when people started using residual networks in these systems.

CITYSCAPES CHALLENGE: PIXEL-LEVEL SEMANTIC LABELING TASK

Source: Papers with Code, 2020 | Chart: 2021 AI Index Report

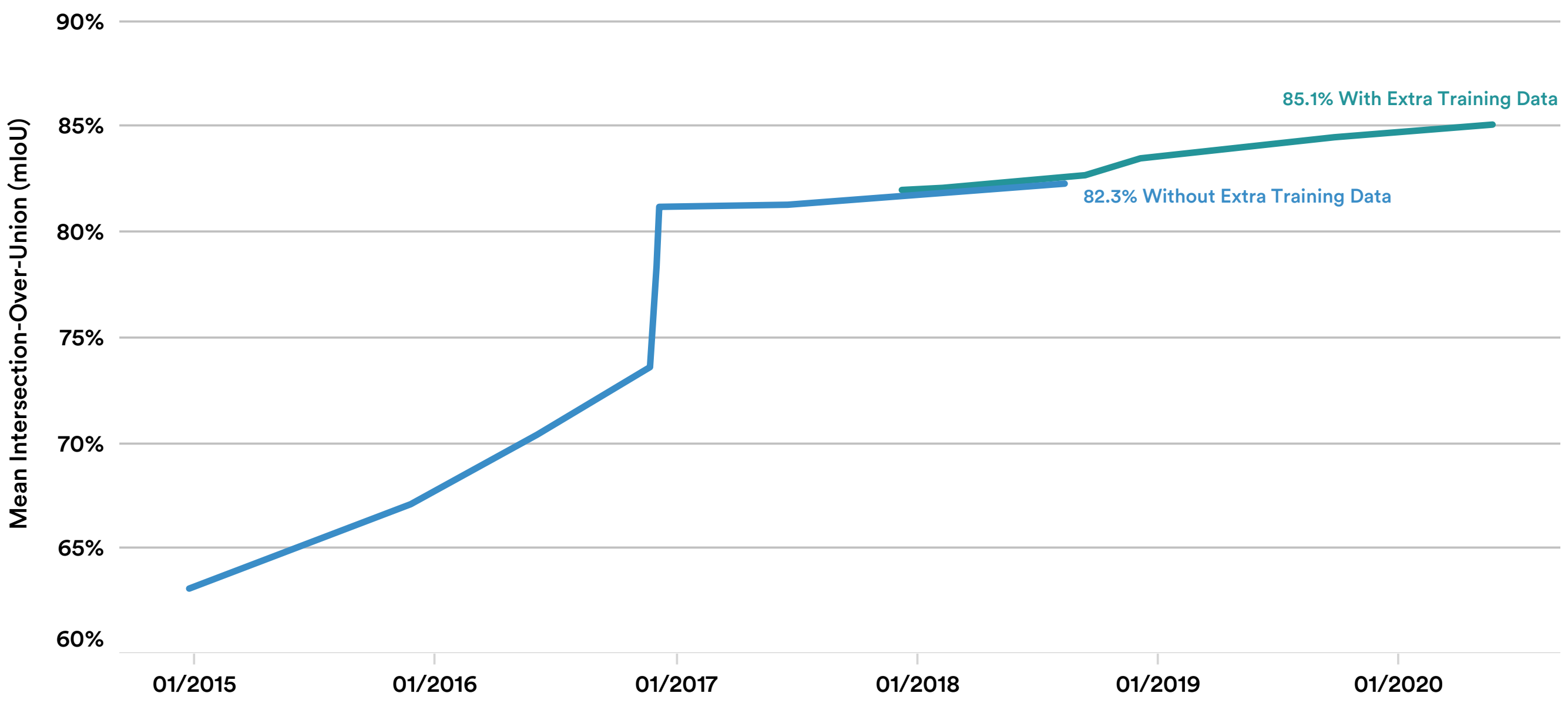

Figure 2.1.11

## EMBODIED VISION

The performance data so far shows that computer vision systems have advanced tremendously in recent years. Object recognition, semantic segmentation, and human pose estimation, among others, have now achieved significant levels of performance. Note that these visual tasks are somewhat passive or disembodied. That is, they can operate on images or videos taken from camera systems that are not physically able to interact with the surrounding environment. As a consequence of the continuous improvement in those passive tasks, researchers have now started to develop more advanced AI systems that can be interactive or embodied—that is, systems that can physically interact with and modify the surrounding environment in which they operate: for example, a robot that can visually survey a new building and autonomously navigate it, or a robot that can learn to assemble pieces by watching visual demonstrations instead of being manually programmed for this.

Progress in this area is currently driven by the development of sophisticated simulation environments, where researchers can deploy robots in virtual spaces, simulate what their cameras would see and capture, and develop AI algorithms for navigation, object search, and object grasping, among other interactive tasks. Because of the relatively early nature of this field, there are few standardized metrics to measure progress. Instead, here are brief highlights of some of the available simulators, their year of release, and any other significant feature.

- **Thor** (AI2, **2017**) focuses on sequential abstract reasoning with predefined "magic" actions that are applicable to objects.
- **Gibson** (Stanford, **2018**) focuses on visual navigation in photorealistic environments obtained with 3D scanners.
- **iGibson** (Stanford, **2019**) focuses on full interactivity in large realistic scenes mapped from real houses and made actable: navigation + manipulation (known in robotics as "mobile manipulation").
- **AI Habitat** (Facebook, **2019**) focuses on visual navigation with an emphasis on much faster execution, enabling more computationally expensive approaches.
- **ThreeDWorld** (MIT and Stanford, **2020**) focuses on photorealistic environments through game engines, as well as adds simulation of flexible materials, fluids, and sounds.
- **SEAN-EP** (Yale, **2020**) is a human-robot interaction environment with simulated virtual humans that enables the collection of remote demonstrations from humans via a web browser.
- **Robosuite** (Stanford and UT Austin, **2020**) is a modular simulation framework and benchmark for robot learning.

Video analysis is the task of making inferences over sequential image frames, sometimes with the inclusion of an audio feed. Though many AI tasks rely on single-image inferences, a growing body of applications require computer vision machines to reason about videos. For instance, identifying a specific dance move benefits from seeing a variety of frames connected in a temporal sequence; the same is true of making inferences about an individual seen moving through a crowd, or a machine carrying out a sequence of movements over time.

# 2.2 COMPUTER VISION—VIDEO

## ACTIVITY RECOGNITION

The task of activity recognition is to identify various activities from video clips. It has many important everyday applications, including surveillance by video cameras and autonomous navigation of robots. Research on video understanding is still focused on short events, such as videos that are a few seconds long. Longer-term video understanding is slowly gaining traction.

### ActivityNet

Introduced in 2015, ActivityNet is a large-scale video benchmark for human-activity understanding. The benchmark tests how well algorithms can label and categorize human behaviors in videos. By improving performance on tasks like ActivityNet, AI researchers are developing systems that can categorize more complex behaviors than those that can be contained in a single image, like characterizing the behavior of pedestrians on a self-driving car's video feed or providing better labeling of specific movements in sporting events.

### ActivityNet: Temporal Action Localization Task

The temporal action localization task in the ActivityNet challenge asks machines to detect time segments in a 600-hour, untrimmed video sequence that contains several activities. Evaluation on this task focuses on (1) localization: how well can the system localize the interval with the precise start time and end time; and (2) recognition: how well can the system recognize the activity and classify it into the correct category (such as throwing, climbing, walking the dog, etc.). Figure 2.2.1 shows that the highest mean average precision of the temporal action localization task among submissions has grown by 140% in the last five years.

ACTIVITYNET: TEMPORAL ACTION LOCALIZATION TASK
Source: ActivityNet, 2020 | Chart: 2021 AI Index Report

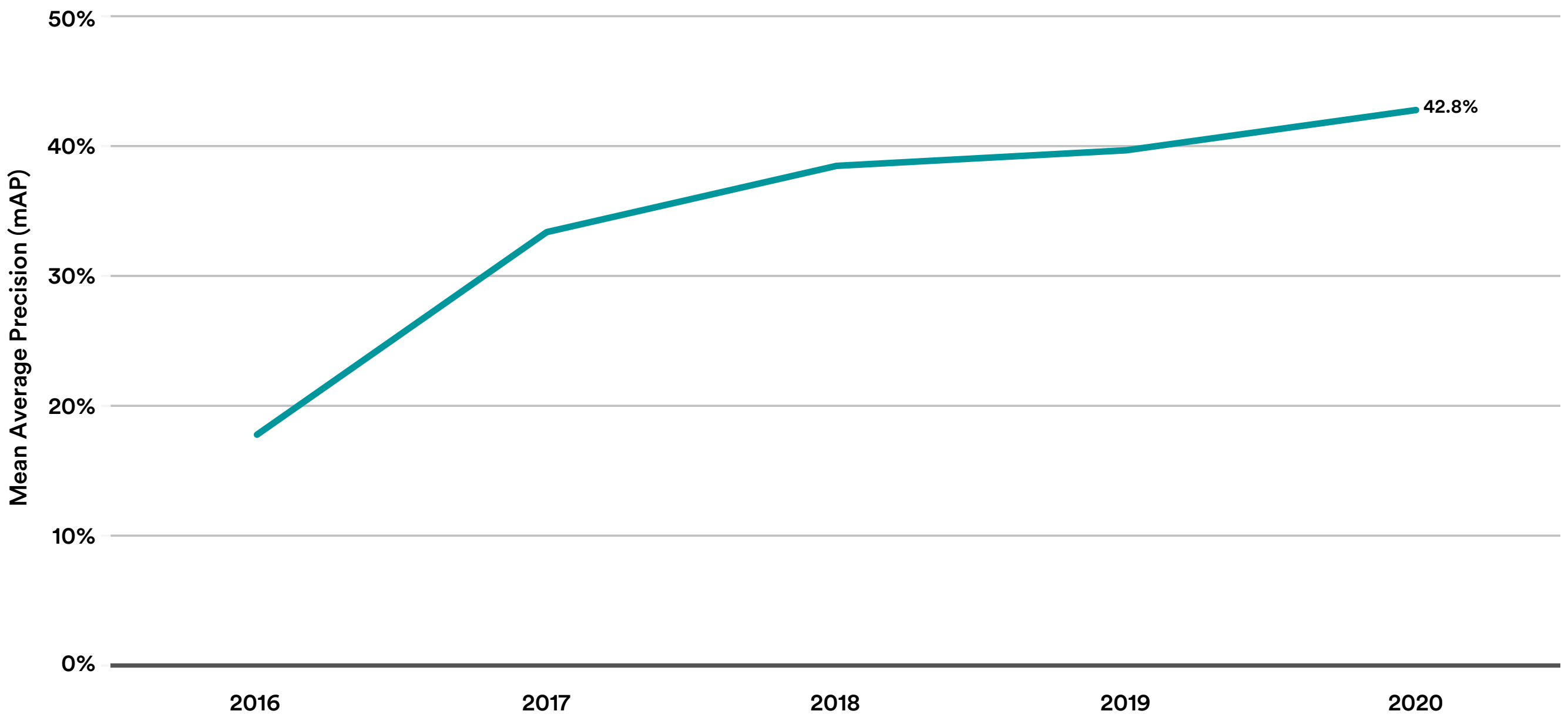

Figure 2.2.1

## ActivityNet: Hardest Activity

Figure 2.2.2 shows the hardest activities of the temporal action location task in 2020 and how their mean average precision compares with the 2019 result. Drinking coffee remained the hardest activity in 2020. Rock-paper-scissors, though still the 10th hardest activity, saw the greatest improvement among all activities, increasing by 129.2%—from 6.6% in 2019 to 15.22% in 2020.

ACTIVITYNET: HARDEST ACTIVITIES, 2019-20

Source: ActivityNet, 2020 | Chart: 2021 AI Index Report

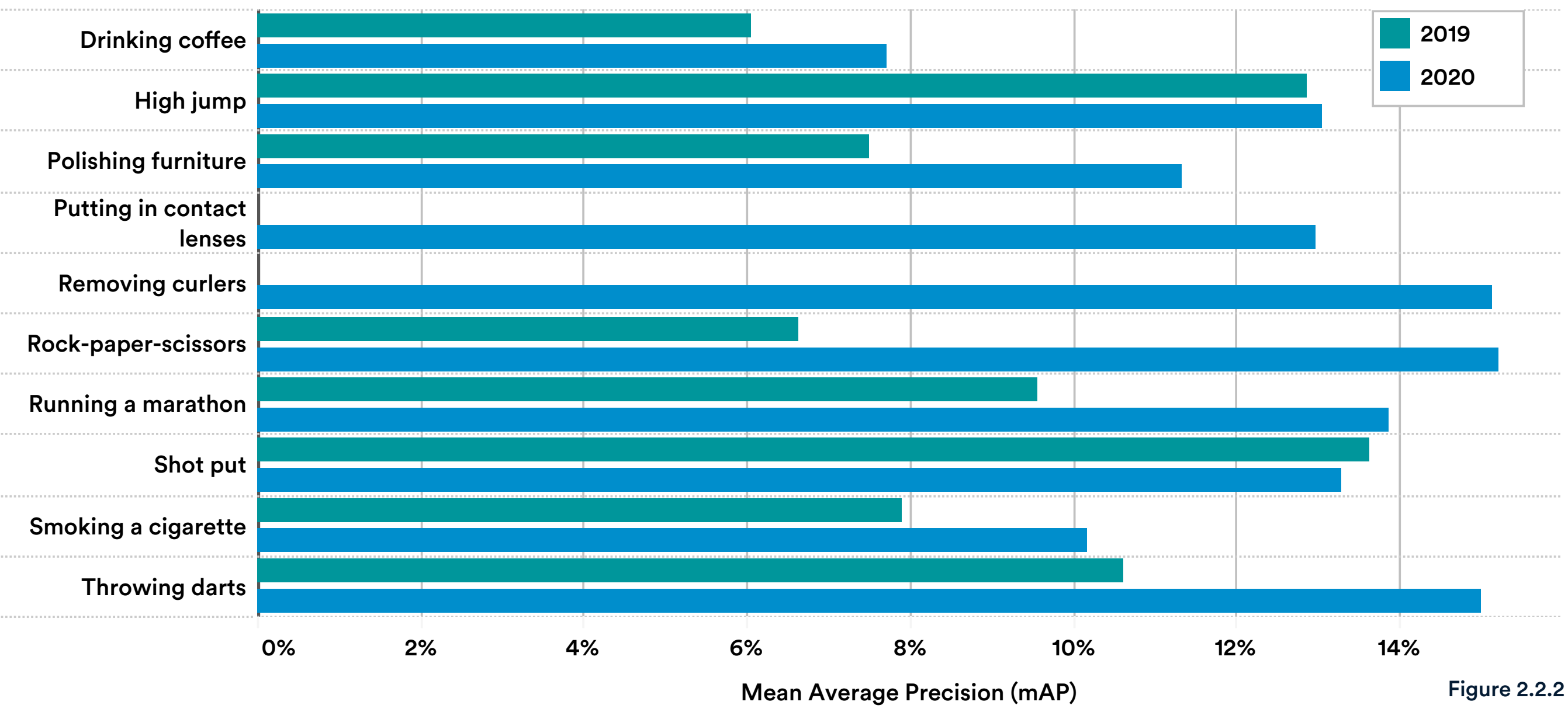

Figure 2.2.2

## OBJECT DETECTION

Object detection is the task of identifying a given object in an image. Frequently, image classification and image detection are coupled together in deployed systems. One way to get a proxy measure for the improvement in deployed object recognition systems is to study the advancement of widely used object detection systems.

### You Only Look Once (YOLO)

You Only Look Once (YOLO) is a widely used open source system for object detection, so its progress has been included on a standard task on YOLO variants to give a sense of how research percolates into widely used open source tools. YOLO has gone through multiple iterations since it was first published in 2015. Over time, YOLO has been optimized along two constraints: performance and inference latency, as shown in Figure 2.2.3. What this means, specifically, is that by measuring YOLO, one can measure the advancement of systems that might not have the best absolute performance but are designed around real-world needs, like low-latency inference over video streams. Therefore, YOLO systems might not always contain the absolute best performance as defined in the research literature, but they will represent good performance when faced with trade-offs such as inference time.

YOU ONLY LOOK ONCE (YOLO): MEAN AVERAGE PRECISION

Source: Redmon & Farhadi (2016 & 2018), Bochkovskiy et al. (2020), Long et al. (2020) | Chart: 2021 AI Index Report

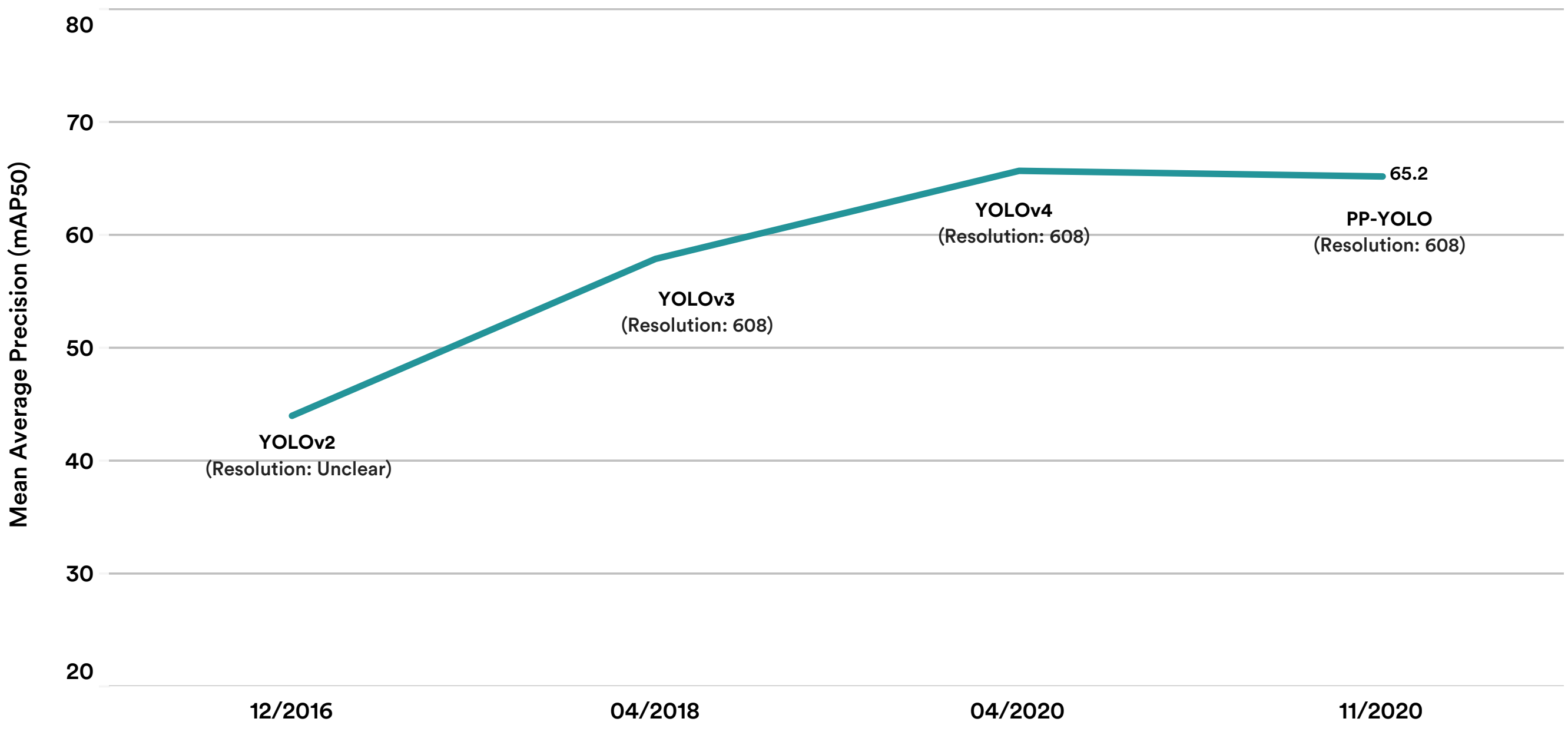

Figure 2.2.3

## FACE DETECTION AND RECOGNITION

Facial detection and recognition is one of the use-cases for AI that has a sizable commercial market and has generated significant interest from governments and militaries. Therefore, progress in this category gives us a sense of the rate of advancement in economically significant parts of AI development.

### National Institute of Standards and Technology (NIST) Face Recognition Vendor Test (FRVT)

The Face Recognition Vendor Tests (FRVT) by the National Institute of Standards and Technology (NIST) provide independent evaluations of commercially available and prototype face recognition technologies. FRVT measures the performance of automated face recognition technologies used for a wide range of civil and governmental tasks (primarily in law enforcement and homeland security), including verification of visa photos, mug shot images, and child abuse images.

Figure 2.2.4 shows the results of the top-performing 1:1 algorithms measured on false non-match rate (FNMR) across several different datasets. FNMR refers to the rate at which the algorithm fails when attempting to match the image with the individual. Facial recognition technologies on mug-shot-type and visa photos have improved the most significantly in the past four years, falling from error rates of close to 50% to a fraction of a percent in 2020.[4]

NIST FRVT 1:1 VERIFICATION ACCURACY by DATASET, 2017-20

Source: National Institute of Standards and Technology, 2020 | Chart: 2021 AI Index Report

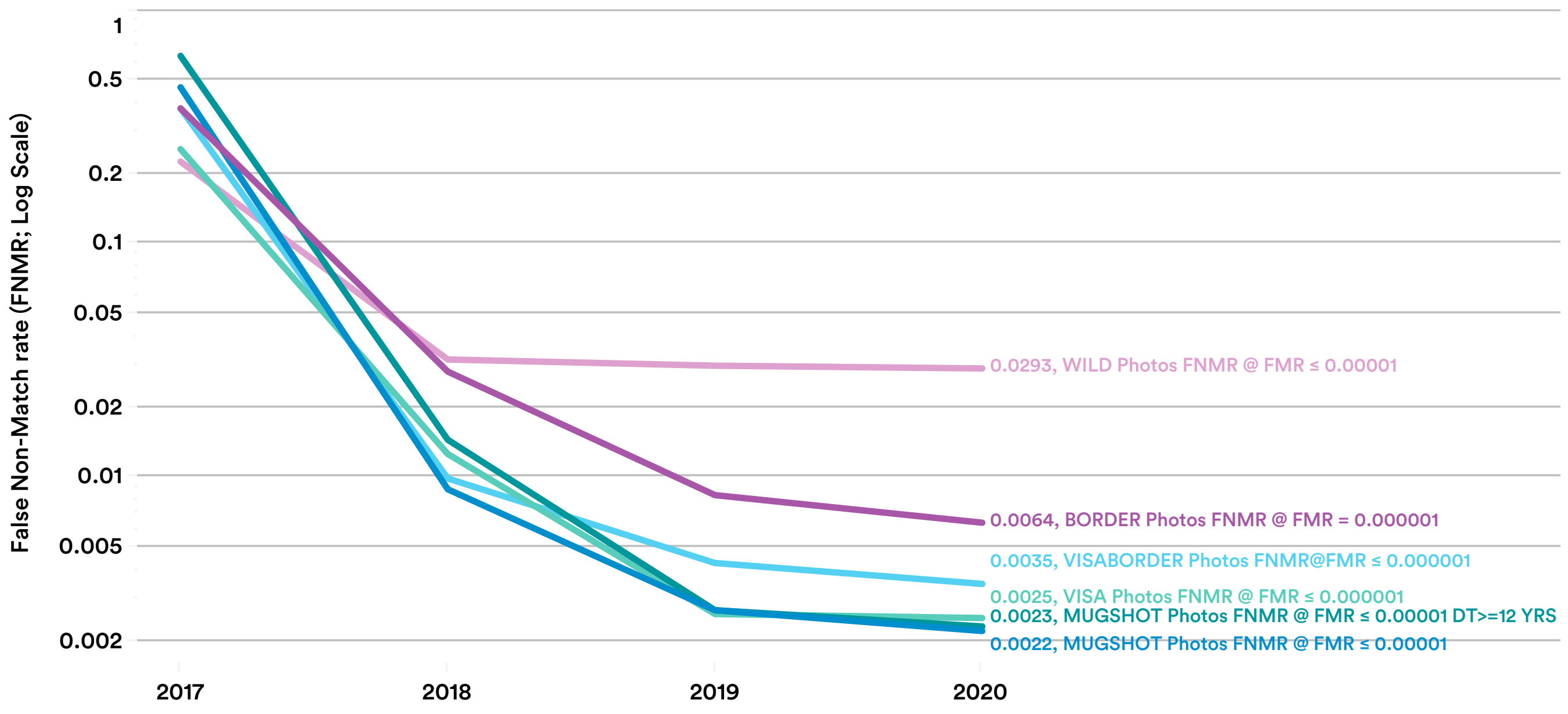

Figure 2.2.4

4 You can view details and examples of various datasets on periodically updated FRVT 1:1 verification reports.

Natural language processing (NLP) involves teaching machines to interpret, classify, manipulate, and generate language. From the early use of handwritten rules and statistical techniques to the recent adoption of generative models and deep learning, NLP has become an integral part of our lives, with applications in text generation, machine translation, question answering, and other tasks.

# 2.3 LANGUAGE

In recent years, advances in natural language processing technology have led to significant changes in large-scale systems that billions of people access. For instance, in late 2019, Google started to deploy its BERT algorithm into its search engine, leading to what the company said was a significant improvement in its in-house quality metrics. Microsoft followed suit, announcing later in 2019 that it was using BERT to augment its Bing search engine.

## ENGLISH LANGUAGE UNDERSTANDING BENCHMARKS

### SuperGLUE

Launched in May 2019, SuperGLUE is a single-metric benchmark that evaluates the performance of a model on a series of language understanding tasks on established datasets. SuperGLUE replaced the prior GLUE benchmark (introduced in 2018) with more challenging and diverse tasks.

The SuperGLUE score is calculated by averaging scores on a set of tasks. Microsoft's DeBERTa model now tops the SuperGLUE leaderboard, with a score of 90.3, compared with an average score of 89.8 for SuperGLUE's "human baselines." This does not mean that AI systems have surpassed human performance on all SuperGLUE tasks, but it does mean that the average performance across the entire suite has exceeded that of a human baseline. The rapid pace of progress (Figure 2.3.1) suggests that SuperGLUE may need to be made more challenging or replaced by harder tests in the future, just as SuperGLUE replaced GLUE.

SUPERGLUE BENCHMARK
Source: SuperGLUE Leaderboard, 2020 | Chart: 2021 AI Index Report

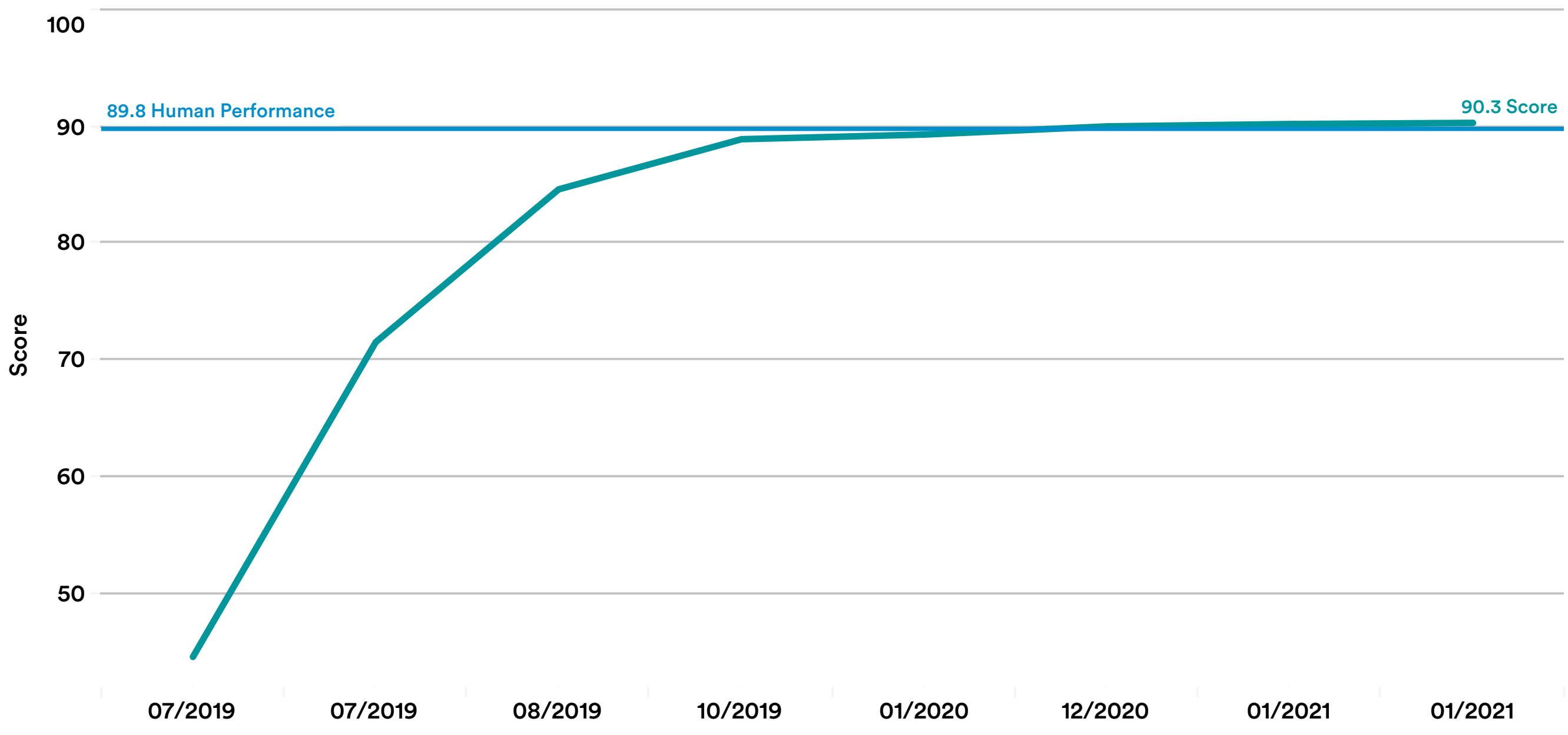

Figure 2.3.1

## SQuAD

The Stanford Question Answering Dataset, or SQuAD, is a reading-comprehension benchmark that measures how accurately a NLP model can provide short answers to a series of questions pertaining to a small article of text. The SQuAD test makers established a human performance benchmark by having a group of people read Wikipedia articles on a variety of topics and then answer multiple-choice questions about those articles. Models are given the same task and are evaluated on the F1 score, or the average overlap between the model prediction and the correct answer. Higher scores indicate better performance.

Two years after the introduction of the original SQuAD, in 2016, SQuAD 2.0 was developed once the initial benchmark revealed increasingly fast performances by the participants (mirroring the trend seen in GLUE and SuperGLUE). SQuAD 2.0 combines the 100,000 questions in SQuAD 1.1 with over 50,000 unanswerable questions written by crowdworkers to resemble answerable ones. The objective is to test how well systems can answer questions and to determine when systems know that no answer exists.

As Figure 2.3.2 shows, the F1 score for SQuAD 1.1 improved from 67.75 in August 2016 to surpass human performance of 91.22 in September 2018—a 25-month period—whereas SQuAD 2.0 took just 10 months to beat human performance (from 66.3 in May 2018 to 89.47 in March 2019). In 2020, the most advanced models of SQuAD 1.1 and SQuAD 2.0 reached the F1 scores of 95.38 and 93.01, respectively.

**SQUAD 1.1 and SQUAD 2.0: F1 SCORE**
Source: CodaLab Worksheets, 2020 | Chart: 2021 AI Index Report

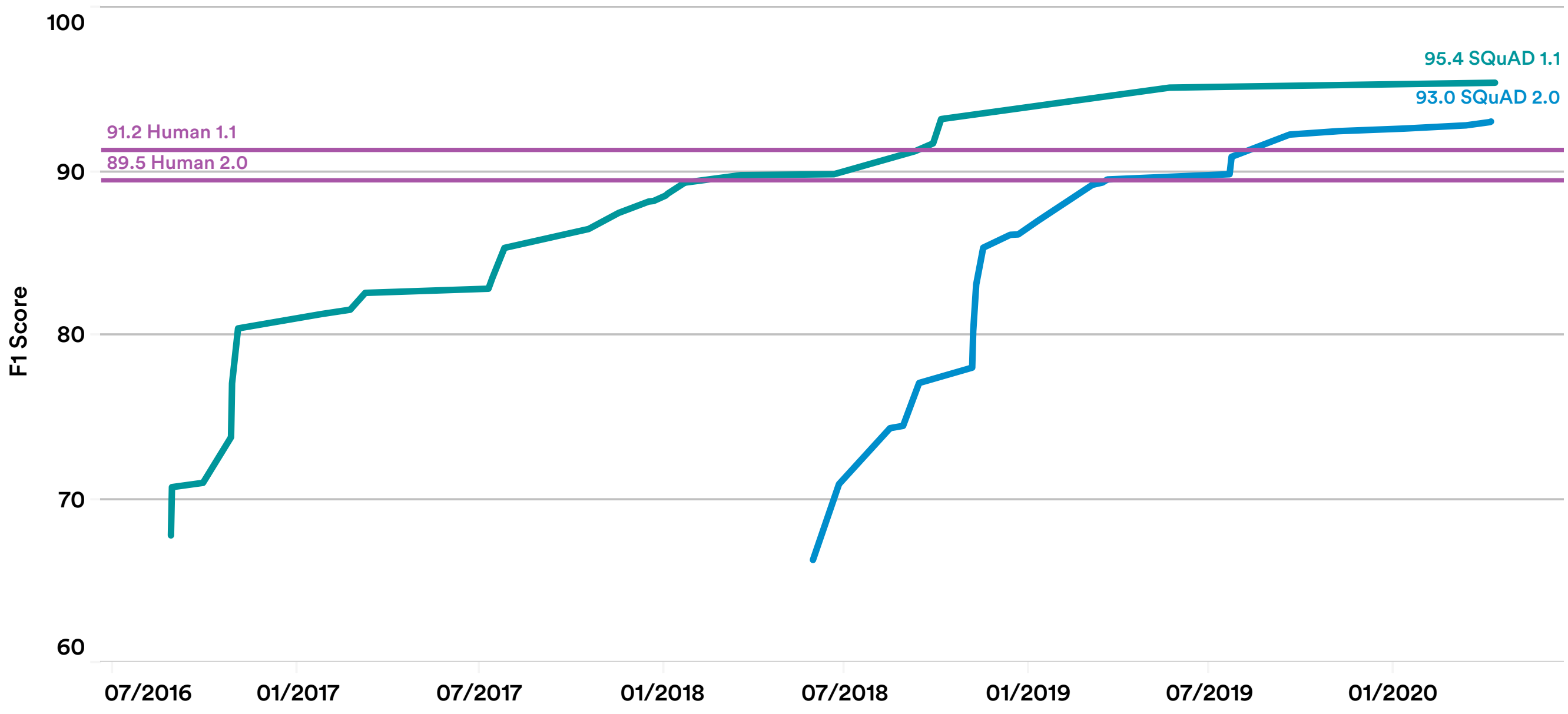

Figure 2.3.2

## COMMERCIAL MACHINE TRANSLATION (MT)

Machine translation (MT), the subfield of computational linguistics that investigates the use of software to translate text or speech from one language to another, has seen significant improvement due to advances in machine learning. Recent progress in MT has prompted developers to shift from symbolic approaches toward ones that use both statistical and deep learning approaches.

### Number of Commercially Available MT Systems

The trend in the number of commercially available systems speaks to the significant growth of commercial machine translation technology and its rapid adoption in the commercial marketplace. In 2020, the number of commercially available independent cloud MT systems with pre-trained models increased to 28, from 8 in 2017, according to Intento, a startup that evaluates commercially available MT services (Figure 2.3.3).

NUMBER of INDEPENDENT MACHINE TRANSLATION SERVICES

Source: Intento, 2020 | Chart: 2021 AI Index Report

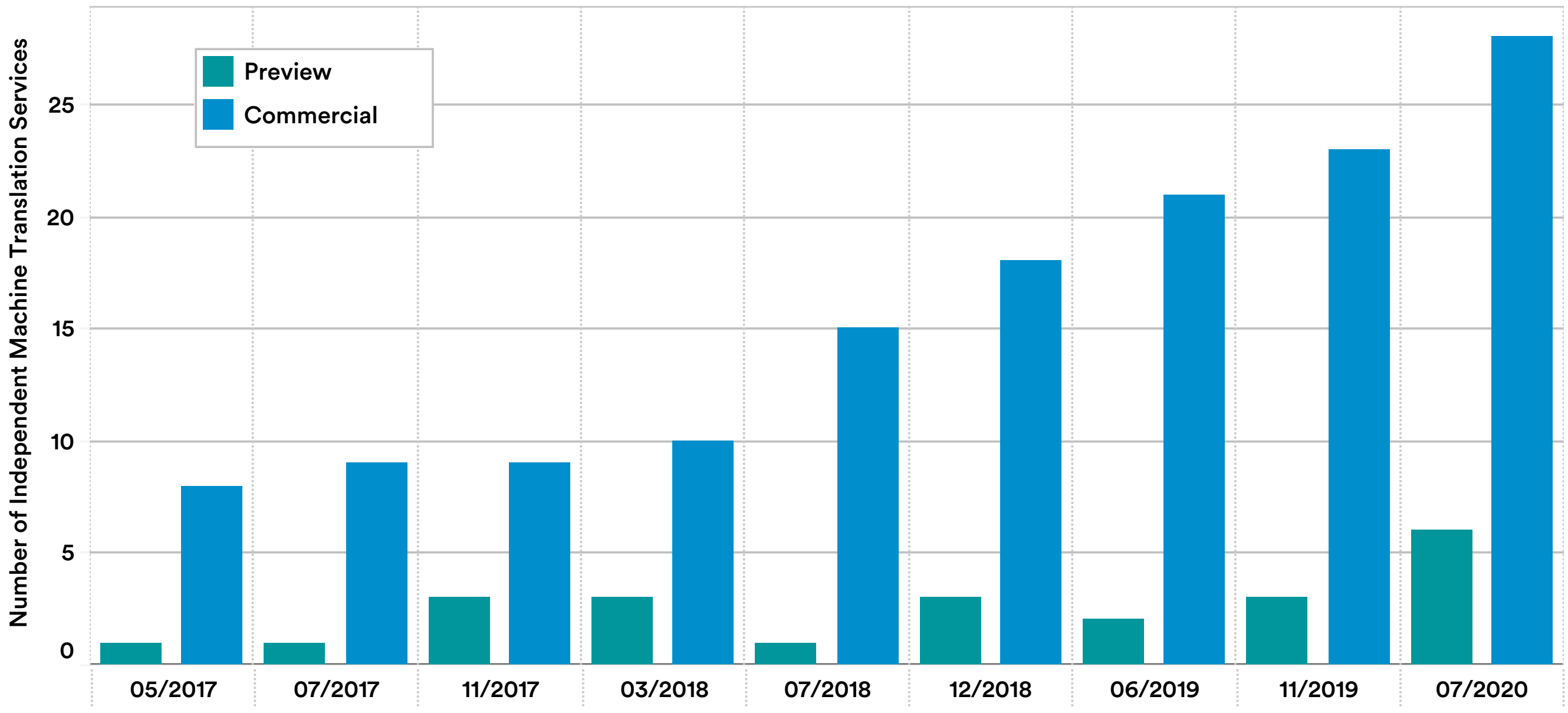

Figure 2.3.3

## GPT-3

In July 2020, OpenAI unveiled GPT-3, the largest known dense language model. GPT-3 has 175 billion parameters and was trained on 570 gigabytes of text. For comparison, its predecessor, GPT-2, was over 100 times smaller, at 1.5 billion parameters. This increase in scale leads to surprising behavior: GPT-3 is able to perform tasks it was not explicitly trained on with zero to few training examples (referred to as zero-shot and few-shot learning, respectively). This behavior was mostly absent in the much smaller GPT-2. Furthermore, for some tasks (but not all; e.g., SuperGLUE and SQuAD2), GPT-3 outperforms state-of-the-art models that were explicitly trained to solve those tasks with far more training examples.

Figure 2.3.4, adapted from the GPT-3 paper, demonstrates the impact of scale (in terms of model parameters) on task accuracy (higher is better) in zero-, one-, and few-shot learning regimes. Each point on the curve corresponds to an average performance accuracy, aggregated across 42 accuracy-oriented benchmarks. As model size increases, average accuracy in all task regimes increases accordingly. Few-shot learning accuracy increases more rapidly with scale, compared with zero-shot learning, which suggests that large models can perform surprisingly well given minimal context.

**GPT-3: AVERAGE PERFORMANCE across 42 BENCHMARKS**

Source: OpenAI (Brown et al.), 2020 | Chart: 2021 AI Index Report

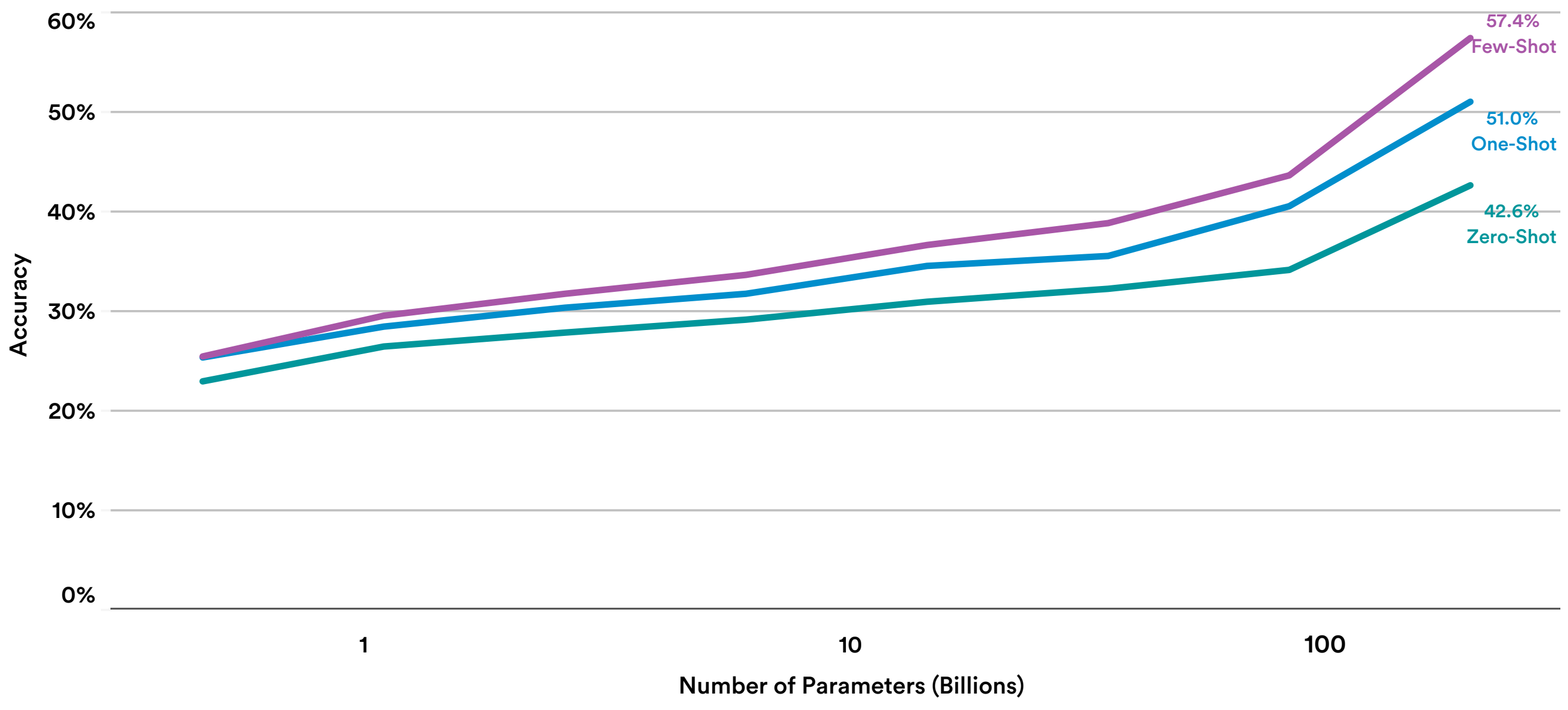

Figure 2.3.4

That a single model can achieve state-of-the-art or close to state-of-the-art performance in limited-training-data regimes is impressive. Most models until now have been designed for a single task, and thus can be evaluated effectively by a single metric. In light of GPT-3, we anticipate novel benchmarks that are explicitly designed to evaluate zero- to few-shot learning performance for language models. This will not be straightforward. Developers are increasingly finding model novel capabilities (e.g., the ability to generate a website from a text description) that will be difficult to define, let alone measure performance on. Nevertheless, the AI Index is committed to tracking performance in this new context as it evolves.

Despite its impressive capabilities, GPT-3 has several shortcomings, many of which are outlined in the original paper. For example, it can generate racist, sexist, and otherwise biased text. Furthermore, GPT-3 (and other language models) can generate unpredictable and factually inaccurate text. Techniques for controlling and "steering" such outputs to better align with human values are nascent but promising. GPT-3 is also expensive to train, which means that only a limited number of organizations with abundant resources can currently afford to develop and deploy such models. Finally, GPT-3 has an unusually large number of uses, from chatbots to computer code generation to search. Future users are likely to discover more applications, both good and bad, making it difficult to identify the range of possible uses and forecast their impact on society.

Nevertheless, research to address harmful outputs and uses is ongoing at several universities and industrial research labs, including OpenAI. For more details, refer to work by Bender and Gebru et al. and the proceedings from a recent Stanford Institute for Human-Centered Artificial Intelligence (HAI) workshop (which included researchers from OpenAI), "Understanding the Capabilities, Limitations, and Societal Impact of Large Language Models."

# 2.4 LANGUAGE REASONING SKILLS

## VISION AND LANGUAGE REASONING

Vision and language reasoning is a research area that addresses how well machines jointly reason about visual and text data.

### Visual Question Answering (VQA) Challenge

The VQA challenge, introduced in 2015, requires machines to provide an accurate natural language answer, given an image and a natural language question about the image based on a public dataset. Figure 2.4.1 shows that the accuracy has grown by almost 40% since its first installment at the International Conference on Computer Vision (ICCV) in 2015. The highest accuracy of the 2020 challenge is 76.4%. This achievement is closer to the human baseline of 80.8% accuracy and represents a 1.1% absolute increase in performance from the top 2019 algorithm.

VISUAL QUESTION ANSWERING (VQA) CHALLENGE: ACCURACY
Source: VQA Challenge, 2020 | Chart: 2021 AI Index Report

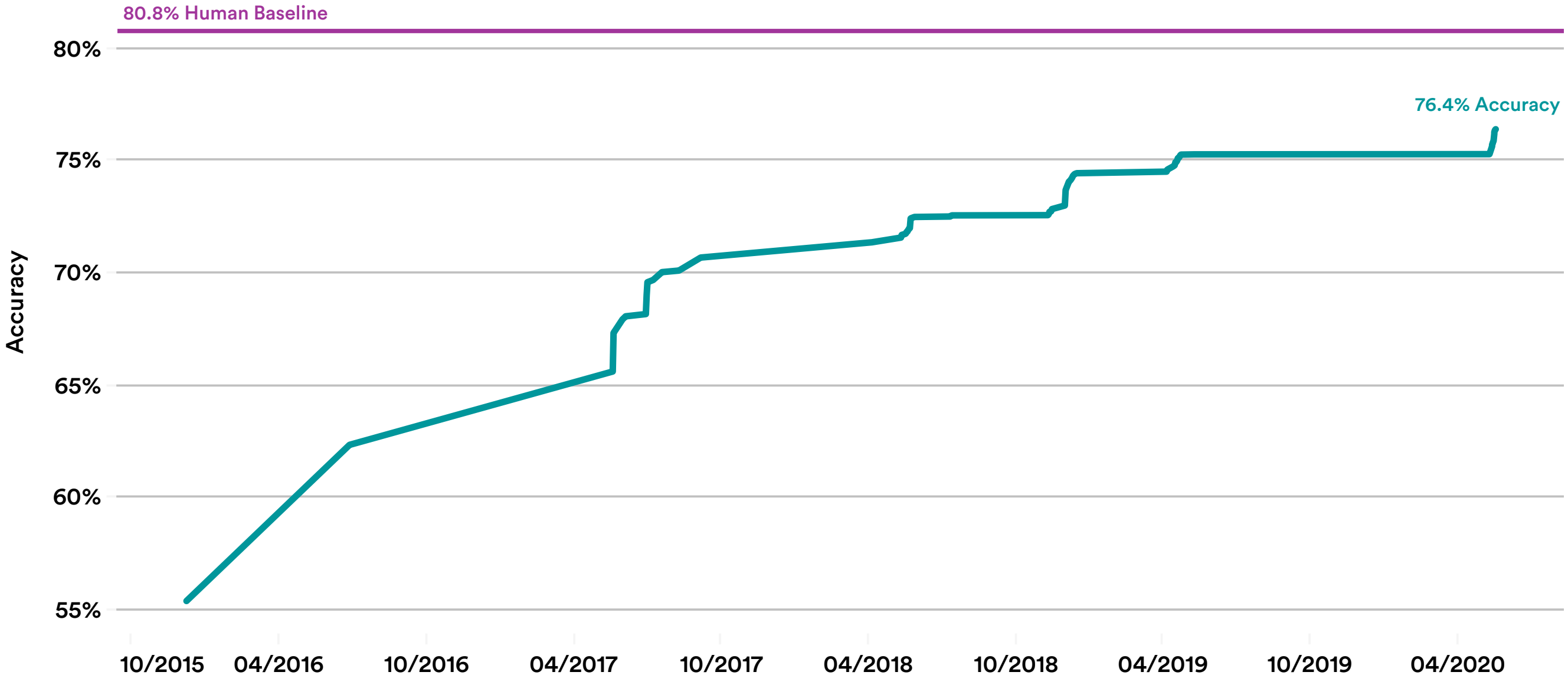

Figure 2.4.1

## Visual Commonsense Reasoning (VCR) Task

The Visual Commonsense Reasoning (VCR) task, first introduced in 2018, asks machines to answer a challenging question about a given image and justify that answer with reasoning (whereas VQA just requests an answer). The VCR dataset contains 290,000 pairs of multiple-choice questions, answers, and rationales, as well as over 110,000 images from movie scenes.

The main evaluation mode for the VCR task is the Q->AR score, requiring machines to first choose the right answer (A) to a question (Q) among four answer choices (Q->A) and then select the correct rationale (R) among four rationale choices based on the answer. A higher score is better, and human performance on this task is measured by a QA->R score of 85. The best-performing machine has improved on the Q->AR score from 44 in 2018 to 70.5 in 2020 (Figure 2.4.2), which represents a 60.2% increase in performance from the top competitor in 2019.

VISUAL COMMONSENSE REASONING (VCR) TASK: Q->AR Score

Source: VCR Leaderboard, 2020 | Chart: 2021 AI Index Report

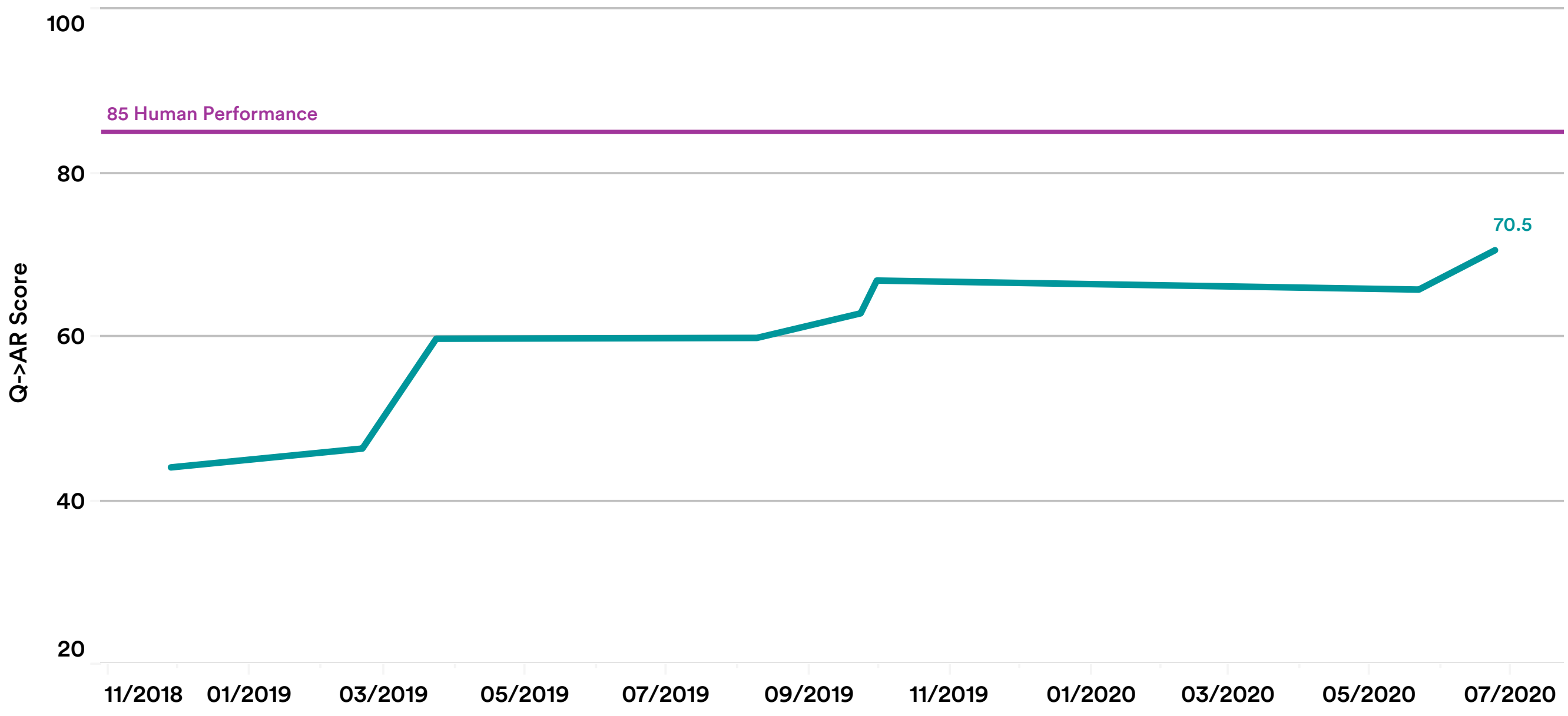

Figure 2.4.2

A major aspect of AI research is the analysis and synthesis of human speech conveyed via audio data. In recent years, machine learning approaches have drastically improved performance across a range of tasks.

# 2.5 SPEECH

## SPEECH RECOGNITION

Speech recognition, or automatic speech recognition (ASR), is the process that enables machines to recognize spoken words and convert them to text. Since IBM introduced its first speech recognition technology in 1962, the technology has evolved with voice-driven applications such as Amazon Alexa, Google Home, and Apple Siri becoming increasingly prevalent. The flexibility and predictive power of deep neural networks, in particular, has allowed speech recognition to become more accessible.

### Transcribe Speech: LibriSpeech

LibriSpeech is a dataset, first introduced in 2015, made up of 1,000 hours of speech from audiobooks. It has become widely used for the development and testing of speech recognition technologies. In recent years, neural-network-based AI systems have started to dramatically improve performance on LibriSpeech, lowering the word error rate (WER; 0% is optimal performance) to around 2% (Figure 2.5.1a and Figure 2.5.1b).

Developers can test out their systems on LibriSpeech in two ways:

- Test Clean determines how well their systems can transcribe speech from a higher-quality subset of the LibriSpeech dataset. This test gives clues about how well AI systems might perform in more controlled environments.
- Test Other determines how systems can deal with lower-quality parts of the LibriSpeech dataset. This test suggests how well AI systems might perform in noisier (and perhaps more realistic) environments.

There has been substantial progress recently on both datasets, with an important trend emerging in the past two years: The gap between performance on Test Clean and Test Other has started to close significantly for frontier systems, shifting from an absolute performance difference of more than seven points in late 2015 to a difference of less than one point in 2020. This reveals dramatic improvements in the robustness of ASR systems over time and suggests that we might be saturating performance on LibriSpeech—in other words, harder tests may be needed.

### Speaker Recognition: VoxCeleb

Speaker identification tests how well machine learning systems can attribute speech to a particular person. The VoxCeleb dataset, first introduced in 2017, contains over a million utterances from 6,000 distinct speakers, and its associated speaker-identification task tests the error rate for systems that try to attribute a particular utterance to a particular speaker. A better (lower) score in VoxCeleb provides a proxy for how well a machine can distinguish one voice among 6,000. Evaluation method for VoxCeleb is Equal Error Rate (EER), a commonly used metric for identity verification systems. EER provides a measure for both the false positive rate (assigning a label incorrectly) and the false negative rate (failing to assign a correct label).

In recent years, progress on this task has come from hybrid systems—systems that fuse contemporary deep learning approaches with more structured algorithms, developed by the broader speech-processing community. As of 2020, error rates have dropped such that computers have a very high (99.4%) ability to attribute utterances to a given speaker (Figure 2.5.2)

Still, obstacles remain: These systems face challenges processing speakers with different accents and in differentiating among speakers when confronted with a large dataset (it is harder to identify one person in a set of a billion people than to pick out one person across the VoxCeleb training set of 6,000).

### LIBRISPEECH: WORD ERROR RATE, TEST CLEAN

Source: Papers with Code, 2020 | Chart: 2021 AI Index Report

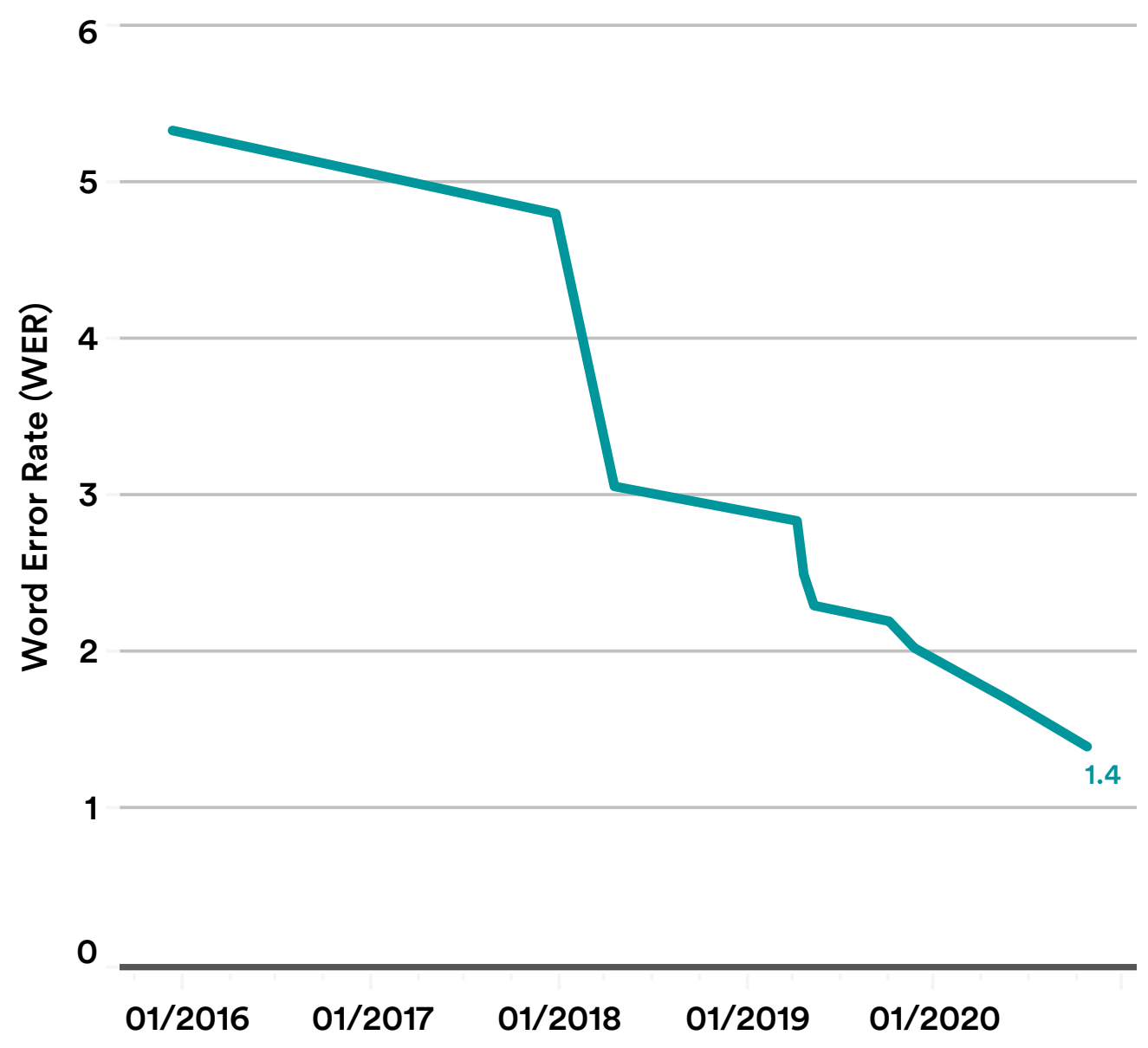

Figure 2.5.1a

### LIBRISPEECH: WORD ERROR RATE, TEST OTHER

Source: Papers with Code, 2020 | Chart: 2021 AI Index Report

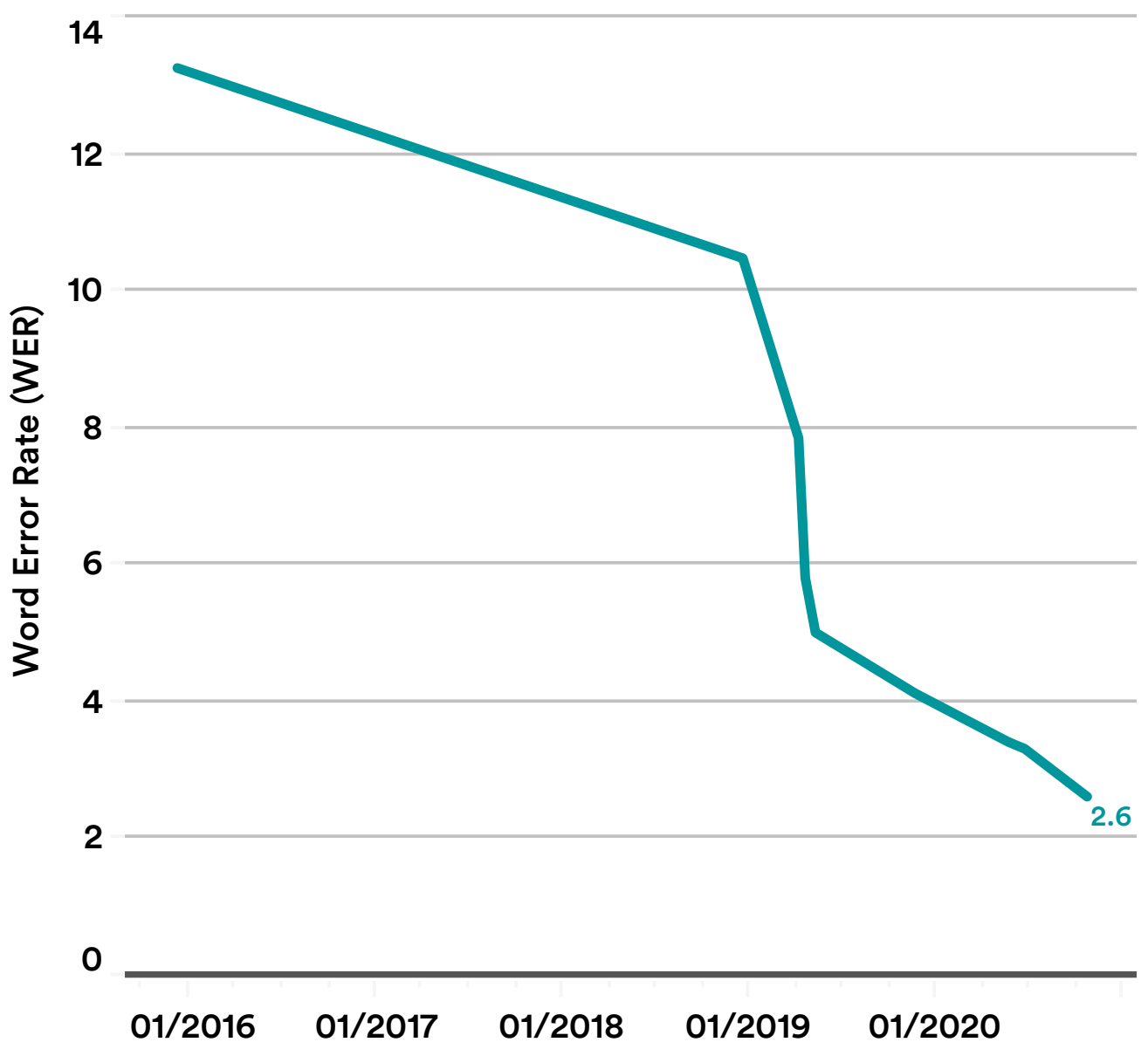

Figure 2.5.1b

### VOXCELEB: EQUAL ERROR RATE

Source: VoxCeleb, 2020 | Chart: 2021 AI Index Report

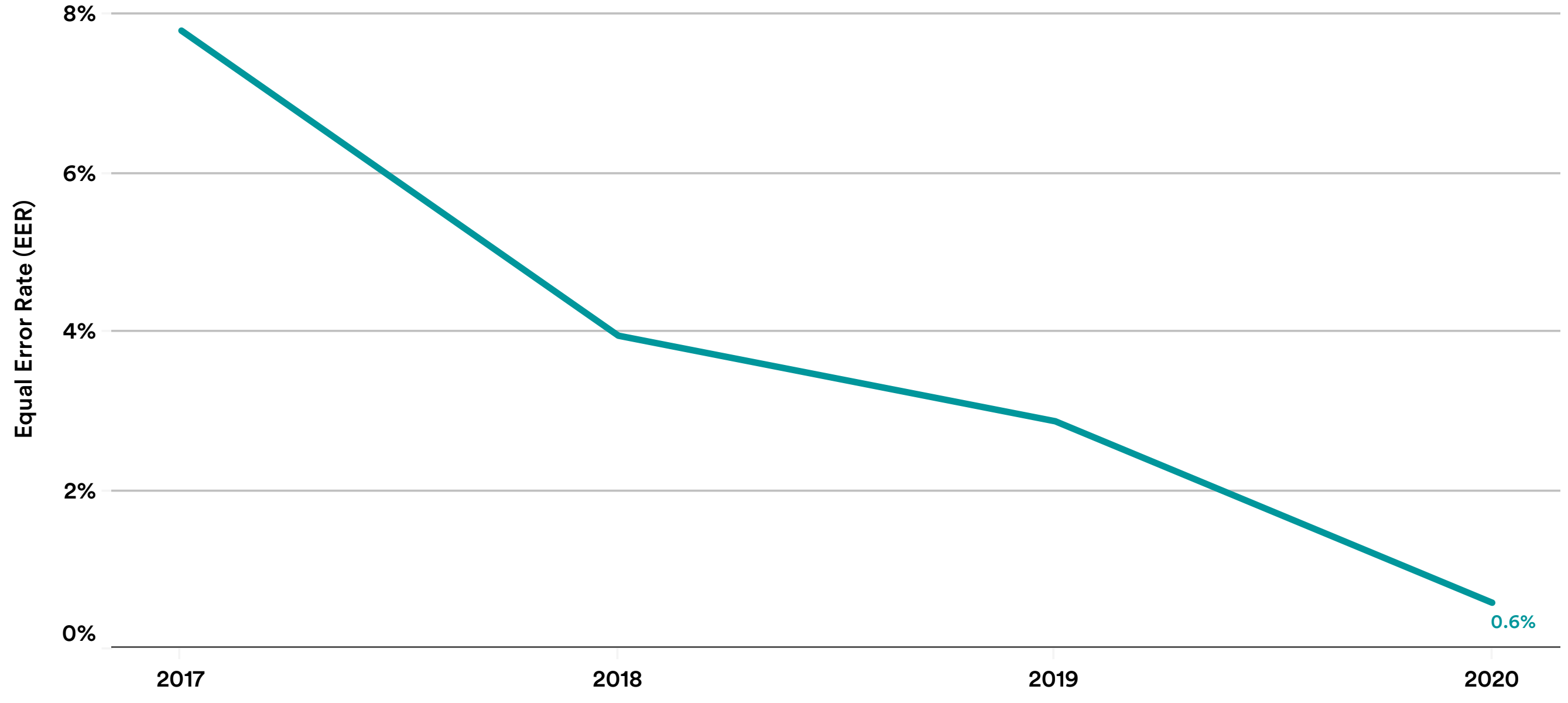

Figure 2.5.2

# The Race Gap in Speech Recognition Technology

Researchers from Stanford University found that state-of-the-art ASR systems exhibited significant racial and gender disparity—they misunderstand Black speakers twice as often as white speakers. In the paper, titled "Racial Disparities in Automated Speech Recognition," authors ran thousands of audio snippets of white and Black speakers, transcribed from interviews conducted with 42 white speakers and 73 Black speakers, through leading speech-to-text services by Amazon, Apple, Google, IBM, and Microsoft.

The results suggest that, on average, systems made 19 errors every hundred words for white speakers and 35 errors for Black speakers—nearly twice as many. Moreover, the systems performed particularly poorly for Black men, with more than 40 errors for every hundred words (Figure 2.5.3). The breakdown by ASR systems shows that gaps are similar across companies (Figure 2.5.4). This research emphasizes the importance of addressing the bias of AI technologies and ensuring equity as they become mature and deployed.

TESTINGS on LEADING SPEECH-to-TEXT SERVICES: WORD ERROR RATE by RACE and GENDER, 2019
Source: Koenecke et al., 2020 | Chart: 2021 AI Index Report

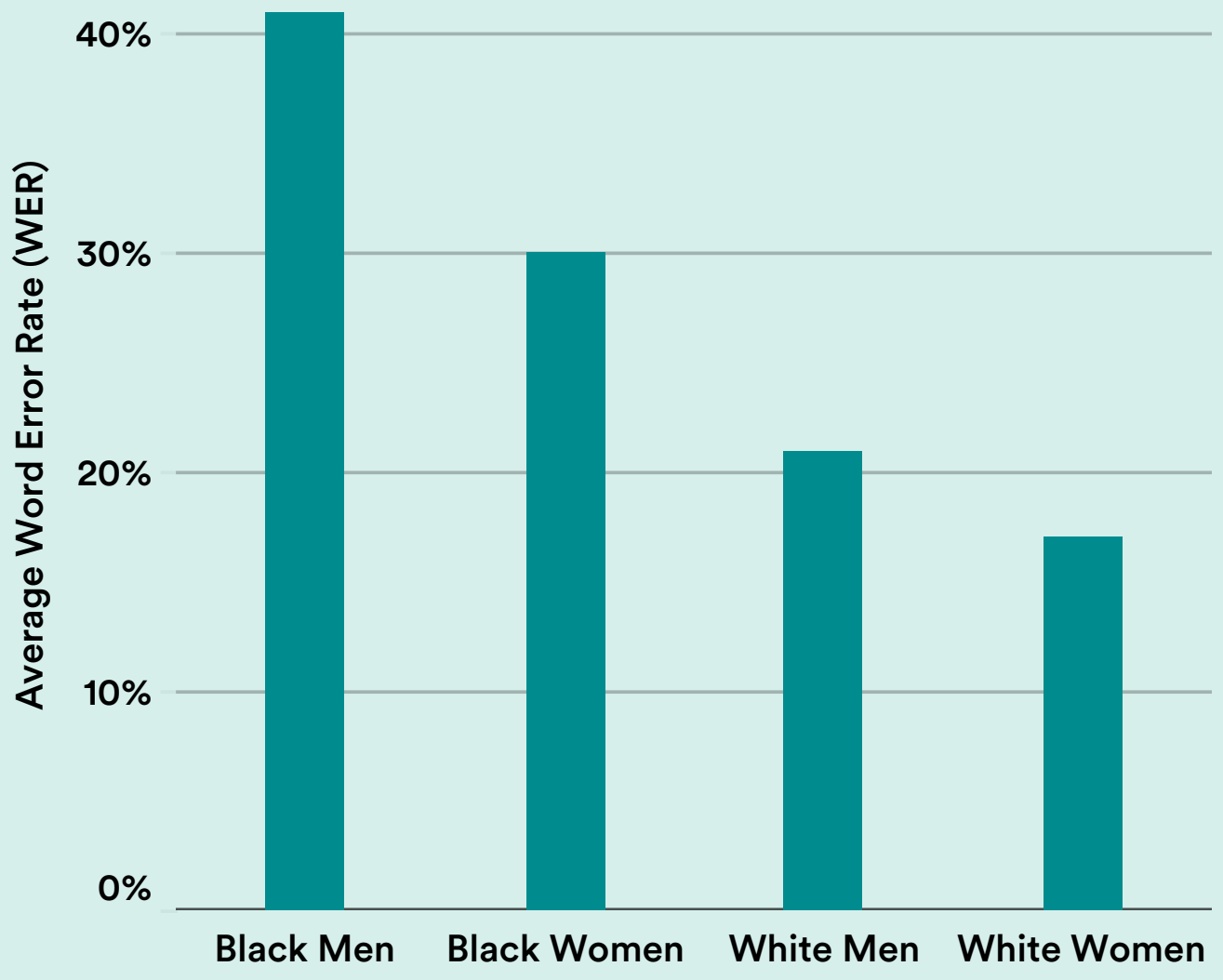

Figure 2.5.3

TESTINGS on LEADING SPEECH-to-TEXT SERVICES: WORD ERROR RATE by SERVICE and RACE, 2019
Source: Koenecke et al., 2020 | Chart: 2021 AI Index Report

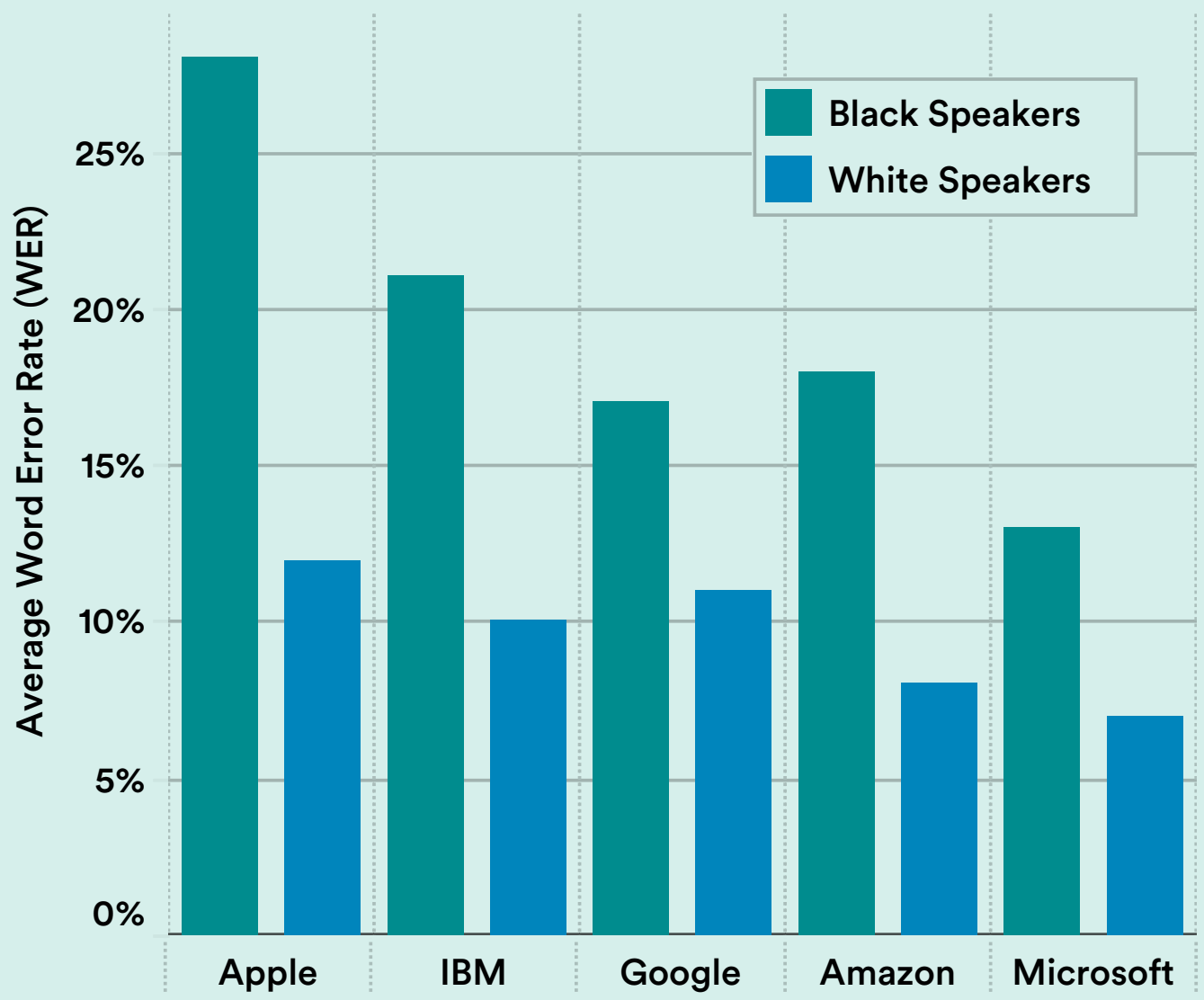

Figure 2.5.4

This section measures progress on symbolic (or logical) reasoning in AI, which is the process of drawing conclusions from sets of assumptions. We consider two major reasoning problems, Boolean Satisfiability (SAT) and Automated Theorem Proving (ATP). Each has real-world applications (e.g., circuit design, scheduling, software verification, etc.) and poses significant measurement challenges. The SAT analysis shows how to assign credit for the overall improvement in the field to individual systems over time. The ATP analysis shows how to measure performance given an evolving test set.

All analyses below are original to this report. Lars Kotthoff wrote the text and performed the analysis for the SAT section. Geoff Sutcliffe, Christian Suttner, and Raymond Perrault wrote the text and performed the analysis for the ATP section. This work had not been published at the time of writing; consequently, a more academically rigorous version of this section (with references, more precise details, and further context) is included in the Appendix.

# 2.6 REASONING

## BOOLEAN SATISFIABILITY PROBLEM

Analysis and text by Lars Kotthoff

The SAT problem considers whether there is an assignment of values to a set of Boolean variables, joined by logical connectives, that makes the logical formula it represents true. Many real-world problems, such as circuit design, automated theorem proving, and scheduling, can be represented and solved efficiently as SAT problems.

The performance of the top-, median-, and bottom-ranked SAT solvers was examined from each of the last five years (2016–2020) of the SAT Competition, which has been running for almost 20 years, to measure a snapshot of state-of-the-art performance. In particular, all 15 solvers were run on all 400 SAT instances from the main track of the 2020 competition and the time (in CPU seconds) it took to solve all instances was measured.[5] Critically, each solver was run on the same hardware, such that comparisons across years would not be confounded by improvements in hardware efficiency over time.

While performance of the best solvers from 2016 to 2018 did not change significantly, large improvements are evident in 2019 and 2020 (Figure 2.6.1). These improvements affect not only the best solvers but also their competitors. The performance of the median-ranked solver in 2019 is better than that of the top-ranked solvers in all previous years, and the performance of the median-ranked solver in 2020 is almost on par with the top-ranked solver in 2019.

Performance improvements in SAT—and more generally, hard computational AI problems—come primarily from two areas of algorithmic improvements: novel techniques and more efficient implementations of existing techniques. Typically, performance improvements arise primarily from novel techniques. However, more efficient implementations (which can arise with performance improvements in hardware over time) can also increase performance. Therefore, it is difficult to assess whether performance improvements arise primarily from novel techniques or more efficient implementations. To address this problem, the temporal Shapley value, which is the contribution of an individual system to state-of-the-art performance over time, was measured (see the Appendix for more details).

Figure 2.6.2 shows the temporal Shapley value contributions of each solver for the different competition years. Note that the contributions of the solvers in 2016 are highest because there is no previous state-of-the-art to compare them with in our evaluation and that their contribution is not discounted.

5 Acknowledgments: The Advanced Research Computing Center at the University of Wyoming provided resources for gathering the computational data. Austin Stephen performed the computational experiments.

TOTAL TIME to SOLVE ALL 400 INSTANCES for EACH SOLVER and YEAR (LOWER IS BETTER), 2016-20
Source: Kotthoff, 2020 | Chart: 2021 AI Index Report

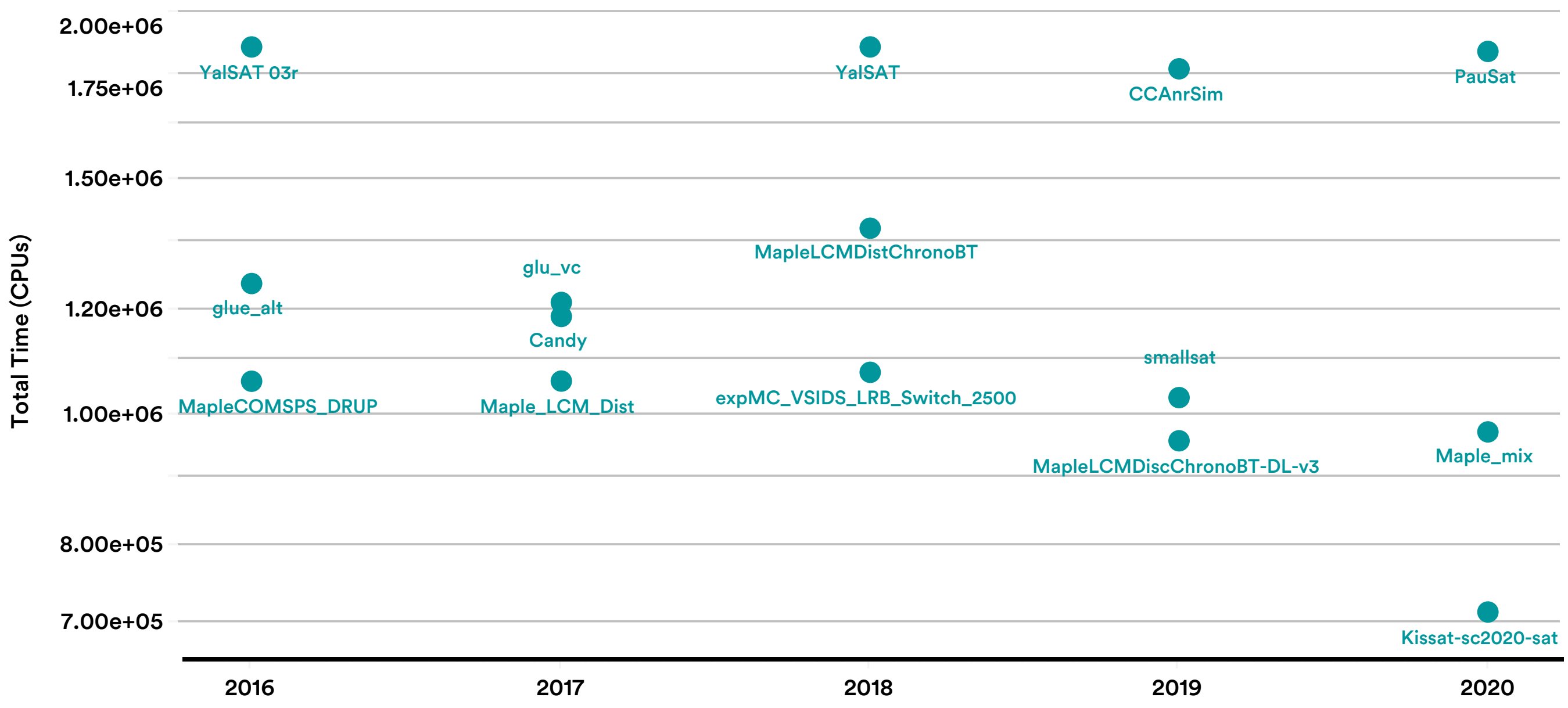

Figure 2.6.1

TEMPORAL SHAPLEY VALUE CONTRIBUTIONS of INDIVIDUAL SOLVERS to the STATE of the ART OVER TIME (HIGHER IS BETTER), 2016-20
Source: Kotthoff, 2020 | Chart: 2021 AI Index Report

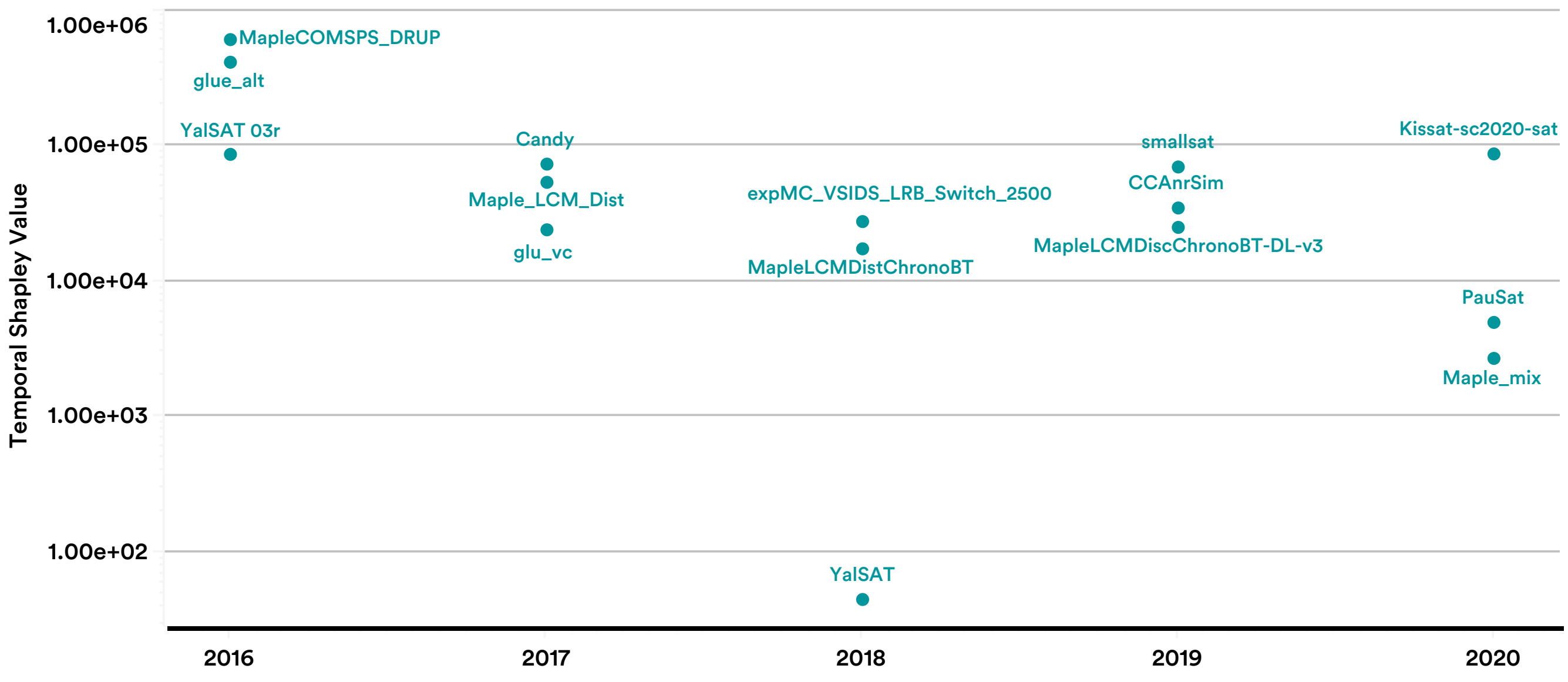

Figure 2.6.2

According to the temporal Shapley value, in 2020 the best solver contributes significantly more than the median- and bottom-ranked solvers do. The 2020 winner, Kissat, has the highest temporal Shapley value of any solvers excluding the first year. The changes it incorporates, compared with those of previous solvers, are almost exclusively more efficient data structures and algorithms; Kissat thus impressively demonstrates the impact of good engineering on the state-of-the-art performance.

By contrast, smallsat, the solver with the largest temporal Shapley value (but not the winner) in 2019, focuses on improved heuristics instead of a more efficient implementation. The same is true of Candy, the solver with the largest temporal Shapley value in 2017, whose main novelty is to analyze the structure of a SAT instance and apply heuristics based on this analysis. Interestingly, neither solver ranked first in their respective years; both were outperformed by versions of the Maple solver, which nevertheless contributes less to the state of the art. This indicates that incremental improvements, while not necessarily exciting, are important for good performance in practice.

Based on our limited analysis of the field, novel techniques and more efficient implementations have made equally important contributions to the state of the art in SAT solving. Incremental improvements of established solvers are as likely to result in top performance as more substantial improvements of solvers without a long track record.

## AUTOMATED THEOREM PROVING (ATP)

Analysis and text by Christian Suttner, Geoff Sutcliffe, and Raymond Perrault

Automated Theorem Proving (ATP) concerns the development and use of systems that automate sound reasoning, or the derivation of conclusions that follow inevitably from facts. ATP systems are at the heart of many computational tasks, including software verification. The TPTP problem library was used to evaluate the performance of ATP algorithms from 1997 to 2020 and to measure the fraction of problems solved by any system over time (see the Appendix for more details).

The analysis extends to the whole TPTP (over 23,000 problems) in addition to four salient subsets (each ranging between 500 and 5,500 problems)—clause normal form (CNF), first-order form (FOF), monomorphic typed first-order form (TF0) with arithmetic, and monomorphic typed higher-order form (TH0) theorems—all including the use of the equality operator.

Figure 2.6.3 shows that the fraction of problems solved climbs consistently, indicating progress in the field. The noticeable progress from 2008 to 2013 included strong progress in the FOF, TF0, and TH0 subsets. In FOF, which has been used in many domains (e.g., mathematics, real-world knowledge, software verification), there were significant improvements in the Vampire, E, and iProver systems. In TF0 (primarily used for solving problems in mathematics and computer science) and TH0 (useful in subtle and complex topics such as philosophy and logic), there was rapid initial progress as systems developed techniques that solved “low-hanging fruit” problems. In 2014–2015, there was another burst of progress in TF0, as the Vampire system became capable of processing TF0 problems. It is noteworthy that, since 2015, progress has continued but slowed, with no indication of rapid advances or breakthroughs in the last few years.

PERCENTAGE of PROBLEMS SOLVED, 1997-2020
Source: Sutcliffe, Suttner & Perrault, 2020 | Chart: 2021 AI Index Report

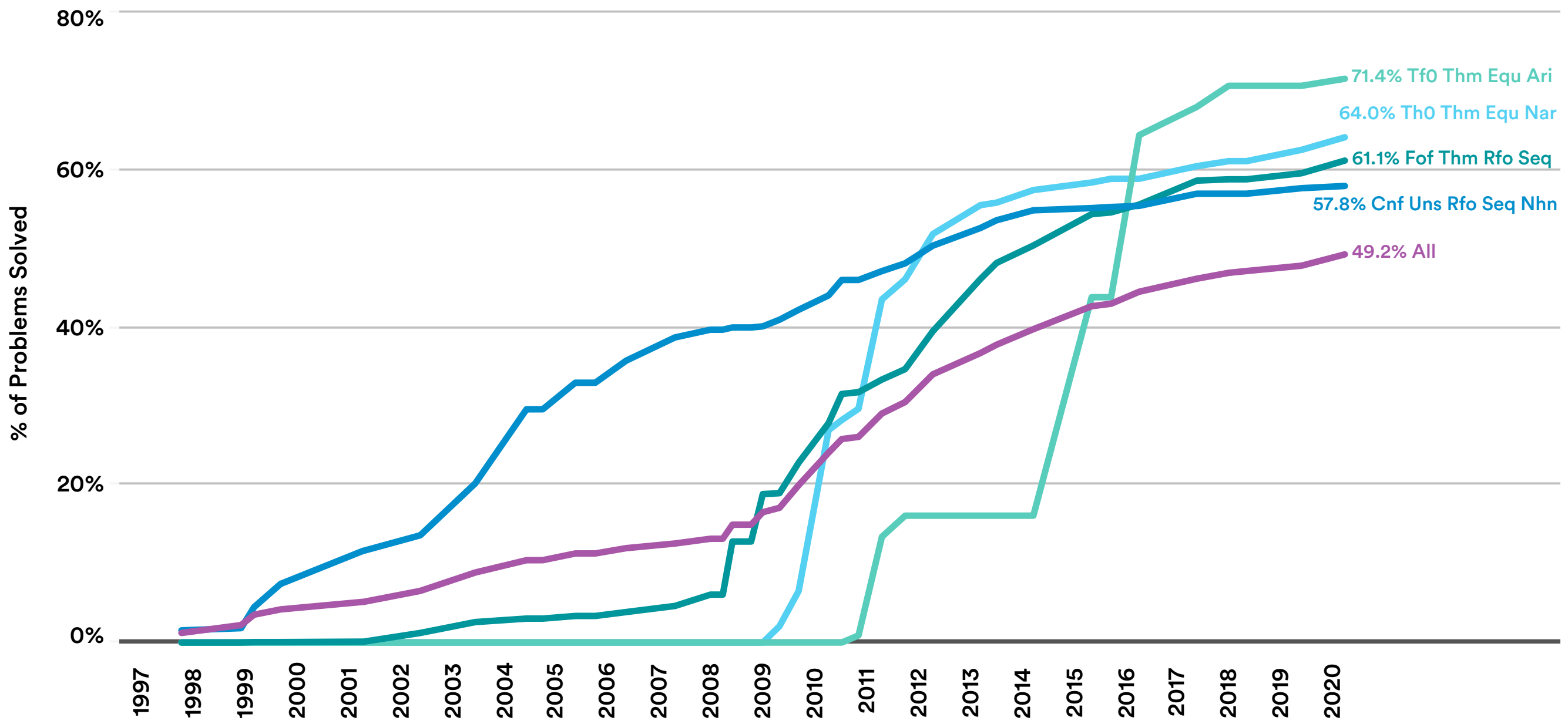

Figure 2.6.3

While this analysis demonstrates progress in ATP, there is obviously room for much more. Two keys to solve ATP problems are axiom selection (given a large set of axioms, only some of which are needed for a proof of the conjecture, how to select an adequate subset of the axioms); and search choice (at each stage of an ATP system's search for a solution, which logical formula(e) should be selected for attention). The latter issue has been at the forefront of ATP research since its inception in the 1960s, while the former has become increasingly important as large bodies of knowledge are encoded for ATP. In the last decade, there has been growing use of machine learning approaches to addressing these two key challenges (e.g., in the MaLARea and Enigma ATP systems). Recent results from the CADE ATP System Competition (CASC) have shown that the emergence of machine learning is a potential game-changer for ATP.

# 2.7 HEALTHCARE AND BIOLOGY

In collaboration with the "State of AI Report"

## MOLECULAR SYNTHESIS

Text by Nathan Benaich and Philippe Schwaller

Over the last 25 years, the pharmaceutical industry has shifted from developing drugs from natural sources (e.g., plants) to conducting large-scale screens with chemically synthesized molecules. Machine learning allows scientists to determine what potential drugs are worth evaluating in the lab and the most effective way of synthesizing them. Various ML models can learn representations of chemical molecules for the purposes of chemical synthesis planning.

A way to approach chemical synthesis planning is to represent chemical reactions with a text notation and cast the task as a machine translation problem. Recent work since 2018 makes use of the transformer architecture trained on large datasets of single-step reactions. Later work in 2020 approached model forward prediction and retrosynthesis as a sequence of graph edits, where the predicted molecules were built from scratch.

Notably, these approaches offer an avenue to rapidly sweep through a list of candidate drug-like molecules in silico and output synthesizability scores and synthesis plans. This enables medicinal chemists to prioritize candidates for empirical validation and could ultimately let the pharmaceutical industry mine the vast chemical space to unearth novel drugs to benefit patients.

### Test Set Accuracy for Forward Chemical Synthesis Planning

Figure 2.7.1 shows the top-1 accuracy of models benchmarked on a freely available dataset of one million reactions in the U.S. patents.[6] Top-1 accuracy means that the product predicted by the model with the highest likelihood corresponds to the one that was reported in the ground truth. Data suggests that progress in chemical synthesis planning has seen steady growth in the last three years, as the accuracy grew by 15.6% in 2020 from 2017. The latest molecular transformer scored 92% on top-1 accuracy in November 2020.

CHEMICAL SYNTHESIS PLANS BENCHMARK: TOP-1 TEST ACCURACY
Source: Schwaller, 2020 | Chart: 2021 AI Index Report

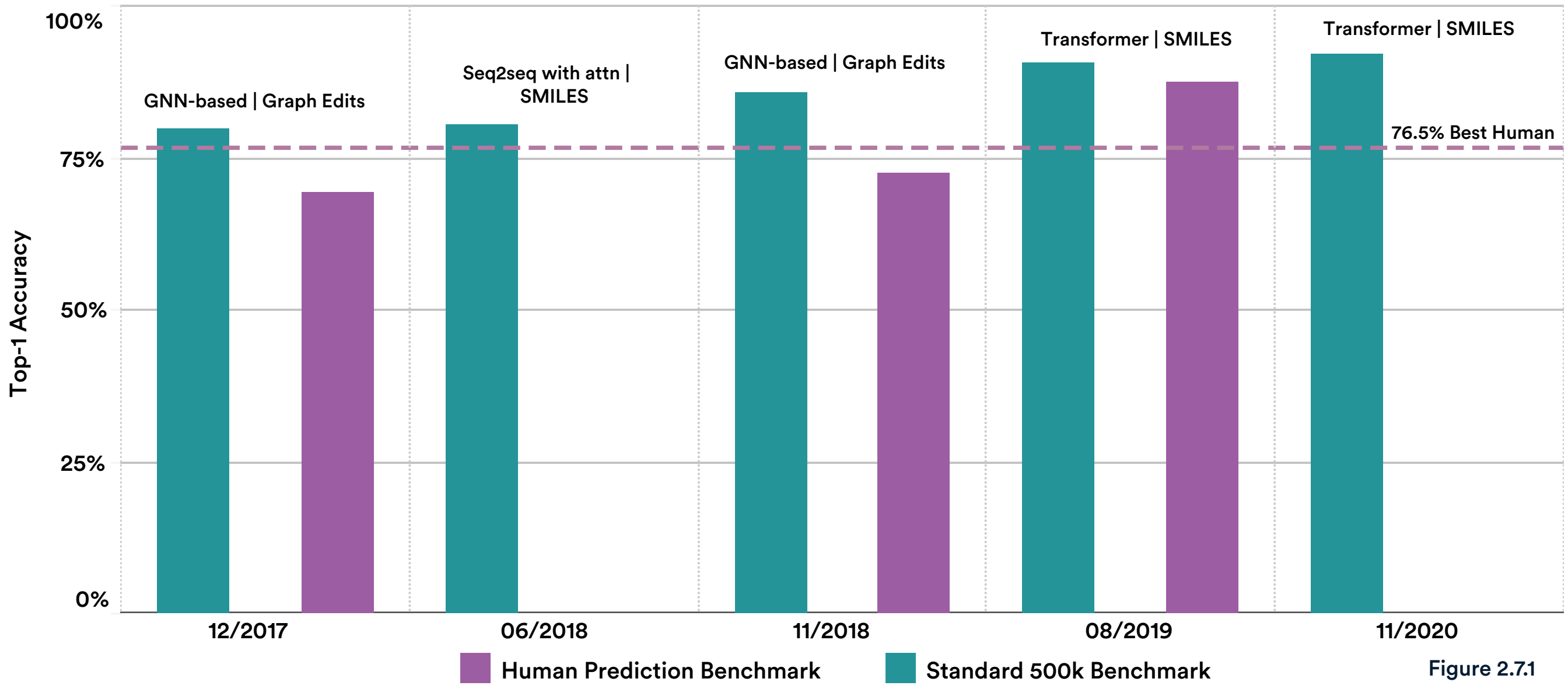

Figure 2.7.1

6 Acknowledgment: Philippe Schwaller at IBM Research–Europe and the University of Bern provided instructions and resources for gathering and analyzing the data.

## COVID-19 AND DRUG DISCOVERY

AI-powered drug discovery has gone open source to combat the COVID-19 pandemic. COVID Moonshot is a crowdsourced initiative joined by over 500 international scientists to accelerate the development of a COVID-19 antiviral. The consortium of scientists submits their molecular designs pro bono, with no claims. PostEra, an AI startup, uses machine learning and computational tools to assess how easily compounds can be made using the submissions from the scientists and generates synthetic routes. After the first week, Moonshot received over 2,000 submissions, and PostEra designed synthetic routes in under 48 hours. Human chemists would have taken three to four weeks to accomplish the same task.

Figure 2.7.2 shows the accumulated number of submissions by scientists over time. Moonshot received over 10,000 submissions from 365 contributors around the world in just four months. Toward the end of August 2020, the crowdsourcing had served its purpose, and the emphasis moved to optimize the lead compounds and set up for animal testing. As of February 2021, Moonshot aims to nominate a clinical candidate by the end of March.

POSTERA: TOTAL NUMBER of MOONSHOT SUBMISSIONS

Source: PostEra, 2020 | Chart: 2021 AI Index Report

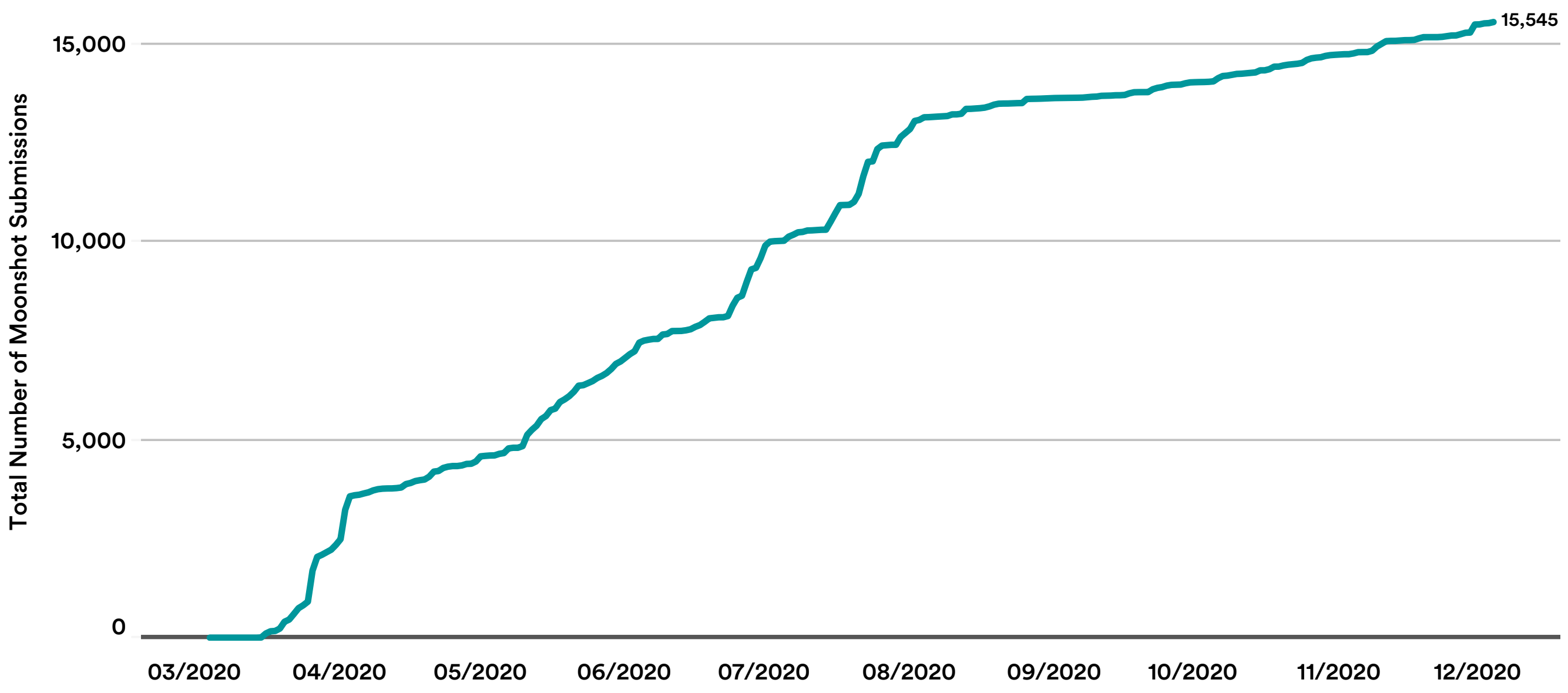

Figure 2.7.2

## ALPHAFOLD AND PROTEIN FOLDING

The protein folding problem, a grand challenge in structural biology, considers how to determine the three-dimensional structure of proteins (essential components of life) from their one-dimensional representations (sequences of amino acids[7]). A solution to this problem can have wide ranging applications—from better understanding the cellular basis of life, to fueling drug discovery, to curing diseases, to engineering de-novo proteins for industrial tasks, and more.

In recent years, machine learning-based approaches have started to make a meaningful difference on the protein folding problem. Most notably, DeepMind's AlphaFold debuted in 2018 at the Critical Assessment of Protein Structure (CASP) competition, a biennial competition to foster and measure progress on protein folding. At CASP, competing teams are given amino acid sequences and tasked to predict the three-dimensional structures of the corresponding proteins, the latter of which are determined through laborious and expensive experimental methods (e.g., nuclear magnetic resonance spectroscopy, X-ray crystallography, cryo-electron microscopy, etc.) and unknown to the competitors. Performance on CASP is commonly measured by the Global Distance Test (GDT) score, a number between 0 and 100, which measures the similarity between two protein structures. A higher GDT score is better.

Figure 2.7.3, adapted from the DeepMind blog post, shows the median GDT scores of the best team on some of the harder types of proteins to predict (the 'free-modelling' category of proteins) at CASP over the last 14 years. In the past, winning algorithms were typically based on physics based models; however, in the last two competitions, Deepmind's AlphaFold and AlphaFold 2 algorithms achieved winning scores through the partial incorporation of deep learning techniques.

CASP: MEDIAN ACCURACY of PREDICTIONS in FREE-MODELING by THE BEST TEAM, 2006-20
Source: DeepMind, 2020 | Chart: 2021 AI Index Report

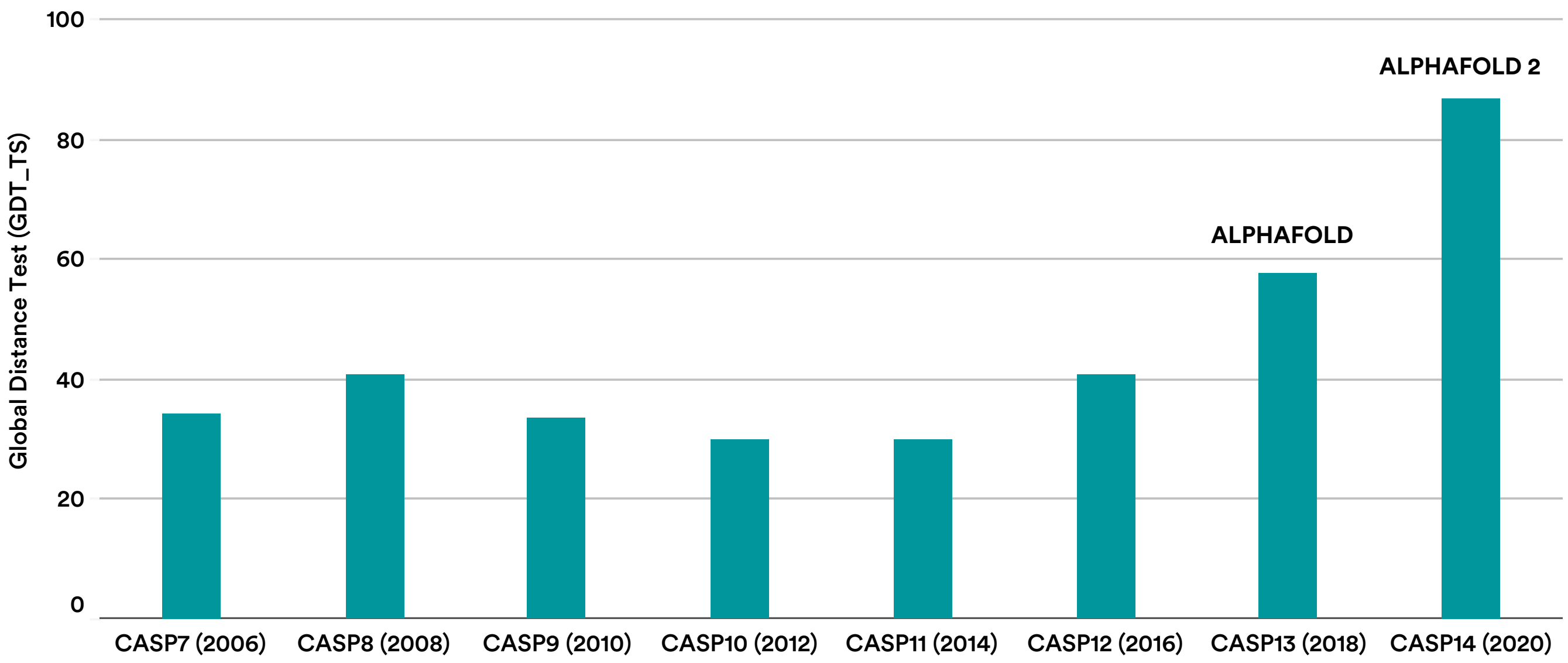

Figure 2.7.3

7 Currently most protein folding algorithms leverage multiple sequence alignments—many copies of a protein sequence representing the same protein across evolution—rather than just a single sequence.

# EXPERT HIGHLIGHTS

This year, the AI Index asked AI experts to share their thoughts on the most significant technical AI breakthroughs in 2020. Here's a summary of their responses, along with a couple of individual highlights.

### What was the single most impressive AI advancement in 2020?

- The two most mentioned systems by a significant margin were AlphaFold (DeepMind), a model for molecular assay, and GPT-3 (OpenAI), a generative text model.

### What single trend will define AI in 2021?

- Experts predict that more advances will be built by using pretrained models. For instance, GPT-3 is a large NLP model that can subsequently be fine-tuned for excellent performance on specific, narrow tasks. Similarly, 2020 saw various computer vision advancements built on top of models pretrained on very large image datasets.

### What aspect of AI technical progress, deployment, and development are you most excited to see in 2021?

- "It's interesting to note the dominance of the Transformers architecture, which started for machine translation but has become the de facto neural network architecture. More broadly, whereas NLP trailed vision in terms of adoption of deep learning, now it seems like advances in NLP are also driving vision." — Percy Liang, Stanford University

- "The incredible recent advancements in language generation have had a profound effect on the fields of NLP and machine learning, rendering formerly difficult research challenges and datasets suddenly useless while simultaneously encouraging new research efforts into the fascinating emergent capabilities (and important failings) of these complex new models." —Carissa Schoenick, Allen Institute of AI Research

# CHAPTER 3:
# The Economy

Artificial Intelligence
Index Report 2021

## CHAPTER 3:

# Chapter Preview

ACCESS THE PUBLIC DATA

# Overview

The rise of artificial intelligence (AI) inevitably raises the question of how much the technologies will impact businesses, labor, and the economy more generally. Considering the recent progress and numerous breakthroughs in AI, the field offers substantial benefits and opportunities for businesses, from increasing productivity gains with automation to tailoring products to consumers using algorithms, analyzing data at scale, and more.

However, the boost in efficiency and productivity promised by AI also presents great challenges: Companies must scramble to find and retain skilled talent to meet their production needs while being mindful about implementing measures to mitigate the risks of using AI. Moreover, the COVID-19 pandemic has caused chaos and continued uncertainty for the global economy. How have private companies relied on and scaled AI technologies to help their business navigate through this most difficult time?

This chapter looks at the increasingly intertwined relationship between AI and the global economy from the perspective of jobs, investment, and corporate activity. It first analyzes the worldwide demand for AI talent using data on hiring rates and skill penetration rates from LinkedIn as well as AI job postings from Burning Glass Technologies. It then looks at trends in private AI investment using statistics from S&P Capital IQ (CapIQ), Crunchbase, and Quid. The third, final section analyzes trends in the adoption of AI capabilities across companies, trends in robot installations across countries, and mentions of AI in corporate earnings, drawing from McKinsey's Global Survey on AI, the International Federation of Robotics (IFR), and Prattle, respectively.

# CHAPTER HIGHLIGHTS

- "Drugs, Cancer, Molecular, Drug Discovery" received the greatest amount of private AI investment in 2020, with more than USD 13.8 billion, 4.5 times higher than 2019.

- Brazil, India, Canada, Singapore, and South Africa are the countries with the highest growth in AI hiring from 2016 to 2020. Despite the COVID-19 pandemic, the AI hiring continued to grow across sample countries in 2020.

- More private investment in AI is being funneled into fewer startups. Despite the pandemic, 2020 saw a 9.3% increase in the amount of private AI investment from 2019—a higher percentage increase than in 2019 (5.7%), though the number of newly funded companies decreased for the third year in a row.

- Despite growing calls to address ethical concerns associated with using AI, efforts to address these concerns in the industry are limited, according to a McKinsey survey. For example, issues such as equity and fairness in AI continue to receive comparatively little attention from companies. Moreover, fewer companies in 2020 view personal or individual privacy risks as relevant, compared with in 2019, and there was no change in the percentage of respondents whose companies are taking steps to mitigate these particular risks.

- Despite the economic downturn caused by the pandemic, half the respondents in a McKinsey survey said that the coronavirus had no effect on their investment in AI, while 27% actually reported increasing their investment. Less than a fourth of businesses decreased their investment in AI.

- The United States recorded a decrease in its share of AI job postings from 2019 to 2020—the first drop in six years. The total number of AI jobs posted in the United States also decreased by 8.2%, from 325,724 in 2019 to 300,999 in 2020.

Attracting and retaining skilled AI talent is challenging. This section examines the latest trend in AI hiring, labor demand, and skill penetration, with data from LinkedIn and Burning Glass.

# 3.1 JOBS

## AI HIRING

How rapidly is the growth of AI jobs in different countries? This section first looks at LinkedIn data that gives the AI hiring rate for different countries. The AI hiring rate is calculated as the number of LinkedIn members who include AI skills on their profile or work in AI-related occupations and who added a new employer in the same month their new job began, divided by the total number of LinkedIn members in the country. This rate is then indexed to the average month in 2016; for example, an index of 1.05 in December 2020 points to a hiring rate that is 5% higher than the average month in 2016. LinkedIn makes month-to-month comparisons to account for any potential lags in members updating their profiles. The index for a year is the average index over all months within that year.

This data suggests that the hiring rate has been increasing across all sample countries in 2020. Brazil, India, Canada, Singapore, and South Africa are the countries with the highest growth in AI hiring from 2016 to 2020 (Figure 3.1.1). Across the 14 countries analyzed, the AI hiring rate in 2020 was 2.2 times higher, on average, than that in 2016. For the top country, Brazil, the hiring index grew by more than 3.5 times. Moreover, despite the COVID-19 pandemic, AI hiring continued its growth across the 14 sampled countries in 2020 (Figure 3.1.2).

For more explorations of cross-country comparisons, see the AI Index Global AI Vibrancy Tool.

AI HIRING INDEX by COUNTRY, 2020

Source: LinkedIn, 2020 | Chart: 2021 AI Index Report

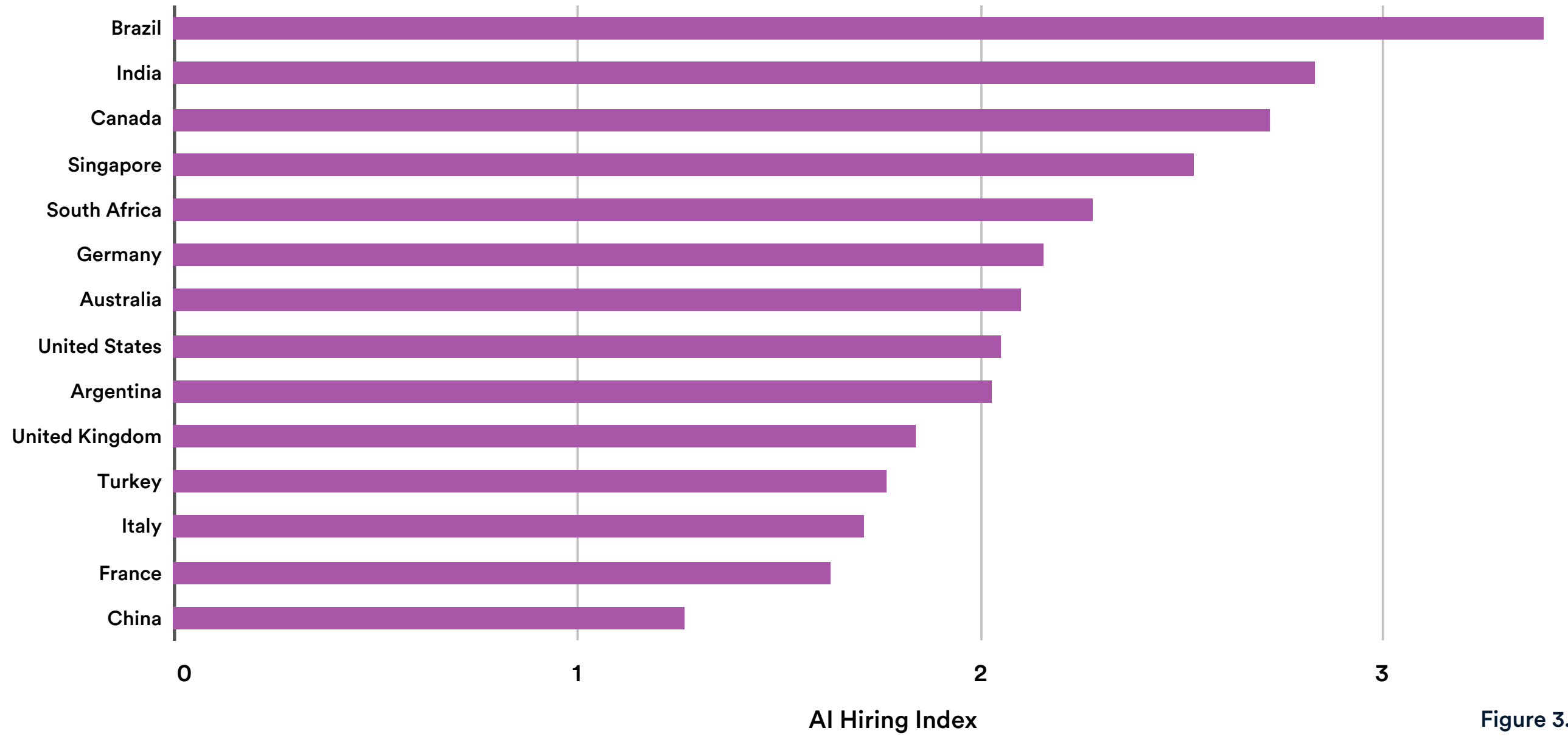

Figure 3.1.1

1 Countries included are a sample of eligible countries with at least 40% labor force coverage by LinkedIn and at least 10 AI hires in any given month. China and India were also included in this sample because of their increasing importance in the global economy, but LinkedIn coverage in these countries does not reach 40% of the workforce. Insights for these countries may not provide as full a picture as in other countries, and should be interpreted accordingly.

AI HIRING INDEX by COUNTRY, 2016-20

Source: LinkedIn, 2020 | Chart: 2021 AI Index Report

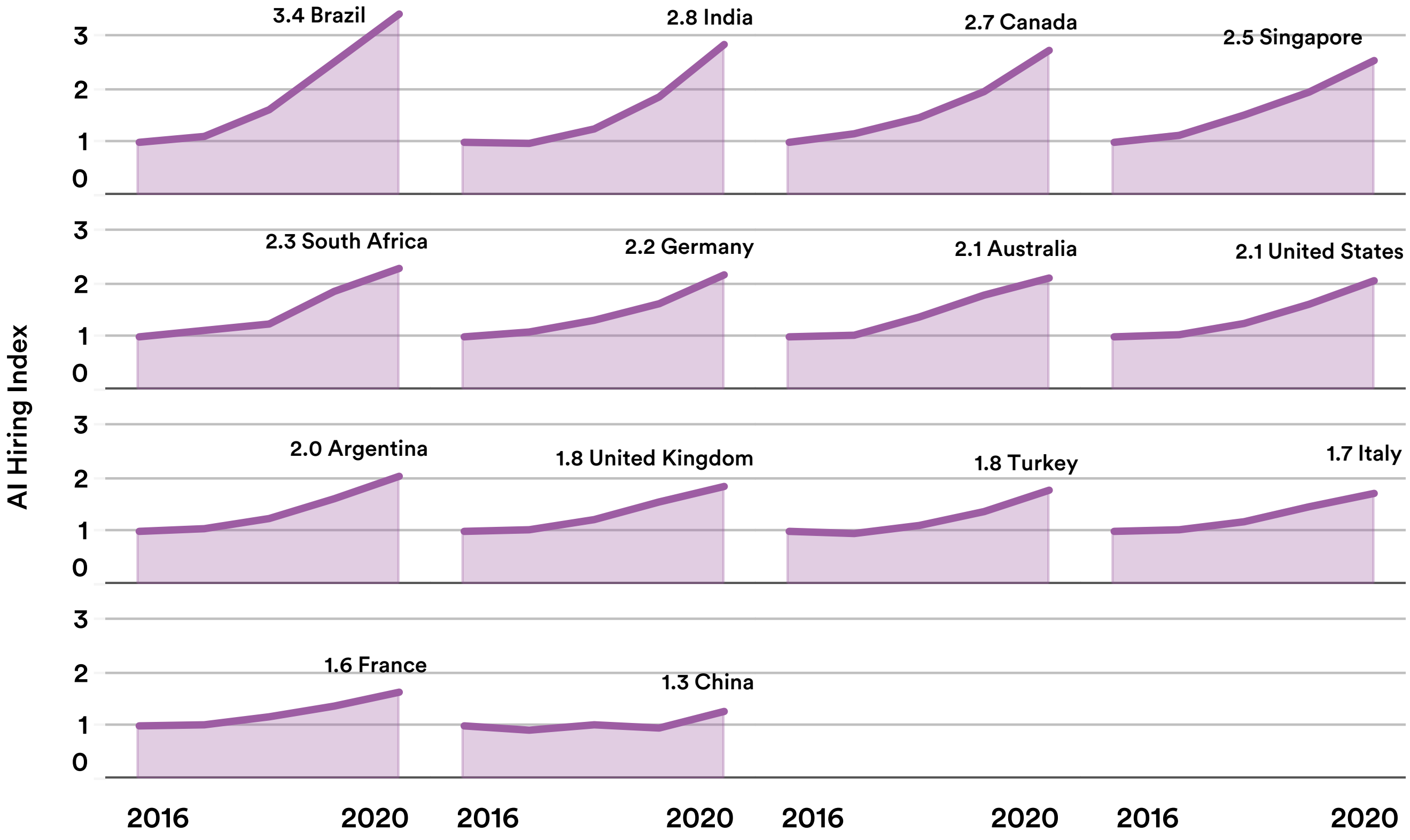

Figure 3.1.2

## AI LABOR DEMAND

This section analyzes the AI labor demand based on data from Burning Glass, an analytics firm that collects postings from over 45,000 online job sites. To develop a comprehensive, real-time portrait of labor market demand, Burning Glass aggregated job postings, removed duplicates, and extracted data from job posting text. Note that Burning Glass updated the data coverage in 2020 with more job sites; as a result, the numbers in this report should not be directly compared with data in the 2019 report.

### Global AI Labor Demand

Demand for AI labor in six countries covered by Burning Glass data—the United States, the United Kingdom, Canada, Australia, New Zealand, and Singapore—has grown significantly in the last seven years (Figure 3.1.3). On average, the share of AI job postings among all job postings in 2020 is more than five times larger than in 2013. Of the six countries, Singapore exhibits the largest growth, as its percentage of AI job postings across all job roles in 2020 is 13.5 times larger than in 2013.

The United States is the only country among the six that recorded a decrease in its share of AI job postings from 2019 to 2020—the first drop in six years. This may be due to the coronavirus pandemic or the country's relatively more mature AI labor market. The total number of AI jobs posted in the United States also decreased by 8.2%, from 325,724 in 2019 to 300,999 in 2020.

AI JOB POSTINGS (% of ALL JOB POSTINGS) by COUNTRY, 2013-20
Source: Burning Glass, 2020 | Chart: 2021 AI Index Report

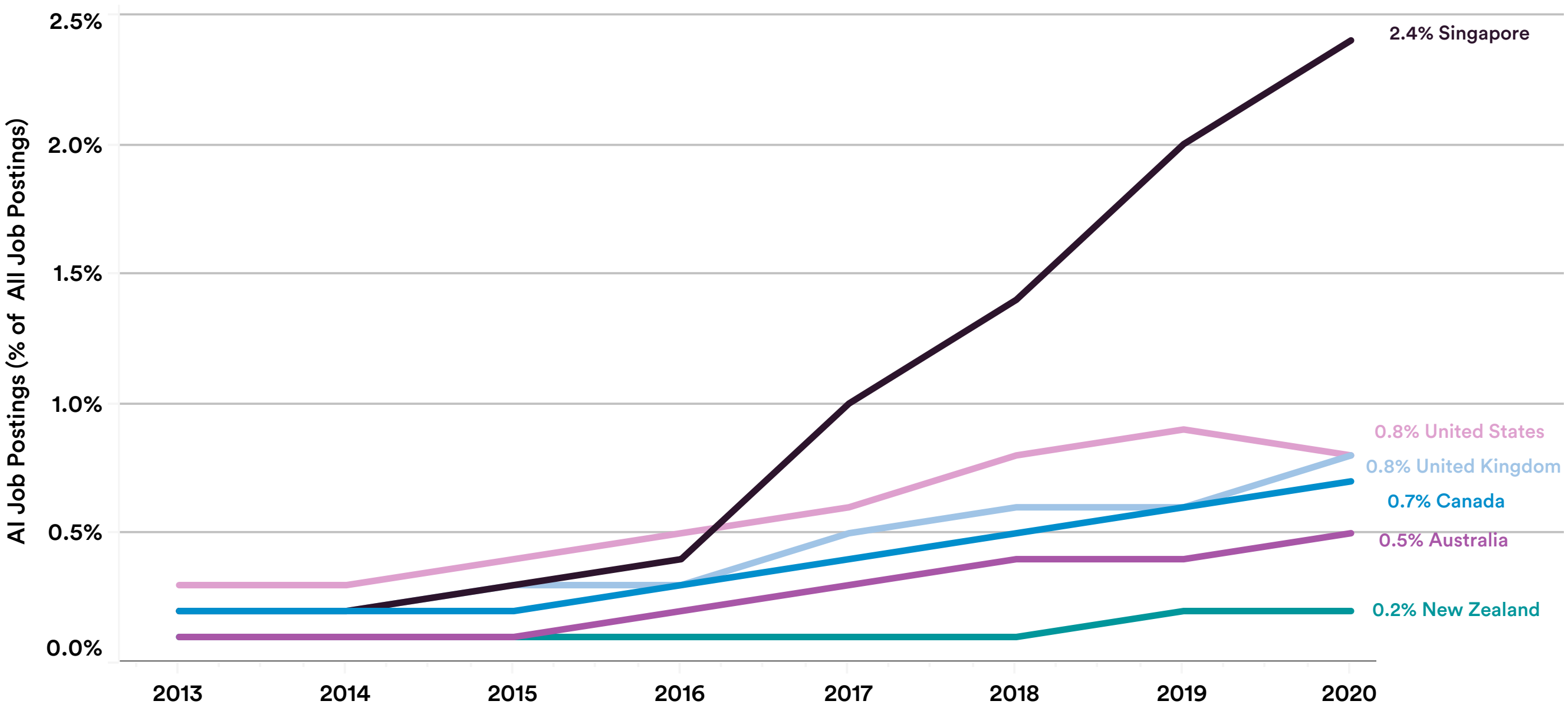

Figure 3.1.3

## U.S. AI Labor Demand: By Skill Cluster

Taking a closer look at the AI labor demand in the United States between 2013 and 2020, Figure 3.1.4 breaks down demand during that period year by year according to skill cluster. Each skill cluster consists of a list of AI-related skills; for example, the neural network skill cluster includes skills like deep learning and convolutional neural network. The Economy chapter appendix provides a complete list of AI skills under each skill cluster.

Between 2013 and 2020, AI jobs related to machine learning and artificial intelligence experienced the fastest growth in online AI job postings in the United States, increasing from 0.1% of total jobs to 0.5% and 0.03% to 0.3%, respectively. As noted earlier, 2020 shows a decrease in the share of AI jobs among overall job postings across all skill clusters.

AI JOB POSTINGS (% of ALL JOB POSTINGS) in the UNITED STATES by SKILL CLUSTER, 2013-20
Source: Burning Glass, 2020 | Chart: 2021 AI Index Report

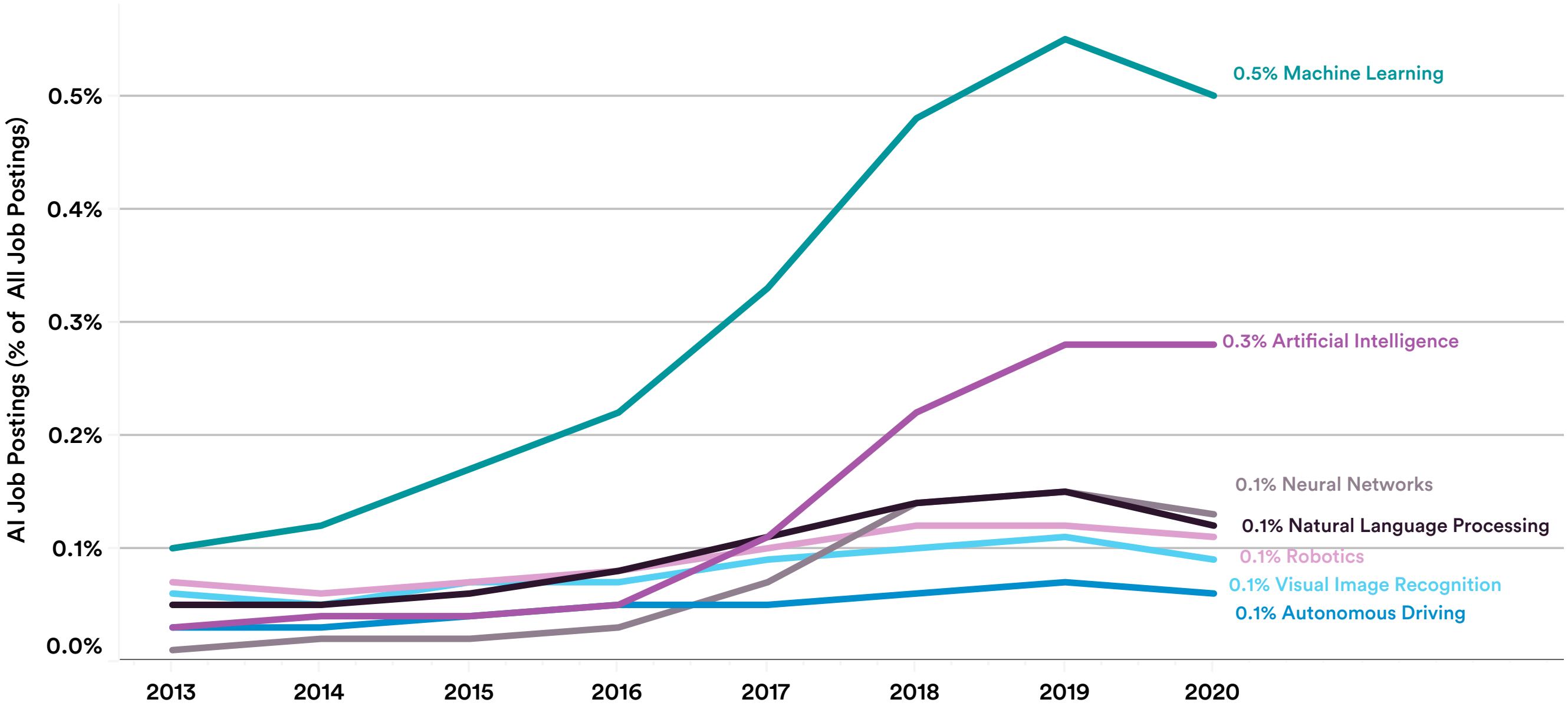

Figure 3.1.4

## U.S. Labor Demand: By Industry

To dive deeper into how AI job demand in the U.S. labor market varies across industries, this section looks at the share of AI job postings across all jobs posted in the United States by industry in 2020 (Figure 3.1.5) as well as the trend in the past 10 years (Figure 3.1.6).

In 2020, industries focused on information (2.8%); professional, scientific, and technical services (2.5%); and agriculture, forestry, fishing, and hunting (2.1%) had the highest share of AI job postings among all job postings in the United States. While the first two have always dominated demand for AI jobs, the agriculture, forestry, fishing, and hunting industry saw the biggest jump—by almost 1 percentage point—in the share of AI jobs from 2019 to 2020.

AI JOB POSTINGS (% of ALL JOB POSTINGS) in the UNITED STATES by INDUSTRY, 2020
Source: Burning Glass, 2020 | Chart: 2021 AI Index Report

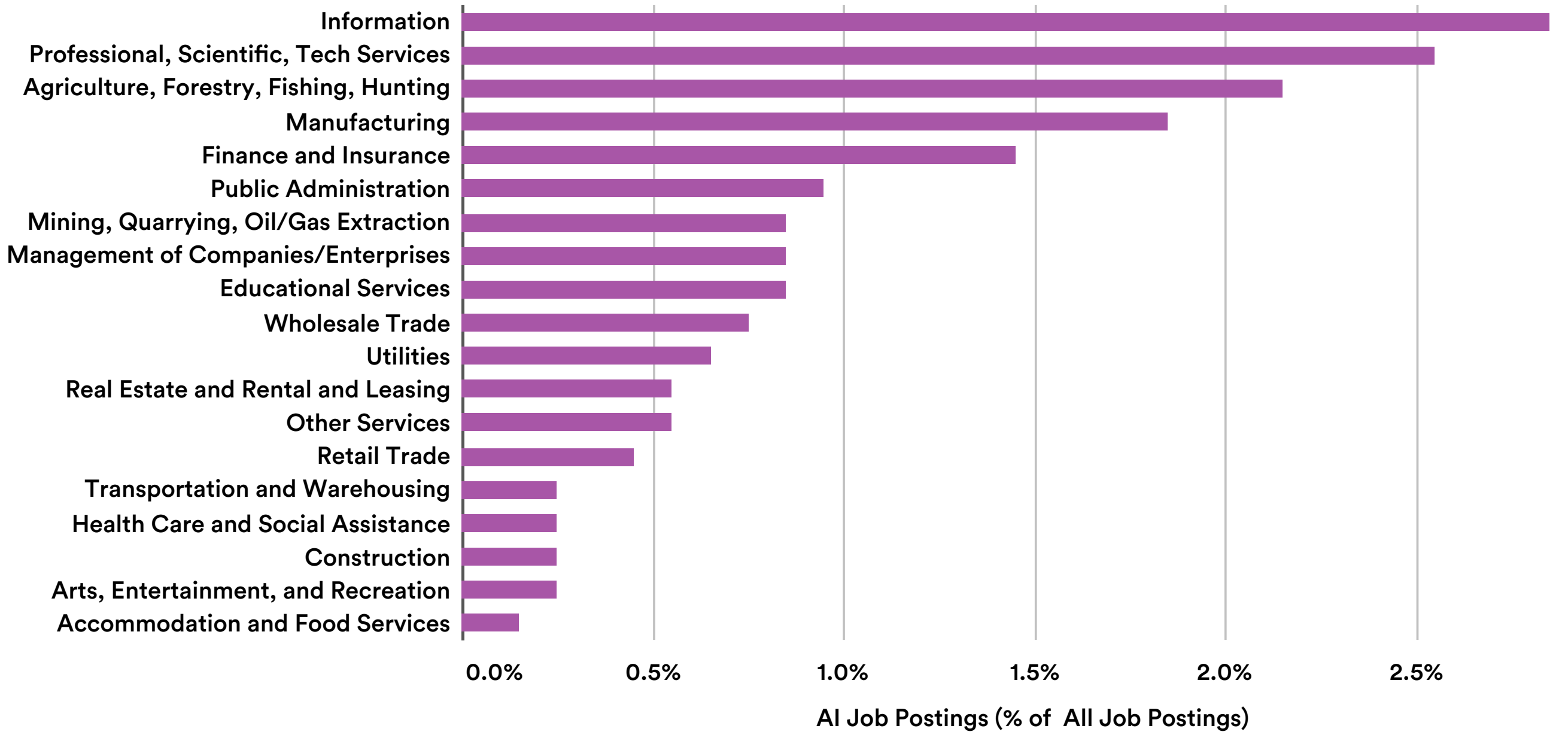

Figure 3.1.5

AI JOB POSTINGS (% of ALL JOB POSTINGS) in the UNITED STATES by INDUSTRY, 2013-20
Source: Burning Glass, 2020 | Chart: 2021 AI Index Report

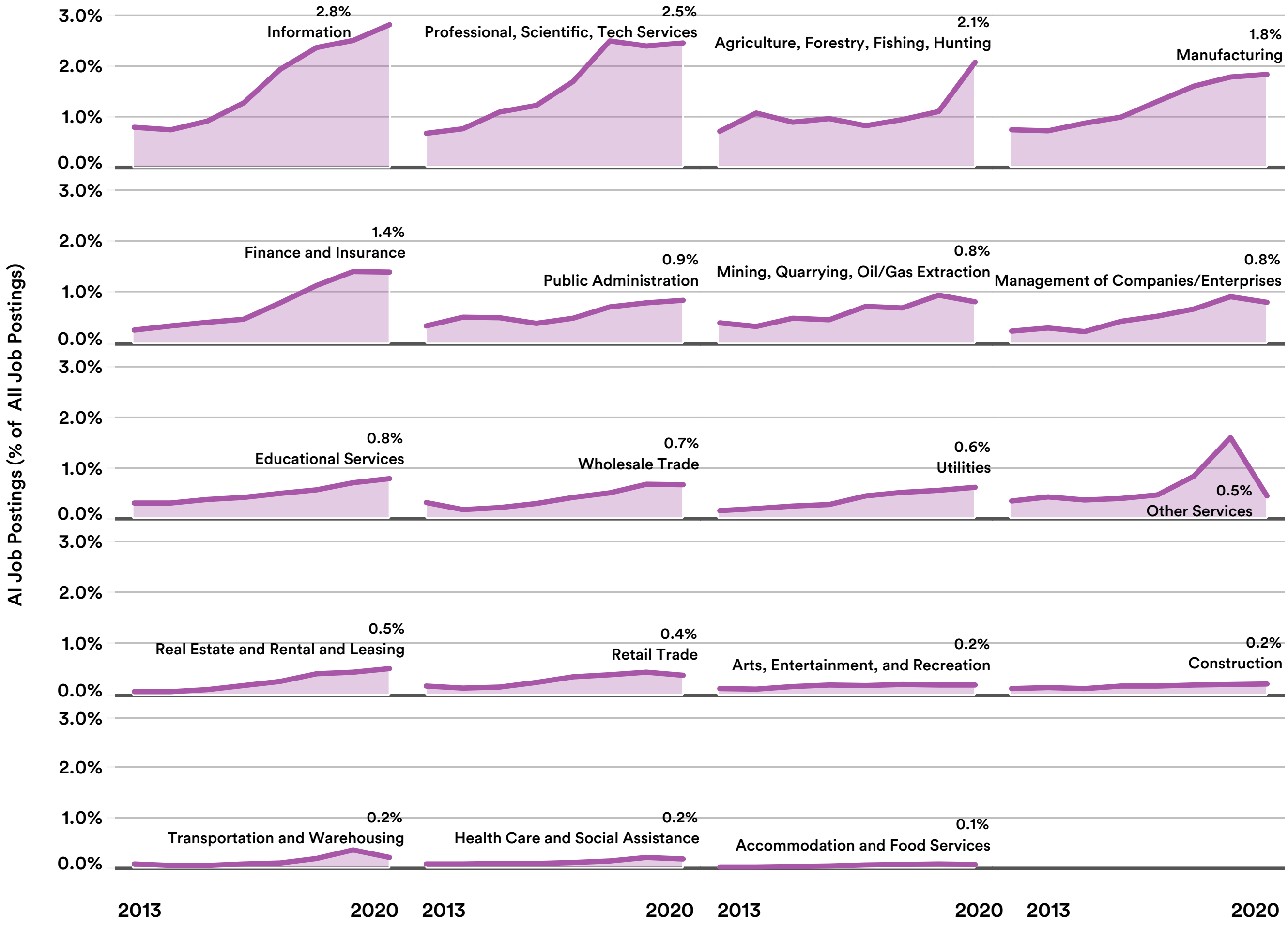

Figure 3.1.6

## U.S. Labor Demand: By State

As the competition for AI talent intensifies, where are companies seeking employees with machine learning, data science, and other AI-related skills within the United States?

Figure 3.1.7 examines the labor demand by U.S. state in 2020, plotting the share of AI job postings across all job postings on the y-axis and the total number of AI jobs posted on a log scale on the x-axis. The chart shows that the District of Columbia has the highest share of AI jobs posted (1.88%), overtaking Washington state in 2019; and California remains the state with the highest number of AI job postings (63,433).

In addition to Washington, D.C., six states registered over 1% of AI job postings among all job postings—Washington, Virginia, Massachusetts, California, New York, and Maryland—compared with five last year. California also has more AI job postings than the next three states combined, which are Texas (22,539), New York (18,580), and Virginia (17,718).

AI JOB POSTINGS (TOTAL and % of ALL JOB POSTINGS) by U.S. STATE and DISTRICT, 2020

Source: Burning Glass, 2020 | Chart: 2021 AI Index Report

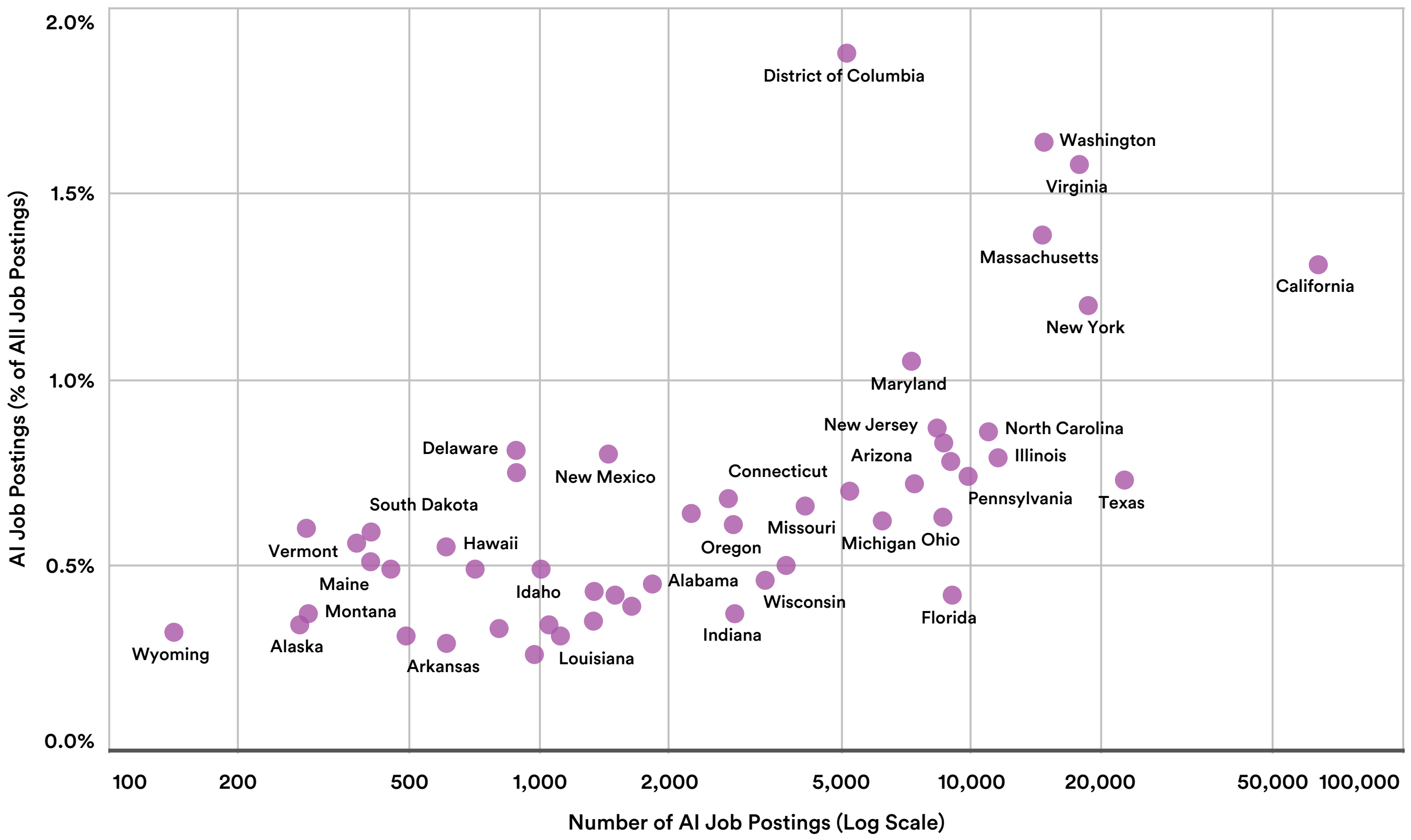

Figure 3.1.7

## AI SKILL PENETRATION

How prevalent are AI skills across occupations? The AI skill penetration metric shows the average share of AI skills among the top 50 skills in each occupation, using LinkedIn data that includes skills listed on a member's profile, positions held, and the locations of the positions.

### Global Comparison

For cross-country comparison, the relative penetration rate of AI skills is measured as the sum of the penetration of each AI skill across occupations in a given country, divided by the average global penetration of AI skills across the same occupations. For example, a relative penetration rate of 2 means that the average penetration of AI skills in that country is 2 times the global average across the same set of occupations.

Among the sample countries shown in Figure 3.1.8, the aggregated data from 2015 to 2020 shows that India (2.83 times the global average) has the highest relative AI skill penetration rate, followed by the United States (1.99 times the global average), China (1.40 times the global average), Germany (1.27 times the global average), and Canada (1.13 times the global average).[2]

**RELATIVE AI SKILL PENETRATION RATE by COUNTRY, 2015-20**

Source: LinkedIn, 2020 | Chart: 2021 AI Index Report

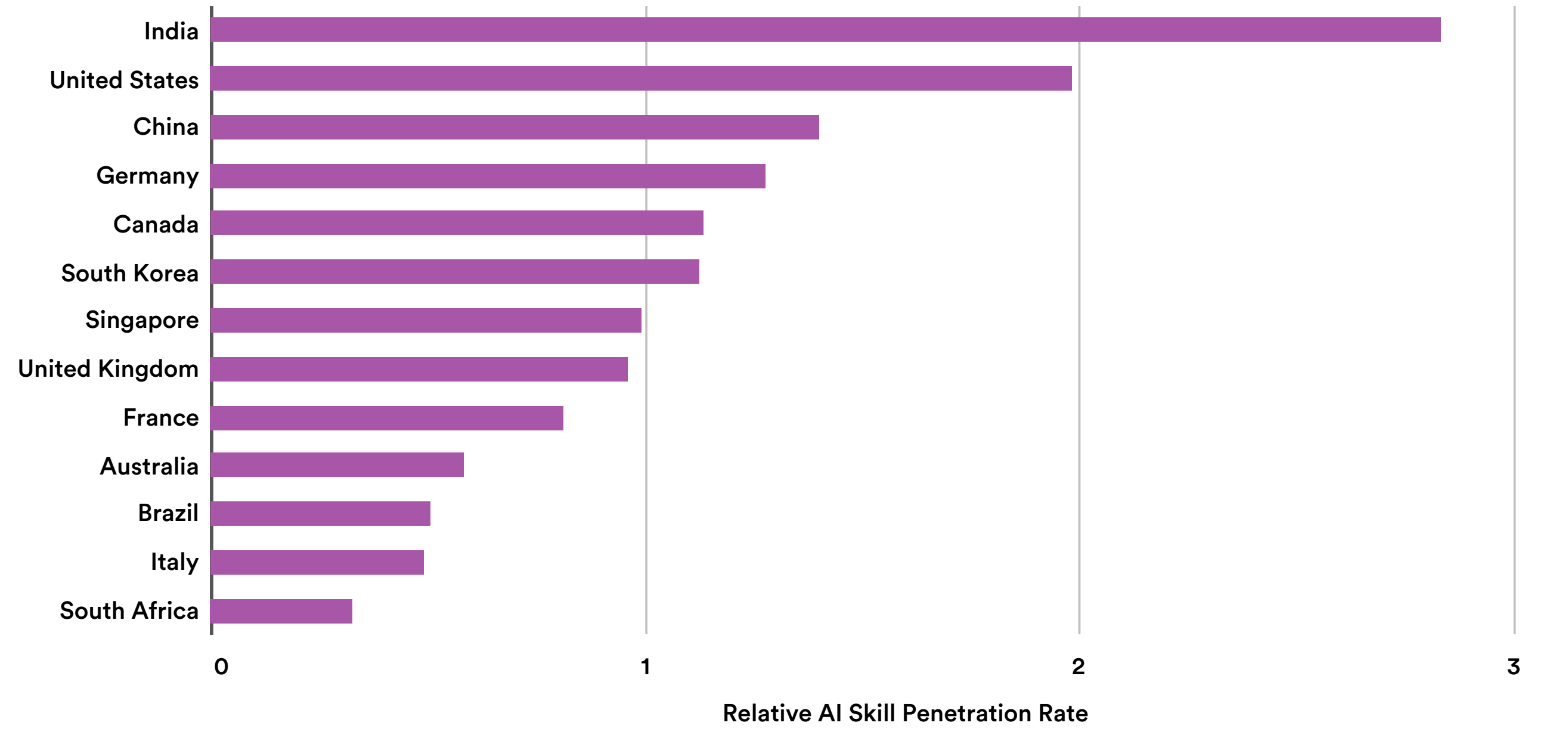

Figure 3.1.8

2 Countries included are a select sample of eligible countries with at least 40% labor force coverage by LinkedIn and at least 10 AI hires in any given month. China and India were included in this sample because of their increasing importance in the global economy, but LinkedIn coverage in these countries does not reach 40% of the workforce. Insights for these countries may not provide as full a picture as other countries, and should be interpreted accordingly.

## Global Comparison: By Industry

To provide an in-depth sectoral decomposition of AI skill penetration across industries and sample countries, Figure 3.1.9 includes the aggregated data of the top five industries with the highest AI skill penetration globally in the last five years: education, finance, hardware and networking, manufacturing, and software and IT.[3] India has the highest relative AI skill penetration across all five industries, while the United States and China frequently appear high up on the list. Other pockets of specialization worth highlighting with relative skill penetration rates of more than 1 include Germany in hardware and networking as well as manufacturing; and Israel in manufacturing and education.

**RELATIVE AI SKILL PENETRATION RATE by INDUSTRY, 2015-20**
Source: LinkedIn, 2020 | Chart: 2021 AI Index Report

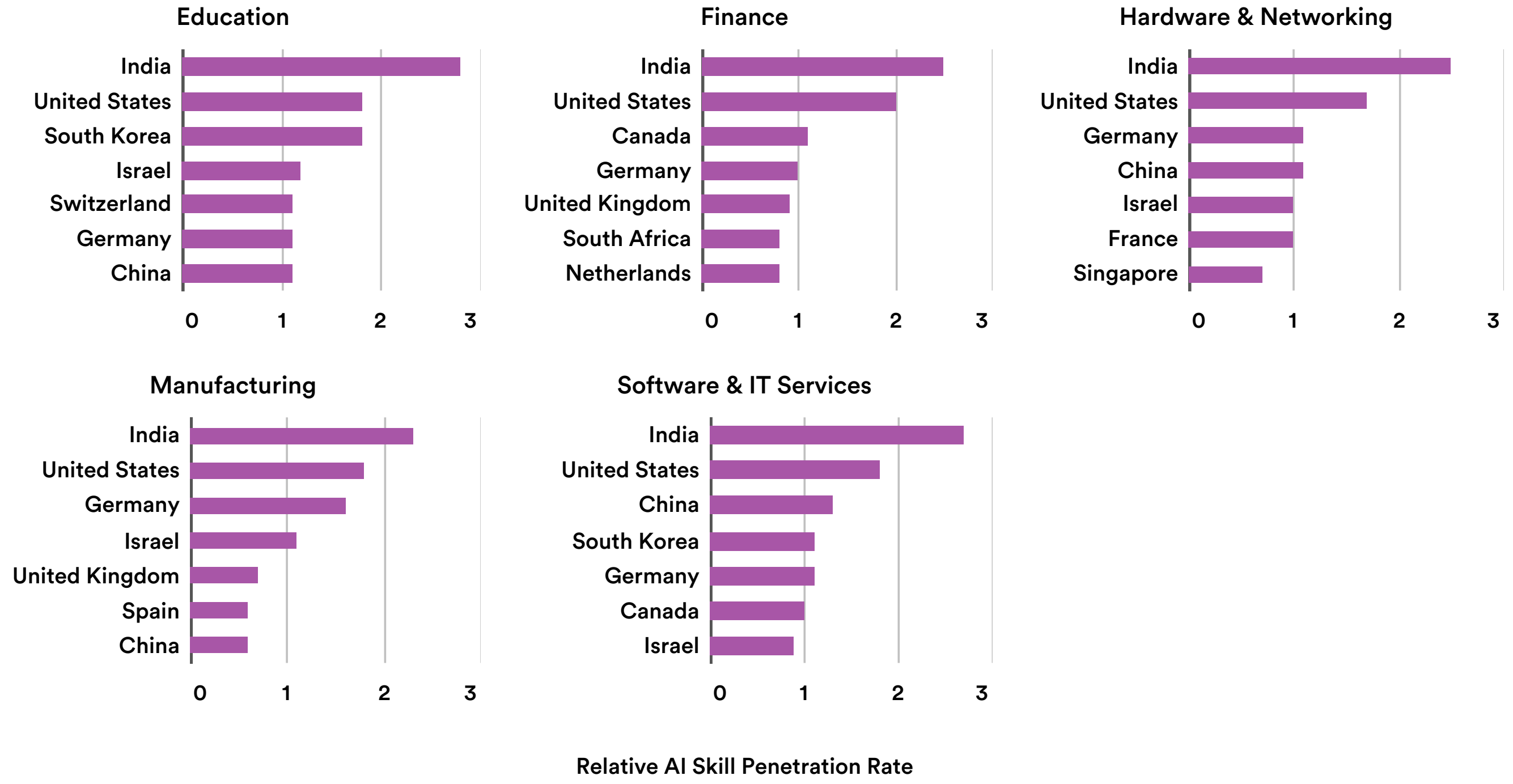

Figure 3.1.9

3 Countries included are a select sample of eligible countries with at least 40% labor force coverage by LinkedIn and at least 10 AI hires in any given month. China and India were included in this sample because of their increasing importance in the global economy, but LinkedIn coverage in these countries does not reach 40% of the workforce. Insights for these countries may not provide as full a picture as other countries, and should be interpreted accordingly.

This section explores the investment activity of private companies by NetBase Quid based on data from CapIQ and Crunchbase. Specifically, it looks at the latest trends in corporate AI investment, such as private investment, public offerings, mergers and acquisitions (M&A), and minority stakes related to AI. The section then focuses on the private investment in AI, or how much private funding goes into AI startups and which sectors are attracting significant investment and in which countries.

# 3.2 INVESTMENT

## CORPORATE INVESTMENT

The total global investment in AI, including private investment, public offerings, M&A, and minority stakes, increased by 40% in 2020 relative to 2019 for a total of USD 67.9 billion (Figure 3.2.1). Given the pandemic, many small businesses have suffered disproportionately. As a result, industry consolidation and increased M&A activity in 2020 are driving up the total corporate investment in AI. M&A made up the majority of the total investment amount in 2020, increasing by 121.7% relative to 2019. Several high-profile acquisitions related to AI took place in 2020, including NVIDIA's acquisition of Mellanox Technologies and Capgemini's of Altran Technologies.

GLOBAL CORPORATE INVESTMENT in AI by INVESTMENT ACTIVITY, 2015-20
Source: CapIQ, Crunchbase, and NetBase Quid, 2020 | Chart: 2021 AI Index Report

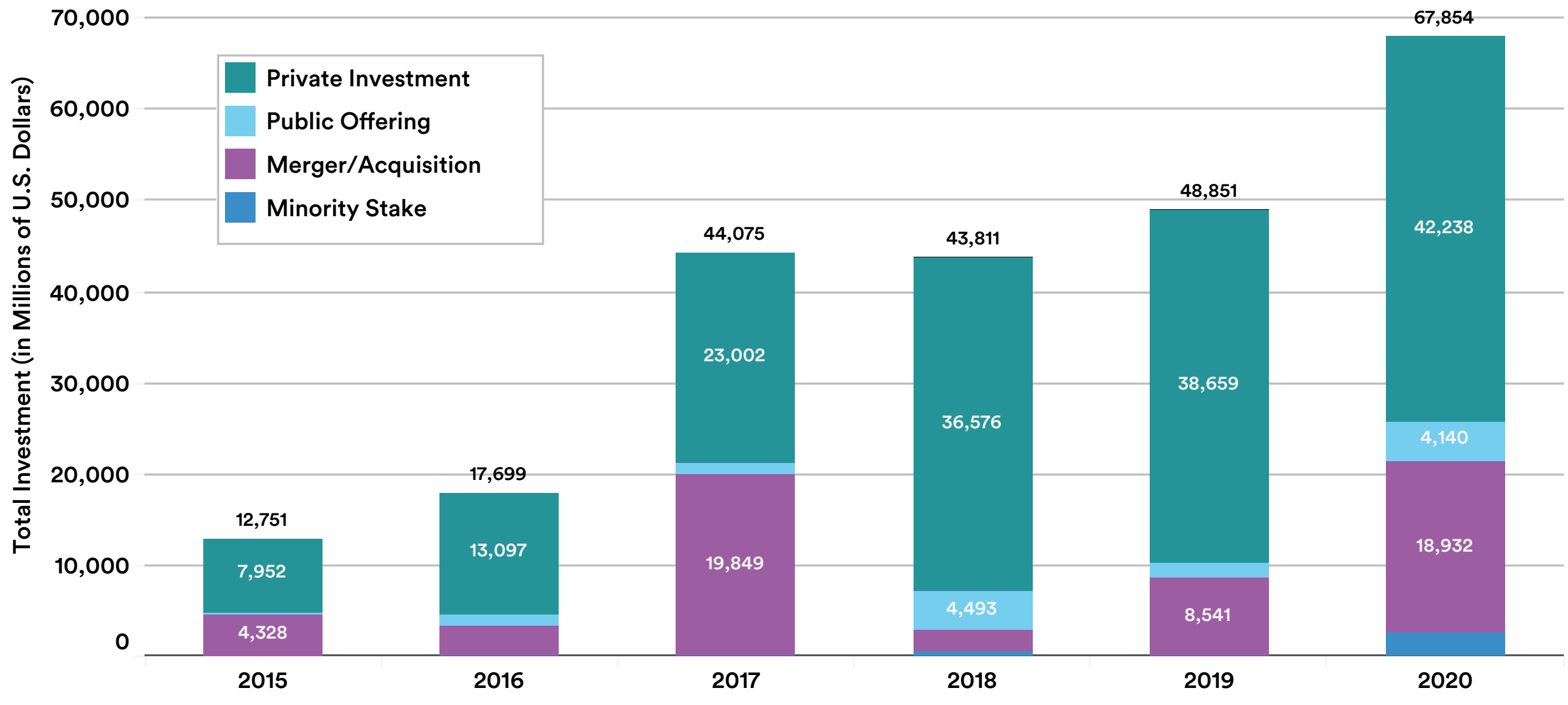

Figure 3.2.1

## STARTUP ACTIVITY

The following section analyzed the trend of private investment in AI startups that have received investments of over USD 400,000 in the last 10 years. While the amount of private investment in AI has soared dramatically in recent years, the rate of growth has slowed.

### Global Trend

More private investment in AI is being funneled into fewer startups. Despite the pandemic, 2020 saw a 9.3% increase in the amount of private AI investment from 2019—a higher percentage than the 5.7% increase in 2019 (Figure 3.2.2), though the number of companies funded decreased for the third year in a row (Figure 3.2.3). While there was a record high of more than USD 40 billion in private investment in 2020, that represents only a 9.3% increase from 2019—compared with the largest increase of 59.0%, observed between 2017 and 2018. Moreover, the number of funded AI startups continued a sharp decline from its 2017 peak.

**PRIVATE INVESTMENT in FUNDED AI COMPANIES, 2015-20**
Source: CapIQ, Crunchbase, and NetBase Quid, 2020 | Chart: 2021 AI Index Report

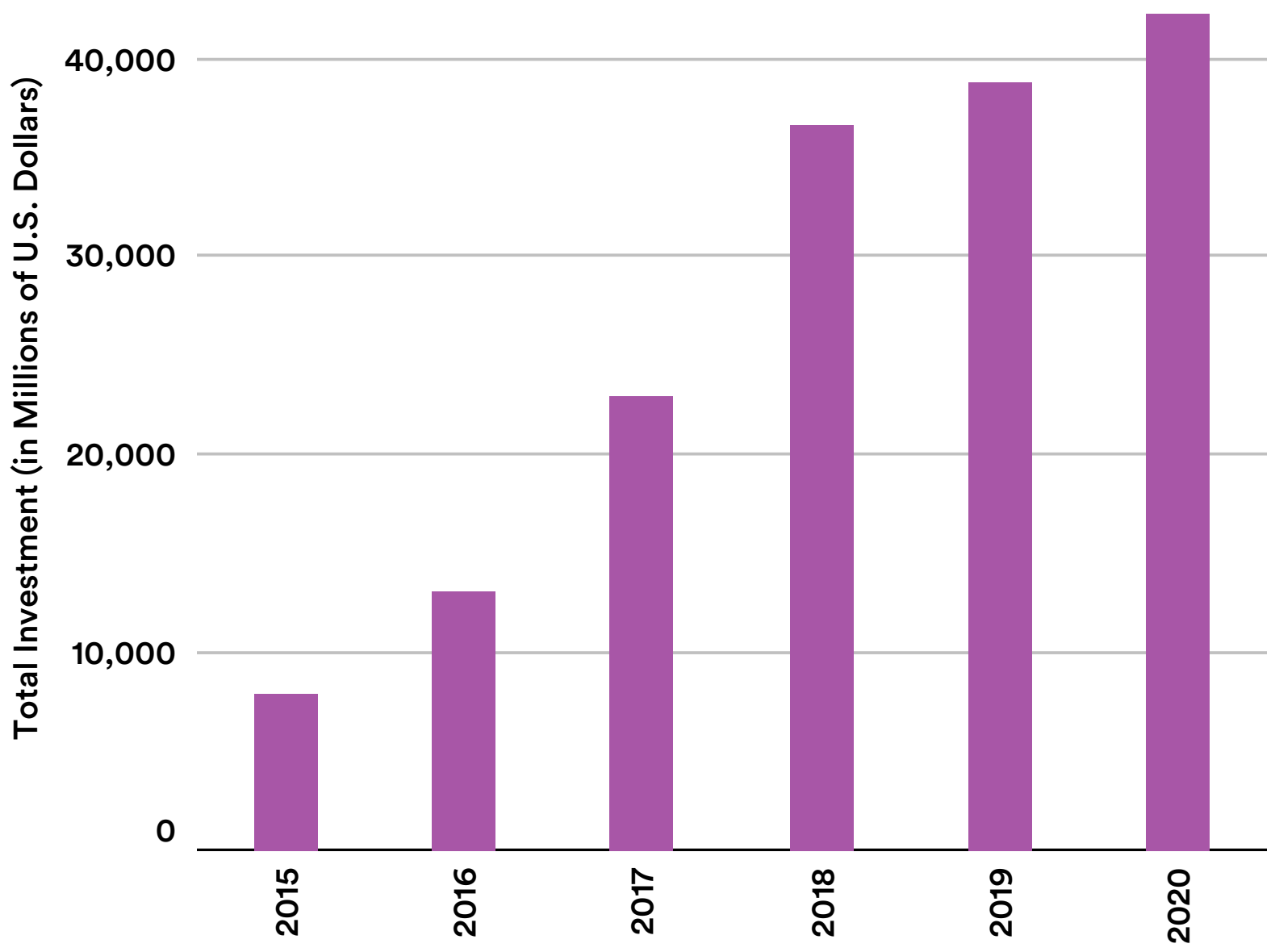

Figure 3.2.2

**NUMBER OF NEWLY FUNDED AI COMPANIES in the WORLD, 2015-20**
Source: CapIQ, Crunchbase, and NetBase Quid, 2020 | Chart: 2021 AI Index Report

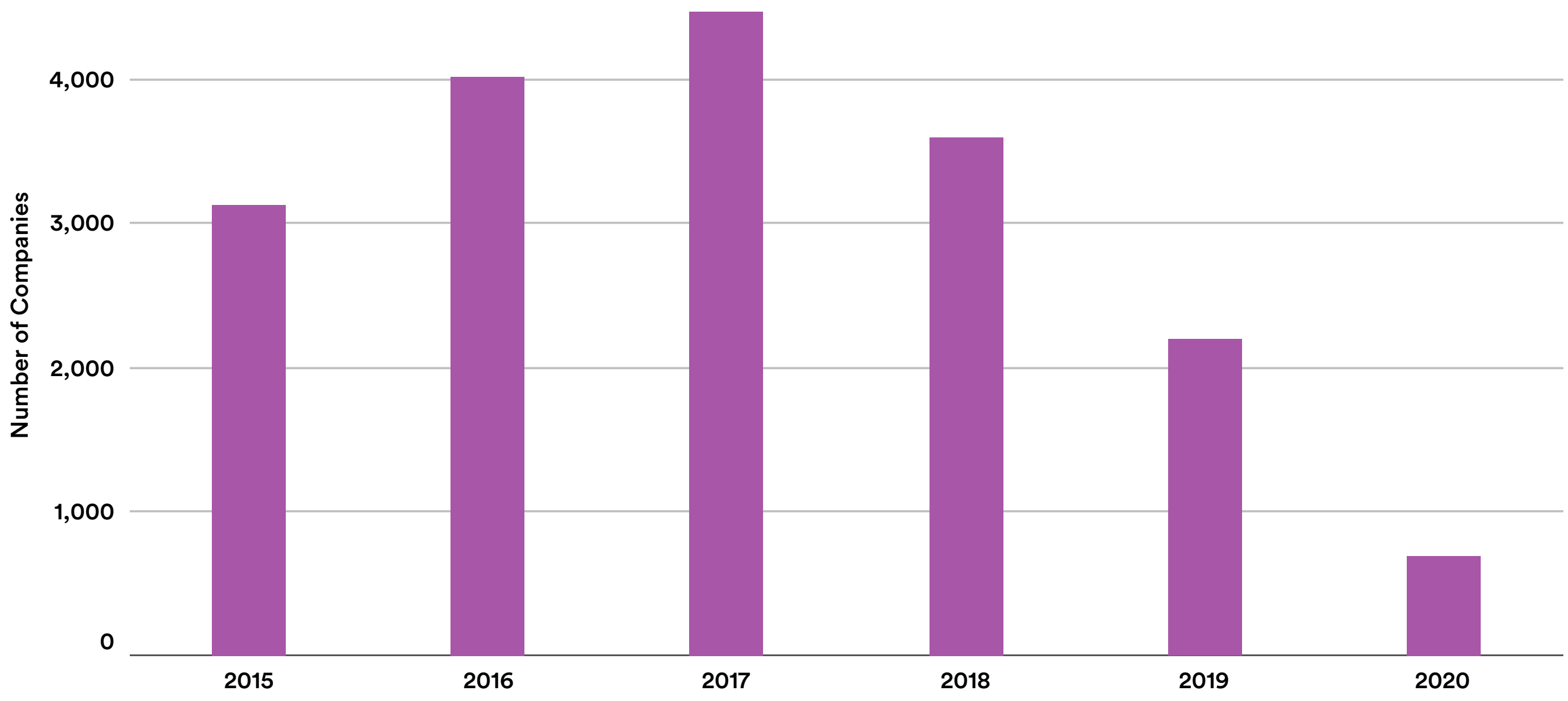

Figure 3.2.3

## Regional Comparison

As shown in Figure 3.2.4, the United States remains the leading destination for private investment, with over USD 23.6 billion in funding in 2020, followed by China (USD 9.9 billion) and the United Kingdom (USD 1.9 billion).

**PRIVATE INVESTMENT in AI by COUNTRY, 2020**

Source: CapIQ, Crunchbase, and NetBase Quid, 2020 | Chart: 2021 AI Index Report

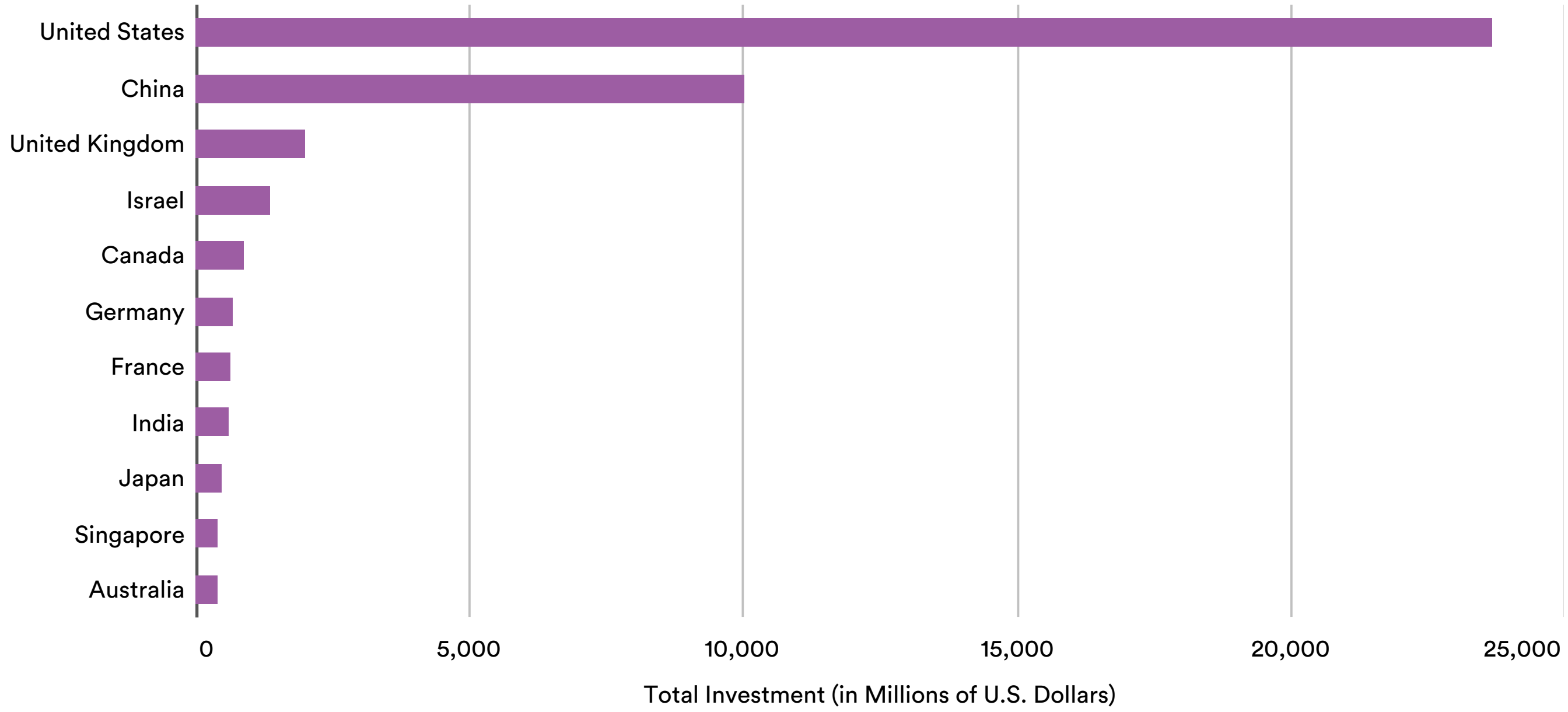

Figure 3.2.4

A closer examination of the three contenders leading the AI race—the United States, China, and the European Union—further validates the United States' dominant position in private AI investment. While China saw an exceptionally high amount of private AI investment in 2018, its investment level in 2020 is less than half that of the United States (Figure 3.2.5). It is important to note, however, that China has strong public investments in AI. Both the central and local governments in China are spending heavily on AI R&D.[4]

4 See "A Brief Examination of Chinese Government Expenditures on Artificial Intelligence R&D" (2020) by the Institute for Defense Analyses for more details.

PRIVATE INVESTMENT in AI by GEOGRAPHIC AREA, 2015-20

Source: CAPIQ, Crunchbase, and NetBase Quid, 2020 | Chart: 2021 AI Index Report

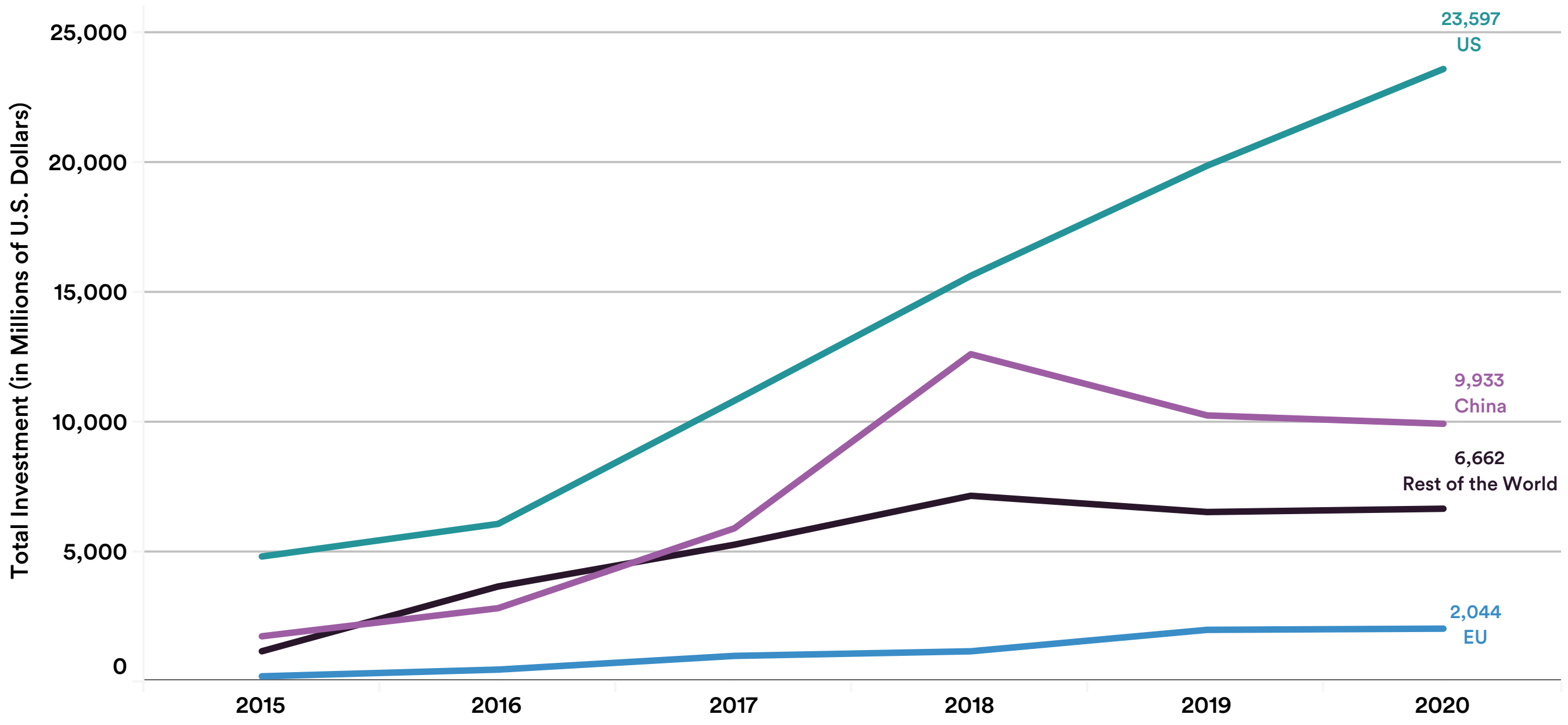

Figure 3.2.5

## Focus Area Analysis

Figure 3.2.6 shows the ranking of the top 10 focus areas that receive the greatest amount of private investment in 2020 as well as their respective investment amount in 2019. The "Drugs, Cancer, Molecular, Drug Discovery" area tops the list, with more than USD 13.8 billion in private AI investment—4.5 times higher than 2019—followed by "Autonomous Vehicles, Fleet, Autonomous Driving, Road" (USD 4.5 billion), and "Students, Courses, Edtech, English Language" (USD 4.1 billion).

In addition to Drugs, Cancer, Molecular, Drug Discovery," both "Games, Fans, Gaming, Football" and "Students, Courses, Edtech, English Language" saw a significant increase in the amount of private AI investment from 2019 to 2020. The former is largely driven by several financing rounds to gaming and sports startups in the United States and South Korea, while the latter is boosted by investments in an online education platform in China.

**GLOBAL PRIVATE INVESTMENT in AI by FOCUS AREA, 2019 vs 2020**
Source: CapIQ, Crunchbase, and NetBase Quid, 2020 | Chart: 2021 AI Index Report

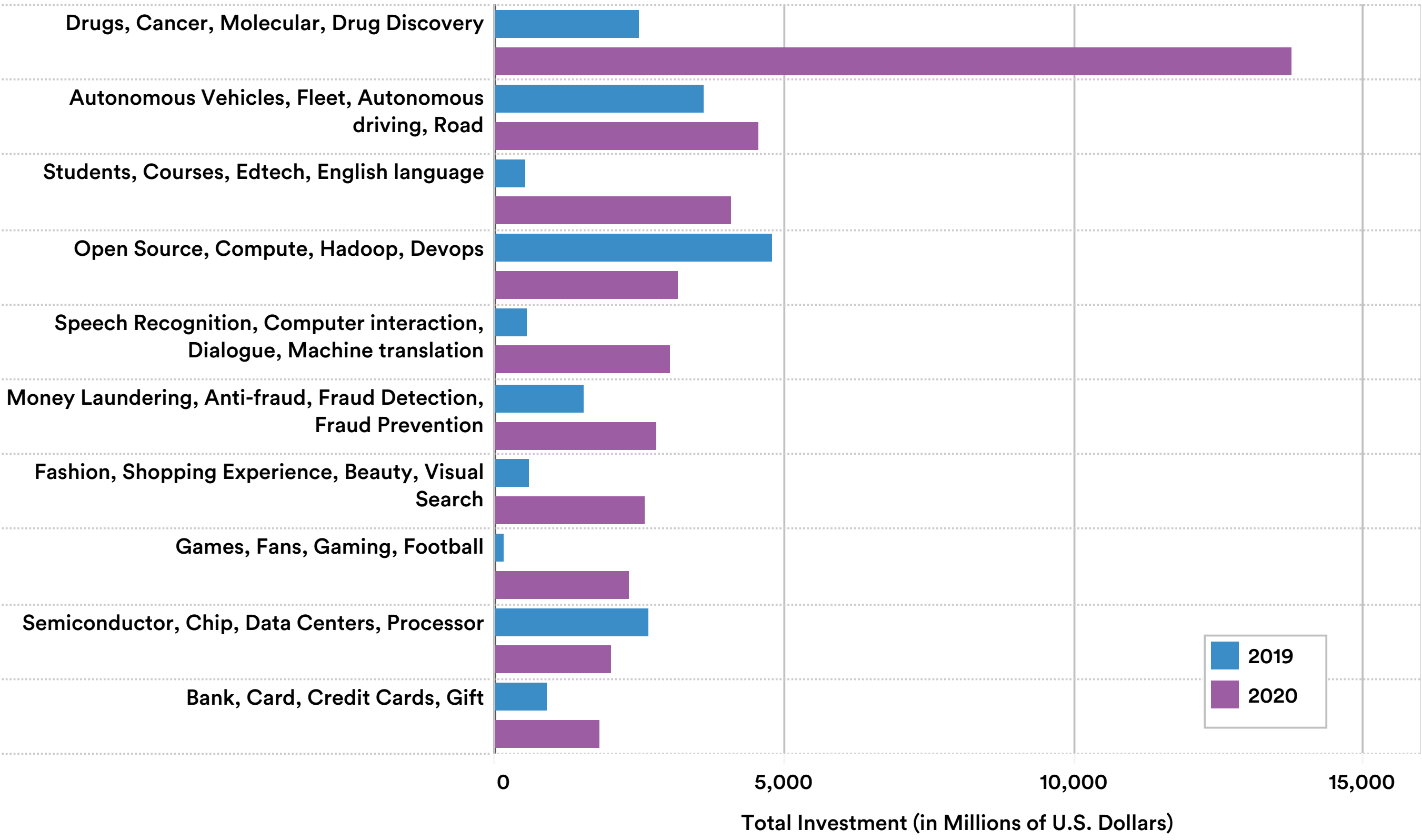

Figure 3.2.6

This section reviews how corporations have capitalized on the advances in AI, using AI and automation to their advantage and generating value at scale. While the number of corporations starting to deploy AI technologies has surged in recent years, the economic turmoil and impact of COVID-19 in 2020 have slowed that rate of adoption. The latest trends in corporate AI activities are examined through data on the adoption of AI capabilities by McKinsey's Global Survey on AI, trends in robot installations across the globe by the International Federation of Robotics (IFR), and mentions of AI in corporate earnings calls by Prattle.

# 3.3 CORPORATE ACTIVITY

## INDUSTRY ADOPTION

This section shares the results of a McKinsey & Company survey of 2,395 respondents: individuals representing companies from a range of regions, industries, sizes, functional specialties, and tenures.

McKinsey & Company's "The State of AI in 2020" report contains the full results of this survey, including insights on how different companies have adopted AI across functions, core best practices shared among the companies that are generating the greatest value from AI, and the impacts of the COVID-19 pandemic on these companies' AI investment plans.

### Global Adoption of AI

The 2020 survey results suggest no increase in AI adoption relative to 2019. Over 50% of respondents say that their organizations have adopted AI in at least one business function (Figure 3.3.1). In 2019, 58% of respondents said their companies adopted AI in at least one function, although the 2019 survey asked about companies' AI adoption differently.

In 2020, companies in developed Asia-Pacific countries led in AI adoptions, followed by those in India and North America. While AI adoption was about equal across regions in 2019, this year's respondents working for companies in Latin America and in other developing countries are much less likely to report adopting AI in at least one business function.

AI ADOPTION by ORGANIZATIONS GLOBALLY, 2020

Source: McKinsey & Company, 2020 | Chart: 2021 AI Index Report

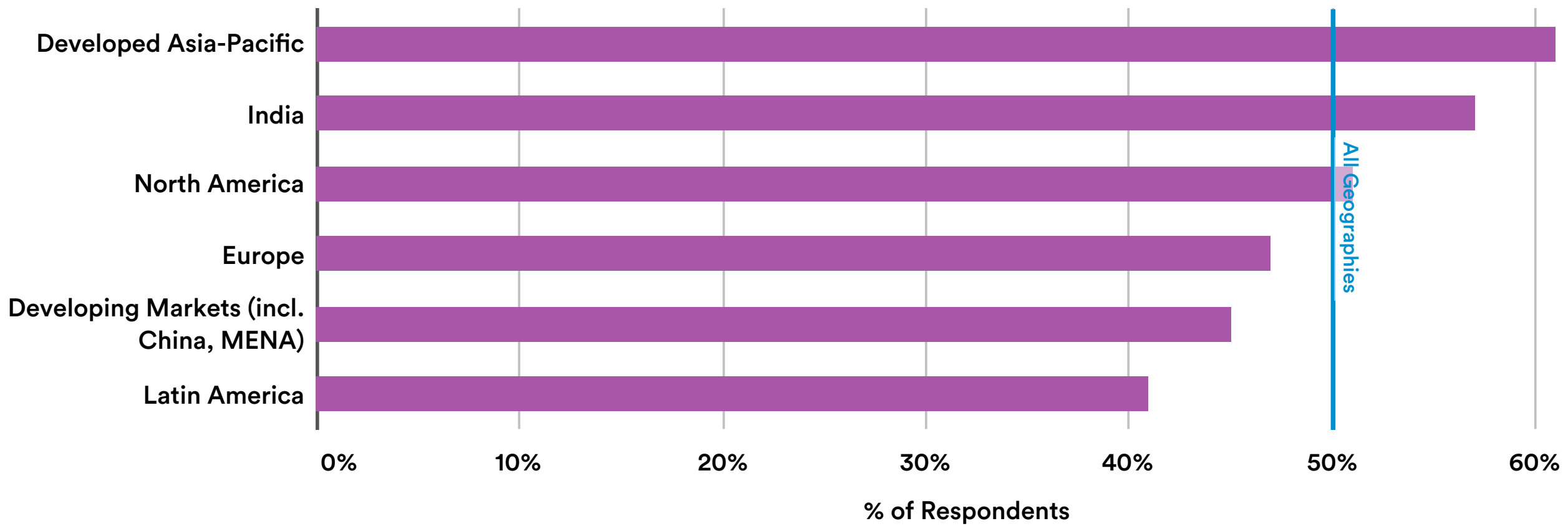

Figure 3.3.1

## AI Adoption by Industry and Function

Respondents representing companies in high tech and telecom were most likely to report AI adoption in 2020, similar to the 2019 results, followed in second place by both financial services and automotive and assembly (Figure 3.3.2).

In another repeat from 2019 (and 2018), the 2020 survey suggests that the functions where companies are most likely to adopt AI vary by industry (Figure 3.3.3). For example, respondents in the automotive and assembly industry report greater AI adoption for manufacturing-related tasks than any other; respondents in financial services report greater AI adoption for risk functions; and respondents in high tech and telecom report greater AI adoption for product and service development functions.

Across industries, companies in 2020 are most likely to report using AI for service operations (such as field services, customer care, back office), product and service development, and marketing and sales, similar to the survey results in 2019.

## Type of AI Capabilities Adopted

By industry, the type of AI capabilities adopted varies (Figure 3.3.4). Across industries, companies in 2020 were most likely to identify other machine learning techniques, robotic process automation, and computer vision as capabilities adopted in at least one business function.

Industries tend to adopt AI capabilities that best serve their core functions. For example, physical robotics, as well as autonomous vehicles, are most frequently adopted by industries where manufacturing and distribution play a large role—such as automotive and assembly, and consumer goods and retail. Natural language processing capabilities, such as text understanding, speech understanding, and text generation, are frequently adopted by industries with high volumes of customer or operational data in text forms; these include business, legal, and professional services, financial services, healthcare, and high tech and telecom.

**AI ADOPTION by INDUSTRY, 2020**
Source: McKinsey & Company, 2020 | Chart: 2021 AI Index Report

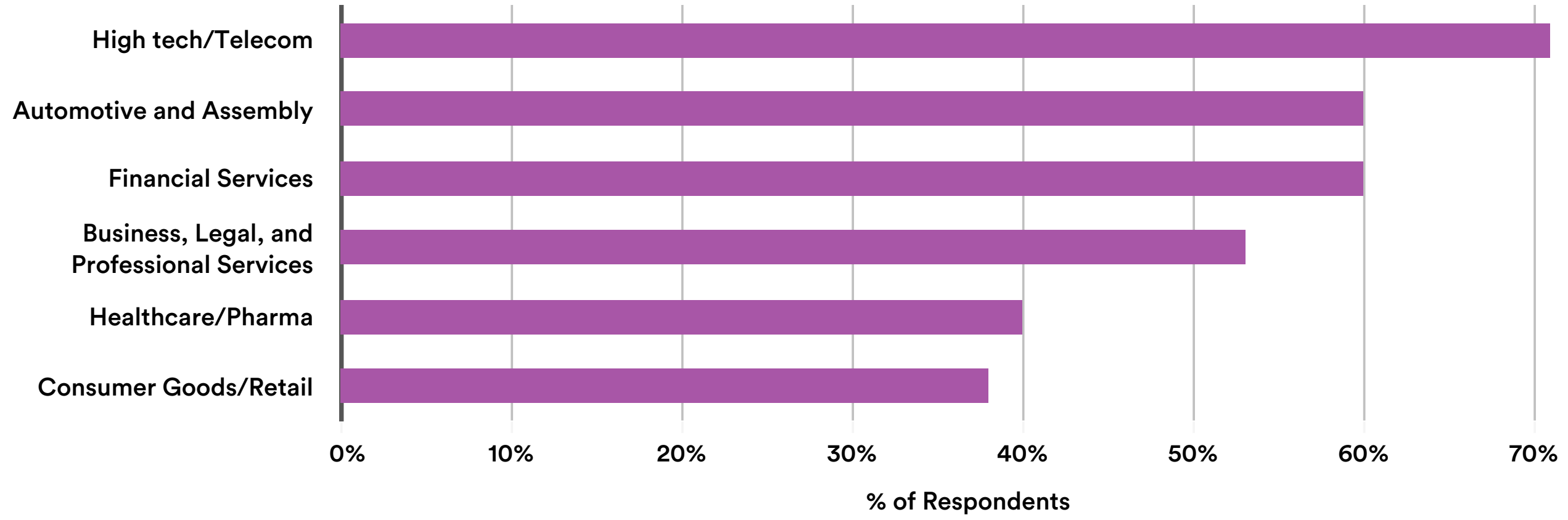

Figure 3.3.2

## AI ADOPTION by INDUSTRY & FUNCTION, 2020

Source: McKinsey & Company, 2020 | Chart: 2021 AI Index Report

| Industry | Human Resources | Manufacturing | Marketing And Sales | Product and/or Service Development | Risk | Service Operations | Strategy and Corporate Finance | Supply-Chain Management |
|---|---|---|---|---|---|---|---|---|
| All Industries | 8% | 12% | 15% | 21% | 10% | 21% | 7% | 9% |
| Automotive and Assembly | 13% | 29% | 10% | 21% | 2% | 16% | 8% | 18% |
| Business, Legal, and Professional Services | 13% | 9% | 16% | 21% | 13% | 20% | 10% | 9% |
| Consumer Goods/Retail | 1% | 19% | 20% | 14% | 3% | 10% | 2% | 10% |
| Financial Services | 5% | 5% | 21% | 15% | 32% | 34% | 7% | 2% |
| Healthcare/Pharma | 3% | 12% | 16% | 15% | 4% | 11% | 2% | 6% |
| High Tech/Telecom | 14% | 11% | 26% | 37% | 14% | 39% | 9% | 12% |

% of Respondents

Figure 3.3.3

## AI CAPABILITIES EMBEDDED in STANDARD BUSINESS PROCESSES, 2020

Source: McKinsey & Company, 2020 | Chart: 2021 AI Index Report

| Industry | Autonomous Vehicles | Computer Vision | Conversational Interfaces | Deep Learning | NL Generation | NL Speech Understanding | NL Text Understanding | Other Machine Learning Techniques | Physical Robotics | Robotic Process Automation |
|---|---|---|---|---|---|---|---|---|---|---|
| All Industries | 7% | 18% | 15% | 16% | 11% | 12% | 13% | 23% | 13% | 22% |
| Automotive and Assembly | 20% | 33% | 16% | 19% | 12% | 14% | 19% | 27% | 31% | 33% |
| Business, Legal, and Professional Services | 7% | 13% | 17% | 19% | 14% | 15% | 18% | 25% | 11% | 13% |
| Consumer Goods/Retail | 13% | 10% | 9% | 6% | 6% | 6% | 9% | 12% | 23% | 14% |
| Financial Services | 6% | 18% | 24% | 19% | 18% | 19% | 26% | 32% | 8% | 37% |
| Healthcare/Pharma | 1% | 15% | 10% | 14% | 12% | 11% | 15% | 19% | 10% | 18% |
| High Tech/Telecom | 9% | 34% | 32% | 30% | 18% | 25% | 33% | 37% | 14% | 34% |

% of Respondents

Figure 3.3.4

## Consideration and Mitigation of Risks from Adopting AI

Only a minority of companies acknowledge the risks associated with AI, and even fewer report taking steps to mitigate those risks (Figure 3.3.5 and Figure 3.3.6). Relative to 2019, the share of survey respondents citing each risk as relevant has largely remained flat; that is, most changes were not statistically significant. Cybersecurity remains the only risk a majority of respondents say their organizations consider relevant. A number of less commonly cited risks, such as national security and political stability, were more likely to be seen as relevant by companies in 2020 than in 2019.

Despite growing calls to attend to ethical concerns associated with the use of AI, efforts to address these concerns in the industry are limited. For example, concerns such as equity and fairness in AI use continue to receive comparatively little attention from companies. Moreover, fewer companies in 2020 view personal or individual privacy as a risk from adopting AI compared with in 2019, and there is no change in the percentage of respondents whose companies are taking steps to mitigate this particular risk.

## RISKS from ADOPTING AI THAT ORGANIZATIONS CONSIDER RELEVANT, 2020

Source: McKinsey & Company, 2020 | Chart: 2021 AI Index Report

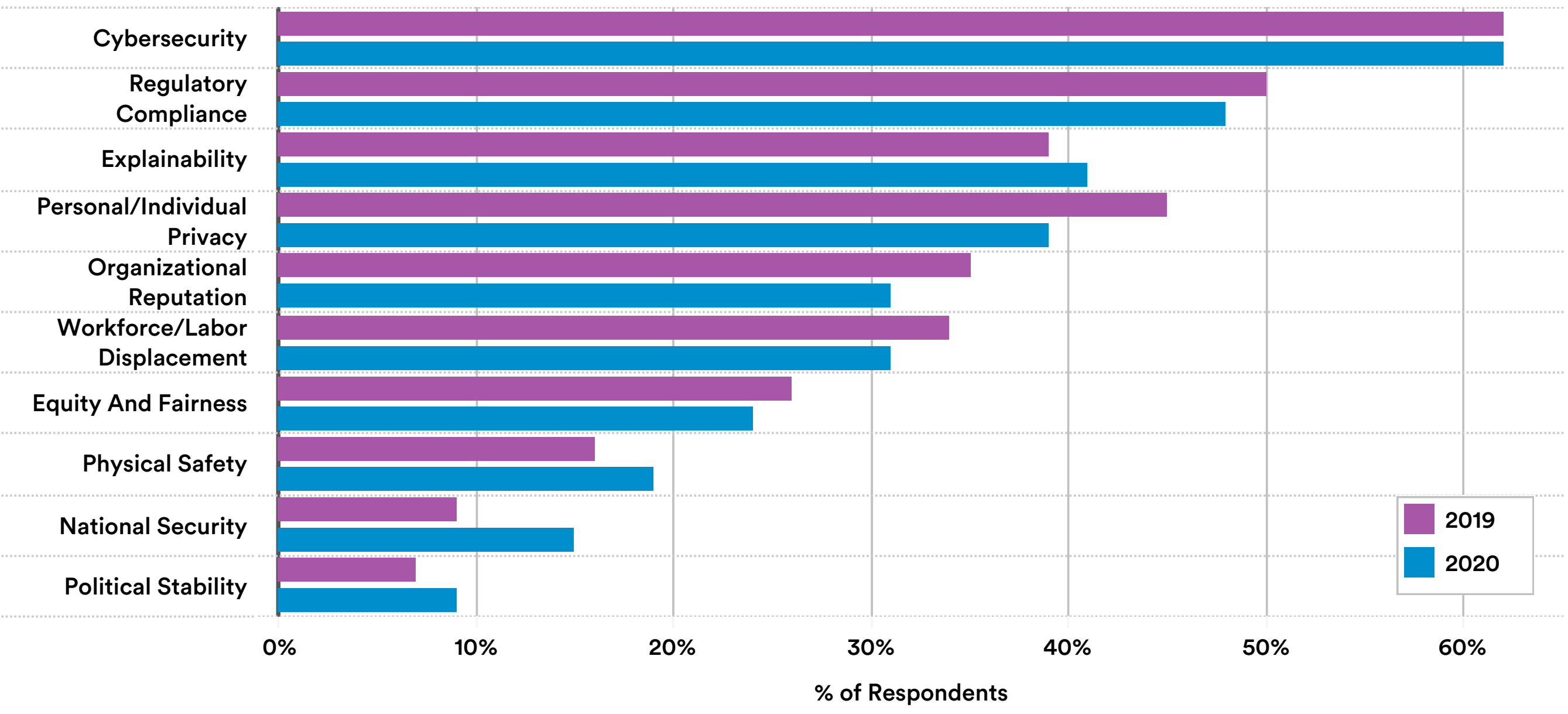

Figure 3.3.5

## RISKS from ADOPTING AI THAT ORGANIZATIONS TAKE STEPS to MITGATE, 2020

Source: McKinsey & Company, 2020 | Chart: 2021 AI Index Report

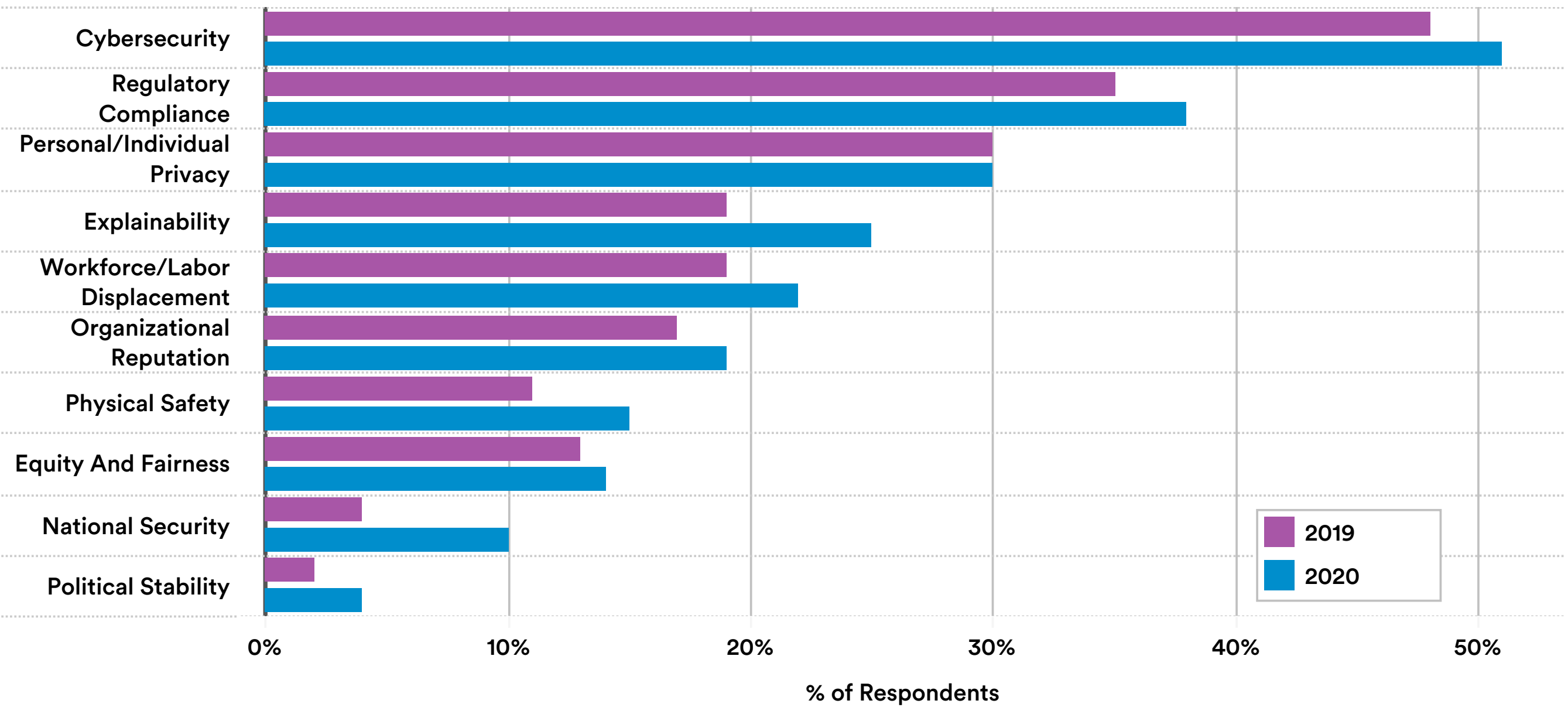

Figure 3.3.6

## The Effect of COVID-19

Despite the economic downturn caused by the pandemic, half of respondents said the pandemic had no effect on their investment in AI, while 27% actually reported increasing their investment. Less than a fourth of businesses decreased their investment in AI (Figure 3.3.7).[5] By industry, respondents in healthcare and pharma as well as automotive and assembly were the most likely to report that their companies had increased investment in AI.

CHANGES in AI INVESTMENTS AMID the COVID-19 PANDEMIC
Source: McKinsey & Company, 2020 | Chart: 2021 AI Index Report

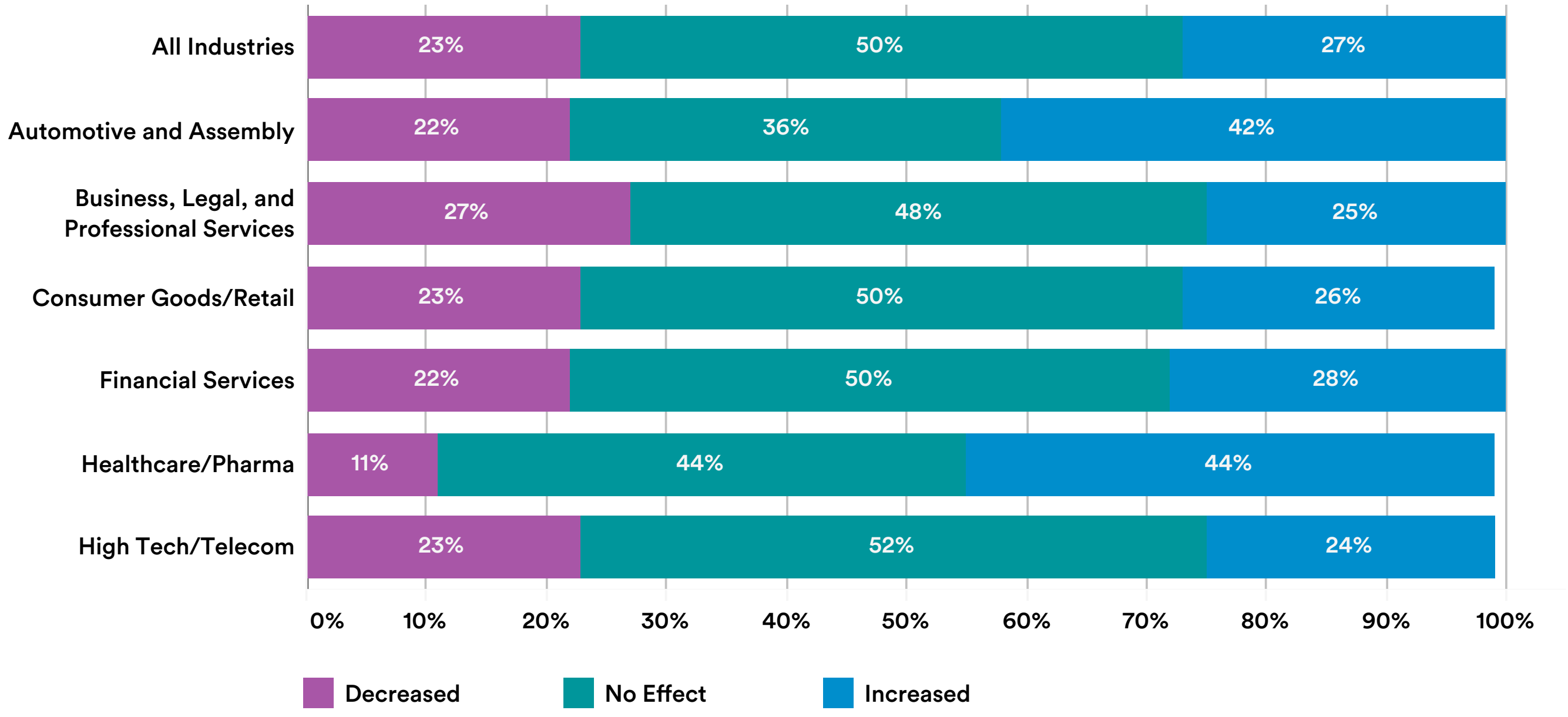

Figure 3.3.7

5 Figures may not sum to 100% because of rounding.

## INDUSTRIAL ROBOT INSTALLATIONS

Right now, AI is being deployed widely onto consumer devices like smartphones and personal vehicles (e.g., self-driving technology). But relatively little AI is deployed on actual robots.[6] That may change as researchers develop software to integrate AI-based approaches with contemporary robots. For now, it is possible to measure global sales of industrial robots to draw conclusions about the amount of AI-ready infrastructure being bought worldwide. While the COVID-19-induced economic crisis will lead to a decline in robot sales in the short term, the International Federation of Robotics (IFR) expects the pandemic to generate global growth opportunities for the robotics industry in the medium term.

### Global Trend

After six years of growth, the number of new industrial robots installed worldwide decreased by 12%, from 422,271 units in 2018 to 373,240 units in 2019 (Figure 3.3.8). The decline is a product of trade tensions between the United States and China as well as challenges faced by the two primary customer industries: automotive and electrical/electronics.

With the automotive industry taking the lead (28% of total installations), followed by electrical/electronics (24%), metal and machinery (12%), plastics and chemical products (5%), and food and beverages (3%).[7] It is important to note that these metrics are a measurement of installed infrastructure that is susceptible to adopting new AI technologies and does not indicate whether every new robot used a significant amount of AI.

**GLOBAL INDUSTRIAL ROBOT INSTALLATIONS, 2012-19**

Source: International Federation of Robotics, 2020 | Chart: 2021 AI Index Report

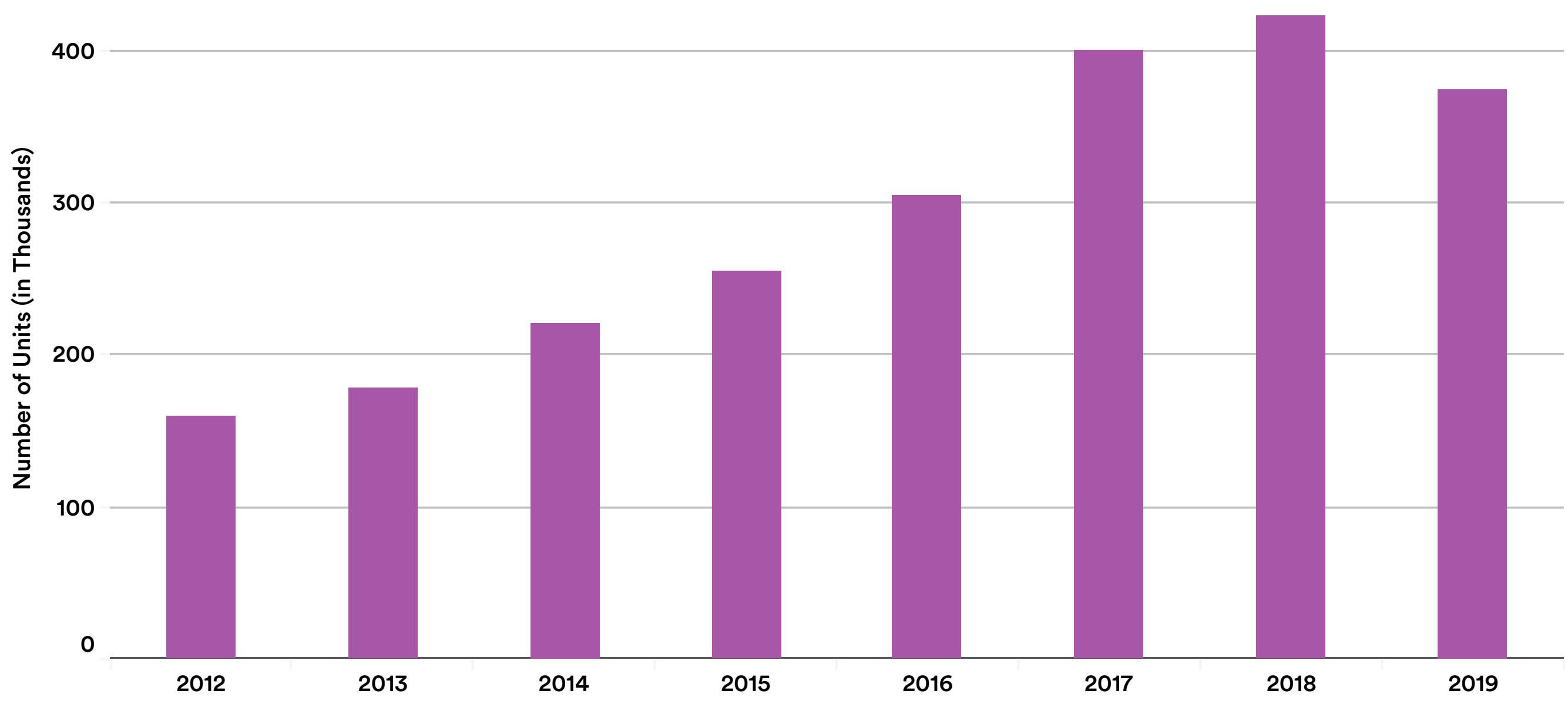

Figure 3.3.8

6 For more insights on the adoption of AI and robots by the industry, read the National Bureau of Economic Research working paper based on the 2018 Annual Business Survey by the U.S. Census Bureau, "Advancing Technologies Adoption and Use by U.S. Firms: Evidence From the Annual Business Survey" (2020).
7 Note that there is no information on the customer industry for approximately 20% of robots installed.

### Regional Comparison

Asia, Europe, and North America—three of the largest industrial robot markets—all witnessed the end of a six-year growth period in robot installations (Figure 3.3.9). North America experienced the sharpest decline, of 16%, in 2019, compared with 5% in Europe and 13% in Asia.

Figure 3.3.10 shows the number of installations in the five major markets for industrial robot markets. All five—accounting for 73% of global robot installations—saw roughly the same decline, except for Germany, which saw a slight bump in installations between 2018 and 2019. Despite the downward trend in China, it is worth noting that the country had more industrial robots in 2019 than the other four countries combined.

NEW INDUSTRIAL ROBOT INSTALLATIONS by REGION, 2017-19
Source: International Federation of Robotics, 2020 | Chart: 2021 AI Index Report

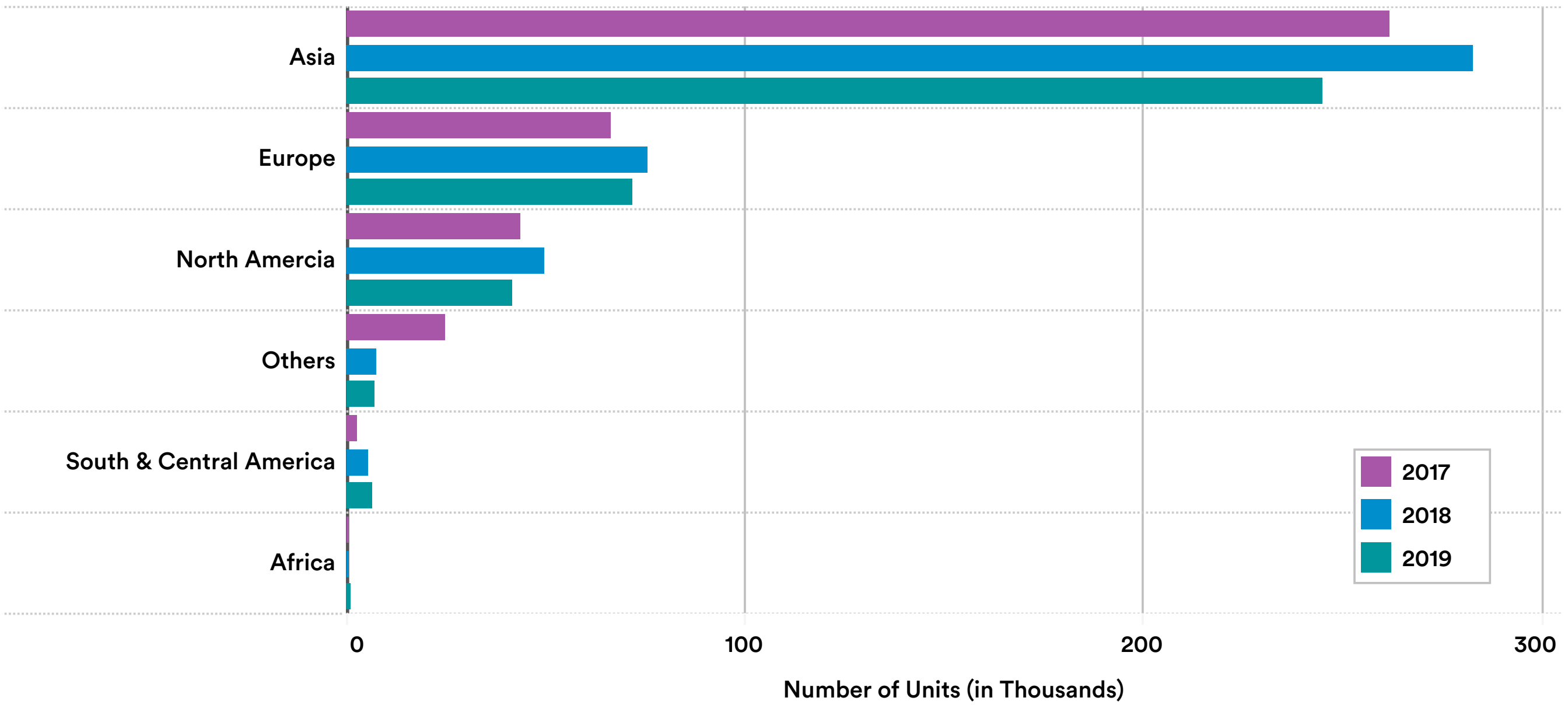

Figure 3.3.9

NEW INDUSTRIAL ROBOT INSTALLATIONS in FIVE MAJOR MARKETS, 2017-19

Source: International Federation of Robotics, 2020 | Chart: 2021 AI Index Report

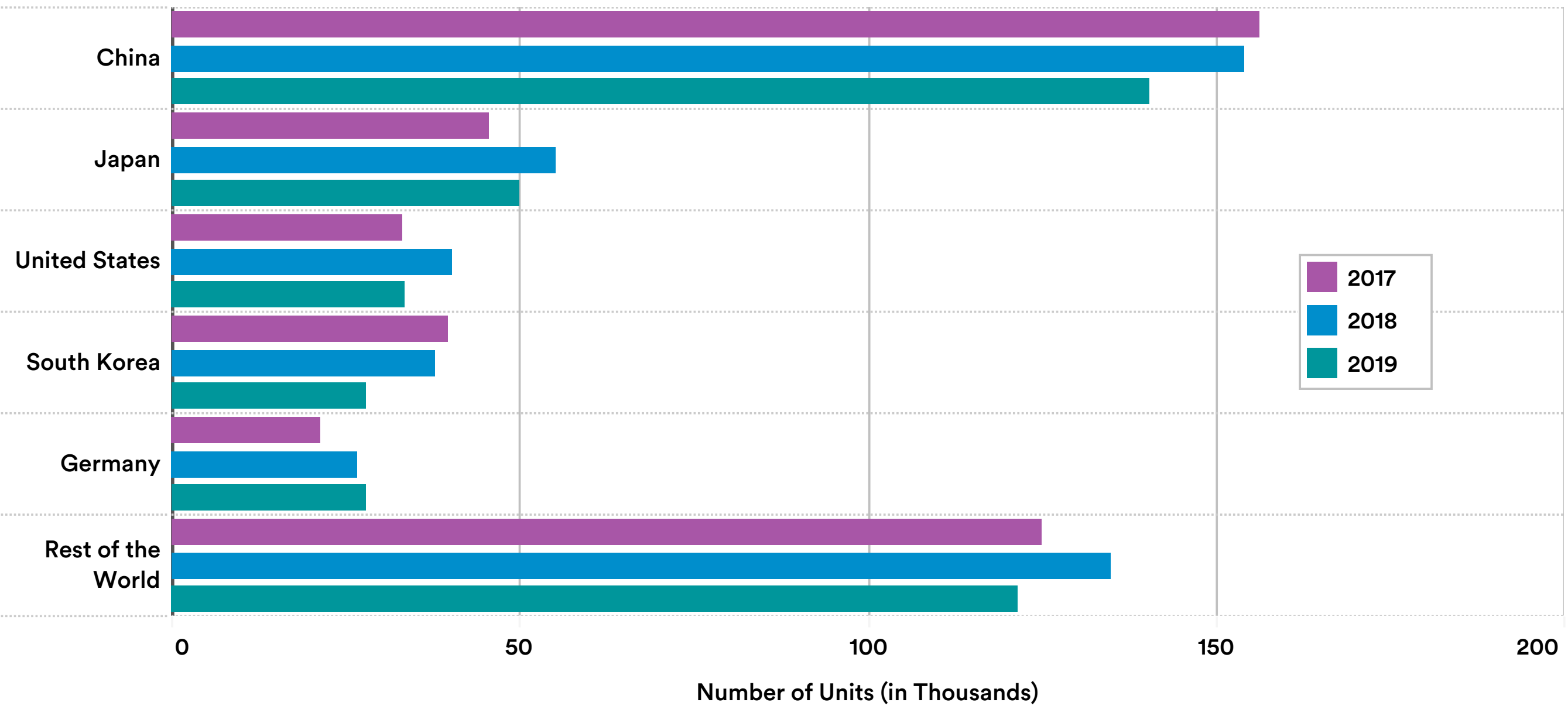

Figure 3.3.10

## EARNINGS CALLS

Mentions of AI in corporate earnings calls have increased substantially since 2013, as Figure 3.3.11 shows. In 2020, the number of mentions of AI in earning calls was two times higher than mentions of big data, cloud, and machine learning combined, though that figure declined by 8.5% from 2019. The mentions of big data peaked in 2017 and have since declined by 57%.

MENTIONS of AI in CORPORATE EARNINGS CALLS, 2011-20

Source: Prattle & Liquidnet, 2020 | Chart: 2021 AI Index Report

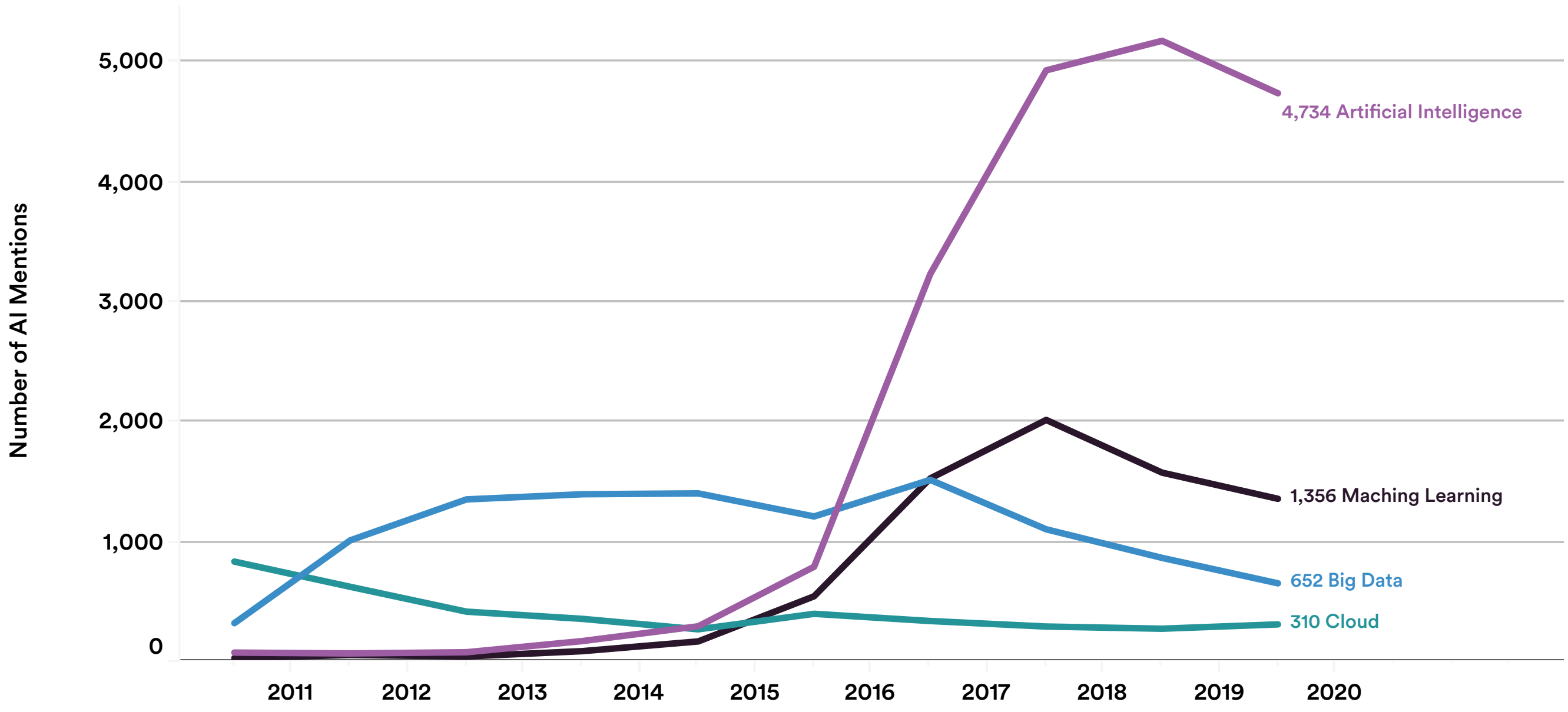

Figure 3.3.11

# CHAPTER 4:
# AI Education

Artificial Intelligence
Index Report 2021

## CHAPTER 4:
# Chapter Preview

ACCESS THE PUBLIC DATA

# Overview

As AI has become a more significant driver of economic activity, there has been increased interest from people who want to understand it and gain the necessary qualifications to work in the field. At the same time, rising AI demands from industry are tempting more professors to leave academia for the private sector. This chapter focuses on trends in the skills and training of AI talent through various education platforms and institutions.

What follows is an examination of data from an AI Index survey on the state of AI education in higher education institutions, along with a discussion on computer science (CS) undergraduate graduates and PhD graduates who specialized in AI-related disciplines, based on the annual Computing Research Association (CRA) Taulbee Survey. The final section explores trends in AI education in Europe, drawing on statistics from the Joint Research Centre (JRC) at the European Commission.

# CHAPTER HIGHLIGHTS

- An AI Index survey conducted in 2020 suggests that the world's top universities have increased their investment in AI education over the past four years. The number of courses that teach students the skills necessary to build or deploy a practical AI model on the undergraduate and graduate levels has increased by 102.9% and 41.7%, respectively, in the last four academic years.

- More AI PhD graduates in North America chose to work in industry in the past 10 years, while fewer opted for jobs in academia, according to an annual survey from the Computing Research Association (CRA). The share of new AI PhDs who chose industry jobs increased by 48% in the past decade, from 44.4% in 2010 to 65.7% in 2019. By contrast, the share of new AI PhDs entering academia dropped by 44%, from 42.1% in 2010 to 23.7% in 2019.

- In the last 10 years, AI-related PhDs have gone from 14.2% of the total of CS PhDs granted in the United States, to around 23% as of 2019, according to the CRA survey. At the same time, other previously popular CS PhDs have declined in popularity, including networking, software engineering, and programming languages. Compilers all saw a reduction in PhDs granted relative to 2010, while AI and Robotics/Vision specializations saw a substantial increase.

- After a two-year increase, the number of AI faculty departures from universities to industry jobs in North America dropped from 42 in 2018 to 33 in 2019 (28 of these are tenured faculty and five are untenured). Carnegie Mellon University had the largest number of AI faculty departures between 2004 and 2019 (16), followed by the Georgia Institute of Technology (14) and University of Washington (12).

- The percentage of international students among new AI PhDs in North America continued to rise in 2019, to 64.3%—a 4.3% increase from 2018. Among foreign graduates, 81.8% stayed in the United States and 8.6% have taken jobs outside the United States.

- In the European Union, the vast majority of specialized AI academic offerings are taught at the master's level; robotics and automation is by far the most frequently taught course in the specialized bachelor's and master's programs, while machine learning (ML) dominates in the specialized short courses.

# 4.1 STATE OF AI EDUCATION IN HIGHER EDUCATION INSTITUTIONS

In 2020, AI Index developed a survey that asked computer science departments or schools of computing and informatics at top-ranking universities around the world and in emerging economies about four aspects of their AI education: undergraduate program offerings, graduate program offerings, offerings on AI ethics, and faculty expertise and diversity. The survey was completed by 18 universities from nine countries.[1] Results from the AI Index survey indicate that universities have increased both the number of AI courses they offer that teach students how to build and deploy a practical AI model and the number of AI-focused faculty.

## UNDERGRADUATE AI COURSE OFFERINGS

Course offerings at the undergraduate level were examined by evaluating trends in courses that teach students the skills necessary to build or deploy a practical AI model, intro-level AI and ML courses, and enrollment statistics.

### Undergraduate Courses That Teach AI Skills

The survey results suggest that CS departments have invested heavily in practical AI courses in the past four academic years (AY).[2] The number of courses on offer that teach students the skills necessary to build or deploy a practical AI model has increased by 102.9%, from 102 in AY 2016–17 to 207 in AY 2019–20, across 18 universities (Figure 4.1.1).

NUMBER of UNDERGRADUATE COURSES THAT TEACH STUDENTS the SKILLS NECESSARY to BUILD or DEPLOY a PRACTICAL AI MODEL, AY 2016-20
Source: AI Index, 2020 | Chart: 2021 AI Index Report

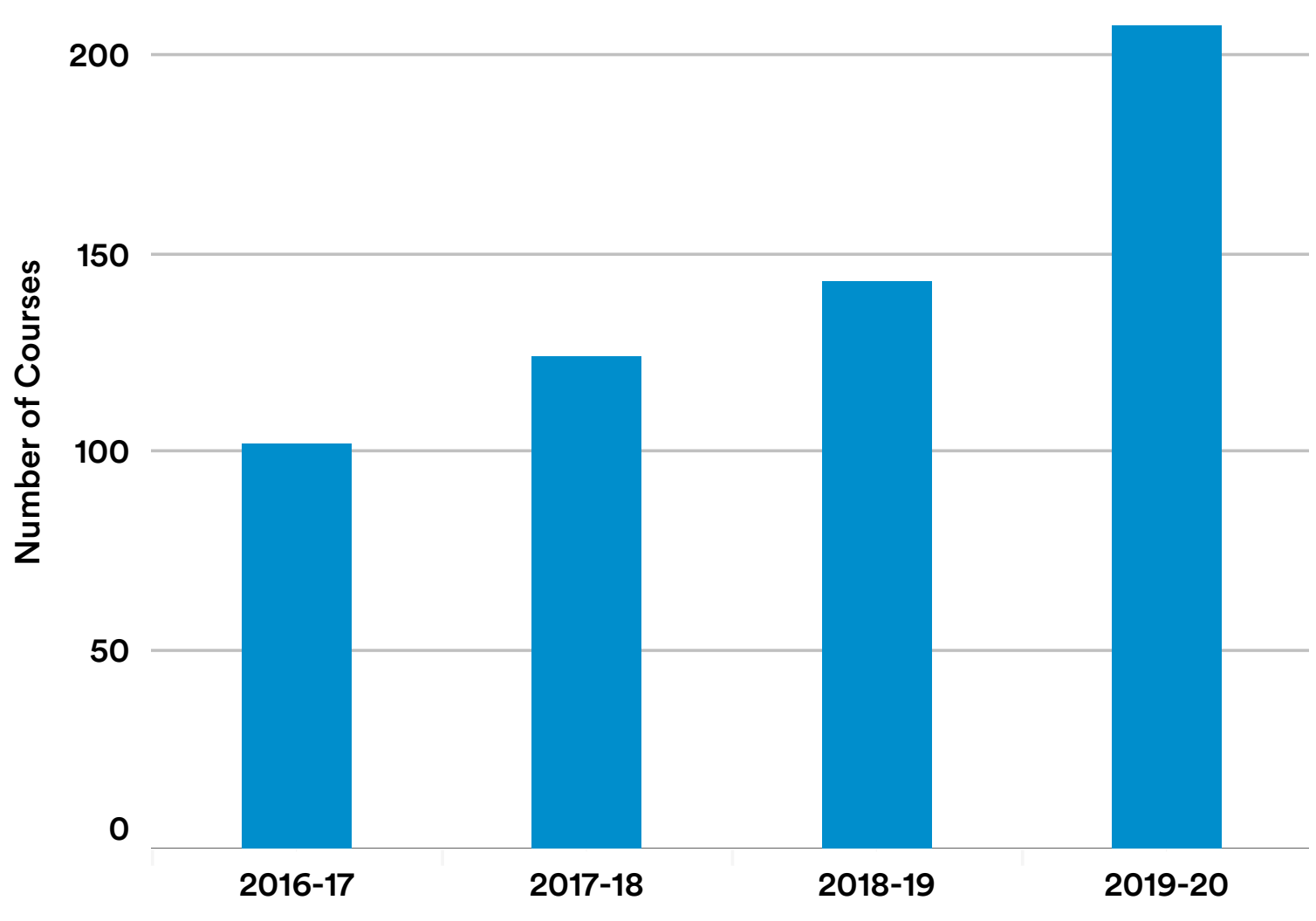

Figure 4.1.1

### Intro-Level AI and ML Courses

The data shows that the number of students who enrolled in or attempted to enroll in an Introduction to Artificial Intelligence course and Introduction to Machine Learning course has jumped by almost 60% in the past four academic years (Figure 4.1.2).[3]

The slight drop in enrollment in the intro-level AI and ML courses in AY 2019–20 is mostly driven by the decrease in the number of course offerings at U.S. universities. Intro-level course enrollment

1 The survey was distributed to 73 universities online over three waves from November 2020 to January 2021 and completed by 18 universities, a 24.7% response rate. The 18 universities are—Belgium: Katholieke Universiteit Leuven; Canada: McGill University; China: Shanghai Jiao Tong University, Tsinghua University; Germany: Ludwig Maximilian University of Munich, Technical University of Munich; Russia: Higher School of Economics, Moscow Institute of Physics and Technology; Switzerland: École Polytechnique Fédérale de Lausanne; United Kingdom: University of Cambridge; United States: California Institute of Technology, Carnegie Mellon University (Department of Machine Learning), Columbia University, Harvard University, Stanford University, University of Wisconsin–Madison, University of Texas at Austin, Yale University.
2 See here for a list of keywords on practical artificial intelligence models provided to the survey respondents. A course is defined as a set of classes that require a minimum of 2.5 class hours (including lecture, lab, TA hours, etc.) per week for at least 10 weeks in total. Multiple courses with the same titles and numbers count as one course.
3 For universities that have a cap on course registration, the number of students who attempted to enroll in the intro-level AI and ML courses are included.

in the European Union has gradually increased by 165% in the past four academic years, while such enrollment in the United States has seen a clear dip in growth in the last academic year (Figure 4.1.3). Six of the eight U.S. universities surveyed say that the number of (attempted) enrollments for the introductory AI and ML courses has decreased within the last year. Some universities cited students taking leaves during the pandemic as the main cause of the drop; others mentioned structural changes in intro-level AI course offerings—such as creating Intro to Data Science last year—that may have driven students away from traditional intro to AI and ML courses.

**NUMBER of STUDENTS WHO ENROLLED or ATTEMPTED to ENROLL in INTRO to AI and INTRO to ML COURSES, AY 2016-20**
Source: AI Index, 2020 | Chart: 2021 AI Index Report

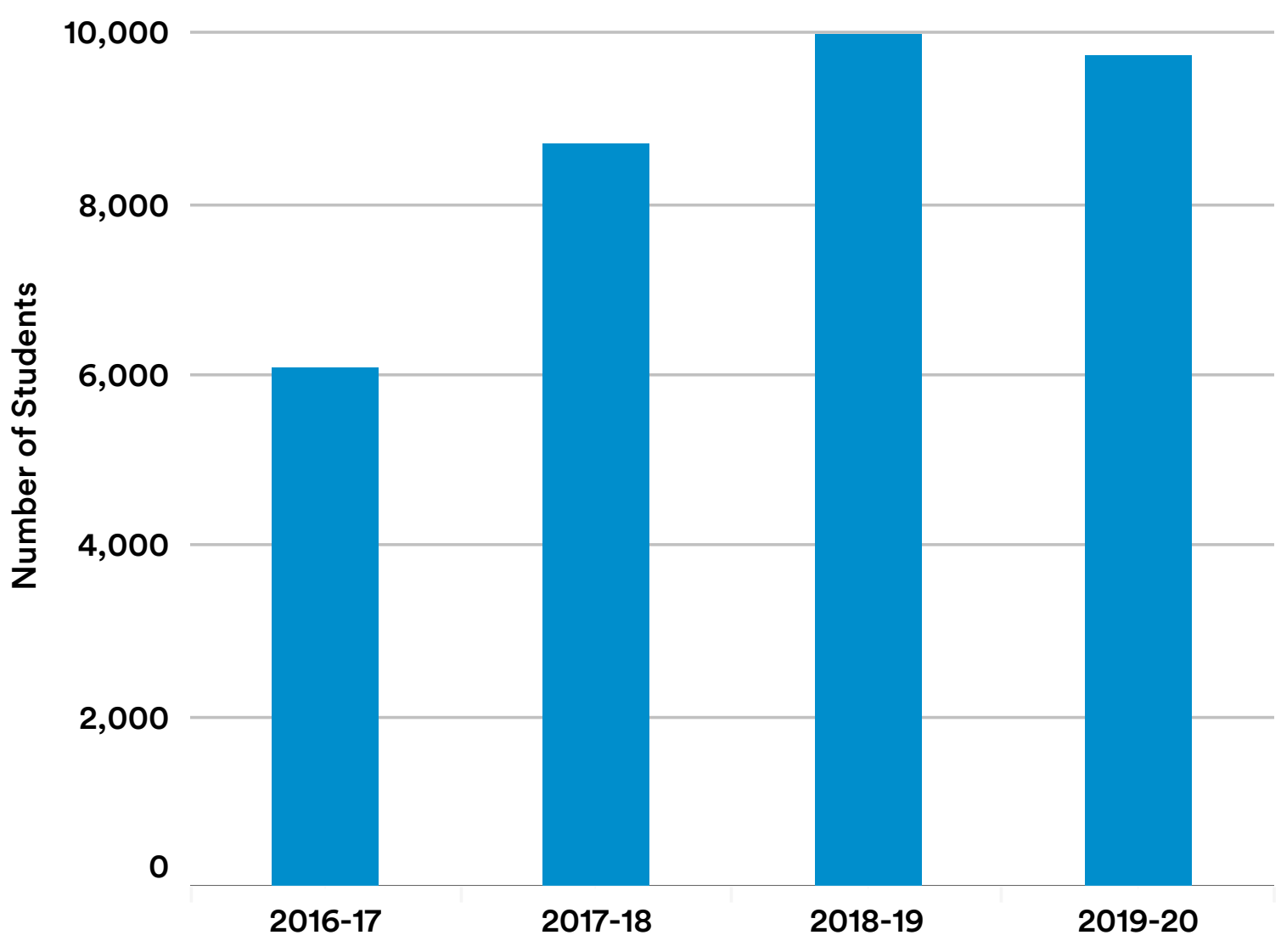

Figure 4.1.2

**PERCENTAGE CHANGE in the NUMBER of STUDENTS WHO ENROLLED or ATTEMPTED to ENROLL in INTRO to AI and INTRO to ML COURSES by GEOGRAPHIC AREA, AY 2016-20**
Source: AI Index, 2020 | Chart: 2021 AI Index Report

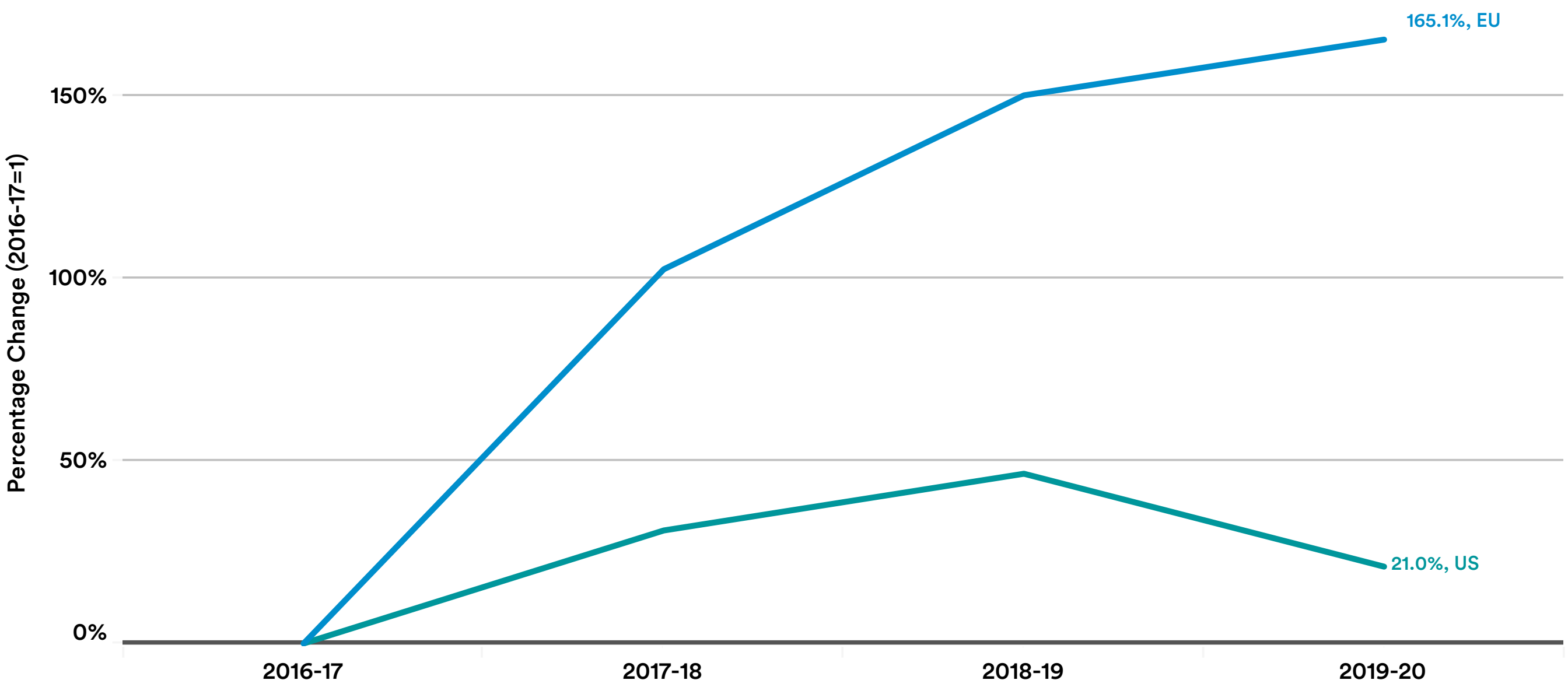

Figure 4.1.3

## GRADUATE AI COURSE OFFERINGS

The survey also looks at course offerings at the graduate or advanced degree level, specifically at graduate courses that teach students the skills necessary to build or deploy a practical AI model.[4]

### Graduate Courses That Focus on AI Skills

Graduate offerings that teach students the skills required to build or deploy a practical AI model increased by 41.7% in the last four academic years, from 151 courses in AY 2016–17 to 214 in AY 2019–20 (Figure 4.1.4).

## FACULTY WHO FOCUS ON AI RESEARCH

As shown in Figure 4.1.5, the number of tenure-track faculty with a primary research focus on AI at the surveyed universities grew significantly over the past four academic years, in keeping with the rising demand for AI classes and degree programs. The number of AI-focused faculty grew by 59.1%, from 105 in AY 2016–17 to 167 in AY 2019–20.

NUMBER of GRADUATE COURSES THAT TEACH STUDENTS the SKILLS NECESSARY to BUILD or DEPLOY a PRACTICAL AI MODEL, AY 2016-20
Source: AI Index, 2020 | Chart: 2021 AI Index Report

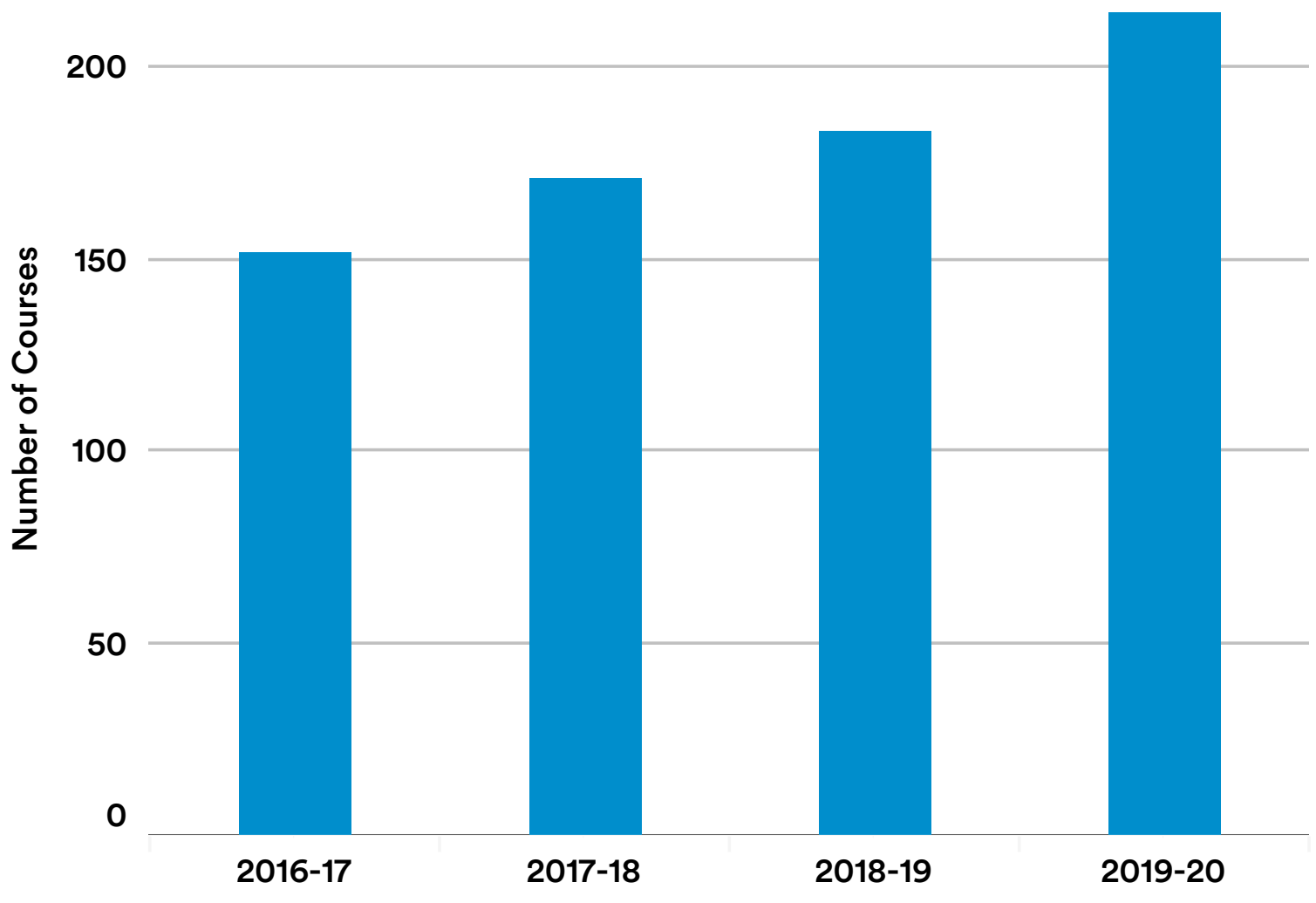

Figure 4.1.4

NUMBER of TENURE-TRACK FACULTY WHO PRIMARILY FOCUS THEIR RESEARCH on AI, AY 2016-20
Source: AI Index, 2020 | Chart: 2021 AI Index Report

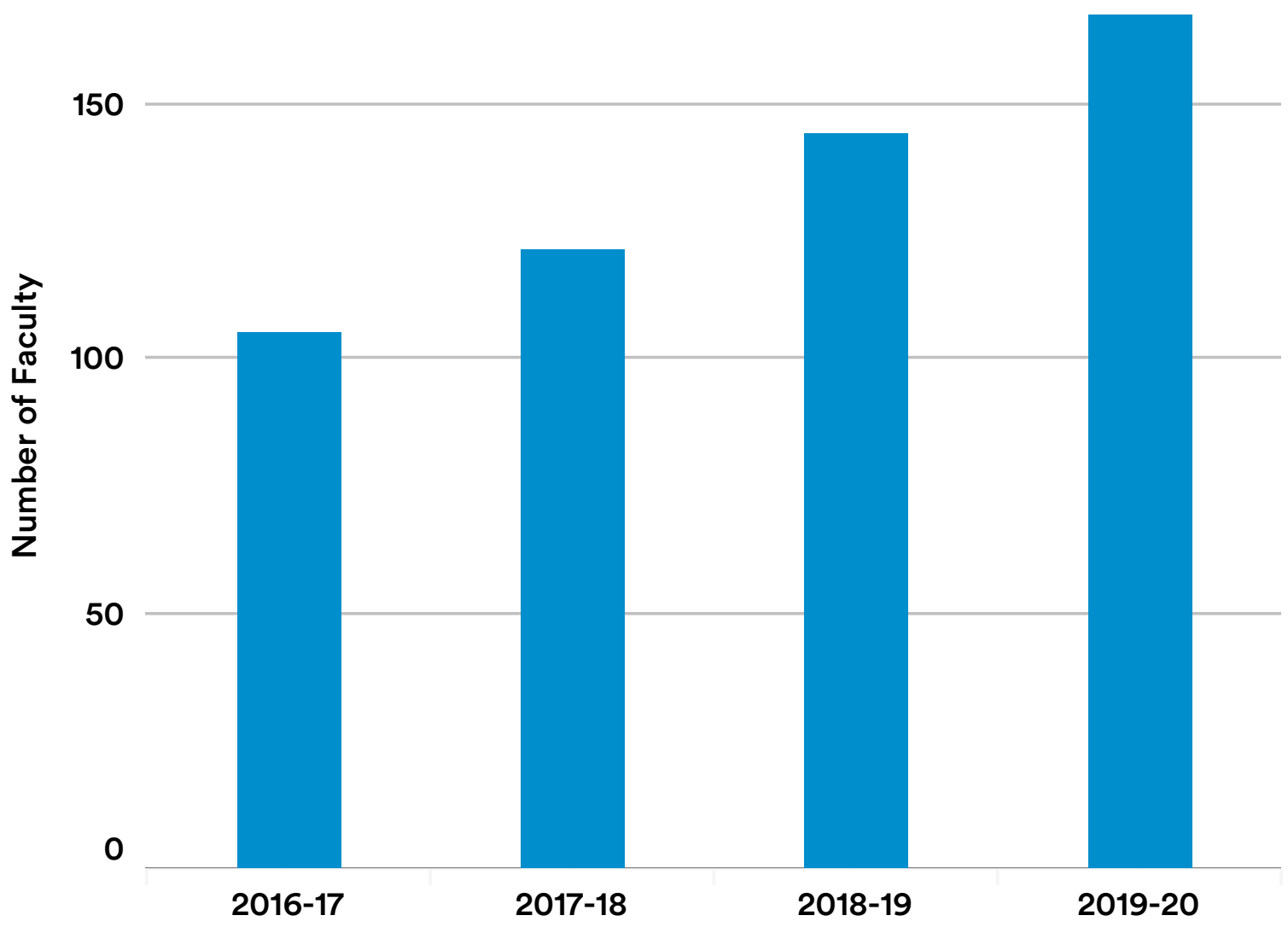

Figure 4.1.5

4 See here for a list of keywords on practical artificial intelligence models provided to the survey respondents. A course is defined as a set of classes that require a minimum of 2.5 class hours (including lecture, lab, TA hours, etc.) per week for at least 10 weeks in total. Multiple courses with the same titles and numbers count as one course.

This section presents findings from the annual Taulbee Survey from the Computing Research Association (CRA). The annual CRA survey documents trends in student enrollment, degree production, employment of graduates, and faculty salaries in academic units in the United States and Canada that grant doctoral degrees in computer science (CS), computer engineering (CE), or information (I). Academic units include departments of computer science and computer engineering or, in some cases, colleges or schools of information or computing.

# 4.2 AI AND CS DEGREE GRADUATES IN NORTH AMERICA

## CS UNDERGRADUATE GRADUATES IN NORTH AMERICA

Most AI-related courses in North America are a part of the CS course offerings at the undergraduate level. The number of new CS undergraduate graduates at doctoral institutions in North America has grown steadily in the last 10 years (Figure 4.2.1). More than 28,000 undergraduates completed CS degrees in 2019, around three times higher than the number in 2010.

## NEW CS PHDS IN THE UNITED STATES

The section examines the trend of CS PhD graduates in the United States with a focus on those with AI-related specialties.[5] The CRA survey includes 20 specialties in total, two of which are directly related to the field of AI, including “artificial intelligence/machine learning” and “robotics/vision.”

NUMBER of NEW CS UNDERGRADUATE GRADUATES at DOCTORAL INSTITUTIONS in NORTH AMERICA, 2010-19

Source: CRA Taulbee Survey, 2020 | Chart: 2021 AI Index Report

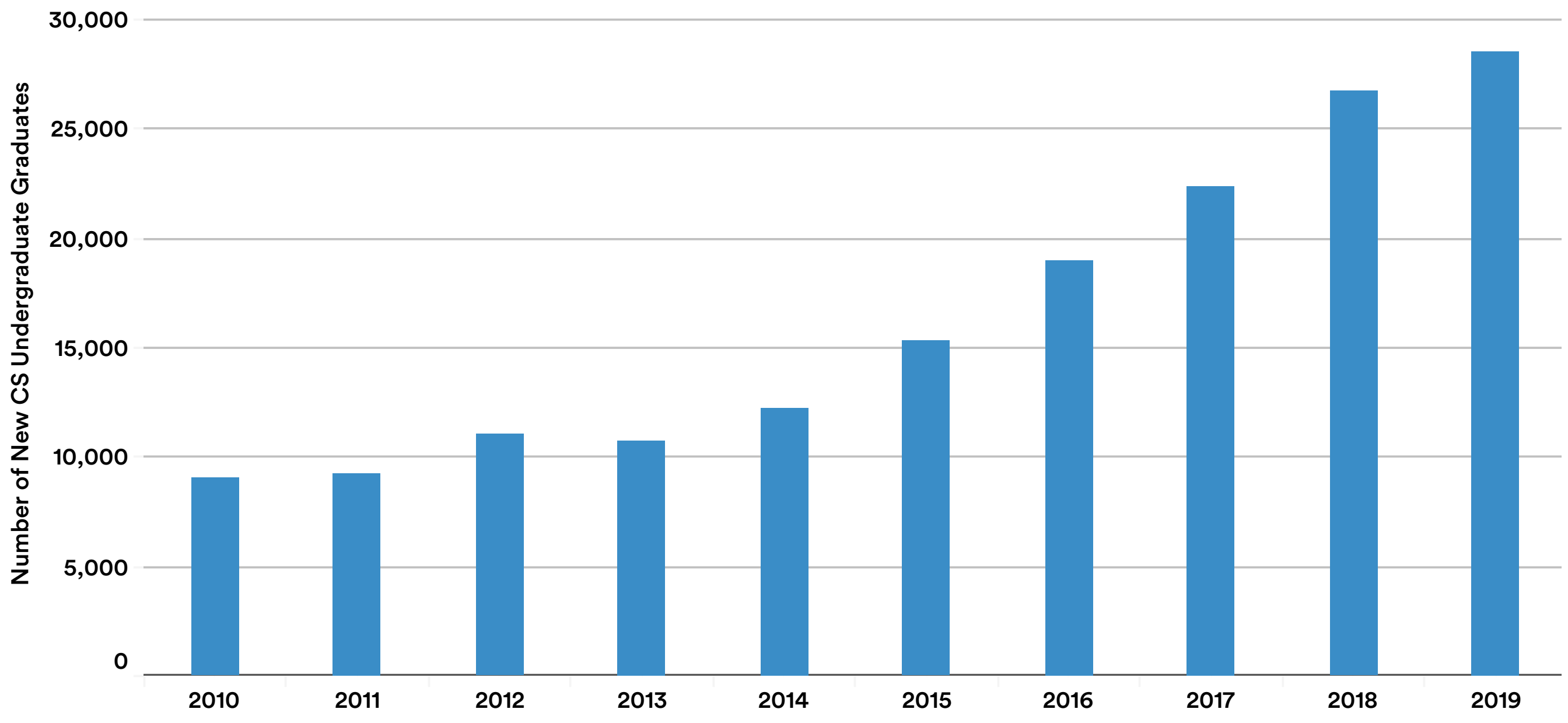

Figure 4.2.1

5 New CS PhDs in this section include PhD graduates from academic units (departments, colleges, or schools within universities) of computer science in the United States.

## NEW CS PHDS BY SPECIALTY

Among all computer science PhD graduates in 2019, those who specialized in artificial intelligence/machine learning (22.8%), theory and algorithms (8.0%), and robotics/vision (7.3%) top the list (Figure 4.2.2). The AI/ML specialty has been the most popular in the past decade, and the number of AI/ML graduates in 2019 is higher than the number of the next five specialties combined. Moreover, robotics/vision jumped from the eighth most popular specialization in 2018 to the third in 2019.

Over the past 10 years, AI/ML and robotics/vision are the CS PhD specializations that exhibit the most significant growth, relative to 18 other specializations (Figure 4.2.3). The percentage of AI/ML-specialized CS PhD graduates among all new CS PhDs in 2020 is 8.6 percentage points (pp) larger than in 2010, followed by robotics/vision-specialized doctorates at 2.4 pp. By contrast, the share of CS PhDs specializing in networks (-4.8 pp), software engineering (-3.6 pp), and programming languages/compilers (-3.0 pp) experienced negative growth in 2020.

NEW CS PHDS (% of TOTAL) in the UNITED STATES by SPECIALITY, 2019
Source: CRA Taulbee Survey, 2020 | Chart: 2021 AI Index Report

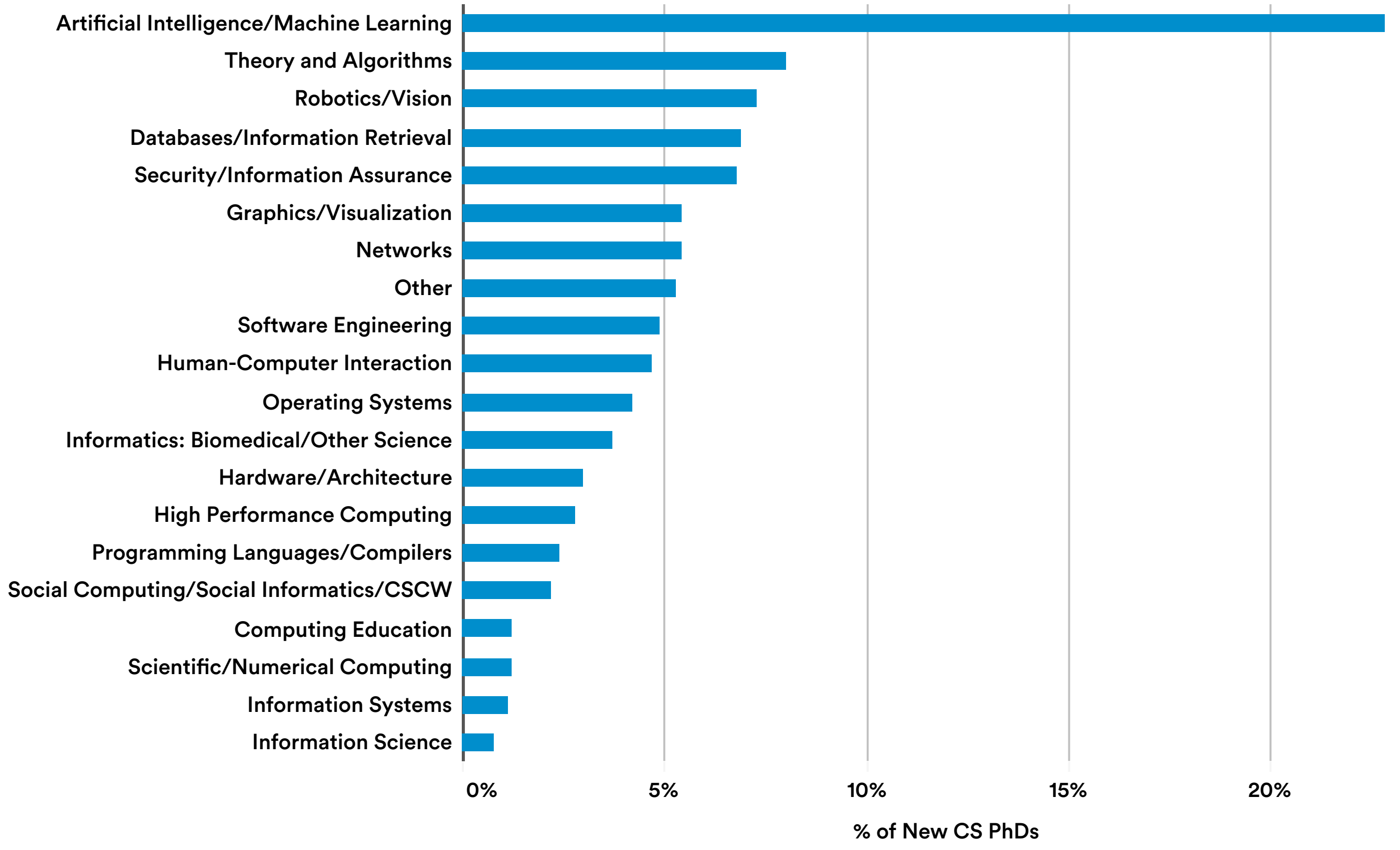

Figure 4.2.2

## PERCENTAGE POINT CHANGE in NEW CS PHDS in the UNITED STATES from 2010 to 2019 by SPECIALTY

Source: CRA Taulbee Survey, 2020 | Chart: 2021 AI Index Report

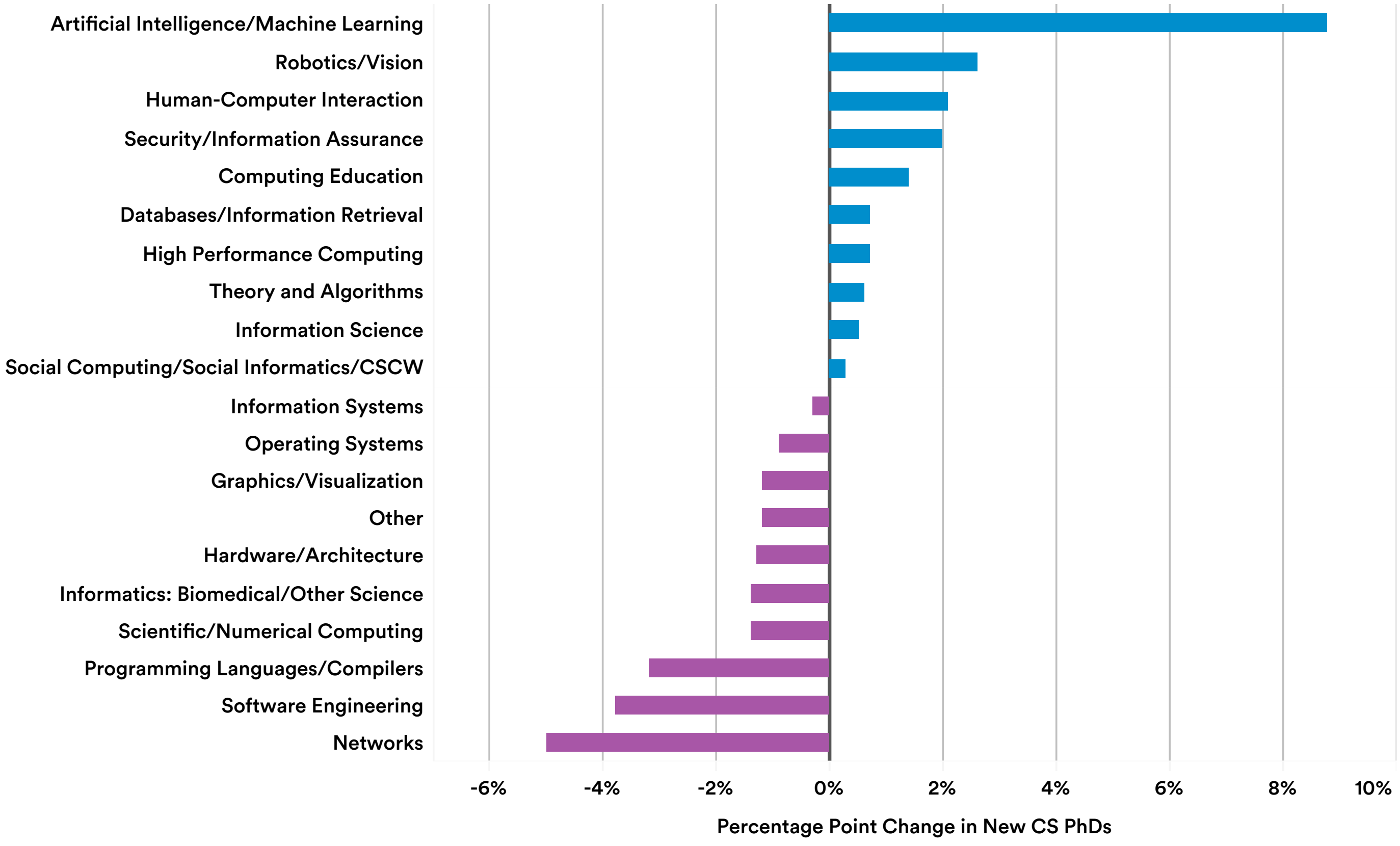

Figure 4.2.3

## NEW CS PHDS WITH AI/ML AND ROBOTICS/VISION SPECIALTIES

Figure 4.2.4a and Figure 4.2.4b take a closer look at the number of recent AI PhDs specializing in AI/ML or robotics/vision in the United States. Between 2010 and 2019, the number of AI/ML-focused graduates grew by 77%, while the percentage of these new PhDs among all CS PhD graduates increased by 61%. The number of both AI/ML and robotics/vision PhD graduates reached an all-time high in 2019.

### NEW CS PHDS with AI/ML and ROBOTICS/VISION SPECIALTY in the UNITED STATES, 2010-19

Source: CRA Taulbee Survey, 2020 | Chart: 2021 AI Index Report

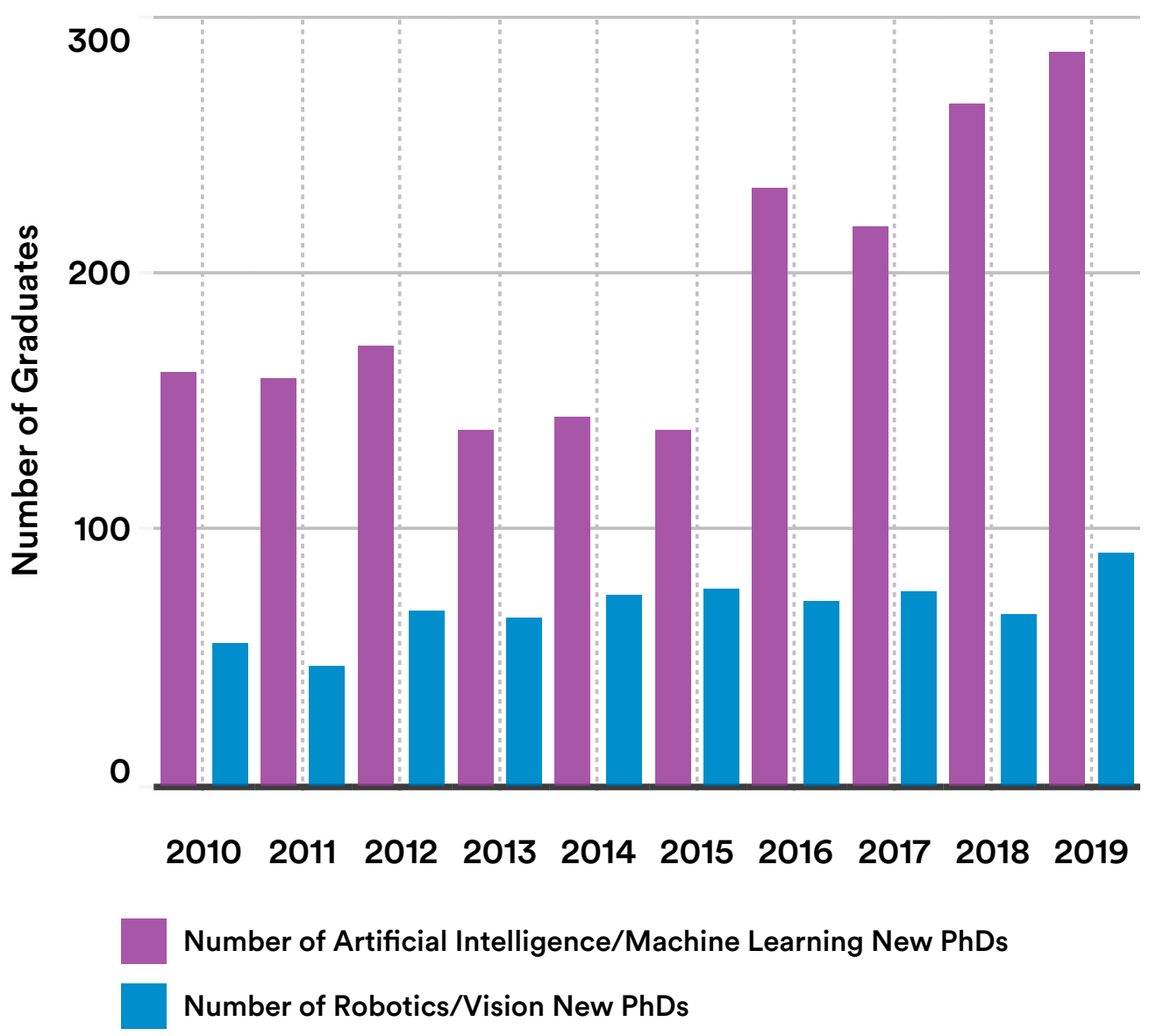

Figure 4.2.4a

### NEW CS PHDS (% of TOTAL) with AI/ML and ROBOTICS/VISION SPECIALTY in the UNITED STATES, 2010-19

Source: CRA Taulbee Survey, 2020 | Chart: 2021 AI Index Report

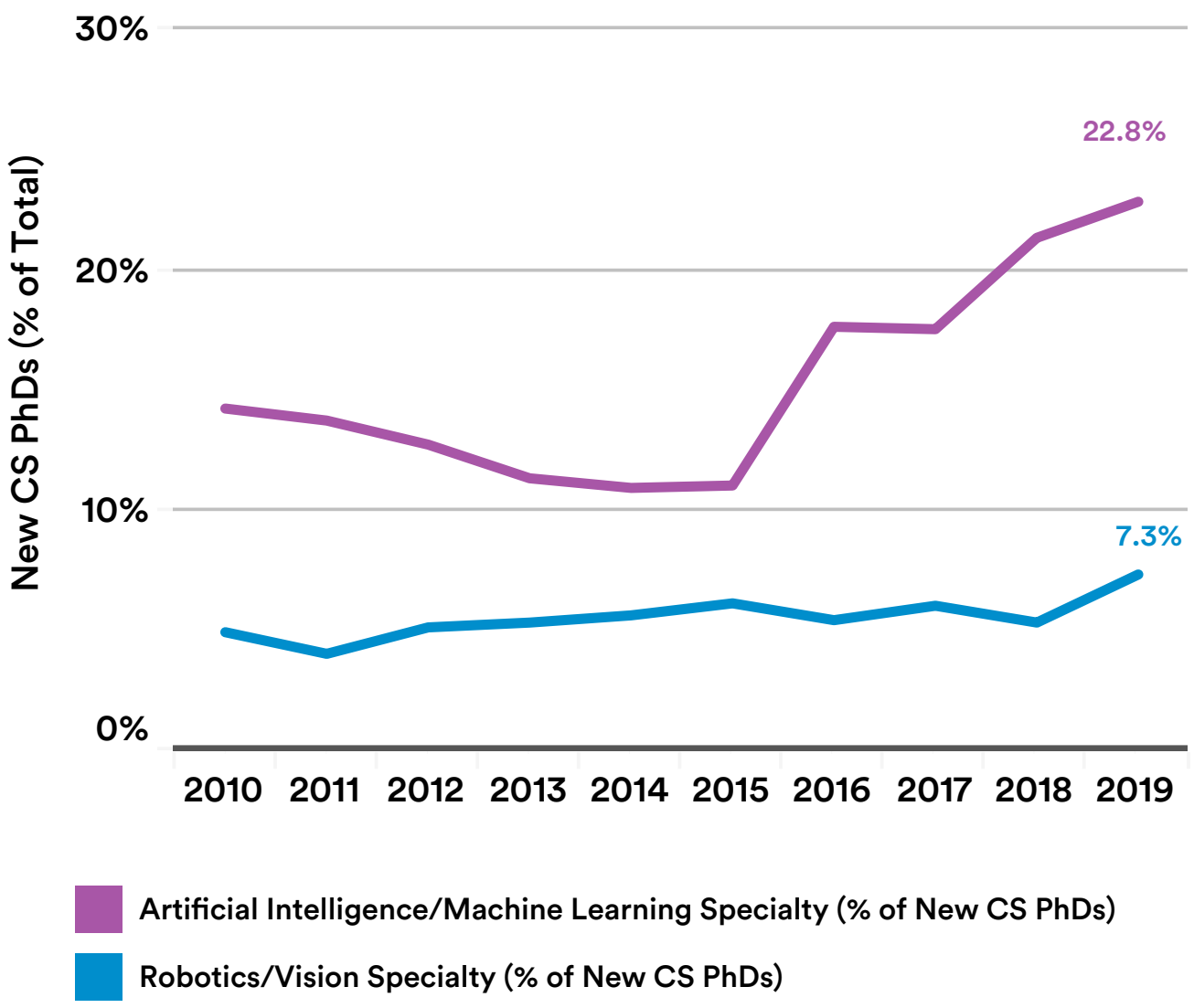

Figure 4.2.4b

## NEW AI PHDS EMPLOYMENT IN NORTH AMERICA

Where do new AI PhD graduates choose to work? This section captures the employment trends of new AI PhDs in academia and industry across North America.[6]

### Industry vs. Academia

In the past 10 years, the number of new AI PhD graduates in North America who chose industry jobs continues to grow, as its share increased by 48%, from 44.4% in 2010 to 65.7% in 2019 (Figure 4.2.5a and Figure 4.2.5b). By contrast, the share of new AI PhDs entering academia dropped by 44%, from 42.1% in 2010 to 23.7% in 2019. As is clear from Figure 4.2.5b, these changes are largely a reflection of the fact that the number of PhD graduates entering academia has remained roughly level through the decade, while the large increase in PhD output is primarily being absorbed by the industry.

EMPLOYMENT of NEW AI PHDS to ACADEMIA or INDUSTRY in NORTH AMERICA, 2010-19

Source: CRA Taulbee Survey, 2020 | Chart: 2021 AI Index Report

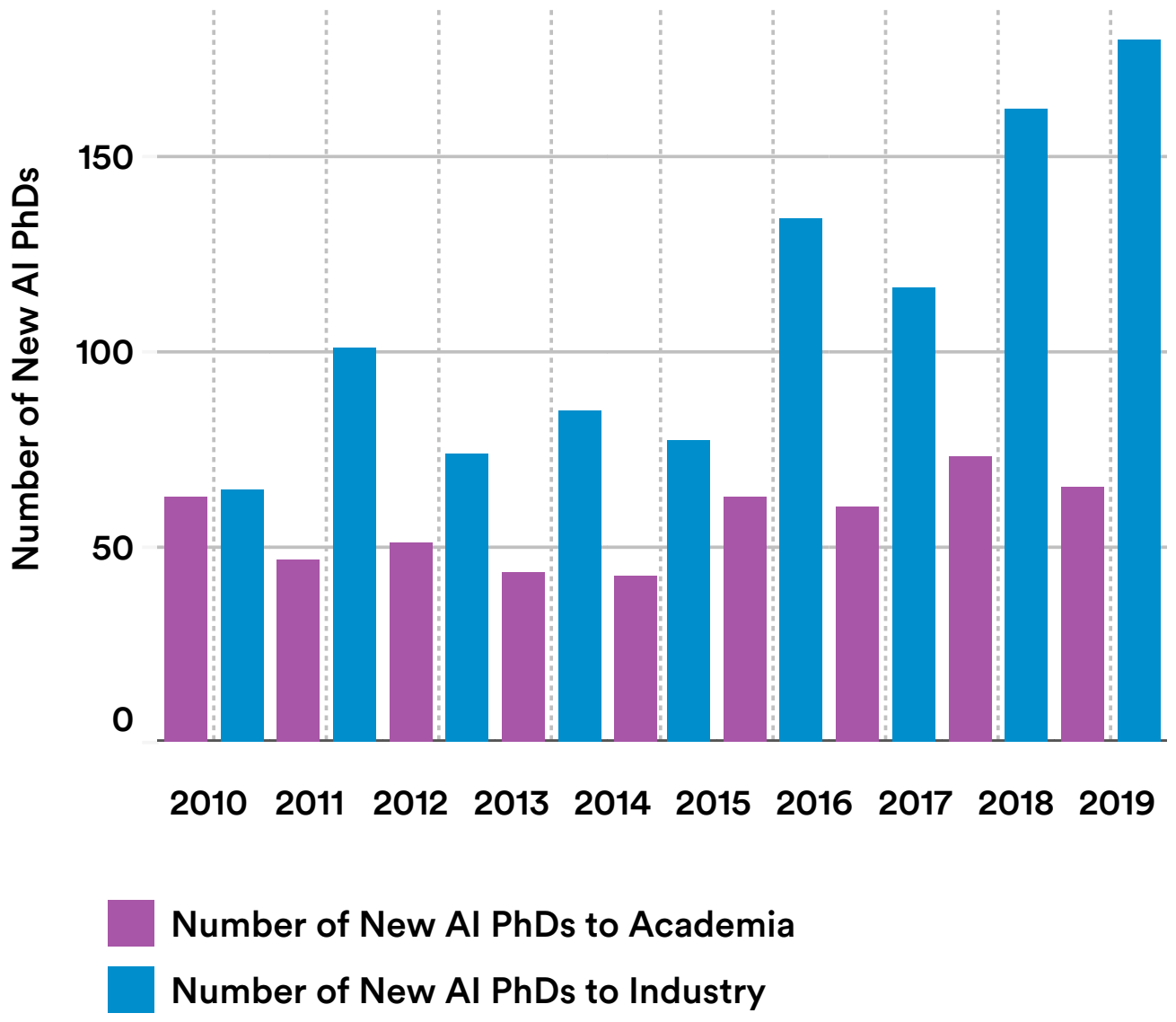

Figure 4.2.5a

EMPLOYMENT of NEW AI PHDS (% of TOTAL) to ACADEMIA or INDUSTRY in NORTH AMERICA, 2010-19

Source: CRA Taulbee Survey, 2020 | Chart: 2021 AI Index Report

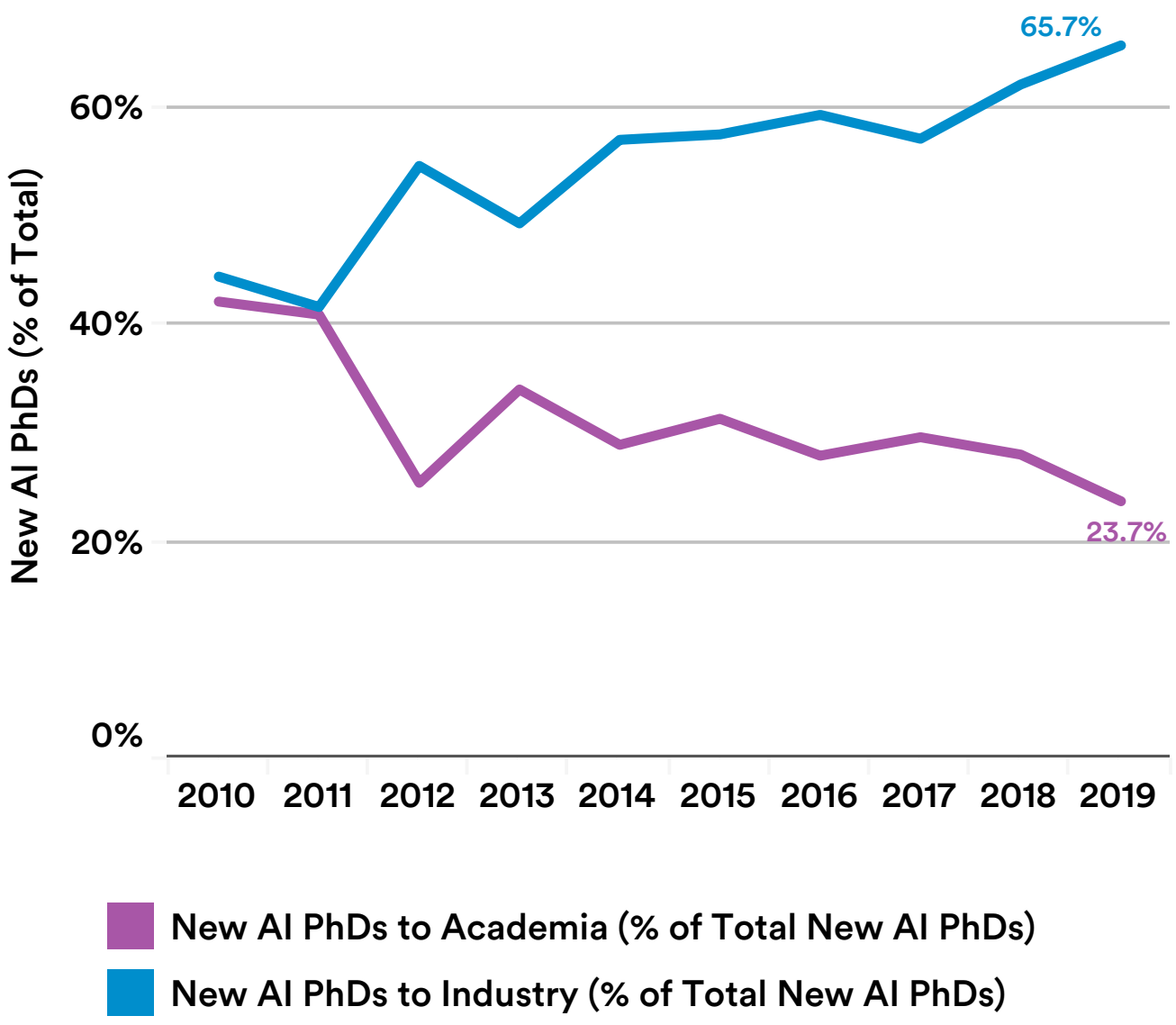

Figure 4.2.5b

6 New AI PhDs in this section include PhD graduates who specialize in artificial intelligence from academic units (departments, colleges, or schools within universities) of computer science, computer engineering, and information in the United States and Canada.

NEW INTERNATIONAL AI PHDS (% of TOTAL NEW AI PHDS) in NORTH AMERICA, 2010-19
Source: CRA Taulbee Survey, 2020 | Chart: 2021 AI Index Report

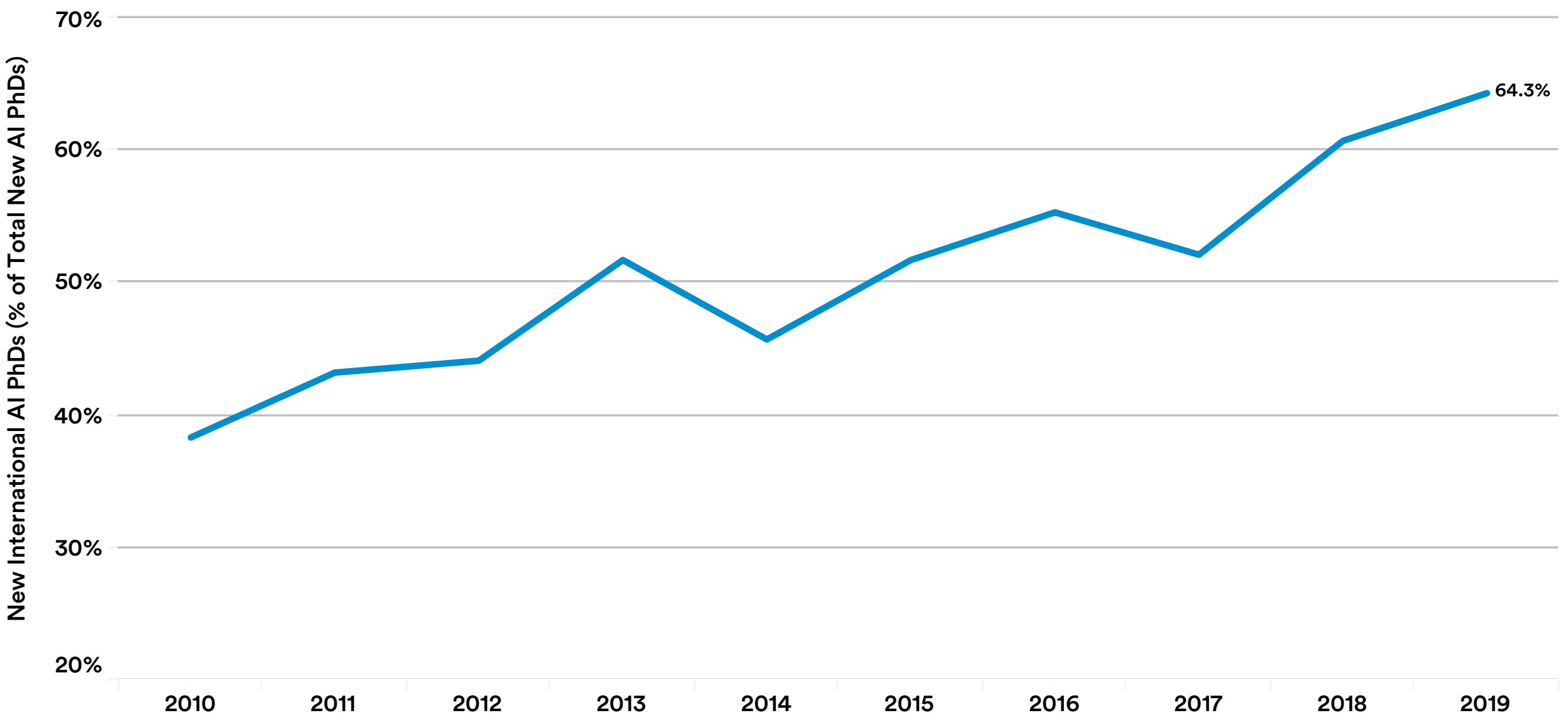

Figure 4.2.6

## NEW INTERNATIONAL AI PHDS

The percentage of international students among new AI PhD graduates in North America continued to rise in 2019, to 64.3%—a 4.3 percentage point increase from 2018 (Figure 4.2.6). For comparison, of all PhDs with a known specialty area, 63.4% of computer engineering, 59.6% of computer science, and 29.5% of information recipients are international students in 2019.

Moreover, among foreign AI PhD graduates in 2019 in the United States specifically, 81.8% stayed in the United States for employment and 8.6% have taken jobs outside the United States (Figure 4.2.7). In comparison, among all international student graduates with known specialties, 77.9% have stayed in the United States while 10.4% were employed elsewhere.

INTERNATIONAL NEW AI PHDS (% of TOTAL) in the UNITED STATES by LOCATION OF EMPLOYMENT, 2019
Source: CRA Taulbee Survey, 2020 | Chart: 2021 AI Index Report

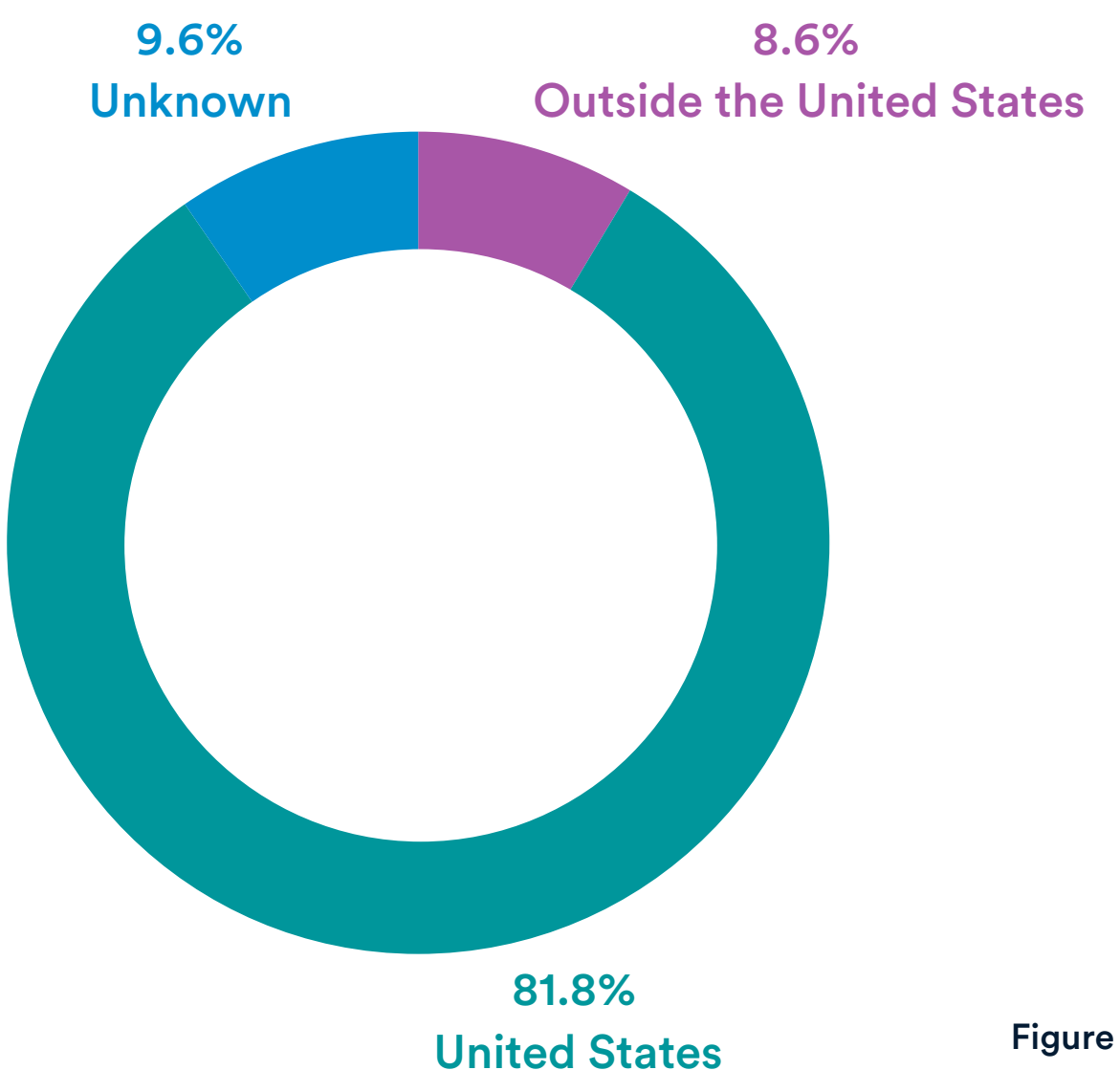

Figure 4.2.7

This section presents research from the Joint Research Center at the European Commission that assessed the academic offerings of advanced digital skills in 27 European Union member states as well as six other countries: the United Kingdom, Norway, Switzerland, Canada, the United States, and Australia. This was the second such study,[7] and the 2020 version addressed four technological domains: artificial intelligence (AI), high performance computing (HPC), cybersecurity (CS), and data science (DS), applying text-mining and machine-learning techniques to extract content related to study programs addressing the specific domains. See the reports "Academic Offer of Advanced Digital Skills in 2019–20. International Comparison" and "Estimation of Supply and Demand of Tertiary Education Places in Advanced Digital Profiles in the EU," for more detail.

# 4.3 AI EDUCATION IN THE EUROPEAN UNION AND BEYOND

## AI OFFERINGS IN EU27

The study revealed a total number of 1,032 AI programs across program scopes and program levels in the 27 EU countries (Figure 4.3.1). The overwhelming majority of specialized AI academic offerings in the EU are taught at the master's level, which leads to a degree that equips students with strong competencies for the workforce. Germany leads the other member nations in offering the most specialized AI programs, followed by the Netherlands, France, and Sweden. France tops the list in offering the most AI programs at the master's level.

NUMBER of SPECIALIZED AI PROGRAMS in EU27, 2019-20
Source: Joint Research Centre, European Commission, 2020 | Chart: 2021 AI Index Report

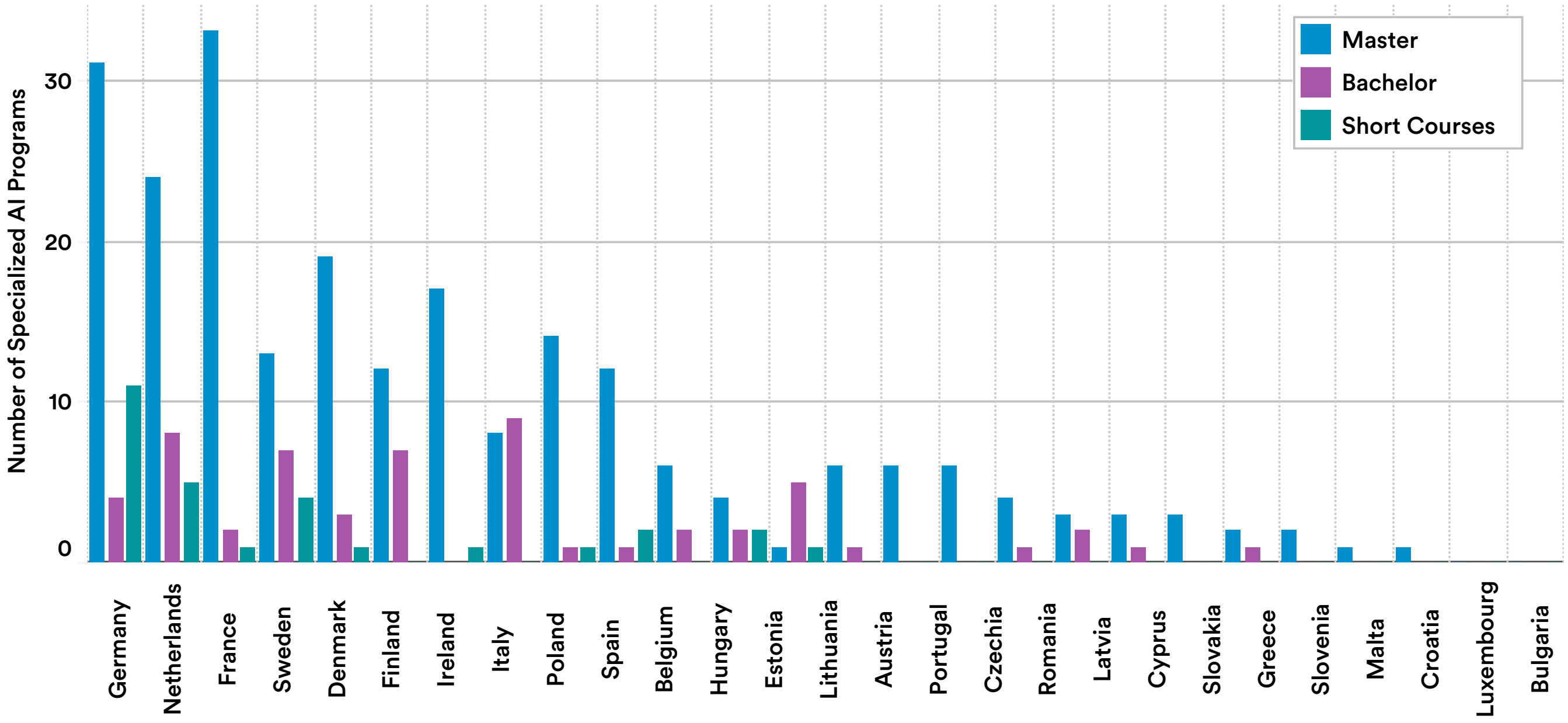

Figure 4.3.1

7 Note that the 2020 report introduced methodological improvements from the 2019 version; therefore, a strict comparison is not possible. Improvements include the removal of certain keywords and the addition of others to identify the programs. Still, more than 90% of all detected programs in the 2020 edition are triggered by keywords present in the 2019 study.

## By Content Taught in AI-Related Courses

What types of AI technologies are the most popular among the course offerings in three levels of specialized AI programs in the European Union? Data suggests that robotics and automation are by far the most frequently taught courses in the specialized bachelor's and master's programs, while machine learning dominates in the specialized short courses (Figure 4.3.2). As short courses cater to working professionals, the trend shows that machine learning has become one of the key competencies in the professional development and implementation of AI.

It is also important to mention the role of AI ethics and AI applications, as both content areas claim a significant share of the education offerings among the three program levels. AI ethics—including courses on security, safety, accountability, and explainability—accounts for 14% of the curriculum on average, while AI applications—such as courses on big data, the internet of things, and virtual reality—take a similar share on average.

SPECIALIZED AI PROGRAMS (% of TOTAL) by CONTENT AREA in EU27, 2019-20
Source: Joint Research Centre, European Commission, 2020 | Chart: 2021 AI Index Report

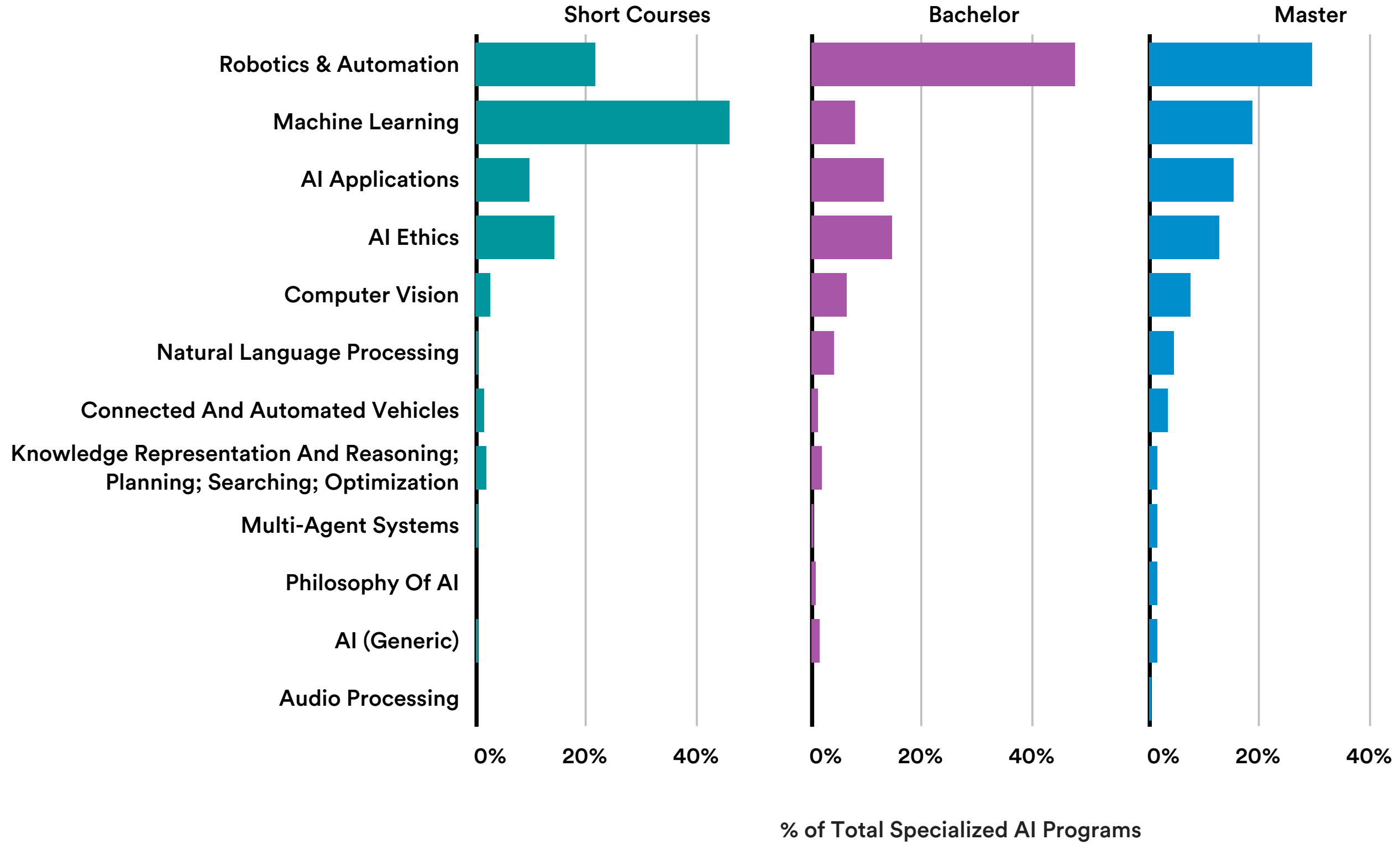

Figure 4.3.2

## INTERNATIONAL COMPARISON

The JRC report compared AI education in the 27 EU member states with other countries in Europe, including Norway, Switzerland, and the United Kingdom, as well as Canada, the United States, and Australia. Figure 4.3.3 shows the total number of 1,680 specialized AI programs in all countries considered in the 2019–20 academic year. The United States appears to have offered more programs specialized in AI than any other geographic area, although EU27 comes in a close second in terms of the number of AI-specialized master's programs.

NUMBER of SPECIALIZED AI PROGRAMS by GEOGRAPHIC AREA and LEVEL, 2019-20

Source: Joint Research Centre, European Commission, 2020 | Chart: 2021 AI Index Report

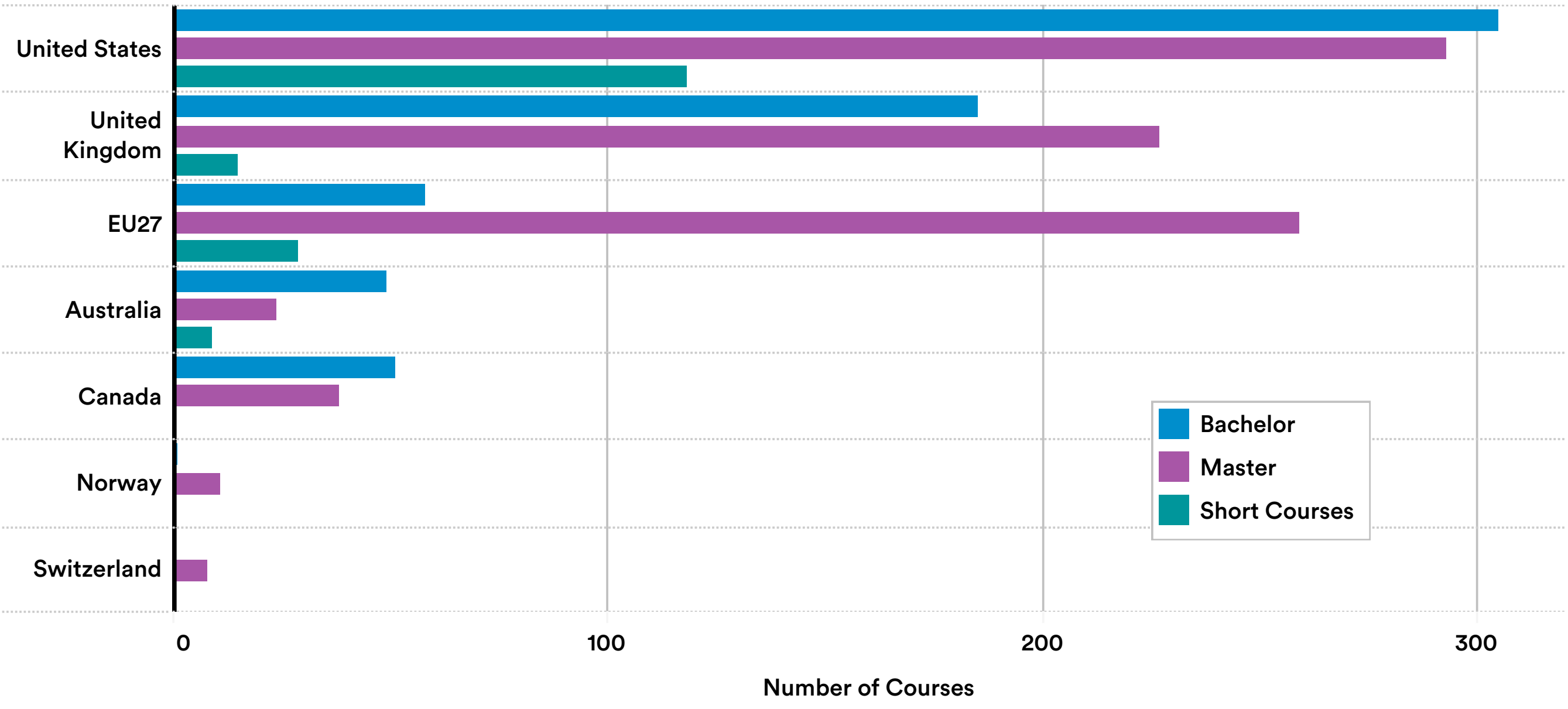

Figure 4.3.3

# AI Brain Drain and Faculty Departure

Michael Gofman and Zhao Jin, researchers from the University of Rochester and Cheung Kong Graduate School of Business, respectively, published a paper titled "Artificial Intelligence, Education, and Entrepreneurship" in 2019 that explores the relationship between domain-specific knowledge of university students and their ability to establish startups and attract funding.[8] For the source of variation in students' AI-specific knowledge, the co-authors used the departure of AI professors—what they referred to as "an unprecedented brain drain"—from universities to industry between 2004 and 2018. They relied on data hand-collected from LinkedIn as well as authors' affiliation from the Scopus database of academic publications and conferences to complement the results from the LinkedIn search.

The paper found that AI faculty departures have a negative effect on AI startups founded by students who graduate from universities where those professors used to work, with the researchers pointing to a chilling effect on future AI entrepreneurs in the years following the faculty departures. PhD students are the most affected, compared with undergraduate

NUMBER of AI FACULTY DEPARTURES in NORTH AMERICA, 2004-19

Source: Gofman and Jin, 2020 | Chart: 2021 AI Index Report

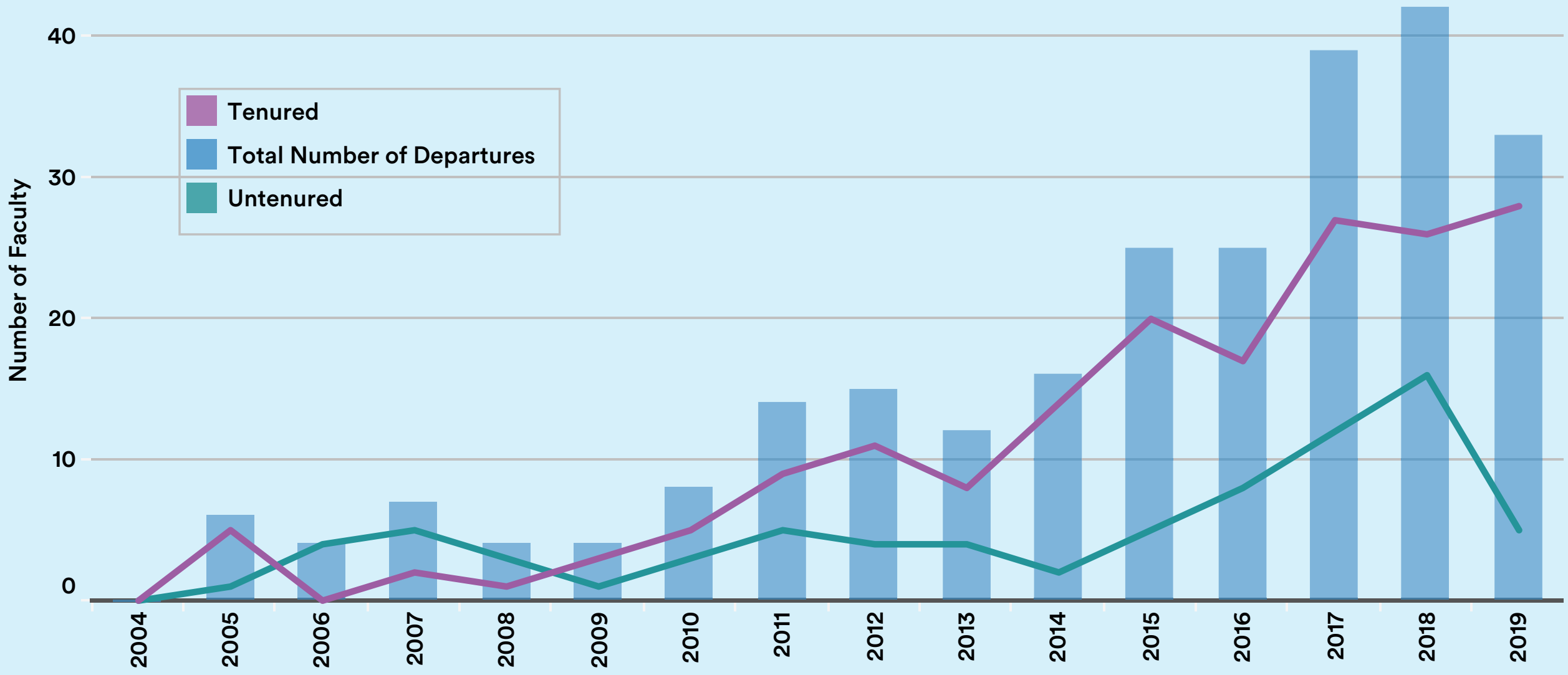

Figure 4.4.1

8 See AI Brain Drain Index for more details.

# AI Brain Drain and Faculty Departure (continued)

and master's students, and the negative impact intensifies when the AI professors who leave are replaced by faculty from lower-ranked schools or untenured AI professors.

With the updated data of 2019 from Gofman and Jin, Figure 4.4.1 shows that after a two-year increase, the total number of AI faculty departures from universities in North America to industry dropped from 42 in 2018 to 33 in 2019 (28 of these are tenured faculty and 5 are untenured). Between 2004 and 2019, Carnegie Mellon University had the largest number of AI faculty departures in 2019 (16), followed by the Georgia Institute of Technology (14) and University of Washington (12), as shown in Figure 4.4.2.

NUMBER of AI FACULTY DEPATURES in NORTH AMERICA (with UNIVERSITY AFFILIATION) by UNIVERSITY, 2004-18
Source: Gofman and Jin, 2020 | Chart: 2021 AI Index Report

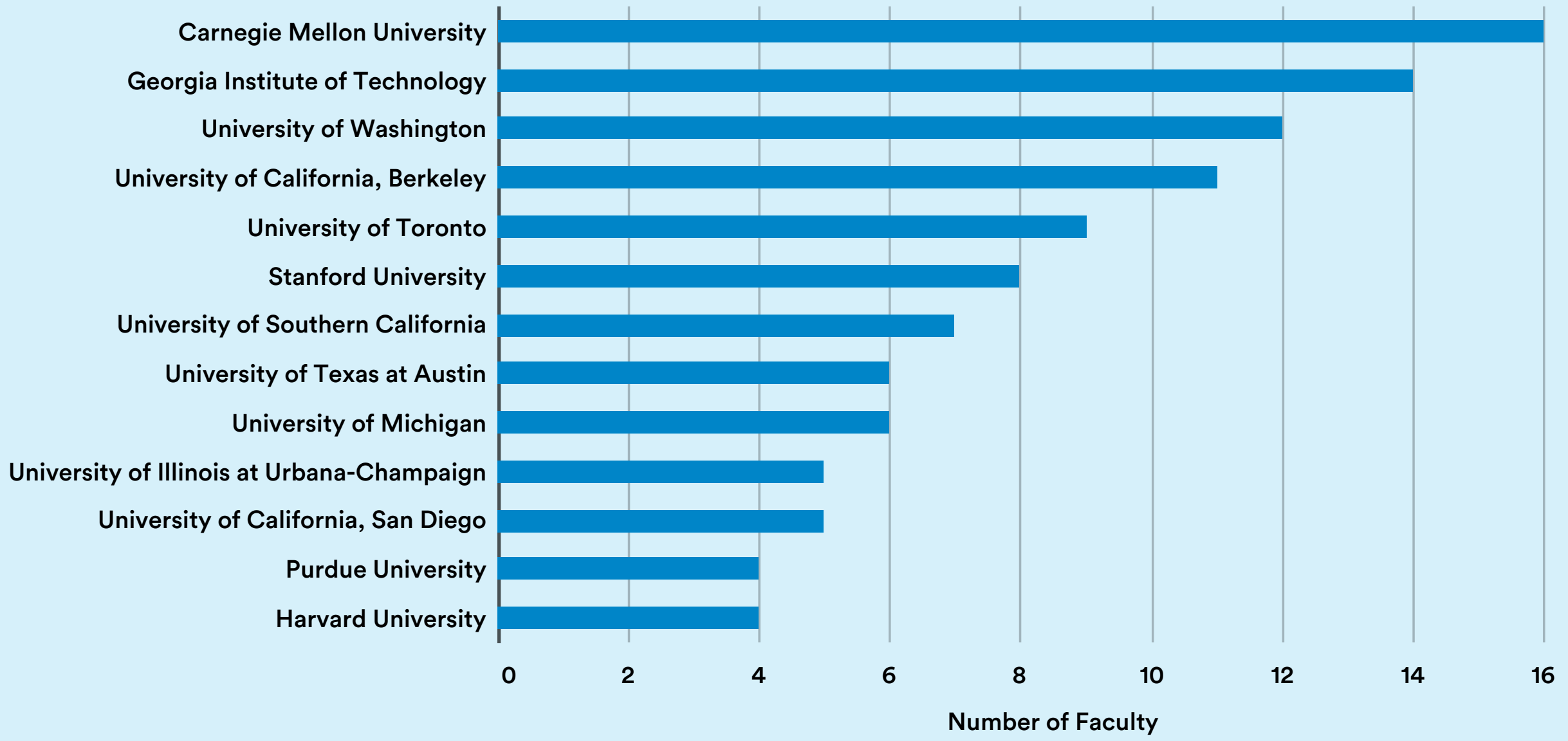

Figure 4.4.2

## CHAPTER 5:

# Ethical Challenges of AI Applications

## CHAPTER 5:

# Chapter Preview

**ACCESS THE PUBLIC DATA**

# Overview

As artificial intelligence–powered innovations become ever more prevalent in our lives, the ethical challenges of AI applications are increasingly evident and subject to scrutiny. As previous chapters have addressed, the use of various AI technologies can lead to unintended but harmful consequences, such as privacy intrusion; discrimination based on gender, race/ethnicity, sexual orientation, or gender identity; and opaque decision-making, among other issues. Addressing existing ethical challenges and building responsible, fair AI innovations before they get deployed has never been more important.

This chapter tackles the efforts to address the ethical issues that have arisen alongside the rise of AI applications. It first looks at the recent proliferation of documents charting AI principles and frameworks, as well as how the media covers AI-related ethical issues. It then follows with a review of ethics-related research presented at AI conferences and what kind of ethics courses are being offered by computer science (CS) departments at universities around the world.

The AI Index team was surprised to discover how little data there is on this topic. Though a number of groups are producing a range of qualitative or normative outputs in the AI ethics domain, the field generally lacks benchmarks that can be used to measure or assess the relationship between broader societal discussions about technology development and the development of the technology itself. One datapoint, covered in the technical performance chapter, is the study by the National Institute of Standards and Technology on facial recognition performance with a focus on bias. Figuring out how to create more quantitative data presents a challenge for the research community, but it is a useful one to focus on. Policymakers are keenly aware of ethical concerns pertaining to AI, but it is easier for them to manage what they can measure, so finding ways to translate qualitative arguments into quantitative data is an essential step in the process.

# CHAPTER HIGHLIGHTS

- The number of papers with ethics-related keywords in titles submitted to AI conferences has grown since 2015, though the average number of paper titles matching ethics-related keywords at major AI conferences remains low over the years.

- The five news topics that got the most attention in 2020 related to the ethical use of AI were the release of the European Commission's white paper on AI, Google's dismissal of ethics researcher Timnit Gebru, the AI ethics committee formed by the United Nations, the Vatican's AI ethics plan, and IBM's exiting the facial-recognition businesses.

# 5.1 AI PRINCIPLES AND FRAMEWORKS

Since 2015, governments, private companies, intergovernmental organizations, and research/professional organizations have been producing normative documents that chart the approaches to manage the ethical challenges of AI applications. Those documents, which include principles, guidelines, and more, provide frameworks for addressing the concerns and assessing the strategies attached to developing, deploying, and governing AI within various organizations. Some common themes that emerge from these AI principles and frameworks include privacy, accountability, transparency, and explainability.

The publication of AI principles signals that organizations are paying heed to and establishing a vision for AI governance. Even so, the proliferation of so-called ethical principles has met with criticism from ethics researchers and human rights practitioners who oppose the imprecise usage of ethics-related terms. The critics also point out that they lack institutional frameworks and are non-binding in most cases. The vague and abstract nature of those principles fails to offer direction on how to implement AI-related ethics guidelines.

Researchers from the AI Ethics Lab in Boston created a ToolBox that tracks the growing body of AI principles. A total of 117 documents relating to AI principles were published between 2015 and 2020. Data shows that research and professional organizations were among the earliest to roll out AI principle documents, and private companies have to date issued the largest number of publications on AI principles among all organization types (Figure 5.1.1). Europe and Central Asia have the highest number of publications as of 2020 (52), followed by North America (41), and East Asia and Pacific (14), according to Figure 5.1.2. In terms of rolling out ethics principles, 2018 was the clear high-water mark for tech companies—including IBM, Google, and Facebook—as well as various U.K., EU, and Australian government agencies.

**Europe and Central Asia have the highest number of publications as of 2020 (44), followed by North America (30), and East Asia and Pacific (14). In terms of rolling out ethics principles, 2018 was the clear high-water mark for tech companies—including IBM, Google, and Facebook—as well as various U.K., EU, and Australian government agencies.**

NUMBER of NEW AI ETHICS PRINCIPLES by ORGANIZATION TYPE, 2015-20

Source: AI Ethics Lab, 2020 | Chart: 2021 AI Index Report

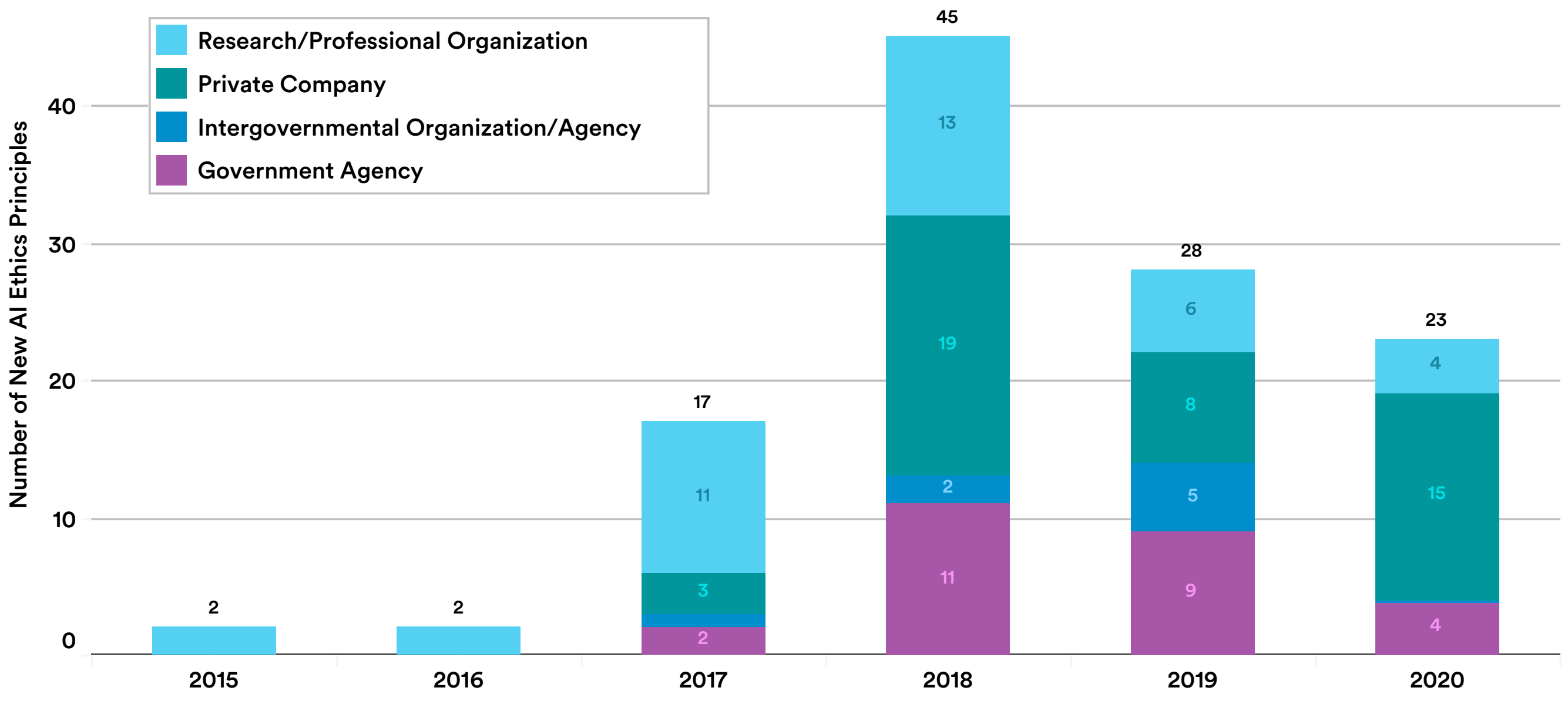

Figure 5.1.1

NUMBER of NEW AI ETHICS PRINCIPLES by REGION, 2015-20

Source: AI Ethics Lab, 2020 | Chart: 2021 AI Index Report

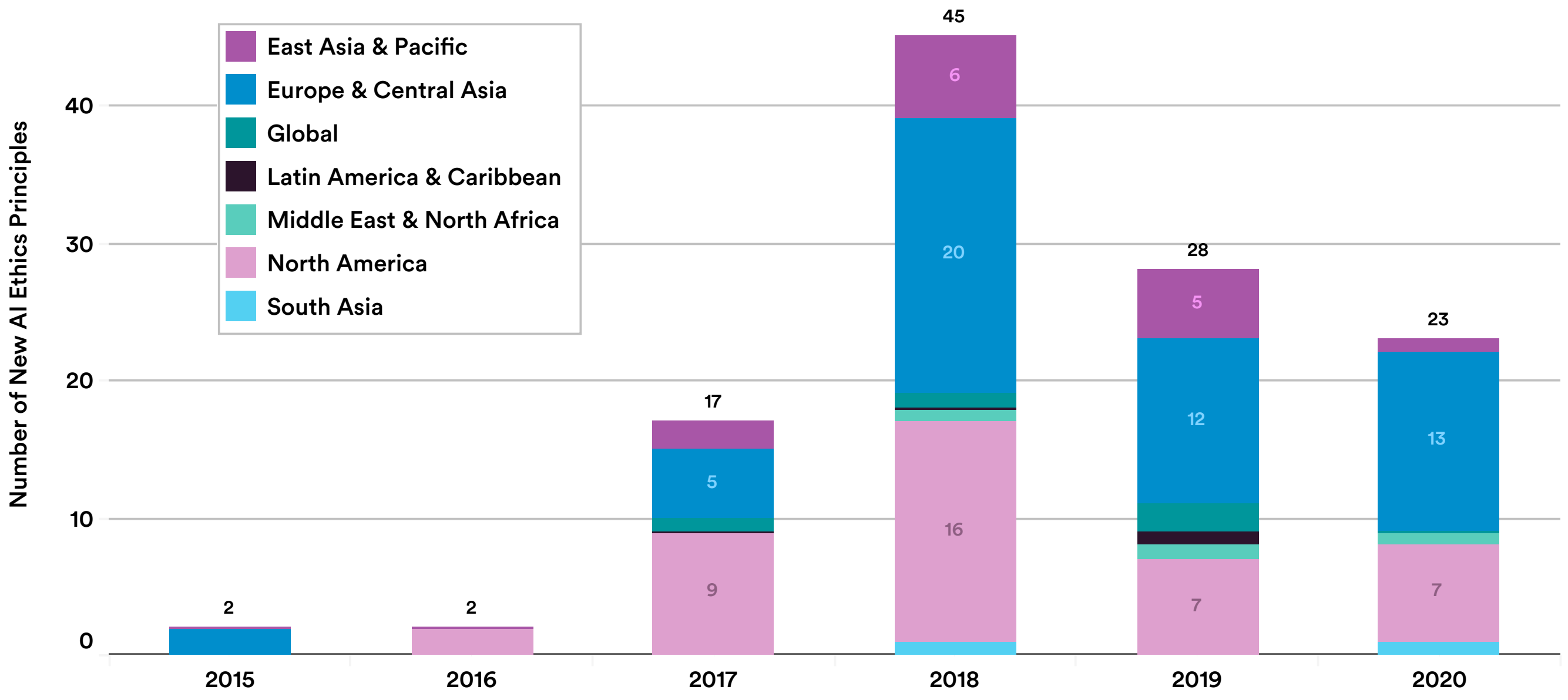

Figure 5.1.2

# 5.2 GLOBAL NEWS MEDIA

How has the news media covered the topic of the ethical use of AI technologies? This section analyzed data from NetBase Quid, which searches the archived news database of LexisNexis for articles that discuss AI ethics[1], analyzing 60,000 English-language news sources and over 500,000 blogs in 2020.

The search found 3,047 articles related to AI technologies that include terms such as "human rights," "human values," "responsibility," "human control," "fairness," "discrimination" or "nondiscrimination," "transparency," "explainability," "safety and security," "accountability," and "privacy." (See the Appendix for more details on search terms.) NetBase Quid clustered the resulting media narratives into seven large themes based on language similarity.

Figure 5.2.1 shows that articles relating to AI ethics guidance and frameworks topped the list of the most covered news topics (21%) in 2020, followed by research and education (20%), and facial recognition (20%).

The five news topics that received the most attention in 2020 related to the ethical use of AI were:

1. The release of the European Commission's white paper on AI (5.9%)
2. Google's dismissal of ethics researcher Timnit Gebru (3.5%)
3. The AI ethics committee formed by the United Nations (2.7%)
4. The Vatican's AI ethics plan (2.6%)
5. IBM exiting the facial-recognition businesses (2.5%).

NEWS COVERAGE on AI ETHICS (% of TOTAL) by THEME, 2020

Source: CAPIQ, Crunchbase, and NetBase Quid, 2020 | Chart: 2021 AI Index Report

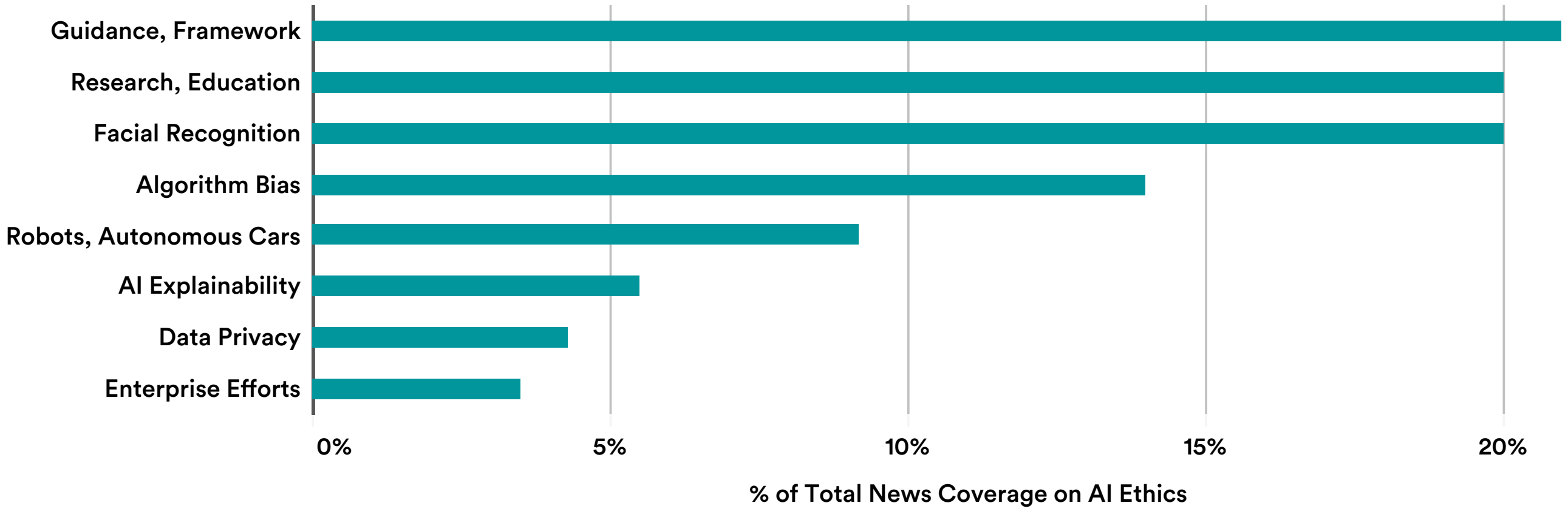

Figure 5.2.1

1 The methodology for this is looking for articles that contain keywords related to AI ethics as determined by a Harvard research study.

# 5.3 ETHICS AT AI CONFERENCES

Researchers are writing more papers that focus directly on the ethics of AI, with submissions in this area more than doubling from 2015 to 2020. To measure the role of ethics in AI research, researchers from the Federal University of Rio Grande do Sul in Porto Alegre, Brazil, searched ethics-related terms in the titles of papers in leading AI, machine learning, and robotics conferences. As Figure 5.3.1 shows, there has been a significant increase in the number of papers with ethics-related keywords in titles submitted to AI conferences since 2015.

Further analysis in Figure 5.3.2 shows the average number of keyword matches throughout all publications among the six major AI conferences. Despite the growing mentions in the previous chart, the average number of paper titles matching ethics-related keywords at major AI conferences remains low over the years.

Changes are coming to AI conferences, though. Starting in 2020, the topic of ethics was more tightly integrated into conference proceedings. For instance, the Neural Information Processing Systems (NeurIPS) conference, one of the biggest AI research conferences in the world, asked researchers to submit "Broader Impacts" statements alongside their work for the first time in 2020, which led to a deeper integration of ethical concerns into technical work. Additionally, there has been a recent proliferation of conferences and workshops that specifically focus on responsible AI, including the new Artificial Intelligence, Ethics, and Society Conference by the Association for the Advancement of Artificial Intelligence and the Conference on Fairness, Accountability, and Transparency by the Association for Computing Machinery.

**There has been a significant increase in the number of papers with ethics-related keywords in titles submitted to AI conferences since 2015. Further analysis shows the average number of keyword matches throughout all publications among the six major AI conferences.**

NUMBER of PAPER TITLES MENTIONING ETHICS KEYWORDS at AI CONFERENCES, 2000-19
Source: Prates et al., 2018 | Chart: 2021 AI Index Report

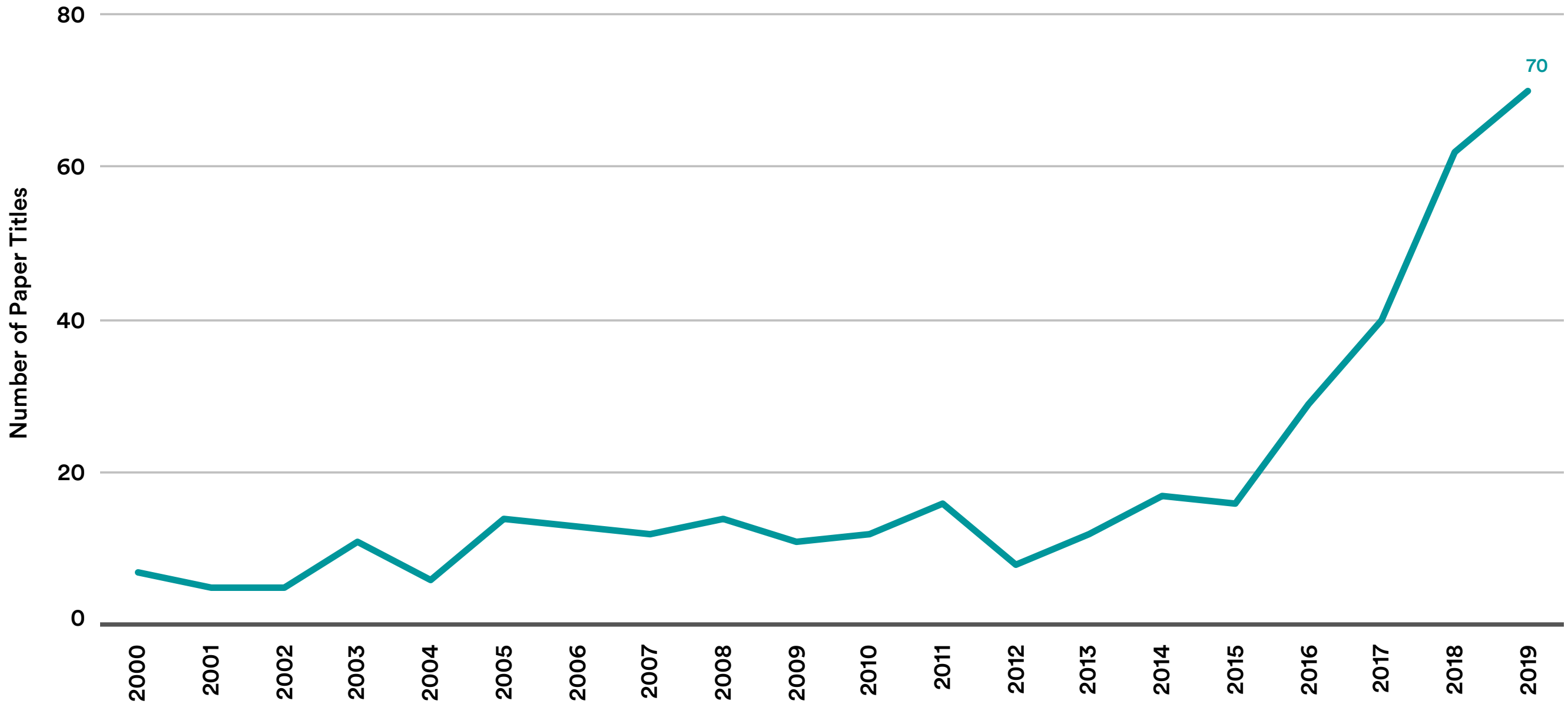

Figure 5.3.1

AVERAGE NUMBER of PAPER TITLES MENTIONING ETHICS KEYWORDS at SELECT LARGE AI CONFERENCES, 2000-19
Source: Prates et al., 2018 | Chart: 2021 AI Index Report

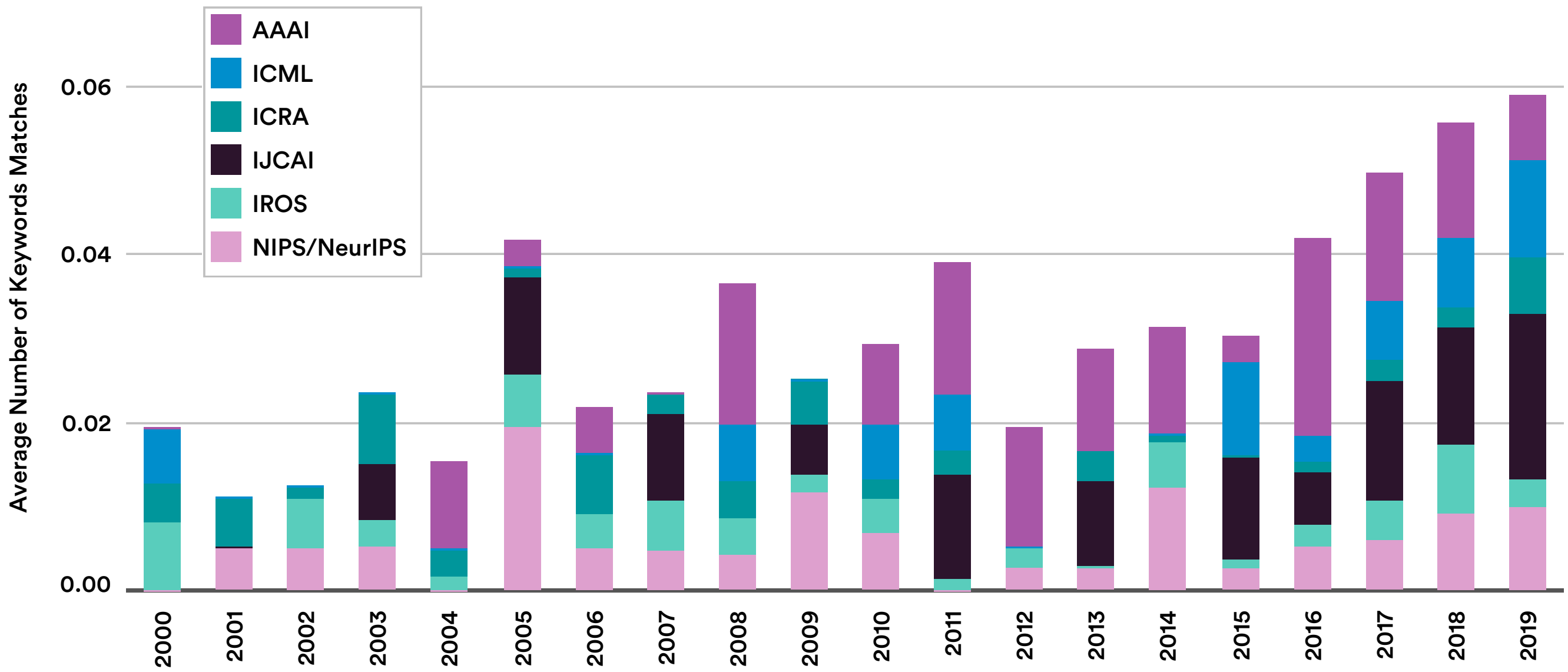

Figure 5.3.2

# 5.4 ETHICS OFFERINGS AT HIGHER EDUCATION INSTITUTIONS

Chapter 4 introduced a survey of computer science departments or schools at top universities around the world in order to assess the state of AI education in higher education institutions.[2] In part, the survey asked whether the CS department or university offers the opportunity to learn about the ethical side of AI and CS. Among the 16 universities that completed the survey, 13 reported some type of relevant offering.

Figure 5.4.1 shows that 11 of the 18 departments report hosting keynote events or panel discussions on AI ethics, while 7 of them offer stand-alone courses on AI ethics in CS or other departments at their university. Some universities also offer classes on ethics in the computer science field in general, including stand-alone CS ethics courses or ethics modules embedded in the CS curriculum offering.[3]

AI ETHICS OFFERING at CS DEPARTMENTS of TOP UNIVERSITIES around the WORLD, AY 2019-20

Source: AI Index, 2020 | Chart: 2021 AI Index Report

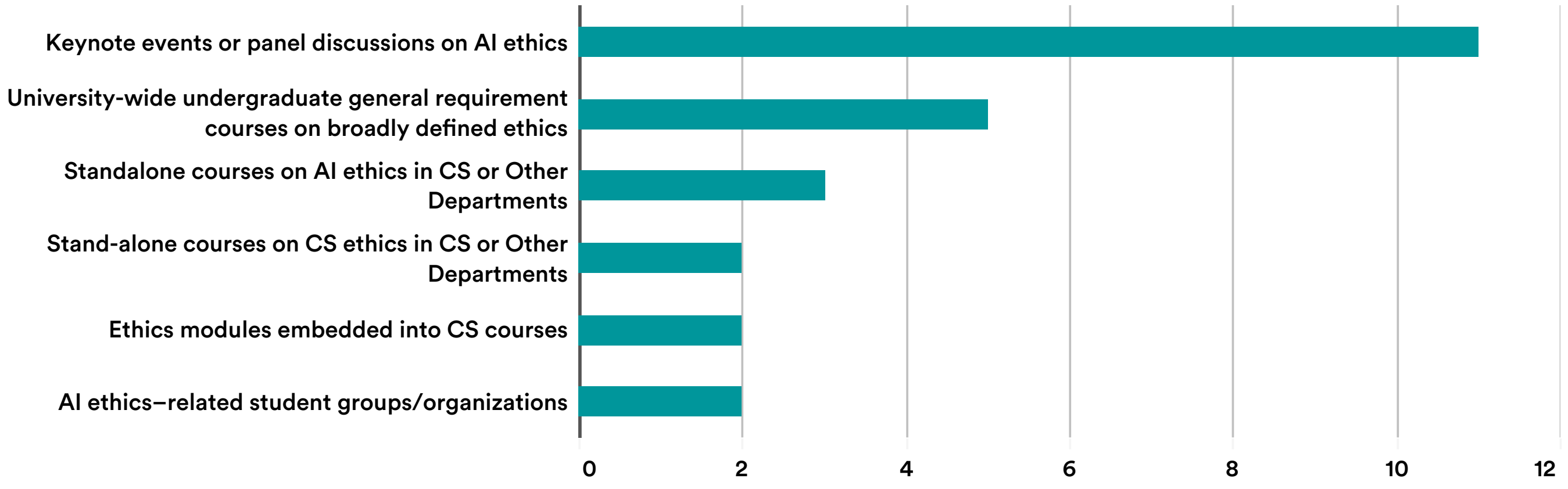

Figure 5.4.1

2 The survey was distributed to 73 universities online over three waves from November 2020 to January 2021 and completed by 18 universities, a 24.7% response rate. The 18 universities are—Belgium: Katholieke Universiteit Leuven; Canada: McGill University; China: Shanghai Jiao Tong University, Tsinghua University; Germany: Ludwig Maximilian University of Munich, Technical University of Munich; Russia: Higher School of Economics, Moscow Institute of Physics and Technology; Switzerland: École Polytechnique Fédérale de Lausanne; United Kingdom: University of Cambridge; United States: California Institute of Technology, Carnegie Mellon University (Department of Machine Learning), Columbia University, Harvard University, Stanford University, University of Wisconsin–Madison, University of Texas at Austin, Yale University.

3 The survey did not explicitly present "Ethics modules embedded into CS courses" as an option. Selections were filled in the "Others" option. This will be included in next year's survey.

# CHAPTER 6:
# Diversity in AI

Artificial Intelligence
Index Report 2021

## CHAPTER 6:
# Chapter Preview

**ACCESS THE PUBLIC DATA**

# Overview

While artificial intelligence (AI) systems have the potential to dramatically affect society, the people building AI systems are not representative of the people those systems are meant to serve. The AI workforce remains predominantly male and lacking in diversity in both academia and the industry, despite many years highlighting the disadvantages and risks this engenders. The lack of diversity in race and ethnicity, gender identity, and sexual orientation not only risks creating an uneven distribution of power in the workforce, but also, equally important, reinforces existing inequalities generated by AI systems, reduces the scope of individuals and organizations for whom these systems work, and contributes to unjust outcomes.

This chapter presents diversity statistics within the AI workforce and academia. It draws on collaborations with various organizations—in particular, Women in Machine Learning (WiML), Black in AI (BAI), and Queer in AI (QAI)— each of which aims to improve diversity in some dimension in the field. The data is neither comprehensive nor conclusive. In preparing this chapter, the AI Index team encountered significant challenges as a result of the sparsity of publicly available demographic data. The lack of publicly available demographic data limits the degree to which statistical analyses can assess the impact of the lack of diversity in the AI workforce on society as well as broader technology development. The diversity issue in AI is well known, and making more data available from both academia and industry is essential to measuring the scale of the problem and addressing it.

There are many dimensions of diversity that this chapter does not cover, including AI professionals with disabilities; nor does it consider diversity through an intersectional lens. Other dimensions will be addressed in future iterations of this report. Moreover, these diversity statistics tell only part of the story. The daily challenges of minorities and marginalized groups working in AI, as well as the structural problems within organizations that contribute to the lack of diversity, require more extensive data collection and analysis.

1 We thank Women in Machine Learning, Black in AI, and Queer in AI for their work to increase diversity in AI, for sharing their data, and for partnering with us.

## CHAPTER HIGHLIGHTS

- The percentages of female AI PhD graduates and tenure-track computer science (CS) faculty have remained low for more than a decade. Female graduates of AI PhD programs in North America have accounted for less than 18% of all PhD graduates on average, according to an annual survey from the Computing Research Association (CRA). An AI Index survey suggests that female faculty make up just 16% of all tenure-track CS faculty at several universities around the world.

- The CRA survey suggests that in 2019, among new U.S. resident AI PhD graduates, 45% were white, while 22.4% were Asian, 3.2% were Hispanic, and 2.4% were African American.

- The percentage of white (non-Hispanic) new computing PhDs has changed little over the last 10 years, accounting for 62.7% on average. The share of Black or African American (non-Hispanic) and Hispanic computing PhDs in the same period is significantly lower, with an average of 3.1% and 3.3%, respectively.

- The participation in Black in AI workshops, which are co-located with the Conference on Neural Information Processing Systems (NeurIPS), has grown significantly in recent years. The numbers of attendees and submitted papers in 2019 are 2.6 times higher than in 2017, while the number of accepted papers is 2.1 times higher.

- In a membership survey by Queer in AI in 2020, almost half the respondents said they view the lack of inclusiveness in the field as an obstacle they have faced in becoming a queer practitioner in the AI/ML field. More than 40% of members surveyed said they have experienced discrimination or harassment as a queer person at work or school.

# 6.1 GENDER DIVERSITY IN AI

## WOMEN IN ACADEMIC AI SETTINGS

Chapter 4 introduced the AI Index survey that evaluates the state of AI education in CS departments at top universities around the world, along with the Computer Research Association's annual Taulbee Survey on the enrollment, production, and employment of PhDs in information, computer science, and computer engineering in North America.

Data from both surveys show that the percentage of female AI and CS PhD graduates as well as tenure-track CS faculty remains low. Female graduates of AI PhD programs and CS PhD programs have accounted for 18.3% of all PhD graduates on average within the past 10 years (Figure 6.1.1). Among the 17 universities that completed the AI Index survey of CS programs globally, female faculty make up just 16.1% of all tenure-track faculty whose primary research focus area is AI (Figure 6.1.2).

TENURE-TRACK FACULTY at CS DEPARTMENTS of TOP UNIVERSITIES around the WORLD by GENDER, AY 2019-20

Source: AI Index, 2020 | Chart: 2021 AI Index Report

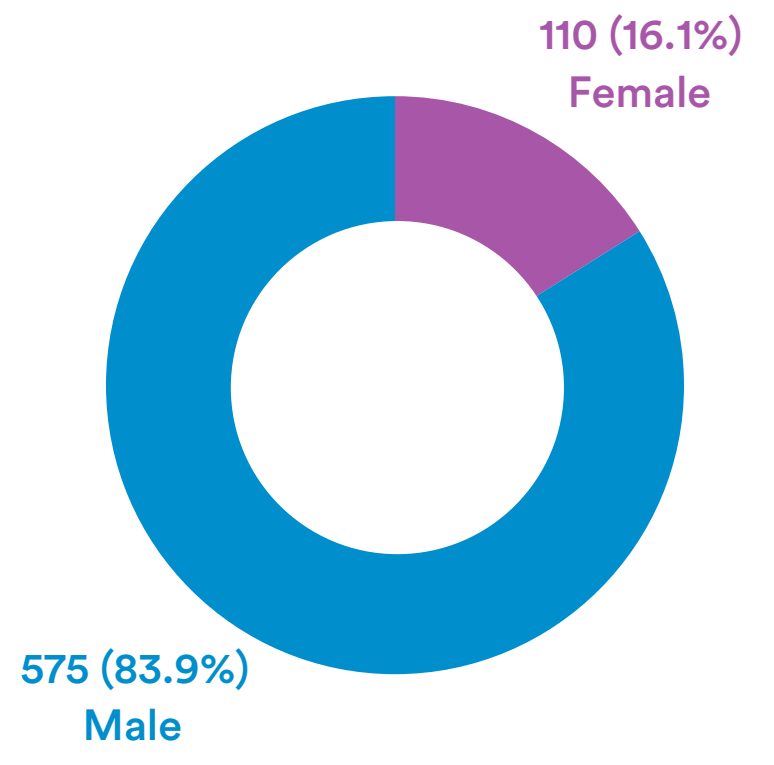

Figure 6.1.2

FEMALE NEW AI and CS PHDS (% of TOTAL NEW AI and CS PHDS) in NORTH AMERICA, 2010-19

Source: CRA Taulbee Survey, 2020 | Chart: 2021 AI Index Report

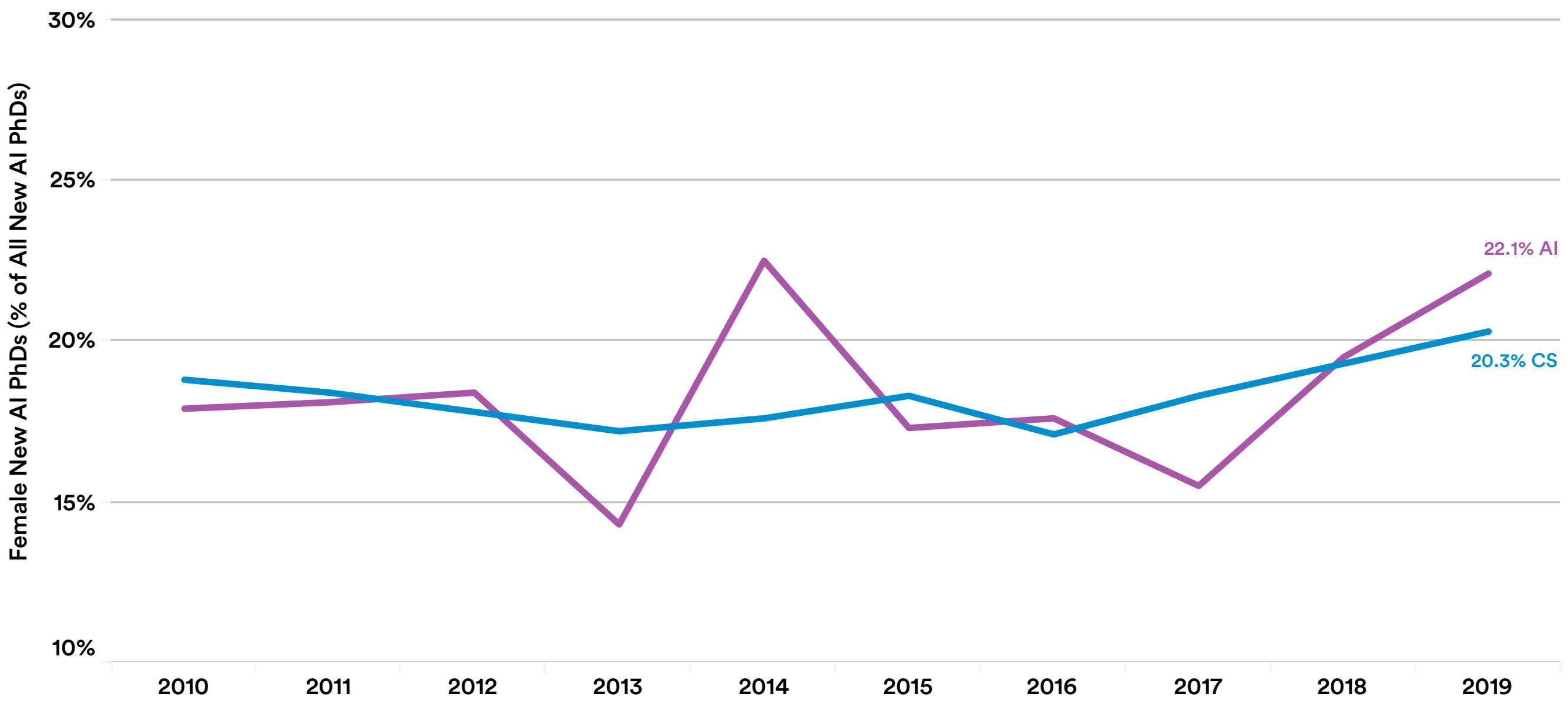

Figure 6.1.1

## WOMEN IN THE AI WORKFORCE

Chapter 3 introduced the "global relative AI skills penetration rate," a measure that reflects the prevalence of AI skills across occupations, or the intensity with which people in certain occupations use AI skills. Figure 6.1.3 shows AI skills penetration by country for female and male labor pools in a set of select countries.[2] The data suggest that across the majority of these countries, the AI skills penetration rate for women is lower than that for men. Among the 12 countries we examined, India, South Korea, Singapore, and Australia are the closest to reaching equity in terms of the AI skills penetration rate of females and males.

**This data suggests that across the majority of select countries, the AI skills penetration rate for women is lower than it is for men.**

RELATIVE AI SKILLS PENETRATION RATE by GENDER, 2015-20

Source: LinkedIn, 2020 | Chart: 2021 AI Index Report

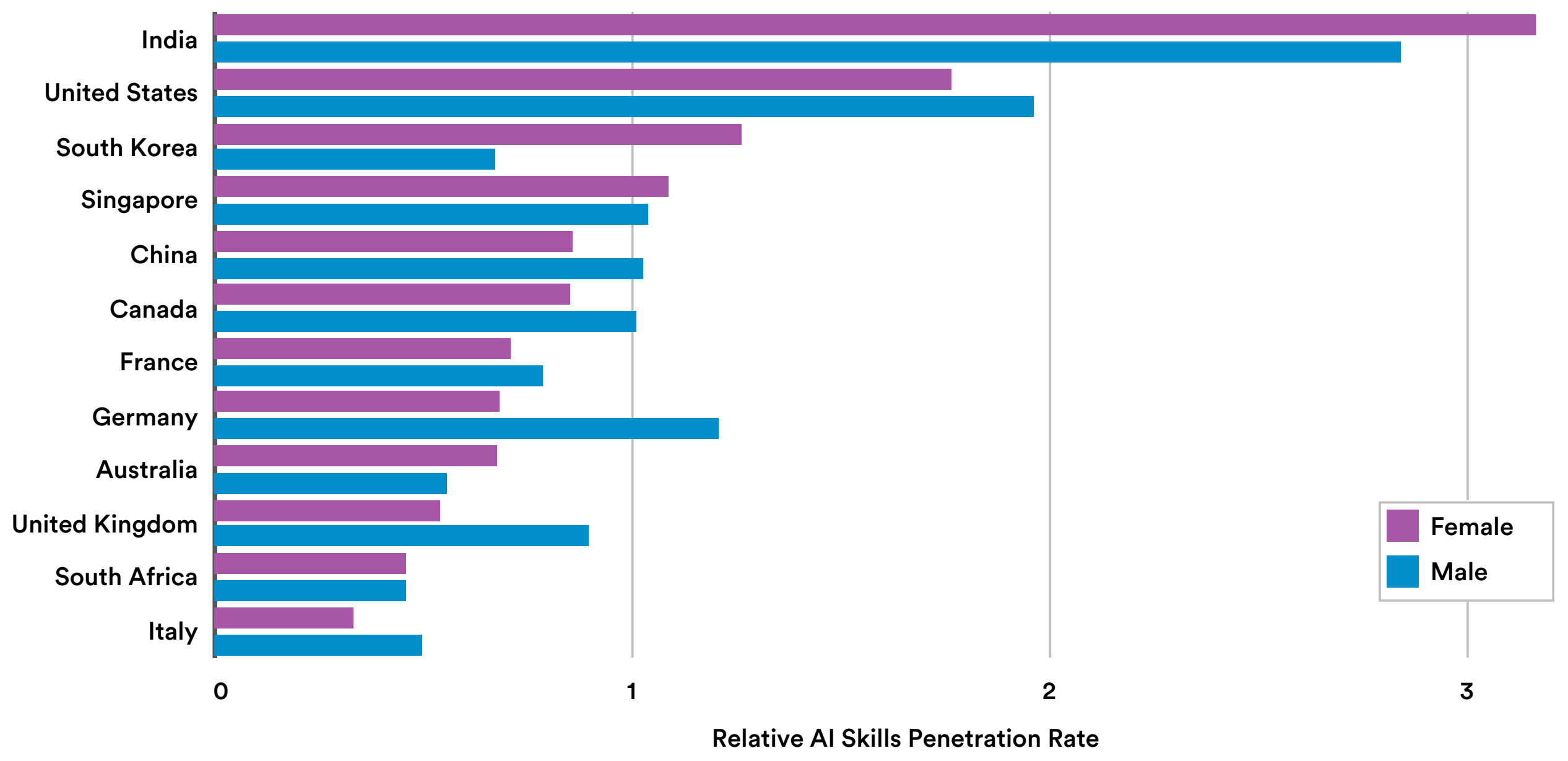

Figure 6.1.3

2 Countries included are a select sample of eligible countries with at least 40% labor force coverage by LinkedIn and at least 10 AI hires in any given month. China and India were included in this sample because of their increasing importance in the global economy, but LinkedIn coverage in these countries does not reach 40% of the workforce. Insights for these countries may not provide as full a picture as other countries, and should be interpreted accordingly.

## WOMEN IN MACHINE LEARNING WORKSHOPS

Women in Machine Learning, founded in 2006 by Hanna Wallach, Jenn Wortman, and Lisa Wainer, is an organization that runs events and programs to support women in the field of machine learning (ML). This section presents statistics from its annual technical workshops, which are held at NeurIPS. In 2020, WiML also hosted for the first time a full-day "Un-Workshop" at the International Conference on Machine Learning 2020, which drew 812 participants.

### Workshop Participants

The number of participants attending WiML workshops at NeurIPS has been steadily increasing since the workshops were first offered in 2006. According to the organization, the WiML workshop in 2020 was completely virtual because of the pandemic and delivered on a new platform (Gather.Town); these two factors may make attendance numbers harder to compare to those of previous years. Figure 6.1.4 shows an estimate of 925 attendees in 2020, based on the number of individuals who accessed the virtual platform.

In the past 10 years, WiML workshops have expanded their programs to include mentoring roundtables, where more senior participants offer one-on-one feedback and professional advice, in addition to the main session that includes keynotes and poster presentations. Similar opportunities may have contributed to the increase in attendance since 2014. Between 2016 and 2019, the WiML workshop attendance is on average about 10% of the overall NeurIPS attendance.

NUMBER of PARTICIPANTS at WIML WORKSHOP at NEURIPS, 2006-20

Source: Women in Machine Learning, 2020 | Chart: 2021 AI Index Report

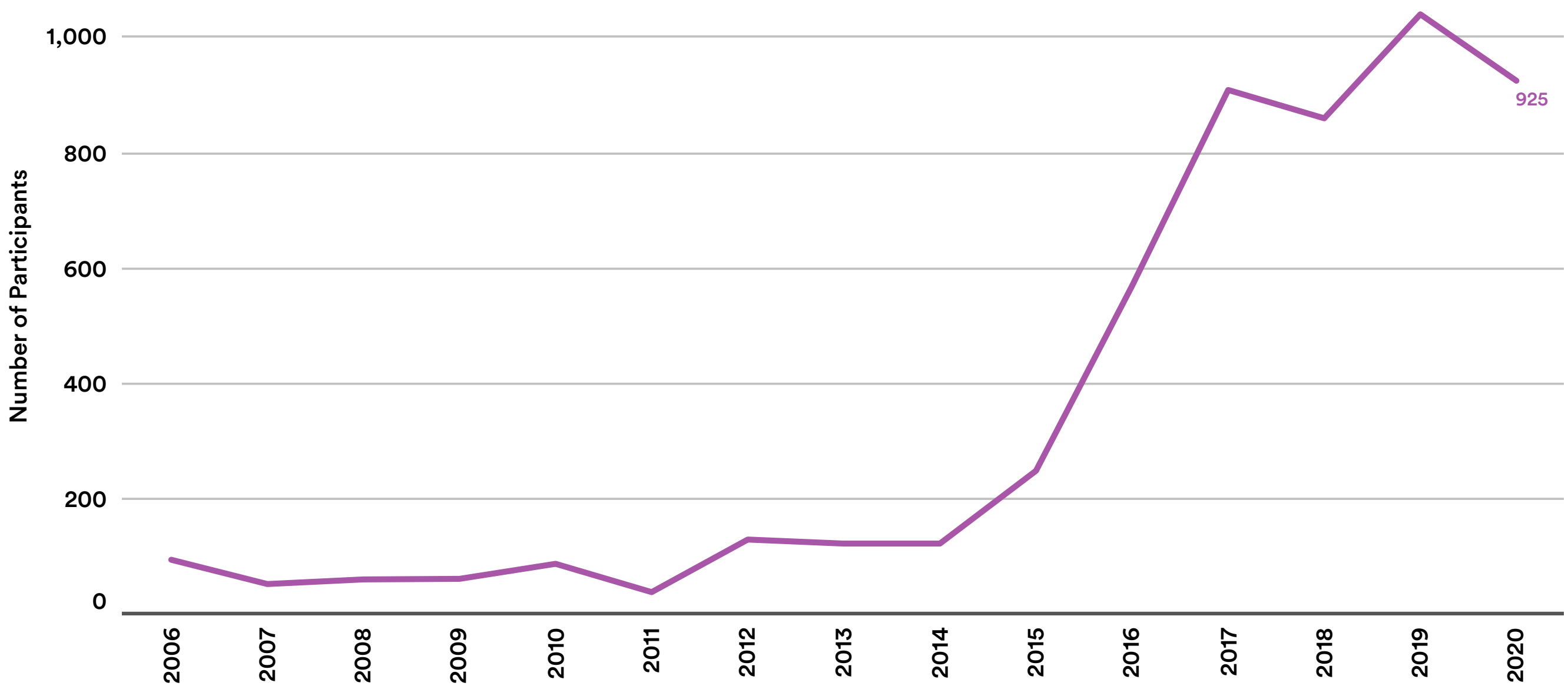

Figure 6.1.4

## Demographics Breakdown

The following geographic, professional position, and gender breakdowns are based only on participants at the 2020 WiML workshop at NeurIPS who consented to having the information aggregated and who spent at least 10 minutes on the virtual platform through which the workshop was offered. Among the participants, 89.5% were women and/or nonbinary, 10.4% were men (Figure 6.1.5), and a large majority were from North America (Figure 6.1.6). Further, as shown in Figure 6.1.7, students—including PhD, master's, and undergraduate students—make up more than half the participants (54.6%). Among participants who work in the industry, research scientist/engineer and data scientist/engineer are the most commonly held professional positions.

**Among the participants, 89.5% were women and/or nonbinary, 10.4% were men, and a large majority were from North America. Further, students—including PhD, master's, and undergraduate students—make up more than half the participants (54.6%).**

**PARTICIPANTS of WIML WORKSHOP at NEURIPS (% of TOTAL) by GENDER, 2020**
Source: Women in Machine Learning, 2020 | Chart: 2021 AI Index Report

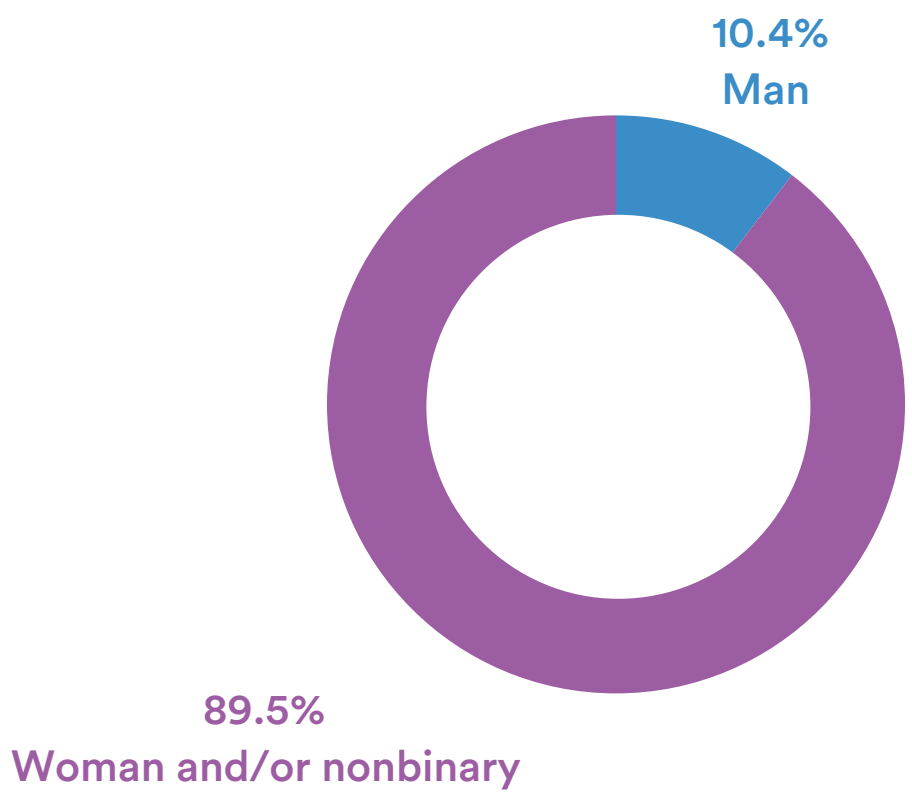

Figure 6.1.5

PARTICIPANTS of WIML WORKSHOP at NEURIPS (% of TOTAL) by CONTINENT of RESIDENCE, 2020
Source: Women in Machine Learning, 2020 | Chart: 2021 AI Index Report

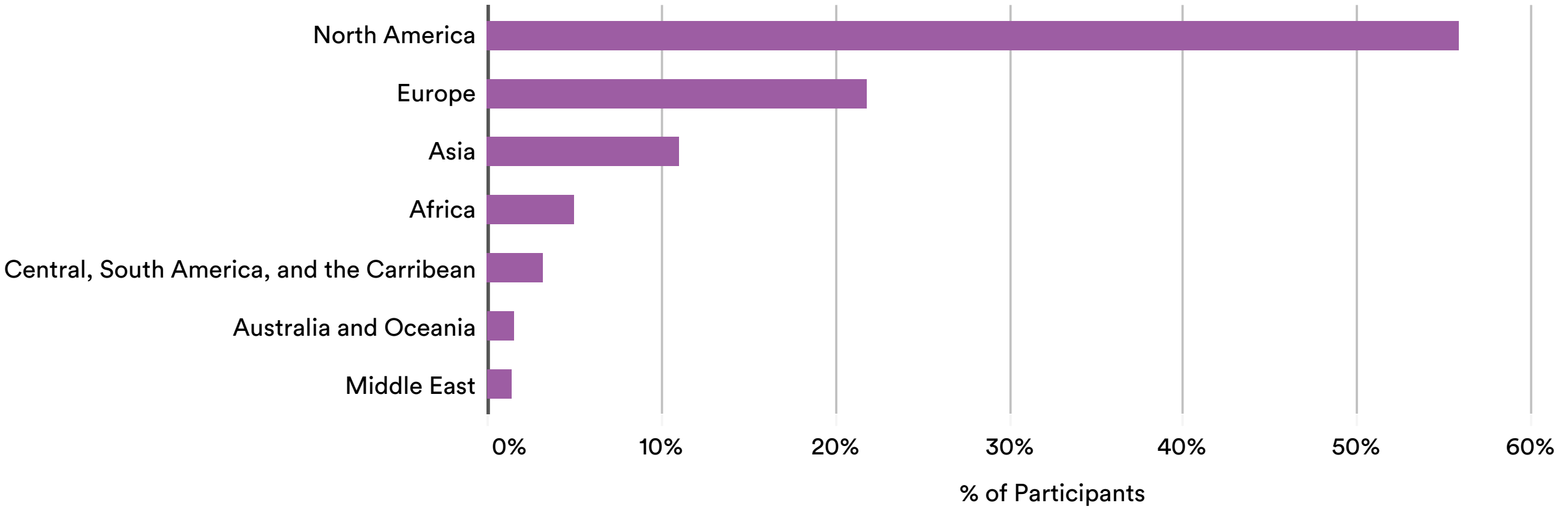

Figure 6.1.6

PARTICIPANTS of WIML WORKSHOP at NEURIPS (% of TOTAL) by TOP 10 PROFESSIONAL POSITIONS, 2020
Source: Women in Machine Learning, 2020 | Chart: 2021 AI Index Report

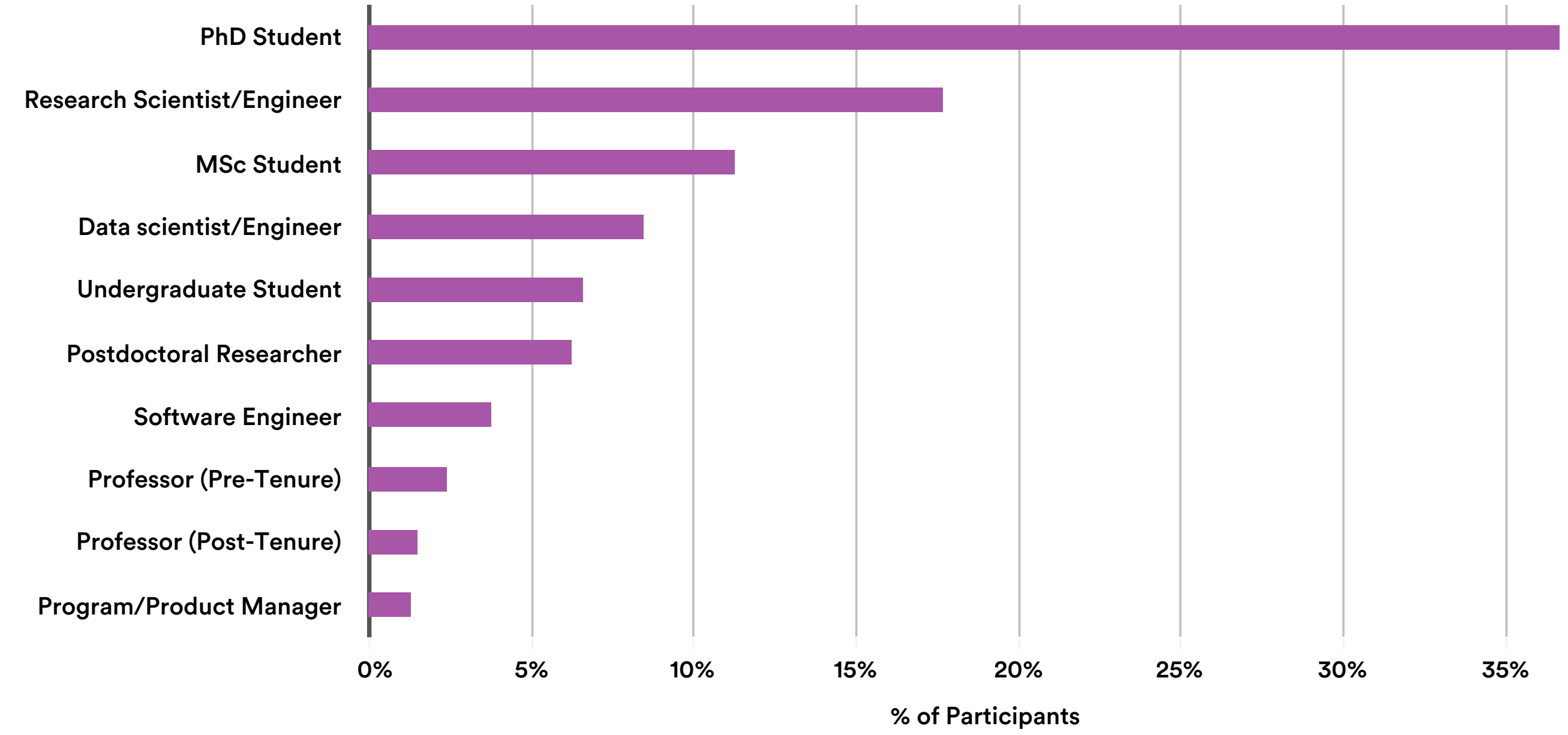

Figure 6.1.7

# 6.2 RACIAL AND ETHNIC DIVERSITY IN AI

## NEW AI PHDS IN THE UNITED STATES BY RACE/ETHNICITY

According to the CRA Taulbee Survey, among the new AI PhDs in 2019 who are U.S. residents, the largest percentage (45.6%) are white (non-Hispanic), followed by Asian (22.4%). By comparison, 2.4% were African American (non-Hispanic) and 3.2% were Hispanic (Figure 6.2.1).

### NEW U.S. RESIDENT AI PHDS (% of TOTAL) by RACE/ETHNICITY, 2019

Source: CRA Taulbee Survey, 2020 | Chart: 2021 AI Index Report

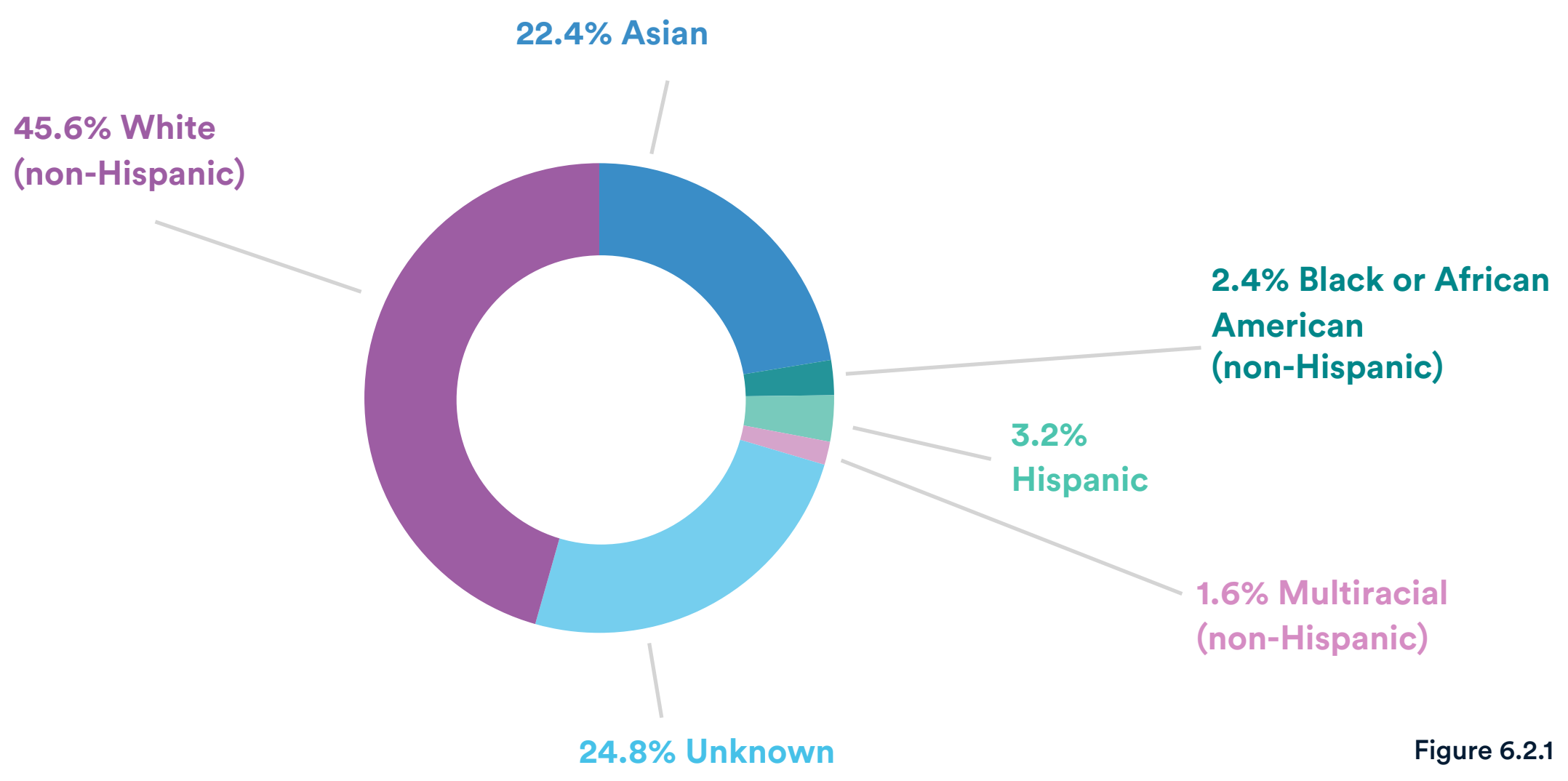

Figure 6.2.1

## NEW COMPUTING PHDS IN THE UNITED STATES BY RACE/ETHNICITY

Figure 6.2.2 shows all PhDs awarded in the United States to U.S. residents across departments of computer science (CS), computer engineering (CE), and information (I) between 2010 and 2019. The CRA survey indicates that the percentage of white (non-Hispanic) new PhDs has changed little over the last 10 years, accounting for 62.7% on average. The share of new Black or African American (non-Hispanic) and Hispanic computing PhDs in the same period is significantly lower, with an average of 3.1% and 3.3%, respectively. We are not able to compare the numbers between new AI and CS PhDs in 2019 because of the number of unknown cases (24.8% for new AI PhDs and 8.5% for CS PhDs).

**NEW COMPUTING PHDS, U.S. RESIDENT (% of TOTAL) by RACE/ETHNICITY, 2010-19**
Source: CRA Taulbee Survey, 2020 | Chart: 2021 AI Index Report

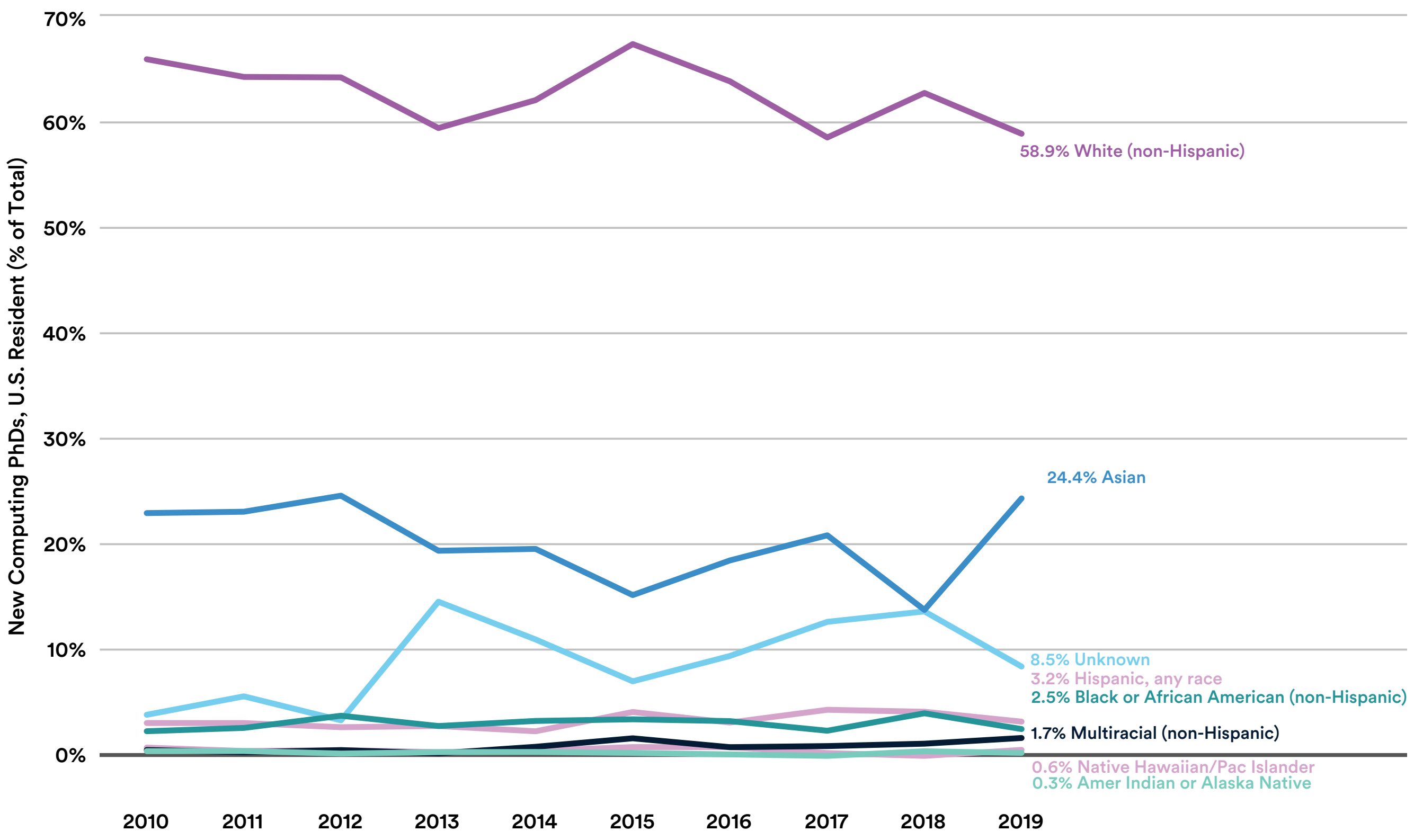

Figure 6.2.2

## CS TENURE-TRACK FACULTY BY RACE/ ETHNICITY

Figure 6.2.3 shows data from the AI Index education survey.[3] Among 15 universities that completed the question pertaining to the racial makeup of their faculty, approximately 67.0% of the tenure-track faculty are white, followed by Asian (14.3%), other races (8.3%), and mixed/other race, ethnicity, or origin (6.3%). The smallest representation among tenure-track faculty are teachers of Black or African and of Hispanic, Latino, or Spanish origins, who account for 0.6% and 0.8%, respectively.

**TENURE-TRACK FACULTY (% of TOTAL) at CS DEPARTMENTS of TOP UNIVERSITIES in the WORLD by RACE/ETHNICITY, 2019-20**
Source: AI Index, 2020 | Chart: 2021 AI Index Report

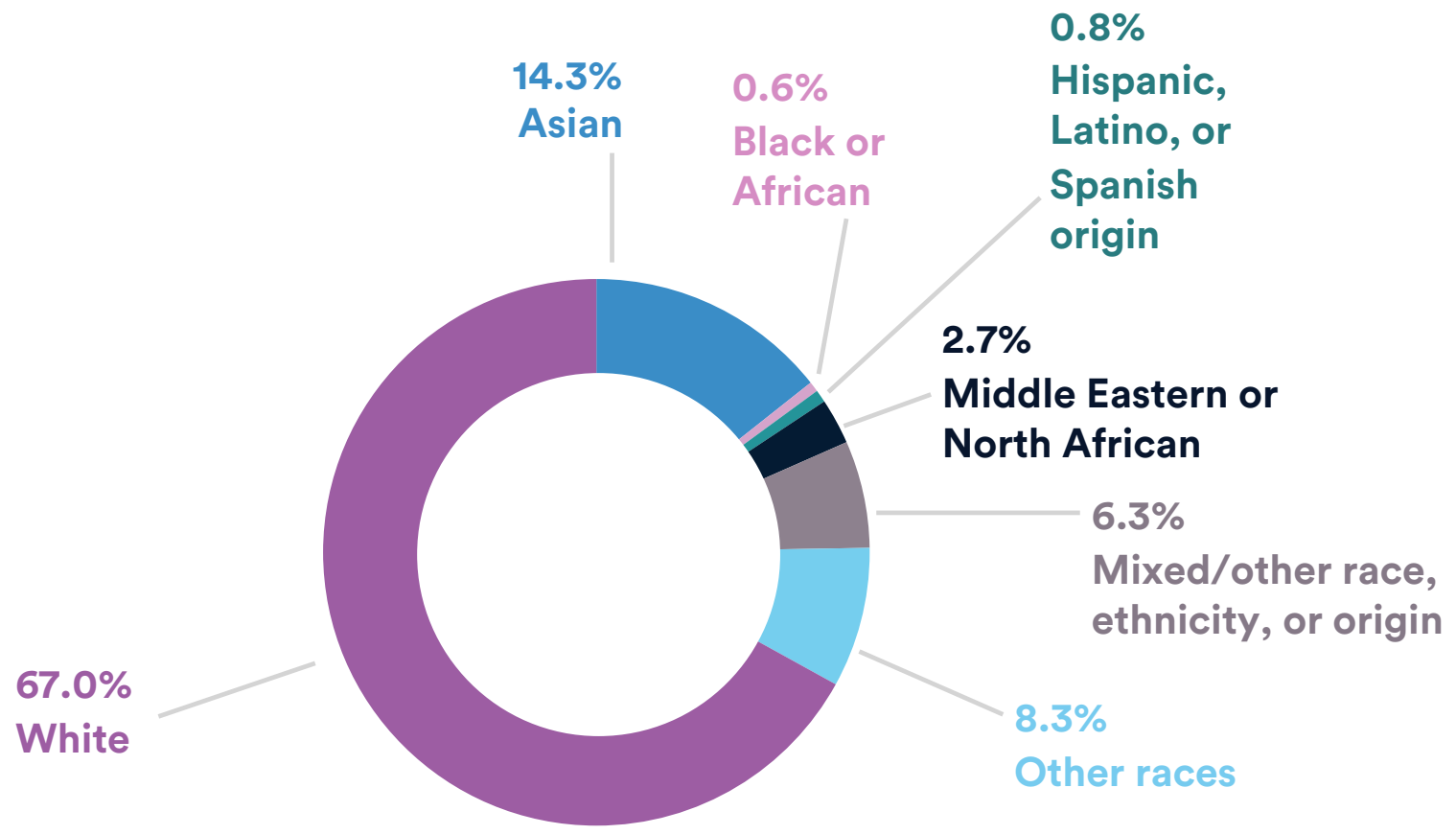

Figure 6.2.3

## BLACK IN AI

Black in AI (BAI), founded in 2017 by Timnit Gebru and Rediet Abebe, is a multi-institutional and transcontinental initiative that aims to increase the presence of Black people in the field of AI. As of 2020, BAI has around 3,000 community members and allies, has held more than 10 workshops at major AI conferences, and has helped increase the number of Black people participating at major AI conferences globally 40-fold. Figure 6.2.4 shows the number of attendees, submitted papers, and accepted papers from the annual Black in AI Workshop, which is co-located with NeurIPS.[4] The numbers of attendees and accepted papers in 2019 are 2.6 times higher than in 2017, while the number of accepted papers is 2.1 times higher.

**NUMBER OF ATTENDEES, SUBMITTED PAPERS, and ACCEPTED PAPERS at BLACK in AI WORKSHOP CO-LOCATED with NEURIPS, 2017-19**
Source: Black in AI, 2020 | Chart: 2021 AI Index Report

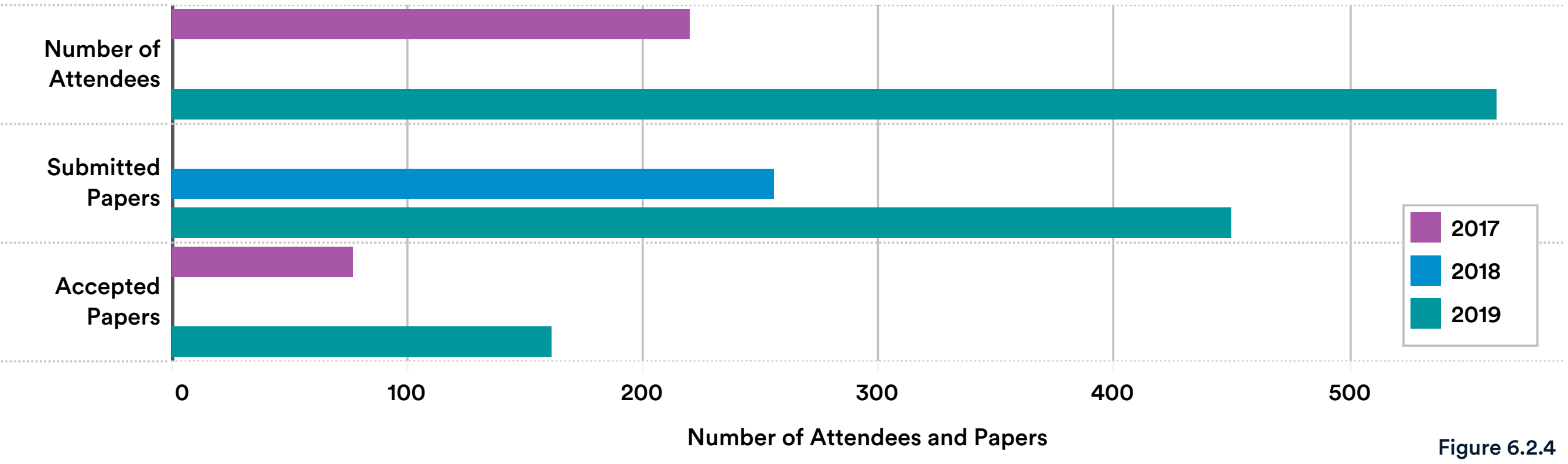

Figure 6.2.4

3 The survey was distributed to 73 universities online over three waves from November 2020 to January 2021 and completed by 18 universities, a 24.7% response rate. The 18 universities are Belgium: Katholieke Universiteit Leuven; Canada: McGill University; China: Shanghai Jiao Tong University, Tsinghua University; Germany: Ludwig Maximilian University of Munich, Technical University of Munich; Russia: Higher School of Economics, Moscow Institute of Physics and Technology; Switzerland: École Polytechnique Fédérale de Lausanne; United Kingdom: University of Cambridge; United States: California Institute of Technology, Carnegie Mellon University (Department of Machine Learning), Columbia University, Harvard University, Stanford University, University of Wisconsin–Madison, University of Texas at Austin, Yale University.
4 The 2020 data are clearly affected by the pandemic and not included as a result. For more information, see the Black in AI impact report.

# 6.3 GENDER IDENTITY AND SEXUAL ORIENTATION IN AI

## QUEER IN AI

This section presents data from a membership survey by Queer in AI (QAI), [5] an organization that aims to make the AI/ML community one that welcomes, supports, and values queer scientists. Founded in 2018 by William Agnew, Raphael Gontijo Lopes, and Eva Breznik, QAI builds a visible community of queer and ally AI/ML scientists through meetups, poster sessions, mentoring, and other initiatives.

### Demographics Breakdown

According to the 2020 survey, with around 100 responses, about 31.5% of respondents identify as gay, followed by bisexual, queer, and lesbian (Figure 6.3.1); around 37.0% and 26.1% of respondents identify as cis male and cis female, respectively, followed by gender queer, gender fluid, nonbinary, and others (Figure 6.3.2). Trans female and male account for 5.0% and 2.5% of total members, respectively. Moreover, the past three years of surveys show that students make up the majority of QAI members—around 41.7% of all respondents on average (Figure 6.3.3), followed by junior-level professionals in academia or industry.

QAI MEMBERSHIP SURVEY: WHAT IS YOUR SEXUAL ORIENTATION, 2020

Source: Queer in AI, 2020 | Chart: 2021 AI Index Report

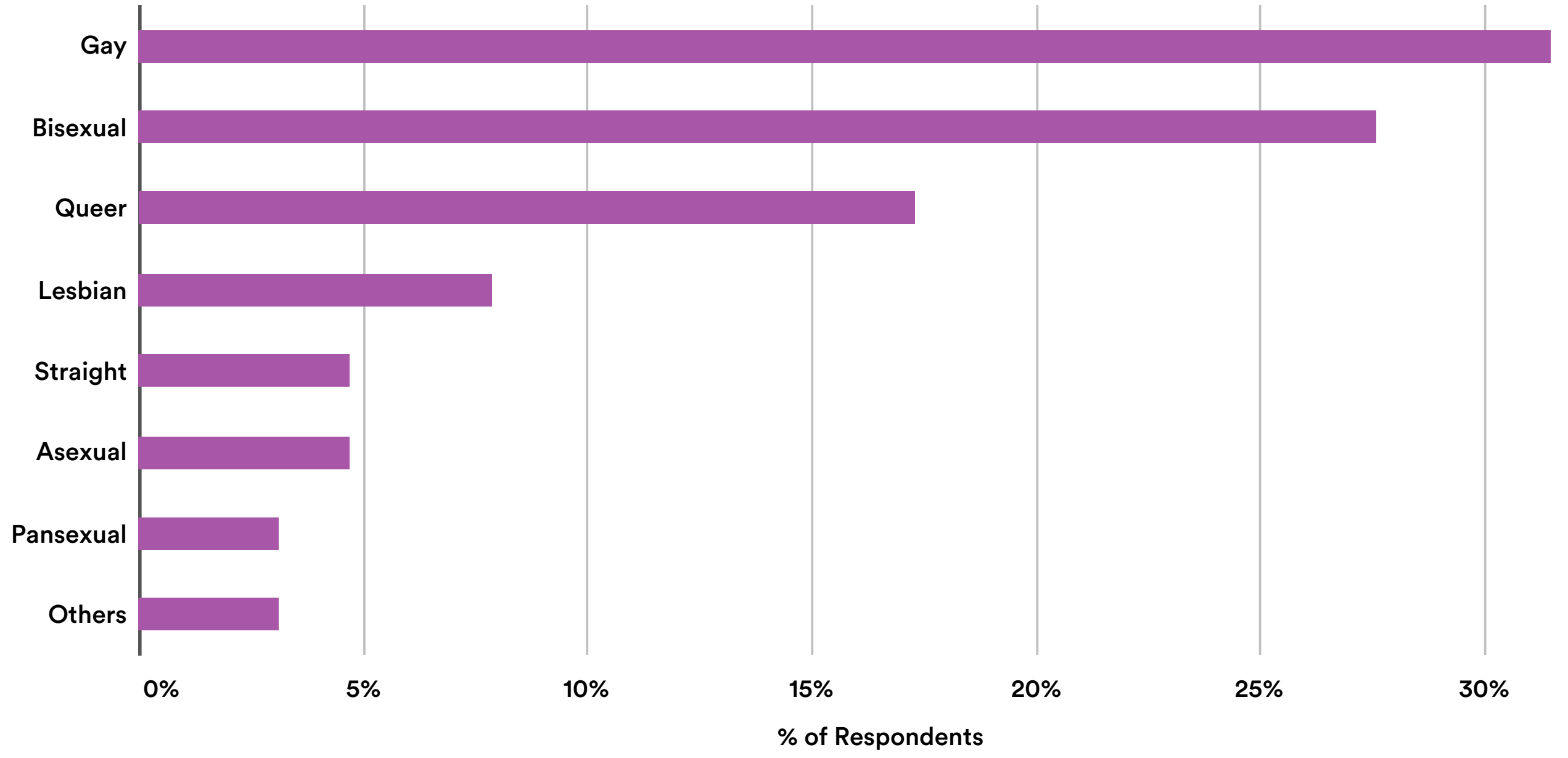

Figure 6.3.1

5 Queer in AI presents the survey results at its workshop at the annual NeurIPS conference.

## QAI MEMBERSHIP SURVEY: WHAT IS YOUR GENDER IDENTITY, 2020

Source: Queer in AI, 2020 | Chart: 2021 AI Index Report

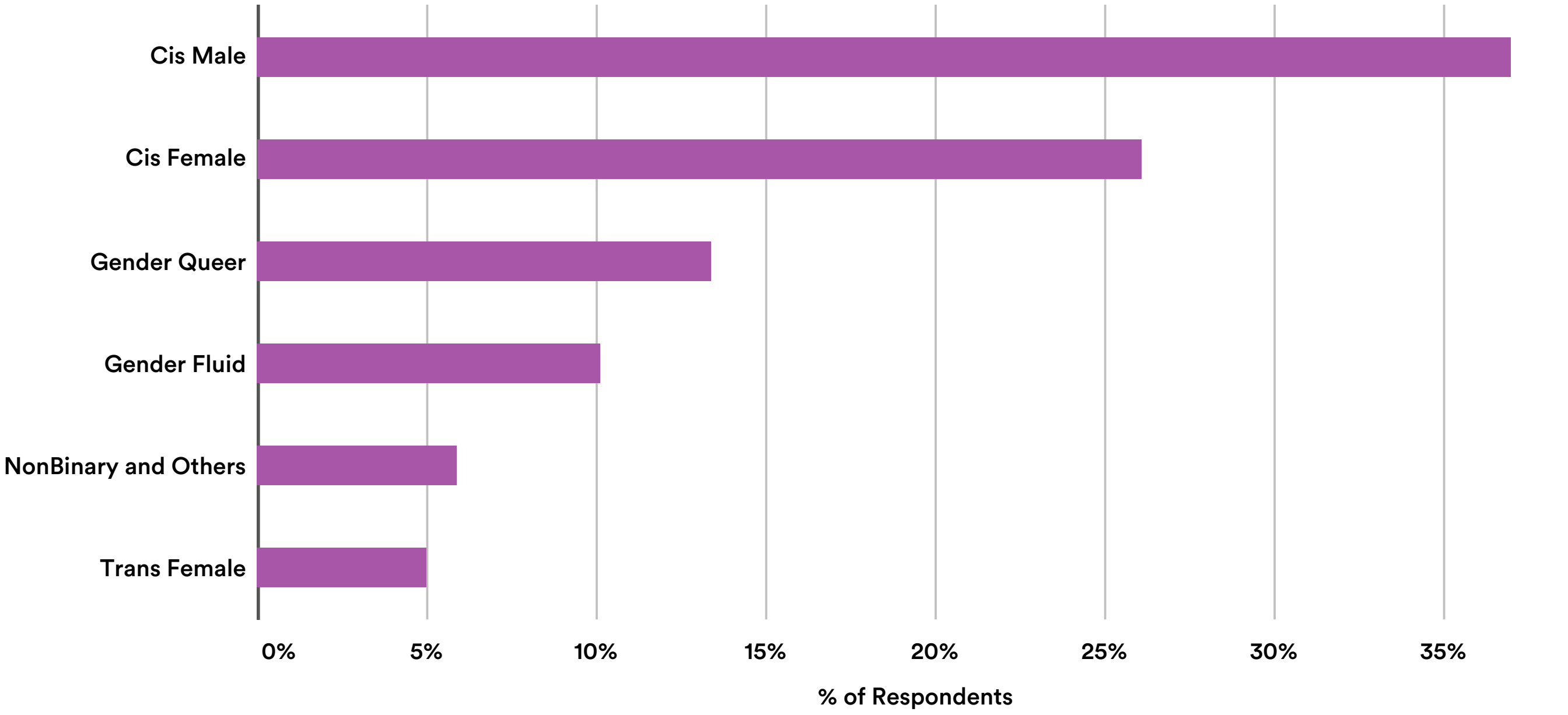

Figure 6.3.2

## QAI MEMBERSHIP SURVEY: HOW WOULD YOU DESCRIBE YOUR POSITION, 2018-20

Source: Queer in AI, 2020 | Chart: 2021 AI Index Report

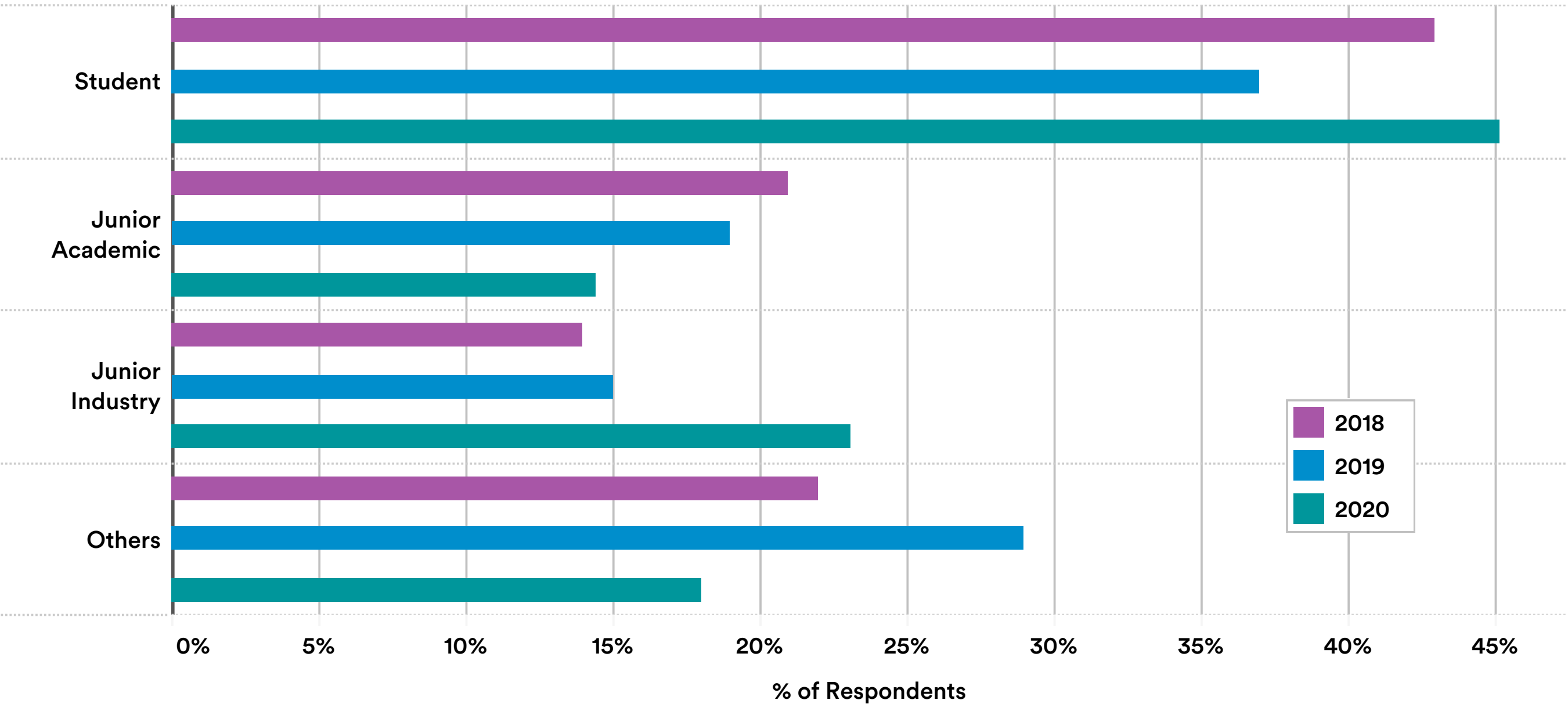

Figure 6.3.3

## Experience as Queer Practitioners

QAI also surveyed its members on their experiences as queer AI/ML practitioners. As shown in Figure 6.3.4, 81.4% regard the lack of role models as being a major obstacle for their careers, and 70.9% think the lack of community contributes to the same phenomenon. Almost half the respondents also view the lack of inclusiveness in the field as an obstacle. Moreover, more than 40% of QAI members have experienced discrimination or harassment as a queer person at work or school (Figure 6.3.5). Around 9.7% have encountered discrimination or harassment on more than five occasions.

**Among surveyed QAI members, 81.4% regard the lack of role models as being a major obstacle for their careers, and 70.9% think the lack of community contributes to the same phenomenon.**

QAI MEMBERSHIP SURVEY: WHAT ARE OBSTACLES YOU HAVE FACED in BECOMING a QUEER AI/ML PRACTITIONER, 2020
Source: Queer in AI, 2020 | Chart: 2021 AI Index Report

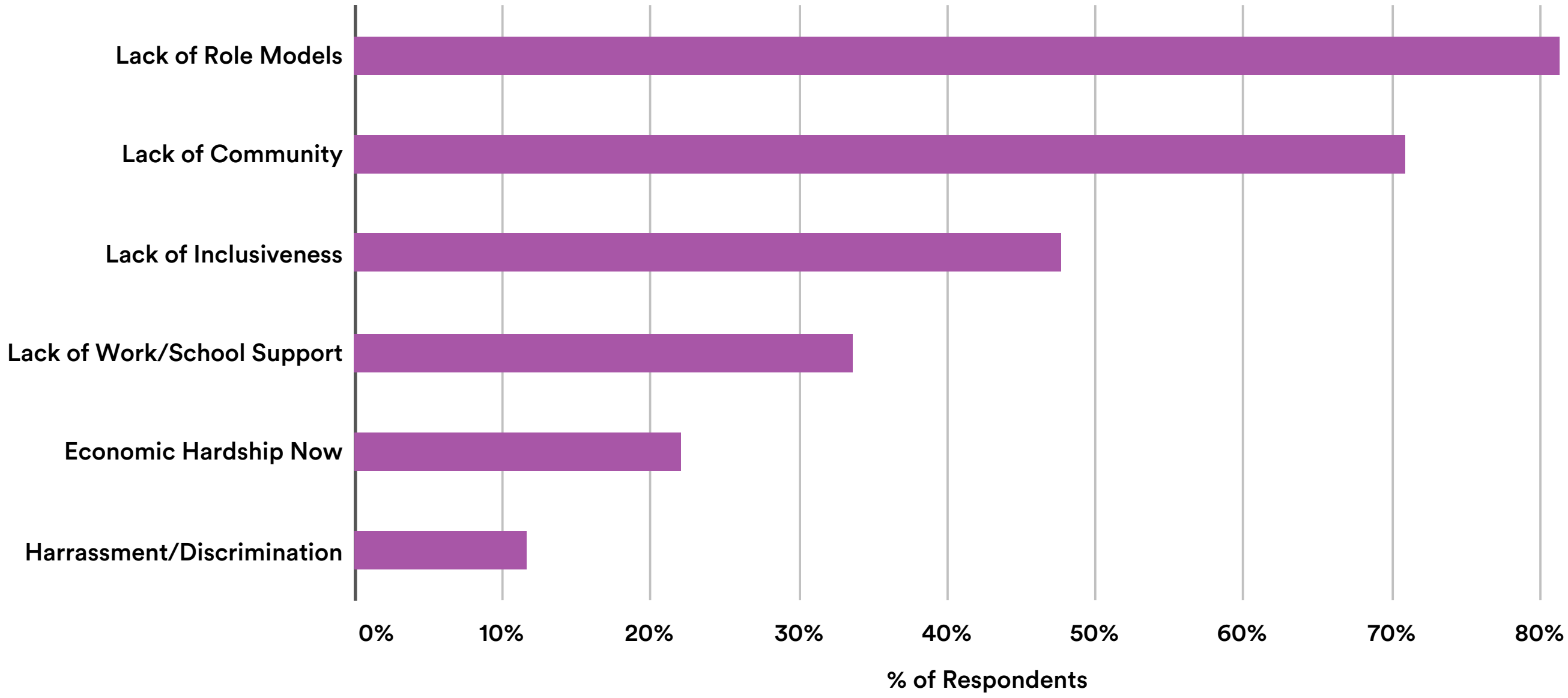

Figure 6.3.4

## More than 40% of QAI members have experienced discrimination or harassment as a queer person at work or school. Around 9.7% have encountered discrimination or harassment on more than five occasions.

QAI MEMBERSHIP SURVEY: HAVE YOU EXPERIENCED DISCRIMINATION/HARASSMENT as a QUEER PERSON at YOUR JOB or SCHOOL, 2020

Source: Queer in AI, 2020 | Chart: 2021 AI Index Report

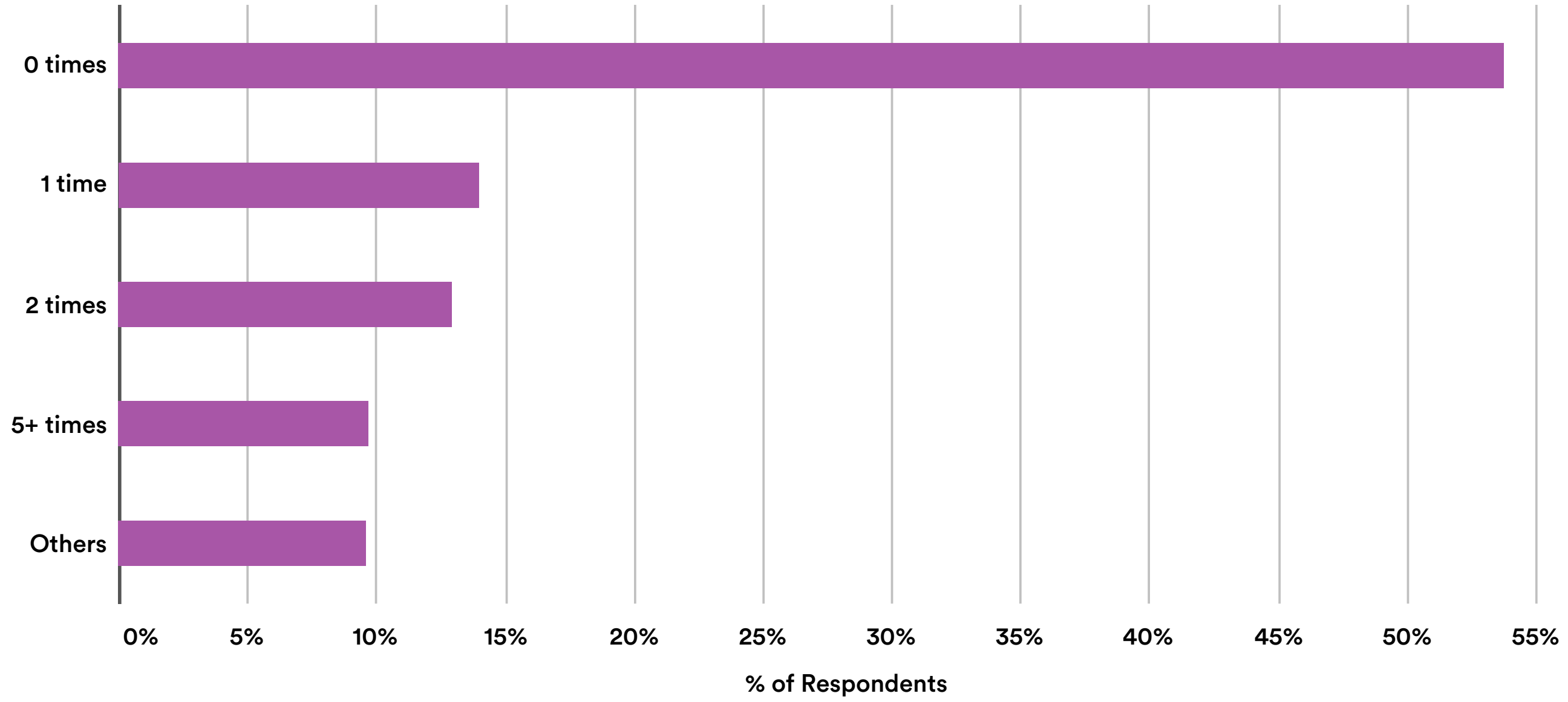

Figure 6.3.5

# CHAPTER 7:
# AI Policy and National Strategies

Artificial Intelligence
Index Report 2021

## CHAPTER 7:
# Chapter Preview

ACCESS THE PUBLIC DATA

# Overview

AI is set to shape global competitiveness over the coming decades, promising to grant early adopters a significant economic and strategic advantage. To date, national governments and regional and intergovernmental organizations have raced to put in place AI-targeted policies to maximize the promise of the technology while also addressing its social and ethical implications.

This chapter navigates the landscape of AI policymaking and tracks efforts taking place on the local, national, and international levels to help promote and govern AI technologies. It begins with an overview of national and regional AI strategies and then reviews activities on the intergovernmental level. The chapter then takes a closer look at public investment in AI in the United States as well as how legislative bodies, central banks, and nongovernmental organizations are responding to the growing need to institute a policy framework for AI technologies.

## CHAPTER HIGHLIGHTS

- Since Canada published the world's first national AI strategy in 2017, more than 30 other countries and regions have published similar documents as of December 2020.

- The launch of the Global Partnership on AI (GPAI) and Organisation for Economic Co-operation and Development (OECD) AI Policy Observatory and Network of Experts on AI in 2020 promoted intergovernmental efforts to work together to support the development of AI for all.

- In the United States, the 116th Congress was the most AI-focused congressional session in history. The number of mentions of AI by this Congress in legislation, committee reports, and Congressional Research Service (CRS) reports is more than triple that of the 115th Congress.

This section presents an overview of select national and regional AI strategies from around the world, including details on the strategies for G20 countries, Estonia, and Singapore as well as links to strategy documents for many others. Sources include websites of national or regional governments, the OECD AI Policy Observatory (OECD.AI), and news coverage. "AI strategy" is defined as a policy document that communicates the objective of supporting the development of AI while also maximizing the benefits of AI for society. Excluded are broader innovation or digital strategy documents which do not focus predominantly on AI, such as Brazil's E-Digital Strategy and Japan's Integrated Innovation Strategy.

# 7.1 NATIONAL AND REGIONAL AI STRATEGIES

To guide and foster the development of AI, countries and regions around the world are establishing strategies and initiatives to coordinate governmental and intergovernmental efforts. Since Canada published the world's first national AI strategy in 2017, more than 30 other countries and regions have published similar documents as of December 2020.

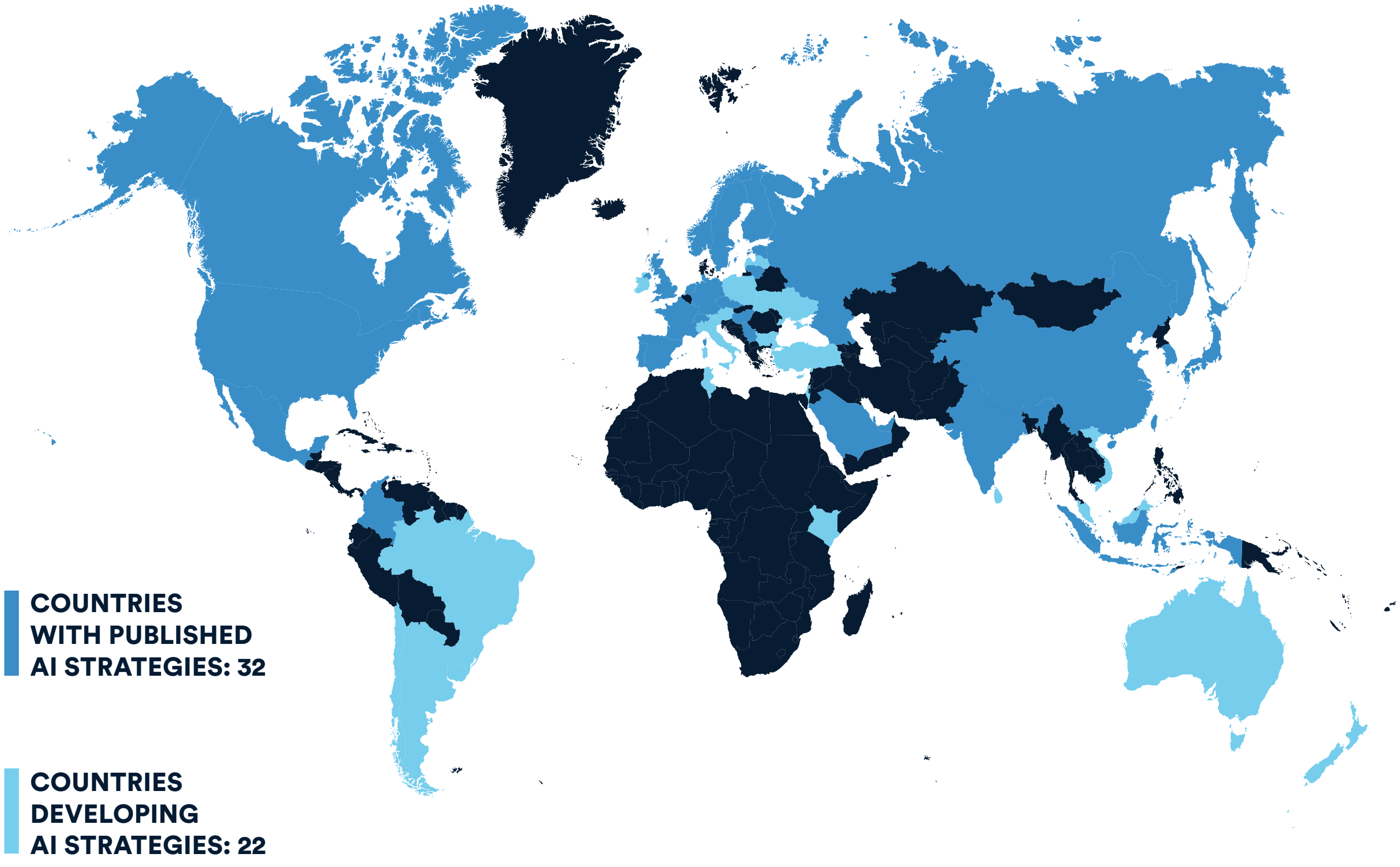

# Published Strategies

## 2017

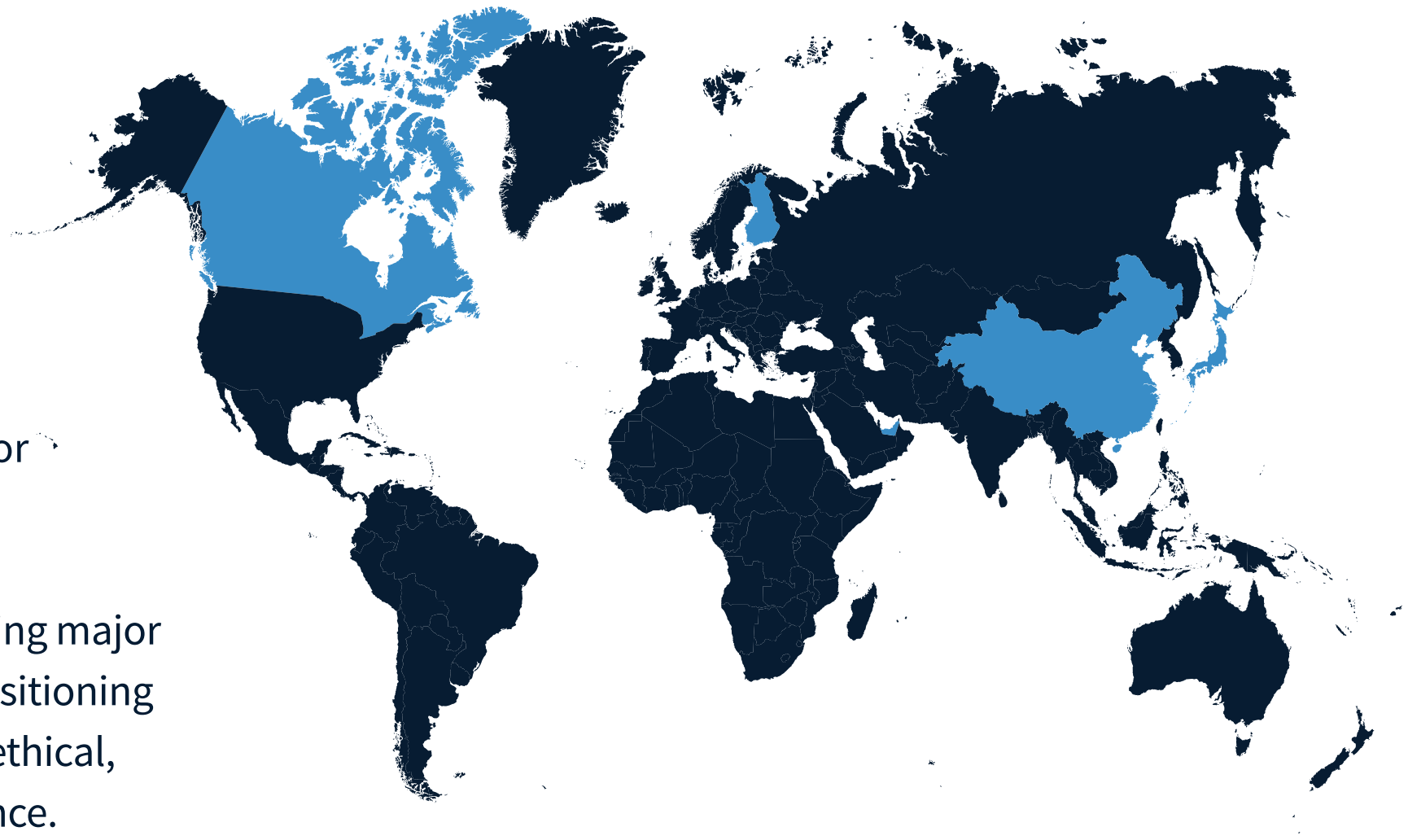

### Canada

- **AI Strategy:** Pan Canadian AI Strategy
- **Responsible Organization:** Canadian Institute for Advanced Research (CIFAR)
- **Highlights:** The Canadian strategy emphasizes developing Canada's future AI workforce, supporting major AI innovation hubs and scientific research, and positioning the country as a thought leader in the economic, ethical, policy, and legal implications of artificial intelligence.
- **Funding (December 2020 conversion rate):** CAD 125 million (USD 97 million)
- In November 2020, CIFAR published its most recent annual report, titled "AICAN," which tracks progress on implementing its national strategy, which highlighted substantial growth in Canada's AI ecosystem, as well as research and activities related to healthcare and AI's impact on society, among other outcomes of the strategy.

### China

- **AI Strategy:** A Next Generation Artificial Intelligence Development Plan
- **Responsible Organization**: State Council for the People's Republic of China
- **Highlights:** China's AI strategy is one of the most comprehensive in the world. It encompasses areas including R&D and talent development through education and skills acquisition, as well as ethical norms and implications for national security. It sets specific targets, including bringing the AI industry in line with competitors by 2020; becoming the global leader in fields such as unmanned aerial vehicles (UAVs), voice and image recognition, and others by 2025; and emerging as the primary center for AI innovation by 2030.
- **Funding:** N/A
- **Recent Updates:** China established a New Generation AI Innovation and Development Zone in February 2019 and released the "Beijing AI Principles" in May 2019 with a multi-stakeholder coalition consisting of academic institutions and private-sector players such as Tencent and Baidu.

### Japan

- **AI Strategy:** Artificial Intelligence Technology Strategy
- **Responsible Organization:** Strategic Council for AI Technology
- **Highlights:** The strategy lays out three discrete phases of AI development. The first phase focuses on the utilization of data and AI in related service industries, the second on the public use of AI and the expansion of service industries, and the third on creating an overarching ecosystem where the various domains are merged.
- **Funding:** N/A
- **Recent Updates:** In 2019, the Integrated Innovation Strategy Promotion Council launched another AI strategy, aimed at taking the next step forward in overcoming issues faced by Japan and making use of the country's strengths to open up future opportunities.

### Others

**Finland:** Finland's Age of Artificial Intelligence
**United Arab Emirates:** UAE Strategy for Artificial Intelligence

# Published Strategies

## 2018

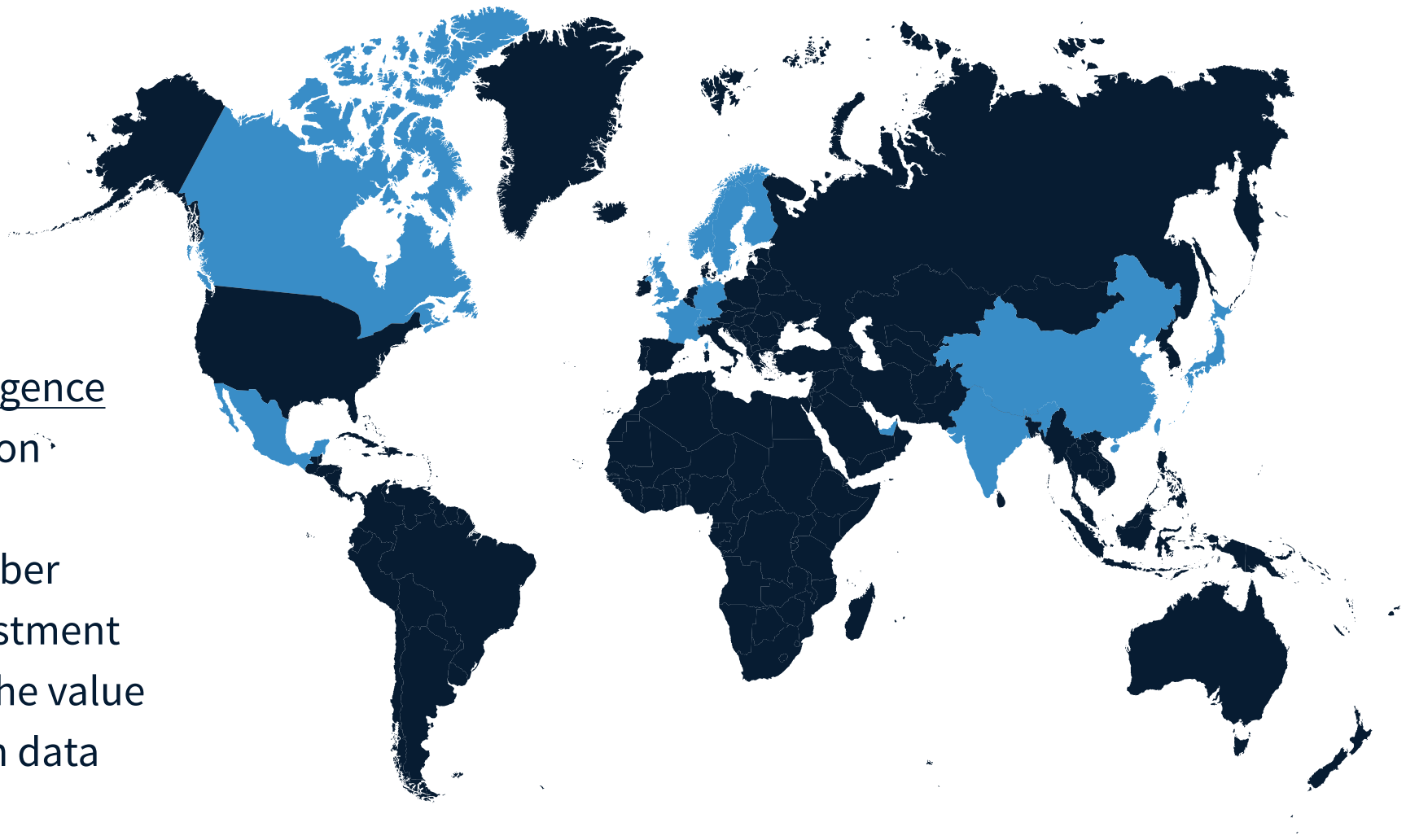

### European Union

- **AI Strategy:** Coordinated Plan on Artificial Intelligence
- **Responsible Organization:** European Commission
- **Highlights:** This strategy document outlines the commitments and actions agreed on by EU member states, Norway, and Switzerland to increase investment and build their AI talent pipeline. It emphasizes the value of public-private partnerships, creating European data spaces, and developing ethics principles.
- **Funding (December 2020 conversation rate)**: At least EUR 1 billion (USD 1.1 billion) per year for AI research and at least EUR 4.9 billion (USD 5.4 billion) for other aspects of the strategy
- **Recent updates:** A first draft of the ethics guidelines was released in June 2018, followed by an updated version in April 2019.

### France

- **AI Strategy:** AI for Humanity: French Strategy for Artificial Intelligence
- **Responsible Organizations:** Ministry for Higher Education, Research and Innovation; Ministry of Economy and Finance; Directorate General for Enterprises; Public Health Ministry; Ministry of the Armed Forces; National Research Institute for Digital Sciences; Interministerial Director of the Digital Technology and the Information and Communication System
- **Highlights:** The main themes include developing an aggressive data policy for big data; targeting four strategic sectors, namely health care, environment, transport, and defense; boosting French efforts in research and development; planning for the impact of AI on the workforce; and ensuring inclusivity and diversity within the field.
- **Funding (December 2020 conversion rate):** EUR 1.5 billion (USD 1.8 billion) up to 2022
- **Recent Updates:** The French National Research Institute for Digital Sciences (Inria) has committed to playing a central role in coordinating the national AI strategy and will report annually on its progress.

### Germany

- **AI Strategy:** AI Made in Germany
- **Responsible Organizations:** Federal Ministry of Education and Research; Federal Ministry for Economic Affairs and Energy; Federal Ministry of Labour and Social Affairs
- **Highlights:** The focus of the strategy is on cementing Germany as a research powerhouse and strengthening the value of its industries. There is also an emphasis on the public interest and working to better the lives of people and the environment.
- **Funding (December 2020 conversion rate):** EUR 500 million (USD 608 million) in the 2019 budget and EUR 3 billion (USD 3.6 billion) for the implementation up to 2025
- **Recent Updates:** In November 2019, the government published an interim progress report on the Germany AI strategy.

## 2018 (continued)

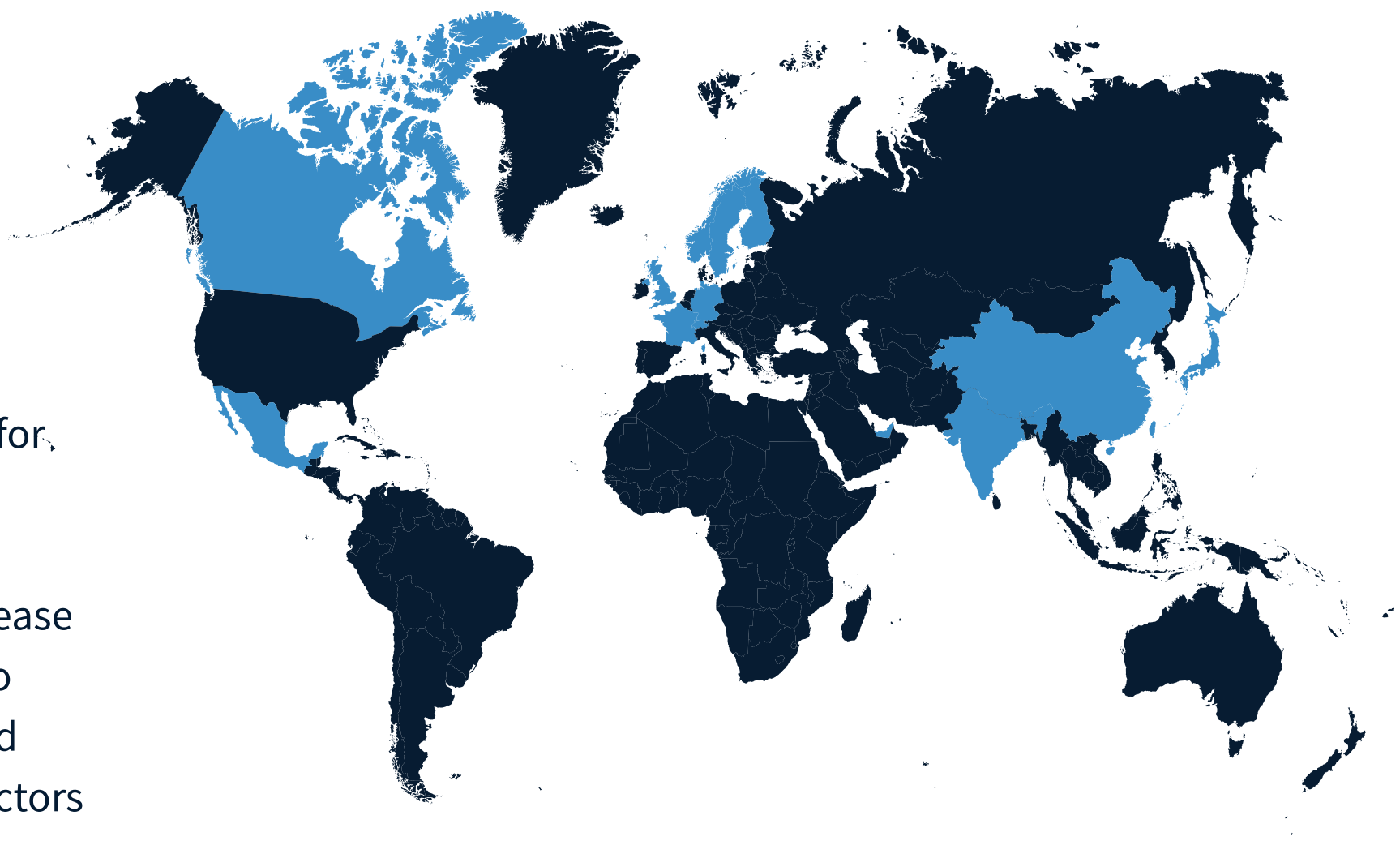

### India

- **AI Strategy:** National Strategy on Artificial Intelligence: #AIforAll
- **Responsible Organization:** National Institution for Transforming India (NITI Ayog)
- **Highlights:** The Indian strategy focuses on both economic growth and ways to leverage AI to increase social inclusion, while also promoting research to address important issues such as ethics, bias, and privacy related to AI. The strategy emphasizes sectors such as agriculture, health, and education, where public investment and government initiative are necessary.
- **Funding (December 2020 conversion rate):** INR 7000 crore (USD 949 million)
- **Recent Updates:** In 2019, the Ministry of Electronics and Information Technology released its own proposal to set up a national AI program with an allocated INR 400 crore (USD 54 million). The Indian government formed a committee in late 2019 to push for an organized AI policy and establish the precise functions of government agencies to further India's AI mission.

### Mexico

- **AI Strategy:** Artificial Intelligence Agenda MX (2019 agenda-in-brief version)
- **Responsible Organization:** IA2030Mx, Economía
- **Highlights:** As Latin America's first strategy, the Mexican strategy focuses on developing a strong governance framework, mapping the needs of AI in various industries, and identifying governmental best practices with an emphasis on developing Mexico's AI leadership.
- **Funding:** N/A
- **Recent Updates:** According to the Inter-American Development Bank's recent fAIr LAC report, Mexico is in the process of establishing concrete AI policies to further implementation.

### United Kingdom

- **AI Strategy:** Industrial Strategy: Artificial Intelligence Sector Deal
- **Responsible Organization:** Office for Artificial Intelligence (OAI)
- **Highlights:** The U.K. strategy emphasizes a strong partnership between business, academia, and the government and identifies five foundations for a successful industrial strategy: becoming the world's most innovative economy, creating jobs and better earnings potential, infrastructure upgrades, favorable business conditions, and building prosperous communities throughout the country.
- **Funding (December 2020 conversion rate):** GBP 950 million (USD 1.3 billion)
- **Recent Updates:** Between 2017 and 2019, the U.K.'s Select Committee on AI released an annual report on the country's progress. In November 2020, the government announced a major increase in defense spending of GBP 16.5 billion (USD 21.8 billion) over four years, with a major emphasis on AI technologies that promise to revolutionize warfare.

### Others

**Sweden:** National Approach to Artificial Intelligence

**Taiwan:** Taiwan AI Action Plan

# Published Strategies

## 2019

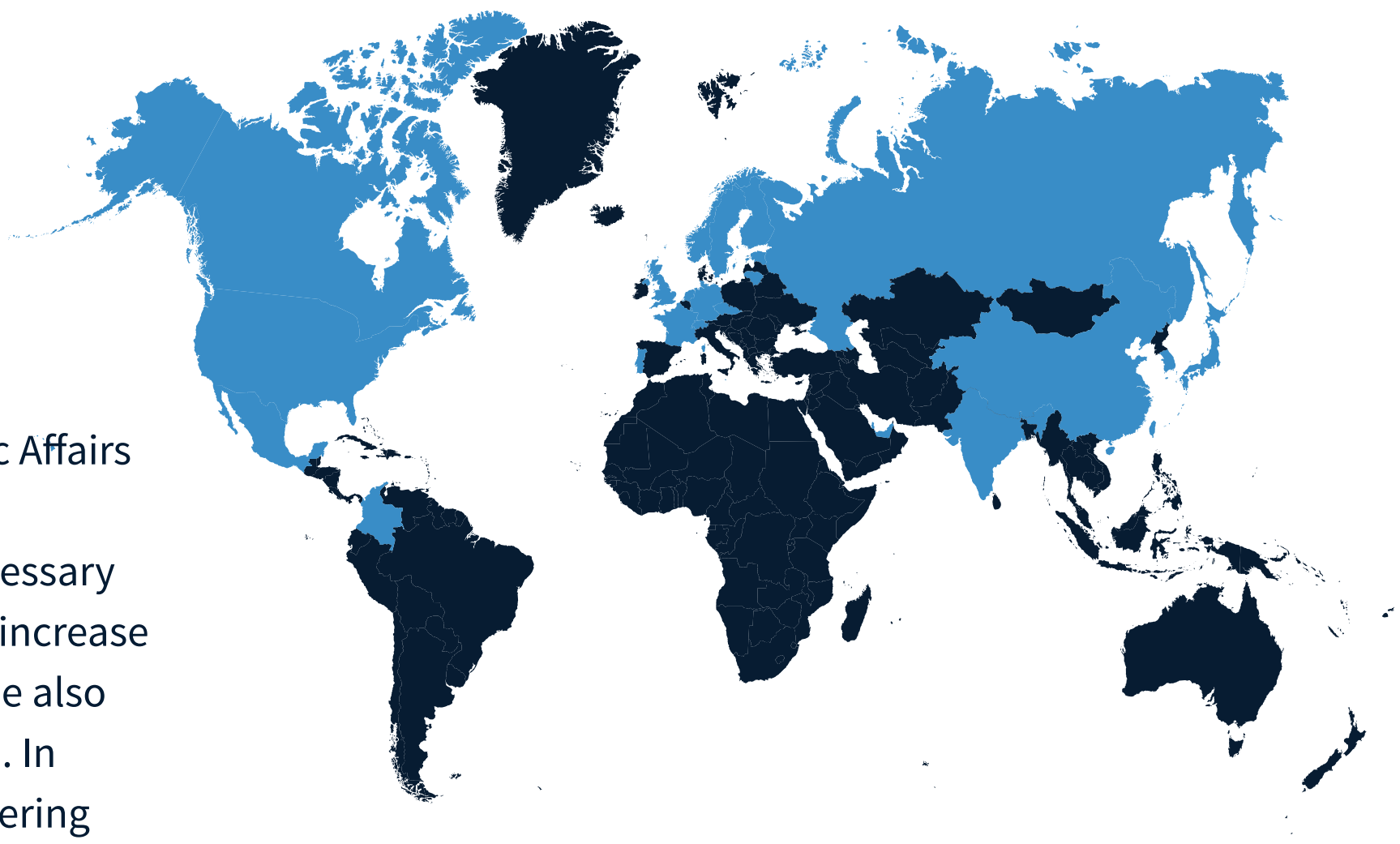

### Estonia

- **AI Strategy:** National AI Strategy 2019–2021
- **Responsible Organization:** Ministry of Economic Affairs and Communications (MKM)
- **Highlights:** The strategy emphasizes actions necessary for both the public and private sectors to take to increase investment in AI research and development, while also improving the legal environment for AI in Estonia. In addition, it hammers out the framework for a steering committee that will oversee the implementation and monitoring of the strategy.
- **Funding (December 2020 conversion rate):** EUR 10 million (USD 12 million) up to 2021
- **Recent Updates:** The Estonian government released an update on the AI taskforce in May 2019.

### Russia

- **AI Strategy:** National Strategy for the Development of Artificial Intelligence
- **Responsible Organizations:** Ministry of Digital Development, Communications and Mass Media; Government of the Russian Federation
- **Highlights:** The Russian AI strategy places a strong emphasis on its national interests and lays down guidelines for the development of an "information society" between 2017 and 2030. These include a national technology initiative, departmental projects for federal executive bodies, and programs such as the Digital Economy of the Russian Federation, designed to implement the AI framework across sectors.
- **Funding:** N/A
- **Recent Updates:** In December 2020, Russian president Vladimir Putin took part in the Artificial Intelligence Journey Conference, where he presented four ideas for AI policies: establishing experimental legal frameworks for the use of AI, developing practical measures to introduce AI algorithms, providing neural network developers with competitive access to big data, and boosting private investment in domestic AI industries.

### Singapore

- **AI Strategy:** National Artificial Intelligence Strategy
- **Responsible Organization:** Smart Nation and Digital Government Office (SNDGO)
- **Highlights:** Launched by Smart Nation Singapore, a government agency that seeks to transform Singapore's economy and usher in a new digital age, the strategy identifies five national AI projects in the following fields: transport and logistics, smart cities and estates, health care, education, and safety and security.
- **Funding (December 2020 conversion rate):** While the 2019 strategy does not mention funding, in 2017 the government launched its national program, AI Singapore, with a pledge to invest SGD 150 million (USD 113 million) over five years.
- **Recent Updates:** In November 2020, SNDGO published its inaugural annual update on the Singaporean government's data protection efforts. It describes the measures taken to date to strengthen public sector data security and to safeguard citizens' private data.

## 2019 (continued)

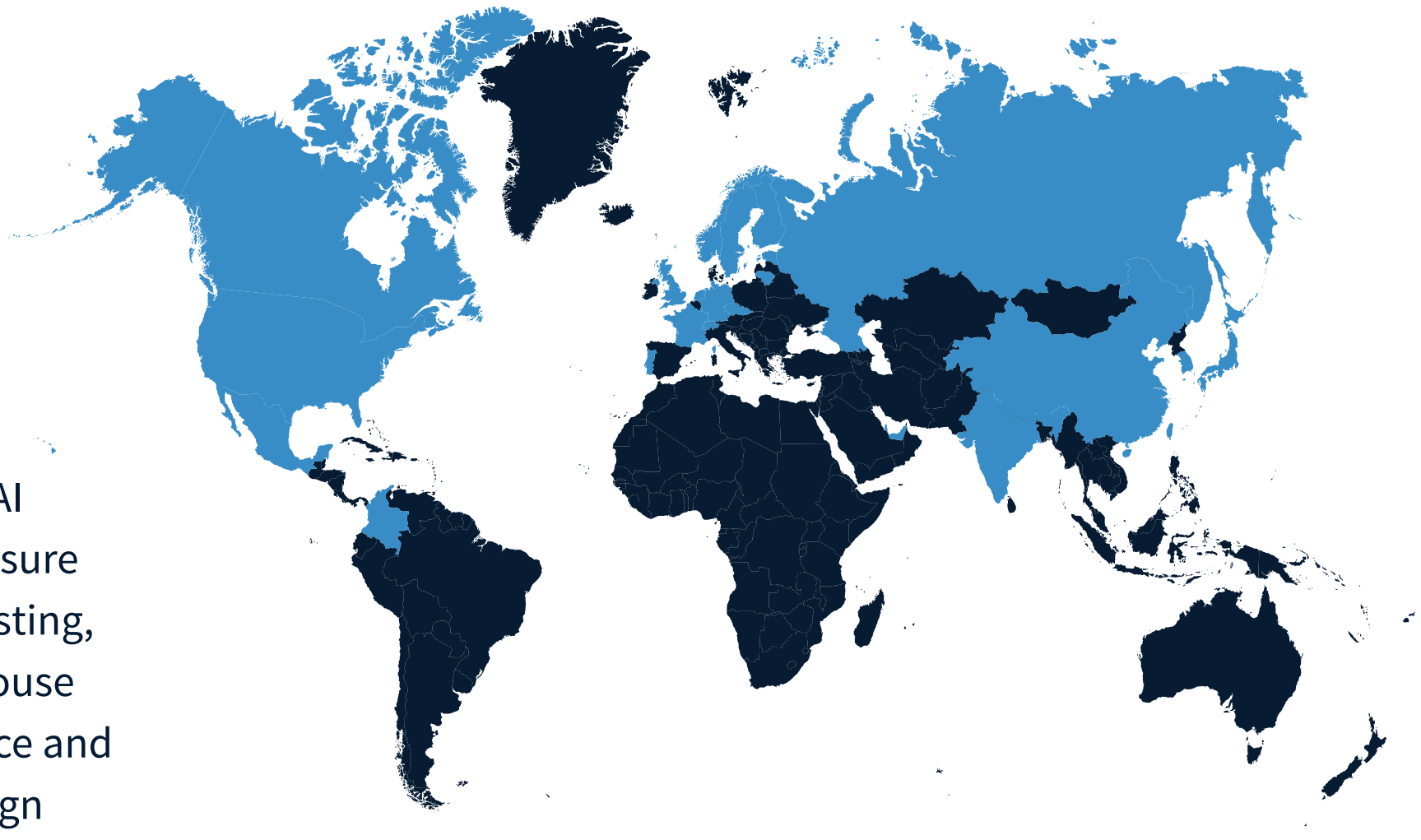

### United States

- **AI Strategy**: American AI Initiative
- **Responsible Organization:** The White House
- **Highlights:** The American AI Initiative prioritizes the need for the federal government to invest in AI R&D, reduce barriers to federal resources, and ensure technical standards for the safe development, testing, and deployment of AI technologies. The White House also emphasizes developing an AI-ready workforce and signals a commitment to collaborating with foreign partners while promoting U.S. leadership in AI. The initiative, however, lacks specifics on the program's timeline, whether additional research will be dedicated to AI development, and other practical considerations.
- **Funding:** N/A
- **Recent Updates:** The U.S. government released its year one annual report in February 2020, followed in November by the first guidance memorandum for federal agencies on regulating artificial intelligence applications in the private sector, including principles that encourage AI innovation and growth and increase public trust and confidence in AI technologies. The National Defense Authorization Act (NDAA) for Fiscal Year 2021 called for a National AI Initiative to coordinate AI research and policy across the federal government.

### South Korea

- **AI Strategy:** National Strategy for Artificial Intelligence
- **Responsible Organization:** Ministry of Science, ICT and Future Planning (MSIP)
- **Highlights:** The Korean strategy calls for plans to facilitate the use of AI by businesses and to streamline regulations to create a more favorable environment for the development and use of AI and other new industries. The Korean government also plans to leverage its dominance in the global supply of memory chips to build the next generation of smart chips by 2030.
- **Funding (December 2020 conversion rate):** KRW 2.2 trillion (USD 2 billion)
- **Recent Updates:** N/A

### Others

**Colombia:** National Policy for Digital Transformation and Artificial Intelligence
**Czech Republic:** National Artificial Intelligence Strategy of the Czech Republic
**Lithuania:** Lithuanian Artificial Intelligence Strategy: A Vision for the Future
**Luxembourg:** Artificial Intelligence: A Strategic Vision for Luxembourg
**Malta:** Malta: The Ultimate AI Launchpad
**Netherlands:** Strategic Action Plan for Artificial Intelligence
**Portugal:** AI Portugal 2030
**Qatar:** National Artificial Intelligence for Qatar

# Published Strategies

## 2020

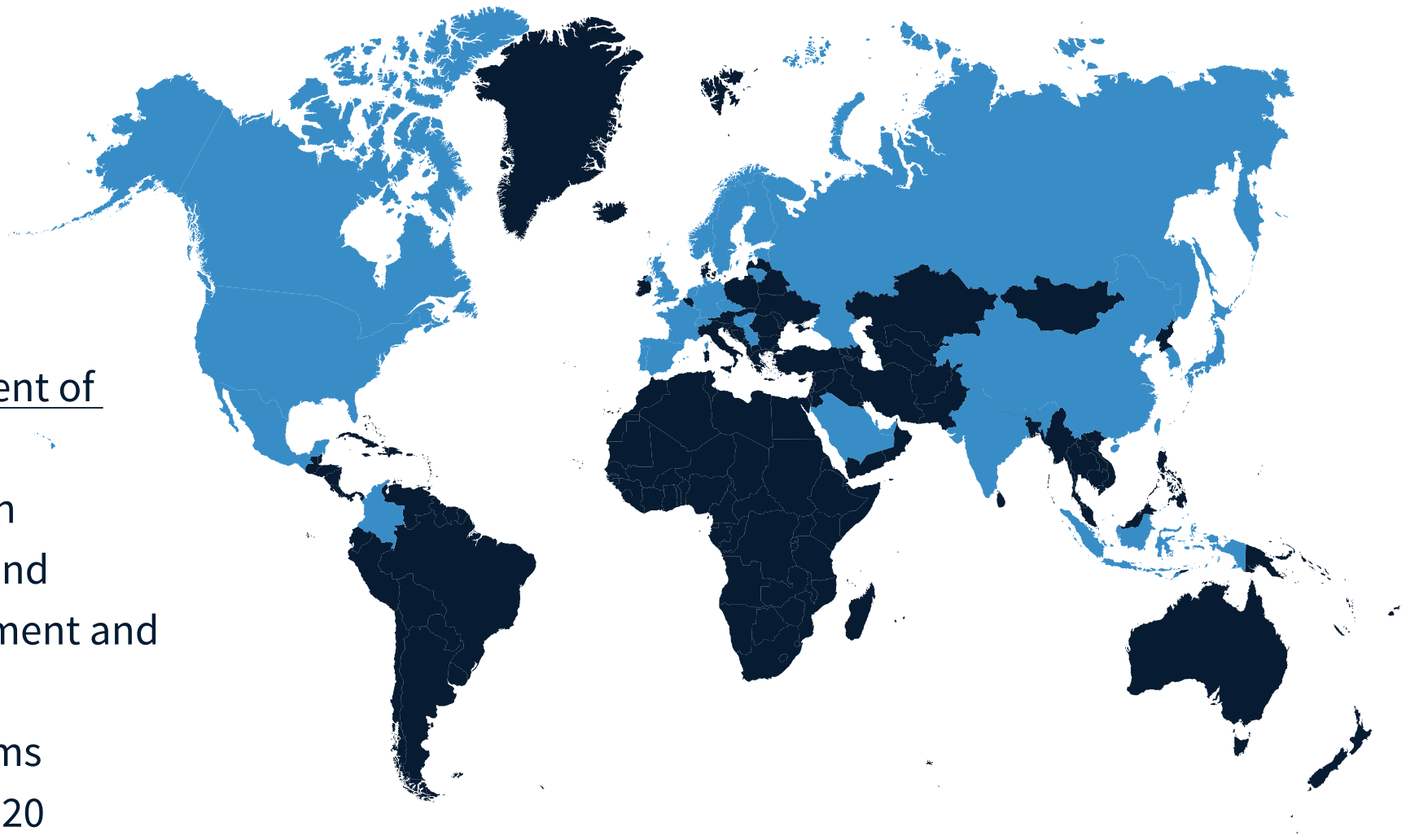

### Indonesia

- **AI Strategy:** National Strategy for the Development of Artificial Intelligence (Stranas KA)
- **Responsible Organizations:** Ministry of Research and Technology (Menristek), National Research and Innovation Agency (BRIN), Agency for the Assessment and Application of Technology (BPPT)
- **Strategy Highlights:** The Indonesian strategy aims to guide the country in developing AI between 2020 and 2045. It focuses on education and research, health services, food security, mobility, smart cities, and public sector reform.
- **Funding:** N/A
- **Recent Updates:** None

### Saudi Arabia

- **AI Strategy:** National Strategy on Data and AI (NSDAI)
- **Responsible Organization:** Saudi Data and Artificial Intelligence Authority (SDAIA)
- **Highlights:** As part of an effort to diversify the country's economy away from oil and boost the private sector, the NSDAI aims to accelerate AI development in five critical sectors: health care, mobility, education, government, and energy. By 2030, Saudi Arabia intends to train 20,000 data and AI specialists, attract USD 20 billion in foreign and local investment, and create an environment that will attract at least 300 AI and data startups.
- **Funding:** N/A
- **Recent Updates:** During the summit where the Saudi government released its strategy, the country's National Center for Artificial Intelligence (NCAI) signed collaboration agreements with China's Huawei and Alibaba Cloud to design AI-related Arabic-language systems.

### Others

**Hungary:** Hungary's Artificial Intelligence Strategy
**Norway:** National Strategy for Artificial Intelligence
**Serbia:** Strategy for the Development of Artificial Intelligence in the Republic of Serbia for the Period 2020–2025
**Spain:** National Artificial Intelligence Strategy

# Strategies in Development
## (AS OF DECEMBER 2020)

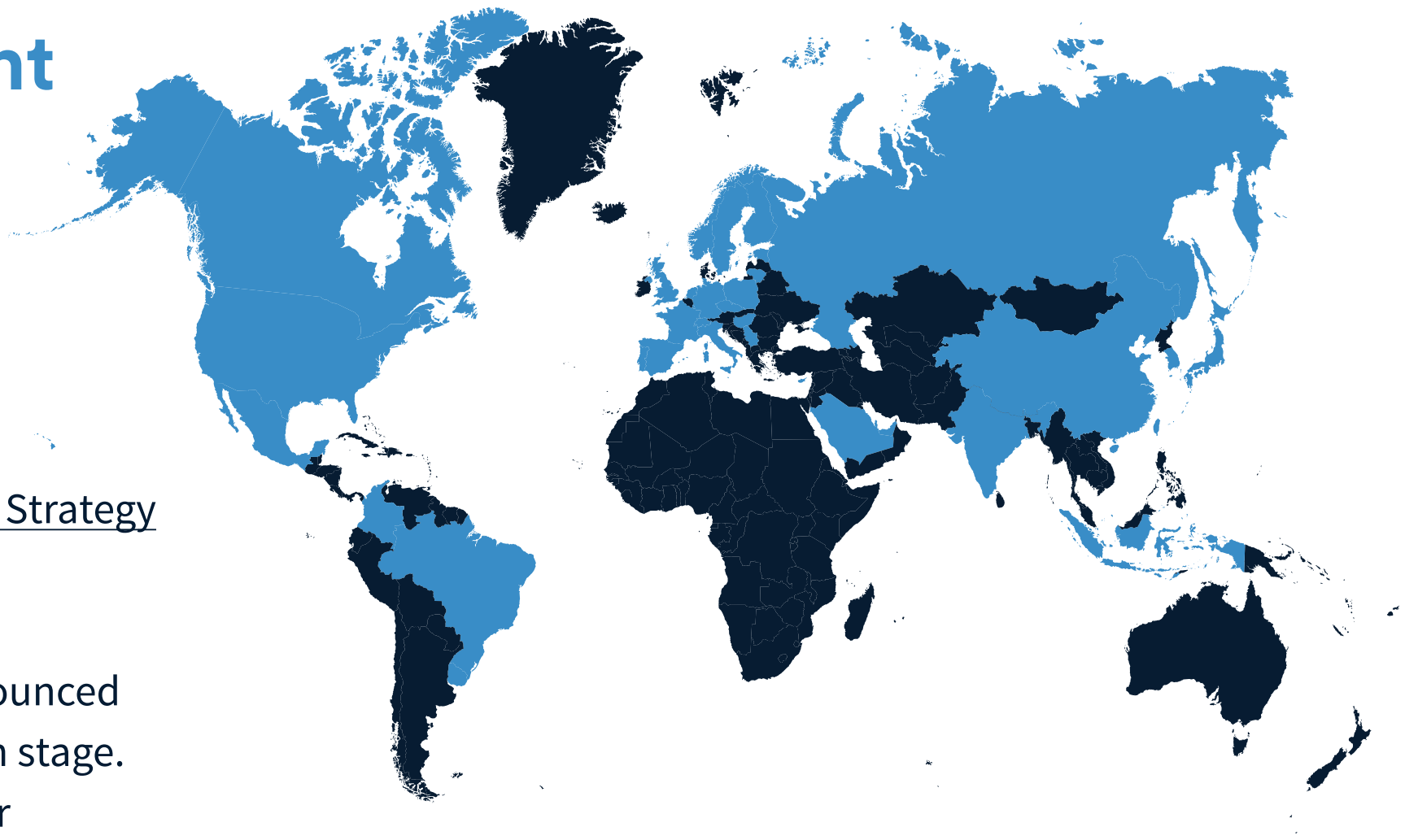

### Strategies in Public Consultation

#### Brazil

- **AI Strategy Draft:** Brazilian Artificial Intelligence Strategy
- **Responsible Organization:** Ministry of Science, Technology and Innovation (MCTI)
- **Highlights:** Brazil's national AI strategy was announced in 2019 and is currently in the public consultation stage. According to the OECD, the strategy aims to cover relevant topics bearing on AI, including its impact on the economy, ethics, development, education, and jobs, and to coordinate specific public policies addressing such issues.
- **Funding:** N/A
- **Recent Updates:** In October 2020, the country's largest research facility dedicated to AI was launched in collaboration with IBM, the University of São Paulo, and the São Paulo Research Foundation.

#### Italy

- **AI Strategy Draft:** Proposal for an Italian Strategy for Artificial Intelligence
- **Responsible Organization:** Ministry of Economic Development (MISE)
- **Highlights:** This document provides the proposed strategy for the sustainable development of AI, aimed at improving Italy's competitiveness in AI. It focuses on improving AI-based skills and competencies, fostering AI research, establishing a regulatory and ethical framework to ensure a sustainable ecosystem for AI, and developing a robust data infrastructure to fuel these developments.
- **Funding (December 2020 conversion rate):** EUR 1 billion (USD 1.1 billion) through 2025 and expected matching funds from the private sector, bringing the total investment to EUR 2 billion.
- **Recent Updates:** None

#### Others

**Cyprus:** National Strategy for Artificial Intelligence
**Ireland:** National Irish Strategy on Artificial Intelligence
**Poland:** Artificial Intelligence Development Policy in Poland
**Uruguay:** Artificial Intelligence Strategy for Digital Government

# Strategies Announced

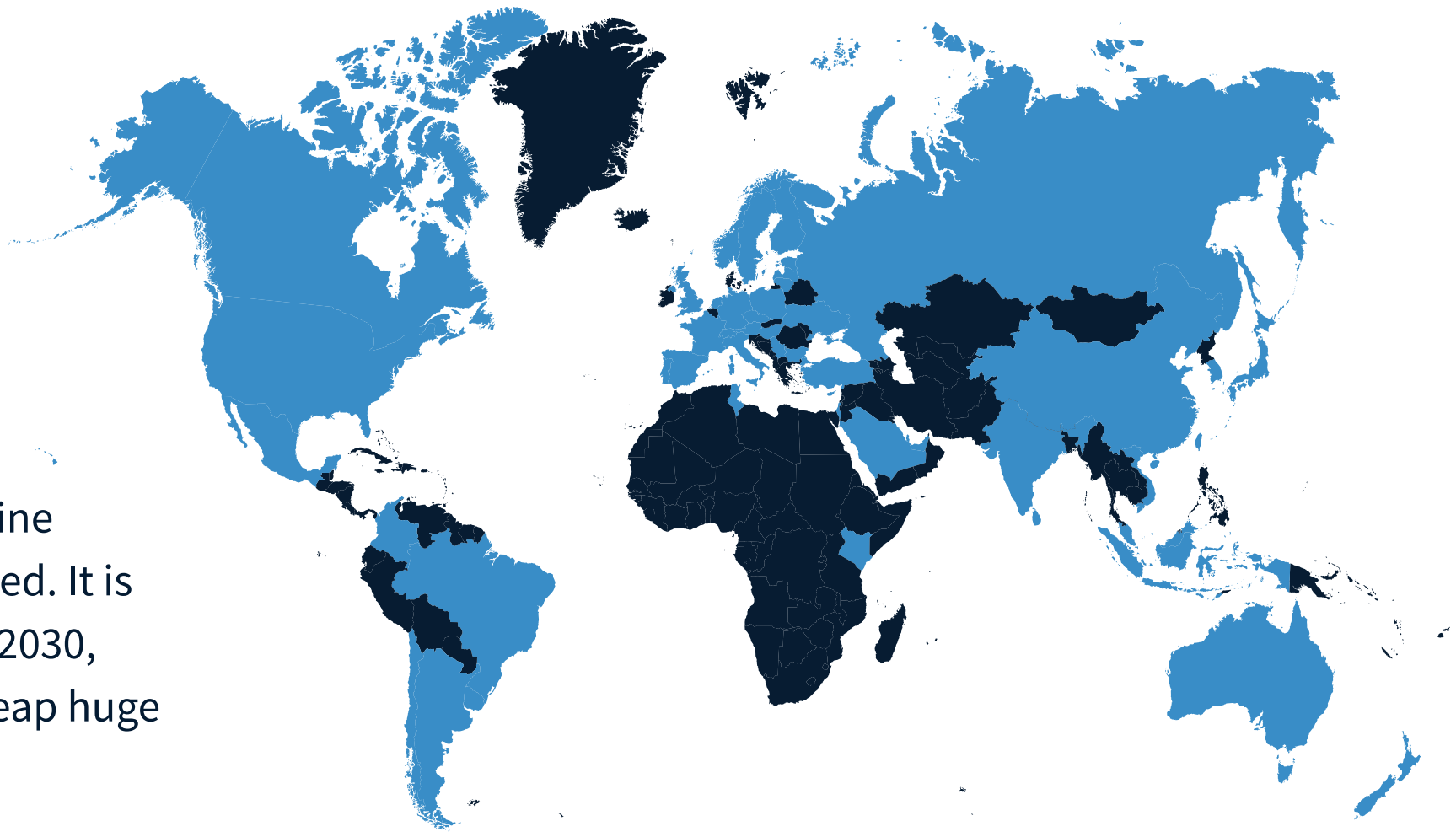

## Argentina

- **Related Document:** N/A
- **Responsible Organization:** Ministry of Science, Technology and Productive Innovation (MINCYT)
- **Status:** Argentina's AI plan is a part of the Argentine Digital Agenda 2030 but has not yet been published. It is intended to cover the decade between 2020 and 2030, and reports indicate that it has the potential to reap huge benefits for the agricultural sector.

## Australia

- **Related Documents:** Artificial Intelligence Roadmap / An AI Action Plan for all Australians
- **Responsible Organizations:** Commonwealth Scientific and Industrial Research Organisation (CSIRO), Data 61, and the Australian government
- **Status:** The Australian government published a road map in 2019 (in collaboration with the national science agency, CSIRO) and a discussion paper of an AI action plan in 2020 as frameworks to develop a national AI strategy. In its 2018–19 budget, the Australian government earmarked AUD 29.9 million (USD 22.2 million [December 2020 conversation rate]) over four years to strengthen the country's capabilities in AI and machine learning (ML). In addition, CSIRO published a research paper on Australia's AI Ethics Framework in 2019 and launched a public consultation, which is expected to produce a forthcoming strategy document.

## Turkey

- **Related Document:** N/A
- **Responsible Organizations:** Presidency of the Republic of Turkey Digital Transformation Office; Ministry of Industry and Technology; Scientific and Technological Research Council of Turkey; Science, Technology and Innovation Policies Council
- **Status:** The strategy has been announced but not yet published. According to media sources, it will focus on talent development, scientific research, ethics and inclusion, and digital infrastructure.

## Others

**Austria:** Artificial Intelligence Mission Austria (official report)
**Bulgaria:** Concept for the Development of Artificial Intelligence in Bulgaria Until 2030 (concept document)
**Chile:** National AI Policy (official announcement)
**Israel:** National AI Plan (news article)
**Kenya:** Blockchain and Artificial Intelligence Taskforce (news article)
**Latvia:** On the Development of Artificial Intelligence Solutions (official report)
**Malaysia:** National Artificial Intelligence (Al) Framework (news article)
**New Zealand:** Artificial Intelligence: Shaping a Future New Zealand (official report)
**Sri Lanka:** Framework for Artificial Intelligence (news article)
**Switzerland:** Artificial Intelligence (official guidelines)
**Tunisia:** National Artificial Intelligence Strategy (task force announced)
**Ukraine:** Concept of Artificial Intelligence Development in Ukraine AI (concept document)
**Vietnam:** Artificial Intelligence Development Strategy (official announcement)

**Read more on AI national strategies:**

- Tim Dutton: An Overview of National AI Strategies
- Organisation for Economic Co-operation and Development: OECD AI Policy Observatory
- Canadian Institute for Advanced Research: Building an AI World, Second Edition
- Inter-American Development Bank: Artificial Intelligence for Social Good in Latin America and the Caribbean: The Regional Landscape and 12 Country Snapshots

## National AI Strategies and Human Rights

In 2020, Global Partners Digital and Stanford's Global Digital Policy Incubator published a report examining governments' national AI strategies from a human rights perspective, titled "National Artificial Intelligence Strategies and Human Rights: A Review." The report assesses the extent to which governments and regional organizations have incorporated human rights considerations into their national AI strategies and made recommendations to policymakers looking to develop or review AI strategies in the future.

The report found that among the 30 states and two regional strategies (from the European Union and the Nordic-Baltic states), a number of strategies refer to the impact of AI on human rights, with the right to privacy as the most commonly mentioned, followed by equality and nondiscrimination (Table 6.1.1). However, very few strategy documents provide deep analysis or concrete assessment of the impact of AI applications on human rights. Specifics as to how and the depth to which human rights should be protected in the context of AI is largely missing, in contrast to the level of specificity on other issues such as economic competitiveness and innovation advantage.

**Table 7.1.1: Mapping human rights referenced in national AI strategies**

| HUMAN RIGHTS MENTIONED | STATES/REGIONAL ORGANIZATIONS |
|---|---|
| The right to privacy | Australia, Belgium, China, Czech Republic, Germany, India, Italy, Luxembourg, Malta, Netherlands, Norway, Portugal, Qatar, South Korea, United States |
| The right to equality/ nondiscrimination | Australia, Belgium, Czech Republic, Denmark, Estonia, EU, France, Germany, Italy, Malta, Netherlands, Norway |
| The right to an effective remedy | Australia (responsibility and ability to hold humans responsible), Denmark, Malta, Netherlands |
| The rights to freedom of thought, expression, and access to information | France, Netherlands, Russia |
| The right to work | France, Russia |

# 7.2 INTERNATIONAL COLLABORATION ON AI

Given the scale of the opportunities and the challenges presented by AI, a number of international efforts have recently been announced that aim to develop multilateral AI strategies. This section provides an overview of those international initiatives from governments committed to working together to support the development of AI for all.

These multilateral initiatives on AI suggest that organizations are taking a variety of approaches to tackle the practical applications of AI and scale those solutions for maximum global impact. Many countries turn to international organizations for global AI norm formulation, while others engage in partnerships or bilateral agreements. Among the topics under discussion, the ethics of AI—or the ethical challenges raised by current and future applications of AI—stands out as a particular focus area for intergovernmental efforts.

Countries such as Japan, South Korea, the United Kingdom, the United States, and members of the European Union are active participants of intergovernmental efforts on AI. A major AI powerhouse, China, on the other hand, has opted to engage in a number of science and technology bilateral agreements that stress cooperation on AI as part of the Digital Silk Road under the Belt and Road (BRI) initiative framework. For example, AI is mentioned in China's economic cooperation under the BRI Initiative with the United Arab Emirates.

## INTERGOVERNMENTAL INITIATIVES

Intergovernmental working groups consist of experts and policymakers from member states who study and report on the most urgent challenges related to developing and deploying AI and then make recommendations based on their findings. These groups are instrumental in identifying and developing strategies for the most pressing issues in AI technologies and their applications.

### Working Groups

**Global Partnership on AI (GPAI)**

- **Participants:** Australia, Brazil, Canada, France, Germany, India, Italy, Japan, Mexico, the Netherlands, New Zealand, South Korea, Poland, Singapore, Slovenia, Spain, the United Kingdom, the United States, and the European Union (as of December 2020)
- **Host of Secretariat:** OECD
- **Focus Areas:** Responsible AI; data governance; the future of work; innovation and commercialization
- **Recent Activities:** Two International Centres of Expertise—the International Centre of Expertise in Montreal for the Advancement of Artificial Intelligence and the French National Institute for Research in Digital Science and Technology (INRIA) in Paris—are supporting the work in the four focus areas and held the Montreal Summit 2020 in December 2020. Moreover, the data governance working group published the beta version of the group's framework in November 2020.

**OECD Network of Experts on AI (ONE AI)**

- **Participants:** OECD countries
- **Host:** OECD
- **Focus Areas:** Classification of AI; implementing trustworthy AI; policies for AI; AI compute
- **Recent Activities:** ONE AI convened its first meeting in February 2020, when it also launched the OECD AI Policy Observatory. In November 2020, the working group on the classification of AI presented the first look at an AI classification framework based on OECD's definition of AI divided into four dimensions (context, data and input, AI model, task and output) that aims to guide policymakers in designing adequate policies for each type of AI system.

**High-Level Expert Group on Artificial Intelligence (HLEG)**

- **Participants:** EU countries
- **Host:** European Commission
- **Focus Areas:** Ethics guidelines for trustworthy AI
- **Recent Activities:** Since its launch at the recommendation

of the EU AI strategy in 2018, HLEG presented the EU Ethics Guidelines for Trustworthy Artificial Intelligence and a series of policy and investment recommendations, as well as an assessment checklist related to the guidelines.

**Ad Hoc Expert Group (AHEG) for the Recommendation on the Ethics of Artificial Intelligence**

- **Participants:** United Nations Educational, Scientific and Cultural Organization (UNESCO) member states
- **Host:** UNESCO
- **Focus Areas:** Ethical issues raised by the development and use of AI
- **Recent Activities:** The AHEG produced a revised first draft Recommendation on the Ethics of Artificial Intelligence, which was transmitted in September 2020 to Member States of UNESCO for their comments by December 31, 2020.

### Summits and Meetings

**AI for Good Global Summit**

- **Participants:** Global (with the United Nations and its agencies)
- **Hosts:** International Telecommunication Union, XPRIZE Foundation
- **Focus Areas:** Trusted, safe, and inclusive development of AI technologies and equitable access to their benefits

**AI Partnership for Defense**

- **Participants:** Australia, Canada, Denmark, Estonia, Finland, France, Israel, Japan, Norway, South Korea, Sweden, the United Kingdom, and the United States
- **Hosts:** Joint Artificial Intelligence Center, U.S. Department of Defense
- **Focus Areas:** AI ethical principles for defense

**China-Association of Southeast Asian Nations (ASEAN) AI Summit**

- **Participants:** Brunei, Cambodia, China, Indonesia, Laos, Malaysia, Myanmar, the Philippines, Singapore, Thailand, and Vietnam
- **Hosts:** China Association for Science and Technology, Guangxi Zhuang Autonomous Region, China
- **Focus Areas:** Infrastructure construction, digital economy, and innovation-driven development

## BILATERAL AGREEMENTS

Bilateral agreements focusing on AI are another form of international collaboration that has been gaining in popularity in recent years. AI is usually included in the broader context of collaborating on the development of digital economies, though India stands apart for investing in developing multiple bilateral agreements specifically geared toward AI.

**India and United Arab Emirates**

Invest India and the UAE Ministry of Artificial Intelligence signed a memorandum of understanding in **July 2018** to collaborate on fostering innovative AI ecosystems and other policy concerns related to AI. Two countries will convene a working committee aimed at increasing investment in AI startups and research activities in partnership with the private sector.

**India and Germany**

It was reported in **October 2019** that India and Germany likely will sign an agreement including partnerships on the use of artificial intelligence (especially in farming).

**United States and United Kingdom**

The U.S. and the U.K. announced a declaration in **September 2020**, through the Special Relationship Economic Working Group, that the two countries will enter into a bilateral dialogue on advancing AI in line with shared democratic values and further cooperation in AI R&D efforts.

**India and Japan**

India and Japan were said to have finalized an agreement in **October 2020** that focuses on collaborating on digital technologies, including 5G and AI.

**French and Germany**

France and Germany signed a road map for a Franco-German Research and Innovation Network on artificial intelligence as part of the Declaration of Toulouse in **October 2019** to advance European efforts in the development and application of AI, taking into account ethical guidelines.

This section examines public investment in AI in the United States based on data from the U.S. Networking and Information Technology Research and Development (NITRD) program and Bloomberg Government.

# 7.3 U.S. PUBLIC INVESTMENT IN AI

## FEDERAL BUDGET FOR NON-DEFENSE AI R&D

In September 2019, the White House National Science and Technology Council released a report attempting to total up all public-sector AI R&D funding, the first time such a figure was published. This funding is to be disbursed as grants for government laboratories or research universities or in the form of government contracts. These federal budget figures, however, do not include substantial AI R&D investments by the Department of Defense (DOD) and the intelligence sector, as they were withheld from publication for national security reasons.

As shown in Figure 7.3.1, federal civilian agencies—those agencies that are not part of the DOD or the intelligence sector—allocated USD 973.5 million to AI R&D for FY 2020, a figure that rose to USD 1.1 billion once congressional appropriations and transfers were factored in. For FY 2021, federal civilian agencies budgeted USD 1.5 billion, which is almost 55% higher than its 2020 request.

U.S. FEDERAL BUDGET for NON-DEFENSE AI R&D, FY 2020-21

Source: U.S. NITRD Program, 2020 | Chart: 2021 AI Index Report

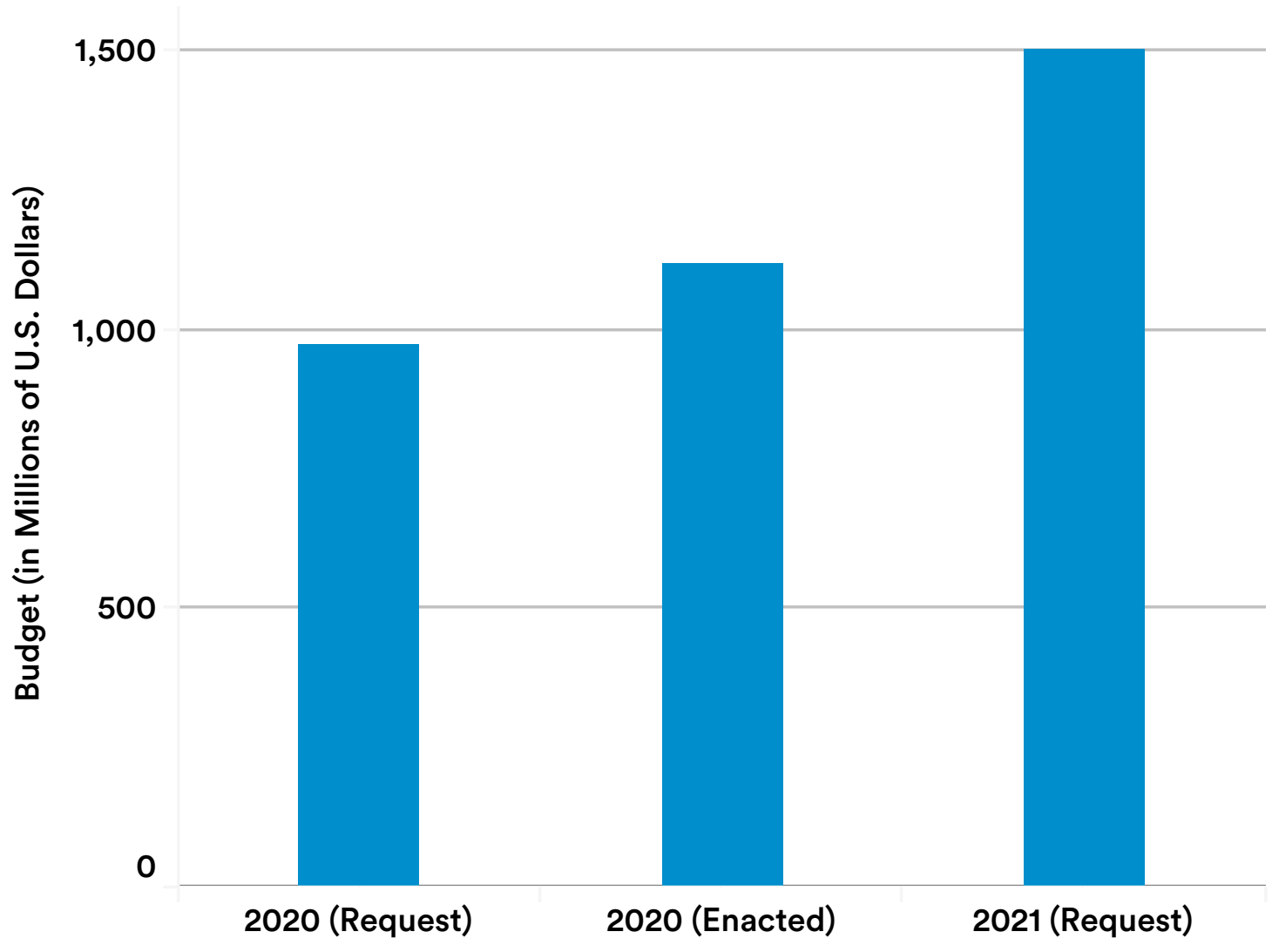

Figure 7.3.1

## U.S. DEPARTMENT OF DEFENSE AI R&D BUDGET REQUEST

While the official DOD budget is not publicly available, Bloomberg Government has analyzed the department's publicly available budget request for research, development, test, and evaluation (RDT&E)— data that sheds light on its spending on AI R&D.

With 305 unclassified DOD R&D programs specifying the use of AI or ML technologies, the combined U.S. military budget for AI R&D in FY 2021 is USD 5.0 billion (Figure 7.3.2). This figure appears consistent with the USD 5.0 billion enacted the previous year. However, the FY 2021 figure reflects a budget request, rather than a final enacted budget. As noted above, once congressional appropriations are factored in, the true level of funding available to DOD AI R&D programs in FY 2021 may rise substantially.

The top five projects set to receive the highest amount of AI R&D investment in FY 2021:

- Rapid Capability Development and Maturation, by the U.S. Army (USD 284.2 million)
- Counter WMD Technologies and Capabilities Development, by the DOD Threat Reduction Agency (USD 265.2 million)
- Algorithmic Warfare Cross-Functional Team (Project Maven), by the Office of the Secretary of Defense (USD 250.1 million)
- Joint Artificial Intelligence Center (JAIC), by the Defense Information Systems Agency (USD 132.1 million)
- High Performance Computing Modernization Program, by the U.S. Army (USD 99.6 million)

In addition, the Defense Advanced Research Projects Agency (DARPA) alone is investing USD 568.4 million in AI R&D, an increase of USD 82 million from FY 2020.

**U.S. DOD BUDGET for AI-SPECIFIC RESEARCH DEVELOPMENT, TEST, and EVALUATION (RDT&E), FY 2018-20**
Sources: Bloomberg Government & U.S. Department of Defense, 2020 | Chart: 2021 AI Index Report

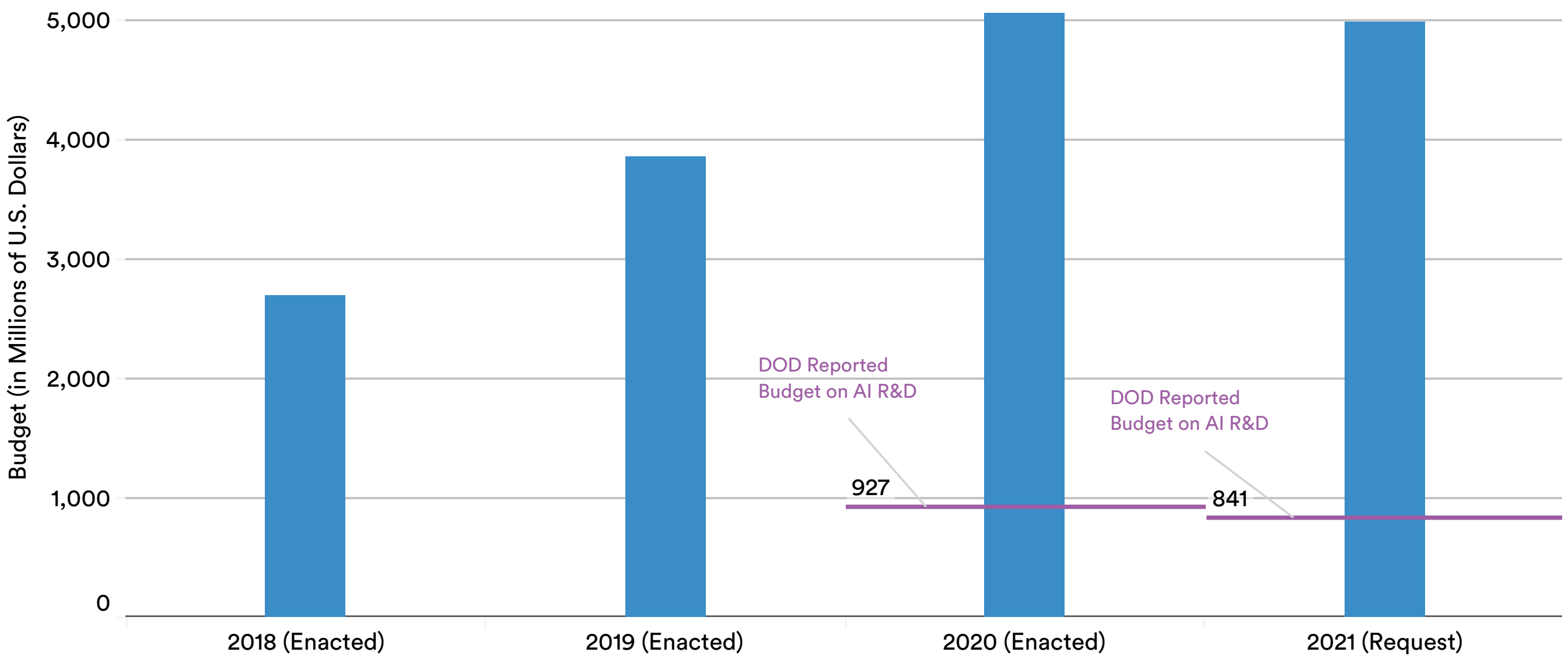

Figure 7.3.2

**Important data caveat:** This chart illustrates the challenge of working with contemporary government data sources to understand spending on AI. By one measure—the requests that include AI-relevant keywords—the DOD is requesting more than USD 5 billion for AI-specific research development in 2021 . However, DOD's own accounting produces a radically smaller number: USD 841 million. This relates to the issue of defining where an AI system ends and another system begins; for instance, an initiative that uses AI for drones may also count hardware-related expenditures for the drones within its "AI" budget request, though the AI software component will be much smaller.

## U.S. GOVERNMENT AI-RELATED CONTRACT SPENDING

Another indicator of public investment in AI technologies is the level of spending on government contracts across the federal government. Contracting for products and services supplied by private businesses typically occupies the largest share of an agency's budget. Bloomberg Government built a model that captures contract spending on AI technologies by adding up all contracting transactions that contain a set of more than 100 AI-specific keywords in their titles or descriptions. The data reveals that the amount the federal government spends on contracts for AI products and services has reached an all-time high and shows no sign of slowing down. However, note that during the procurement process, vendors may add a bunch of keywords into their applications, so some of these things may have a relatively small AI component relative to other parts of technology.

### Total Contract Spending

Federal departments and agencies spent a combined USD 1.8 billion on unclassified AI-related contracts in FY 2020. This represents a more than 25% increase from the USD 1.5 billion agencies spent in FY 2019 (Figure 7.3.3). AI spending in 2020 was more than six times higher than what it was just five years ago—about USD 300 million in FY 2015. However, to put this in perspective, the federal government spent USD 682 billion on contracts in FY 2020, so AI currently represents 0.25% of government spending.

### Contract Spending by Department and Agency

Figure 7.3.4 shows that in FY 2020, the DOD spent more on AI-related contracts than any other federal department or agency (USD 1.4 billion). In second and third place are NASA (USD 139.1 million) and the Department of Homeland Security (USD 112.3 million). DOD, NASA, and the Department of Health and Human Services top the list for the most contract spending on AI over the past 10 years combined (Figure 7.3.5). In fact, DOD's total contract spending on AI from 2001 to 2020 (USD 3.9 billion) is more than what was spent by the other 44 departments and agencies combined (USD 2.9 billion) over the same period.

Looking ahead, DOD spending on AI contracts is only expected to grow as the Pentagon's Joint Artificial Intelligence Center (JAIC), established in June 2018, is

U.S. GOVERNMENT TOTAL CONTRACT SPENDING on AI, FY 2001-20
Source: Bloomberg Government, 2020 | Chart: 2021 AI Index Report

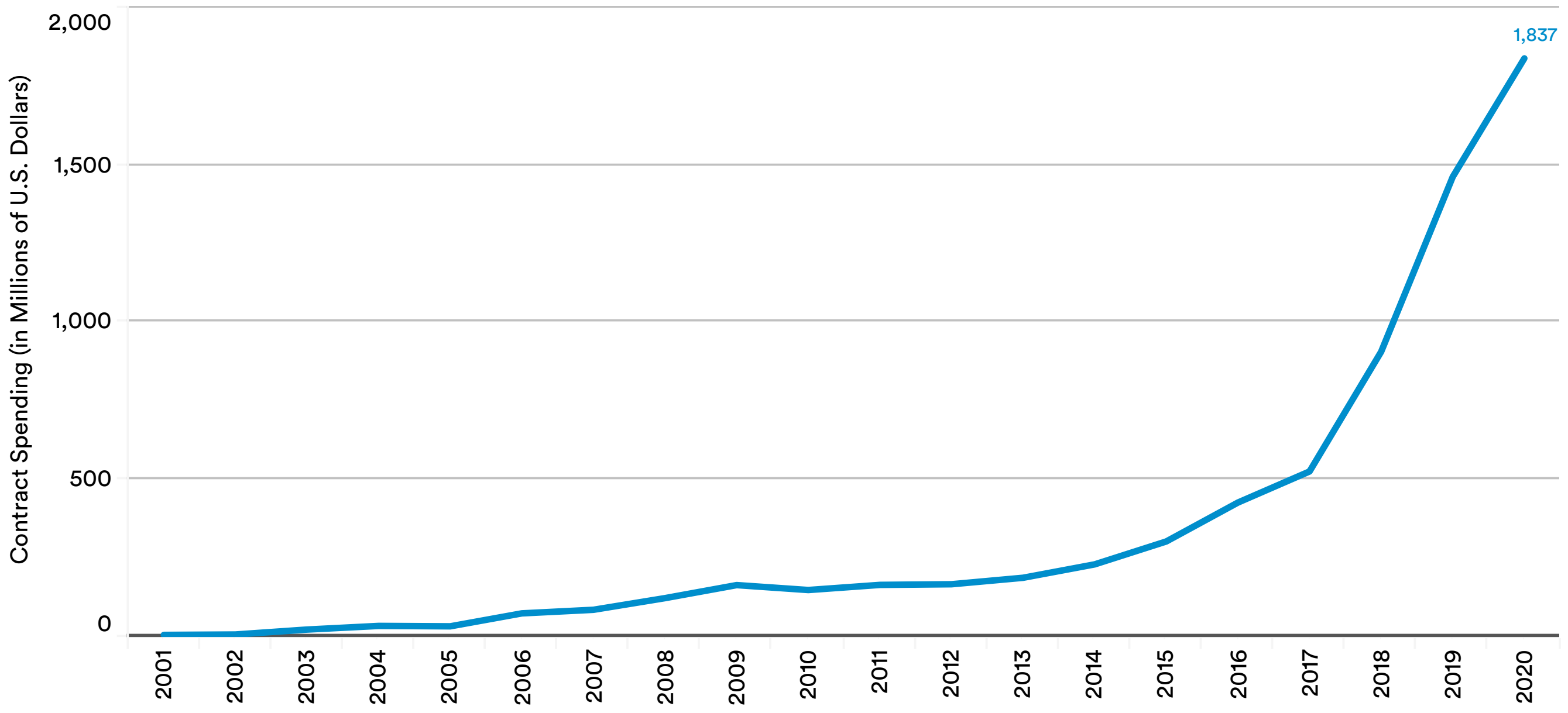

Figure 7.3.3

still in the early stages of driving DOD's AI spending. In 2020, JAIC awarded two massive contracts, one to Booz Allen Hamilton for the five-year, USD 800 million Joint Warfighter program, and another to Deloitte Consulting for a four-year, USD 106 million enterprise cloud environment for the JAIC, known as the Joint Common Foundation.

**TOP 10 CONTRACT SPENDING on AI by U.S. GOVERNMENT DEPARTMENT and AGENCY, 2020**
Source: Bloomberg Government, 2020 | Chart: 2021 AI Index Report

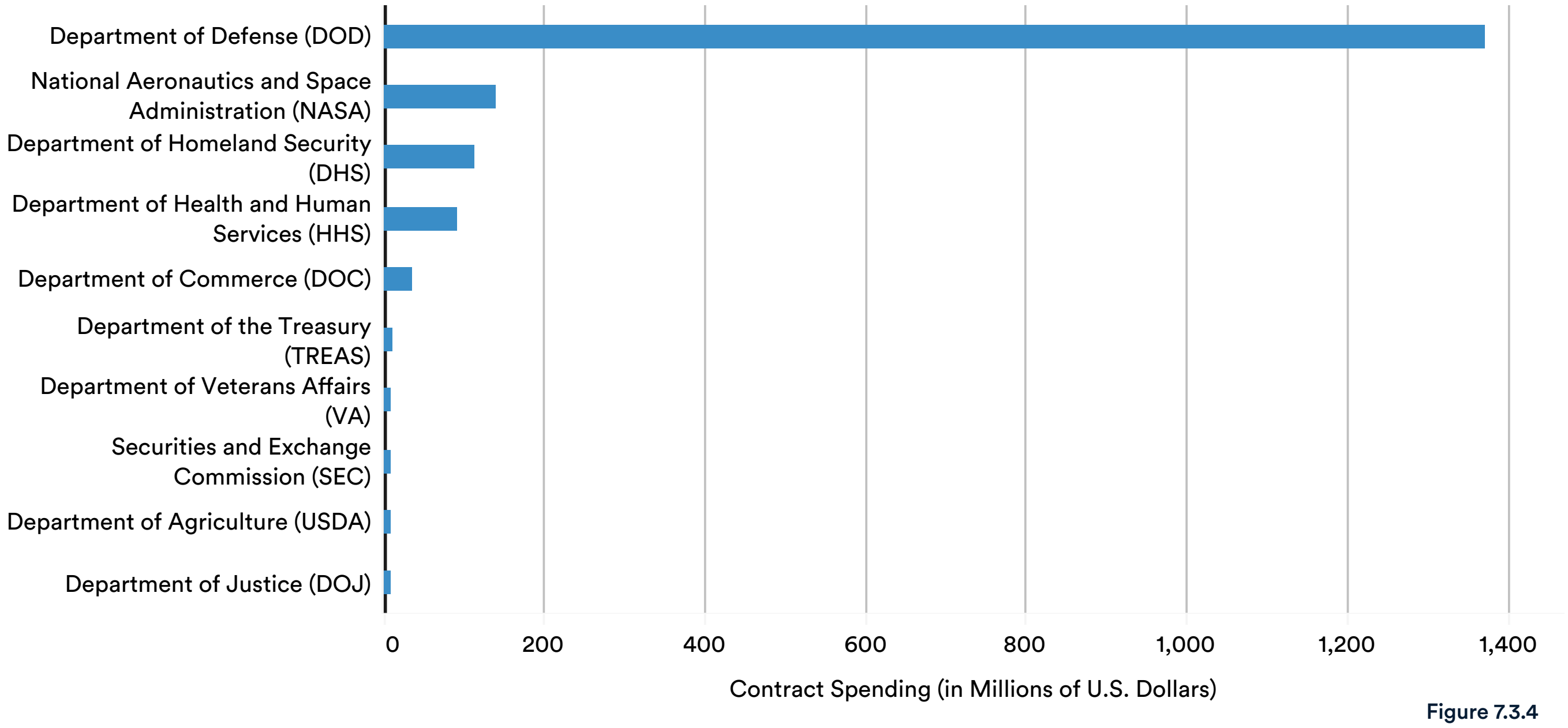

Figure 7.3.4

**TOP 10 CONTRACT SPENDING on AI by U.S. GOVERNMENT DEPARTMENT and AGENCY, 2001-20 (SUM)**
Source: Bloomberg Government, 2020 | Chart: 2021 AI Index Report

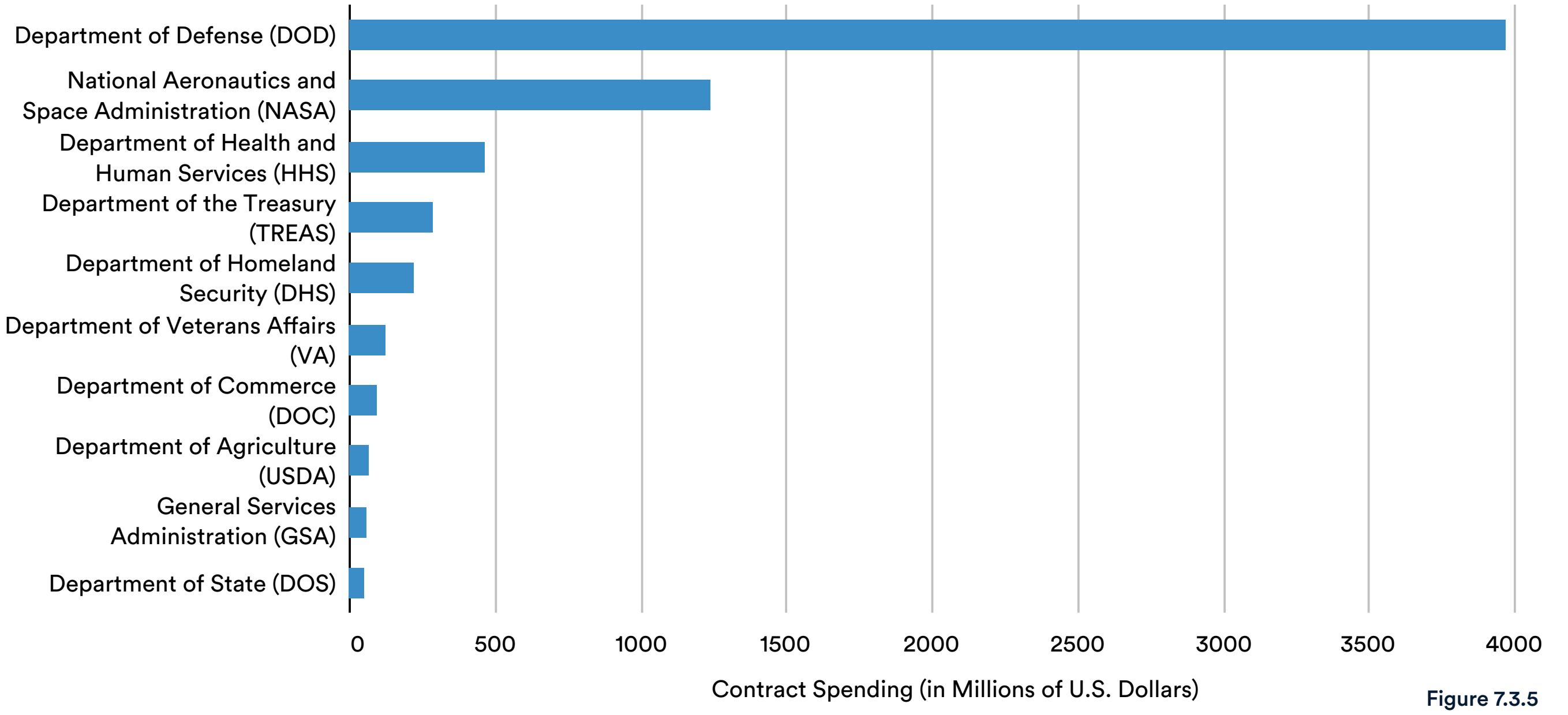

Figure 7.3.5

# 7.4 AI AND POLICYMAKING

As AI gains attention and importance, policies and initiatives related to the technology are becoming higher priorities for governments, private companies, technical organizations, and civil society. This section examines how three of these four are setting the agenda for AI policymaking, including the legislative and monetary authority of national governments, as well as think tanks, civil society, and the technology and consultancy industry.

## LEGISLATION RECORDS ON AI

The number of congressional and parliamentary records on AI is an indicator of governmental interest in developing AI capabilities—and legislating issues pertaining to AI. In this section, we use data from Bloomberg and McKinsey & Company to ascertain the number of these records and how that number has evolved in the last 10 years.

Bloomberg Government identified all legislation (passed or introduced), reports published by congressional committees, and CRS reports that referenced one or more AI-specific keywords. McKinsey & Company searched for the terms "artificial intelligence" and "machine learning" on the websites of the U.S. Congressional Record, the U.K. Parliament, and the Parliament of Canada. For the United States, each count indicates that AI or ML was mentioned during a particular event contained in the Congressional Record, including the reading of a bill; for the U.K. and Canada, each count indicates that AI or ML was mentioned in a particular comment or remark during the proceedings.[1]

MENTIONS of AI in U.S. CONGRESSIONAL RECORD by LEGISLATIVE SESSION, 2001-20
Source: Bloomberg Government, 2020 | Chart: 2021 AI Index Report

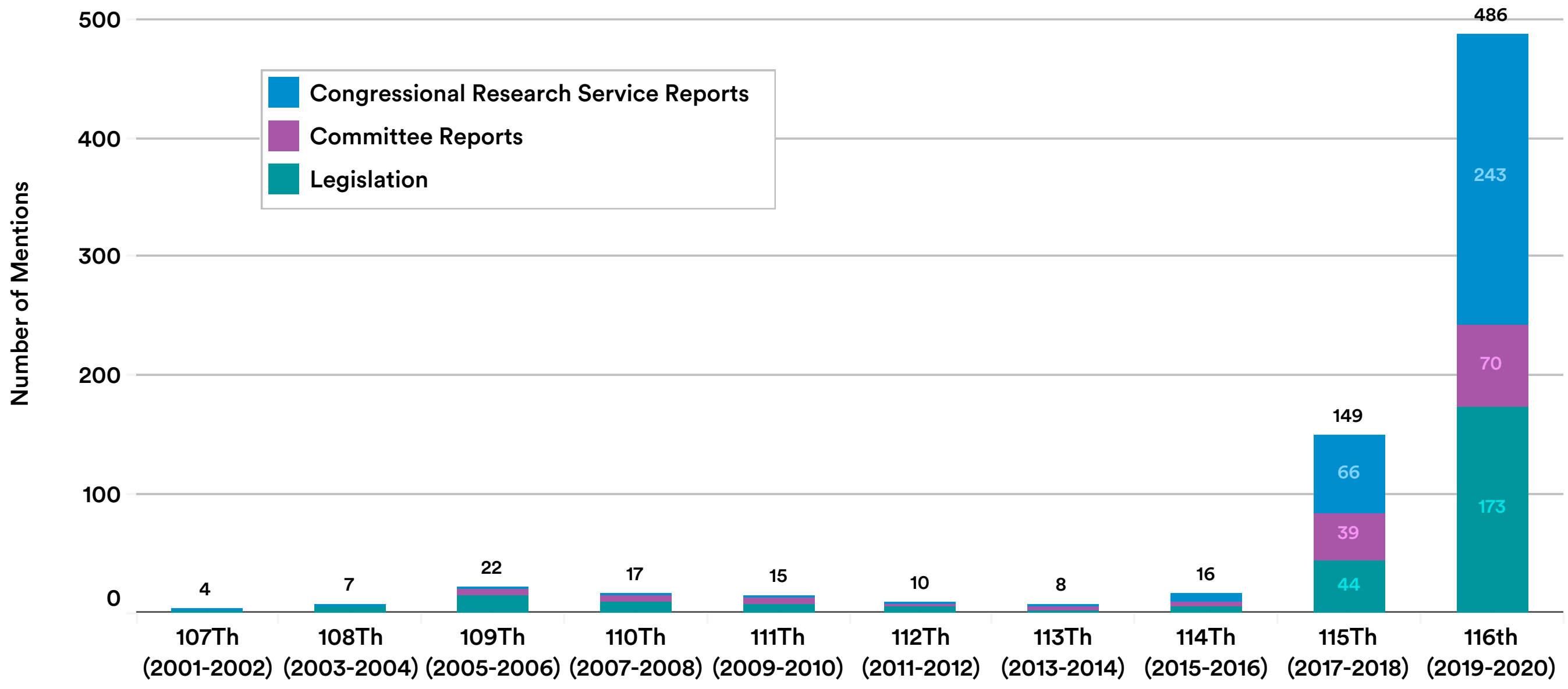

Figure 7.4.1

1 If a speaker or member mentioned artificial intelligence (AI) or machine learning (ML) multiple times within remarks, or multiple speakers mentioned AI or ML within the same event, it appears only once as a result. Counts for AI and ML are separate, as they were conducted in separate searches. Mentions of the abbreviations "AI" or "ML" are not included.

## U.S. Congressional Record

The 116th Congress (January 1, 2019–January 3, 2021) is the most AI-focused congressional session in history. The number of mentions of AI by this Congress in legislation, committee reports, and CRS reports is more than triple that of the 115th Congress. Congressional interest in AI has continued to accelerate in 2020. Figure 7.4.1 shows that during this congressional session, 173 distinct pieces of legislation either focused on or contained language about AI technologies, their development, use, and rules governing them. During that two-year period, various House and Senate committees and subcommittees commissioned 70 reports on AI, while the CRS, tasked as a fact-finding body for members of Congress, published 243 about AI or referencing AI.

## Mentions of AI and ML in Congressional/ Parliamentary Proceedings

As shown in Figures 7.4.2–7.4.5, the number of mentions of artificial intelligence and machine learning in the proceedings of the U.S. Congress and the U.K. parliament continued to rise in 2020, while there were fewer mentions in the parliamentary proceedings of Canada.

MENTIONS of AI and ML in the PROCEEDINGS of U.S. CONGRESS, 2011-20

Sources: U.S. Congressional Record website, the McKinsey Global Institute, 2020 | Chart: 2021 AI Index Report

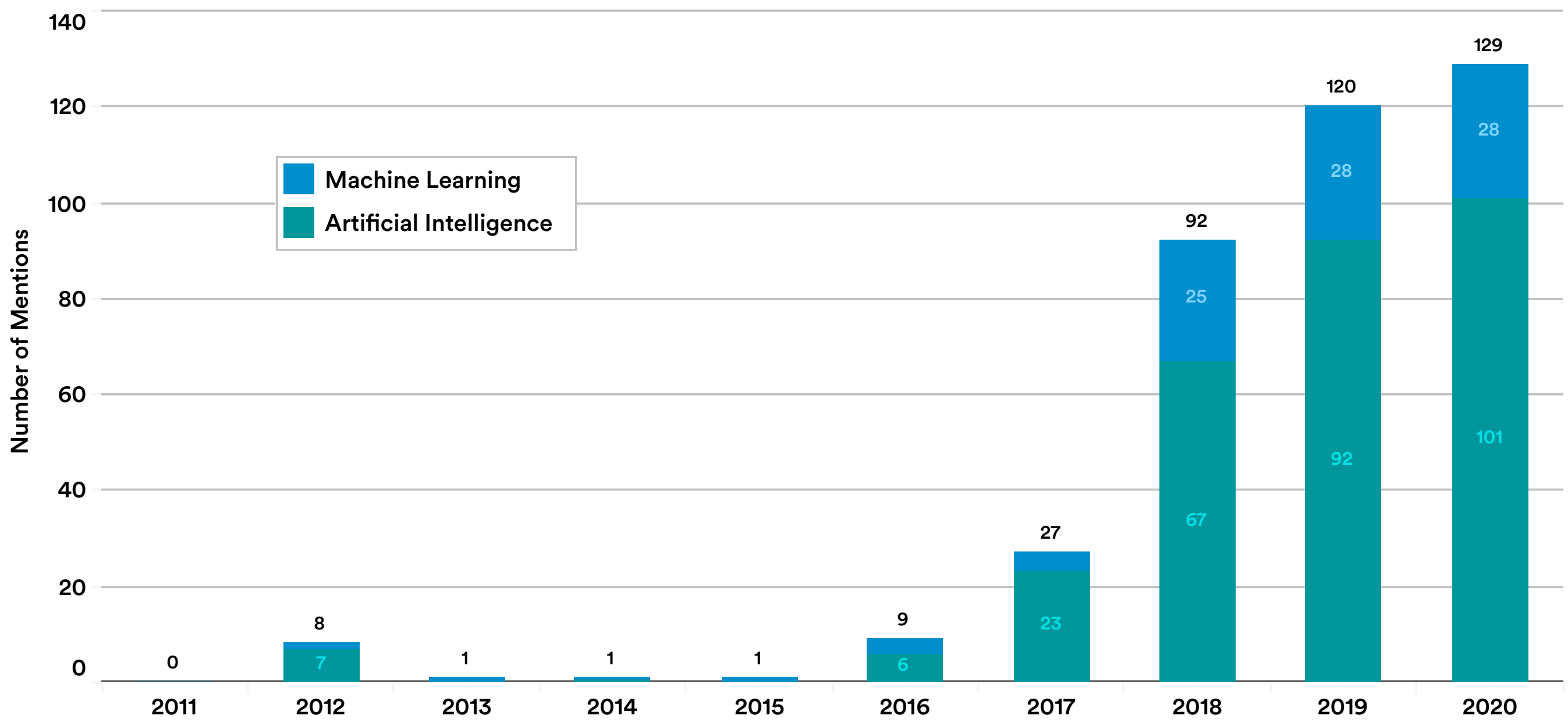

Figure 7.4.2

## MENTIONS of AI and ML in the PROCEEDINGS of U.K. PARLIAMENT, 2011-20

Sources: Parliament of U.K. website, the McKinsey Global Institute, 2020 | Chart: 2021 AI Index Report

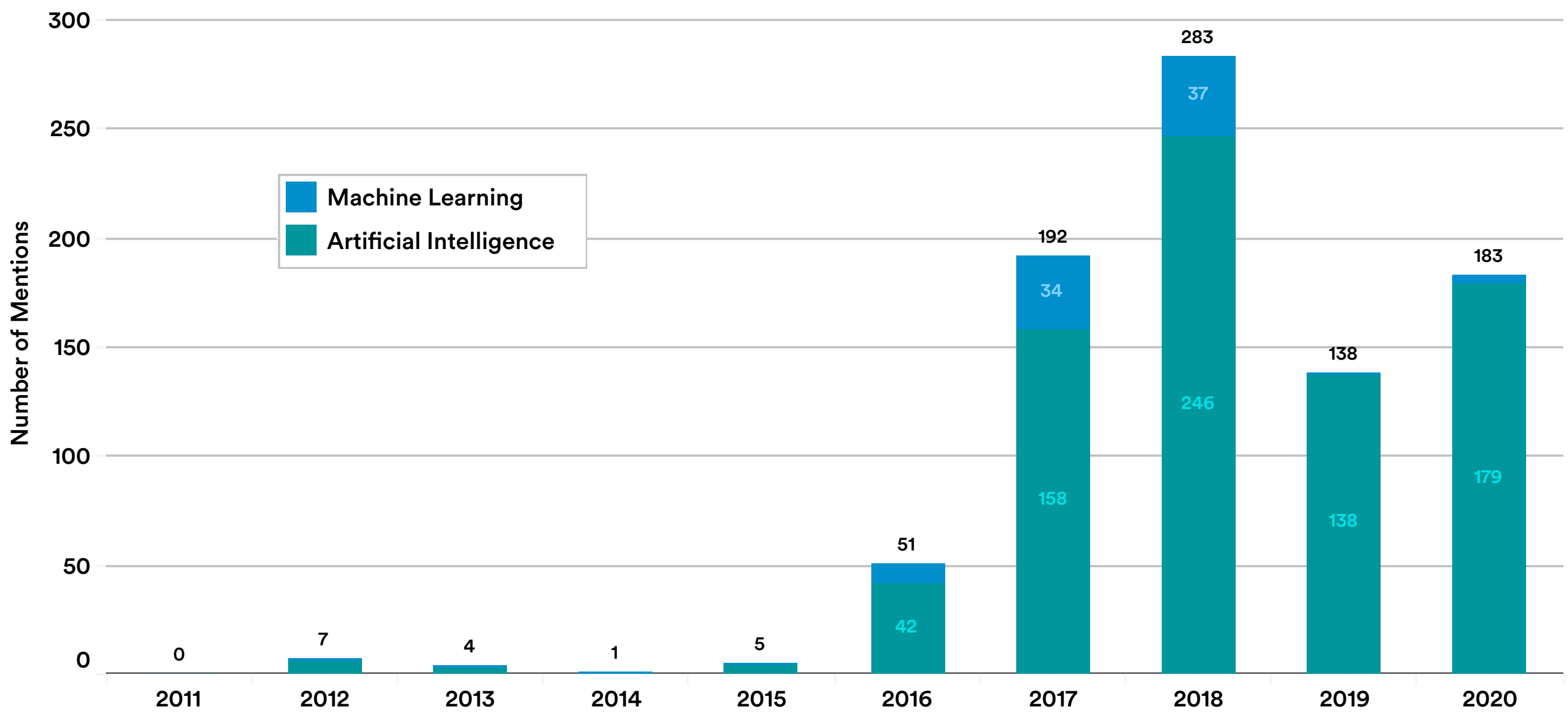

Figure 7.4.3

## MENTIONS of AI and ML in the PROCEEDINGS of CANADIAN PARLIAMENT, 2011-20

Sources: Canadian Parliament website, the McKinsey Global Institute, 2020 | Chart: 2021 AI Index Report

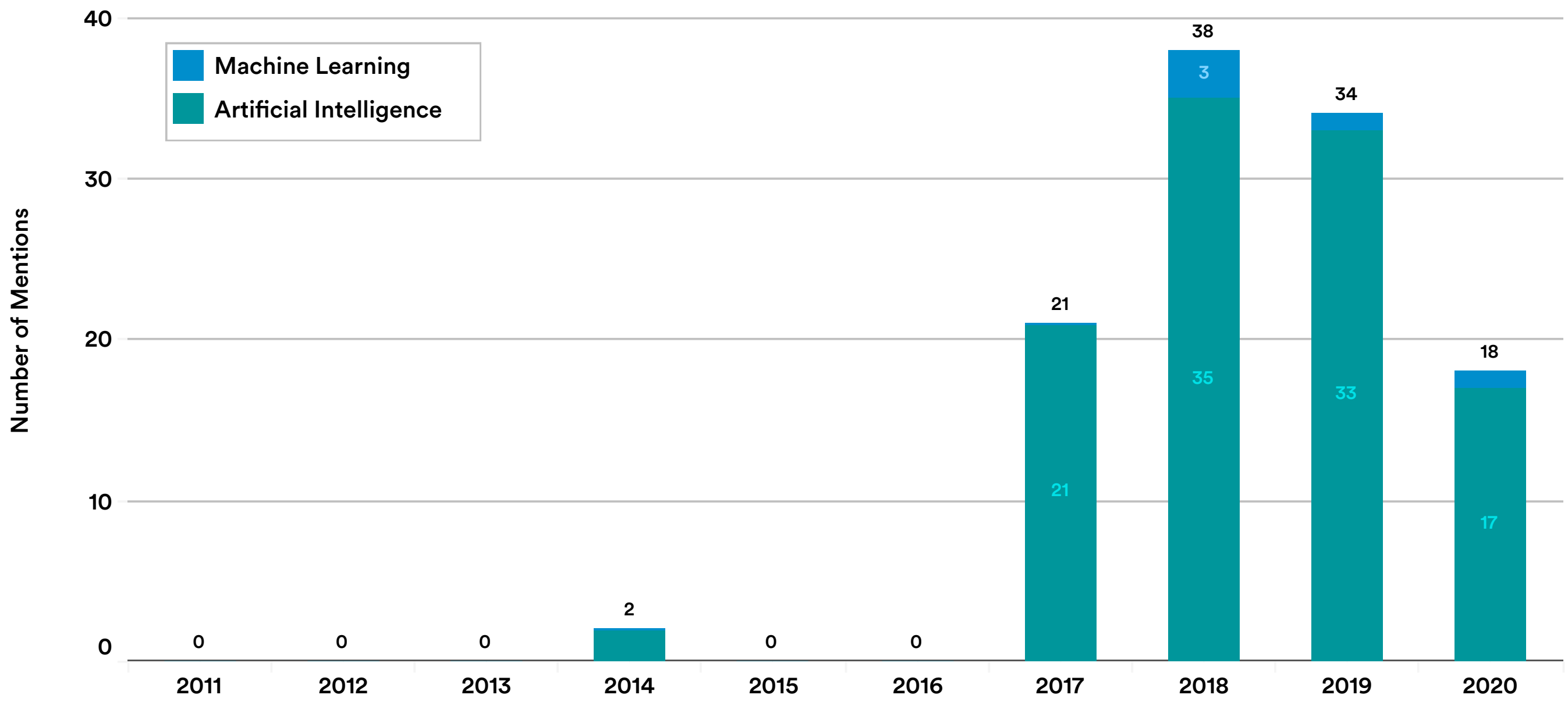

Figure 7.4.4

## CENTRAL BANKS

Central banks play a key role in conducting currency and monetary policy in a country or a monetary union. As with many other institutions, central banks are tasked with integrating AI into their operations and relying on big data analytics to assist them with forecasting, risk management, and financial supervision.

Prattle, a leading provider of automated investment research solutions, monitors mentions of AI in the communications of central banks, including meeting minutes, monetary policy papers, press releases, speeches, and other official publications.

Figure 7.4.5 shows a significant increase in the mention of AI across 16 central banks over the past 10 years, with the number reaching a peak of 1,020 in 2019. The sharp decline in 2020 can be explained by the COVID-19 pandemic as most central bank communications focused on responses to the economic downturn. Moreover, the Federal Reserve in the United States, Norges Bank in Norway, and the European Central Bank top the list for the most aggregated number of AI mentions in communications in the past five years (Figure 7.4.6).

MENTIONS of AI in CENTRAL BANK COMMUNICATIONS around THE WORLD, 2011-20
Source: Prattle/LiquidNet, 2020 | Chart: 2021 AI Index Report

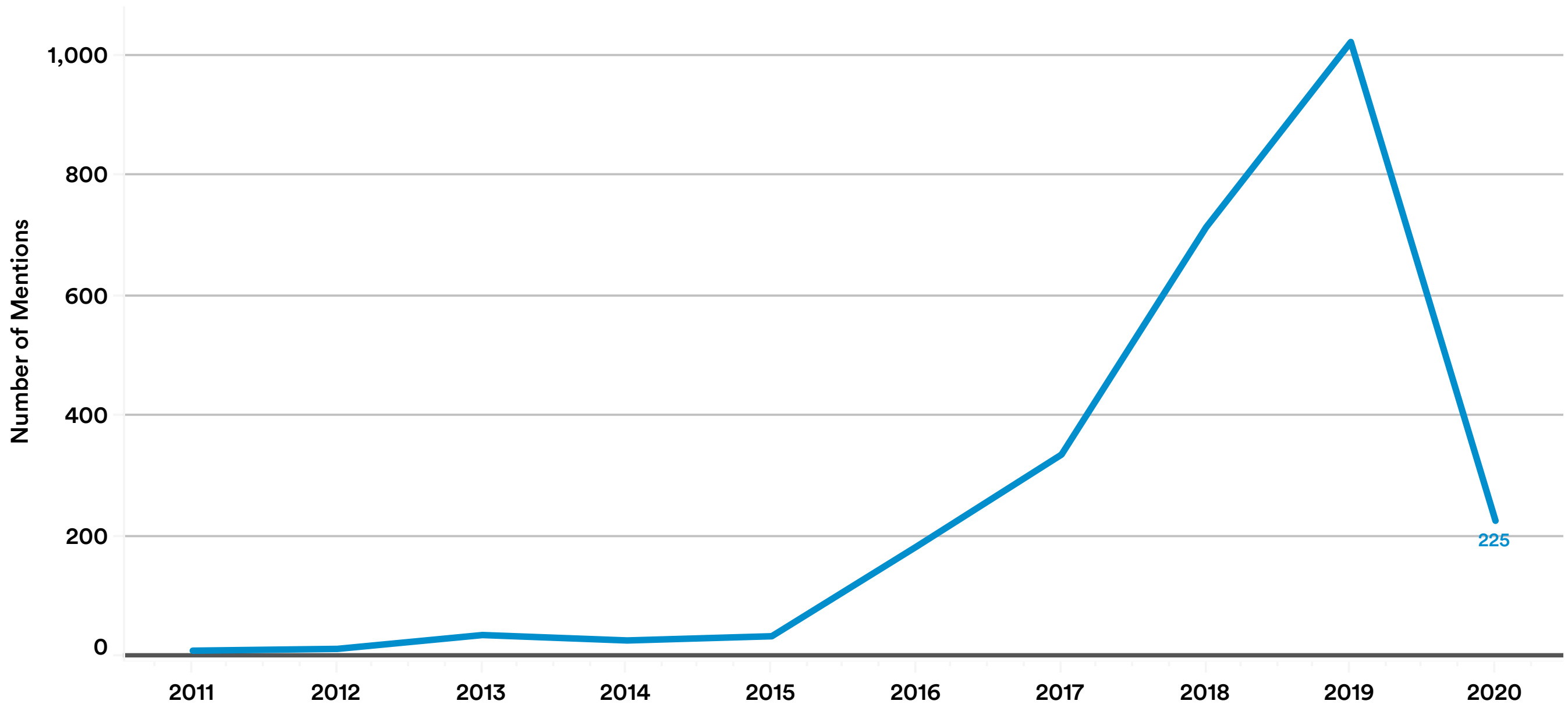

Figure 7.4.5

2 See Science & Technology Review and Scientific American for more details.

MENTIONS of AI in CENTRAL BANK COMMUNICATIONS around THE WORLD by BANK, 2016-20 (SUM)
Source: Prattle/LiquidNet, 2020 | Chart: 2021 AI Index Report

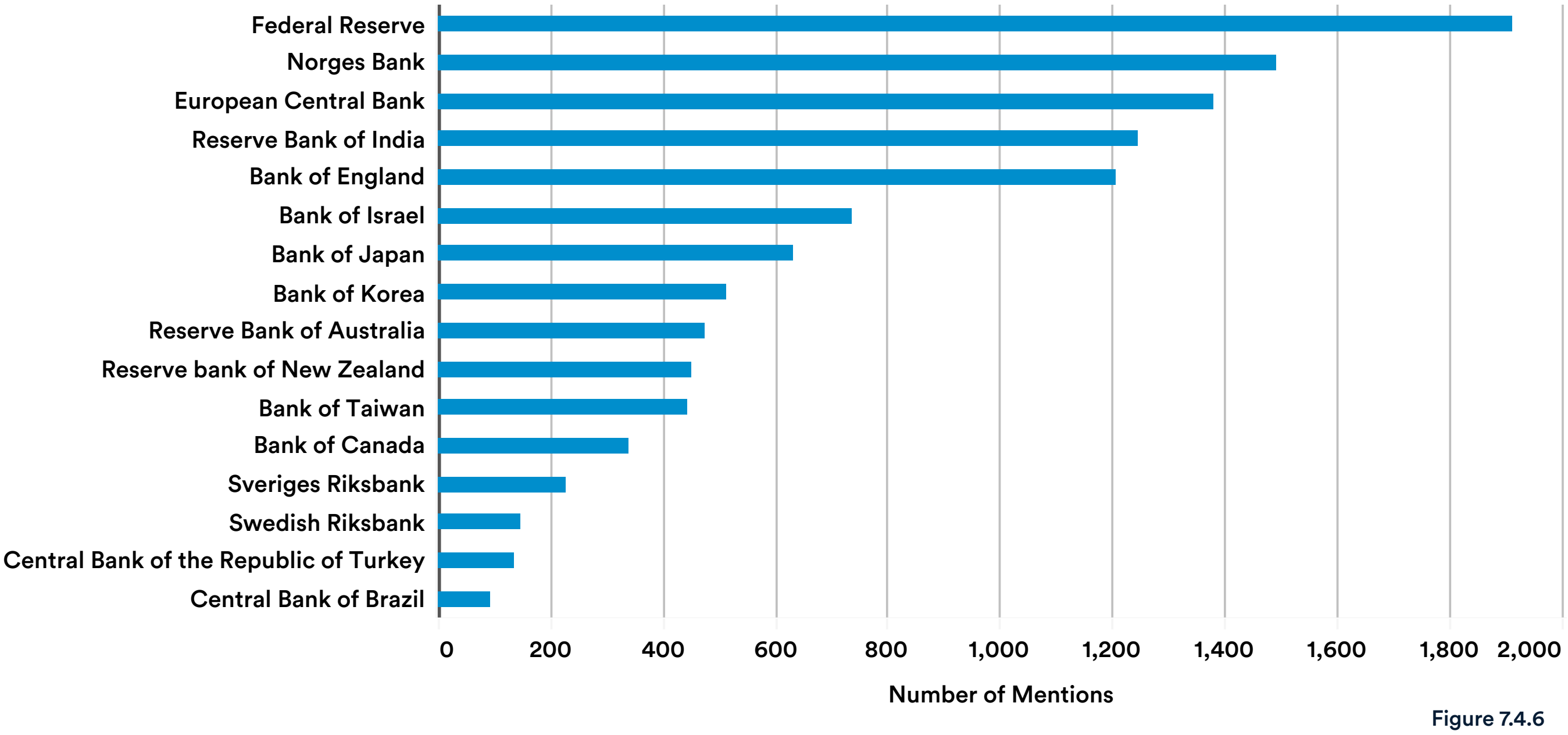

Figure 7.4.6

## U.S. AI POLICY PAPERS

What are the AI policy initiatives outside national and intergovernmental governments? We monitored 42 prominent organizations that deliver policy papers on topics related to AI and assessed the primary topic as well as the secondary topic on policy papers published in 2019 and 2020. (See the Appendix for a complete list of organizations included.) Those organizations are either U.S.-based or have a sizable presence in the United States, and we grouped them into three categories: think tanks, policy institutes and academia (27); civil society organizations, associations and consortiums (9); and industry and consultancy (6).

AI policy papers are defined as research papers, research reports, blog posts, and briefs that focus on a specific policy issue related to AI and provide clear recommendations for policymakers. Primary topics mean that such a topic is the main focus of the policy paper, while secondary topics mean that the policy paper either briefly touches on the topic or the topic is a sub-focus of the paper.

Combined data for 2019 and 2020 suggests that the topics of innovation and technology, international affairs and international security, and industry and regulation are the main focuses of AI policy papers in the United States (Figure 7.4.7). Fewer documents placed a primary focus on topics related to AI ethics—such as ethics, equity and inclusion; privacy, safety and security; and justice and law enforcement—which have largely been secondary topics. Moreover, topics bearing on the physical sciences, energy and environment, humanities, and democracy have received the least attention in U.S. AI policy papers.

U.S. AI POLICY PRODUCTS by TOPIC, 2019-20 (SUM)
Source: Stanford HAI & AI Index, 2020 | Chart: 2021 AI Index Report

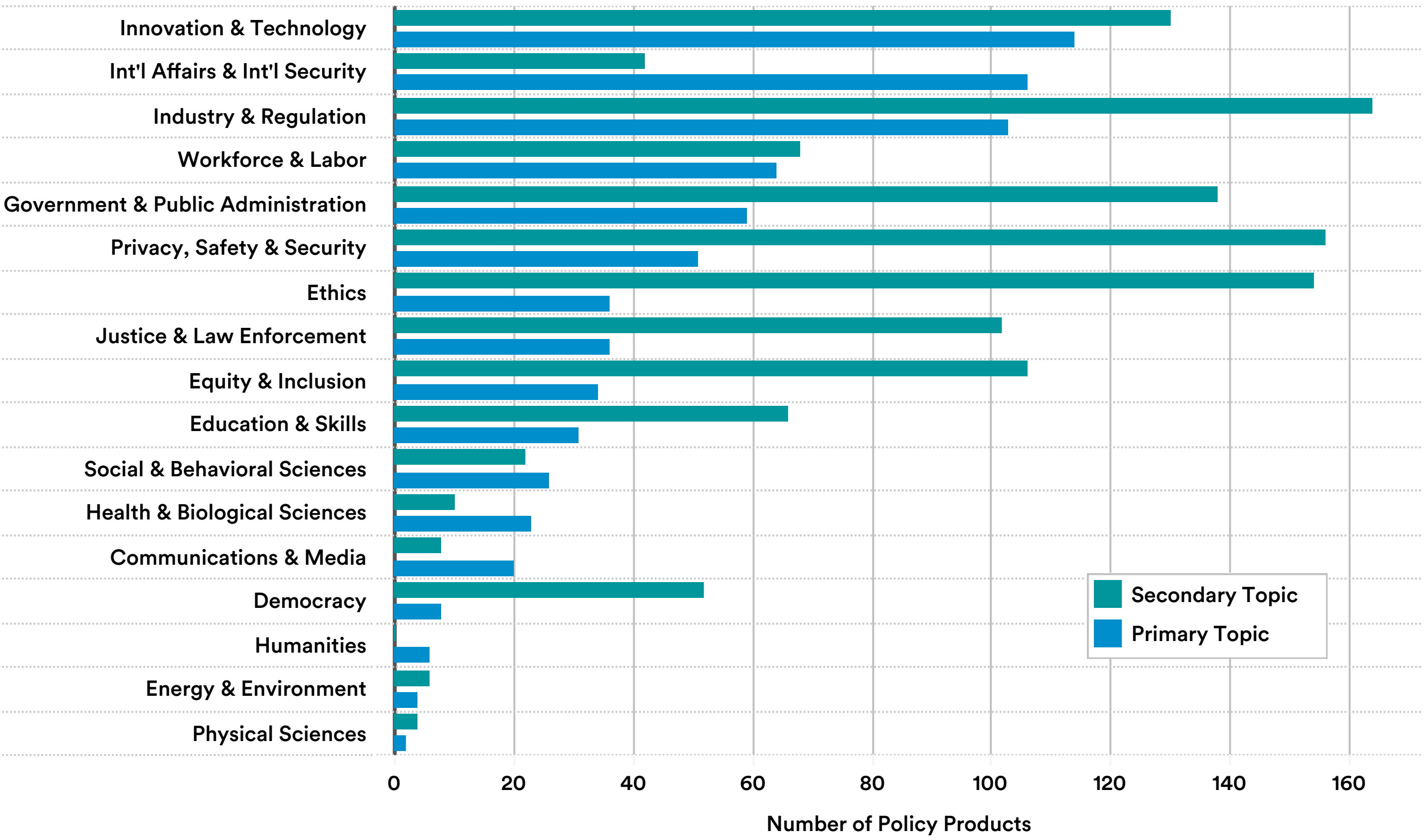

Figure 7.4.7

# Appendix

# Appendix

# CHAPTER 1: RESEARCH & DEVELOPMENT

## ELSEVIER

Prepared by Jörg Hellwig and Thomas A. Collins

### Source

Elsevier's Scopus database of scholarly publications has indexed more than 81 million peer-reviewed documents. This data was compiled by Elsevier.

### Methodology

Scopus tags its papers with keywords, publication dates, country affiliations, and other bibliographic information.

The Elsevier AI Classifier leveraged the following features extracted from the Scopus records that were returned as a result of querying against the provided approximately 800 AI search terms. Each record fed into the feature creation also maintained a list of each search term that hit for that particular record:

- hasAbs: Boolean value whether or not the record had an abstract text section in the record (e.g., some records are only title and optional keywords)
- coreCnt: number of core-scored search terms present for the record
- mediumCnt: number of medium-scored search terms present for the record
- lowCnt: number of low-scored search terms present for the record
- totalCnt: total number of search terms present for the record
- pcntCore: coreCnt/totalCnt
- pcntMedium: mediumCnt/totalCnt
- pcntLow: lowCnt/totalCnt
- totalWeight = 5*coreCnt + 3*mediumCnt + 1*lowCnt
- normWeight = if (has Abs) { totalWeight / (title.length + abstract.length) } else
- { totalWeight/title.length}
- hasASJC: Boolean value: does the record have an associated ASJC list?
- isAiASJC: does ASJC list contain 1702?
- isCompSciASJC does ASJC list contain a 17XX ASJC code ("1700," "1701," "1702," "1703," "1704," "1705," "1706," "1707," "1708," "1709," "1710," "1711," "1712")
- isCompSubj: does the Scopus record have a ComputerScience subject code associated with it? This should track 1:1 to isCompSciASJC. Scopus has 27 major subject areas of which one is Computer Science. The feature checks, if the publication is within Computer Science or not. This is no exclusion.pcntCompSciASJC: percentage of ASJC codes for record that are from the CompSci ASJC code list

Details on Elsevier's dataset defining AI, country affiliations, and AI subcategories can be found in the 2018 AI Index Report Appendix.

### Nuance

- The Scopus system is retroactively updated. As a result, the number of papers for a given query may increase over time.
- Members of the Elsevier team commented that data on papers published after 1995 would be the most reliable. The raw data has 1996 as the starting year for Scopus data.

### Nuances specific to AI publications by region

- Papers are counted utilizing whole counting rather than fractional counting. Papers assigned to multiple countries (or regions) due to collaborations are counted toward each country (or region). This explains why top-line numbers in a given year may not match individual country numbers. For example, a paper assigned to Germany, France, and the United States will appear on each country's count, but only once for Europe (plus once for the U.S.) as well as being counted only at the global level.
- "Other" includes all other countries that have published one or more AI papers on Scopus.

## Nuances specific to publications by topic

- The 2017 AI Index Report showed only AI papers within the CS category. In the 2018 and 2019 reports, all papers tagged as AI were included, regardless of whether they fell into the larger CS category.
- Scopus has a subject category called AI, which is a subset of CS, but this is relevant only for a subject-category approach to defining AI papers. The methodology used for the report includes all papers, since increasingly not all AI papers fall under CS.

## Nuances specific to methodology

- The entire data collection process was done by Elsevier internally. The AI Index was not involved in the keyword selection process or the counting of relevant papers.
- The boundaries of AI are difficult to establish, in part because of the rapidly increasing applications in many fields, such as speech recognition, computer vision, robotics, cybersecurity, bioinformatics, and healthcare. But limits are also difficult to define because of AI's methodological dependency on many areas, such as logic, probability and statistics, optimization, photogrammetry, neuroscience, and game theory—to name just a few. Given the community's interest in AI bibliometrics, it would be valuable if groups producing these studies strived for a level of transparency in their methods, which would support the reproducibility of results, particularly on different underlying bibliographic databases.

## AI Training Set

A training set of approximately 1,500 publications defines the AI field. The set is only the EID (the Scopus identifier of the underlying publications). Publications can be searched and downloaded either from Scopus directly or via the API.The training set is a set of publications randomly selected from the initial seven mio publications. After running the algorithm we verify the results of the training set with the gold set (expert hand-checked publications which are definitely AI).

## MICROSOFT ACADEMIC GRAPH: METHODOLOGY

Prepared by Zhihong Shen, Boya Xie, Chiyuan Huang, Chieh-Han Wu, and Kuansan Wang

### Source

The Microsoft Academic Graph[1] is a heterogeneous graph containing scientific publication records and citation relationships between those publications, as well as authors, institutions, journals, conferences, and fields of study. This graph is used to power experiences in Bing, Cortana, Word, and Microsoft Academic. The graph is currently being updated on a weekly basis. Learn more about MAG here.

### Methodology

**MAG Data Attribution:** Each paper is counted exactly once. When a paper has multiple authors or regions, the credit is equally distributed to the unique regions. For example, if a paper has two authors from the United States, one from China, and one from the United Kingdom, then the United States, China, and the United Kingdom each get one-third credit.

**Metrics:** Total number of published papers (journal papers, conference papers, patents, repository[2]); total number of citations of published papers.

**Definition:** The citation and reference count represents the number of respective metrics for AI papers collected from all papers. For example, in "OutAiPaperCitationCountryPairByYearConf.csv," a row stating "China, United States, 2016, 14955" means that China's conference AI papers published in 2016 received 14,955 citations from (all) U.S. papers indexed by MAG.

**Curating the MAG Dataset and References:** Generally speaking, the robots sit on top of a Bing crawler to read everything from the web and have access to the entire web index. As a result, MAG is able to program the robots to conduct more web searches than a typical human can complete. This helps disambiguate entities with the same names. For example, for authors, MAG gets to additionally use all the CVs and institutional homepages on the web as signals to recognize and verify claims[3]. MAG has found this approach to be superior to the results of the best of the KDD Cup 2013 competition, which uses only data from within all publication records and Open Researcher and Contributor Identifiers (ORCIDs).

### Notes About the MAG Data

**Conference Papers:** After the contents and data sources were scrutinized, it was determined that some of the 2020 conference papers were not properly tagged with their conference venue. Many conference papers in the MAG system are under arXiv papers, but due to issues arising from some data sources (including delays in DBLP and web form changes on the ACM website), they were possibly omitted as 2020 conference papers (ICML-PKDD, IROS, etc.). However, the top AI conferences (selected not in terms of publication count, but rather considering both publication and citation count as well as community prestige) are complete. In 2020, the top 20 conferences presented 103,000 papers, which is 13.7% of all AI conference papers, and they received 7.15 million citations collectively, contributing 47% of all citations received for all AI conference papers. The number of 2020 conference publications is slightly lower than in 2019. Data is known to be missing for ICCV and NAACL. About 100 Autonomous Agents and Multiagent Systems (AAMAS) conference papers are erroneously attributed to an eponymous journal.

**Unknown Countries for Journals and Conferences:** For the past 20 to 30 years, 30% of journal and conference affiliation data lacks affiliation by country or region, due to errors in paper format, data source, and PDF parsing, among others.

1 See "A Review of Microsoft Academic Services for Science of Science Studies" and "Microsoft Academic Graph: When Experts Are Not Enough" for more details.
2 Repository as a publication type in MAG refers to both preprints and postprints. In the AI domain, it predominantly comes from arXiv. See "Is Preprint the Future of Science? A Thirty Year Journey of Online Preprint Services" for details.
3 See "Machine Verification for Paper and Author Claims" and "How Microsoft Academic Uses Knowledge to Address the Problem of Conflation/Disambiguation" for details.

## MICROSOFT ACADEMIC GRAPH: PATENT DATA CHALLENGE

As mentioned in the report, the patent data—especially the affiliation information—is incomplete in the MAG database. The reason for the low coverage is twofold. First, applications published by the patent offices often identify the inventors by their residencies not affiliations. While patent applications often have the information about the "assignees" of a patent, they do not necessarily mean the underlying inventions originate from the assignee institutions. Therefore, detected affiliations may be inaccurate. In case a patent discloses the scholarly publications underlying the invention, MAG can infer inventors' affiliations through the scholarly publications.

Second, to maximize intellectual property protection around the globe, institutions typically file multiple patent applications on the same invention under various jurisdictions. These multiple filings, while appear very different because the titles and inventor names are often translated into local languages, are in fact the result of a single invention. Raw patent counts therefore inflate the inventions in their respective domains. To remediate this issue, MAG uses the patent family ID feature to combine all filings with the original filing, which allows the database to count filings all around the world of the same origin only once.[4] Conflating the multiple patent applications of the same invention is not perfect, and over-conflations of patents are more noticeable in MAG than scholarly articles.

These challenges raise questions about the reliability of data on the share of AI patent publications by both region and geographic area. Those charts are included below.

### By Region

AI PATENT PUBLICATIONS (% of WORLD TOTAL) by REGION, 2000-20

Source: Microsoft Academic Graph, 2020 | Chart: 2021 AI Index Report

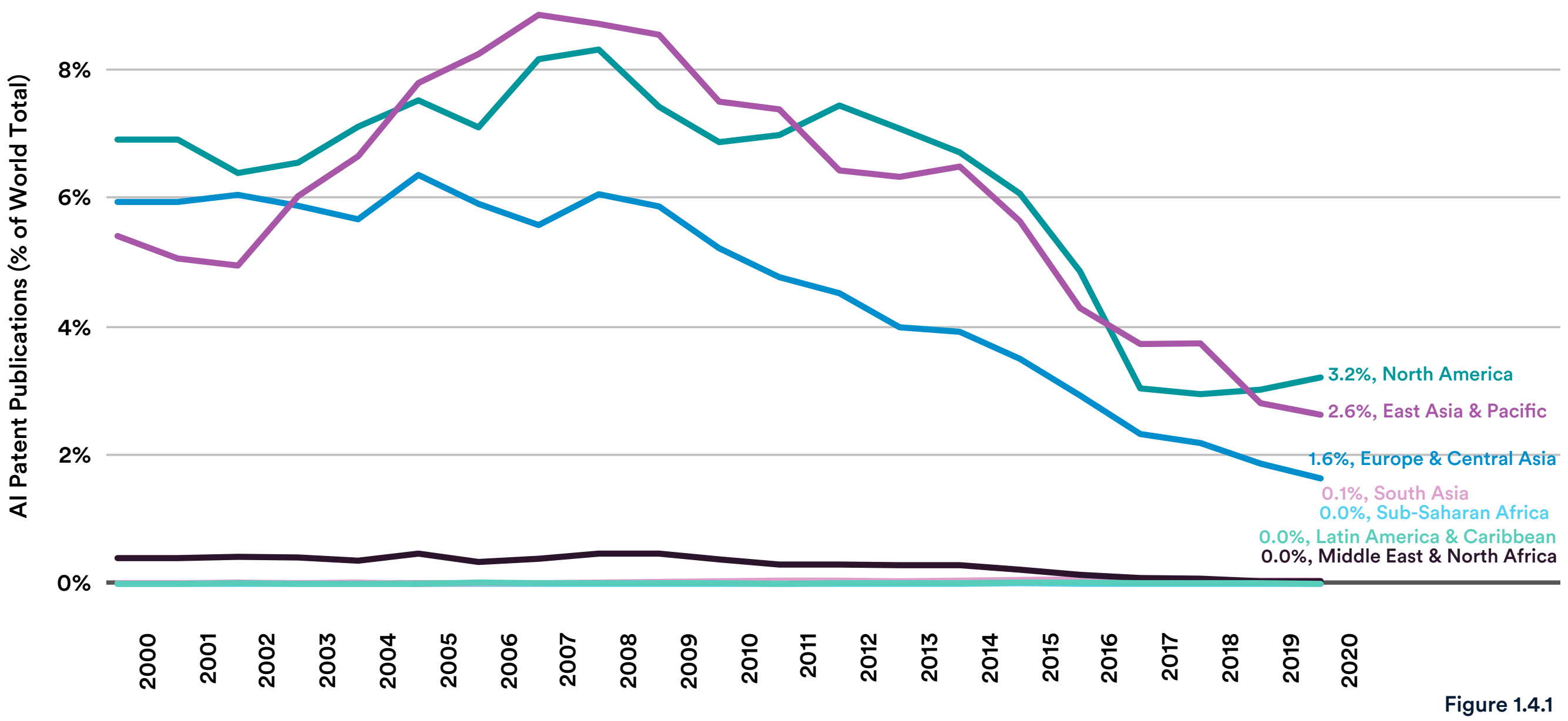

Figure 1.4.1

4 Read "Sharpening Insights into the Innovation Landscape with a New Approach to Patents" for more details.

## By Geographic Area

AI PATENT PUBLICATIONS (% of WORLD TOTAL) by GEOGRAPHIC AREA, 2000-20

Source: Microsoft Academic Graph, 2020 | Chart: 2021 AI Index Report

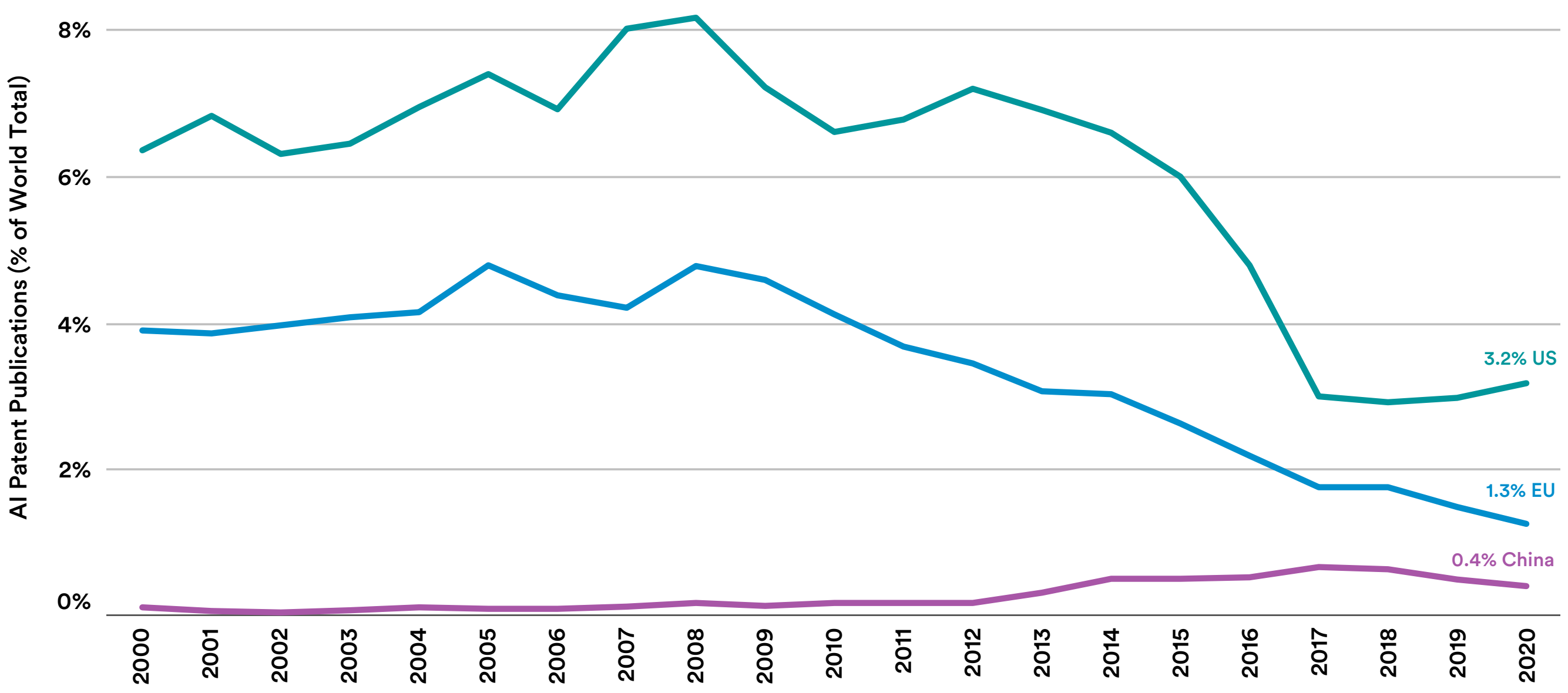

Figure 1.4.2

## Citation

AI PATENT CITATIONS (% of WORLD TOTAL) by GEOGRAPHIC AREA, 2000-20

Source: Microsoft Academic Graph, 2020 | Chart: 2021 AI Index Report

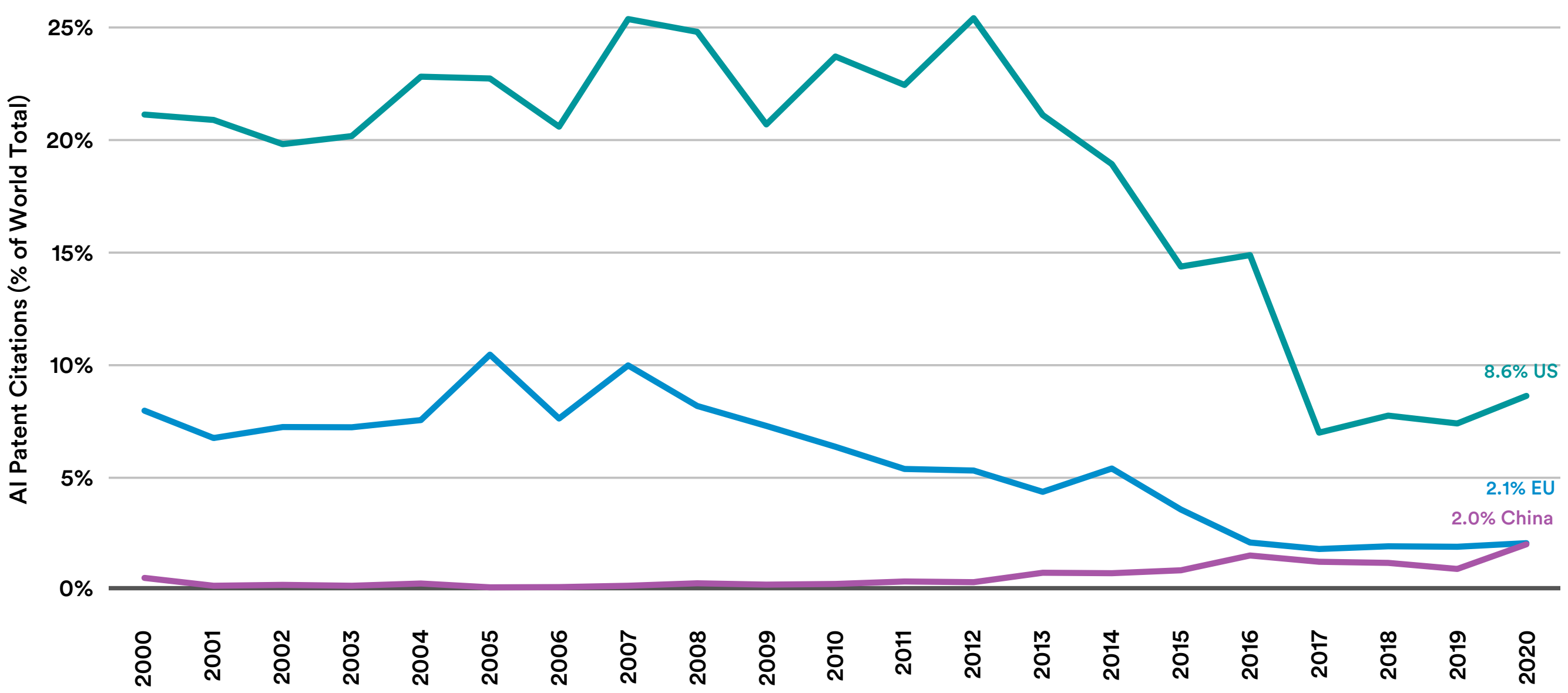

Figure 1.4.3

## MICROSOFT ACADEMIC GRAPH: MEASUREMENT CHALLENGES AND ALTERNATIVE DEFINITION OF AI

As the AI Index team discussed in the paper "Measurement in AI Policy: Opportunities and Challenges," choosing how to define AI and correctly capture relevant bibliometric data remain challenging. Data in the main report is based on a restricted definition of AI, adopted by MAG, that aligns with what has been used in previous AI Index reports. One consequence is that such a definition excludes many AI publications from venues considered to be core AI venues. For example, only 25% of conference publications in the 2020 AAAI conference are included in the original conference dataset.

To spur discussion on this important topic, this section presents the MAG data with an alternative definition of AI used by the Organisation for Economic Co-operation and Development (OECD). OECD defines AI publications as papers in the MAG database tagged with a field of study that is categorized in either the "artificial intelligence" or the "machine learning" field of study as well as their subtopics in the MAG taxonomy.[5] This is a more liberal definition than the one used by MAG, which considers only those publications tagged with "artificial intelligence" as AI publications. For example, an application paper in biology that uses ML techniques will be counted as an AI publication under the OECD definition, but not under the MAG definition unless the paper is specifically tagged in the AI category.

Charts corresponding to those in the main text but using the OECD definition are presented below. The overall trends are very similar.

### AI Journal Publications (OECD Definition)

OECD DEFINITION: NUMBER of AI JOURNAL PUBLICATIONS, 2000-20
Source: Microsoft Academic Graph, 2020 | Chart: 2021 AI Index Report

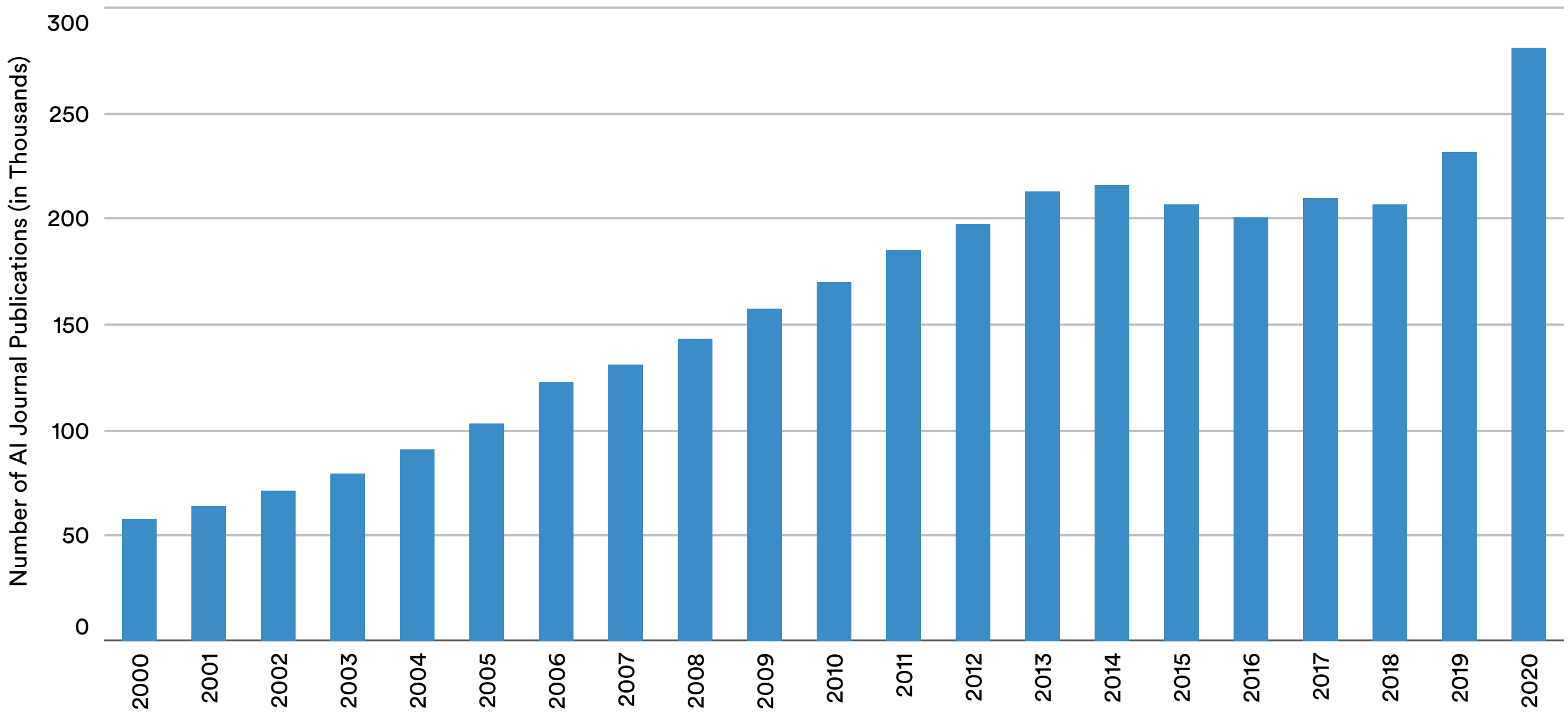

Figure 1.5.1a

5 Read the OECD.AI Policy Observatory MAG methodological note for more details on the MAG-OECD definition of AI and "A Web-scale System for Scientific Knowledge Exploration" on the MAG Taxonomy.

## OECD DEFINITION: AI JOURNAL PUBLICATIONS (% of ALL JOURNAL PUBLICATIONS), 2000-20

Source: Microsoft Academic Graph, 2020 | Chart: 2021 AI Index Report

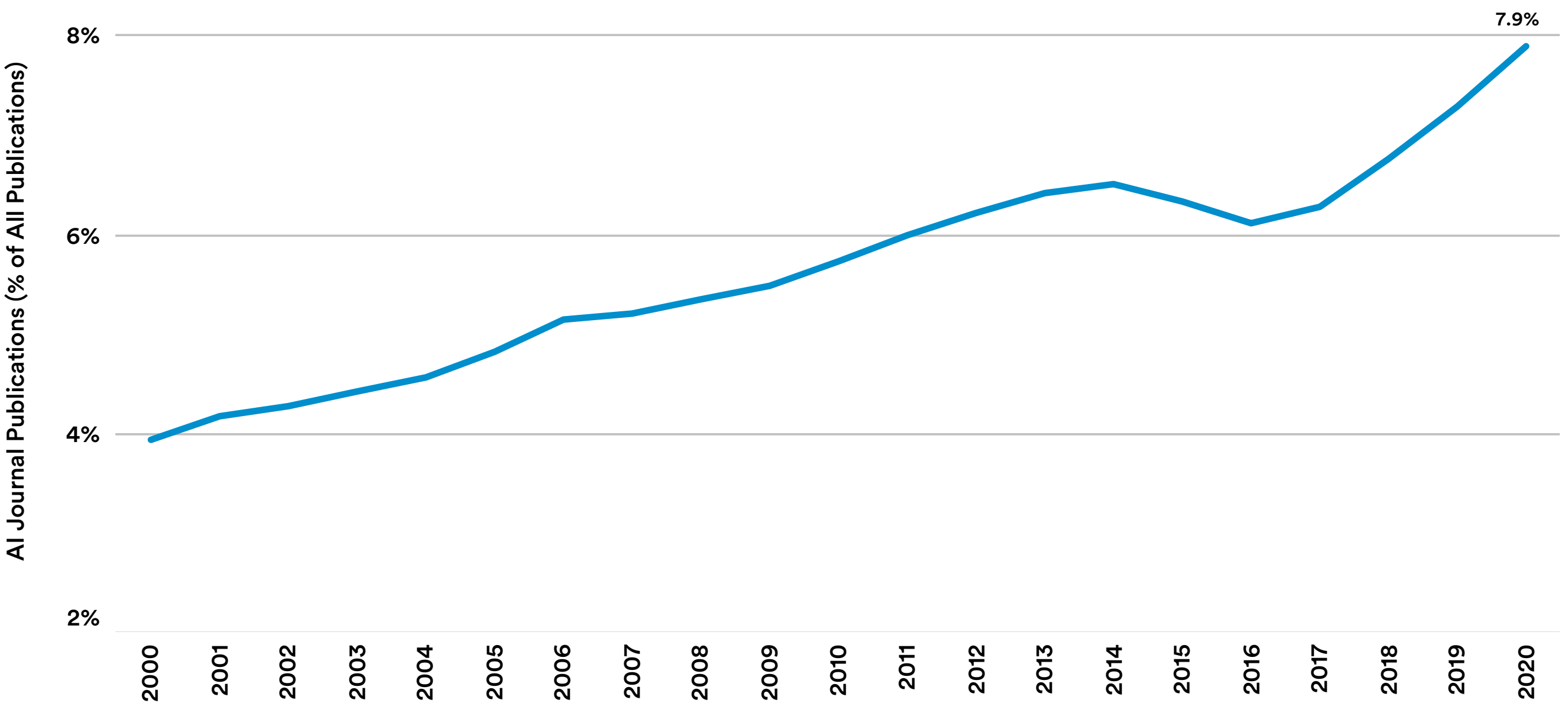

Figure 1.5.1b

## OECD DEFINITION: AI JOURNAL PUBLICATION (% of WORLD TOTAL) by REGION, 2000-20

Source: Microsoft Academic Graph, 2020 | Chart: 2021 AI Index Report

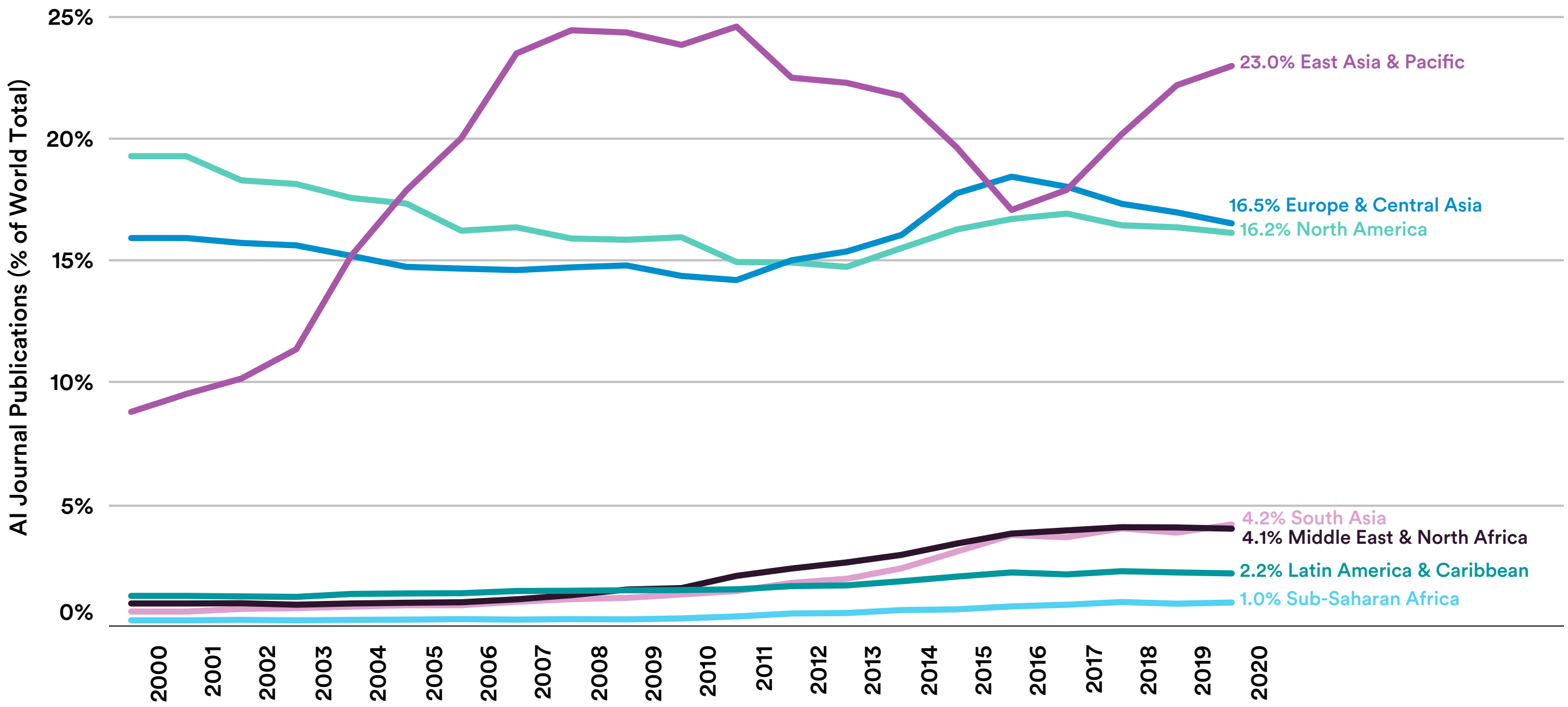

Figure 1.5.2

## OECD DEFINITION: AI JOURNAL PUBLICATION (% of WORLD TOTAL) by GEOGRAPHIC AREA, 2000-20

Source: Microsoft Academic Graph, 2020 | Chart: 2021 AI Index Report

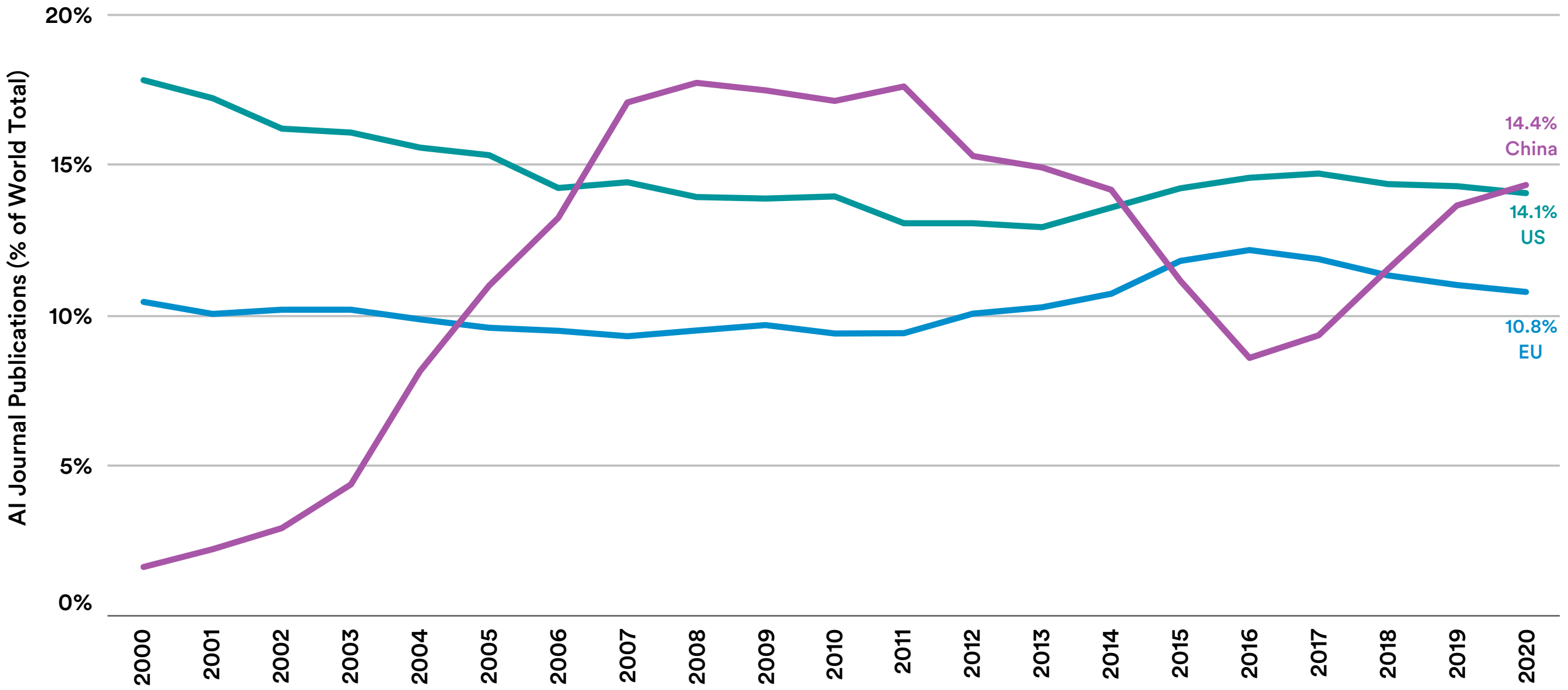

Figure 1.5.3

## OECD DEFINITION: AI JOURNAL CITATIONS (% of WORLD TOTAL) by GEOGRAPHIC AREA, 2000-20

Source: Microsoft Academic Graph, 2020 | Chart: 2021 AI Index Report

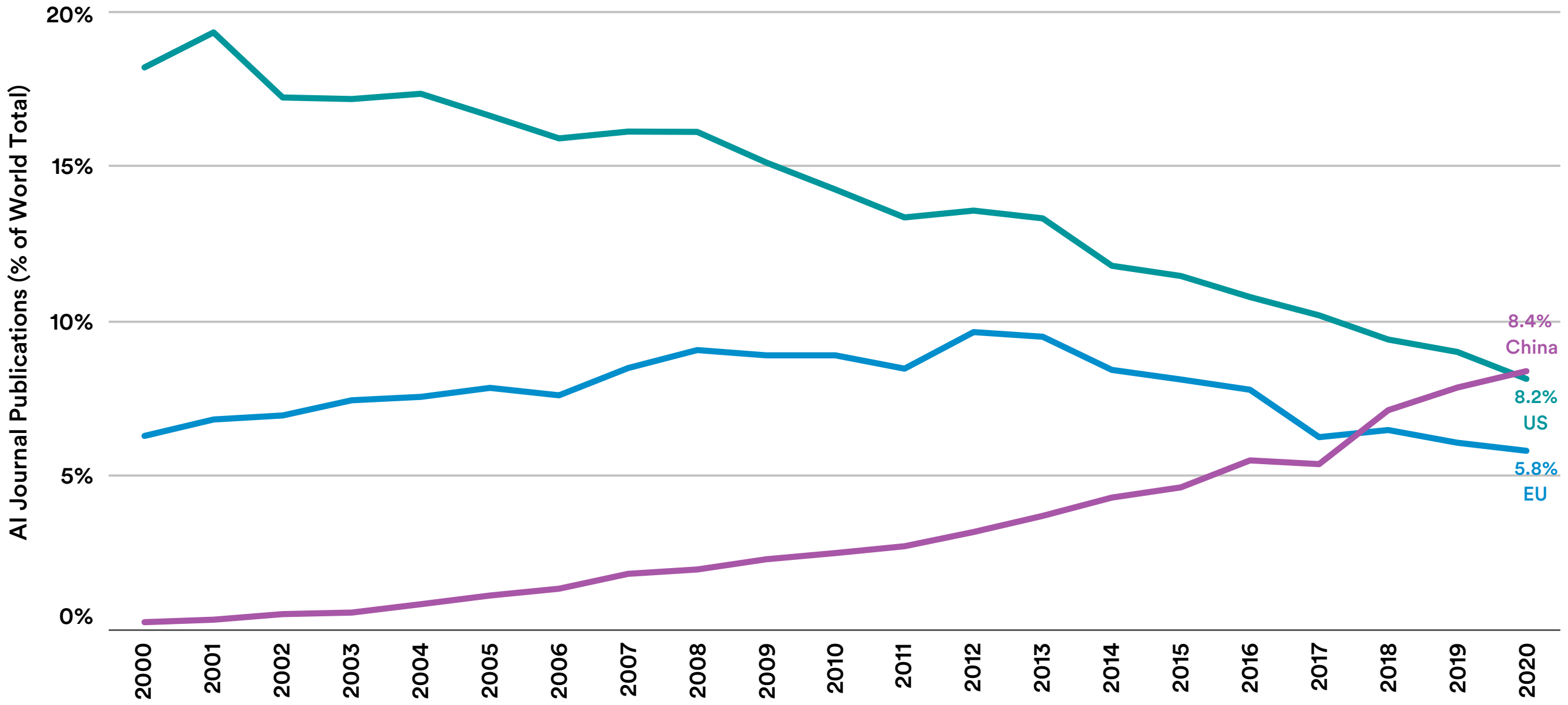

Figure 1.5.4

## AI Conference Publications (OECD Definition)

### OECD DEFINITION: NUMBER of AI CONFERENCE PUBLICATIONS, 2000-20

Source: Microsoft Academic Graph, 2020 | Chart: 2021 AI Index Report

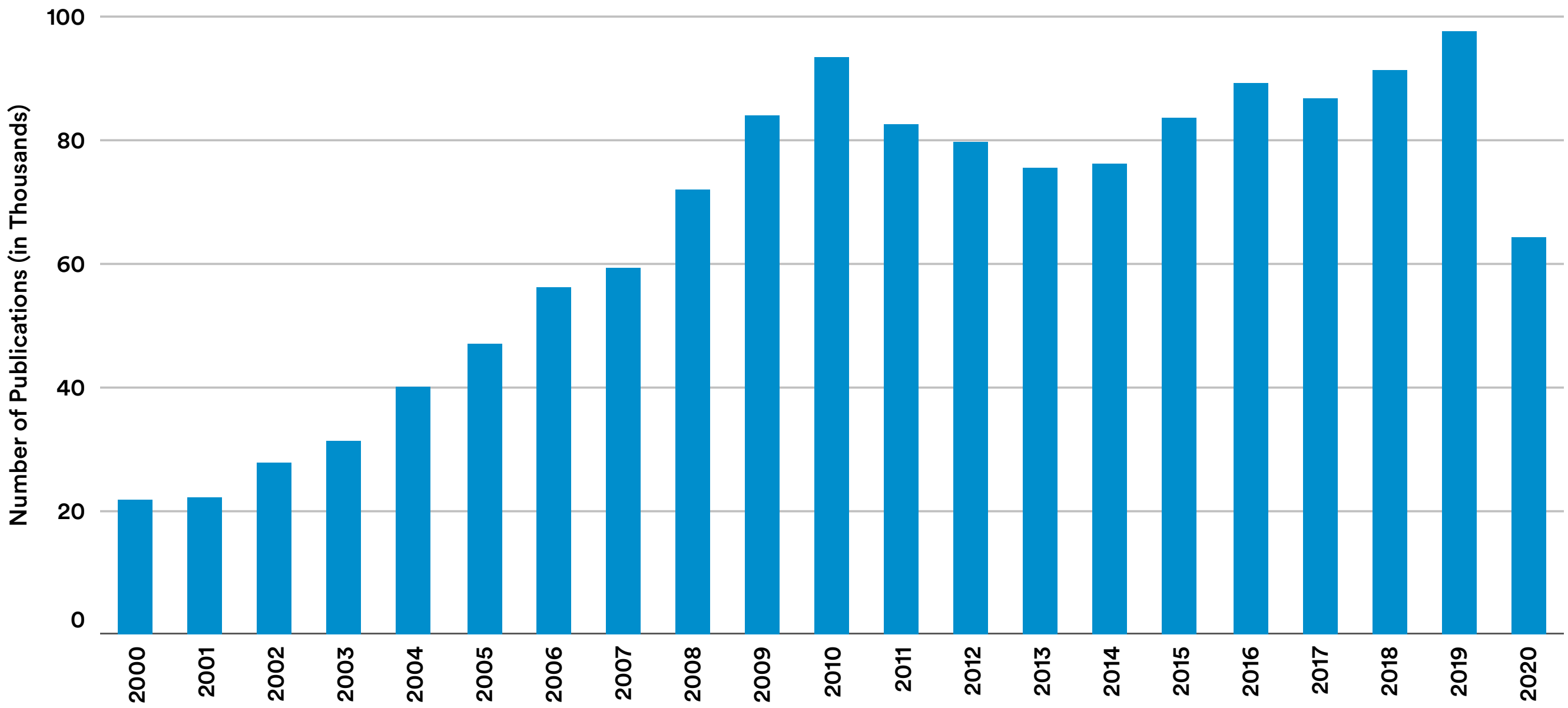

Figure 1.5.5a

### OECD DEFINITION: AI CONFERENCE PUBLICATIONS (% of ALL CONFERENCE PUBLICATIONS), 2000-20

Source: Microsoft Academic Graph, 2020 | Chart: 2021 AI Index Report

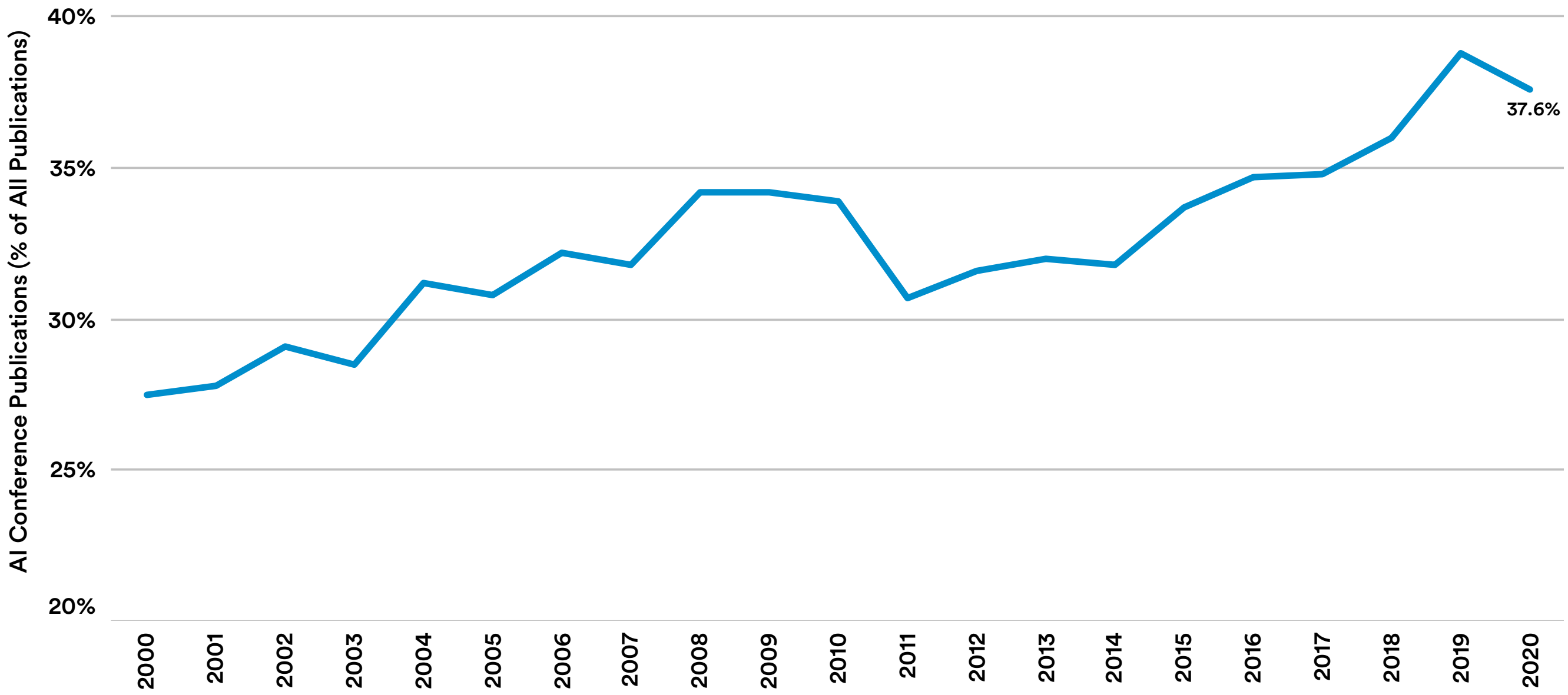

Figure 1.5.5b

## OECD DEFINITION: AI CONFERENCE PUBLICATION (% of WORLD TOTAL) by REGION, 2000-20

Source: Microsoft Academic Graph, 2020 | Chart: 2021 AI Index Report

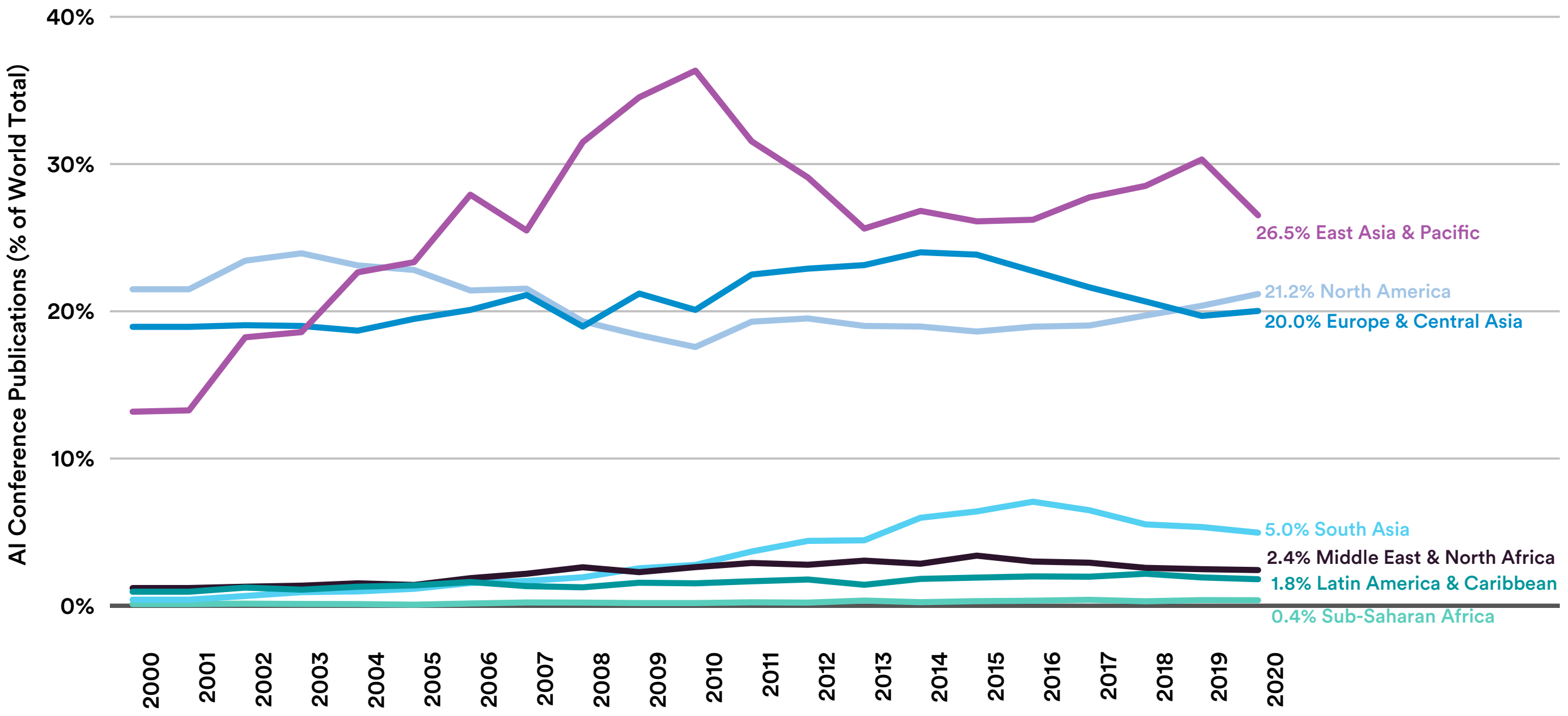

Figure 1.5.6

## OECD DEFINITION: AI CONFERENCE PUBLICATION (% of WORLD TOTAL) by GEOGRAPHIC AREA, 2000-20

Source: Microsoft Academic Graph, 2020 | Chart: 2021 AI Index Report

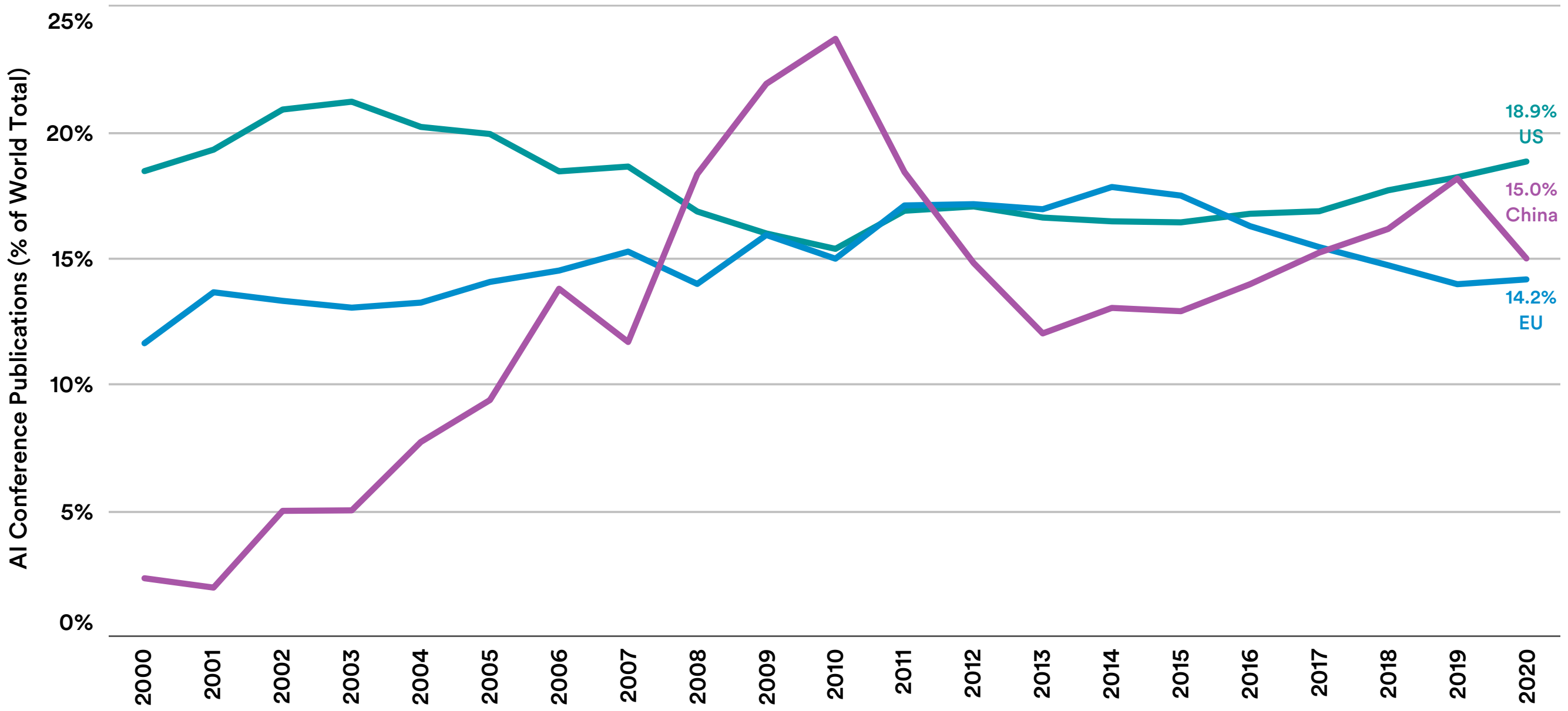

Figure 1.5.7

OECD DEFINITION: AI CONFERENCE CITATION (% of WORLD TOTAL) by GEOGRAPHIC AREA, 2000-20
Source: Microsoft Academic Graph, 2020 | Chart: 2021 AI Index Report

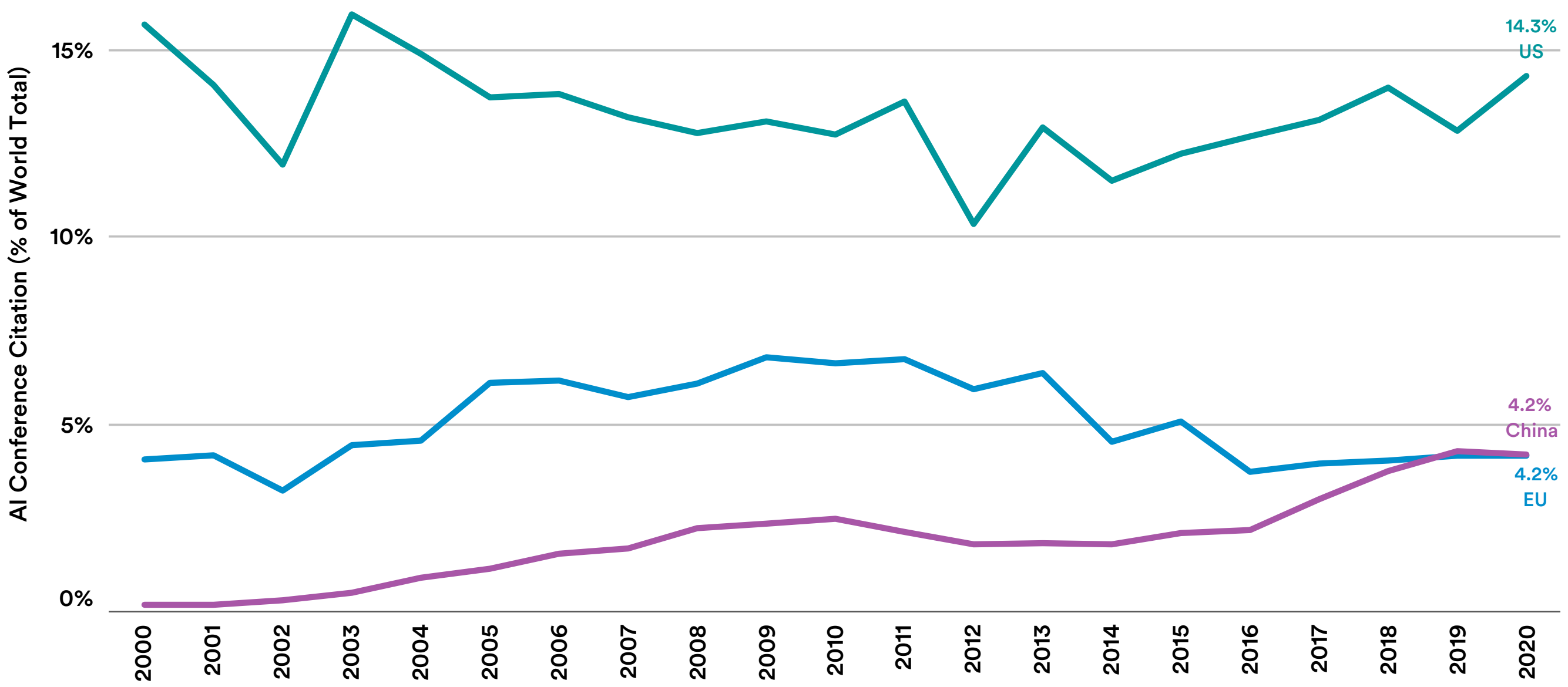

Figure 1.5.8

## AI Patent Publications (OECD Definition)

OECD DEFINITION: NUMBER of AI PATENT PUBLICATIONS, 2000-20
Source: Microsoft Academic Graph, 2020 | Chart: 2021 AI Index Report

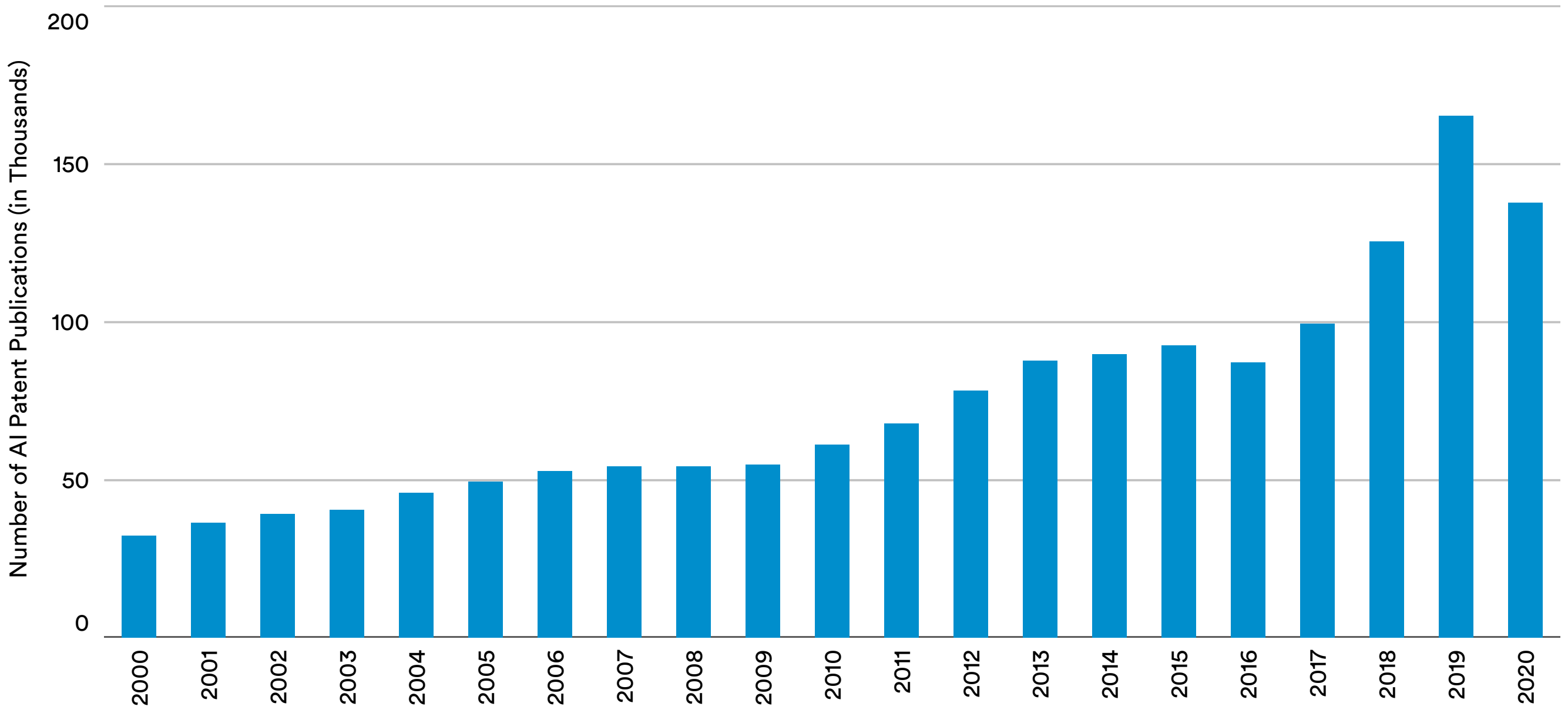

Figure 1.5.9a

## OECD DEFINITION: AI PATENT PUBLICATIONS (% of ALL PATENT PUBLICATIONS), 2000-20

Source: Microsoft Academic Graph, 2020 | Chart: 2021 AI Index Report

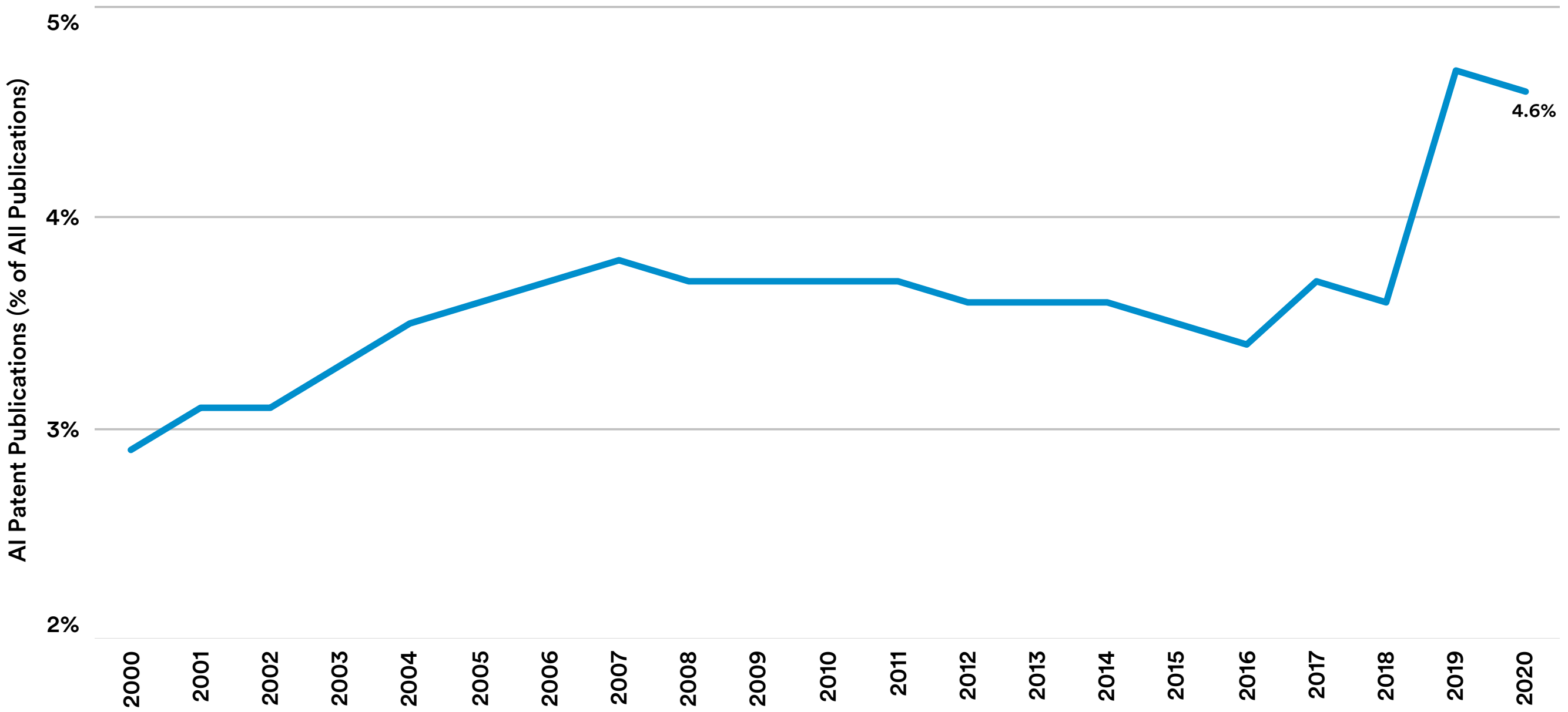

Figure 1.5.9b

## OECD DEFINITION: AI PATENT PUBLICATION (% of WORLD TOTAL) by REGION, 2000-20

Source: Microsoft Academic Graph, 2020 | Chart: 2021 AI Index Report

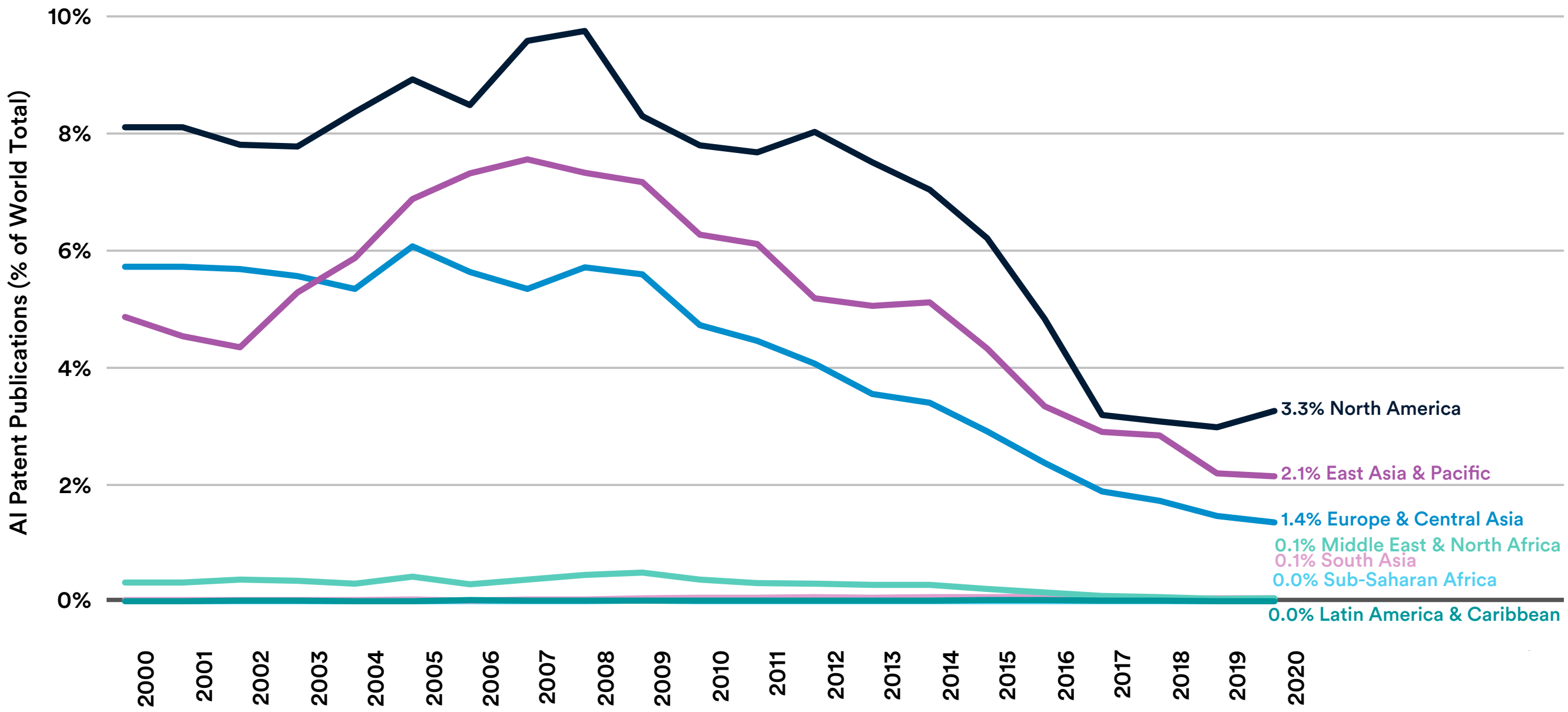

Figure 1.5.10

OECD DEFINITION: AI PATENT PUBLICATIONS (% of WORLD TOTAL) by GEOGRAPHIC AREA, 2000-20
Source: Microsoft Academic Graph, 2020 | Chart: 2021 AI Index Report

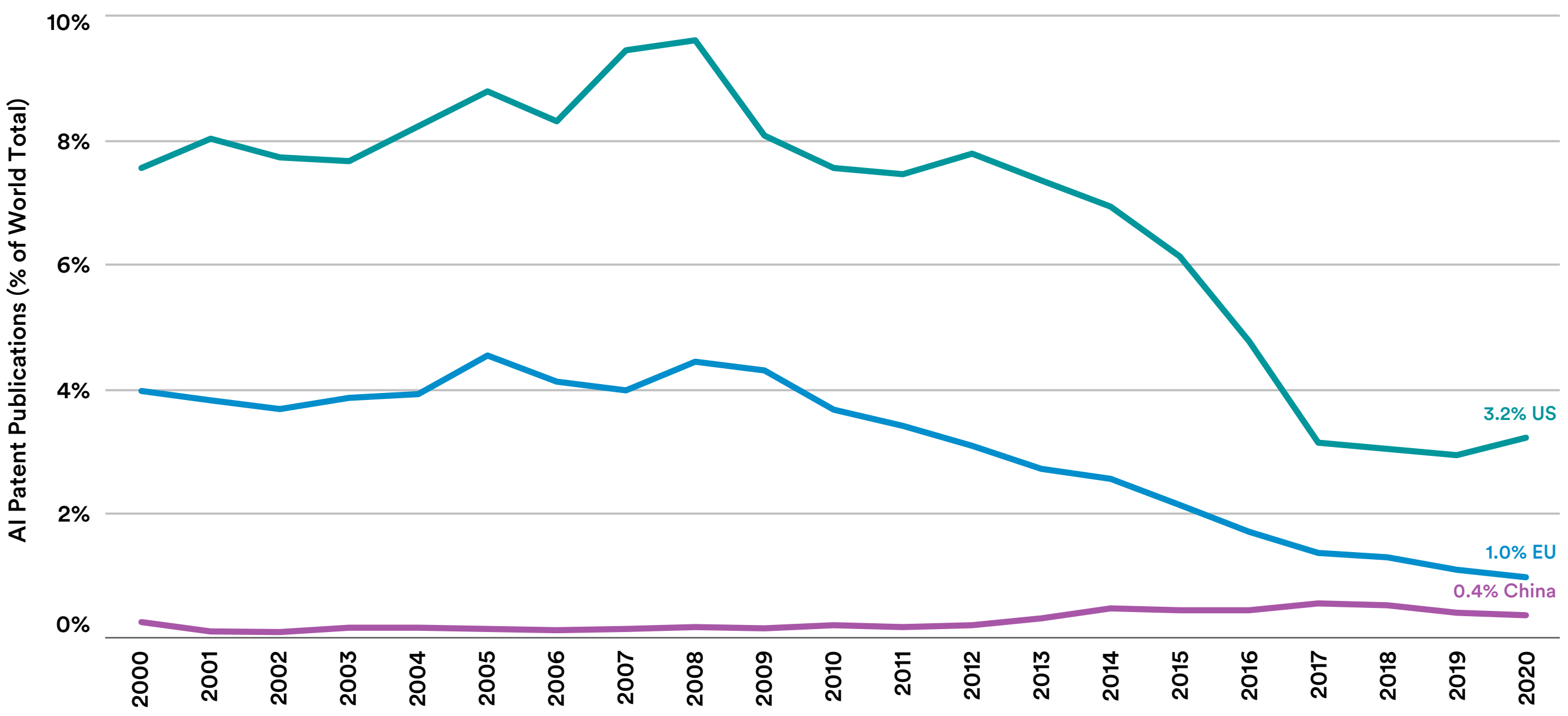

Figure 1.5.11

OECD DEFINITION: AI PATENT CITATION (% of WORLD TOTAL) by GEOGRAPHIC AREA, 2000-20
Source: Microsoft Academic Graph, 2020 | Chart: 2021 AI Index Report

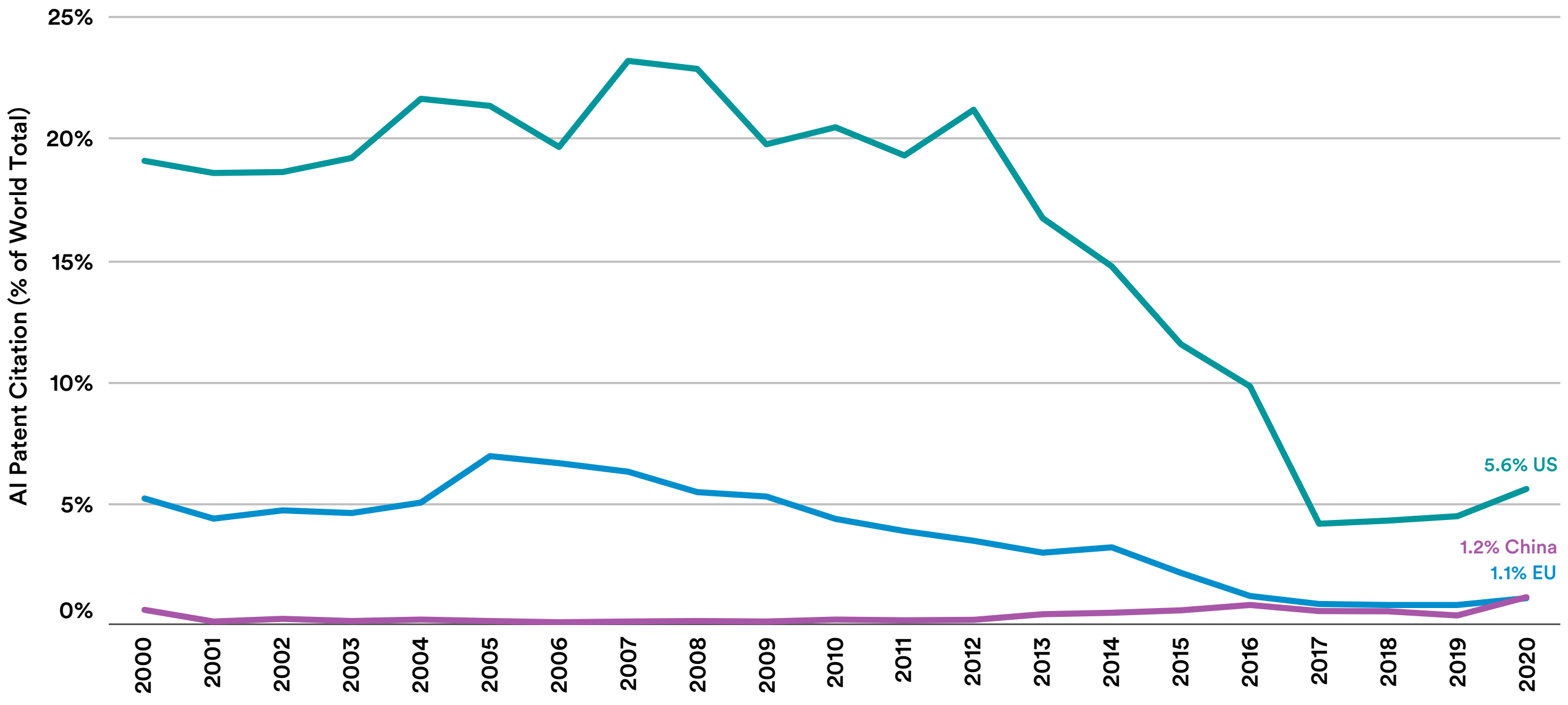

Figure 1.5.12

## PAPERS ON ARXIV

Prepared by Jim Entwood and Eleonora Presani

### Source

arXiv.org is an online archive of research articles in the fields of physics, mathematics, computer science, quantitative biology, quantitative finance, statistics, electrical engineering and systems science, and economics. arXiv is owned and operated by Cornell University. See more information on arXiv.org.

### Methodology

Raw data for our analysis was provided by representatives at arXiv.org. The keywords we selected, and their respective categories, are below:

Artificial intelligence (cs.AI)
Computation and language (cs.CL)
Computer vision and pattern recognition (cs.CV)
Machine learning (cs.LG)
Neural and evolutionary computing (cs.NE)
Robotics (cs.RO)
Machine learning in stats (stats.ML)

For most categories, arXiv provided data for 2015–2020. To review other categories' submission rates on arXiv, see arXiv.org's submission statistics.

The arXiv team has been expanding the publicly available submission statistics. This is a tableau-based application with tabs at the top for various displays of submission stats and filters on the side bar to drill down by topic. (Hover over the charts to view individual categories.) The data is meant to be displayed on a monthly basis with download options.

arXiv is actively looking at ways to improve how it can better support AI/ML researchers as the field grows and discovering content becomes more challenging. For example, there may be ways to create finer grained categories in arXiv for machine learning to help researchers in subfields share and find work more easily. The other rapidly expanding area is computer vision, where there is considerable overlap for ML applications of computer vision.

### Nuance

- Categories are self-identified by authors—those shown are selected as the "primary" category. Thus there is not a single automated categorization process. Additionally, the artificial intelligence or machine learning categories may be categorized by other subfields or keywords.
- arXiv team members suggest that participation on arXiv can breed greater participation, meaning that an increase in a subcategory on arXiv could drive over-indexed participation by certain communities.

## NESTA

Prepared by Joel Kliger and Juan Mateos-Garcia

### Source

Details can be found in the following publication: Deep Learning, Deep Change? Mapping the Development of the Artificial Intelligence General Purpose Technology

### Methodology

Deep learning papers were identified through a topic modeling analysis of the abstracts of arXiv papers in the CS (computer science) and stats.ML (statistics: machine learning category) arXiv categories. The data was enriched with institutional affiliation and geographic information from the Microsoft Academic Graph and the Global Research Identifier. Nesta's arXlive tool is available here.

### Access the Code

The code for data collection and processing can be found here; or, without the infrastructure overhead here.

## GITHUB STARS

### Source

GitHub: star-history (available at star history website) was used to retrieve the data.

### Methodology

The visual in the report shows the number of stars for various GitHub repositories over time. The repositories include the following:
apache/incubator-mxnet, BVLC/cafe, cafe2/cafe2, dmlc/mxnet, fchollet/keras, Microsoft/CNTK, pytorch/pytorch, scikit-learn/scikit-learn, tensorflow/tensorflow, Theano/Theano, Torch/Torch7.

### Nuance

The GitHub Archive currently does not provide a way to count when users remove a star from a repository. Therefore, the reported data slightly overestimates the number of stars. A comparison with the actual number of stars for the repositories on GitHub reveals that the numbers are fairly close and that the trends remain unchanged.

# CHAPTER 2: TECHNICAL PERFORMANCE

## IMAGENET: ACCURACY

Prepared by Jörg Hellwig and Thomas A. Collins

### Source

Data on ImageNet accuracy was retrieved through an arXiv literature review. All results reported were tested on the LSRVC 2012 validation set, as the results on the test set, which are not significantly different, are not public. Their ordering may differ from the results reported on the LSRVC website, since those results were obtained on the test set. Dates we report correspond to the day when a paper was first published to arXiv, and top-1 accuracy corresponds to the result reported in the most recent version of each paper. We selected a top result at any given point in time from 2012 to Nov. 17, 2019. Some of the results we mention were submitted to LSRVC competitions over the years. Image classification was part of LSRVC through 2014; in 2015, it was replaced with an object localization task, where results for classification were still reported but no longer a part of the competition, having instead been replaced by more difficult tasks.

For papers published in 2014 and later, we report the best result obtained using a single model (we did not include ensembles) and using single-crop testing. For the three earliest models (AlexNet, ZFNet, Five Base), we reported the results for ensembles of models.

While we report the results as described above, due to the diversity in models, evaluation methods, and accuracy metrics, there are many other ways to report ImageNet performance. Some possible choices include:

- Evaluation set: validation set (available publicly) or test set (available only to LSRVC organizers)
- Performance metric: Top-1 accuracy (whether the correct label was the same as the first predicted label for each image) or top-5 accuracy (whether the correct label was present among the top five predicted labels for each image)
- Evaluation method: single-crop or multi-crop

To highlight progress here in top-5 accuracy, we have taken scores from the following papers, without extra training data:

Fixing the Train-Test Resolution Discrepancy: FixEfficientNet
Adversarial Examples Improve Image Recognition
OverFeat: Integrated Recognition, Localization and Detection Using Convolutional Networks
Local Relation Networks for Image Recognition
Densely Connected Convolutional Networks
Revisiting Unreasonable Effectiveness of Data in Deep Learning Era
Squeeze-and-Excitation Networks
EfficientNet: Rethinking Model Scaling for Convolutional Neural Networks
MultiGrain: A Unified Image Embedding for Classes and Instances
EfficientNet: Rethinking Model Scaling for Convolutional Neural Networks
Billion-Scale Semi-Supervised Learning for Image Classification
GPipe: Efficient Training of Giant Neural Networks Using Pipeline Parallelism
RandAugment: Practical Data Augmentation with No Separate Search
Fixing the Train-Rest Resolution Discrepancy

To highlight progress here in top-5 accuracy, we have taken scores from the following papers, with extra training data:
Meta Pseudo Labels
Self-Training with Noisy Student Improves ImageNet Classification
Big Transfer (BiT): General Visual Representation Learning
ImageNet Classification with Deep Convolutional Neural Networks
ESPNetv2: A Light-Weight, Power Efficient, and General Purpose Convolutional Neural Network
Xception: Deep Learning with Depthwise Separable Convolutions
EfficientNet: Rethinking Model Scaling for Convolutional Neural Networks
Self-training with Noisy Student Improves ImageNet Classification

To highlight progress here in top-1 accuracy, we have taken scores from the following papers, without extra training data:
Fixing the Train-Test Resolution Discrepancy: FixEfficientNet
Adversarial Examples Improve Image Recognition
OverFeat: Integrated Recognition, Localization and Detection using Convolutional Networks
Densely Connected Convolutional Networks
Revisiting Unreasonable Effectiveness of Data in Deep Learning Era
Dual Path Networks
Res2Net: A New Multi-Scale Backbone Architecture
Billion-Scale Semi-Supervised Learning for Image Classification
Squeeze-and-Excitation Networks
EfficientNet: Rethinking Model Scaling for Convolutional Neural Networks
MultiGrain: A Unified Image Embedding for Classes and Instances
EfficientNet: Rethinking Model Scaling for Convolutional Neural Networks
Billion-Scale Semi-Supervised Learning for Image Classification
EfficientNet: Rethinking Model Scaling for Convolutional Neural Networks
RandAugment: Practical Data Augmentation with No Separate Search
Fixing the Train-Test Resolution Discrepancy

To highlight progress here in top-1 accuracy, we have taken scores from the following papers, without extra training data:
Meta Pseudo Labels
Sharpness-Aware Minimization for Efficiently Improving Generalization
An Image Is Worth 16x16 Words: Transformers for Image Recognition at Scale
Fixing the Train-Test Resolution Discrepancy: FixEfficientNet
Self-training with Noisy Student Improves ImageNet Classification
Big Transfer (BiT): General Visual Representation Learning
ImageNet Classification with Deep Convolutional Neural Networks
ESPNetv2: A Light-Weight, Power Efficient, and General Purpose Convolutional Neural Network
Xception: Deep Learning with Depthwise Separable Convolutions
EfficientNet: Rethinking Model Scaling for Convolutional Neural Networks
Self-training with Noisy Student Improves ImageNet Classification

The estimate of human-level performance is from Russakovsky et al, 2015. Learn more about the LSVRC ImageNet competition and the ImageNet data set.

## IMAGENET: TRAINING TIME

Trends can also be observed by studying research papers that discuss the time it takes to train ImageNet on *any* infrastructure. To gather this data, we looked at research papers from the past few years that tried to optimize for training ImageNet to a standard accuracy level while competing on reducing the overall training time.

### Source

The data is sourced from MLPerf. Detailed data for runs for specific years are available:
2020: MLPerf Training v0.7 Results
2019: MLPerf Training v0.6 Results
2018: MLPerf Training v0.5 Results

### Notes

Data from MLPerf is available in cloud systems for rent. Available On Premise systems contain only components that are available for purchase. Preview systems must be submittable as Available In Cloud or Available on Premise in the next submission round. Research, Development, or Internal (RDI) contain experimental, in development, or internal-use hardware or software. Each row in the results table is a set of results produced by a single submitter using the same software stack and hardware platform. Each row contains the following information:

Submitter: the organization that submitted the results
System: general system description
Processor and count: the type and number of CPUs used, if CPUs perform the majority of ML compute
Accelerator and count: the type and number of accelerators used, if accelerators perform the majority of ML compute
Software: the ML framework and primary ML hardware library used
Benchmark results: training time to reach a specified target quality, measured in minutes
Details: link to metadata for submission
Code: link to code for submission
Notes: arbitrary notes from the submitter

## IMAGENET: TRAINING COST

### Source

DAWNBench is a benchmark suite for end-to-end, deep-learning training and inference. Computation time and cost are critical resources in building deep models, yet many existing benchmarks focus solely on model accuracy. DAWNBench provides a reference set of common deep-learning workloads for quantifying training time, training cost, inference latency, and inference cost across different optimization strategies, model architectures, software frameworks, clouds, and hardware. More details available at DawnBench.

### Note

The DawnBench data source has been deprecated for the period after March 2020, and MLPerf is the most reliable and updated source for AI compute measurements.

## COCO: KEYPOINT DETECTION

The data for COCO keypoint detection data is sourced from COCO keypoints leaderboard.

## COCO: DENSEPOSE ESTIMATION

We gathered data from the CODALab 2020 challenge and read arXiv repository papers to build comprehensive data on technical progress in this challenge. The detailed list of papers and sources used in our survey include:
DensePose: Dense Human Pose Estimation In the Wild
COCO-DensePose 2018 CodaLab
Parsing R-CNN for Instance-Level Human Analysis
Capture Dense: Markerless Motion Capture Meets Dense Pose Estimation
Slim DensePose: Thrifty Learning from Sparse Annotations and Motion Cues
COCO-DensePose 2020 CodaLab
Transferring Dense Pose to Proximal Animal Classes
Making DensePose Fast and Light
SimPose: Effectively Learning DensePose and Surface Normals of People from Simulated Data

## ACTIVITYNET: TEMPORAL LOCALIZATION TASK

In the challenge, there are three separate tasks, but they focus on the main problem of temporally localizing where activities happen in untrimmed videos from the ActivityNet benchmark. We have compiled several attributes for the task of temporal localization at the challenge over the last four rounds. Below is a link to the overall stats and trends for this task, as well as some detailed analysis (e.g., how has the performance for individual activity classes improved over the years? Which are the hardest and easiest classes now? Which classes have the most improvement over the years?). See the Performance Diagnosis (2020) tab for a detailed trends update. Please see ActivityNet Statisticsin the public data folder for more details.

## YOLO (YOU ONLY LOOK ONCE)

YOLO is a neural network model mainly used for the detection of objects in images and in real-time videos. mAP (mean average precision) is a metric that is used to measure the accuracy of object detectors. It is a combination of precision and recall. mAP is the average of the precision and recall calculated over a document. The performance of YOLO has increased gradually with the development of new architectures and versions in past years. With the increase in size of model, its mean average precision increases as well, with a corresponding decrease in FPS of the video.

We conducted a detailed survey of arXiv papers and GitHub repository to segment progress in YOLO across its various versions. Below are the references for original sources:

YOLOv1:
You Only Look Once: Unified, Real-Time Object Detection

YOLOv2:
YOLO9000: Better, Faster, Stronger
YOLO: Real-Time Object Detection

YOLOv3:
YOLOv3: An Incremental Improvement
Learning Spatial Fusion for Single-Shot Object Detection
GitHub: ultralytics/yolov3

YOLOv4:
YOLOv4: Optimal Speed and Accuracy of Object Detection
GitHub: AlexeyAB/darknet

YOLOv5:
GitHub: ultralytics/yolov5

PP-YOLO:
PP-YOLO: An Effective and Efficient Implementation of Object Detector

POLY-YOLO:
Poly-YOLO: Higher Speed, More Precise Detection and Instance Segmentation for YOLOV3

## VISUAL QUESTION ANSWERING (VQA)

VQA accuracy data was provided by the VQA team. Learn more about VQA here. More details on VQA 2020 are available here.

### Methodology

Given an image and a natural language question about the image, the task is to provide an accurate natural language answer. The challenge is hosted on the VQA Challenge website. The challenge is hosted on EvalAI. The challenge link is here.

The VQA v2.0 training, validation, and test sets, containing more than 250,000 images and 1.1 million questions, are available on the download page. All questions are annotated with 10 concise, open-ended answers each. Annotations on the training and validation sets are publicly available.

VQA Challenge 2020 is the fifth edition of the VQA Challenge. Results from previous versions of the VQA Challenge were announced at the VQA Challenge Workshop in CVPR 2019, CVPR 2018, CVPR 2017, and CVPR 2016. More details about past challenges can be found here: VQA Challenge 2019, VQA Challenge 2018, VQA Challenge 2017, VQA Challenge 2016.

VQA had 10 humans answer each question. More details about the VQA evaluation metric and human accuracy can be found here (see Evaluation Code section) and in sections three ("Answers") and four ("Inter-Human Agreement") of the paper.

See slide 56 for the progress graph in VQA in the 2020 Challenge. The values corresponding to the progress graph are available in a sheet. Here is the information about the teams that participated in the 2020 challenge and their accuracies. For more details about the teams, please refer to the VQA website.

## PAPERS WITH CODE: PAPER AND CODE LINKING

We used paperswithcode (PWC) for referencing technical progress where available. Learn more about PWC here and see the public link here.

### Methodology

For papers, we follow specific ML-related categories on arxiv (see [1] below for the full list) and the major ML conferences (NeurIPS, ICML, ICLR, etc.). For code, we follow GitHub repositories mentioning papers. We have good coverage of core ML topics but are missing some applications—for instance, applications of ML in medicine or bioinformatics, which are usually in journals behind paywalls. For code, the dataset is fairly unbiased (as long as the paper is freely available).

For tasks (e.g., "image classification"), the dataset has annotated those on 1,600 state-of-the-art papers from the database, published in 2018 Q3.

For state-of-the-art tables (e.g., "image classification on ImageNet"), the data has been scraped from different sources (see the full list here), and a large number focusing on CV and NLP were hand-annotated. A significant portion of our data was contributed by users, and they have added data based on their own preferences and interests. Arxiv categories we follow:
ARXIV_CATEGORIES = "cs.CV", "cs.AI", "cs.LG", "cs.CL", "cs.NE", "stat.ML","cs.IR"}

### Process of Extracting Dataset at Scale

1) Follow various paper sources (as described above) for new papers.
2) Conduct a number of predefined searches on GitHub (e.g., for READMEs containing links to arxiv).
3) Extract GitHub links from papers.
4) Extract paper links from GitHub.
5) Run validation tests to decide if links from 3) and 4) are bona fide links or false positives.
6) Let the community fix any errors and/or add any missing values.

## NIST FRVT

### Source

There are two FRVT evaluation leaderboards available here: 1:1 Verification and 1:N Identification

### Nuances about FRVT evaluation metrics

Wild Photos have some identity labeling errors as the best algorithm has a low false non-match rate (FNMR), but obtaining complete convergence is difficult. This task will be retired in the future. The data became public in 2018 and has become easier over time. Wild is coming from public web sources. So it is possible those same images have been scrapped from the web by developers. There is no training in the FRVT data, only test data.

The 1:1 and 1:N should be studied separately. The differences include algorithmic approaches, particularly fast search algorithms are especially useful in 1:N whereas speed is not a factor in 1:1.

## SUPERGLUE

The SuperGLUE benchmark data was pulled from the SuperGLUE leaderboard. Details about the SuperGLUE benchmark are in the SuperGLUE paper and SuperGLUE software toolkit. The tasks and evaluation metrics for SuperGLUE are:

| NAME | IDENTIFIER | METRIC |
|---|---|---|
| Broad Coverage Diagnostics | AX-b | Matthew's Corr |
| CommitmentBank | CB | Avg. F1 / Accuracy |
| Choice of Plausible Alternatives | COPA | Accuracy |
| Multi-Sentence Reading Comprehension | MultiRC | F1a / EM |
| Recognizing Textual Entailment | RTE | Accuracy |
| Words in Context | WiC | Accuracy |
| The Winograd Schema Challenge | WSC | Accuracy |
| BoolQ | BoolQ | Accuracy |
| Reading Comprehension with Commonsense Reasoning | ReCoRD | F1 / Accuracy |
| Winogender Schema Diagnostics | AX-g | Gender Parity / Accuracy |

## VISUAL COMMONSENSE REASONING (VCR)

Technical progress for VCR is taken from the VCR leaderboard. VCR has two different subtasks:

- Question Answering (Q->A): A model is provided a question and has to pick the best answer out of four choices. Only one of the four is correct.
- Answer Justification (QA->R): A model is provided a question, along with the correct answer, and it must justify it by picking the best rationale among four choices.

The two parts with the Q->AR metrics are combined in which a model only gets a question right if it answers correctly and picks the right rationale. Models are evaluated in terms of accuracy (%).

## VOXCELEB

VoxCeleb is an audio-visual dataset consisting of short clips of human speech, extracted from interview videos uploaded to YouTube. VoxCeleb contains speech from 7,000-plus speakers spanning a wide range of ethnicities, accents, professions, and ages—amounting to over a million utterances (face-tracks are captured "in the wild," with background chatter, laughter, overlapping speech, pose variation, and different lighting conditions) recorded over a period of 2,000 hours (both audio and video). Each segment is at least three seconds long. The data contains an audio dataset based on celebrity voices, shorts, films, and conversational pieces (e.g., talk shows). The initial VoxCeleb 1 (100,000 utterances taken from 1,251 celebrities on YouTube) was expanded to VoxCeleb 2 (1 million utterances from 6,112 celebrities).

However, in earlier years of the challenge, top-1 and top-5 scores were also reported. For top-1 score, the system is correct if the target label is the class to which it assigns the highest probability. For top-5 score, the system is correct if the target label is one of the five predictions with the highest probabilities. In both cases, the top score is computed as the number of times a predicted label matches the target label, divided by the number of data points evaluated.

The data is extracted from different years of the submission challenges, including:

- 2017: VoxCeleb: A Large-Scale Speaker Identification Dataset
- 2018: VoxCeleb2: Deep Speaker Recognition
- 2019: Voxceleb: Large-Scale Speaker Verification in the Wild
- 2020: Query ExpansionSystem for the VoxCeleb Speaker Recognition Challenge 2020

## BOOLEAN SATISFIABILITY PROBLEM

Analysis and text by Lars Kotthoff

### Primary Source and Data Sets

The Boolean Satisfiability Problem (SAT) determines whether there is an assignment of values to a set of Boolean variables joined by logical connectives that makes the logical formula it represents true. SAT was the first problem to be proven NP-complete, and the first algorithms to solve it were developed in the 1960s. Many real-world problems, such as circuit design, automated theorem proving, and scheduling, can be represented and solved efficiently as SAT. The annual SAT competition is designed to present a snapshot of the state-of-the-art and has been running for almost 20 years.

We took the top-ranked, median-ranked, and bottom-ranked solvers from each of the last five years (2016-2020) of the SAT competition. We ran all 15 solvers on all 400 SAT instances from the main track of the 2020 competition. More information on the competition, as well as the solvers and instances, is available at the SAT competition website.

### Results

We ran each solver on each instance on the same hardware, with a time limit of 5,000 CPU seconds per instance, and measured the time it took a solver to solve an instance in CPU seconds. Ranked solvers always return correct results, hence we do not consider correctness as a metric. Except for the 2020 competition solvers, we evaluated the performance of the SAT solvers on a set of instances different from the set of instances they competed on. Further, our hardware is different from what was used for the SAT competition. The results we report here will therefore differ from the exact results reported for the respective SAT competitions.

The Shapley value is a concept from cooperative game theory that assigns a contribution to the total value that a coalition generates to each player. It quantifies how important each player is for the coalition and has several desirable properties that make the distribution of the total value to the individual players fair. For example, the Shapley value is used to distribute airport costs to its users, allocate funds to different marketing campaigns, and in machine learning, where it helps render complex black-box models more explainable.

In our context, it quantifies the contribution of a solver to the state-of-the-art through the average performance improvement it provides over a set of other solvers and over all subsets of solvers (Fréchette et al. (2016)). For a given set of solvers, we choose the respective best for each instance to solve. By including another solver and being able to choose it, overall solving performance improves, with the difference to the original set of solvers being the marginal contribution of the added solver. The average marginal contribution to all sets of solvers is the Shapley value.

Quantifying the contribution of a solver through the Shapley value compares solvers from earlier competitions to solvers in later competitions. This is often not a fair comparison, as later solvers are often improved versions of earlier solvers, and the contribution of the solver to the future state-of-the-art will always be low. The temporal Shapley value (Kotthoff et al. (2018)) solves this problem by considering the time a particular solver was introduced when quantifying its contribution to the state-of-the-art.

## AUTOMATED THEOREM PROVING

Analysis and text by Christian Suttner, Geoff Sutcliffe, and Raymond Perrault

### 1. Motivation

Automated Theorem Proving (ATP) (also referred to as Automated Deduction) is a subfield of automated reasoning, concerned with the development and use of systems that automate sound reasoning: the derivation of conclusions that follow inevitably from facts. ATP systems are at the heart of many computational tasks and are used commercially, e.g., for integrated circuit design and computer program verification. ATP problems are typically solved by showing that a conjecture is or is not a logical consequence of a set of axioms. ATP problems are encoded in a chosen logic, and an ATP system for that logic is used to (attempt to) solve the problem. A key concern of ATP research is the development of more powerful systems, capable of solving more difficult problems within the same resource limits. In order to assess the merits of new techniques, sound empirical evaluations of ATP systems are key.

### 2. Analysis

For the evaluation of ATP systems, there exists a large and growing collection of problems called the TPTP problem library. The current release v7.4.0 (released June 10, 2020) contains 23,291 ATP problems, structured into 54 topic domains (e.g., Set Theory, Software Verification, Philosophy, etc.). Orthogonally, the TPTP is divided into Specialist Problem Classes (SPCs), each of which contains problems with a specified set of logical, language, and syntactic characteristics (e.g. first-order logic theorems with some use of equality). The SPCs allow ATP system developers to select problems and evaluate their systems appropriately. Since its first release in 1993, many researchers have used the TPTP as an appropriate and convenient basis for ATP system evaluation. Over the years, the TPTP has also increasingly been used as a conduit for ATP users to contribute samples of their problems to ATP system developers. This exposes the problems to ATP system developers, who can then improve their systems' performances on the problems, which completes a cycle to provide users with more effective tools.

Associated with the TPTP is the TSTP solution library, which maintains updated results from running all current versions of ATP systems (available to the maintainer) on all the TPTP problems. One use of the TSTP is to compute TPTP problem difficulty ratings: Easy problems, which are solved by all ATP systems, have a rating of 0.0; difficult problems, which are solved by some ATP systems, have ratings between 0.0 and 1.0; unsolved problems, which are not solved by any ATP system, have a rating of 1.0. Note that the rating for a problem is not strictly decreasing, as different ATP systems and versions become available for populating the TSTP. The history of each TPTP problem's ratings is saved with the problem, which makes it possible to tell when the problem was first solved by any ATP system (the point at which its rating dropped below 1.0). That information has been used here to obtain an indication of progress in the field.

The simplest way to measure progress takes a fixed set of problems that has been available (and unchanged) in the TPTP from some chosen initial TPTP release, and then for the TPTP releases from then on, counts how many of the problems had been solved from that release. The analysis reports the fraction of problems solved for each release. This simple approach is unambiguous, but it does not take into account new problems that are added to the TPTP after the initial release.

The analysis used here extends the "Fixed Set" analysis, taking into account new problems added after the initial release. As it is not possible to run all previously available ATP systems on new problems when they are added, this approach assumes that if a problem is unsolved by current ATP systems when it is added to the TPTP, then it would have been unsolved by previously available ATP systems. Under that assumption, the new problem is retrospectively "added" to prior TPTP releases for the analysis. If a problem is solved when it is added to the TPTP, it is ignored because it may have been solved in prior versions as well, and therefore should not serve as an indicator of progress. This analysis reports the fraction of problems solved for each release, but note that the fraction is with respect to both the number of problems actually in the release and also the problems retrospectively "added."

The growing set analysis is performed on the whole TPTP and on four SPCs. These were chosen because many ATP problems in those forms have been contributed to the TPTP, and correspondingly there are many ATP systems that can attempt them; they represent the "real world" demand for ATP capability.

The table here in the public data folder shows the breakdown of TPTP problems by content fields, as well as by SPCs used in the analysis. The totals are slightly larger than those shown in the analysis, as some problems were left out for technical reasons (no scores available, problems revised over time, etc.).

# CHAPTER 3: ECONOMY

## LINKEDIN

Prepared by Mar Carpanelli, Ramanujam MV, and Nathan Williams

### Country Sample

Included countries represent a select sample of eligible countries with at least 40% labor force coverage by LinkedIn and at least 10 AI hires in any given month. China and India were included in this sample because of their increasing importance in the global economy, but LinkedIn coverage in these countries does not reach 40% of the workforce. Insights for these countries may not provide as full a picture as other countries, and should be interpreted accordingly.

### Skills

LinkedIn members self-report their skills on their LinkedIn profiles. Currently, more than 35,000 distinct, standardized skills are identified by LinkedIn. These have been coded and classified by taxonomists at LinkedIn into 249 skill groupings, which are the skill groups represented in the dataset. The top skills that make up the AI skill grouping are machine learning, natural language processing, data structures, artificial intelligence, computer vision, image processing, deep learning, TensorFlow, Pandas (software), and OpenCV, among others.

Skill groupings are derived by expert taxonomists through a similarity-index methodology that measures skill composition at the industry level. Industries are classified according to the ISIC 4 industry classification (Zhu et al., 2018).

### AI Skills Penetration

The aim of this indicator is to measure the intensity of AI skills in an entity (in a particular country, industry, gender, etc.) through the following methodology:

- Compute frequencies for all self-added skills by LinkedIn members in a given entity (occupation, industry, etc.) in 2015–2020.
- Re-weight skill frequencies using a TF-IDF model to get the top 50 most representative skills in that entity. These 50 skills compose the “skill genome” of that entity.
- Compute the share of skills that belong to the AI skill group out of the top skills in the selected entity.

**Interpretation:** The AI skill penetration rate signals the prevalence of AI skills across occupations, or the intensity with which LinkedIn members utilize AI skills in their jobs. For example, the top 50 skills for the occupation of engineer are calculated based on the weighted frequency with which they appear in LinkedIn members’ profiles. If four of the skills that engineers possess belong to the AI skill group, this measure indicates that the penetration of AI skills is estimated to be 8% among engineers (e.g., 4/50).

### Relative AI Skills Penetration

To allow for skills penetration comparisons across countries, the skills genomes are calculated and a relevant benchmark is selected (e.g., global average). A ratio is then constructed between a country’s and the benchmark’s AI skills penetrations, controlling for occupations.

**Interpretation:** A country’s relative AI skills penetration of 1.5 indicates that AI skills are 1.5 times as frequent as in the benchmark, for an overlapping set of occupations.

### Global Comparison

For cross-country comparison, we present the relative penetration rate of AI skills, measured as the sum of the penetration of each AI skill across occupations in a given country, divided by the average global penetration of AI skills across the overlapping occupations in a sample of countries.

**Interpretation:** A relative penetration rate of 2 means that the average penetration of AI skills in that country is two times the global average across the same set of occupations.

### Global Comparison: By Industry

The relative AI skills penetration by country for industry provides an in-depth sectoral decomposition of AI skill penetration across industries and sample countries.

**Interpretation:** A country's relative AI skill penetration rate of 2 in the education sector means that the average penetration of AI skills in that country is two times the global average across the same set of occupations in that sector.

### LinkedIn AI Hiring Index

The LinkedIn AI hiring rate is calculated as the total number of LinkedIn members who are identified as AI talent and added a new employer in the same month the new job began, divided by the total number of LinkedIn members in the country. By analyzing only the timeliest data, it is possible to make month-to-month comparisons and account for any potential lags in members updating their profiles.

The baseline time period is typically a year, and it is indexed to the average month/period of interest during that year. The AI hiring rate is indexed against the average annual hiring in 2016; for example, an index of 3.5 for Brazil in 2020 indicates that the AI hiring rate is 3.5 times higher in 2020 than the average in 2016.

**Interpretation:** The hiring index means the rate of hiring in the AI field, specifically how fast each country is experiencing growth in AI hiring.

### Top AI Skills

AI skills most frequently added by members during 2015–2020 period.

## BURNING GLASS TECHNOLOGIES

Prepared by Bledi Taska, Layla O'Kane, and Zhou Zhou

Burning Glass Technologies delivers job market analytics that empower employers, workers, and educators to make data-driven decisions. The company's artificial intelligence technology analyzes hundreds of millions of job postings and real-life career transitions to provide insight into labor market patterns. This real-time strategic intelligence offers crucial insights, such as what jobs are most in demand, the specific skills employers need, and the career directions that offer the highest potential for workers. For more information, visit burning-glass.com.

### Job Posting Data

To support these analyses, Burning Glass mined its dataset of millions of job postings collected since 2010. Burning Glass collects postings from over 45,000 online job sites to develop a comprehensive, real-time portrait of labor market demand. It aggregates job postings, removes duplicates, and extracts data from job postings text. This includes information on job title, employer, industry, and region, as well as required experience, education, and skills.

Job postings are useful for understanding trends in the labor market because they allow for a detailed, real-time look at the skills employers seek. To assess the representativeness of job postings data, Burning Glass conducts a number of analyses to compare the distribution of job postings to the distribution of official government and other third-party sources in the United States. The primary source of government data on U.S. job postings is the Job Openings and Labor Turnover Survey (JOLTS) program, conducted by the Bureau of Labor Statistics.

To understand the share of job openings captured by Burning Glass data, it is important to first note that Burning Glass and JOLTS collect data on job postings differently. Burning Glass data captures new postings: A posting appears in the data only on the first month it is found and is considered a duplicate and removed in subsequent months. JOLTS data captures active postings: A posting appears in the data for every month that it is still actively posted, meaning the same posting can be counted in two or more consecutive months if it has not been filled. To allow for apples-to-apples volume comparison in postings, the Burning Glass data needs to be inflated to account for active postings, not only new postings. The number of postings from Burning Glass can be inflated using the ratio of new jobs to active jobs in Help Wanted OnLine™ (HWOL), a method used in Carnevale, Jayasundera and Repnikov (2014). Based on this calculation, the share of jobs online as captured by Burning Glass is roughly 85% of the jobs captured in JOLTS in 2016.

The labor market demand captured by Burning Glass data represents over 85% of the total labor demand. Jobs not posted online are usually in small businesses (the classic example being the "Help Wanted" sign in the restaurant window) and union hiring halls.

## Measuring Demand for AI

In order to measure the demand by employers of AI skills, Burning Glass uses its skills taxonomy of over 17,000 skills. The list of AI skills from Burning Glass data are shown below, with associated skill clusters. While some skills are considered to be in the AI cluster specifically, for the purposes of this report, all skills below were considered AI skills. A job posting was considered an AI job if it requested one or more of these skills.

**Artificial Intelligence:** Expert System, IBM Watson, IPSoft Amelia, Ithink, Virtual Agents, Autonomous Systems, Lidar, OpenCV, Path Planning, Remote Sensing

**Natural Language Processing (NLP):** ANTLR, Automatic Speech Recognition (ASR), Chatbot, Computational Linguistics, Distinguo, Latent Dirichlet Allocation, Latent Semantic Analysis, Lexalytics, Lexical Acquisition, Lexical Semantics, Machine Translation (MT), Modular Audio Recognition Framework (MARF), MoSes, Natural Language Processing, Natural Language Toolkit (NLTK), Nearest Neighbor Algorithm, OpenNLP, Sentiment Analysis/Opinion Mining, Speech Recognition, Text Mining, Text to Speech (TTS), Tokenization, Word2Vec

**Neural Networks:** Caffe Deep Learning Framework, Convolutional Neural Network (CNN), Deep Learning, Deeplearning4j, Keras, Long Short-Term Memory (LSTM), MXNet, Neural Networks, Pybrain, Recurrent Neural Network (RNN), TensorFlow

**Machine Learning:** AdaBoost algorithm, Boosting (Machine Learning), Chi Square Automatic Interaction Detection (CHAID), Classification Algorithms, Clustering Algorithms, Decision Trees, Dimensionality Reduction, Google Cloud Machine Learning Platform, Gradient boosting, H2O (software), Libsvm, Machine Learning, Madlib, Mahout, Microsoft Cognitive Toolkit, MLPACK (C++ library), Mlpy, Random Forests, Recommender Systems, Scikit-learn, Semi-Supervised Learning, Supervised Learning (Machine Learning), Support Vector Machines (SVM), Semantic Driven Subtractive Clustering Method (SDSCM), Torch (Machine Learning), Unsupervised Learning, Vowpal, Xgboost

**Robotics:** Blue Prism, Electromechanical Systems, Motion Planning, Motoman Robot Programming, Robot Framework, Robotic Systems, Robot Operating System (ROS), Robot Programming, Servo Drives / Motors, Simultaneous Localization and Mapping (SLAM)

**Visual Image Recognition:** Computer Vision, Image Processing, Image Recognition, Machine Vision, Object Recognition

# NETBASE QUID

Prepared by Julie Kim

NetBase Quid is a big data analytics platform that inspires full-picture thinking by drawing connections across massive amounts of unstructured data. The software applies advanced natural language processing technology, semantic analysis, and artificial intelligence algorithms to reveal patterns in large, unstructured datasets and to generate visualizations that allow users to gain actionable insights. NetBase Quid uses Boolean query to search for focus areas, topics, and keywords within the archived news and blogs, companies, and patents database, as well as any custom uploaded datasets. This can filter out the search by published date time frame, source regions, source categories, or industry categories on the news—and by regions, investment amount, operating status, organization type (private/public), and founding year within the companies database. NetBase Quid then visualizes these data points based on the semantic similarity.

## Search, Data Sources, and Scope

Here 3.6 million public and private company profiles from multiple data sources are indexed in order to search across company descriptions, while filtering and including metadata ranging from investment information to firmographic information, such as founded year, HQ location, and more. Company information is updated on a weekly basis. Quid algorithm reads a big amount of text data from each document (news article, company descriptions, etc.) to make links between different documents based on their similar language. This process is repeated at an immense scale, which produces a network with different clusters identifying distinct topics or focus areas. Trends are identified based on keywords, phrases, people, companies, institutions that Quid identifies, and the other metadata that is put into the software.

## Data

Organization data is embedded from CapIQ and Crunchbase. These companies include all types of companies (private, public, operating, operating as a subsidiary, out of business) throughout the world. The investment data includes private investments, M&A, public offerings, minority stakes made by PE/VCs, corporate venture arms, governments, and institutions both within and outside the United States. Some data is simply unreachable—for instance, when the investors are undisclosed or the funding amounts by investors are undisclosed. Quid also embeds firmographic information such as founded year and HQ location.

NetBase Quid embeds CapIQ data as a default and adds in data from Crunchbase for the ones that are not captured in CapIQ. This yields not only comprehensive and accurate data on all global organizations, but it also captures early-stage startups and funding events data. Company information is uploaded on a weekly basis.

## Methodology

Boolean query is used to search for focus areas, topics, and keywords within the archived company database, within their business descriptions and websites. We can filter out the search results by HQ regions, investment amount, operating status, organization type (private/public), and founding year. Quid then visualizes these companies. If there are more than 7,000 companies from the search result, Quid selects the 7,000 most relevant companies for visualization based on the language algorithm.

**Boolean Search:** “artificial intelligence” or “AI” or “machine learning” or “deep learning”

**Companies:**

- Chart 3.2.1: Global AI & ML companies that have been invested (private, IPO, M&A) from 01/01/2011 to 12/31/2020.
- Chart 3.2.2–3.2.6: Global AI & ML companies that have invested over USD 400,000 for the last 10 years (January 1, 2011 to December 31, 2020)—7,000 companies out of 7,500 companies have been selected through Quid’s relevance algorithm.

### Target Event Definitions

- Private investments: A private placement is a private sale of newly issued securities (equity or debt) by a company to a selected investor or a selected group of investors. The stakes that buyers take in private placements are often minority stakes (under 50%), although it is possible to take control of a company through a private placement as well, in which case the private placement would be a majority stake investment.
- Minority investment: These refer to minority stake acquisitions in Quid, which take place when the buyer acquires less than 50% of the existing ownership stake in entities, asset product, and business divisions.
- M&A: This refers to a buyer acquiring more than 50% of the existing ownership stake in entities, asset product, and business divisions.

## MCKINSEY & COMPANY

### SOURCE

This survey was written, filled, and analyzed by McKinsey & Company. You can find additional results from the Global AI Survey here.

### Methodology

The survey was conducted online and was in the field from June 9, 2020, to June 19, 2020, and garnered responses from 2,395 participants representing the full range of regions, industries, company sizes, functional specialties, and tenures. Of those respondents, 1,151 said their organizations had adopted AI in at least one function and were asked questions about their organizations' AI use. To adjust for differences in response rates, the data are weighted by the contribution of each respondent's nation to global GDP. McKinsey also conducted interviews with executives between May and August 2020 about their companies' use of AI. All quotations from executives were gathered during those interviews.

### Note

Survey respondents are limited by their perception of their organization's AI adoption.

## INTERNATIONAL FEDERATION OF ROBOTICS

### Source

Data was received directly from the International Federation of Robotics' (IFR) 2020 World Robotics Report. Learn more about IFR.

### Methodology

The data displayed is the number of industrial robots installed by country. Industrial robots are defined by the ISO 8373:2012 standard. See more information on IFR's methodology.

### Nuance

- It is unclear how to identify what percentage of robot units run software that would be classified as "AI," and it is unclear to what extent AI development contributes to industrial robot usage.
- This metric was called "robot imports" in the 2017 AI Index Report.

## PRATTLE (EARNING CALLS ONLY)

Prepared by Jeffrey Banner and Steven Nichols

### Source

Liquidnet provides sentiment data that predicts the market impact of central bank and corporate communications. Learn more about Liquidnet here.

# CHAPTER 4: AI EDUCATION

## CRA TAULBEE SURVEY

Prepared by Betsy Bizot (CRA senior research associate) and Stu Zweben (CRA survey chair, professor emeritus at The Ohio State University)

### Source

Computing Research Association (CRA) members are 200-plus North American organizations active in computing research: academic departments of computer science and computer engineering; laboratories and centers in industry, government, and academia; and affiliated professional societies (AAAI, ACM, CACS/AIC, IEEE Computer Society, SIAM USENIX). CRA's mission is to enhance innovation by joining with industry, government, and academia to strengthen research and advanced education in computing. Learn more about CRA here.

### Methodology

CRA Taulbee Survey gathers survey data during the fall of each academic year by reaching out to over 200 PhD-granting departments. Details about the Taulbee Survey can be found here. Taulbee does not directly survey the students. The department identifies each new PhD's area of specialization as well as their type of employment. Data is collected from September to January of each academic year for PhDs awarded in the previous academic year. Results are published in May after data collection closes. So the 2019 data points were newly available last spring, and the numbers provided for 2020 will be available in May 2021.

The CRA Taulbee Survey is sent only to doctoral departments of computer science, computer engineering, and information science/systems. Historically, (a) Taulbee covers 1/4 to 1/3 of total BS CS recipients in the United States; (b) the percent of women earning bachelor's degrees is lower in the Taulbee schools than overall; and (c) Taulbee tracks the trends in overall CS production.

### Nuances

- Of particular interest in PhD job market trends are the metrics on the AI PhD area of specialization. The categorization of specialty areas changed in 2008 and was clarified in 2016. From 2004-2007, AI and robotics were grouped; from 2008-present, AI is separate; 2016 clarified to respondents that AI includes ML.
- Notes about the trends in new tenure-track hires (overall and particularly at AAU schools): In the 2018 Taulbee Survey, for the first time, we asked how many new hires had come from the following sources: new PhD, postdoc, industry, and other academic. Results indicated that 29% of new assistant professors came from another academic institution.
- Some may have been teaching or research faculty rather than tenure-track, but there is probably some movement between institutions, meaning the total number hired overstates the total who are actually new.

## AI INDEX EDUCATION SURVEY

Prepared by Daniel Zhang (Stanford Institute for Human-Centered Artificial Intelligence)

### Methodology

The survey was distributed to 73 universities online over three waves from November 2020 to January 2021 and completed by 18 universities, a 24.7% response rate. The selection of universities is based on the World University Rankings 2021 and Emerging Economies University Rankings 2020 by The Times Higher Education.

The 18 universities are:

- Belgium: Katholieke Universiteit Leuven
- Canada: McGill University
- China: Shanghai Jiao Tong University, Tsinghua University
- Germany: Ludwig Maximilian University of Munich, Technical University of Munich
- Russia: Higher School of Economics, Moscow Institute of Physics and Technology
- Switzerland: École Polytechnique Fédérale de Lausanne
- United Kingdom: University of Cambridge
- United States: California Institute of Technology, Carnegie Mellon University (Department of Machine Learning), Columbia University, Harvard University, Stanford University, University of Wisconsin–Madison, University of Texas at Austin, Yale University

### Key Definitions

- **Major or a study program:** a set of required and elective courses in an area of discipline—such as AI—that leads to a bachelor's degree upon successful completion.
- **Course:** a set of classes that require a minimum of 2.5 class hours (including lecture, lab, TA hours, etc.) per week for at least 10 weeks in total. Multiple courses with the same titles and numbers count as one course.
- **Practical Artificial Intelligence Models - Keywords:** Adaptive learning, AI Application, Anomaly detection, Artificial general intelligence, Artificial intelligence, Audio processing, Automated vehicle, Automatic translation, Autonomous system, Autonomous vehicle, Business intelligence, Chatbot, Computational creativity, Computational linguistics, Computational neuroscience, Computer vision, Control theory, Cyber physical steam, Deep learning, Deep neural network, Expert system, Face recognition, Human-AI interaction, Image processing, Image recognition, Inductive programming, Intelligence software, Intelligent agent, Intelligent control, Intelligent software development, Intelligence system, Knowledge representation and reasoning, Machine learning, Machine translation, Multi-agent system, Narrow artificial intelligence, Natural language generation, Natural language processing, Natural language understanding, Neural network, Pattern recognition, Predictive analysis, Recommender system, Reinforcement learning, Robot system, Robotics, Semantic web, Sentiment analysis, Service robot, Social robot, Sound synthesis, Speaker identification, Speech processing, Speech recognition, Speech synthesis, Strong artificial intelligence, Supervised learning, Support vector machine, Swarm intelligence, Text mining, Transfer learning, Unsupervised learning, Voice recognition, Weak artificial intelligence (Adapted from: Joint Research Centre, European Commission, p.68)
- **AI Ethics - Keywords:** Accountability, Consent, Contestability, Ethics, Equality, Explainability, Fairness, Non-discrimination, Privacy, Reliability, Safety, Security, Transparency, Trustworthy ai, Uncertainty, Well-being (Adapted from: Joint Research Centre, European Commission, p.68)

## EU ACADEMIC OFFERING, JOINT RESEARCH CENTER, EUROPEAN COMMISSION

Prepared by Giuditta De-Prato, Montserrat López Cobo, and Riccardo Righi

### Source

The Joint Research Centre (JRC) is the European Commission's science and knowledge service. The JRC employs scientists to carry out research in order to provide independent scientific advice and support to EU policy. Learn more about JRC here.

### Methodology

By means of text-mining techniques, the study identifies AI-related education programs from the programs' descriptions present in JRC's database. To query the database, a list of domain-specific keywords is obtained through a multistep methodology involving (i) selection of top keywords from AI-specific scientific journals; (ii) extraction of representative terms of the industrial dimension of the technology; (iii) topic modeling; and (iv) validation by experts. In this edition, the list of keywords has been enlarged to better cover certain AI subdomains and to expand to related transversal domains, such as philosophy and ethics in AI. Then the keywords are grouped into categories, which are used to analyze the content areas taught in the identified programs. The content areas used are adapted from the JRC report "Defining Artificial Intelligence: Towards an Operational Definition and Taxonomy of Artificial Intelligence," conducted in the context of AI Watch.

The education programs are classified into specialized and broad, according to the depth with which they address artificial intelligence. Specialized programs are those with a strong focus in AI, e.g., "automation and computer vision" or "advanced computer science (computational intelligence)." Broad programs target the addressed domain, but in a more generic way, usually aiming at building wider profiles or making reference to the domain in the framework of a program specialized in a different discipline (e.g., biomedical engineering).

Due to some methodological improvements introduced in this edition, namely the addition of new keywords, a strict comparison is not possible. Still, more than 90% of all detected programs in this edition are triggered by keywords present in the 2019 study.

The original source on which queries are performed is the Studyportals' database, which is made up of over 207,000 programs from 3,700 universities in over 120 countries. Studyportals collects information from institutions' websites, and their database is regularly updated. This source, although offering the widest coverage among all those identified, still suffers from some lack of coverage, mostly because it only tracks English-language programs. This poses a comparability issue between English-native-speaking countries and the rest, but also between countries with differing levels of incorporation of English as a teaching language in higher education. Bachelor's-level studies are expected to be more affected by this fact, where the offer is mostly taught in a native language, unlike master's, which attracts more international audiences and faculties. As a consequence, this study may be showing a partial picture of the level of inclusion of advanced digital skills in bachelor's degree programs.

# CHAPTER 5: ETHICAL CHALLENGES OF AI APPLICATIONS

## NETBASE QUID

Prepared by Julie Kim

Quid is a data analytics platform within the NetBase Quid portfolio that applies advanced natural language processing technology, semantic analysis, and artificial intelligence algorithms to reveal patterns in large, unstructured datasets and generate visualizations to allow users to gain actionable insights. Quid uses Boolean query to search for focus areas, topics, and keywords within the archived news and blogs, companies, and patents database, as well as any custom uploaded datasets. Users can then filter their search by published date time frame, source regions, source categories, or industry categories on the news; and by regions, investment amount, operating status, organization type (private/public), and founding year within the companies' database. Quid then visualizes these data points based on the semantic similarity.

### Network

Searched for [AI technology keywords + Harvard ethics principles keywords] global news from January 1, 2020, to December 31, 2020.

Search Query: (AI OR ["artificial intelligence"]("artificial intelligence" OR "pattern recognition" OR algorithms) OR ["machine learning"]("machine learning" OR "predictive analytics" OR "big data" OR "pattern recognition" OR "deep learning") OR ["natural language"]("natural language" OR "speech recognition") OR NLP OR "computer vision" OR ["robotics"]("robotics" OR "factory automation") OR "intelligent systems" OR ["facial recognition"]("facial recognition" OR "face recognition" OR "voice recognition" OR "iris recognition") OR ["image recognition"]("image recognition" OR "pattern recognition" OR "gesture recognition" OR "augmented reality") OR ["semantic search"]("semantic search" OR "data-mining" OR "full-text search" OR "predictive coding") OR "semantic web" OR "text analytics" OR "virtual assistant" OR "visual search") AND (ethics OR "human rights" OR "human values" OR "responsibility" OR "human control" OR "fairness" OR discrimination OR non-discrimination OR "transparency" OR "explainability" OR "safety and security" OR "accountability" OR "privacy" )

### News Dataset Data Source

Quid indexes millions of global-source, English-language news articles and blog posts from LexisNexis. The platform has archived news and blogs from August 2013 to the present, updating every 15 minutes. Sources include over 60,000 news sources and over 500,000 blogs.

### Visualization in Quid Software

Quid uses Boolean query to search for topics, trends, and keywords within the archived news database, with the ability to filter results by the published date time frame, source regions, source categories, or industry categories. (In this case, we only looked at global news published from January 1, 2020, to December 31, 2020.) Quid then selects the 10,000 most relevant stories using its NLP algorithm and visualizes de-duplicated unique articles.

## ETHICS IN AI CONFERENCES

Prepared by Marcelo Prates, Pedro Avelar, and Luis C. Lamb

### Source

Prates, Marcelo, Pedro Avelar, Luis C. Lamb. 2018. On Quantifying and Understanding the Role of Ethics in AI Research: A Historical Account of Flagship Conferences and Journals. September 21, 2018.

### Methodology

The percent of keywords has a straightforward interpretation: For each category (classical/trending/ethics), the number of papers for which the title (or abstract, in the case of the AAAI and NeurIPS figures) contains at least one keyword match. The percentages do not necessarily add up to 100% (e.g, classical/trending/ethics are not mutually exclusive). One can have a paper with matches on all three categories.

To achieve a measure of how much Ethics in AI is discussed, ethics-related terms are searched for in the titles of papers in flagship AI, machine learning, and robotics conferences and journals.

The ethics keywords used were the following: Accountability, Accountable, Employment, Ethic, Ethical, Ethics, Fool, Fooled, Fooling, Humane, Humanity, Law, Machine Bias, Moral, Morality, Privacy, Racism, Racist, Responsibility, Rights, Secure, Security, Sentience, Sentient, Society, Sustainability, Unemployment, and Workforce.

The classical and trending keyword sets were compiled from the areas in the most cited book on AI by Russell and Norvig [2012] and from curating terms from the keywords that appeared most frequently in paper titles over time in the venues.

The keywords chosen for the classical keywords category were:
Cognition, Cognitive, Constraint Satisfaction, Game Theoretic, Game Theory, Heuristic Search, Knowledge Representation, Learning, Logic, Logical, Multiagent, Natural Language, Optimization, Perception, Planning, Problem Solving, Reasoning, Robot, Robotics, Robots, Scheduling, Uncertainty, and Vision.

The curated trending keywords were:
Autonomous, Boltzmann Machine, Convolutional Networks, Deep Learning, Deep Networks, Long Short Term Memory, Machine Learning, Mapping, Navigation, Neural, Neural Network, Reinforcement Learning, Representation Learning, Robotics, Self Driving, Self-Driving, Sensing, Slam, Supervised/Unsupervised Learning, and Unmanned.

The terms searched for were based on the issues exposed and identified in papers below, and also on the topics called for discussion in the First AAAI/ACM Conference on AI, Ethics, and Society.

## Conference and Public Venue - Sample

The AI group contains papers from the main artificial intelligence and machine learning conferences such as AAAI, IJCAI, ICML, and NIPS and also from both the *Artificial Intelligence Journal* and the *Journal of Artificial Intelligence Research* (JAIR).

The robotics group contains papers published in the IEEE Transactions on Robotics and Automation (now known as IEEE Transactions on Robotics), ICRA, and IROS.

The CS group contains papers published in the mainstream computer science venues such as the Communications of the ACM, IEEE Computer, ACM Computing Surveys, and the ACM and IEEE Transactions.

## Codebase

The code and data are hosted in this GitHub repository.

# CHAPTER 6: DIVERSITY IN AI

## LINKEDIN

### AI Skills Penetration

The aim of this indicator is to measure the intensity of AI skills in an entity (in a particular country, industry, gender, etc.) through the following methodology:

- Compute frequencies for all self-added skills by LinkedIn members in a given entity (occupation, industry, etc.) in 2015–2020.
- Re-weight skill frequencies using a TF-IDF model to get the top 50 most representative skills in that entity. These 50 skills compose the “skill genome” of that entity.
- Compute the share of skills that belong to the AI skill group out of the top skills in the selected entity.

**Interpretation:** The AI skill penetration rate signals the prevalence of AI skills across occupations, or the intensity with which LinkedIn members utilize AI skills in their jobs. For example, the top 50 skills for the occupation of engineer are calculated based on the weighted frequency with which they appear in LinkedIn members’ profiles. If four of the skills that engineers possess belong to the AI skill group, this measure indicates that the penetration of AI skills is estimated to be 8% among engineers (e.g., 4/50).

### Relative AI Skills Penetration

To allow for skills penetration comparisons across countries, the skills genomes are calculated and a relevant benchmark is selected (e.g., global average). A ratio is then constructed between a country’s and the benchmark’s AI skills penetrations, controlling for occupations.

**Interpretation:** A country’s relative AI skills penetration of 1.5 indicates that AI skills are 1.5 times as frequent as in the benchmark, for an overlapping set of occupations.

### Global Comparison: By Gender

The relative AI skill penetration by country for gender provides an in-depth decomposition of AI skills penetration across female and male labor pools and sample countries.

**Interpretation:** A country’s relative AI skill penetration rate of 2 for women means that the average penetration of AI skills among women in that country is two times the global average across the same set of occupations among women. If, in the same country, the relative AI skill penetration rate for men is 1.9, this indicates that the average penetration of AI skills among women in that country is 5% higher than that of men (calculated by dividing 1.9 by 2 and then subtracting 1, or 2/1.9-1) for the same set of occupations.

# CHAPTER 7: AI POLICY AND NATIONAL STRATEGIES

## BLOOMBERG GOVERNMENT

Bloomberg Government (BGOV) is a subscription-based market intelligence service designed to make U.S. government budget and contracting data more accessible to business development and government affairs professionals. BGOV's proprietary tools ingest and organize semi-structured government data sets and documents, enabling users to track and forecast investment in key markets.

### Methodology

The BGOV data included in this section was drawn from three original sources:

**Contract Spending**: BGOV's Contracts Intelligence Tool ingests on a twice-daily basis all contract spending data published to the beta.SAM.gov Data Bank, and structures the data to ensure a consistent picture of government spending over time. For the section "U.S. Government Contract Spending," BGOV analysts used FPDS-NG data, organized by the Contracts Intelligence Tool, to build a model of government spending on artificial intelligence-related contracts in the fiscal years 2000 through 2021. BGOV's model used a combination of government-defined produce service codes and more than 100 AI-related keywords and acronyms to identify AI-related contract spending.

**Defense RDT&E Budget:** BGOV organized all 7,057 budget line items included in the RDT&E budget request based on data available on the DOD Comptroller website. For the section "U.S. Department of Defense (DOD) Budget," BGOV used a set of more than a dozen AI-specific keywords to identify 305 unique budget activities related to artificial intelligence and machine learning worth a combined USD 5.0 billion in FY 2021.

**Congressional Record** (available on Congressional Record website): BGOV maintains a repository of congressional documents, including bills, amendments, bill summaries, Congressional Budget Office assessments, reports published by congressional committees, Congressional Research Service (CRS), and others. For the section "U.S. Congressional Record," BGOV analysts identified all legislation (passed or introduced), congressional committee reports, and CRS reports that referenced one or more of a dozen AI-specific keywords. Results are organized by a two-year congressional session.

## LIQUIDNET

Prepared by Jeffrey Banner and Steven Nichols

### Source

Liquidnet provides sentiment data that predicts the market impact of central bank and corporate communications. Learn more about Liquidnet here.

### Examples of Central Bank Mentions

Here are some examples of how AI is mentioned by central banks: In the first case, China uses a geopolitical environment simulation and prediction platform that works by crunching huge amounts of data and then providing foreign policy suggestions to Chinese diplomats or the Bank of Japan use of AI prediction models for foreign exchange rates. For the second case, many central banks are leading communications through either official documents—for example, on July 25, 2019, the Dutch Central Bank (DNB) published Guidelines for the use of AI in financial services and launched its six "SAFEST" principles for regulated firms to use AI responsibly—or a speech on June 4, 2019, by the Bank of England's Executive Director of U.K. Deposit Takers Supervision James Proudman, titled "Managing Machines: The Governance of Artificial Intelligence," focused on the increasingly important strategic issue of how boards of regulated financial services should use AI.

## MCKINSEY GLOBAL INSTITUTE

### Source

Data collection and analysis was performed by the McKinsey Global Institute (MGI).

### Canada (House of Commons)

Data was collected using the Hansard search feature on Parliament of Canada website. MGI searched for the terms "*Artificial Intelligence*" and "*Machine Learning*" (quotes included) and downloaded the results as a CSV. The date range was set to "all debates." Data is as of Dec. 31, 2020. Data are available online from Aug. 31, 2002.

Each count indicates that *Artificial Intelligence* or *Machine Learning* was mentioned in a particular comment or remark during the proceedings of the House of Commons. This means that within an event or conversation, if a member mentions *AI* or *ML* multiple times within their remarks, it will appear only once. However if, during the same event, the speaker mentions *AI* or *ML* in separate comments (with other speakers in between), it will appear multiple times. Counts for *Artificial Intelligence* or *Machine Learning* are separate, as they were conducted in separate searches. Mentions of the abbreviations *AI* or *ML* are not included.

### United Kingdom (House of Commons, House of Lords, Westminster Hall, and Committees)

Data was collected using the Find References feature of the Hansard website of the U.K. Parliament. MGI searched for the terms "*Artificial Intelligence*" and "*Machine Learning*" (quotes included) and catalogued the results. Data is as of Dec. 31, 2020. Data are available online from January 1, 1800 onward. Contains Parliamentary information licensed under the Open Parliament Licence v3.0.

As in Canada, each count indicates that *Artificial Intelligence* or *Machine Learning* was mentioned in a particular comment or remark during a proceeding. Therefore, if a member mentions *AI* or *ML* multiple times within their remarks, it will appear only once. However if, during the same event, the same speaker mentions *AI* or *ML* in separate comments (with other speakers in between), it will appear multiple times. Counts for *Artificial Intelligence* or *Machine Learning* are separate, as they were conducted in separate searches. Mentions of the abbreviations *AI* or *ML* are not included.

### United States (Senate and House of Representatives)

Data was collected using the advanced search feature of the U.S. Congressional Record website. MGI searched the terms "*Artificial Intelligence*" and "*Machine Learning*" (quotes included) and downloaded the results as a CSV. The "word variant" option was not selected, and proceedings included Senate, House of Representatives, and Extensions of Remarks, but did not include the Daily Digest. Data is as of Dec. 31, 2020, and data is available online from the 104th Congress onward (1995).

Each count indicates that *Artificial Intelligence* or *Machine Learning* was mentioned during a particular event contained in the Congressional Record, including the reading of a bill. If a speaker mentioned *AI* or *ML* multiple times within remarks, or multiple speakers mentioned *AI* or *ML* within the same event, it would appear only once as a result. Counts for *Artificial Intelligence* or *Machine Learning* are separate, as they were conducted in separate searches. Mentions of the abbreviations *AI* or *ML* are not included.

## U.S. AI POLICY PAPER

### Source

Data collection and analysis was performed by Stanford Institute of Human-Centered Artificial Intelligence and AI Index.

### Organizations

To develop a more nuanced understanding of the thought leadership that motivates AI policy, we tracked policy papers published by 36 organizations across three broad categories including:

Think Tanks, Policy Institutes & Academia: This includes organizations where experts (often from academia and the political sphere) provide information and advice on specific policy problems. We included the following 27 organizations: AI PULSE at UCLA Law, American Enterprise Institute, Aspen Institute, Atlantic Council, Berkeley Center for Long-Term Cybersecurity, Brookings

Institution, Carnegie Endowment for International Peace, Cato Institute, Center for a New American Security, Center for Strategic and International Studies, Council on Foreign Relations, Georgetown Center for Security and Emerging Technology (CSET), Harvard Belfer Center, Harvard Berkman Klein Center, Heritage Foundation, Hudson Institute, MacroPolo, MIT Internet Policy Research Initiative, New America Foundation, NYU AI Now Institute, Princeton School of Public and International Affairs, RAND Corporation, Rockefeller Foundation, Stanford Institute for Human-Centered Artificial Intelligence (HAI), Stimson Center, Urban Institute, Wilson Center.

Civil Society, Associations & Consortiums: Not-for profit institutions including community-based organizations and NGOs advocating for a range of societal issues. We included the following nine organizations: Algorithmic Justice League, Alliance for Artificial Intelligence in Healthcare, Amnesty International, EFF, Future of Privacy Forum, Human Rights Watch, IJIS, Institute for Electrical and Electronics Engineers, Partnership on AI

Industry & Consultancy: Professional practices providing expert advice to clients and large industry players. We included six prominent organizations in this space: Accenture, Bain & Co., BCG, Deloitte, Google AI, McKinsey & Company

## Methodology

Each broad topic area is based on a collection of underlying keywords that describes the content of the specific paper. We included 17 topics that represented the majority of discourse related to AI between 2019-2020. These topic areas and the associated keywords are listed below.

- Health & Biological Sciences: medicine, healthcare systems, drug discovery, care, biomedical research, insurance, health behaviors, COVID-19, global health
- Physical Sciences: chemistry, physics, astronomy, earth science
- Energy & Environment: Energy costs, climate change, energy markets, pollution, conservation, oil & gas, alternative energy
- International Affairs & International Security: international relations, international trade, developing countries, humanitarian assistance, warfare, regional security, national security, autonomous weapons
- Justice & Law Enforcement: civil justice, criminal justice, social justice, police, public safety, courts
- Communications & Media: social media, disinformation, media markets, deepfakes
- Government & Public Administration: federal government, state government, local government, public sector efficiency, public sector effectiveness, government services, government benefits, government programs, public works, public transportation
- Democracy: elections, rights, freedoms, liberties, personal freedoms
- Industry & Regulation: economy, antitrust, M&A, competition, finance, management, supply chain, telecom, economic regulation, technical standards, autonomous vehicle industry & regulation
- Innovation & Technology: advancements and improvements in AI technology, R&D, intellectual property, patents, entrepreneurship, innovation ecosystems, startups, computer science, engineering
- Education & Skills: early childhood, K-12, higher education, STEM, schools, classrooms, reskilling
- Workforce & Labor: labor supply and demand, talent, immigration, migration, personnel economics, future of work
- Social & Behavioral Sciences: sociology, linguistics, anthropology, ethnic studies, demography, geography, psychology, cognitive science
- Humanities: arts, music, literature, language, performance, theater, classics, history, philosophy, religion, cultural studies
- Equity & Inclusion: biases, discrimination, gender, race, socioeconomic inequality, disabilities, vulnerable populations
- Privacy, Safety & Security: anonymity, GDPR, consumer protection, physical safety, human control, cybersecurity, encryption, hacking
- Ethics: transparency, accountability, human values, human rights, sustainability, explainability, interpretability, decision-making norms

# GLOBAL AI VIBRANCY

## OVERVIEW

The tables below show the high-level pillar, sub-pillars, and indicators covered by the Global AI Vibrancy Tool. Each sub-pillar is composed of individual indicators reported in the Global AI Vibrancy Codebook. There are 22 metrics in total, with 14 metrics under Research and Development (R&D) pillar, 6 metrics under the Economy pillar, and 2 metrics available under the Inclusion pillar specific to gender diversity. To aid data-driven decision-making to design national policy strategies, the Global AI Vibrancy is available as a web tool.

### R&D

| SUB-PILLAR | VARIABLE |
|---|---|
| Conference Publications | Number of AI conference papers* |
| Conference Publications | Number of AI conference papers per capita |
| Conference Publications | Number of AI conference citations* |
| Conference Publications | Number of AI conference citations per capita |
| Journal Publications | Number of AI journal papers* |
| Journal Publications | Number of AI journal papers per capita |
| Journal Publications | Number of AI journal citations* |
| Journal Publications | Number of AI journal citations per capita |
| Innovation > Patents | Number of AI patents* |
| Innovation > Patents | Number of AI patents per capita |
| Innovation > Patents | Number of AI patent citations* |
| Innovation > Patents | Number of AI patent citations per capita |
| Journal Publications > Deep Learning | Number of Deep Learning papers* |
| Journal Publications > Deep Learning | Number of Deep Learning papers per capita |

### ECONOMY

| SUB-PILLAR | VARIABLE |
|---|---|
| Skills | Relative Skill Penetration |
| Skills | Number of unique AI occupations (job titles) |
| Labor | AI hiring index |
| Investment | Total AI Private Investment* |
| Investment | AI Private Investment per capita |
| Investment | Number of Startups Funded* |
| Investment | Number of funded startups per capita |

### INCLUSION

| SUB-PILLAR | VARIABLE |
|---|---|
| Gender Diversity | AI Skill Penetration (female) |
| Gender Diversity | Number of unique AI occupations (job titles), female |

The webtool allows users to adjust weights to each metric based on their individual preference. The default settings of the tool allow the user to select between three weighting options:

### All weights to midpoint

This button assigns equal weights to all indicators.

### Only absolute metrics

This button assigns maximum weights to absolute metrics. Per capita metrics are not considered.

### Only per capita metrics

This button assigns maximum weights to per capita metrics. Absolute metrics are not considered.

The user can adjust the weights to each metric based on their preference. The charts automatically update when any weight is changed.

The user can select "Global" or "National" view to visualize the results. The "Global" view offers a cross-country comparative view based on the weights selected by the user. The "National" view offers a country deep dive to assess which AI indicators a given country is relatively better at. The country-metric specific values are scaled (0-100), where 100 indicates that a given country has the highest number in the global distribution for that metric, and conversely small numbers like 0 or 1 indicates relatively low values in the global distribution This can help identify areas for improvement and identify national policy strategies to support a vibrant AI ecosystem.

## CONSTRUCTION OF THE GLOBAL AI VIBRANCY: COMPOSITE MEASURE

### Source

The data is collected by AI Index using diverse datasets that are referenced in the 2020 AI Index Report chapters.

### Methodology

Step 1: Obtain, harmonize, and integrate data on individual attributes across countries and time.
Step 2: Use Min-Max Scalar to normalize each country-year specific indicator between 0-100.
Step 3: Take arithmetic Mean per country-indicator for a given year.
Step 4: Build modular weighted for available pillars and individual indicators.

### Aggregate Measure

The AI Vibrancy Composite Index can be expressed in the following equation:

$$AI\ Vibrancy_{c,t} = \left( \Psi_{pillar} * \left[ \alpha_{c,t} * \Psi_{indicator} \right] \right) \div N$$

where *c* represents a country and *t* represents year, $\alpha_{c,t}$ is the scaled (0-100) individual indicator, $\Psi_{indicator}$ is the weight assigned to individual indicators, $\Psi_{pillar}$ is weight specific to one of the three high-level pillars and *N* is the number of indicators available for a given country for a specific year.

### Normalization

To adjust for differences in units of measurement and ranges of variation, all 22 variables were normalized into the [0, 100] range, with higher scores representing better outcomes. A minimum-maximum normalization method was adopted, given the minimum and maximum values of each variable respectively. Higher values indicate better outcomes. The normalization formula is:

$$Min - max\ scalar\ (MS100) = 100 * \frac{(value) - (min)}{(max) - (min)}$$

### Coverage and Nuances

A threshold of 73% coverage was chosen to select the final list of countries based on an average of available data between 2015-2020. Russia and South Korea were added manually due to their growing importance in the global AI landscape, even though they did not pass the 73% threshold.

## RESEARCH AND DEVELOPMENT INDICATORS

| ID | PILLAR | SUB-PILLAR | NAME | DEFINITION | SOURCE |
|---|---|---|---|---|---|
| 1 | Research and Development | Conference Publications | Number of AI conference papers* | Total count of published AI conference papers attributed to institutions in the given country. | Microsoft Academic Graph (MAG) |
| 2 | Research and Development | Conference Publications | Number of AI conference papers per capita | Total count of published AI conference papers attributed to institutions in the given country in per capita terms. The denominator is the population (in tens of millions) for a given year to obtain scaled values. | Microsoft Academic Graph (MAG) |
| 3 | Research and Development | Conference Publications | Number of AI conference citations* | Total count of AI conference citations attributed to institutions in the given country. | Microsoft Academic Graph (MAG) |
| 4 | Research and Development | Conference Publications | Number of AI conference citations per capita | Total count of AI conference citations attributed to institutions in the given country in per capita terms. The denominator is the population (in tens of millions) for a given year to obtain scaled values. | Microsoft Academic Graph (MAG) |
| 5 | Research and Development | Journal Publications | Number of AI journal papers* | Total count of published AI journal papers attributed to institutions in the given country. | Microsoft Academic Graph (MAG) |
| 6 | Research and Development | Journal Publications | Number of AI journal papers per capita | Total count of published AI journal papers attributed to institutions in the given country in per capita terms. The denominator is the population (in tens of millions) for a given year to obtain scaled values. | Microsoft Academic Graph (MAG) |
| 7 | Research and Development | Journal Publications | Number of AI journal citations* | Total count of AI journal citations attributed to institutions in the given country. | Microsoft Academic Graph (MAG) |
| 8 | Research and Development | Journal Publications | Number of AI journal citations per capita | Total count of AI journal citations attributed to institutions in the given country in per capita terms. The denominator is the population (in tens of millions) for a given year to obtain scaled values. | Microsoft Academic Graph (MAG) |
| 9 | Research and Development | Innovation > Patents | Number of AI patents* | Total count of published AI patents attributed to institutions in the given country. | Microsoft Academic Graph (MAG) |
| 10 | Research and Development | Innovation > Patents | Number of AI patents per capita | Total count of published AI patents attributed to institutions in the given country in per capita terms. The denominator is the population (in tens of millions) for a given year to obtain scaled values. | Microsoft Academic Graph (MAG) |
| 11 | Research and Development | Innovation > Patents | Number of AI patent citations* | Total count of published AI patents citations attributed to institutions of originating patent filing. | Microsoft Academic Graph (MAG) |
| 12 | Research and Development | Innovation > Patents | Number of AI patent citations per capita | Total count of published AI patent citations attributed to institutions in the given country of originating patent filing, in per capita terms. The denominator is the population (in tens of millions) for a given year to obtain scaled values. | Microsoft Academic Graph (MAG) |
| 13 | Research and Development | Journal Publications > Deep Learning | Number of deep learning papers* | Total count of arXiv papers on Deep Learning attributed to institutions in the given country. | arXiv, NESTA |
| 14 | Research and Development | Journal Publications > Deep Learning | Number of deep learning papers per capita | Total count of arXiv papers on Deep Learning attributed to institutions in the given country in per capita terms. The denominator is the population (in tens of millions) for a given year to obtain scaled values. | arXiv, NESTA |

## ECONOMY INDICATORS

| ID | PILLAR | SUB-PILLAR | NAME | DEFINITION | SOURCE |
|---|---|---|---|---|---|
| 15 | Economy | Skills | Relative skill penetration | Relative skill penetration rate measure is based on a method to compare how prevalent AI skills are at the average occupation in each country against a benchmark (e.g. the global average), controlling for the same set of occupations. | LinkedIn Economic Graph |
| 16 | Economy | Labor | AI hiring index | AI hiring rate is the percentage of LinkedIn members who had any AI skills (see the Appendix for the AI skill grouping) on their profile and added a new employer to their profile in the same month the new job began, divided by the total number of LinkedIn members in the country. This rate is then indexed to the average month in 2015-2016; for example, an index of 1.05 indicates a hiring rate that is 5% higher than the average month in 2015-2016. | LinkedIn Economic Graph |
| 17 | Economy | Investment | Total Amount of Funding* | Total amount of private investment funding received for AI startups (nominal USD). | Crunchbase, CapIQ, NetBase Quid |
| 18 | Economy | Investment | Total per capita funding | Total amount of private investment funding received for AI startups in per capita terms. The denominator is the population (in tens of millions) for a given year to obtain appropriately scaled values. | Crunchbase, CapIQ, NetBase Quid |
| 19 | Economy | Investment | Number of companies funded* | Total number of AI companies founded in the given country. | Crunchbase, CapIQ, NetBase Quid |
| 20 | Economy | Investment | Number of companies funded per capita | Total number of AI companies founded in the given country in per capita terms. The denominator is the population (in tens of millions) for a given year to obtain appropriately scaled values. | Crunchbase, CapIQ, NetBase Quid |

## INCLUSION INDICATORS

| ID | PILLAR | SUB-PILLAR | NAME | DEFINITION | SOURCE |
|---|---|---|---|---|---|
| 21 | Inclusion | Gender Diversity | AI skill penetration (female) | Relative skill penetration rate measure is based on a method to compare how prevalent AI skills are at the average occupation in each country against a benchmark (e.g. the global average), controlling for the same set of occupations. | LinkedIn Economic Graph |
| 22 | Inclusion | Gender Diversity | Number of unique AI occupations (job titles), female | Number of unique AI occupations (or job titles) with high AI skill penetration for females in a given country. | LinkedIn Economic Graph |

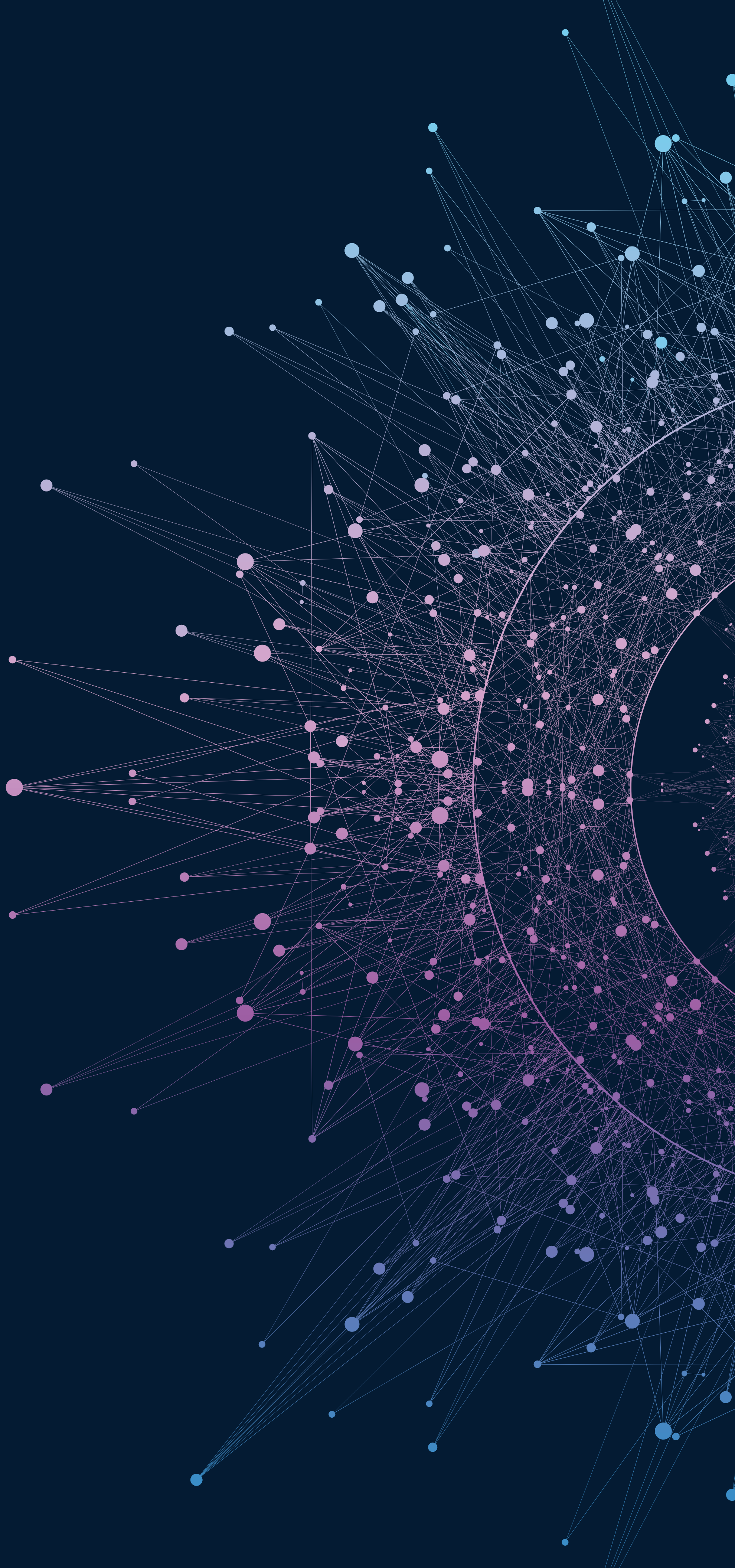

Artificial
Intelligence
Index Report 2021

**Stanford University**
Human-Centered
Artificial Intelligence